%% file: main.tex
\documentclass[11pt,fleqn]{book} 

\usepackage{algorithm}
\usepackage{algpseudocode}

\newfloat{algorithm}{t}{lop}
\floatname{algorithm}{Algorithm}

\input{structure} 

\begin{document}
	
	
	\begingroup
	\thispagestyle{empty}
	\begin{tikzpicture}[remember picture,overlay]
		\coordinate [below=12cm] (midpoint) at (current page.north);
		\node at (current page.north west)
		{\begin{tikzpicture}[remember picture,overlay]
				\node[anchor=north west,inner sep=0pt] at (0,0) {\includegraphics[width=1.04\paperwidth]{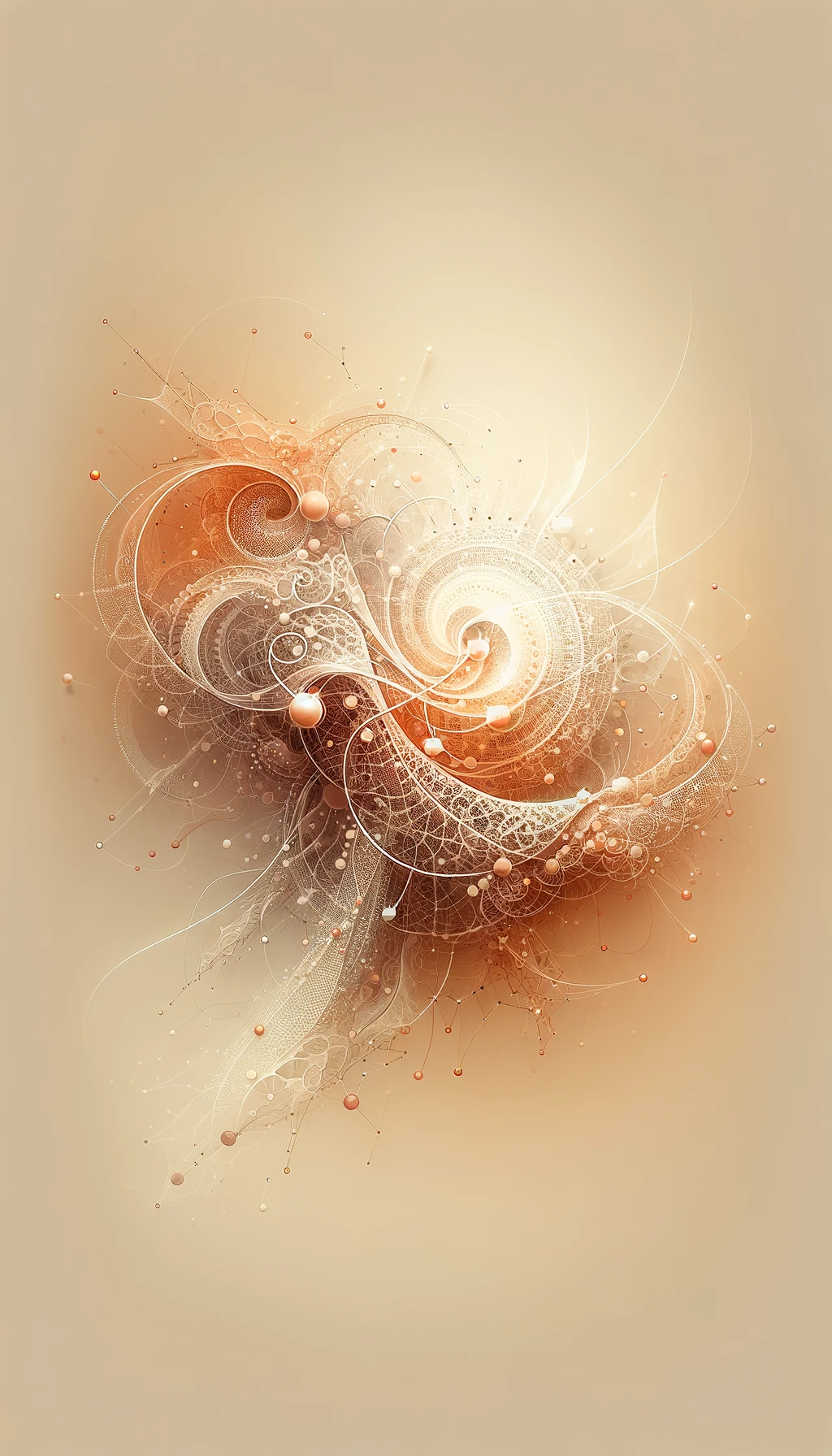}}; 
				\draw[anchor=north] (midpoint) node [fill=ocre!30!white,fill opacity=0.7,text opacity=1,inner sep=1cm]{\Huge\centering\bfseries\sffamily\parbox[c][][t]{\paperwidth}{\centering Introduction to Algogens\\[15pt] 
						{\Large Integrating Generative AI with Algorithmic Frameworks}\\[20pt] 
						{\huge Amir Shachar}}}; 
		\end{tikzpicture}};
	\end{tikzpicture}
	\vfill
	\endgroup
	
	\newpage
	~\vfill
	\thispagestyle{empty}
	
	\noindent Copyright \copyright\ 2024 Amir Shachar\\ 
	
	\noindent \textsc{Published by Amir Shachar}\\ 
	
	\noindent \textsc{www.amirshachar.com}\\ 
	
	\noindent Licensed under the Creative Commons Attribution-NonCommercial 3.0 Unported License (the ``License''). You may not use this file except in compliance with the License. You may obtain a copy of the License at \url{http://creativecommons.org/licenses/by-nc/3.0}. Unless required by applicable law or agreed to in writing, software distributed under the License is distributed on an \textsc{``as is'' basis, without warranties or conditions of any kind}, either express or implied. See the License for the specific language governing permissions and limitations under the License.\\ 
	
	\noindent \textit{First printing, February 2024} 
	
	\clearpage
	\begin{center}
		\thispagestyle{empty}
		\vspace*{\fill}
		To my beloved parents:
		
		Sarit, an artisan in parenting, as inventive as a symphony;\linebreak
		Yaron, an expert in nurturing, as precise as clockwork;\linebreak
		Together, embodying the perfect blend of creativity and accuracy.
		\vspace*{\fill}
	\end{center}
	\clearpage
	
	
	
	\chapterimage{pngs/table_of_contents.png} 
	
	\pagestyle{empty} 
	
	\tableofcontents 
	
	\cleardoublepage 
	
	\pagestyle{fancy} 
	
	\part{Prologue}
	
	
	\chapterimage{pngs/overview.png} 
	
	\chapter{Overview}\index{Overview}
	
	\section{Abstract}\index{Abstract}
	This book introduces the concept of Algogens, a blend of generative artificial intelligence and traditional algorithms, as a new tool for solving complex problems across various fields. Algogens symbolize a significant change in tackling challenges, combining the creativity of AI with the precision of algorithms to overcome the limitations of each approach alone. The book covers everything from the basics of Algogens, their development, practical uses, and the benefits they offer, such as enhanced problem-solving, adaptability, and efficiency.

	Starting with an explanation of what Algogens are and their core principles, the book sets the stage for understanding how they work and why they're important. It then explores how Algogens represent an evolution in technology, driven by the need to address increasingly complex issues in areas like cybersecurity, healthcare, and environmental science. Through examples and case studies, the reader will see how Algogens are being used today to foster innovation and improve efficiency across different industries.

	The discussion also touches on the challenges and future possibilities for Algogens, including technical obstacles and ethical considerations. The book aims to provide a balanced view of where Algogens are headed and how they could further change the technological landscape.

	A significant focus is placed on why Algogens are revolutionary, detailing the specific benefits of integrating generative AI into algorithmic frameworks. It looks at what makes an algorithm suitable for enhancement with Algogens and considers the importance of compatibility and quality control.

	Ending on a forward-looking note, the book highlights the potential of Algogens to reshape how we solve problems and innovate, inviting readers and experts alike to contribute to this emerging field. It's not just an academic or professional guide but a call to be part of a new era in computational technology, emphasizing Algogens' role in pushing the boundaries of what's possible in the 21st century.

	
	\section{Introduction}\index{Introduction}
	\subsection{What are Algogens?}
	\subsubsection{Definition and Core Concepts}
	Algogens, a portmanteau of "algorithm" and "genetic" (from generative AI), represent a cutting-edge framework integrating the precision of algorithmic processes with the creativity and adaptability of generative artificial intelligence. This integration aims to harness the best of both worlds: the reliability and predictability of algorithms with the innovative potential of AI to generate novel solutions and ideas. By doing so, Algogens offer a holistic approach to solving complex problems that either domain alone could not fully address.
	
	\paragraph{The Essence of Algogens}
	At its heart, an Algogen is a hybrid tool designed to tackle challenges requiring both structured analytical strategies and creative, outside-the-box thinking. It embodies a synthesis approach where algorithms provide a solid foundation for problem-solving by breaking down tasks into manageable steps, ensuring consistency and reliability. In parallel, generative AI introduces a layer of dynamism and innovation, capable of generating new data, scenarios, or solutions that might not be immediately apparent or accessible through conventional algorithmic methods.
	
	\paragraph{Core Principles}
	The core principles of Algogens revolve around synergy, adaptability, and enhancement. Synergy refers to the seamless integration of algorithms and AI, where each complements the other to improve overall performance. Adaptability highlights the ability of Algogens to adjust to new information, learn from outcomes, and evolve over time, making them particularly suited for dynamic and complex environments. Enhancement underscores the value added by merging these technologies, such as increased efficiency, creativity, and the ability to solve previously intractable problems.
	
	\paragraph{Operational Framework}
	An Algogen operates on a dual-framework basis. First, it leverages algorithmic methods to organize, analyze, and process data, establishing a structured approach to understanding the problem space. This step ensures that the AI component has a solid foundation from which to generate innovative solutions. Following this, generative AI takes the reins, using the insights and parameters defined by the algorithms to explore a vast landscape of potential solutions, including those that might not be immediately obvious or traditionally considered.
	
	\paragraph{Generative AI's Role}
	Generative AI's role within Algogens is pivotal. It not only generates content but also simulates various scenarios and predicts outcomes based on vast datasets that it has been trained on. This capability allows Algogens to anticipate challenges and identify solutions across a broad spectrum of applications, from designing complex systems to solving nuanced problems in healthcare, finance, and environmental science.
	
	\paragraph{The Algorithmic Backbone}
	The algorithmic backbone of Algogens ensures that the generative AI's creativity is grounded in logical and systematic methodologies. Algorithms act as the guide rails, defining the problem space, setting boundaries, and providing a framework within which AI operates. This ensures that the solutions proposed by the AI are feasible, practical, and aligned with the objectives at hand.
	
	In summary, Algogens embody a novel approach to problem-solving that leverages the strengths of both algorithmic methods and generative AI. By doing so, they offer a robust, versatile, and innovative solution framework capable of addressing the multifaceted challenges of the modern world.
	
	\begin{figure}
		\centering
		\includegraphics[width=0.7\textwidth]{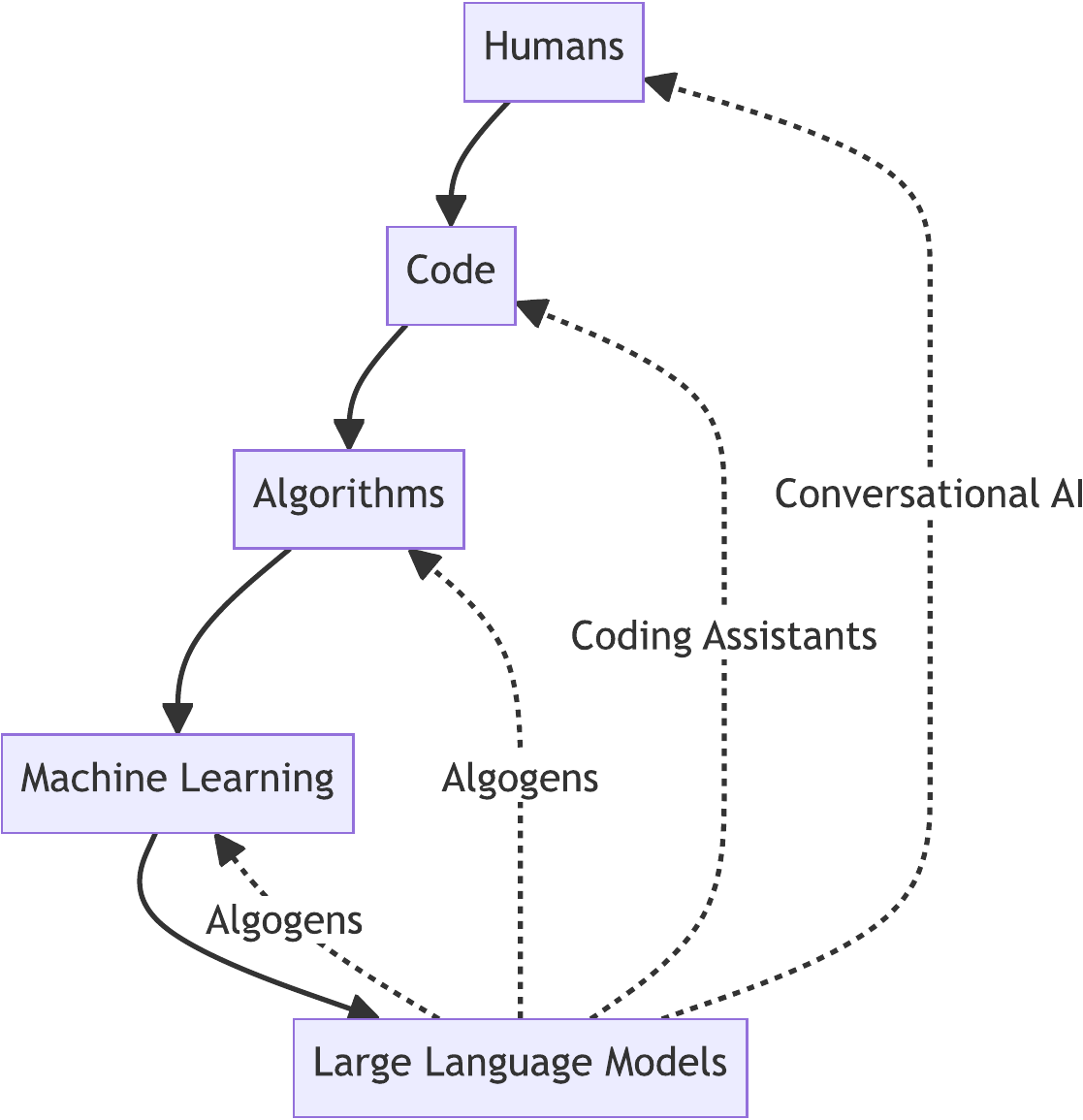} 
		\caption{Diagram illustrating the concept of Algogens: applications of Large Language Models to enhance algorithms and machine learning frameworks. Similar to how conversational AI applications interact with humans and coding assistants aid in software development, Algogens represent the feedback loop where insights from LLMs are utilized to iteratively refine and improve the underlying algorithms and machine learning models, driving advancements in AI capabilities and efficiency.}
		\label{fig:Algogens_vs_llm_applications}
	\end{figure}	
	
	\subsubsection{Components of Algogens}
	Algogens are distinguished by their innovative integration of algorithmic frameworks with generative artificial intelligence, merging the strengths of both to forge a powerful instrument for tackling complex problems. This fusion brings together the structured, rule-based logic of algorithms with the dynamic, creative potential of AI, enabling a multifaceted approach to problem-solving that surpasses the capabilities of either component in isolation.
	
	\paragraph{Algorithmic Frameworks}
	At the core of an Algogen lies its algorithmic frameworks, which establish a robust foundation for systematic problem analysis and solution formulation. These frameworks consist of precise procedures and sets of rules that direct the handling and processing of data, as well as the execution of tasks. They are instrumental in ensuring that the solutions derived are consistent, reliable, and predictable. Through tasks such as data preprocessing, structuring, and analysis, these algorithms prepare the ground for the AI component, making complex datasets accessible and amenable to further exploration. Additionally, they apply logical strategies to navigate and resolve problems within a defined parameter space, emphasizing efficiency and the pursuit of the most resource-effective solutions.
	
	\paragraph{Generative Artificial Intelligence}
	Complementing the deterministic nature of algorithmic frameworks, generative artificial intelligence introduces the capacity for innovation and adaptability. By generating new data, ideas, and solutions from learned patterns and examples, generative AI expands the realm of possibility, offering fresh perspectives and unforeseen approaches to intricate challenges. It excels in creating novel solutions by reinterpreting and extrapolating from its vast reservoir of training data, simulating diverse scenarios to anticipate outcomes and generate actionable insights. Moreover, its ability to adapt to evolving data and feedback enables a process of continual learning and refinement, ensuring that the solutions it proposes remain relevant and effective over time.
	
	\paragraph{Integrative Mechanisms}
	The essence of an Algogen's efficacy lies in the seamless integration of its algorithmic and AI components, a synergy that amplifies the strengths of each. This harmonization is achieved through sophisticated feedback loops that facilitate mutual learning between the algorithmic processes and the AI, fostering an environment of continuous improvement. The decision-making within Algogens embodies a hybrid approach, leveraging the innovative capabilities of AI within the structured, logical framework provided by algorithms, ensuring that solutions are both imaginative and grounded. This dynamic balance allows Algogens to adjust their operations to suit the specific demands of a task, optimizing performance across a diverse array of applications.
	
	\paragraph{Operational Synergy}
	The operational synergy between the algorithmic frameworks and generative AI is the defining characteristic of Algogens, enabling a comprehensive approach to problem-solving that leverages the best of both worlds. This collaboration not only enhances the creative and adaptive potential of AI with the stability and reliability of algorithms but also fosters a level of problem-solving proficiency that is innovative, efficient, and highly adaptable. Through this integrated approach, Algogens are equipped to confront a wide spectrum of challenges, delivering solutions that are both inventive and pragmatically viable, thereby setting a new standard in the field of technological problem-solving.

	\subsection{Previous Work: Algogens in the AI Literature}
	\subsubsection{Machine Learning}
	In the domain of Machine Learning, several papers offer innovative approaches for leveraging Large Language Models to enhance algorithmic processes.
	
	One paper introduces a significant stride towards addressing the gap in algorithm selection techniques by proposing an approach that integrates algorithm representation into the selection process. By employing distinct modules to extract representations of both problems and algorithms, the proposed model leverages the capabilities of pre-trained LLMs in code comprehension. Through comprehensive experiments, the authors validate the effectiveness of the proposed model and showcase the potential of LLMs in improving algorithm selection accuracy \cite{wu2024large}.
	
	On a similar note, another paper introduces the Algorithm of Thoughts, a novel strategy that propels LLMs through algorithmic reasoning pathways. By employing algorithmic examples, the approach exploits the innate recurrence dynamics of LLMs, expanding their idea exploration with minimal queries. The paper outperforms earlier single-query methods and stands on par with recent multi-query strategies, highlighting LLMs' inherent ability to optimize searches and weave intuition into algorithmic reasoning \cite{sel2023algorithm}. These papers collectively demonstrate the potential of LLMs in revolutionizing algorithmic processes within the domain of Machine Learning.
	
	\subsubsection{Verification}
	In the realm of Verification, \cite{zhang2024algo} introduces ALGO, a framework that synthesizes algorithmic programs with LLM-generated oracles to guide generation and verify their correctness. ALGO first generates a reference oracle by prompting an LLM to enumerate all combinations of relevant variables, which is then utilized to guide an arbitrary search strategy and verify synthesized algorithms. The study demonstrates that LLM-generated oracles are correct for a significant portion of cases, enhancing the performance of existing code generation models. Through experiments, ALGO achieves improved pass rates over state-of-the-art models on code contests, highlighting the potential of combining LLMs with verification frameworks to enhance algorithmic synthesis and correctness validation. This work showcases the promising intersection between Large Language Models and the field of Verification, offering insights into more robust algorithmic solutions.

	\subsubsection{Graph Computing}
	
	In the realm of graph computing, researchers have proposed innovative methods leveraging Large Language Models to tackle graph-related challenges.
	
	One paper introduces Graph Neural Prompting (GNP), a novel method to assist pre-trained LLMs in learning beneficial knowledge from knowledge graphs (KGs). GNP encompasses various designs, including a standard graph neural network encoder and a self-supervised link prediction objective. Extensive experiments demonstrate the superiority of GNP on commonsense and biomedical reasoning tasks, showcasing its potential across different LLM sizes and settings \cite{tian2023graph}.
	
	Similarly, another paper explores the potential of LLMs in graph machine learning, focusing on the node classification task. The authors investigate two possible pipelines: LLMs-as-Enhancers and LLMs-as-Predictors. Through comprehensive studies, they make original observations and find new insights, suggesting promising directions to leverage LLMs for learning on graphs \cite{chen2024exploring}. These papers contribute valuable insights into the integration of LLMs with graph computing, paving the way for advancements in this domain.

	\subsubsection{Medical Imaging}
	In the domain of medical imaging, researchers have proposed innovative approaches leveraging Large Language Models to enhance diagnostic processes and improve clinical decision-making.
	
	One paper provides a comprehensive review and tutorial for researchers in the field of medical imaging, emphasizing the potential of LLMs to improve tasks such as image captioning, report generation, and visual question answering. The authors highlight ChatGPT as a tool for exploring potential applications and discuss the benefits of accurate and efficient language models for medical imaging analysis \cite{hu2023advancing}.
	
	Similarly, another paper introduces ChatCAD, a framework for interactive computer-aided diagnosis on medical images using LLMs. The proposed method integrates LLMs with computer-aided diagnosis networks to enhance the interpretation of medical images by summarizing and reorganizing information in natural language text format. The goal is to merge the domain knowledge of LLMs with the vision understanding capability of existing medical-image CAD models, creating a more user-friendly and understandable system for clinical decision-making \cite{wang2023chatcad}. These papers contribute to the ongoing exploration of LLMs' potential in improving diagnostic processes and enhancing medical imaging analysis.

	\subsubsection{Optimization}
	In the realm of optimization, researchers have explored novel methods leveraging Large Language Models to enhance traditional optimization techniques and address complex optimization problems.
	
	One paper introduces Large Language Models as Optimizers (OPRO), a simple and effective approach that leverages LLMs to solve optimization problems described in natural language. OPRO generates new solutions iteratively based on prompts containing previously generated solutions, demonstrating promising results on linear regression, traveling salesman problems, and prompt optimization tasks \cite{yang2023large}.
	
	Another paper investigates the potential of using LLMs to generate novel metaheuristic optimization algorithms. The authors propose a framework that combines LLMs with genetic and evolutionary computation to synthesize hybrid swarm intelligence optimization algorithms. By leveraging the language processing capabilities of LLMs, the framework identifies and decomposes well-performing swarm algorithms to create new hybrid algorithms, demonstrating promising results across various optimization tasks \cite{pluhacek2023leveraging}.
	
	Furthermore, researchers have explored the integration of LLMs with Bayesian optimization (BO) to enhance optimization processes. One paper presents a novel approach that frames the BO problem in natural language terms, enabling LLMs to propose promising solutions conditioned on historical evaluations. By combining contextual understanding and few-shot learning proficiency of LLMs with domain knowledge, the approach improves surrogate modeling and candidate sampling, particularly in the early stages of search when observations are sparse \cite{liu2024large}.
	
	Similarly, another paper investigates the use of LLMs for real-world urban-delivery route optimization. The authors propose a novel approach based on LLMs to optimize delivery routes by learning from drivers' historical experiences and integrating implicit knowledge of complex delivery operating environments. By analogizing delivery routes to natural language sentences and leveraging LLMs to learn drivers' delivery behaviors, the approach demonstrates improved performance compared to traditional optimization-based approaches \cite{liu2023can}. These papers collectively contribute valuable insights into the integration of LLMs with optimization techniques, offering new avenues for addressing complex optimization problems.
	
	\subsubsection{Natural Languge Processing}
	In the domain of Natural Language Processing (NLP), one paper explores the potential of language models in solving graph problems presented in natural language descriptions. The authors propose NLGraph, a benchmark for evaluating language models on graph reasoning tasks described in natural language. NLGraph contains various graph reasoning tasks with different levels of complexity, ranging from simple connectivity and shortest path tasks to more challenging problems like maximum flow and graph neural network simulation. The authors evaluate the performance of large language models, particularly GPT-3 and GPT-4, on the NLGraph benchmark using different prompting approaches. They find that while language models demonstrate preliminary graph reasoning abilities, advanced prompting approaches are less effective on complex graph problems. To enhance language models' performance on natural language graph problems, the authors propose two instruction-based approaches: Build-a-Graph Prompting and Algorithmic Prompting. These approaches leverage language models to generate better solutions for graph reasoning tasks, improving performance across multiple tasks and settings. The NLGraph benchmark and the proposed prompting approaches offer valuable insights into the capabilities of language models for solving graph problems described in natural language \cite{wang2024can}.
	
	\subsubsection{Evolutionary Computing}
	
	In the realm of evolutionary computing, the following papers shed light on innovative approaches leveraging large language models and evolutionary algorithms (EAs). The first paper proposes EvoPrompt, a framework for discrete prompt optimization, which integrates LLMs with EAs to automate prompt generation for various tasks. EvoPrompt employs EAs to iteratively generate and improve prompts based on LLM outputs, significantly outperforming human-engineered prompts and existing methods for automatic prompt generation. This novel framework demonstrates the synergies between LLMs and EAs, paving the way for further research in combining language models with conventional optimization algorithms \cite{guo2023connecting}. The second paper explores the strong consistency between LLMs and EAs, highlighting key characteristics shared between text sequence generation and evolutionary processes. The authors illustrate how this consistency can be leveraged to develop evolved artificial agents that excel in complex problem-solving tasks. By analyzing existing coupling studies and outlining future research directions, the paper provides valuable insights into harnessing the collective power of LLMs and EAs for solving real-world problems \cite{chao2024match}.
	
	\subsubsection{Robotics}
	In the field of robotics, researchers are exploring innovative methods to empower embodied agents using large language models. One notable paper introduces LLM-Planner, a novel approach for few-shot planning in visually-perceived environments. This method harnesses the power of LLMs to enable embodied agents to follow natural language instructions and complete complex tasks. LLM-Planner leverages large language models to generate and update plans grounded in the current environment, achieving competitive few-shot performance even with limited training data. By bridging the gap between language understanding and embodied action, this work opens new possibilities for developing versatile and sample-efficient embodied agents that can quickly adapt to new tasks in real-world settings \cite{song2023llm}.

	\subsection{The Rise of Algogens}
	\subsubsection{Technological Evolution and the Need for Algogens}
	
	The rapid pace of technological evolution has catalyzed the emergence of increasingly complex and dynamic challenges across various domains, from cybersecurity and healthcare to environmental management and beyond. Traditional computational methods, while robust, often fall short in addressing these challenges with the required agility and innovation. This gap between existing capabilities and emerging demands underscores the critical need for a new paradigm in problem-solving technologies, leading to the development of Algogens.
	
	Algogens, as a concept, arise from a recognition of the limitations inherent in both standalone algorithmic approaches and generative AI. Algorithms, for all their precision and reliability, are constrained by their predefined parameters and lack the capacity for creativity and adaptability. Conversely, generative AI offers unparalleled potential for innovation and flexibility but can be prone to unpredictability and lacks the structured approach necessary for many applications. The evolving landscape of technology, with its emphasis on speed, efficiency, and adaptability, calls for a solution that combines the best of both worlds.
	
	The advent of Algogens represents a direct response to this need, offering a synergistic framework that leverages the structured logic of algorithms with the dynamic problem-solving capabilities of generative AI. This integration facilitates a more nuanced approach to computational challenges, enabling solutions that are both innovative and grounded in logical precision. The ability of Algogens to adapt to new data, learn from evolving scenarios, and generate creative solutions positions them as a pivotal innovation in the era of rapid technological change.
	
	Moreover, the emergence of Algogens reflects a broader trend in the technological evolution towards systems that are not only intelligent but also highly adaptable and capable of dealing with the complexities of the real world. By bridging the gap between deterministic algorithms and the generative power of AI, Algogens offer a promising pathway to addressing the multifaceted challenges of the 21st century. They embody the convergence of human ingenuity with the potential of artificial intelligence, marking a significant milestone in our quest to harness technology for more effective problem-solving.

	\subsubsection{Advantages of Algogens Over Traditional Approaches}
	
	The development and implementation of Algogens herald a transformative shift in how complex problems are approached and solved, offering distinct advantages over traditional computational methods. These advantages underscore the reasons why Algogens are becoming increasingly prominent in various fields, from technology and science to business and environmental management.
	
	\paragraph{Complex Problem-Solving}
	Algogens excel in tackling complex problems that traditional algorithms and standalone AI systems often struggle with. By integrating the deterministic nature of algorithms with the creative capabilities of generative AI, Algogens can dissect and understand multifaceted issues in ways that were previously unattainable. This approach allows for a nuanced analysis of complex scenarios, enabling the identification of innovative solutions that are both effective and efficient. The capacity of Algogens to process vast amounts of data, recognize patterns, and generate novel insights is particularly valuable in domains where complexity and unpredictability are the norms, such as in big data analytics, climate modeling, and healthcare diagnostics.
	
	\paragraph{Adaptability and Innovation}
	One of the hallmark advantages of Algogens is their inherent adaptability and capacity for innovation. Unlike traditional algorithms, which operate within a fixed set of rules, Algogens can learn from new data, adapt to changing conditions, and generate creative solutions to unforeseen problems. This flexibility is crucial in today’s rapidly changing technological landscape, where the ability to pivot and innovate is key to maintaining relevance and effectiveness. Algogens' adaptability not only enhances their problem-solving capabilities but also ensures that they can evolve over time, staying at the forefront of technological advancements and emerging challenges.
	
	\paragraph{Efficiency and Scalability}
	Algogens also offer significant benefits in terms of efficiency and scalability. By automating the process of data analysis and solution generation, they reduce the need for intensive manual intervention, thereby saving time and resources. Furthermore, the scalability of Algogens allows them to be applied to problems of varying sizes and complexities, from small-scale operational issues to large-scale strategic challenges. This scalability is complemented by the efficiency with which Algogens can process information and generate solutions, making them an ideal choice for organizations looking to enhance their operational efficiencies and strategic capabilities. The ability to scale and adapt without a proportional increase in resources or complexity is a critical advantage in an era characterized by rapid growth and ever-expanding data volumes.

	\subsection{Structure of the Book}
	\subsubsection{Understanding Algogens}
	
	Understanding Algogens involves delving into their theoretical foundations and the technological mechanisms that enable their operation. This exploration sheds light on how Algogens represent a significant advancement over traditional computational methods, offering a nuanced approach that combines the strengths of algorithmic precision with the generative capabilities of artificial intelligence.
	
	\paragraph{Theoretical Foundations}
	The theoretical underpinnings of Algogens are rooted in both computer science and artificial intelligence disciplines. At their core, Algogens are based on the principle of leveraging deterministic algorithms to structure and process data, while simultaneously employing generative AI to explore a vast space of potential solutions that human intuition might not readily conceive. This dual approach allows Algogens to tackle problems with a level of depth and creativity that neither purely algorithmic nor solely AI-based systems can achieve on their own. The theoretical foundations also draw from complexity science, which studies how relationships between parts give rise to the collective behaviors of a system and how the system interacts and forms relationships with its environment. This interdisciplinary approach is crucial for understanding how Algogens can model complex systems and predict their behavior in a way that is both accurate and adaptable.
	
	\paragraph{Technological Mechanisms}
	The technological mechanisms that enable Algogens to function effectively involve a combination of advanced data processing, machine learning models, and algorithmic decision-making processes. Initially, Algogens utilize algorithms to organize and analyze data, setting the stage for the generative AI component to generate innovative solutions. This generative component is powered by machine learning models, particularly those capable of understanding and creating complex patterns and predictions, such as neural networks and deep learning techniques. The interaction between the algorithmic and AI components is facilitated through a feedback loop, where generated solutions are evaluated, refined, and tested against real-world criteria to ensure their viability and effectiveness. This iterative process allows Algogens to continuously learn and improve, adapting to new data and evolving requirements. The technological mechanisms of Algogens thus embody a dynamic and flexible approach to problem-solving that is both rigorous and innovative, enabling them to address a wide range of challenges with unprecedented efficiency and creativity.

	\subsubsection{Applications of Algogens}
	
	The applications of Algogens span a wide array of industries, showcasing their versatility and the broad impact they can have on solving complex problems. This section highlights the applications of Algogens and provides insights into case studies and real-world examples where Algogens already have been, and potentially can be, successfully implemented.
	
	\paragraph{Enhancing Established Algorithms with Algogens}
	We will survey a varied set of algorithms from many different domains, observe their interplay with classical machine learning and AI in the literature, and suggest that they can be further enhanced with generative AI, particularly large language models. These advanced models can help to further enhance efficiency and overcome existing limitations.
	
	\paragraph{Industry-Specific Applications}
	Algogens are uniquely positioned to revolutionize multiple sectors by offering innovative solutions to longstanding challenges. In healthcare, for example, Algogens can be used to analyze patient data, identify patterns, and predict health outcomes, enabling personalized medicine and early detection of diseases. In the realm of finance, they can process vast amounts of market data to forecast trends, manage risks, and optimize investment strategies. Algogens also have significant applications in cybersecurity, where they can anticipate and mitigate potential threats through predictive modeling and simulation. Additionally, in environmental science, Algogens contribute to climate change modeling and natural resource management, helping to predict environmental impacts and inform conservation efforts. These industry-specific applications demonstrate the adaptability of Algogens to different domains, leveraging their dual strengths in algorithmic analysis and AI-driven innovation to address diverse challenges.
	
	\paragraph{Case Studies and Real-World Examples}
	Several case studies underscore the effectiveness of Algogens in practical scenarios. One notable example is in the healthcare industry, where an Algogen-based system was developed to predict the outbreak of infectious diseases by analyzing patterns in healthcare data and social media. This system enabled early intervention and significantly reduced the spread of disease. Another example can be found in the finance sector, where Algogens were used to develop a predictive model for stock market trends, combining historical data analysis with real-time market sentiment analysis. This model provided investors with more accurate forecasts, leading to better investment decisions. In cybersecurity, Algogens have been deployed to create dynamic defense mechanisms that adapt to evolving threats, significantly enhancing the security of information systems. These real-world examples illustrate the practical benefits of Algogens, showcasing their ability to not only solve complex problems but also to innovate and improve upon traditional methods. Through these applications, Algogens prove to be a transformative force across industries, heralding a new era of problem-solving capabilities.

	\subsubsection{Challenges and Future Directions}
	
	While Algogens represent a significant advancement in the integration of generative AI with algorithmic methods, they are not without challenges. Addressing these challenges is crucial for their continued development and adoption. This section explores the technical challenges and ethical considerations associated with Algogens and discusses future trends and evolutions that may influence their trajectory.
	
	\paragraph{Technical Challenges}
	The implementation of Algogens involves several technical challenges, primarily related to computational demands, data integration, and algorithmic complexity. The computational power required to run advanced AI models, especially when processing large datasets, poses a significant challenge, necessitating efficient algorithms and optimized hardware. Furthermore, integrating diverse data sources while ensuring data quality and relevance is critical for the effective functioning of Algogens. Algorithmic complexity, particularly in designing systems that can dynamically adapt to changing conditions and inputs, also presents a challenge. Addressing these technical hurdles requires ongoing research and development, with a focus on enhancing computational efficiency, improving data processing capabilities, and refining algorithmic models to ensure Algogens can operate effectively across different domains and scales.
	
	\paragraph{Ethical Considerations}
	As with any technology that leverages AI, ethical considerations are paramount. The use of Algogens raises questions about data privacy, bias in AI-generated solutions, and the potential for misuse in sensitive applications. Ensuring that Algogens operate within ethical boundaries involves implementing strict data governance policies, conducting bias audits to identify and mitigate potential biases in AI models, and establishing clear guidelines for the ethical use of Algogens, particularly in areas such as healthcare and law enforcement. Addressing these ethical challenges is essential for maintaining public trust and ensuring that the benefits of Algogens are realized without compromising individual rights or societal values.
	
	\paragraph{Future Trends and Evolutions}
	Looking ahead, the evolution of Algogens is likely to be influenced by advancements in AI and algorithmic research, as well as by changing societal needs and technological landscapes. One key trend is the increasing integration of quantum computing, which could significantly enhance the computational capabilities of Algogens, allowing for the processing of complex datasets at unprecedented speeds. Additionally, the growing emphasis on ethical AI and responsible technology use is expected to shape the development of Algogens, with a focus on transparency, fairness, and accountability. As Algogens continue to evolve, they are poised to become even more integral to solving complex problems, driving innovation across industries, and contributing to societal progress. The future of Algogens, while promising, will depend on how effectively these challenges are addressed and how the technology adapts to emerging trends and ethical considerations.

	\subsection{Algorithmic Advancements: Generative AI versus Traditional AI/ML}
	\subsubsection{The Unique Value of Generative AI in Algogens}
	
	Generative AI holds a distinctive place within the Algogen framework, offering unique advantages that traditional algorithmic approaches alone cannot provide. This section delves into the contributions of generative AI to the Algogen methodology, emphasizing its role in enhancing creativity, flexibility, and problem-solving capabilities.
	
	Generative AI introduces an unparalleled level of creativity and innovation to the problem-solving process. Unlike traditional algorithms that follow predefined paths and rules, generative AI can produce novel ideas, scenarios, and solutions that might not be immediately apparent or even conceivable to human designers or traditional computational models. This creative capability allows Algogens to explore a wider solution space, generating unique approaches to complex problems.
	
	Flexibility is another critical advantage brought by generative AI. It enables Algogens to adapt to new, unforeseen challenges without the need for extensive reprogramming or manual intervention. This adaptability is crucial in dynamic environments where conditions and requirements can change rapidly. Generative AI empowers Algogens to learn from these changes, adjusting its strategies and solutions in real-time, thereby maintaining relevance and effectiveness over time.
	
	Moreover, generative AI enhances the problem-solving capabilities of Algogens by enabling them to handle ambiguity and uncertainty more effectively. Traditional algorithms require clear, well-defined problems and input parameters. In contrast, generative AI can work with incomplete, ambiguous, or noisy data, making inferences and generating solutions that can be refined and improved through iterative processes. This ability is particularly valuable in real-world applications where data may be imperfect or incomplete, allowing Algogens to deliver practical solutions under less-than-ideal conditions.
	
	The integration of generative AI into Algogens does not come without its challenges, such as ensuring the accuracy and reliability of AI-generated solutions and managing the computational resources required. However, the unique value it adds in terms of creativity, flexibility, and enhanced problem-solving capabilities makes it an indispensable component of the Algogen approach. As generative AI continues to evolve, its role in Algogens is expected to become even more significant, driving innovation and offering new possibilities for tackling the complex challenges of the modern world.

	\subsubsection{Benefits and Advantages}
	
	Within the broader discourse on the integration of artificial intelligence with traditional algorithmic approaches, the question arises as to why the focus has shifted specifically towards Algogens, rather than AlgoDeep or AlgoML. This distinction is not merely terminological but reflects a strategic choice underpinned by the unique benefits and advantages that generative AI, as employed in Algogens, offers over more conventional deep learning (AlgoDeep) or machine learning approaches (AlgoML).
	
	The primary benefit of utilizing generative AI within Algogens lies in its ability to generate new data points, scenarios, or solutions that did not previously exist in the training data. This is a departure from the typical machine learning models of AlgoML, which are generally more predictive in nature, focusing on identifying patterns within existing datasets rather than creating new ones. Similarly, while deep learning models (AlgoDeep) have demonstrated remarkable achievements in areas such as image and speech recognition, they still largely operate within the confines of pattern recognition and classification, rather than the creation of new content or solutions.
	
	Generative AI, by contrast, excels in areas where innovation and creativity are paramount. It can imagine new designs, hypothesize novel solutions to complex problems, and simulate outcomes for scenarios that have not been explicitly programmed into its algorithms. This capacity for generative creativity is not just an incremental improvement but a fundamental shift in how problems can be approached and solved, offering a broader spectrum of solutions beyond the limitations of existing data or precedents.
	
	Furthermore, Algogens benefit from the synergy between the structured, rule-based reasoning of traditional algorithms and the dynamic, generative capabilities of AI. This hybrid approach ensures that the creativity and innovation brought by generative AI are grounded in logical, systematic frameworks provided by algorithmic methods, resulting in solutions that are both inventive and reliable. The advantage here is a more balanced, holistic approach to problem-solving that leverages the best of both worlds: the creativity of AI and the dependability of algorithms.
	
	In terms of practical applications, the Algogen framework is designed to be more adaptable and efficient in solving complex, real-world problems compared to AlgoDeep or AlgoML. This adaptability is crucial in environments where conditions are constantly changing, and solutions need to evolve in response to new challenges. Algogens, with their generative AI component, are inherently more flexible, allowing them to quickly generate and test new solutions as the situation demands, without the need for extensive retraining or manual adjustments.
	
	In summary, the preference for Algogens over AlgoDeep or AlgoML is rooted in the unique benefits that generative AI brings to the table: creativity, innovation, and the ability to generate novel solutions. Combined with the structured, logical frameworks of traditional algorithms, Algogens offer a potent approach to tackling complex problems with an efficiency, adaptability, and breadth of perspective that other approaches cannot match.
	
	\subsubsection{Caveats and Considerations}
	
	While the benefits of integrating generative AI into algorithmic frameworks through Algogens are substantial, it is crucial to acknowledge and address the caveats and considerations that accompany this innovative approach. These challenges underscore the importance of a cautious and informed application of Algogens, ensuring that their implementation maximizes benefits while mitigating potential drawbacks.
	
	One significant caveat of employing generative AI within Algogens is the potential for generating unrealistic or impractical solutions. Unlike traditional algorithms, which operate within a defined set of parameters and rules, generative AI can produce outcomes that, while novel and creative, may not always align with real-world constraints or practicality. This challenge necessitates rigorous validation and testing processes to ensure that the generated solutions are not only innovative but also applicable and feasible in the intended context.
	
	Another consideration is the complexity and unpredictability of the generative AI models themselves. The internal workings of these models can be opaque, leading to what is often referred to as the "black box" problem. This lack of transparency can make it difficult to understand how and why certain solutions are generated, complicating efforts to debug or refine the system. Moreover, the unpredictability of generative outputs requires a robust framework for monitoring and controlling the AI's creative processes, ensuring that they remain aligned with the system's goals and ethical standards.
	
	Data dependency represents a further caveat. Generative AI's effectiveness is heavily reliant on the quality and quantity of the data it is trained on. Biases or inaccuracies in the training data can lead to skewed or discriminatory outcomes, highlighting the need for careful data curation and an ongoing assessment of the AI's outputs for unintended biases or errors.
	
	Additionally, the integration of generative AI into Algogens raises ethical considerations, particularly concerning the autonomy of the AI's generative processes. As these systems become capable of generating novel solutions, questions arise about the ownership of these ideas and the ethical implications of AI-generated content. This aspect demands clear guidelines and ethical frameworks to govern the use and application of generative AI within Algogens, ensuring that its capabilities are used responsibly and beneficially.
	
	Lastly, there is the issue of resource intensity. Generative AI models, especially those capable of producing complex and novel outputs, can be computationally demanding. This requires significant computational resources, which can be a limiting factor for organizations with limited access to such resources. Balancing the benefits of generative AI's innovative potential with the practicalities of computational costs and energy consumption is a critical consideration for the sustainable and efficient use of Algogens.
	
	In conclusion, while Algogens offer a promising avenue for leveraging the benefits of generative AI in conjunction with traditional algorithms, it is imperative to approach their development and application with a mindful consideration of these caveats. Addressing these challenges through thoughtful design, ethical guidelines, and ongoing evaluation will be key to realizing the full potential of Algogens while navigating the complexities of their implementation.

	\subsection{Qualifying Algorithms for Algogen Enhancement}
	\subsubsection{Criteria for Selection}
	
	Identifying the right algorithmic frameworks for enhancement through Algogens involves a nuanced understanding of several key criteria. These criteria ensure that the integration of generative AI not only enhances the existing capabilities of an algorithm but also addresses specific needs that are unmet by traditional algorithmic approaches alone. The selection process is crucial for maximizing the effectiveness and applicability of Algogens in solving complex, dynamic problems.
	
	\paragraph{Complexity and Non-Linearity}
	
	One of the primary criteria for selecting an algorithm for Algogen enhancement is the inherent complexity and non-linearity of the problem it addresses. Algorithms that tackle problems characterized by complex relationships and interactions, where traditional linear approaches fail to capture the full scope or nuances, are prime candidates for Algogen integration. The addition of generative AI can introduce a level of adaptability and depth in processing that enables these algorithms to navigate and solve multifaceted, non-linear problems more effectively. This criterion recognizes that certain challenges require beyond-linear thinking and solutions, which generative AI can facilitate by exploring a broader solution space and uncovering non-obvious, innovative pathways to problem resolution.
	
	\paragraph{Creativity and Innovation Needs}
	
	Another critical criterion is the need for creativity and innovation in problem-solving. Algorithms operating in fields where solutions benefit from novel approaches or where traditional methods have plateaued in effectiveness can significantly benefit from Algogen enhancement. Generative AI's capability to ideate and generate creative solutions can invigorate these algorithmic frameworks, pushing the boundaries of what is possible and discovering new solutions that traditional algorithms might not conceive. This criterion emphasizes the value of generative AI in contributing a creative dimension to problem-solving, particularly in areas requiring innovation, such as design, content generation, or complex decision-making processes.
	
	\paragraph{Dynamic and Evolving Environments}
	
	Lastly, the suitability of an algorithm for Algogen enhancement is often determined by its application within dynamic and evolving environments. Algorithms that operate in settings characterized by rapid changes or unpredictability can greatly benefit from the incorporation of generative AI. Such environments require solutions that can adapt in real-time and anticipate future variations. Generative AI enhances algorithmic frameworks by introducing the ability to generate predictive models and adaptively respond to changes, ensuring solutions remain relevant and effective over time. This criterion highlights the importance of flexibility and foresight in algorithmic problem-solving, qualities that generative AI can significantly augment.
	
	In summary, the selection of algorithmic frameworks for Algogen enhancement is guided by the complexity and non-linearity of the problem, the need for creativity and innovation, and the dynamic nature of the application environment. These criteria help identify areas where the integration of generative AI can provide substantial benefits, ensuring that Algogen applications are both impactful and relevant to the challenges they aim to address.

	\subsubsection{Caveats in Integration}
	
	While the integration of generative AI with algorithmic frameworks promises significant advancements, it is essential to approach this fusion with a clear understanding of potential caveats. Successful integration requires careful consideration of how these two distinct technologies align and interact, ensuring that the combined approach enhances rather than complicates problem-solving efforts. Addressing these caveats is crucial for realizing the full potential of Algogens.
	
	\paragraph{Alignment and Compatibility}
	
	The first caveat concerns the alignment and compatibility between the generative AI components and the existing algorithmic frameworks. Not all algorithms are designed to seamlessly integrate with AI-driven approaches, and without proper alignment, the integration can lead to inefficiencies or unintended consequences. It is crucial to ensure that the generative AI's capabilities complement the algorithm's objectives, operating within the same problem domain and leveraging the AI to fill gaps or extend the algorithm's capabilities. This requires a detailed understanding of the algorithm's structure, its limitations, and how AI can enhance its performance without disrupting its core functionality. The process often involves iterative adjustments and fine-tuning to achieve a harmonious integration that leverages the strengths of both components effectively.
	
	\paragraph{Quality Control and Validation}
	
	Another significant caveat is the need for rigorous quality control and validation mechanisms throughout the integration process. The inclusion of generative AI introduces new variables and potential sources of error into the algorithmic framework, making robust validation essential to ensure the reliability and accuracy of the outcomes. Quality control measures must be implemented to monitor the AI's contributions, verifying that generated solutions or enhancements align with expected standards and real-world applicability. This involves developing comprehensive testing protocols and evaluation criteria that can assess the performance of the integrated system under various conditions. Additionally, continuous monitoring is necessary to identify and address any issues that arise post-integration, ensuring that the Algogen maintains its effectiveness and reliability over time.
	
	Addressing these caveats requires a thoughtful and systematic approach to integration, emphasizing the need for compatibility between technologies and stringent quality control. By acknowledging and proactively managing these challenges, developers can enhance the synergy between generative AI and algorithmic frameworks, unlocking the full potential of Algogens in solving complex problems.
	
	\subsection{Motivation}
	
	The motivation behind introducing and focusing on the \textit{Algogens} concept is manifold and rooted in both the current trends and future potentials of technological advancements.
	
	\paragraph{Clear Conceptual Framework}
	
	Establishing \textit{Algogens} provides a much-needed conceptual scaffold, delineating this hybrid approach with precision. This clarity fosters a conducive environment for systematic exploration, enabling researchers to navigate the intricacies of integrating AI with algorithmic methods more effectively.
	
	\paragraph{Focus and Organization}
	
	The increasing introduction of hybrid generative AI and traditional algorithmic methodologies in the literature underscores the evolution and significance of such approaches, thereby strengthening the notion that a distinct name, like "\textit{Algogens}," is needed. This emerging trend reflects a growing recognition within the scientific and technological communities of the unique benefits and challenges associated with integrating these two powerful paradigms. By formalizing this approach under a specific term, we not only acknowledge its growing prevalence and importance but also facilitate a more organized and focused discourse.
	
	\paragraph{Focus on Innovation}
	
	The \textit{Algogens} framework inherently emphasizes innovation, spotlighting the synergistic potential of combining generative AI with algorithmic problem-solving. This focus nurtures a fertile ground for breakthroughs that might otherwise remain unexplored within broader research paradigms.
	
	\paragraph{Facilitates Collaboration}
	
	By offering a common platform, \textit{Algogens} facilitates seamless collaboration across disciplines. This interdisciplinary approach is paramount for addressing complex, multifaceted problems, ensuring a holistic and comprehensive solution strategy.
	
	\paragraph{Enhances Communication}
	
	A distinct terminology enhances communication clarity, significantly impacting the reception and adoption of new technologies. It simplifies conveying complex ideas, making it easier for diverse audiences to grasp and engage with the innovations brought forth by \textit{Algogens}.
	
	\paragraph{Standardization and Best Practices}
	
	As the \textit{Algogens} concept gains traction, it paves the way for the development of standardized methodologies, best practices, and ethical guidelines. Such standardization is crucial for ensuring compatibility across different platforms and industries, promoting responsible and sustainable technological advancement.
	
	\paragraph{Inspires New Applications}
	
	The focused framework of \textit{Algogens} inspires novel applications, encouraging innovators to envision and realize new solutions that leverage the strengths of both AI and traditional algorithms. This has the potential to transform industries by introducing groundbreaking products, services, and processes.
	
	\paragraph{Educational Tool}
	
	Finally, the introduction of \textit{Algogens} into the academic and professional lexicon serves as a potent educational tool. It facilitates the development of targeted curricula and training programs, equipping the next generation of professionals with the knowledge and skills to thrive in this emerging field.
	
	In essence, the motivation behind defining and championing the concept of \textit{Algogens} is to provide a robust foundation for future innovations. It is a call to the research community and industry practitioners to embrace this integrative approach, fostering advancements that could redefine our technological capabilities and address some of the most pressing challenges of our time.

	\subsection{Conclusion}
	
	The journey through the realm of Algogens has illuminated a path towards a future where the synergy between generative AI and algorithmic frameworks redefines the landscape of problem-solving across various industries. This concluding section encapsulates the essence of Algogens, their transformative potential, and extends an invitation to delve deeper into this fascinating domain.
	
	\subsubsection{Recapitulating the Essence of Algogens}
	
	Algogens stand as a testament to the power of integration, marrying the precision and reliability of algorithmic methods with the creativity and adaptability of generative AI. This fusion not only enhances the capabilities of each component but also introduces a new paradigm in addressing complex and dynamic problems. The essence of Algogens lies in their ability to offer nuanced, innovative solutions that traditional approaches might overlook, signaling a significant leap in how we approach technological problem-solving.
	
	\subsubsection{The Promise and Potential of Algogens}
	
	The promise of Algogens extends far beyond the immediate benefits of enhanced problem-solving. It represents a shift towards more intelligent, efficient, and adaptable systems capable of navigating the complexities of the modern world. The potential applications are vast, ranging from healthcare and cybersecurity to environmental management and beyond, each offering a glimpse into a future where Algogens play a pivotal role in driving progress and innovation. As this technology evolves, its impact is expected to grow, opening new avenues for exploration and discovery.
	
	\subsubsection{Invitation to Explore Further}
	
	As we conclude this introductory exploration of Algogens, we extend an invitation to readers to delve deeper into this compelling field. The journey ahead promises to be one of discovery, challenge, and immense reward, as we uncover new ways to harness the power of Algogens. Whether you are a researcher, practitioner, or simply a technology enthusiast, the exploration of Algogens offers a unique opportunity to contribute to the shaping of our technological future. We encourage you to engage with this emerging field, participate in its development, and explore the myriad ways in which Algogens can transform our world.

	
	\part{Foundations}
	
	
	\chapterimage{pngs/theoretical_framework.png} 
	
	\chapter{Theoretical Framework}\index{Theoretical Framework}
	
	\section{Overview of Generative AI}\index{Overview of Generative AI}
	
	Generative Artificial Intelligence has emerged as a transformative force in AI, representing a significant shift from traditional, deterministic algorithms to models capable of generating new data and insights. This subsection delves into generative AI's core concepts, methodologies, and advancements, highlighting its profound impact across various domains.
	
	\subsection{Foundational Concepts of Generative AI}
	Generative Artificial Intelligence stands as a paradigmatic shift in the realm of machine learning, distinguished by its unique capacity to not only interpret and process existing data but also to generate entirely novel data or scenarios based on the patterns and structures it has discerned. At its foundational level, generative AI operates on the principle of learning from available data to create new instances that exhibit similar characteristics to the original dataset. This departure from traditional discriminative models, which focus primarily on classification or prediction tasks based on input data, underscores the transformative potential of generative models.
	
	Central to the concept of generative AI is its ability to extrapolate beyond the confines of existing data, effectively venturing into uncharted territory to produce outputs that are not mere permutations or combinations of known inputs but rather novel creations in their own right. This creative aspect of generative AI introduces a dimension of innovation and ingenuity, enabling AI systems to go beyond mere replication or mimicry of observed patterns and instead contribute fresh insights and solutions to complex problems.
	
	Moreover, generative AI's foundational concepts entail a deep understanding of probabilistic modeling, where the system learns the underlying probability distribution of the data it encounters. By grasping the nuances and intricacies of this distribution, generative models can then generate new samples that adhere to the learned distribution, thereby exhibiting coherence and fidelity to the original data while also introducing variations and novel combinations.
	
	In essence, the foundational concepts of generative AI herald a new era of machine intelligence characterized not only by its ability to interpret and comprehend data but also by its capacity to imagine and create, paving the way for unprecedented advancements in fields ranging from art and design to scientific discovery and problem-solving.
	
	\subsection{Learning Mechanisms in Generative AI}
	The learning mechanisms employed within the realm of generative AI represent a sophisticated interplay between computational algorithms and vast datasets. At the heart of this process lies the endeavor to glean intricate distributions and relationships inherent within the data, thereby empowering AI systems to extrapolate beyond mere replication towards the creation of entirely novel outputs.
	
	Prominent among these mechanisms are cutting-edge techniques such as Generative Adversarial Networks (GANs) and Variational Autoencoders (VAEs), which stand as pillars of innovation in the field. GANs, for instance, operate on a dual-network architecture, where one network, known as the generator, endeavors to produce synthetic data, while the other, the discriminator, evaluates the authenticity of these outputs. Through an iterative process of adversarial training, wherein the generator strives to deceive the discriminator, and the discriminator refines its discernment capabilities, GANs engender a symbiotic relationship between the two networks, resulting in the continual enhancement of the quality and fidelity of generated outputs.
	
	Conversely, Variational Autoencoders (VAEs) adopt a distinct approach, focusing on the encoding and decoding of data to facilitate generative processes. VAEs begin by encoding input data into a compressed latent space representation, effectively capturing the underlying structure of the data. Subsequently, through a decoding process, VAEs are capable of reconstructing and generating new data points, thereby traversing the continuum between replication and innovation. This compressed representation serves as a latent manifold from which novel data samples can be derived, imbuing VAEs with the capacity for creative exploration within the confines of learned data distributions.
	
	In essence, the learning mechanisms inherent in generative AI epitomize the convergence of theoretical ingenuity and computational prowess, propelling the field towards ever-greater heights of innovation and discovery.

	\subsection{Capabilities in Pattern Recognition and Predictive Modeling}
	Demonstrating remarkable proficiency in pattern recognition, generative AI possesses the uncanny ability to discern and replicate intricate patterns embedded within datasets, often eluding human observation. Through sophisticated algorithms and advanced learning mechanisms, these AI systems sift through vast troves of data, uncovering latent patterns and structures that serve as the bedrock for subsequent predictive modeling endeavors.
	
	In the domain of predictive modeling, generative AI emerges as a formidable ally, leveraging its adeptness in pattern recognition to anticipate future data points or scenarios with remarkable accuracy. By extrapolating from learned patterns, these systems can forecast a myriad of phenomena, ranging from meteorological events in weather forecasting to shifts in market trends and fluctuations in financial markets. In the healthcare sector, generative AI's predictive modeling capabilities find application in medical diagnosis, where early detection of anomalies or trends can significantly impact patient outcomes and treatment strategies.
	
	The inherent versatility of generative AI in pattern recognition and predictive modeling transcends disciplinary boundaries, offering invaluable insights and foresight across a spectrum of industries and domains. Its capacity to navigate complex datasets and extract actionable intelligence underscores its indispensability in addressing contemporary challenges and shaping the trajectory of future advancements.

	\subsection{Advancements in Natural Language Processing}
	Within the realm of natural language processing (NLP), the integration of generative AI has heralded a new era of innovation and progress. Spearheading these advancements are cutting-edge models such as GPT (Generative Pretrained Transformer), which have garnered widespread acclaim for their unprecedented capabilities in generating text that is both coherent and contextually relevant. These models transcend mere replication, delving into the realm of creative expression by simulating conversational dynamics and even producing imaginative compositions that rival human-generated content.
	
	The transformative impact of these advancements reverberates across various applications within the NLP landscape, revolutionizing industries reliant on language-driven technologies. In the realm of chatbots, for instance, generative AI's prowess in natural language generation imbues virtual assistants with enhanced conversational abilities, enabling more seamless interactions with users and fostering deeper engagement. Language translation services also stand to benefit significantly from these advancements, with AI-driven models capable of producing translations that exhibit greater fidelity to the nuances of human speech and expression.
	
	Moreover, the advent of generative AI in natural language processing has profound implications for automated content creation, where AI-generated text can serve as a foundational tool for generating diverse and engaging content across platforms. From news articles to marketing copy, generative AI empowers content creators with the ability to produce high-quality, contextually relevant content at scale, streamlining workflows and enhancing productivity.
	
	In summary, the advancements in natural language processing facilitated by generative AI represent a paradigm shift in our approach to language-driven technologies. By pushing the boundaries of linguistic creativity and expression, these innovations pave the way for a future where human-machine interaction is characterized by unprecedented levels of sophistication and nuance.

	\subsection{Innovations in Image Generation}
	In the realm of image generation and editing, generative AI stands as a beacon of innovation, reshaping the landscape of visual creativity in profound ways. One of the hallmark techniques in this domain is neural style transfer, a method wherein the stylistic elements of one image are seamlessly merged with the content of another, resulting in visually striking compositions that blend artistic flair with technical precision. This fusion of styles exemplifies the creative potential inherent within generative AI models, transcending traditional notions of image manipulation and ushering in a new era of artistic expression.
	
	Moreover, the advent of AI-generated art and deepfake technologies represents a paradigm shift in the realm of visual media and entertainment. Through sophisticated algorithms and advanced neural networks, generative AI systems can produce hyper-realistic images and videos that challenge our perceptions of reality. Deepfake technologies, in particular, have garnered significant attention for their ability to seamlessly superimpose facial expressions and gestures onto existing footage, blurring the line between fact and fiction. While these innovations showcase the remarkable capabilities of generative AI, they also raise profound ethical considerations regarding the authenticity and integrity of visual media.
	
	Indeed, the intersection of generative AI and image generation presents a myriad of opportunities and challenges, shaping the future of visual storytelling and artistic expression. As these technologies continue to evolve, it becomes imperative to navigate the ethical implications inherent within AI-generated content, ensuring responsible usage and stewardship of these powerful tools.

	\subsection{Impact on Automated Decision-Making}
	The burgeoning role of generative AI in automated decision-making underscores its growing significance in shaping organizational strategies and operational frameworks. Rather than merely executing predefined instructions, these AI models assume a proactive stance by generating diverse arrays of potential scenarios and outcomes. This proactive approach empowers decision-makers to navigate through a labyrinth of possibilities, thereby facilitating more informed and nuanced decision-making processes.
	
	This paradigm shift in decision-making is particularly salient in contexts characterized by complexity and uncertainty, where traditional decision-making approaches may falter. In strategic planning, for instance, generative AI enables decision-makers to anticipate and evaluate various strategic alternatives, each contingent upon a distinct set of circumstances and assumptions. By illuminating the potential consequences and trade-offs associated with different courses of action, these AI-driven insights arm organizations with the foresight needed to chart a course towards sustainable growth and resilience.
	
	Moreover, the impact of generative AI on automated decision-making extends beyond strategic planning to encompass policy development and implementation. In policy formulation, AI models facilitate the exploration of diverse policy options and their respective ramifications, enabling policymakers to weigh the potential benefits and risks associated with each alternative. This iterative process of scenario analysis and simulation empowers policymakers to craft policies that are adaptive, responsive, and attuned to the evolving needs of society.
	
	Furthermore, in complex problem-solving scenarios, generative AI serves as a catalyst for innovation and creativity. By generating novel solutions and exploring unconventional approaches, these AI-driven insights challenge conventional wisdom and spark new avenues for exploration. Whether addressing challenges in healthcare, finance, or environmental sustainability, generative AI augments human decision-making capabilities, fostering a symbiotic relationship between human expertise and machine intelligence.
	
	In essence, the impact of generative AI on automated decision-making transcends mere optimization or efficiency gains; it represents a paradigm shift in how organizations conceptualize, evaluate, and enact decisions. By harnessing the power of AI-driven insights, decision-makers are empowered to navigate through complexity with confidence, embracing uncertainty as an opportunity for growth and innovation.

	In conclusion, the advancements in generative AI mark a paradigm shift in the capabilities of artificial intelligence. From enhancing creativity and innovation to improving predictive accuracy and decision-making, the potential applications of generative AI are vast and continually evolving, presenting exciting opportunities and new challenges to explore.

	\section{Overview of Algorithmic Methods}\index{Overview of Algorithmic Methods}
	
	Algorithms form the bedrock of computational problem-solving, offering systematic and logical frameworks for processing data and making decisions. This expanded subsection delves into algorithms' fundamental principles, evolution, and diverse applications in modern computing.
	
	\subsection{Fundamental Principles of Algorithms}
	At its essence, an algorithm serves as a meticulously crafted roadmap, guiding the computational journey towards problem resolution or computation execution. These sequences of instructions are imbued with clarity and precision, ensuring that each step is unambiguously defined and comprehensible to both humans and machines alike. In their pursuit of efficiency, algorithms are meticulously engineered to minimize the temporal and resource overhead associated with their execution, thus optimizing computational performance and scalability.
	
	One of the defining characteristics of algorithms lies in their deterministic nature, wherein they consistently yield identical outputs when provided with the same input. This predictability fosters a sense of reliability and trustworthiness, essential attributes in domains where accuracy and repeatability are paramount. Moreover, algorithms exhibit a remarkable versatility, finding application across a diverse array of problem domains ranging from mathematical computations to data processing and beyond.
	
	Central to the design and implementation of algorithms are the foundational principles that underpin their efficacy and functionality. These principles emphasize the importance of logical coherence, wherein each step in the algorithmic sequence logically follows from its predecessor, ensuring the integrity and correctness of the overall solution. Furthermore, algorithms are imbued with adaptability, capable of accommodating varying inputs and scenarios while maintaining their efficacy and efficiency.
	
	The iterative refinement of algorithms represents a continuous journey towards optimization and improvement, driven by insights gleaned from real-world applications and theoretical advancements. Through meticulous analysis and experimentation, algorithms evolve to address emerging challenges and capitalize on newfound opportunities, thereby perpetuating a cycle of innovation and refinement.
	
	In summary, the fundamental principles of algorithms constitute the bedrock upon which the edifice of computational problem-solving is erected. By embodying clarity, precision, efficiency, determinism, and adaptability, algorithms empower practitioners to navigate the complexities of modern computing landscapes with confidence and efficacy.

	\subsection{Types and Characteristics of Algorithms}
	Within the vast landscape of computing, an array of algorithms flourishes, each meticulously crafted to address a distinct set of challenges and problem domains. From the intricate task of sorting data to the nuanced exploration of graphs, algorithms exhibit a diverse range of types and characteristics, each imbued with its own unique strengths and applications.
	
	Among the myriad categories of algorithms, sorting algorithms stand as stalwarts of computational efficiency, exemplifying precision and speed in arranging data elements in a predetermined order. Algorithms such as QuickSort and MergeSort epitomize this category, offering elegant solutions to the perennial task of organizing datasets with optimal efficiency.
	
	In the realm of search algorithms, where the quest for specific data points within a dataset reigns supreme, algorithms like binary search emerge as quintessential tools of exploration. Through a process of iterative refinement and binary partitioning, these algorithms navigate vast datasets with unparalleled efficiency, culminating in the rapid identification of target elements.
	
	Additionally, graph algorithms occupy a prominent position within the algorithmic pantheon, offering indispensable solutions to a myriad of network-related challenges. Dijkstra's algorithm for shortest paths stands as a beacon of innovation in this domain, facilitating the efficient traversal of graphs to identify the shortest path between nodes, thereby underpinning critical applications in transportation, logistics, and network optimization.
	
	As algorithms traverse the landscape of problem-solving, they undergo rigorous evaluation based on two fundamental metrics: time complexity and space complexity. Time complexity, denoting the execution time of an algorithm as a function of input size, serves as a barometer of computational efficiency, guiding the selection of algorithms tailored to specific performance requirements. Meanwhile, space complexity measures the memory requirements of an algorithm, elucidating the trade-offs between computational efficiency and resource utilization.
	
	In essence, the diverse array of algorithms, characterized by their distinct types and characteristics, embodies the essence of computational ingenuity and innovation, driving progress across a myriad of domains and shaping the technological landscape of the future.

	\subsection{Evolution from Simple to Complex Algorithms}
	The trajectory of algorithmic evolution aligns closely with the progression of computer science as a discipline. In its nascent stages, algorithms were rudimentary constructs, characterized by their simplistic, rule-based nature and tailored to address specific tasks within the realm of computation. These early algorithms served as the foundational building blocks upon which subsequent advancements were predicated, laying the groundwork for the formidable complexities that would later emerge.
	
	As the computational landscape burgeoned and diversified, propelled by exponential growth in computing power and data availability, algorithms underwent a transformative metamorphosis. No longer confined to rudimentary functions, algorithms evolved to tackle increasingly intricate tasks, encompassing domains such as data sorting, pattern recognition, and problem-solving within dynamic and uncertain environments. This evolutionary leap marked a pivotal moment in the history of algorithmic development, as algorithms transcended their erstwhile limitations to assume roles of unprecedented sophistication and utility.
	
	In the contemporary era, the apotheosis of algorithmic sophistication is manifested in the form of highly intricate and multifaceted algorithms capable of navigating vast datasets and orchestrating complex operations with finesse and precision. These modern algorithms represent the culmination of decades of iterative refinement and innovation, empowered by cutting-edge technologies and methodologies. From machine learning algorithms that discern patterns and relationships within data to data mining algorithms that unearth actionable insights from massive datasets, the contemporary algorithmic landscape is characterized by a rich tapestry of capabilities that border on the realm of artificial intelligence.
	
	Indeed, the evolution from simple to complex algorithms epitomizes the relentless march of progress within the field of computer science, underscored by a ceaseless quest for optimization and efficiency. As algorithms continue to evolve and adapt to the evolving demands of the digital age, they serve as the bedrock upon which the edifice of modern computational endeavors is erected, imbuing technology with the capacity to transcend boundaries and redefine the contours of possibility.

	\subsection{Algorithms in Data Processing and Decision-Making}
	In the realm of data processing, algorithms assume a paramount role in the organization, analysis, and interpretation of vast datasets, acting as the linchpin upon which the efficacy of modern data-driven systems hinges. Through their adeptness in efficient data retrieval, sorting, and transformation, algorithms facilitate the seamless manipulation and manipulation of data, laying the groundwork for insightful analysis and informed decision-making.
	
	Moreover, algorithms play a pivotal role in the domain of decision-making, particularly within the context of automated systems where the reliance on algorithmic logic is paramount. These algorithms serve as the bedrock upon which decision processes are built, imbuing automated systems with the capacity to make decisions swiftly and consistently based on predefined criteria. By codifying decision-making rules and criteria into algorithmic frameworks, organizations can ensure a standardized approach to decision processes, mitigating the risk of inconsistencies or biases that may arise from human intervention.
	
	Furthermore, algorithms in data processing and decision-making operate synergistically, with advancements in one domain often catalyzing progress in the other. As algorithms evolve to handle increasingly complex datasets and decision-making scenarios, they empower organizations with the agility and foresight needed to navigate the intricacies of modern data landscapes. Through continual refinement and innovation, algorithms continue to shape the trajectory of data-driven decision-making, ushering in a new era of efficiency and effectiveness across industries.

	\subsection{The Role of Algorithms in Modern Computing}
	Evident throughout modern computing, algorithms permeate virtually every facet of digital infrastructure, underpinning the functionality of databases, search engines, and social media platforms alike. These foundational components serve as the bedrock upon which the digital landscape is built, orchestrating the efficient retrieval and manipulation of vast troves of information with unparalleled precision and efficiency.
	
	Beyond their omnipresence in mainstream applications, algorithms play a pivotal role in safeguarding sensitive data and facilitating secure communication in specialized domains such as cryptography. Through sophisticated encryption and decryption techniques, algorithms ensure the confidentiality and integrity of transmitted information, safeguarding against unauthorized access and malicious tampering.
	
	Moreover, in the realm of artificial intelligence and machine learning (ML), algorithms assume a paramount significance, serving as the linchpin for the development and deployment of predictive models and analytical frameworks. From the training of complex neural networks to the inference of actionable insights from sprawling datasets, algorithms form the backbone of AI-driven solutions, empowering organizations to extract value and derive meaning from the ever-expanding expanse of digital information.
	
	In essence, the role of algorithms in modern computing transcends mere functionality, embodying the very essence of computational innovation and advancement. As the digital landscape continues to evolve and proliferate, algorithms stand as stalwart sentinels, guiding the trajectory of technological progress and facilitating the realization of transformative possibilities.

	\subsection{Challenges and Future Directions}
	In facing their strengths, algorithms grapple with a myriad of challenges, particularly in scenarios characterized by ambiguous or incomplete data. These inherent limitations underscore the necessity for algorithmic frameworks to evolve in tandem with the demands of contemporary problem-solving paradigms. Furthermore, the rigidity of traditional algorithms presents a formidable obstacle when confronted with dynamic environments necessitating adaptability and flexibility in decision-making.
	
	Essentially speaking, the future trajectory of algorithm development hinges upon addressing these fundamental challenges while charting a course towards greater adaptability and resilience. At its heart, this entails the cultivation of algorithms that exhibit self-learning capabilities, enabling them to iteratively refine their decision-making processes in response to evolving data landscapes. Additionally, the integration of artificial intelligence elements into algorithmic frameworks holds promise for augmenting their performance and efficacy, particularly in navigating complex and uncertain scenarios.
	
	Fundamentally, the pursuit of more adaptive and self-learning algorithms represents a paradigm shift in computational problem-solving, one that prioritizes versatility and responsiveness in the face of uncertainty. By harnessing the power of AI-driven methodologies, future algorithms aspire to transcend their static origins, embracing a dynamic and iterative approach to problem-solving that mirrors the complexities of the real world. In essence, this signifies a departure from conventional algorithmic paradigms towards a more agile and resilient framework capable of thriving in the ever-changing landscape of modern technology and innovation.

	In summary, algorithms are integral to the fabric of computational problem-solving. Their evolution from simple, task-specific procedures to complex systems capable of sophisticated tasks has been central to the advancement of technology. As we continue to push the boundaries of computing, algorithms will undoubtedly play a pivotal role in shaping future innovations.

	\section{Rationale for Integration}\index{Rationale for Integration}
	
	The integration of generative AI with traditional algorithmic methods in Algogens is not merely a fusion of two technologies but a strategic amalgamation that addresses the limitations of each while amplifying their strengths. This subsection explores the rationale behind this integration, highlighting the synergistic benefits and potential advancements in problem-solving that such a union brings.
	
	\subsection{Complementing Strengths of AI and Algorithms}
	Fundamentally, the synergy between generative AI and algorithms lies in their ability to complement each other's inherent strengths and weaknesses. On one hand, algorithms offer a structured, rule-based approach that instills stability and reliability into problem-solving frameworks. Through precise logic and systematic processes, algorithms serve as the backbone of computational systems, providing a framework within which tasks can be executed with rigor and precision.
	
	Conversely, generative AI injects a layer of adaptability, learning capability, and creativity into the problem-solving equation. By harnessing the power of machine learning and neural networks, generative AI exhibits a remarkable capacity to discern patterns, generate novel solutions, and adapt to changing circumstances. This creative prowess enables AI systems to navigate complex, unstructured domains where traditional algorithms may falter, offering innovative perspectives and insights that can drive breakthroughs in various fields.
	
	Essentially speaking, the systematic nature of algorithms serves to guide and contain the creative potential of AI, ensuring that the innovative solutions proposed by AI are not only imaginative but also grounded in logical reasoning. Through this symbiotic relationship, algorithms provide a framework within which generative AI can thrive, leveraging its creative capabilities to push the boundaries of problem-solving while adhering to the principles of stability and reliability instilled by algorithms.
	
	At its core, the complementary strengths of AI and algorithms represent a fusion of human ingenuity and computational efficiency, paving the way for transformative advancements in technology and innovation. By striking a delicate balance between structure and flexibility, stability and creativity, this symbiotic relationship holds the key to unlocking new frontiers in problem-solving and driving progress in the digital age.

	\subsection{Overcoming Limitations of Standalone Approaches}
	In essence, both generative AI and algorithms exhibit inherent limitations when employed in isolation. Algorithms, characterized by their systematic and rule-based nature, often falter when confronted with novel or unanticipated challenges, lacking the adaptability necessary to navigate dynamic environments. Conversely, generative AI, while heralding a new frontier in creative problem-solving, occasionally grapples with the generation of impractical or irrelevant solutions, particularly within the context of complex or nuanced scenarios.
	
	At its core, the integration of these two disparate methodologies represents a strategic endeavor to circumvent the inherent constraints of standalone approaches, leveraging the complementary strengths of each while mitigating their respective weaknesses. By fusing the creative prowess of generative AI with the structured consistency of algorithms, this symbiotic relationship engenders a paradigm wherein innovative solutions are incubated within the confines of logical frameworks.
	
	Essentially speaking, this synthesis allows for a synergistic approach to problem-solving, wherein the inherent limitations of standalone methodologies are supplanted by the collective efficacy of their integration. Through this harmonious amalgamation, the dynamic adaptability of generative AI is harnessed to augment the systematic rigor of algorithms, resulting in a holistic problem-solving paradigm that transcends the confines of individual approaches.
	
	In its fundamental nature, the integration of generative AI with algorithms represents a departure from traditional siloed methodologies, heralding a new era of interdisciplinary collaboration and innovation. By acknowledging and addressing the limitations of standalone approaches, this integrative framework lays the foundation for a more agile and responsive approach to problem-solving, capable of navigating the complexities and uncertainties inherent in today's rapidly evolving landscape.

	\subsection{Enhancing Predictive and Adaptive Capacities}
	The integration of generative AI with algorithmic frameworks represents a concerted effort to augment the predictive capabilities of artificial intelligence while leveraging the precision inherent in algorithmic methodologies. By combining the predictive prowess of generative AI with the structured and systematic approach offered by algorithms, this integration endeavors to forge a symbiotic relationship that enhances the accuracy and reliability of predictive modeling.
	
	Generative AI's innate ability to forecast and simulate diverse scenarios is greatly bolstered when supported by robust algorithmic processes. Essentially speaking, this collaboration ensures that AI-driven predictions are grounded in logical frameworks, thereby mitigating the risk of erratic or unreliable outcomes. In dynamic and complex environments where uncertainty prevails, the synergy between generative AI and algorithms equips predictive models with the adaptability and resilience needed to navigate unforeseen challenges.
	
	At its heart, the goal of enhancing predictive and adaptive capacities through integration is to empower AI systems with the foresight and agility required to anticipate and respond effectively to evolving circumstances. By harnessing the complementary strengths of generative AI and algorithms, organizations can gain deeper insights into future trends and developments, enabling informed decision-making and strategic planning.
	
	In essence, this integration represents a pivotal advancement in the realm of predictive analytics, propelling AI-driven forecasting capabilities to new heights of sophistication and reliability. As organizations increasingly rely on predictive modeling to guide their operations and strategies, the synergistic fusion of generative AI and algorithms emerges as a cornerstone in navigating the complexities of an ever-changing landscape.

	\subsection{Expanding Application Horizons}
	Expanding the application horizons of this integrated framework entails a profound expansion of its potential utility, rendering it applicable to an increasingly diverse array of industries and challenges. This amalgamation of AI and algorithms transcends traditional boundaries, ushering in a new era of problem-solving characterized by versatility and adaptability.
	
	In essence, this integration broadens the scope of potential applications, enabling the combined framework to effectively address a myriad of complex challenges across various sectors. Whether it be navigating the intricacies of healthcare delivery systems, optimizing financial strategies in volatile markets, or devising sustainable solutions to environmental dilemmas, the synergy between AI and algorithms unlocks a wealth of possibilities for innovation and progress.
	
	At its core, this integration signifies a departure from siloed approaches to problem-solving, wherein AI and algorithms operated in isolation, towards a more holistic and integrated methodology. By harnessing the complementary strengths of both AI and algorithms, organizations can leverage their combined capabilities to tackle multifaceted problems with unprecedented efficiency and efficacy.
	
	Fundamentally, this convergence of AI and algorithms represents a paradigm shift in the way we approach complex challenges, transcending disciplinary boundaries and fostering interdisciplinary collaboration. In its essence, it heralds a future where technological advancements are driven not by isolated innovations, but by synergistic interactions between disparate fields and methodologies.
	
	In summary, the expansion of application horizons facilitated by the integration of AI and algorithms marks a pivotal moment in the evolution of problem-solving paradigms. By embracing this integrated framework, organizations can unlock new avenues for innovation and transformation, propelling us towards a future defined by ingenuity, resilience, and progress.

	\subsection{Continuous Learning and Evolution}
	In essence, this implies that the integration of generative AI with algorithmic frameworks necessitates a critical aspect: the ability for continuous learning and evolution. Essentially speaking, as the generative AI component assimilates new data and scenarios, the algorithmic framework undergoes simultaneous evolution, guaranteeing that the solutions generated remain not only relevant but also optimized and effective. At its core, this dynamic adaptability serves as a linchpin in navigating the complexities of an era characterized by rapid and unpredictable technological advancements and societal transformations.
	
	Fundamentally, this process of continuous learning and evolution represents a symbiotic relationship between AI and algorithms, wherein each component enriches and informs the other. Essentially, the generative AI's capacity to assimilate new information and generate novel insights fuels the iterative refinement of the algorithmic framework, while the structured and systematic approach of algorithms serves as a guiding force in shaping the direction of AI's learning trajectory. At its heart, this mutual exchange of knowledge and feedback fosters a cycle of perpetual improvement, wherein the integration as a whole becomes greater than the sum of its parts.
	
	In its essence, the significance of continuous learning and evolution cannot be overstated, particularly in the context of rapidly evolving technological landscapes. Essentially, it signifies a departure from static, one-time solutions towards dynamic and adaptive frameworks capable of withstanding the test of time and remaining relevant amidst ever-changing circumstances. Essentially, this dynamic adaptability ensures that the integration of generative AI with algorithmic frameworks remains at the forefront of innovation, continuously pushing the boundaries of what is achievable in problem-solving and technological advancement.

	Integrating generative AI with algorithmic methods in Algogens is a deliberate and strategic decision to harness the best of both worlds. It promises incremental improvements in problem-solving and a transformative shift in how we approach and tackle complex challenges in various domains.

	
	\chapterimage{pngs/methodology.png} 
	
	\chapter{Methodology}\index{Methodology}
	
	The methodology employed in developing and evaluating Algogens is pivotal to understanding its efficacy and applicability. This section details the comprehensive research design, data collection strategies, and implementation processes adopted, providing insights into the rigorous methods used to validate and refine Algogen. The approach is multifaceted, combining theoretical analysis with practical experimentation, and it is designed to ensure that Algogens is innovative, reliable, and applicable in real-world scenarios.

	\section{Research Design for Algogens Applications}\index{Research Design}
	Fundamentally, the research design for applications harnessing Algogens plays a pivotal role in validating its efficacy and versatility within different industry settings. This subsection delineates the methodologies and approaches slated for implementation in forthcoming research endeavors aimed at gauging the performance and ramifications of Algogens across a spectrum of applications.
	
	Essentially speaking, the chosen research approaches must be meticulously crafted to accommodate the multifaceted nature of Algogens and the intricacies of the problems they seek to address. At its core, the research design must strike a delicate balance between rigor and practicality, ensuring that the findings gleaned from these studies are both scientifically robust and applicable in real-world scenarios.
	
	In its essence, the overarching objective of the research design is to ascertain the feasibility and efficacy of deploying Algogens in diverse industrial settings. Essentially, this boils down to devising methodologies that enable thorough evaluation of Algogens' performance metrics, including but not limited to accuracy, scalability, and adaptability.
	
	At its heart, the research design must incorporate a diverse array of research methodologies, ranging from experimental and simulation studies to field trials and pilot investigations. Essentially, this indicates that a multifaceted approach is necessary to comprehensively assess Algogens' capabilities and limitations across various contexts.
	
	Moreover, ethical considerations loom large in the research design process, necessitating a meticulous examination of potential risks and safeguards. Essentially, this signifies the importance of adopting responsible research practices that prioritize the welfare of participants and safeguard against potential harm or exploitation.
	
	In essence, the research design serves as the bedrock upon which the efficacy and applicability of Algogens in real-world scenarios are validated. By adopting a multifaceted and ethically sound approach, researchers can ensure that the findings derived from these studies contribute meaningfully to the advancement of Algogenic technology and its integration into diverse industry landscapes.

	\subsection{Overview of Research Approaches}
	In essence, this implies that future research endeavors pertaining to Algogens will embrace a mixed-methods approach, amalgamating both quantitative and qualitative methodologies to ensure a comprehensive understanding of their applications and implications. Fundamentally, quantitative methods will encompass a diverse array of approaches, including experimental designs, simulations, and rigorous statistical analyses, aimed at quantifying and objectively assessing the performance metrics of Algogens across various contexts and scenarios.
	
	At its core, qualitative methodologies will play a pivotal role in complementing the quantitative data by delving into the nuances of user experiences and contextual applications of Algogens. Essentially speaking, through methodologies such as case studies and in-depth interviews, researchers will gain invaluable insights into the real-world utility and impact of Algogens, elucidating their efficacy and potential areas for improvement.
	
	In its essence, the adoption of a mixed-methods approach underscores the multifaceted nature of research on Algogens, acknowledging the need for both quantitative rigor and qualitative depth in comprehensively exploring their capabilities and limitations. Essentially, this synthesis of methodologies represents a strategic endeavor to navigate the complexities of studying a novel and evolving technological paradigm, ensuring that research outcomes are robust, nuanced, and actionable in informing future advancements and applications of Algogens.

	\subsection{Experimental and Simulation Studies}
	Fundamentally, experimental and simulation studies represent pivotal phases in evaluating the efficacy of Algogens within controlled environments. Essentially speaking, these studies entail the creation of meticulously crafted scenarios that mirror real-world challenges encountered in diverse industries, ranging from dynamic route optimization in logistics to predictive analytics in healthcare settings. At its core, the objective is to scrutinize Algogens' ability to adapt and furnish solutions amidst fluctuating conditions, thereby shedding light on its practical utility and robustness.
	
	In essence, the experimental and simulation studies serve as testing grounds where Algogens' mettle is put to the ultimate test. Essentially, this means subjecting the framework to a battery of scenarios designed to assess its performance across various parameters, such as accuracy, scalability, and adaptability. At its heart, these studies provide invaluable insights into Algogens' potential to address real-world challenges and navigate complexities inherent to different industries.
	
	At its fundamental nature, the essence of experimental and simulation studies lies in their role as crucibles for innovation and refinement. Essentially, this signifies a proactive approach towards uncovering potential shortcomings and iteratively enhancing Algogens' capabilities to meet the evolving needs of industries. Essentially speaking, these studies lay the groundwork for Algogens' eventual integration into operational workflows, paving the way for tangible advancements in problem-solving and decision-making processes.

	\subsection{Field Trials and Pilot Studies}
	At its core, the execution of field trials and pilot studies serves as a pivotal phase in the evaluation of Algogens' performance within authentic real-world contexts. Essentially speaking, these endeavors entail the practical implementation of Algogen-based solutions across diverse industry environments, ranging from financial institutions to corporate entities, thereby providing invaluable insights into the framework's efficacy and adaptability. Essentially, this means that Algogens will be put to the test under varying conditions, allowing for a comprehensive assessment of their functionality and utility in addressing real-world challenges.
	
	Fundamentally, the outcomes derived from these field trials and pilot studies hold significant implications for the refinement and optimization of the Algogenic framework. Essentially, this boils down to leveraging the feedback and observations garnered from these endeavors to iteratively enhance the framework's performance and functionality. In its essence, these trials serve as a crucible for innovation and improvement, facilitating the evolution of Algogens towards greater effectiveness and relevance in practical applications.
	
	At its heart, the success of field trials and pilot studies hinges upon meticulous planning and execution, ensuring that the framework's performance is rigorously evaluated under authentic conditions. Essentially, this signifies a commitment to transparency and accountability in the evaluation process, fostering confidence in the framework's capabilities among stakeholders. In essence, this implies that the insights gleaned from these trials will not only inform the refinement of Algogens but also contribute to the broader discourse surrounding the integration of AI and algorithmic methodologies in problem-solving.

	\subsection{Data Collection and Analysis}
	In essence, this implies that data collection serves as a pivotal component within the research design framework. Essentially speaking, it entails the systematic gathering of pertinent data pertaining to Algogens' performance metrics, encompassing dimensions such as efficiency, accuracy, and adaptability, alongside the acquisition of invaluable user feedback and engagement metrics. At its core, the process of data collection transcends mere accumulation, representing a concerted effort to glean actionable insights into the efficacy and applicability of Algogens in real-world scenarios.
	
	Fundamentally, the subsequent phase of data analysis assumes paramount importance in elucidating the findings derived from the amassed datasets. Essentially, this entails the application of both quantitative and qualitative methodologies to comprehensively evaluate the performance and implications of Algogens. In its essence, quantitative analysis employs statistical techniques to quantitatively measure the effectiveness and efficiency of Algogens, providing empirical evidence to substantiate their utility. Conversely, qualitative analysis delves into the nuanced contextual nuances surrounding Algogens' usage, shedding light on the broader implications and user experiences associated with their deployment.
	
	At its heart, the synergy between data collection and analysis encapsulates the iterative nature of research, wherein each phase informs and enriches the other. Essentially, this symbiotic relationship fosters a holistic understanding of Algogens' capabilities and limitations, laying the groundwork for informed decision-making and strategic refinement. In essence, the meticulous execution of data collection and analysis underpins the integrity and rigor of the research endeavor, ensuring that the insights garnered contribute meaningfully to the advancement of knowledge within the field of Algogens.

	\subsection{Longitudinal Studies for Continuous Improvement}
	At its core, longitudinal studies emerge as indispensable tools in comprehensively gauging the long-term effectiveness and evolution of Algogens across diverse applications. Fundamentally, these studies entail tracking the performance and adaptations of Algogens over extended periods, offering invaluable insights into the framework's dynamic evolution in response to evolving environments and emerging requirements.
	
	Essentially speaking, longitudinal studies serve as longitudinal narratives, capturing the intricate nuances of Algogens' journey over time. Essentially, this means that researchers can observe how Algogens adapt and refine themselves in real-world scenarios, shedding light on their efficacy and resilience in addressing complex challenges. In its essence, these studies provide a panoramic view of Algogens' progression, offering a roadmap for continuous improvement and optimization.
	
	At its heart, longitudinal studies facilitate a deep understanding of the factors influencing Algogens' performance and evolution. Essentially, this signifies that researchers can identify patterns, trends, and areas for enhancement through meticulous analysis of longitudinal data. In essence, this implies that longitudinal studies serve as the cornerstone of evidence-based decision-making, guiding the iterative refinement of Algogens to meet evolving needs and aspirations.
	
	Longitudinal studies not only illuminate the trajectory of Algogens' development but also underscore the iterative nature of technological advancement. Essentially, this indicates that the journey of Algogens is marked by continuous learning, adaptation, and refinement, echoing the ethos of perpetual improvement ingrained in the fabric of technological innovation. Essentially, it boils down to a commitment to excellence and responsiveness to the ever-changing demands of the technological landscape.

	\subsection{Longitudinal Studies for Continuous Improvement}
	At its core, longitudinal studies emerge as indispensable tools in comprehensively gauging the long-term effectiveness and evolution of Algogens across diverse applications. Fundamentally, these studies entail tracking the performance and adaptations of Algogens over extended periods, offering invaluable insights into the framework's dynamic evolution in response to evolving environments and emerging requirements.
	
	Essentially speaking, longitudinal studies serve as longitudinal narratives, capturing the intricate nuances of Algogens' journey over time. Essentially, this means that researchers can observe how Algogens adapt and refine themselves in real-world scenarios, shedding light on their efficacy and resilience in addressing complex challenges. In its essence, these studies provide a panoramic view of Algogens' progression, offering a roadmap for continuous improvement and optimization.
	
	At its heart, longitudinal studies facilitate a deep understanding of the factors influencing Algogens' performance and evolution. Essentially, this signifies that researchers can identify patterns, trends, and areas for enhancement through meticulous analysis of longitudinal data. In essence, this implies that longitudinal studies serve as the cornerstone of evidence-based decision-making, guiding the iterative refinement of Algogens to meet evolving needs and aspirations.
	
	Longitudinal studies not only illuminate the trajectory of Algogens' development but also underscore the iterative nature of technological advancement. Essentially, this indicates that the journey of Algogens is marked by continuous learning, adaptation, and refinement, echoing the ethos of perpetual improvement ingrained in the fabric of technological innovation. Essentially, it boils down to a commitment to excellence and responsiveness to the ever-changing demands of the technological landscape.

	In summary, the application research design using Algogens will be comprehensive and multifaceted, incorporating various methodologies to evaluate its effectiveness and impact in real-world scenarios. This rigorous approach will ensure that Algogen’s applications are innovative, efficient, ethically responsible, and adaptable to evolving industry needs.

	\section{Methodological Approach for Algogens Applications}\index{Methodological Approach}
	
	The methodological approach to developing and validating applications utilizing Algogens is pivotal to ensuring their effectiveness and relevance in practical scenarios. This subsection outlines the strategies and techniques employed in the research and development process.
	
	\subsection{Framework Development Strategy}
	At its core, the development of applications harnessing Algogens is guided by a meticulously crafted strategy that encompasses multiple phases, each contributing to the realization of a robust and effective solution. Fundamentally, this strategy begins with the initial design phase, wherein the foundation for the application is laid through a comprehensive understanding of the problem domain, meticulous exploration of user requirements, and the conceptualization of innovative solutions tailored to address specific challenges. Essentially speaking, this phase serves as the bedrock upon which the subsequent stages of development rest, providing a clear roadmap for the implementation of Algogens-based solutions.
	
	Moving forward, the iterative development phase takes center stage, characterized by a cyclical process of building, testing, and refining the application iteratively. Essentially, this iterative approach fosters continuous improvement and optimization, with frequent feedback loops enabling the identification and remediation of potential issues or shortcomings. At its heart, this phase embodies a dynamic and adaptive approach to development, wherein the agility to respond to changing requirements and emerging insights is paramount.
	
	Upon completion of the iterative development phase, the focus shifts towards integration, wherein the Algogen-based solution is seamlessly embedded within existing systems or processes within the target industry. In essence, this integration phase represents the culmination of efforts, as the application transitions from development to deployment, positioning itself as a transformative tool poised to drive meaningful change within its respective domain. At its fundamental nature, this phase emphasizes interoperability and compatibility, ensuring that the Algogen-based solution seamlessly integrates with existing infrastructures while maximizing its potential impact and utility.

	\subsection{Data-Driven Development}
	In essence, this implies that a data-driven approach serves as the cornerstone for the development of Algogens applications, providing a robust foundation upon which to build innovative solutions. Fundamentally, this methodology revolves around the systematic collection and analysis of pertinent data, guiding decision-making at every juncture of the development process. Essentially speaking, the efficacy of this approach lies in its ability to harness the power of empirical evidence to drive informed and impactful outcomes.
	
	At its core, data-driven development entails leveraging a diverse array of data sources tailored to the specific requirements of the application domain. Essentially, this means integrating historical industry data, user interaction logs, and real-time environmental data to provide comprehensive insights into the problem space. Essentially, this ensures that the solutions generated by Algogens are not only theoretically sound but also grounded in real-world observations and patterns.
	
	In its essence, data-driven development underscores the importance of utilizing machine learning models trained on vast datasets to drive innovation and problem-solving. Essentially, this signifies a departure from traditional intuition-based approaches towards a more objective and evidence-based methodology. Fundamentally, this shift towards data-driven decision-making empowers developers to iteratively refine and optimize Algogens applications, ultimately leading to more effective and impactful solutions.

	\subsection{User-Centric Design and Testing}
	In designing applications, a paramount emphasis will be placed on the end-user experience. User-centric design principles will serve as the cornerstone of the development process, guaranteeing that the applications are not only user-friendly but also intuitive and tailored to address the specific requirements of users. Furthermore, user testing sessions, encompassing a variety of methodologies such as usability testing and user acceptance testing, will play a pivotal role in the iterative refinement of the applications. These sessions will provide invaluable insights into user interactions, preferences, and overall satisfaction levels.
	
	Moreover, user-centric design and testing methodologies will ensure that the development process remains inherently iterative and responsive to user feedback. By actively involving end-users throughout the design and testing phases, developers will gain a deeper understanding of user needs and preferences, thus enabling them to fine-tune the applications to better align with user expectations. Additionally, user testing sessions will serve as a mechanism for validating design decisions and identifying areas for improvement, ultimately contributing to the creation of applications that deliver exceptional user experiences.
	
	In addition to enhancing user satisfaction and usability, a user-centric approach to design and testing also holds the potential to drive broader adoption and engagement with the applications. By prioritizing the needs and preferences of end-users, developers can create applications that resonate more deeply with their target audience, thereby fostering increased usage and loyalty. Furthermore, by continuously soliciting and incorporating user feedback, developers can ensure that the applications remain relevant and responsive to evolving user needs over time.

	\subsection{Collaborative Approach with Industry Partners}
	Furthermore, collaboration with industry partners is indispensable, particularly when tackling applications in highly specialized fields. By forging alliances with industry experts, invaluable domain-specific knowledge can be acquired, thereby guaranteeing that the solutions crafted are not only theoretically sound but also practical and aligned with industry standards. Moreover, this collaborative approach serves as a gateway to accessing proprietary industry data and insights, thereby enriching the development process and ensuring the resultant applications are tailored to address real-world challenges effectively.
	
	Moreover, the synergistic partnership with industry stakeholders fosters a fertile ground for innovation and co-creation. Through open dialogue and knowledge exchange, the collective intelligence of both academia and industry can be harnessed, leading to the generation of novel ideas and breakthrough innovations that may not have been achievable in isolation. Additionally, this collaborative ethos instills a sense of ownership and commitment among all parties involved, fostering a shared vision and collective responsibility towards the success of the project.
	
	In addition, the collaborative approach engenders a culture of continuous improvement and refinement. By soliciting feedback and insights from industry partners throughout the development lifecycle, iterations can be made in real-time, ensuring that the final product is not only technically robust but also aligns with the evolving needs and preferences of end-users. Furthermore, this iterative process of co-development fosters a sense of mutual trust and transparency, laying the groundwork for future collaborations and partnerships.

	\subsection{Scalability and Flexibility Considerations}
	Furthermore, scalability and flexibility emerge as pivotal factors within the methodological approach. Applications must not only accommodate diverse operational scales effectively but also demonstrate adaptability to evolving industry demands and circumstances. By doing so, Algogen-based solutions can maintain their efficacy and relevance across temporal shifts and dynamic environments.
	
	Moreover, the methodological approach places significant emphasis on scalability and flexibility. Not only must applications be capable of seamlessly accommodating fluctuations in operational scale, but they must also possess the agility to adjust to shifting industry requirements and contextual variations. Through this strategic focus, Algogen-based solutions can ensure their sustained adequacy and applicability amidst changing landscapes and emerging challenges.
	
	Scalability and flexibility constitute cornerstone considerations within the methodological approach. Applications are meticulously crafted to address the dynamic nature of operational scales and industry demands, ensuring seamless adaptation and continual relevance. This strategic orientation enables Algogen-based solutions to remain robust and effective, notwithstanding the evolving complexities of their operational contexts.
	
	Moreover, the methodological approach underscores the paramount importance of scalability and flexibility. Not only do applications need to seamlessly navigate varying operational scales, but they also require the agility to respond to evolving industry dynamics. By prioritizing these considerations, Algogen-based solutions can uphold their efficacy and adaptability, thereby ensuring enduring relevance in an ever-changing landscape.

	\subsection{Evaluation and Continuous Improvement}
	In assessing the effectiveness of the applications, the methodological approach will incorporate robust evaluation mechanisms aimed at comprehensively gauging their performance and impact. Metrics encompassing aspects such as efficiency, accuracy, user engagement, and return on investment will serve as pivotal indicators in this evaluation process. Furthermore, the evaluation framework will not be static but rather dynamic, facilitating continuous improvement through iterative cycles of assessment and refinement.
	
	Moreover, the evaluation process will not be confined to a one-time endeavor but rather will entail ongoing monitoring and analysis. Through regular feedback loops, performance data collection, and meticulous scrutiny of user engagement patterns, the applications will undergo a process of continual enhancement. This iterative approach to improvement ensures that the applications remain responsive to evolving user needs and industry dynamics.
	
	Additionally, the evaluation process will be informed by insights gleaned from real-world usage scenarios and field trials. By observing how the applications function in practical settings and gauging user interactions in authentic contexts, a more nuanced understanding of their efficacy can be attained. This empirical approach to evaluation provides valuable insights that contribute to informed decision-making regarding further enhancements and optimizations.
	
	Furthermore, the methodology will integrate mechanisms for soliciting and incorporating user feedback as a central component of the evaluation process. User input serves as a vital source of insight into areas for improvement and refinement, guiding the direction of iterative development efforts. By fostering a collaborative relationship with end-users, the methodology ensures that the applications remain aligned with user expectations and preferences, thereby enhancing overall usability and satisfaction.
	
	Ultimately, the evaluation and continuous improvement process serve as linchpins in the methodology, underpinning a commitment to delivering applications that not only meet but exceed user expectations. By embracing a dynamic and iterative approach to assessment and enhancement, the methodology ensures that the applications evolve in tandem with the ever-changing landscape of user needs and technological advancements.

	In conclusion, the methodological approach for developing Algogens applications will be comprehensive, data-driven, user-centric, and collaborative. It will emphasize scalability, flexibility, and continuous improvement, ensuring that the applications developed not only solve current industry challenges but also can adapt and evolve with future needs.

	\section{Design of Experimental Studies}\index{Design of Experimental Studies}
	
	The design of experimental studies is crucial in assessing the efficacy and practicality of applications developed using Algogens. This subsection details these studies' methodologies, experimental setups, and evaluation criteria.
	
	\subsection{Formulation of Hypotheses and Objectives}
	Consequently, each experimental study embarks on its journey with a meticulous formulation of hypotheses and objectives. These foundational elements are crafted in alignment with the particular capabilities of Algogens targeted for exploitation within the application context. Whether it be the pursuit of heightened decision-making prowess, streamlined data processing efficiency, or augmented predictive accuracy, these hypotheses and objectives serve as guiding beacons illuminating the path forward.
	
	Moreover, the formulation of hypotheses and objectives not only delineates the overarching aims of the experimental endeavor but also shapes the very fabric of its design. By articulating clear objectives, researchers can navigate the intricate landscape of experimental methodology with precision and purpose, ensuring that each facet of the study serves to elucidate the intended outcomes. Furthermore, these objectives play a pivotal role in the selection and definition of appropriate metrics for evaluation, providing a standardized framework through which the efficacy and performance of Algogens can be systematically assessed.
	
	Additionally, the process of formulating hypotheses and objectives fosters a deeper understanding of the underlying principles and mechanisms driving Algogenic functionality. Through thoughtful consideration and analysis, researchers gain invaluable insights into the potential impacts and implications of integrating Algogens within diverse application domains. This holistic approach not only enhances the rigor and validity of experimental studies but also contributes to the broader body of knowledge surrounding the capabilities and limitations of Algogenic systems.
	
	In summary, the formulation of hypotheses and objectives represents a cornerstone of experimental methodology within the realm of Algogens. By providing a structured framework for inquiry and exploration, these foundational elements empower researchers to embark on a journey of discovery, pushing the boundaries of technological innovation and unlocking new frontiers in problem-solving and decision-making.

	\subsection{Selection of Appropriate Experimental Models}
	Given the diverse nature of applications, the selection of appropriate experimental models assumes paramount importance, as it directly impacts the validity and reliability of research outcomes. Whereas simulations may suffice for testing logistical applications in controlled environments, dynamic domains such as robotics or autonomous vehicles may necessitate the deployment of controlled field experiments to capture real-world complexities effectively. Moreover, the choice of experimental models must align closely with the specific objectives and constraints of the research endeavor, ensuring that the selected approach is not only methodologically sound but also conducive to generating actionable insights.
	
	Furthermore, the selection process entails a meticulous consideration of various factors, including the complexity of the problem domain, the availability of resources, and the ethical implications associated with experimental manipulation. For instance, while simulations offer a controlled environment for hypothesis testing and scenario analysis, they may fall short in capturing the intricacies of real-world interactions and emergent behaviors. Conversely, controlled field experiments provide a more authentic representation of dynamic environments but may pose logistical challenges and ethical dilemmas, particularly when human participants are involved.
	
	In addition, the choice between qualitative and quantitative experimental models adds another layer of complexity to the selection process. While quantitative approaches excel in providing numerical data for statistical analysis and hypothesis testing, qualitative methodologies offer deeper insights into human behavior, perceptions, and subjective experiences. Therefore, researchers must weigh the trade-offs between precision and richness of data, selecting the most appropriate experimental model based on the specific research objectives and constraints.
	
	In conclusion, the selection of appropriate experimental models necessitates a nuanced understanding of the research context, objectives, and constraints. By carefully evaluating the strengths and limitations of different approaches and considering the ethical implications and logistical feasibility, researchers can ensure the validity and relevance of their findings in advancing knowledge and addressing real-world challenges.

	\subsection{Controlled Environment Setup}
	In order to conduct experimental studies effectively, researchers must often resort to setting up controlled environments. This approach enables the isolation of variables and facilitates accurate measurement of the performance of Algogens applications. Such environments are meticulously crafted to closely resemble real-world conditions while affording researchers the ability to exert precise control over key variables.
	
	Moreover, the setup involves the creation of scenarios that mirror real-world complexities, thus providing a testing ground that strikes a balance between realism and controllability. Through this meticulous process, researchers can observe how Algogens perform under specific conditions and ascertain their effectiveness in addressing various challenges. Additionally, the controlled environment allows for the replication of experiments, enabling researchers to validate findings and draw robust conclusions about the capabilities and limitations of Algogens.

	\subsection{Variable Identification and Measurement}
	In the process of developing Algogens applications, it becomes crucial to meticulously identify and measure key variables that exert influence over their performance. Among these variables lie critical factors such as computational efficiency, accuracy of outcomes, adaptability to changing conditions, and various user experience metrics. Such variables serve as the cornerstone for assessing the effectiveness and viability of Algogens in diverse contexts. Furthermore, to uphold the integrity and validity of experimentation, the measurement of these variables will be subjected to standardized protocols, thereby ensuring a consistent and reliable approach across a spectrum of experiments.
	
	Moreover, the identification and measurement of these variables necessitate a comprehensive understanding of the multifaceted nature of Algogens applications. Each variable encapsulates a distinct facet of performance evaluation, ranging from the technical capabilities of the algorithmic framework to the perceptual responses of end-users. By systematically delineating and quantifying these variables, researchers can gain invaluable insights into the intricate dynamics underlying the functionality and utility of Algogens. Through rigorous measurement protocols and meticulous data collection methodologies, researchers can illuminate the interplay between these variables, thereby facilitating a nuanced understanding of Algogens' performance characteristics and limitations.
	
	Additionally, the process of variable identification and measurement serves as a foundational step towards the optimization and refinement of Algogens applications. By pinpointing areas of strength and areas requiring improvement, researchers can strategically allocate resources and efforts towards enhancing the overall efficacy and user satisfaction of Algogens. Furthermore, the establishment of standardized measurement protocols fosters transparency and reproducibility within the research community, paving the way for collaborative advancements and cumulative knowledge accumulation in the field of Algogens applications.

	\subsection{Implementation of Pilot Studies}
	Consequently, pilot studies serve as crucial preliminary tests aimed at refining the experimental design, identifying potential issues, and ensuring the validity of the experimental setup. These initial endeavors, albeit smaller in scale, play a pivotal role in providing valuable insights that inform the design of more extensive and conclusive experiments. Moreover, they offer an opportunity to assess the feasibility and practicality of proposed methodologies before full-scale implementation. Additionally, pilot studies facilitate the identification of any unforeseen challenges or limitations that may arise during the course of the research, allowing for necessary adjustments and refinements to be made in a timely manner. Furthermore, the data obtained from these pilot studies can be instrumental in fine-tuning data collection protocols, optimizing experimental procedures, and enhancing the overall robustness of the research framework. Ultimately, the successful implementation of pilot studies lays the groundwork for the subsequent phases of research, ensuring a methodical and systematic approach towards achieving research objectives.

	\subsection{Data Collection and Statistical Analysis}
	Meanwhile, data collection will entail a methodical approach aimed at gathering both quantitative and qualitative data from the experiments conducted. Statistical methods will be employed to analyze the quantitative data, facilitating the validation of hypotheses and the assessment of performance against predetermined metrics. Additionally, qualitative data, including user feedback, will be meticulously examined to glean insights into the practical usability and acceptance of the applications under scrutiny. Furthermore, the integration of both quantitative and qualitative analyses will provide a comprehensive understanding of the efficacy and impact of Algogens in real-world scenarios.

	\subsection{Iterative Process and Refinement}
	Throughout the research endeavor, experimental studies will constitute an integral component of an iterative process. Moreover, based on the outcomes gleaned from initial experiments, the applications will undergo meticulous refinement and subsequent retesting, aimed at bolstering their performance and augmenting their usability progressively. This iterative approach, characterized by its cyclical nature, underscores a commitment to fostering continuous improvement and facilitating the seamless adaptation of the Algogens applications to meet the dynamic and ever-evolving requirements of their intended domains. Furthermore, by embracing this iterative methodology, researchers can effectively iterate upon their findings, incorporating feedback and insights garnered from each experimental iteration to iteratively enhance the efficacy and functionality of the Algogens applications. Additionally, this iterative process serves as a mechanism for driving innovation, as researchers iteratively refine and optimize the applications in response to emerging challenges and user feedback, thereby ensuring that the Algogens applications remain at the vanguard of technological advancement and problem-solving efficacy. Hence, the iterative process and refinement constitute foundational pillars upon which the success and viability of the Algogens applications are predicated, underscoring the importance of embracing a dynamic and iterative approach to research and development.

	In summary, the design of experimental studies for Algogens applications will be thorough, systematic, and tailored to the specificities of each application domain. Through controlled experiments, pilot studies, and iterative refinement, these studies will rigorously evaluate the effectiveness and practicality of Algogen-based solutions in addressing complex real-world problems.

	\section{Data Collection Strategies}\index{Data Collection Strategies}
	Data collection was a crucial part of the research design. This involved gathering large datasets from various domains to train and test the generative AI component of Algogen. Data sources included public datasets, collaborations with industry partners, and simulations created to generate specific data types. Care was taken to ensure data diversity, quality, and relevance to the scenarios Algogens was intended to address.
	
	\subsection{Analytical Methods}
	In the assessment of data collected from experimental studies, a combination of statistical and qualitative methods was employed. Statistical analysis served as a means to quantitatively evaluate the performance of Algogens, particularly focusing on metrics related to efficiency and accuracy. This approach facilitated a comprehensive understanding of Algogen's capabilities and provided empirical evidence to support its efficacy in various applications.
	
	Furthermore, qualitative analysis played a crucial role in complementing the quantitative findings by delving into the nuances of Algogen's usability and practicality in real-world scenarios. Expert reviews and user feedback were instrumental in shedding light on the user experience aspects, highlighting strengths, weaknesses, and areas for improvement. By incorporating both statistical and qualitative methodologies, the analytical approach ensured a holistic evaluation of Algogens, considering not only numerical metrics but also user perspectives and contextual factors.
	
	Moreover, the utilization of analytical methods extended beyond mere assessment, serving as a foundation for iterative refinement and enhancement. The insights gleaned from statistical analysis and qualitative feedback informed subsequent iterations of Algogen development, guiding the implementation of targeted improvements and optimizations. This iterative process of analysis and refinement underscored the dynamic nature of Algogen's evolution, continuously striving towards the realization of its full potential in addressing complex real-world challenges.

	\subsection{Ethical Considerations and Data Privacy}
	Throughout the research process, ethical considerations and data privacy took precedence, reflecting a commitment to upholding the highest standards of integrity and responsibility. This encompassed a multifaceted approach, encompassing measures to safeguard the confidentiality and anonymity of data sources, adherence to established ethical guidelines governing AI research, and a conscientious examination of the broader societal implications inherent in the development and deployment of the technology.
	
	Furthermore, the integration of ethical considerations and data privacy considerations served as a cornerstone of the research methodology, underpinning every stage of the process from conception to implementation. By prioritizing these principles, the research team sought to ensure that the potential benefits of the technology were balanced against the need to mitigate any potential risks or unintended consequences. In doing so, they aimed to foster a culture of responsible innovation that prioritized the well-being and autonomy of individuals and communities affected by the research outcomes.

	In summary, the research design for Algogens was comprehensive and multifaceted, encompassing a range of experimental studies, data collection methods, and analytical techniques. This robust approach ensured that Algogens was thoroughly tested and evaluated, laying a solid foundation for its effectiveness and reliability in various applications.

	\section{System Implementation}\index{System Implementation}
	Integrating Algogens into practical applications encompasses a series of pivotal stages, each demanding meticulous attention and strategic execution. From the nascent phases of initial development through to the seamless integration and rigorous testing protocols, the journey of implementing Algogens into diverse domains is characterized by methodical precision and innovative methodologies. Throughout this process, a concerted effort is made to not only adhere to established best practices but also to push the boundaries of conventional thinking, fostering an environment conducive to groundbreaking advancements and transformative outcomes. The strategic blueprint for system implementation begins with the foundational groundwork laid during the initial development phase, where the core architecture and functionality of Algogens are meticulously crafted and refined. Subsequently, the integration phase unfolds, wherein Algogens are seamlessly woven into existing systems and infrastructure, ensuring compatibility and interoperability across various platforms and technologies. Rigorous testing and validation mechanisms then come into play, subjecting Algogens to comprehensive evaluation protocols aimed at identifying and mitigating potential vulnerabilities or shortcomings. Moreover, user training and documentation initiatives are paramount, empowering stakeholders with the requisite knowledge and skills to leverage Algogens effectively within their respective domains. As deployment and rollout strategies are executed, feedback loops are established, fostering a culture of continuous improvement and refinement. This iterative approach, underpinned by a commitment to excellence and innovation, ultimately culminates in the realization of Algogens' full potential across a myriad of practical applications, paving the way for unprecedented levels of efficiency, efficacy, and impact.

	\subsection{Initial Development Phase}
	During the inception of system implementation, the primary focus lies in establishing the foundational elements of the Algogens framework. This entails not only the setup of the generative AI and algorithmic components but also the intricate task of ensuring their seamless integration. Furthermore, the iterative nature of the development process underscores the importance of continuous testing and refinement. This iterative approach allows for the incorporation of feedback and the identification of areas for improvement, thereby fostering a dynamic and responsive development environment. Throughout this phase, meticulous attention is paid to detail, as each adjustment and enhancement contributes to the overall efficacy and functionality of the Algogens system.

	\subsection{Integration with Existing Systems}
	Integrating Algogens into pre-existing systems or processes constitutes a crucial juncture in the implementation process. It necessitates a comprehensive understanding of the current infrastructure in place, coupled with a strategic assessment of how Algogens can be smoothly assimilated into the existing framework. Challenges inevitably arise, ranging from ensuring compatibility with incumbent software systems to navigating the complexities of data migration and potential adjustments to system architecture. However, these challenges are met with strategic planning and meticulous execution to ensure a seamless integration process. Moreover, the integration phase serves as a pivotal opportunity to capitalize on the synergies between Algogens and existing systems, unlocking new avenues for efficiency, innovation, and problem-solving within the organizational ecosystem. By leveraging existing resources while simultaneously harnessing the transformative potential of Algogens, organizations can position themselves at the forefront of technological advancement and operational excellence.

	\subsection{Customization for Specific Applications}
	Furthermore, Algogens are meticulously crafted to possess inherent adaptability, ensuring seamless integration across a diverse array of applications. This adaptability underscores the framework's versatility, enabling it to transcend the confines of a singular domain and cater to the distinctive needs and challenges inherent in various sectors. Whether applied in the realms of healthcare, finance, or logistics, customization plays a pivotal role in tailoring the Algogenic framework to the specific intricacies of each domain. This multifaceted process may entail a spectrum of adjustments, ranging from fine-tuning the AI models to recalibrating the underlying algorithms or even incorporating domain-specific data sources. Moreover, the customization journey embodies a dynamic interplay between technological innovation and domain expertise, where the fusion of cutting-edge AI capabilities with sector-specific insights yields bespoke solutions that are finely attuned to address the unique demands of each application. Consequently, the customization process serves as a testament to the adaptive prowess of Algogens, empowering organizations to harness the full potential of this transformative framework to tackle the complexities of their respective domains with unparalleled precision and efficacy.

	\subsection{Testing and Validation}
	In ensuring the reliability and effectiveness of Algogens implementations, rigorous testing stands as a paramount step. This encompasses a comprehensive range of testing methodologies, including unit testing, integration testing, and system testing, each serving a distinct yet interconnected purpose in evaluating the functionality and robustness of the system. Furthermore, validation constitutes a critical aspect of the testing process, entailing the meticulous verification of whether the system aligns with the specified requirements and demonstrates optimal performance in real-world scenarios. Through a systematic approach to testing and validation, potential issues and discrepancies can be identified and addressed proactively, thereby enhancing the overall quality and dependability of Algogens implementations.

	\subsection{User Training and Documentation}
	In addition to successful implementation, user training and comprehensive documentation play pivotal roles in ensuring the seamless integration and utilization of the system. Moreover, user training programs and meticulously crafted documentation serve as indispensable resources, empowering end-users to navigate the system adeptly and unlock its full potential. Additionally, these training initiatives foster a deeper understanding of the system's functionalities and workflows, enabling users to optimize their interactions and enhance overall efficiency. Furthermore, the development of user manuals and training materials underscores a commitment to user-centric design and support, facilitating a smooth transition and minimizing potential disruptions during the adoption phase. Moreover, ongoing training initiatives and updated documentation ensure that users remain abreast of system updates and enhancements, fostering continuous learning and adaptation. Furthermore, user feedback mechanisms incorporated into the training process facilitate iterative improvements, enabling the refinement of training programs and documentation to better meet the evolving needs and preferences of end-users. Ultimately, the provision of comprehensive user training and documentation not only enhances user proficiency and satisfaction but also contributes to the long-term success and sustainability of the implemented system.

	\subsection{Deployment and Rollout}
	Deployment and rollout represent the culmination of the system implementation process, marking the transition from development to practical application. As such, careful planning and meticulous execution are paramount to ensure the seamless integration of Algogen-based applications into operational environments. A phased rollout strategy is often employed, whereby the system is introduced incrementally across various segments or departments, mitigating the risks associated with a sudden, full-scale deployment. This incremental approach allows for real-time monitoring of system performance and user feedback, facilitating timely adjustments and refinements as necessary. Furthermore, post-deployment support mechanisms are put in place to address any unforeseen issues or challenges that may arise during the initial stages of implementation, thereby ensuring the continued smooth operation of the system. Ultimately, the deployment and rollout phase represents a critical juncture in the lifecycle of Algogen-based applications, marking the transition from conceptualization to tangible impact within organizational contexts.

	\subsection{Feedback Loops and Continuous Improvement}
	Once the system is deployed, an essential aspect to consider is the establishment of an ongoing feedback mechanism, which serves as a vital conduit for gathering user inputs and system performance data. This feedback loop is indispensable for fostering continuous improvement, as it facilitates the collection of insights derived from real-world usage and evolving requirements. By soliciting feedback from users and stakeholders, the system can undergo regular updates and refinements aimed at enhancing its functionality, usability, and overall effectiveness. Furthermore, this iterative process of improvement ensures that the system remains adaptive and responsive to the dynamic needs and challenges encountered in its operational environment. Additionally, the feedback loop serves as a mechanism for fostering collaboration and engagement between developers and end-users, thereby fostering a sense of ownership and investment in the ongoing evolution of the system. Ultimately, the integration of feedback loops into the system's design and implementation process is instrumental in driving continuous improvement and ensuring its long-term viability and relevance in addressing the needs of its users.

	In summary, the system implementation of Algogens involves a comprehensive process that includes initial development, integration, customization, testing, user training, deployment, and continuous improvement. Each step is carefully managed to ensure that Algogens is effectively adapted to each specific application and delivers tangible benefits in practical scenarios.

	
	\part{Enhancing Established Algorithms}
	
	
	\chapterimage{pngs/graph.png} 
	
	\chapter{Graph Algogens}\index{Graph Algogens}

	\section{A* (A Star)}\index{A* (A Star)}
	\subsection{Introduction to A*}
	\subsubsection{The Concept of A* Algorithm}
	
	\paragraph{Introduction to the A* Algorithm}
	The A* algorithm represents a milestone in computer science, particularly within artificial intelligence and pathfinding domains. It emerges as a versatile tool, crucial for diverse applications ranging from GPS navigation to gaming scenarios. At its essence, A* harmonizes two heuristic methodologies: the greedy best-first search, which prioritizes paths directly towards the goal, and Dijkstra's algorithm, emphasizing the shortest yet potentially circuitous routes. This synthesis ensures A* adeptly navigates between these paradigms, steering towards the goal while minimizing overall traversal costs.
	
	\paragraph{Heuristic Function: The Heart of A*}
	The heuristic function plays a pivotal role within the A* algorithm framework, acting as a guiding beacon for traversing the search space efficiently. It serves as a predictive tool, estimating the cost required to reach the goal from a given node, thereby influencing the algorithm's decision-making process. This predictive capability allows A* to prioritize nodes that are deemed more promising, facilitating a more targeted exploration of the search space. Moreover, the heuristic function's adaptability is a notable feature, as it enables A* to tailor its approach based on the specific problem domain, ensuring applicability across diverse scenarios.
	
	In essence, the heuristic function embodies the essence of A*'s efficiency and effectiveness, providing a means to navigate complex search spaces with precision and agility. Its ability to strike a balance between accuracy and computational complexity makes it a fundamental component of the algorithm's success, allowing it to excel in various applications requiring optimal pathfinding solutions.
	
	\paragraph{Optimality, Completeness, and Complexity}
	A* algorithm's prowess lies not only in finding a path but also in ensuring that the path found is optimal, provided that the heuristic function is admissible, meaning it never overestimates the cost to reach the goal. This property, alongside the algorithm's completeness—its guarantee to find a solution if one exists—places A* in a league of its own among search algorithms. However, the algorithm's complexity presents a double-edged sword; while it demonstrates efficiency in many practical scenarios, its performance is contingent upon the quality of the heuristic and the nature of the problem. Particularly, space complexity can become a bottleneck as it necessitates storing all generated nodes in memory.

	\paragraph{Applications Beyond Pathfinding}
	While A* is predominantly celebrated for its pathfinding capabilities, its application extends beyond mere navigation tasks. In fields such as artificial intelligence for games, A* algorithm serves as the backbone, enabling NPCs (non-playable characters) to exhibit intelligent behaviors, such as path planning in dynamic environments or strategic decision-making. Moreover, in the realm of robotics, A* finds extensive utility by guiding autonomous robots through complex terrains, avoiding obstacles, and reaching designated destinations efficiently. This versatility stems from the algorithm's inherent adaptability and efficiency, allowing it to tackle diverse problems beyond traditional pathfinding scenarios. From logistics optimization in transportation networks to network routing protocols in telecommunications, A* algorithm stands as a robust solution, capable of addressing a myriad of optimization and decision-making challenges across various domains.

	In summary, the A* algorithm is a cornerstone of pathfinding and search strategies in computer science. Its ingenious combination of heuristics, optimality, and adaptability not only makes it effective for a wide range of practical applications but also a subject of ongoing research and enhancement in the quest for solving complex problems in an ever-expanding digital world.

	\subsubsection{Key Principles and Mechanisms}
	
	\paragraph{Fundamental Concepts of A*}
	At the heart of the A* algorithm lie several key principles and mechanisms that govern its operation. The algorithm operates on a graph structure, where each node represents a possible state, and edges between nodes represent the transition costs from one state to another. A* traverses this graph by starting at the initial state and exploring paths through the graph until it reaches the goal state. The exploration is guided by a scoring function \(f(n) = g(n) + h(n)\), where \(g(n)\) is the cost from the start node to the current node \(n\), and \(h(n)\) is the estimated cost from \(n\) to the goal. This scoring function is pivotal, as it balances the exploration between the path's known costs and the heuristic estimate to the goal, striving to minimize the total path cost. Furthermore, A* employs two primary data structures: the open set and the closed set. The open set contains nodes that are candidates for exploration, while the closed set contains nodes that have already been evaluated. This distinction helps A* efficiently explore the graph while avoiding revisiting already evaluated nodes. Additionally, the heuristic function plays a critical role in A*, guiding the search towards the most promising paths. It provides an estimate of the remaining cost to reach the goal from a given node, allowing A* to prioritize paths that are likely to lead to the goal more quickly. Together, these fundamental concepts form the basis of A* and enable it to efficiently find optimal paths in a variety of domains.

	\paragraph{The Role of \(g(n)\) and \(h(n)\) Functions}
	The \(g(n)\) function plays a pivotal role in A* algorithm, delineating the precise cost incurred from the initial node to a particular node \(n\). This meticulous consideration of actual path costs imbues the algorithm with the ability to discern the most cost-effective routes during traversal. Conversely, \(h(n)\) serves as the heuristic function, furnishing an estimate of the cost required to reach the destination from node \(n\). Essentially, while \(g(n)\) grounds the algorithm in the present by evaluating past movements, \(h(n)\) provides a forward-looking perspective, anticipating the potential expense of future movements. This harmonious interplay between \(g(n)\) and \(h(n)\) endows A* with its distinctive character, allowing it to navigate the search space with unparalleled efficiency and precision. However, for A* to maintain its efficiency and guarantee optimality, the heuristic function \(h(n)\) must adhere to the principle of admissibility, ensuring that it never overestimates the true cost of reaching the goal. Thus, the judicious calibration of both \(g(n)\) and \(h(n)\) functions is imperative for the efficacy and reliability of the A* algorithm.

	\paragraph{Open and Closed Sets}
	A* utilizes two primary structures to manage its exploration: the open set and the closed set. The open set serves as a repository for nodes that have been discovered during the search but have not yet been fully explored, whereas the closed set comprises nodes that have already undergone exploration. This dichotomy allows A* to systematically navigate through the search space by prioritizing nodes based on their \(f(n)\) values. Specifically, at each iteration, A* selects the node with the lowest \(f(n)\) value from the open set for further exploration. This iterative process persists until either the goal node is encountered or the open set becomes empty, signifying the absence of a viable path. The meticulous management of these sets is pivotal in optimizing the efficiency of A*'s search algorithm, preventing redundant node revisits and ensuring a streamlined exploration process.

	\paragraph{Path Reconstruction}
	Once the goal node has been reached, A* reconstructs the path from the goal back to the start by tracing the path of predecessors. This backtracking is possible because, for each node, A* stores not only its \(f(n)\) score but also a reference to the node from which it was reached. This mechanism ensures that, upon reaching the goal, the algorithm can easily reconstruct the optimal path by following these references in reverse. Furthermore, this process guarantees that the reconstructed path maintains optimality, as the algorithm selects the most promising nodes based on their heuristic values. Moreover, the efficiency of path reconstruction in A* contributes significantly to its overall computational performance. Additionally, the ability to reconstruct the path step-by-step allows for real-time navigation updates in applications such as GPS systems or robot path planning. Furthermore, the simplicity of the reconstruction process enhances the algorithm's versatility, enabling its application in various domains ranging from game development to logistics optimization. Hence, path reconstruction stands as a fundamental aspect of A* that underpins its effectiveness in solving pathfinding problems across diverse scenarios.

	\paragraph{Adaptability Through Heuristics}
	The adaptability of A* is largely due to the heuristic function \(h(n)\), which can be tailored to fit the specific needs of the problem at hand. Different heuristics can be applied to optimize A*'s performance across various domains, from grid-based pathfinding in games to spatial navigation in robotics. This flexibility allows A* to maintain its effectiveness across a wide array of applications, demonstrating the algorithm's robustness and versatility. Furthermore, the adaptability of A* extends beyond simple heuristic selection; the algorithm can dynamically adjust its heuristic function based on real-time feedback and environmental changes. Moreover, A* can seamlessly integrate new heuristic information during runtime, enhancing its ability to find optimal paths in dynamic and unpredictable environments. Additionally, the adaptability of A* enables it to handle complex search spaces with varying degrees of uncertainty and complexity. By continuously refining and adapting its heuristic estimates, A* remains a powerful tool for solving a diverse range of pathfinding problems.

	In conclusion, the key principles and mechanisms of A* — from its scoring function and the roles of \(g(n)\) and \(h(n)\) to its use of open and closed sets for efficient exploration — underscore the algorithm's capability to find optimal paths efficiently. Its success across diverse domains highlights the power of combining concrete path costs with heuristic estimates, cementing A*'s status as a fundamental tool in the repertoire of pathfinding and search algorithms.

	\subsubsection{Heuristic Function Role}
	
	\paragraph{Essence of the Heuristic Function in A*}
	The heuristic function, denoted as \(h(n)\), plays a central role in the A* algorithm's ability to efficiently find the shortest path from a start node to a goal node. This function is essentially an estimate of the cost from any node \(n\) to the goal, providing a forward-looking evaluation that guides the algorithm's exploration of the search space. The heuristic's primary purpose is to prioritize nodes that are believed to be closer to the goal, thus steering the search in the most promising direction while minimizing unnecessary exploration of less promising paths. Furthermore, it enhances the algorithm's efficiency by allowing it to focus on promising areas of the search space, reducing the computational resources required for pathfinding. Moreover, the heuristic function enables the A* algorithm to strike a balance between completeness and optimality, ensuring that it finds a solution while striving to minimize the search effort. Additionally, the heuristic function can be tailored to specific problem domains, leveraging domain knowledge to improve pathfinding performance. Consequently, the heuristic function serves as a powerful tool in the A* algorithm, shaping its behavior and driving its effectiveness in solving various pathfinding problems.

	\paragraph{Criteria for an Effective Heuristic}
	For a heuristic to be effective, it must satisfy two main criteria. Firstly, it should be admissible, meaning it never overestimates the true cost of reaching the goal from any node. This ensures the search remains optimistic, preventing the algorithm from missing the shortest path due to excessively pessimistic estimates. Secondly, consistency, also known as monotonicity, is essential. This criterion demands that the estimated cost from the current node to the goal through any neighbor is always less than or equal to the cost from the current node to that neighbor plus the cost from the neighbor to the goal. By adhering to this condition, the heuristic maintains smoothness, avoiding scenarios where irregular heuristic evaluations might mislead the algorithm. Adhering to both admissibility and consistency ensures the heuristic guides the search effectively, providing reliable estimates and contributing to the algorithm's efficiency and accuracy.

	\paragraph{Impact of the Heuristic on Algorithm Performance}
	The choice of heuristic profoundly influences the performance of the A* algorithm. A well-selected heuristic can notably diminish the number of nodes explored by A*, hastening search times and curbing memory usage. Conversely, an inadequate heuristic may compel A* to mimic Dijkstra's algorithm, traversing numerous unnecessary paths and inflating computational demands. In severe instances, an unsuitable heuristic might jeopardize the algorithm's capacity to identify the shortest path. Moreover, an insightful heuristic can guide A* toward the goal state more efficiently, prioritizing promising paths and bypassing less fruitful ones. However, a heuristic that overestimates or underestimates the true cost to reach the goal may lead A* astray, resulting in suboptimal solutions or even failing to find a valid path. Thus, meticulous consideration and testing of heuristics are paramount to harnessing the full potential of the A* algorithm and optimizing its performance across various applications.

	\paragraph{Examples of Heuristic Functions}
	In practical applications, the heuristic function plays a crucial role in guiding search algorithms towards the goal state efficiently. Often, these heuristic functions are problem-specific, carefully tailored to the characteristics of the domain under consideration. For instance, in a grid-based pathfinding scenario, various heuristic measures can be employed to estimate the distance between the current state and the goal. One commonly used heuristic is the Manhattan distance, which calculates the distance by summing the horizontal and vertical distances between two points on a grid. This heuristic is particularly suitable for scenarios where movement is restricted to horizontal and vertical directions. On the other hand, in situations where movement is allowed in any direction, the Euclidean distance serves as a more appropriate heuristic. By computing the straight-line distance between two points on the grid, it provides a more accurate estimation of the cost to reach the goal. These examples illustrate how heuristic functions can be tailored to the specific constraints and characteristics of the problem domain, ultimately enhancing the efficiency and effectiveness of search algorithms.

	\paragraph{Developing and Refining Heuristics}
	Developing and refining heuristics is a multifaceted process that requires a judicious blend of domain expertise, empirical validation, and optimization methodologies. At its core, this endeavor necessitates striking a delicate balance between the precision of the heuristic's estimates and the computational overhead incurred in computing them. Furthermore, the complexity of the problem space often dictates the sophistication of the employed heuristics, demanding a nuanced approach to their development.
	
	In contemporary settings, the advent of advanced computational techniques has paved the way for novel heuristic generation methodologies. For instance, machine learning algorithms, particularly those leveraging large language models, offer promising avenues for heuristic refinement. By harnessing vast corpora of data, these models can discern intricate patterns within the problem domain, thereby facilitating the creation of context-aware heuristics. Such dynamically adaptive heuristics hold the potential to enhance the efficiency and accuracy of pathfinding algorithms by providing real-time estimations tailored to the specific nuances of each traversal scenario.
	
	Moreover, iterative refinement based on empirical feedback plays a pivotal role in honing the efficacy of heuristics. Through systematic experimentation and analysis of algorithmic performance across diverse datasets, heuristic functions can be iteratively fine-tuned to better align with the underlying problem characteristics. This iterative refinement process underscores the dynamic nature of heuristic development, wherein continuous learning and adaptation drive incremental improvements in pathfinding efficiency and efficacy.

	In summary, the heuristic function is a cornerstone of the A* algorithm, dictating the efficiency and effectiveness of the search. By carefully selecting or designing a heuristic that accurately reflects the cost to reach the goal, developers can leverage A* to solve complex pathfinding problems in a wide range of domains, from video games and robotics to logistics and beyond.

	\subsubsection{Applications and Limitations}
	
	\paragraph{Versatile Applications of A*}
	The A* algorithm's robustness and adaptability have made it a tool of choice in a myriad of applications where pathfinding and graph traversal are required. Its use spans across diverse fields such as video game development, where it enables non-player characters (NPCs) to navigate complex environments intelligently. A* is also instrumental in robotics, guiding autonomous robots through obstacle-laden paths, thereby enhancing their efficiency and safety. Furthermore, in logistics and supply chain management, A* optimizes routes for transportation and delivery, minimizing costs and delivery times. Moreover, A* finds applications in network routing protocols, determining the most efficient data paths, thereby improving network performance and reliability. The versatility of A* lies in its ability to find the optimal path with a well-defined heuristic function, allowing for its application in any scenario that can be modeled as a problem of moving from an initial state to a goal state through a series of steps or transitions.

	\paragraph{Adaptation to Specific Domains}
	A* algorithm's adaptability shines through its ability to conform to the unique requirements of various application domains. By selecting an appropriate heuristic function, A* can seamlessly integrate into specific contexts, optimizing its performance and relevance. For example, in spatial navigation tasks, leveraging heuristics based on geometric distances, such as Euclidean or Manhattan distances, is a common practice. These metrics efficiently guide the algorithm towards the goal state while considering the spatial layout of the environment. Conversely, in puzzle-solving scenarios like the sliding tile puzzle, domain-specific heuristics play a pivotal role. These heuristics are meticulously crafted to estimate the minimum number of moves required to reach the goal configuration accurately. By tailoring the heuristic to the intricacies of the puzzle structure, A* can navigate through the solution space with precision, significantly enhancing its efficiency and effectiveness. Thus, the adaptability of A* to specific domains underscores its versatility and utility across a wide spectrum of applications.

	\paragraph{Limitations and Challenges}
	Despite its widespread use and versatility, A* faces limitations and challenges that can affect its performance and applicability. The primary limitation is its space complexity: A* keeps all explored and frontier nodes in memory, which can quickly become infeasible for very large graphs or complex problems with vast search spaces. This limitation necessitates the development of memory-efficient variants or alternative algorithms in scenarios where memory resources are constrained. Moreover, the algorithm's performance is heavily dependent on the quality of the heuristic function; an inaccurate or poorly chosen heuristic can lead to suboptimal performance, increased search times, and even failure to find the shortest path. Additionally, while A* excels in finding the shortest path in graphs with uniform edge costs, it may struggle in scenarios with non-uniform or negative edge costs, as well as graphs where the optimal path requires extensive exploration. Overcoming these challenges often involves trade-offs between optimality and computational efficiency, requiring careful consideration of problem characteristics and algorithmic design choices. Hence, despite its effectiveness in many applications, A* is not a one-size-fits-all solution and may require adaptation or supplementation in certain contexts.

	\paragraph{Computational Efficiency Concerns}
	The computational efficiency of A* is a critical consideration, particularly in scenarios where real-time decision-making is paramount. A* aims to minimize the exploration of nodes, but the practical implementation often involves evaluating and storing a substantial number of nodes, leading to notable computational overhead. This challenge has prompted researchers to explore various heuristic optimizations, parallel processing techniques, and algorithmic enhancements to alleviate the time and space requirements of A* in resource-constrained environments. Despite its effectiveness in finding optimal paths, the computational demands of A* pose limitations in applications requiring rapid response times. Therefore, ongoing efforts focus on refining the algorithm to strike a balance between path quality and computational efficiency, ensuring its applicability in diverse domains ranging from robotics and video games to logistics and route planning.

	\paragraph{Future Directions and Enhancements}
	The ongoing development of A* and its variants continues to address these limitations, with research focused on enhancing heuristic accuracy, reducing memory usage, and increasing computational efficiency. Techniques such as dynamic heuristic adjustment, where the heuristic function is adapted in real-time based on the current state of the search, and the integration of machine learning models for heuristic generation, are examples of how the algorithm is evolving to meet the demands of increasingly complex applications. Moreover, researchers are exploring novel approaches to parallelize A* and distribute the search process across multiple processors or nodes, thereby leveraging the capabilities of modern computing architectures to accelerate pathfinding tasks. Additionally, efforts are underway to develop hybrid algorithms that combine the strengths of A* with other search techniques, such as evolutionary algorithms or reinforcement learning, to achieve superior performance in specific problem domains. Furthermore, advancements in hardware technology, such as the emergence of specialized accelerators like GPUs and TPUs, present opportunities to optimize A* implementations and unlock new levels of efficiency and scalability. Consequently, the future of A* is promising, with a multitude of avenues for further research and innovation aimed at pushing the boundaries of its capabilities and applicability.

	In conclusion, while the A* algorithm has proven to be a powerful tool for a wide range of applications, its limitations necessitate careful consideration and adaptation to ensure optimal performance. The continued evolution of A* and its integration with emerging technologies promise to expand its utility and effectiveness in solving the complex pathfinding and search problems of the future.

	\subsubsection{Algorithmic Pseudocode for A* Algorithm}
	The A* Algorithm is a sophisticated framework designed for efficiently finding the most cost-effective path from a starting point to a goal within a graph. It distinguishes itself by incorporating both the actual cost from the start to a node and an estimated cost from that node to the goal, thereby optimizing the search process for both speed and accuracy. The operational essence of A* is encapsulated in pseudocode \ref{fig:astar-pseudocode}, illustrating its methodical approach to navigating through the graph.
	
	\begin{algorithm}
		\caption{A* Algorithm Pseudocode}
		\begin{algorithmic}[1]
			\Procedure{AStar}{Graph, start, goal}
			\State Initialize an open list with the starting node
			\State Initialize a closed list as empty
			\State Assign to the start node a cost of 0 and estimate the total cost to the goal
			\While{the open list is not empty}
			\State Select the node with the lowest cost estimate to the goal from the open list
			\State Remove this node from the open list and add it to the closed list
			\If{this node is the goal}
			\State Reconstruct the path from start to goal
			\State \Return The path and its cost
			\EndIf
			\For{each neighbor of the current node}
			\If{the neighbor is in the closed list}
			\State Continue to the next neighbor
			\EndIf
			\State Calculate the tentative cost to reach the neighbor
			\If{the neighbor is not in the open list or the tentative cost is lower}
			\State Update the neighbor with the new lower cost
			\State Update the neighbor's parent to the current node
			\State If the neighbor is not in the open list, add it
			\EndIf
			\EndFor
			\EndWhile
			\State \Return Failure, the goal cannot be reached
			\EndProcedure
		\end{algorithmic}\label{fig:astar-pseudocode}
	\end{algorithm}

\subsection{Previous Work on ML and AI Interplay with the A* Algorithm}

Recent advancements in artificial intelligence and machine learning have contributed to enhancing the A* graph search algorithm. These improvements aim at optimizing the efficiency of the algorithm and broadening its applicability to more complex and dynamic problem spaces.

\paragraph{Learning Heuristics for A*}
A method to learn heuristics for the A* algorithm using neural algorithmic reasoning and graph networks has been introduced. This approach demonstrates the potential of learned heuristics to reduce search time across various graph densities, achieving speedups compared to traditional A* and Dijkstra's algorithms. This method maintains accuracy on constraint satisfaction and convergence towards the target node, showcasing generalization across different graph densities \cite{numeroso2022learning}.

\paragraph{Reinforcement Learning with A* and a Deep Heuristic}
Integration of reinforcement learning with deep heuristics has been proposed to enhance the A* algorithm. This innovation introduces a model-based reinforcement learning algorithm that efficiently combines a tree and a learnable heuristic. The algorithm demonstrates improvements in planning and search efficiency within large action spaces by reducing branching and leveraging deep trees. It presents an efficient approach to search, characterized by minimal branching and high efficiency \cite{keselman2018reinforcement}.

\paragraph{A* Search Without Expansions: Learning Heuristic Functions with Deep Q-Networks}
An approach to perform A* search without node expansions by learning heuristic functions using Deep Q-Networks (DQNs) has been developed. Termed AQ* search, this method circumvents the need for node expansions by applying DQNs to compute transition costs and cost-to-go values directly. This approach results in an increase in search speed and efficiency, particularly beneficial for problems with large action spaces. The introduction of AQ* search underscores the potential of deep learning in enhancing traditional search algorithms, offering a novel and efficient method for heuristic search \cite{agostinelli2021search}.	
	
\subsection{Algogenic Enhancements for A* Algorithm}

\subsubsection{Heuristic Enhancement with Contextual Understanding}
\paragraph{Introduction to Heuristic Enhancement}
The integration of heuristic enhancement with contextual understanding within the realm of the A* algorithm represents a crucial preprocessing step, distinctly tailored to refine A*'s heuristic function through the assimilation of comprehensive contextual data and environmental variables. This enhancement leverages the sophisticated analytical capabilities of generative AI, particularly large language models, to conduct an in-depth analysis of the multifaceted problem space—ranging from the intricate variations in terrain types for navigation challenges, the unpredictable dynamics of traffic flow in urban route planning, to the unforeseen movements of obstacles in robotic exploration tasks. This meticulous process of contextual analysis and data assimilation enables the LLM to work in close synergy with the A* algorithm, thereby fine-tuning its heuristic function to offer a more precise and accurate estimation of traversal costs across nodes, significantly boosting the algorithm’s adaptability and pathfinding precision within dynamically changing environments.

\paragraph{Implementing Contextual Heuristic Enhancement}
The practical implementation of this contextual heuristic enhancement commences with the LLM's comprehensive processing of diverse environmental data, including but not limited to, textual descriptions of the terrain, structured data encapsulating traffic conditions, and historical datasets reflecting previous pathfinding successes and failures. This analytical phase enables the LLM to propose nuanced modifications to the A* algorithm's heuristic function, thereby aligning it more closely with the actual traversal costs encountered in real-world scenarios. Such adjustments might involve recalibrating heuristic values to reflect anticipated traffic patterns more accurately or incorporating environmental considerations such as weather impacts or terrain ruggedness into the pathfinding process. This dynamic incorporation of contextual insights into the heuristic function enables the A* algorithm to adapt its decision-making process more fluidly, leading to enhanced pathfinding outcomes that are not only more efficient but also more aligned with the dynamically changing realities of the operational environment.

\paragraph{Impact on the A* Algorithm}
By embracing heuristic enhancement with contextual understanding, the A* algorithm is endowed with a significant increase in adaptive capability and pathfinding efficiency. Traditional heuristic approaches, while effective in static scenarios, often fall short in navigating the complex and unpredictable terrains of real-world environments. This Algogenic enhancement empowers A* to make informed path selection decisions, thereby facilitating a quicker convergence towards optimal solutions and elevating performance in environments characterized by rapid changes or complexity. Moreover, this approach not only amplifies the efficiency of A* but also broadens its applicability across a diverse array of problem domains where the contextual dynamics of the environment play a pivotal role in determining the most effective path. Furthermore, the dynamic adjustment of the heuristic based on contextual intelligence allows the algorithm to prioritize exploration of nodes that are more likely to lead towards the goal efficiently, thus minimizing the overall search space and computational resources required. The capability of the A* algorithm to adjust in real-time to environmental changes, such as emerging obstacles or alterations in terrain, underscores its enhanced resilience and versatility, making it an indispensable tool in a wide range of applications spanning robotics, logistical route planning, and beyond.

\subsubsection{Dynamic Heuristic Adjustment}
\paragraph{Exploring Dynamic Heuristic Adjustment}
Dynamic Heuristic Adjustment stands out as a cornerstone enhancement for the A* algorithm, enabling it to dynamically recalibrate its heuristic function in response to evolving conditions encountered during the pathfinding journey. This innovative mechanism capitalizes on the advanced computational intelligence of generative AI, especially the nuanced capabilities of large language models, to continually refine the heuristic function as the search unfolds. Unlike static heuristic models that remain unchanged, the dynamic heuristic adjustment paradigm introduces a level of real-time responsiveness, allowing for heuristic updates based on the emergence of new obstacles, shifts in environmental conditions, or fresh insights into the goal state.

\paragraph{Implementation of Real-time Heuristic Updates}
The implementation of Dynamic Heuristic Adjustment initiates a seamless integration of real-time data flow between the operational environment and the A* algorithm, mediated by the Learning and Logic Module (LLM). This dynamic interaction facilitates an ongoing process of environmental assessment and heuristic recalibration. As the A* algorithm progresses in its search and encounters new information or changes within the environment, these updates are promptly analyzed by the LLM. The LLM, in turn, assesses the current landscape of the search terrain and suggests immediate heuristic adjustments to reflect newly discovered obstacles or changes in environmental conditions accurately. This iterative recalibration process ensures that the heuristic function remains optimally aligned with the latest state of the environment, thereby enhancing the algorithm's efficiency and responsiveness in navigating complex search spaces. The real-time nature of these updates equips the A* algorithm with the agility to adapt its search strategy on-the-fly, leveraging the most current and relevant information to navigate through challenging terrains with increased efficiency and precision. Additionally, this dynamic heuristic adjustment mechanism imbues the A* algorithm with the capability to respond promptly to environmental dynamics, such as the sudden appearance or removal of obstacles, ensuring robustness and adaptability in facing unpredictable and evolving search scenarios.

\paragraph{Benefits to Pathfinding Efficiency and Accuracy}
The integration of Dynamic Heuristic Adjustment within the A* algorithm brings forth substantial enhancements in its pathfinding efficiency and accuracy. By enabling the heuristic function to adapt in real time, A* can more effectively prioritize exploration of nodes that are promisingly closer to the goal, even as environmental conditions evolve. This capacity for dynamic adaptation reduces the incidence of redundant explorations and recalculations, thereby accelerating the search process and diminishing computational demands. Moreover, this enhancement ensures that the chosen path remains optimally aligned with the goal despite dynamic changes, bolstering the algorithm's reliability and efficacy in real-time applications where conditions are prone to fluctuate. Additionally, the dynamic nature of heuristic adjustment empowers the algorithm to navigate complex and shifting environments with greater finesse, as it continuously refines its heuristic estimates in light of the evolving search space. By incorporating feedback from ongoing search iterations, the algorithm fosters a learning mechanism that progressively enhances its performance and resilience. The ability to dynamically adjust the heuristic function renders the algorithm adept at accommodating varying levels of uncertainty or incomplete information, thus broadening its applicability across a spectrum of challenging scenarios. Collectively, the adoption of Dynamic Heuristic Adjustment signifies a leap forward in pathfinding technology, offering augmented efficiency, precision, and adaptability for sophisticated real-world applications.

\subsubsection{Predictive Path Analysis}
\paragraph{Introduction to Predictive Path Analysis}
Predictive Path Analysis heralds a transformative approach to augmenting the A* algorithm by infusing it with the capability to proactively anticipate future environmental conditions and strategically optimize path selection in anticipation of those conditions. This forward-looking methodology leverages cutting-edge predictive analytics and generative AI techniques, such as those embodied by large language models, to endow the A* algorithm with the foresight required to navigate dynamically changing environments with unprecedented efficiency and adaptability. At the core of Predictive Path Analysis lies its unparalleled ability to not only react to current environmental states but to also proactively forecast impending changes, thereby enabling preemptive adjustments to the pathfinding strategy for optimal navigation. By extrapolating future scenarios and considering potential environmental dynamics, Predictive Path Analysis significantly amplifies the A* algorithm’s efficiency, reliability, and adaptability across a wide array of domains and application contexts.

\paragraph{Implementing Predictions in Pathfinding}
The implementation of Predictive Path Analysis within the A* algorithm framework involves the strategic utilization of a diverse array of data sources and predictive models, meticulously curated by the LLMs. This ensemble includes, but is not limited to, historical data repositories, real-time sensor feeds, and bespoke predictive algorithms specifically tailored to the domain of application. For instance, within the urban navigation context, the system might leverage traffic prediction models, meteorological forecasts, and local event schedules to accurately forecast future conditions that could impact route efficiency. By integrating these predictive insights into the heuristic function of the A*, the LLM transcends mere consideration of the current environmental state to incorporate anticipated future changes. This forward-looking approach empowers the algorithm to adjust its path selections proactively, optimizing route planning and navigation in light of expected developments.

\paragraph{Enhancing Path Selection with Future Insights}
The strategic incorporation of Predictive Path Analysis into the A* algorithm enriches the pathfinding process with a forward-looking perspective, enabling A* to select paths that are not merely optimal based on the current environmental configuration but are also resilient to expected future changes. Predictive Path Analysis furnishes the A* algorithm with the ability to foresee upcoming obstacles, anticipate areas of traffic congestion, or identify other dynamic factors likely to influence the chosen path. By factoring in these future insights, A* is positioned to make more informed and strategic decisions, prioritizing paths that are less susceptible to disruption or obstruction, thereby optimizing the overall efficiency and effectiveness of the pathfinding process.

\subsubsection{Semantic Path Tagging and Prioritization}
\paragraph{Delving into Semantic Path Tagging and Prioritization}
Semantic Path Tagging and Prioritization emerges as a novel Algogenic enhancement specifically devised for the A* algorithm, enriching the pathfinding process with an added layer of depth and nuanced decision-making capabilities. This enhancement capitalizes on the interpretive prowess of generative AI, notably large language models, to evaluate and prioritize paths not solely on conventional metrics such as distance or cost but also by incorporating qualitative considerations like safety, aesthetic appeal, or alignment with specific user preferences. By assigning semantic tags to paths that encapsulate these qualitative attributes, this approach introduces a nuanced layer of prioritization that more accurately mirrors the complex objectives often inherent in real-world navigation and exploration endeavors.

\paragraph{Operationalizing Semantic Insights in Pathfinding}
The operationalization of Semantic Path Tagging and Prioritization entails a sophisticated process of leveraging advanced machine learning algorithms and data analysis techniques to meticulously analyze extensive datasets related to paths and their contextual attributes. This analytical endeavor aims to assign meaningful semantic tags to paths, extending beyond mere geometric or topological attributes to encompass aspects such as scenic beauty, safety levels, or environmental impact. For example, within urban contexts, paths may be tagged based on proximity to landmarks, green spaces, or cultural hotspots, thereby enriching the user experience by offering routes that are not only efficient but also aesthetically pleasing or culturally enriching. Similarly, in logistic applications, semantic insights might prioritize paths based on criteria like road conditions, congestion trends, or safety considerations, optimizing delivery routes for both efficiency and safety.

\paragraph{Implications for Pathfinding Strategy and Outcome}
The incorporation of Semantic Path Tagging and Prioritization into the A* algorithm significantly expands its scope and efficacy, enabling it to cater to a wide spectrum of user needs and preferences by integrating a rich array of qualitative factors into the path selection process. This enhancement not only augments the algorithm's ability to generate paths that are optimally efficient but also ensures that these paths are profoundly aligned with users' unique objectives and preferences, thereby elevating the relevance and satisfaction derived from the pathfinding outcomes.

\subsubsection{Path Interpretation and Explanation}
\paragraph{Unveiling Path Interpretation and Explanation}
Path Interpretation and Explanation stands as a pivotal post-processing Algogenic enhancement for the A* algorithm, meticulously designed to bridge the gap between the algorithm's computational output and the user's comprehension of that output. This enhancement employs the sophisticated capabilities of generative AI, particularly large language models, to generate accessible, detailed, and meaningful explanations of the path selected by the A* algorithm. These explanations delve into the myriad factors and constraints that influenced the algorithm's decision-making process, not only enhancing transparency but also fostering trust, especially in applications where comprehending the rationale behind path selection is as crucial as the selection itself.

\paragraph{Implementing Explanatory Mechanisms}
The implementation of Path Interpretation and Explanation within the A* algorithm involves a meticulously structured process that unfolds once the optimal path has been identified. Following the path's determination, the Local Logic Module (LLM) embarks on a comprehensive analysis, considering a wide range of factors such as the heuristic functions applied, semantic tags associated with the nodes, and any dynamic adjustments that were made during the pathfinding journey. This thorough examination forms the foundation for crafting a coherent narrative that elucidates the rationale behind the selected path, presented in a format that is both accessible and highly relevant to the end user.

\paragraph{Enhancing User Experience and Algorithm Utility}
Integrating Path Interpretation and Explanation into the Algogenic framework of the A* algorithm fundamentally transforms the pathfinding experience into a transparent and informative journey. Users are afforded deep insights into the reasoning behind the algorithm's decisions, instilling a robust sense of confidence and trust in the technology. This level of transparency is particularly vital in intricate decision-making contexts, where users must critically evaluate the algorithm's recommendations against their own knowledge and preferences. Moreover, these explanations serve as a critical feedback mechanism for system designers and engineers, providing invaluable insights into the algorithm's real-world behavior and highlighting areas for further refinement and adaptation.

\subsubsection{Adaptive Learning from Path Successes and Failures}
\paragraph{Exploring Adaptive Learning Mechanisms}
Adaptive Learning from Path Successes and Failures introduces a dynamic learning mechanism aimed at continuously enhancing the A* algorithm's performance through the principle of iterative refinement. This post-processing Algogenic enhancement leverages the insights gained from both successful and unsuccessful pathfinding endeavors, orchestrating a learning process that iteratively refines the algorithm's decision-making capabilities. By analyzing a broad spectrum of pathfinding outcomes, the LLM identifies patterns, nuances, and critical factors that distinguish successful paths from those that encountered challenges, using these insights to guide the algorithm's evolutionary development.

\paragraph{Operationalizing Adaptive Learning}
The operationalization of Adaptive Learning involves establishing a comprehensive feedback loop wherein data pertaining to each pathfinding attempt's outcome is meticulously collected and analyzed. This process encompasses a detailed examination of metrics related to efficiency, safety, and user satisfaction, as well as an assessment of various environmental factors that may influence the efficacy of different paths. While the primary focus centers on the success of pathfinding attempts, the algorithm also evaluates the computational resources expended during each attempt, aiming to strike a balance between efficiency and accuracy.

\paragraph{Benefits of Continuous Learning and Adaptation}
Incorporating Adaptive Learning from Path Successes and Failures into the A* algorithm metamorphoses it from a static solution-finding tool into a dynamic, evolving system that enhances its effectiveness and reliability over time. This Algogenic enhancement enables the A* algorithm to become increasingly adept and responsive, as it continuously adapts to changing conditions and learns from its interactions with the environment. Such a capability is invaluable in dynamic or complex environments, where initial conditions and available data may not fully encapsulate the challenges encountered during pathfinding. Through the mechanism of adaptive learning, the A* algorithm is equipped to offer optimized and informed path selections, significantly improving both its performance and utility across a diverse range of applications.

	\subsubsection{Pseudocode for Algogenic A*}
	
	The Algogenic A* approach utilizes AI to enhance traditional numerical integration methods by dynamically adjusting integration parameters and strategies based on the observed behavior of the function and real-time error estimates. This pseudocode, available in \ref{fig:astar-Algogen-pseudocode}, outlines an advanced framework incorporating AI-driven enhancements for adaptive scheme selection, domain partitioning, error estimation, and real-time parameter optimization.
	
	\begin{algorithm}
		\caption{Algogenic A* Pseudocode}
		\begin{algorithmic}[1]
			\Procedure{AlgogenicAStar}{Graph, Start, Goal}
			\State PreprocessGraph(Graph) \Comment{Heuristic enhancement based on context}
			\State Initialize all nodes with $g = \infty$ and $f = \infty$ except $Start.g = 0$ and $Start.f = $ Heuristic(Start, Goal)
			\State OpenSet $\gets$ \{Start\}
			\While{OpenSet is not empty}
			\State Current $\gets$ Node in OpenSet with the lowest $f$
			\If{Current == Goal}
			\State \Return ReconstructPath(Current)
			\EndIf
			\State OpenSet.Remove(Current)
			\For{each Neighbor of Current}
			\State TentativeGScore $\gets$ Current.g + dist(Current, Neighbor)
			\If{TentativeGScore < Neighbor.g}
			\State Neighbor.CameFrom $\gets$ Current
			\State Neighbor.g $\gets$ TentativeGScore
			\State Neighbor.f $\gets$ Neighbor.g + Heuristic(Neighbor, Goal)
			\If{Neighbor not in OpenSet}
			\State OpenSet.Add(Neighbor)
			\EndIf
			\State DynamicHeuristicAdjustment(Neighbor, Goal) \Comment{Adjust heuristics dynamically}
			\EndIf
			\EndFor
			\State PredictivePathAnalysis(Current, Goal) \Comment{Adjust for future states}
			\EndWhile
			\State Path $\gets$ ReconstructPath(Goal)
			\State Path $\gets$ SemanticPathTagging(Path) \Comment{Tag path with semantic info}
			\State ExplainPath(Path) \Comment{Generate explanation for the chosen path}
			\State AdaptiveLearning(Path) \Comment{Learn from path success or failure}
			\State \Return Path
			\EndProcedure
		\end{algorithmic}\label{fig:astar-Algogen-pseudocode}
	\end{algorithm}

	\begin{figure}
		\centering
		\includegraphics[width=0.7\textwidth]{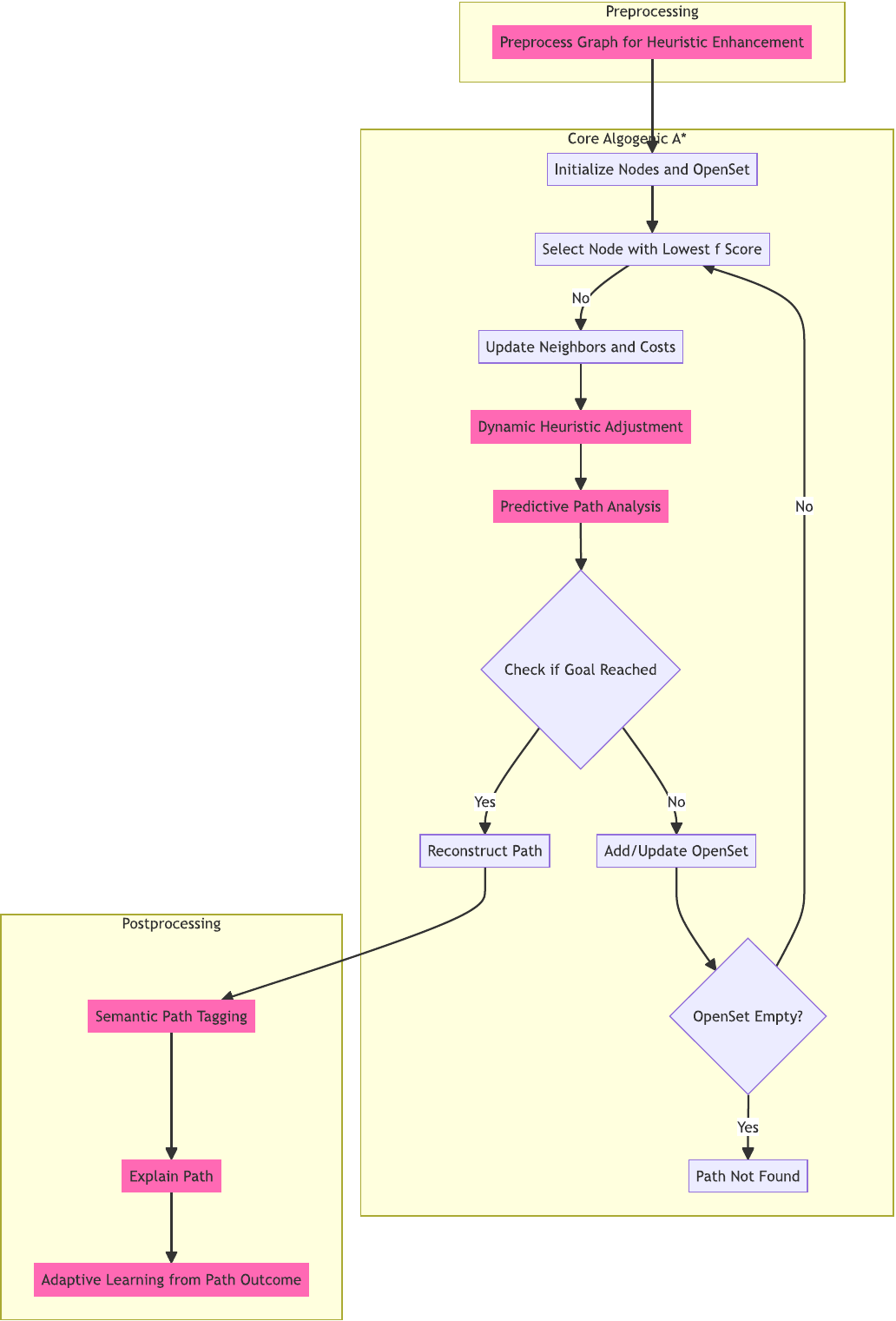} 
		\caption{Integrating Algogenic Enhancements with A*: This diagram visualizes the comprehensive integration of generative AI enhancements within the A* pathfinding framework, divided into preprocessing, core, and postprocessing phases. In the preprocessing phase, heuristic enhancement is performed to tailor the heuristic function to the specific context of the problem. The core phase illustrates a detailed interplay between traditional A* algorithm steps and Algogenic enhancements, including dynamic heuristic adjustment and predictive path analysis, highlighting how these enhancements improve real-time adaptability and decision-making. The postprocessing phase emphasizes the role of semantic path tagging, explanation of path choices, and adaptive learning from outcomes, showcasing how generative AI can enhance the interpretability, effectiveness, and continuous improvement of the A* algorithm in navigating complex environments.}
		\label{fig:a_star}
	\end{figure}

	\section{Dijkstra's Algorithm}\index{Dijkstra's Algorithm}
	\subsection{Introduction to Dijkstra's Algorithm}
	\subsubsection{The Concept of Dijkstra's Algorithm}
	\paragraph{Definition and Purpose}
	Dijkstra's Algorithm, coined after its inventor Edsger W. Dijkstra, stands as a cornerstone method within the realm of computer science, primarily employed for determining the most efficient path between nodes within a graph structure. These graphs may denote diverse systems, spanning from transportation networks to digital communication infrastructures, encompassing any system amenable to graphical representation. The algorithm's principal objective lies in resolving the single-source shortest path quandary, wherein the primary aim is to ascertain the most concise routes originating from a designated source vertex to all other vertices dispersed throughout the graph. This algorithmic paradigm plays a pivotal role in various real-world applications, furnishing indispensable insights into optimal routing strategies and resource allocation schemes. Through meticulous examination and evaluation of potential pathways, Dijkstra's Algorithm empowers decision-making processes across multifarious domains, fostering enhanced efficiency, resource utilization, and operational efficacy.

	\paragraph{Core Mechanism}
	The essence of Dijkstra's Algorithm lies in its systematic exploration of the graph, starting from the source vertex. It iteratively selects the vertex with the minimum distance from the source, updates the distances of its adjacent vertices, and repeats the process until all vertices have been visited. The algorithm employs a priority queue to efficiently identify the next vertex to process, which significantly optimizes its running time. This prioritization ensures that the algorithm always explores the shortest paths first, gradually expanding its search outward. By maintaining a set of visited vertices and their shortest distances from the source, Dijkstra's Algorithm guarantees the discovery of the shortest paths in weighted graphs without negative cycles. Furthermore, its greedy nature ensures that once a vertex is included in the set of visited vertices, its shortest path is finalized, contributing to the algorithm's efficiency. Additionally, Dijkstra's Algorithm exhibits a locality property, focusing its attention on nearby vertices before exploring distant ones, which further enhances its performance, particularly in dense graphs. Thus, through its systematic and efficient exploration strategy, Dijkstra's Algorithm stands as a fundamental tool for solving shortest path problems in various applications.

	\paragraph{Algorithmic Process}
	At the outset of the algorithm, distances to all vertices are initialized to infinity, except for the source vertex, which is assigned a distance of zero. Dijkstra's Algorithm operates by maintaining a set of vertices whose minimum distance from the source is currently known, initially comprising solely the source vertex. During each iteration, the algorithm selects the vertex \(v\) with the smallest distance from the source that has not yet been included in the set. Subsequently, it updates the distances of vertices adjacent to \(v\) if a shorter path is discovered through \(v\). This iterative process persists until distances to all vertices in the graph are ultimately determined.

	\paragraph{Mathematical Representation}
	The update of distances in Dijkstra's Algorithm is represented by the formula \(d\left(u\right) = \min\left(d\left(u\right), d\left(v\right) + \text{wt}\left(v, u\right)\right)\). Here, \(d\left(u\right)\) signifies the current distance from the source to vertex \(u\), while \(d\left(v\right)\) represents the distance from the source to vertex \(v\). The term \(\text{wt}\left(v, u\right)\) denotes the weight of the edge connecting vertices \(v\) and \(u\). This formula embodies the essence of Dijkstra's Algorithm, ensuring that the algorithm always considers the shortest path discovered so far to each vertex, thereby progressively updating the distances as it traverses the graph. Moreover, this mathematical representation underscores the algorithm's efficiency in finding the shortest paths in weighted graphs, as it systematically evaluates and updates the distances based on the accumulated weights along the explored paths. Additionally, the use of the minimum function in the formula emphasizes the algorithm's greedy nature, prioritizing the exploration of the shortest paths to reach each vertex from the source.

	\paragraph{Significance and Utility}
	Dijkstra's Algorithm is celebrated for its clarity, efficiency, and broad applicability in various domains requiring the optimization of paths. It stands as a cornerstone in computer science and graph theory, offering a straightforward approach to finding the shortest path between nodes in a graph. Its elegance lies in its simplicity, making it accessible even to those new to algorithmic concepts. Moreover, its efficient implementation using priority queues ensures quick computation, enabling real-time pathfinding in dynamic environments. The algorithm finds extensive use in practical applications such as routing protocols, geographic mapping services, and network design, where determining the most efficient path is essential. By prioritizing nodes based on their distance from the source, Dijkstra's Algorithm efficiently explores the graph, guaranteeing optimal paths with non-negative edge weights. Its adaptability to various scenarios and straightforward implementation make it a staple in the toolkit of algorithms for graph analysis.

	\subsubsection{Key Principles and Mechanisms}
	\paragraph{Optimality and Greediness}
	Dijkstra's Algorithm operates on the principle of optimality, ensuring that the shortest path to any vertex found at a given step is indeed the shortest path to that vertex. It achieves this through a greedy strategy, selecting the most promising vertex (the one with the smallest known distance from the source) at each step. This greedy choice guarantees that the path being extended is always optimal with respect to the part of the graph already processed. Furthermore, this greedy approach simplifies the algorithm's implementation and allows for efficient computation, as it focuses on immediate gains without considering potential future consequences. However, it's important to note that while Dijkstra's Algorithm guarantees optimality in finding the shortest path to each explored vertex, it does not necessarily ensure that the overall path from the source to the destination is the shortest. This limitation arises due to the greedy nature of the algorithm, which may overlook alternative paths that could potentially yield a shorter overall distance. Nevertheless, in practice, Dijkstra's Algorithm remains a highly effective and widely used method for finding shortest paths in various applications, ranging from network routing to transportation logistics.

	\paragraph{Data Structures for Efficiency}
	Efficient execution of Dijkstra's Algorithm relies heavily on the use of appropriate data structures. A priority queue, often implemented as a binary heap, min-priority queue, or Fibonacci heap, is critical for maintaining and retrieving the next vertex with the minimum distance in logarithmic time. Additionally, arrays or dictionaries are used to store distances from the source to each vertex and to keep track of whether a vertex has been visited. Using a priority queue ensures that the algorithm selects the vertex with the shortest distance efficiently, preventing the need to iterate through all vertices to find the minimum. This significantly reduces the time complexity of the algorithm, making it suitable for large-scale graphs. Arrays or dictionaries provide constant-time access to the distance of each vertex from the source and enable efficient updates to these distances as the algorithm progresses. By utilizing these data structures, Dijkstra's Algorithm achieves its goal of finding the shortest paths in a graph while maintaining optimal time complexity.

	\paragraph{Edge Relaxation}
	Edge relaxation is a fundamental mechanism in Dijkstra's Algorithm that iteratively updates and tightens the bounds on the shortest path distances. It plays a pivotal role in determining the most efficient routes within a graph. When an edge \(\left(v, u\right)\) is relaxed, it involves scrutinizing whether the current shortest path to vertex \(u\) can be further optimized by passing through vertex \(v\). This assessment is guided by comparing the sum of the shortest path distance to \(v\) and the weight of the edge \(\left(v, u\right)\) with the current shortest distance to \(u\). If the former sum is lesser than the latter, then the distance to \(u\) is updated accordingly, ensuring the shortest path is accurately represented. This process encapsulates the essence of Dijkstra's Algorithm, where each edge relaxation contributes to the refinement of shortest path estimations until the optimal paths are determined. Mathematically, this relaxation condition can be expressed as: if \(d\left(u\right) > d\left(v\right) + \text{wt}\left(v, u\right)\), then \(d\left(u\right)\) is updated to \(d\left(v\right) + \text{wt}\left(v, u\right)\), where \(d(u)\) represents the current shortest path distance to vertex \(u\) and \(wt(v, u)\) denotes the weight of the edge \(\left(v, u\right)\).

	\paragraph{Convergence through Locality}
	The algorithm progresses by expanding a frontier of explored vertices outward from the source, ensuring that the shortest path to any vertex in the frontier is known before moving on. This local optimality ensures global optimality by the time all vertices are reached. This property is guaranteed by the algorithm's careful selection of vertices based on their current distance estimates. Moreover, as the algorithm advances, it continually updates the distance estimates of vertices within its frontier, ensuring that the shortest path to each vertex is accurately represented. This iterative refinement process, coupled with the algorithm's reliance on locality, contributes to its efficiency and effectiveness in finding the shortest paths in weighted graphs. Additionally, the convergence through locality allows the algorithm to scale well with larger graphs, as it focuses computational efforts on nearby vertices before extending to farther reaches of the graph. Thus, by prioritizing local information over global considerations, Dijkstra's algorithm achieves its goal of efficiently finding the shortest paths from a single source vertex to all other vertices in the graph.

	\paragraph{Non-negative Weights Requirement}
	A critical assumption underlying Dijkstra's Algorithm is that all edge weights in the graph must be non-negative. This requirement is essential because the algorithm's selection process depends on the fact that adding a new edge to a path cannot decrease the total path length. If negative weights were allowed, the algorithm could potentially overlook shorter paths that become available only after including edges that initially increase the path length. Furthermore, negative weights could lead to unexpected behavior, such as cycles with negative total weight, causing the algorithm to enter an infinite loop. Thus, ensuring non-negative weights is crucial for the correctness and efficiency of Dijkstra's Algorithm. Moreover, the algorithm's mathematical foundation relies on the assumption of non-negative weights, as it utilizes the principle of dynamic programming to determine the shortest paths. Consequently, violating this requirement would invalidate the underlying mathematical reasoning and compromise the algorithm's ability to find optimal solutions. Hence, adhering to the non-negative weights requirement is not just a practical consideration but a fundamental aspect of Dijkstra's Algorithm.

	\subsubsection{The Role of Priority Queues}
	\paragraph{Priority Queue as a Core Component}
	The priority queue is an indispensable data structure in Dijkstra's Algorithm, serving as the backbone for efficiently managing the set of vertices to be processed. Its primary role is to keep track of all vertices that have been discovered but not yet finalized. This means their shortest distance from the source vertex is not yet confirmed, and the priority queue quickly selects the vertex with the minimum distance from the source at each step of the algorithm. The efficiency of Dijkstra's Algorithm heavily relies on the priority queue's ability to efficiently retrieve and update the vertex with the minimum distance. This is typically achieved through heap data structures, such as binary heaps or Fibonacci heaps, which offer logarithmic time complexity for key operations like insertion, deletion, and finding the minimum element. Moreover, the priority queue facilitates the greedy nature of Dijkstra's Algorithm by ensuring that the vertex with the smallest tentative distance is always selected for expansion, thus guaranteeing the algorithm's correctness. Additionally, the priority queue plays a crucial role in optimizing the algorithm's time complexity, enabling Dijkstra's Algorithm to efficiently find the shortest paths in large-scale graphs with millions of vertices and edges.

	\paragraph{Facilitating Efficient Vertex Selection}
	Utilizing a priority queue is pivotal for Dijkstra's Algorithm to efficiently determine which vertex to process next without the need to scan all vertices. This approach significantly enhances the algorithm's performance, particularly in dense graphs or those with a large number of vertices. The priority queue automatically organizes vertices based on their current distance from the source, thereby ensuring that the algorithm proceeds with the closest vertex not yet finalized. This prioritization mechanism streamlines the selection process, reducing the computational burden and enabling quicker convergence towards the optimal solution. Moreover, by maintaining a sorted order of vertices, the priority queue optimizes the overall runtime complexity of the algorithm, making it well-suited for real-world applications where efficiency is paramount. The mathematical representation of the priority queue's functionality can be expressed as follows:
	
	\[
	\text{priority}(v) = \text{distance}(v)
	\]
	
	where \( v \) represents a vertex, and \(\text{distance}(v)\) denotes the current shortest distance from the source to vertex \( v \). This prioritization strategy ensures that the algorithm always prioritizes vertices with shorter distances, effectively guiding the exploration process towards the target destination.

	\paragraph{Implementation Variants and Their Impact}
	The choice of priority queue implementation has a significant impact on the algorithm's overall time complexity. Basic implementations like binary heaps offer a good balance between ease of implementation and performance, with operations like insert and extract-min running in \(O(\log n)\) time. More advanced structures like Fibonacci heaps can reduce the amortized cost of decrease-key operations, which is critical in Dijkstra's Algorithm, potentially leading to even more efficient runtime characteristics. Additionally, pairing heaps provide a simpler alternative to Fibonacci heaps while still achieving comparable performance in practice. Moreover, other variants such as binomial heaps or Brodal queues offer different trade-offs in terms of space complexity and operation efficiency. Furthermore, the choice of implementation may also depend on the specific requirements of the application and the characteristics of the input graph. For example, in scenarios where the graph is sparse, certain implementations may outperform others due to their better handling of memory usage and cache efficiency. Hence, careful consideration of implementation variants is essential for optimizing the performance of Dijkstra's Algorithm in diverse contexts.

	\paragraph{Optimization of Update Operations}
	Dijkstra's Algorithm heavily relies on efficiently updating the distances of vertices adjacent to the currently processed vertex, a pivotal operation for its overall performance. The priority queue plays a central role in facilitating these updates by swiftly locating and adjusting the priorities (distances) of adjacent vertices. When a shorter path to a vertex is uncovered during the algorithm's execution, the decrease-key operation within the priority queue enables the rapid reordering of that vertex based on its newly calculated, shorter distance. This operation involves replacing the current priority of a vertex with a new, lower priority, reflecting the discovery of a more optimal path. Mathematically, this operation can be represented as follows:
	
	\[ \text{{new\_priority}}(v) = \text{{min}}(\text{{new\_priority}}(v), \text{{distance}}(u) + \text{{weight}}(u, v)) \]
	
	Here, \( v \) represents the vertex being updated, \( u \) denotes the vertex from which the shorter path is discovered, and \( \text{{distance}}(u) \) represents the distance to vertex \( u \). This update operation ensures that the algorithm maintains the shortest distances to each vertex as it progresses through the graph, contributing significantly to its efficiency and correctness.

	\paragraph{Critical for Algorithm's Correctness and Performance}
	The use of a priority queue in Dijkstra's Algorithm is critical for ensuring both its correctness and performance. Without a priority queue, the algorithm lacks the ability to systematically select the next vertex for processing based on its distance from the source. Consequently, the algorithm may fail to identify the shortest paths accurately, leading to incorrect results. The priority queue facilitates the selection of vertices with the shortest distances, ensuring that the algorithm progresses in a manner consistent with its objective of finding the shortest paths. Moreover, the efficiency of Dijkstra's Algorithm heavily depends on the priority queue's ability to efficiently extract the vertex with the minimum distance, enabling the algorithm to explore the graph in a systematic and optimized manner. Therefore, the priority queue serves as a fundamental component that underpins both the correctness and efficiency of Dijkstra's Algorithm, making it indispensable for its successful implementation in various applications. In mathematical terms, the priority queue ensures that the vertex with the minimum distance from the source is always selected for processing, adhering to the algorithm's greedy strategy for finding the shortest paths.

	\subsubsection{Applications and Limitations}
	\paragraph{Wide Range of Applications}
	Dijkstra's Algorithm has found extensive applications across various fields due to its robust and versatile nature in solving shortest path problems. It is pivotal in network routing protocols where determining the most efficient path is critical, such as in OSPF (Open Shortest Path First) and in GPS navigation systems to calculate the quickest route between locations. Moreover, it plays a significant role in planning and optimization problems within logistics, urban planning, and even in the realm of electronic design automation for laying out circuits on silicon chips. Additionally, Dijkstra's Algorithm is widely utilized in telecommunications for call routing, in social network analysis for determining influential nodes, and in biology for analyzing metabolic pathways. Its applications extend to computer graphics for pathfinding in video games, to robotics for motion planning, and to traffic management systems for optimizing traffic flow. The algorithm's simplicity and efficiency make it a go-to choice in various real-world scenarios, contributing significantly to the advancement of diverse fields and technologies.

	\paragraph{Limitations and Constraints}
	Despite its widespread use, Dijkstra's Algorithm is not without limitations. The requirement for all edge weights to be non-negative is a significant constraint, as it cannot correctly process graphs with negative weight edges, a scenario that might arise in applications involving cost adjustments, rebates, or certain types of financial modeling. This limitation stems from the algorithm's reliance on the greedy approach, where it always selects the vertex with the shortest known distance, assuming non-negative weights ensure optimality. However, negative weights can lead to incorrect shortest path calculations, as the algorithm might prematurely terminate the search. Furthermore, its performance can be less than optimal for graphs with a very large number of vertices or edges, due to the computational complexity associated with maintaining the priority queue and updating the distances of adjacent vertices. Despite these constraints, Dijkstra's Algorithm remains a powerful tool in various fields, especially when applied to scenarios where non-negative edge weights and relatively small graphs are prevalent.

	\paragraph{Performance Considerations}
	The computational efficiency of Dijkstra's Algorithm is heavily influenced by the implementation of the priority queue. With a simple array or linked list, the time complexity can degrade to \(O(v^2)\), where \(v\) is the number of vertices. However, using a binary heap improves this to \(O((v + e) \log v)\), where \(e\) is the number of edges, and employing a Fibonacci heap can further optimize it. Additionally, careful consideration must be given to the data structure used for representing the graph, as inefficient representations can significantly impact performance. Furthermore, the algorithm may encounter challenges in extremely large or dynamic graphs, where frequent updates necessitate recalculating paths. Moreover, the presence of negative weights or cycles can also affect the algorithm's performance, as it is designed for graphs with non-negative weights. Hence, while Dijkstra's Algorithm is highly efficient in certain scenarios, its performance can be influenced by various factors, requiring careful analysis and optimization for optimal results.

	\paragraph{Innovations and Adaptations}
	To overcome some of these limitations, numerous variations and improvements on Dijkstra's Algorithm have been proposed. Techniques such as A* incorporate heuristics to guide the search process, significantly reducing the number of vertices explored in applications like pathfinding in video games or robotics. Additionally, algorithms like Bellman-Ford and Floyd-Warshall offer alternatives that can handle negative weights, albeit with different trade-offs in terms of complexity and applicability. Moreover, recent advancements in parallel computing have led to parallelized versions of Dijkstra's Algorithm, allowing for faster execution on modern multi-core processors and distributed systems. Furthermore, hybrid algorithms that combine elements of Dijkstra's Algorithm with other techniques, such as genetic algorithms or simulated annealing, have emerged to address specific challenges in complex optimization problems. These adaptations aim to enhance the algorithm's efficiency, scalability, and versatility in various real-world scenarios, ranging from transportation networks to telecommunications infrastructure.

	\paragraph{Conclusion on Applicability}
	The applications and limitations of Dijkstra's Algorithm highlight its importance in computational theory and practice. Dijkstra's Algorithm stands as a fundamental pillar in graph theory and network analysis, offering a robust method for finding the shortest path between nodes in a graph. Its simplicity and efficiency make it a go-to choice for a wide range of applications, including network routing, transportation planning, and resource allocation. However, despite its versatility, Dijkstra's Algorithm is not without its limitations. One key constraint is its inability to handle negative edge weights, which restricts its applicability in certain scenarios where negative weights are present. Additionally, its computational complexity can become prohibitive for large-scale graphs, necessitating the exploration of alternative algorithms for more efficient solutions. Nevertheless, with careful consideration of its strengths and weaknesses, Dijkstra's Algorithm remains a valuable tool in the algorithmic toolbox, providing insights into graph traversal and optimization that continue to shape the development of new algorithms and methodologies in the field of computer science.

	\subsubsection{Pseudocode for Dijkstra's Algorithm}
	The Dijkstra Algorithm is a well-known method utilized for finding the shortest path from a starting node to all other nodes within a graph. It operates by iteratively selecting the node with the lowest tentative distance from the source node, updating the distances of its neighboring nodes accordingly. Unlike the A* algorithm, Dijkstra's Algorithm does not consider any heuristic or estimated cost to reach the goal. Instead, it simply focuses on minimizing the cumulative distance traveled from the source node to all other nodes. The procedural details of the Dijkstra Algorithm are outlined in the pseudocode \ref{fig:dijkstra-pseudocode}, depicting its systematic procedure for traversing the graph and determining the shortest paths.
	
	\begin{algorithm}
		\caption{Dijkstra's Algorithm Pseudocode}
		\begin{algorithmic}[1]
			\Procedure{Dijkstra}{Graph, source}
			\State Initialize all distances to infinity except source to zero
			\State Set all vertices as unvisited
			\State Create a priority queue to hold vertices by distance
			\State Insert the source vertex into the queue with distance zero
			\While{priority queue is not empty}
			\State Extract the vertex with the minimum distance from the queue
			\State Mark the vertex as visited
			\For{each neighbor of the extracted vertex}
			\If{the neighbor is not visited and the new path is shorter}
			\State Update the distance to the neighbor
			\State Update the neighbor's entry in the priority queue
			\EndIf
			\EndFor
			\EndWhile
			\State \Return The array of distances from source to all vertices
			\EndProcedure
		\end{algorithmic}\label{fig:dijkstra-pseudocode}
	\end{algorithm}

	\subsection{Previous Work on ML and AI Interplay with Dijkstra's Algorithm}
	Building upon the foundation of Dijkstra's algorithm, recent advancements have explored the potential of machine learning and artificial intelligence to enhance its efficiency and applicability.
	
	\paragraph{Incorporating Machine Learning Predictions}
	
	One approach involves incorporating machine learning predictions to potentially accelerate the search process. A study proposed training a model to predict edge distances within the graph \cite{feijen2021using}. These predictions are then utilized to prioritize the exploration space, focusing on areas with a higher likelihood of containing the shortest path. This strategy aims to reduce computational cost while maintaining result accuracy.
	
	\paragraph{Integration of Learned Heuristics}
	
	Another direction investigates the integration of learned heuristics within the algorithm's framework. A recent work introduced a variant that employs a deep neural network to predict the remaining distance to the target node from any point in the graph \cite{feijen2021dijkstras}. These predictions guide the search in both forward and backward directions, potentially accelerating the convergence towards the optimal path compared to the traditional Dijkstra algorithm.

\subsection{Algogenic Enhancements for Dijkstra's Algorithm}
\subsubsection{Graph Structure Optimization}
\paragraph{Introduction to Graph Structure Optimization}
Enhancing Dijkstra's algorithm through graph structure optimization involves using generative AI to refine the graph's layout and connectivity for improved efficiency. This technique, focusing specifically on Dijkstra's application, analyzes the graph's topology and historical path data to identify and eliminate redundancies, streamline connections, and highlight crucial paths. The process aims to maintain the integrity of shortest path calculations while simplifying the graph's structure to facilitate quicker pathfinding.

\paragraph{Implementing Graph Structure Optimization}
Implementing this optimization requires a detailed analysis of the graph, considering metrics like node centrality and edge density. The AI may suggest merging frequently co-occurring nodes to simplify the topology, or eliminating rarely used edges to reduce complexity. This preprocessing step enhances Dijkstra's algorithm by reducing computation times and improving path quality through a more streamlined graph.

\paragraph{Impact on Dijkstra's Algorithm}
Optimizing the graph structure can significantly improve Dijkstra's algorithm's performance, especially in complex networks. By refining the graph, the algorithm can navigate more efficiently, leading to quicker and more resource-effective shortest path determinations. This Algogenic approach ensures Dijkstra's algorithm remains effective in dynamic and densely connected environments.

\subsubsection{Dynamic Weight Adjustment}
\paragraph{Exploring Dynamic Weight Adjustment}
Dynamic Weight Adjustment involves real-time modification of edge weights in Dijkstra's algorithm based on changing conditions, like traffic in routing systems. This adaptation reflects current conditions, optimizing pathfinding outcomes by rerouting around congestion or disruptions, enhancing the algorithm's responsiveness and efficiency in dynamic environments.

\paragraph{Implementation of Weight Adjustments}
Implementing dynamic adjustments involves a feedback loop where real-time data predicts necessary weight changes. For instance, increasing weights on congested routes can guide the algorithm to select more efficient paths, optimizing route selection dynamically in response to evolving conditions.

\paragraph{Benefits to Pathfinding Efficiency and Accuracy}
Dynamic weight adjustment enhances Dijkstra's efficiency and adaptability, allowing it to respond to network changes. This ensures optimal path selection even under fluctuating conditions, enhancing the algorithm's robustness and reliability in real-time applications.

\subsubsection{Predictive Path Prioritization}
\paragraph{Introduction to Predictive Path Prioritization}
Predictive Path Prioritization enhances Dijkstra's algorithm by forecasting future changes in the graph's environment, such as traffic conditions. This foresight allows the algorithm to adjust its path selection strategy proactively, prioritizing routes likely to remain optimal, enhancing efficiency and adaptability in dynamic scenarios.

\paragraph{Operationalizing Predictive Prioritization}
This involves analyzing historical data and current trends with predictive models to forecast changes. The algorithm dynamically adjusts path priorities based on these predictions, ensuring more efficient exploration and improved route selection in real-time.

\paragraph{Enhancing Path Selection with Predictive Insights}
Incorporating predictive insights allows Dijkstra's algorithm to navigate dynamic graphs more effectively, selecting paths that account for future conditions. This proactive approach enhances the algorithm's resilience and efficiency, offering a robust solution to dynamic pathfinding challenges.

\subsubsection{Heuristic-Guided Exploration}
\paragraph{Redefining Exploration with Heuristics}
Integrating heuristic guidance in Dijkstra's algorithm enables more effective exploration by prioritizing nodes and paths based on contextual factors and historical data. This approach helps mitigate the algorithm's limitations, such as unnecessary explorations, enhancing efficiency and adaptability in dynamic environments.

\paragraph{Implementing Heuristic-Guided Exploration}
This involves developing a context-aware heuristic function that evaluates node potential, incorporating probabilistic models to account for uncertainty. The heuristic function dynamically adjusts exploration strategy, optimizing pathfinding in real-time based on changing conditions.

\paragraph{Impact on Algorithmic Efficiency and Effectiveness}
Heuristic guidance transforms Dijkstra's algorithm into a more efficient and intelligent tool, focusing on promising areas of the graph. This reduces unnecessary computations and enhances the quality of solutions, making the algorithm more adaptable and effective in complex scenarios.

\subsubsection{Path Reliability Analysis}
\paragraph{Introducing Path Reliability Analysis}
Path Reliability Analysis enhances Dijkstra's algorithm by assessing the dependability of identified paths, considering factors like historical stability and environmental influences. This analysis prioritizes paths not only for their efficiency but also for their reliability in varying conditions, introducing a probabilistic optimization perspective.

\paragraph{Executing Reliability Analysis}
This involves analyzing a broad set of data to assign reliability scores to paths, prioritizing those offering greater stability over time. The analysis adapts to changing conditions and user feedback, refining path selection for robustness and reliability.

\paragraph{Enhancing Path Selection with Reliability Insights}
Incorporating reliability insights into Dijkstra's algorithm improves decision-making, prioritizing routes that balance efficiency with reliability. This Algogenic enhancement fosters a more resilient and adaptable navigation strategy, enhancing the pathfinding process's overall effectiveness.

\subsubsection{Semantic Path Enhancement}
\paragraph{Expanding Pathfinding with Semantic Context}
Semantic Path Enhancement leverages generative AI to integrate qualitative attributes into Dijkstra's algorithm, aligning path recommendations with user preferences for a more personalized route planning experience. This approach enriches pathfinding by considering factors like scenic value and safety, enhancing user engagement and satisfaction.

\paragraph{Implementing Semantic Enhancements}
Implementing this involves analyzing datasets to tag paths with semantic attributes and utilizing natural language processing to extract contextual information. Machine learning predicts user preferences, tailoring path recommendations to individual needs and enhancing the navigation experience.

\paragraph{Benefits to Path Selection and User Experience}
Semantic Path Enhancement transforms pathfinding into a user-centric endeavor, prioritizing routes based on individual preferences and enhancing the overall user experience. This customization capability makes Dijkstra's algorithm more applicable across various domains, enriching user engagement and satisfaction.

\subsubsection{Adaptive Learning from Path Performance}
\paragraph{Enhancing Pathfinding through Adaptive Learning}
Adaptive learning from path performance uses feedback to refine Dijkstra's algorithm, analyzing outcomes to improve future pathfinding. This continuous improvement cycle adapts to changes and optimizes decision-making, enhancing the algorithm's efficiency, reliability, and responsiveness to user needs.

\paragraph{Operationalizing Feedback for Continuous Improvement}
This involves collecting and analyzing data on path performance and environmental conditions, utilizing feedback loops for dynamic adjustments. Adaptive learning mechanisms iteratively refine the algorithm's behavior, leading to more efficient and effective pathfinding solutions.

\paragraph{Impact on Dijkstra's Algorithm Efficiency and Reliability}
Adaptive learning enhances Dijkstra's long-term efficiency and reliability, ensuring it adapts to changing conditions and evolves to meet user needs. This continuous improvement process identifies and addresses inefficiencies, making the algorithm more robust and dependable in diverse scenarios.

	\subsubsection{Pseudocode for Algogenic Dijkstra's}
	The Algogenic Dijkstra approach utilizes AI to enhance traditional Dijkstra's algorithm by dynamically adjusting algorithmic parameters and strategies based on the observed behavior of the graph and real-time error estimates. This pseudocode, available in \ref{fig:dijkstra-Algogen-pseudocode}, outlines an advanced framework incorporating AI-driven enhancements for adaptive path selection, node exploration, error estimation, and real-time parameter optimization.
	
	\begin{algorithm}
		\caption{Algogenic Dijkstra's Pseudocode}
		\begin{algorithmic}[1]
			\Procedure{AlgogenicDijkstra}{Graph, Source}
			\State PreprocessGraph(Graph) \Comment{Graph structure optimization}
			\State InitializeDistances(Graph, Source)
			\State PriorityQueue $\gets$ InitializePriorityQueue(Source)
			\While{PriorityQueue not empty}
			\State CurrentNode $\gets$ ExtractMin(PriorityQueue)
			\If{DynamicWeightAdjustment(CurrentNode)} \Comment{Adjust weights dynamically}
			\State UpdateNeighborWeights(CurrentNode)
			\EndIf
			\For{each Neighbor of CurrentNode}
			\If{NewPathShorter(Neighbor)}
			\State UpdatePathToNeighbor(Neighbor)
			\State PriorityQueue.Update(Neighbor)
			\EndIf
			\If{PredictivePathPrioritization(Neighbor)} \Comment{Prioritize based on predictions}
			\State ReorderPriorityQueue(PriorityQueue, Neighbor)
			\EndIf
			\EndFor
			\EndWhile
			\State Path $\gets$ ReconstructPath(Source, Goal)
			\State Path $\gets$ SemanticPathEnhancement(Path) \Comment{Enhance path semantically}
			\State AnalyzePathReliability(Path) \Comment{Assess path reliability}
			\State AdaptiveLearning(Path) \Comment{Learn from path performance}
			\State \Return Path
			\EndProcedure
		\end{algorithmic}\label{fig:dijkstra-Algogen-pseudocode}
	\end{algorithm}

	\begin{figure}
		\centering
		\includegraphics[width=0.7\textwidth]{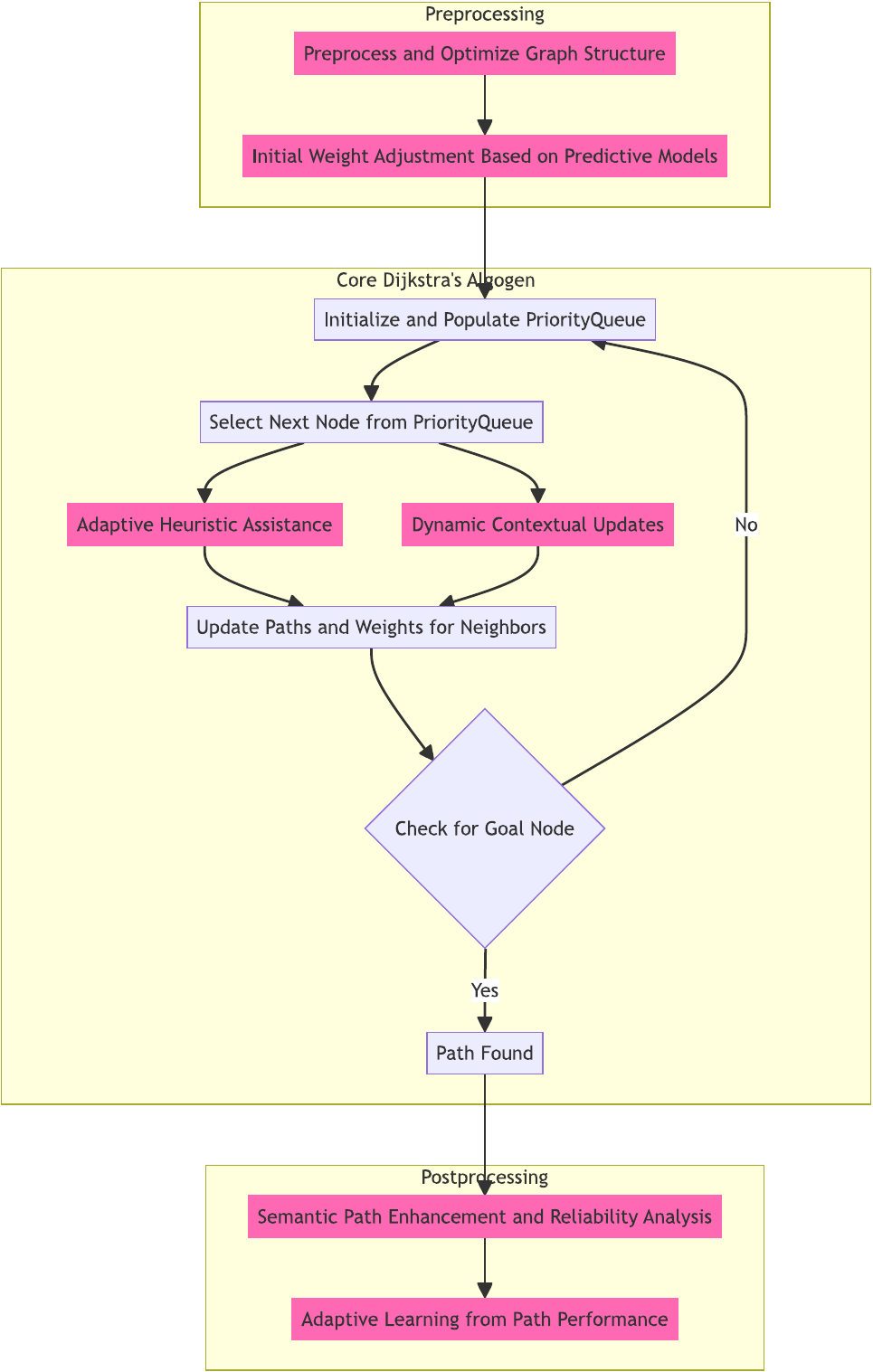}
		\caption{Integrating Algogenic Enhancements into Dijkstra's Algorithm: This diagram presents a comprehensive view of the Algogenic framework applied to Dijkstra's algorithm, emphasizing the strategic incorporation of generative AI at various stages of the algorithm. In the preprocessing phase, 'Preprocess and Optimize Graph Structure' combines initial graph analysis with structural optimizations, leveraging generative AI to refine the graph based on predictive insights, ensuring the initial setup is primed for efficient pathfinding. This is followed by 'Initial Weight Adjustment Based on Predictive Models', where generative AI adjusts edge weights to reflect anticipated conditions, enhancing the algorithm's foresight and adaptability. The core phase introduces 'Adaptive Heuristic Assistance' and 'Dynamic Contextual Updates', both powered by generative AI, to dynamically inform path selection and adjustment processes with real-time data and heuristic guidance, ensuring optimal paths are chosen even as conditions evolve. The postprocessing phase with 'Semantic Path Enhancement and Reliability Analysis' employs generative AI to enrich the identified paths with semantic context and evaluate their reliability, providing a deeper understanding of path choices. Finally, 'Adaptive Learning from Path Performance' closes the loop, using outcomes from executed paths to refine the algorithm's predictive models and adjustments, fostering continuous improvement. This Algogenic approach significantly augments Dijkstra's algorithm, offering a pathfinding solution that is not only efficient and accurate but also dynamically responsive to changing environments and user needs.}
		\label{fig:dijkstra_Algogenic}
	\end{figure}

	\section{Bellman-Ford}\index{Bellman-Ford}
	
	\subsection{Introduction to the Bellman-Ford Algorithm}
	\paragraph{Overview of the Bellman-Ford Algorithm}
	The Bellman-Ford Algorithm stands as a cornerstone in computer science, particularly within the realm of graph theory, owing to its remarkable capacity to precisely compute the shortest paths originating from a single source vertex to all other vertices within a weighted graph, even when negative edge weights are present. This algorithmic approach holds immense significance due to its versatility and robustness in handling various graph structures and edge weight configurations. Essentially, it iteratively relaxes edges in a graph for \(|V| - 1\) rounds, where \(|V|\) denotes the number of vertices, ensuring the convergence of shortest path estimates. Furthermore, the Bellman-Ford Algorithm gracefully accommodates graphs with negative weight cycles by detecting and flagging them during execution. Its wide-ranging applications encompass diverse domains, including network routing protocols, resource allocation in distributed systems, and critical path analysis in project management. Additionally, the Bellman-Ford Algorithm serves as a foundational building block for more intricate pathfinding algorithms, underscoring its pivotal role in computational problem-solving paradigms.

	\paragraph{Fundamental Mechanics}
	At its core, the Bellman-Ford Algorithm iterates over all edges of the graph, performing relaxations that update the cost of the shortest path to each vertex if a shorter path is found. Unlike Dijkstra's Algorithm, which optimistically proceeds by exploring the nearest unvisited vertices first, Bellman-Ford methodically relaxes all edges, allowing it to effectively handle negative edge weights and correctly identify the shortest path even when a graph contains cycles that decrease path costs. Additionally, Bellman-Ford's approach guarantees convergence by repeating the relaxation process for a number of iterations equal to the number of vertices minus one. This iterative nature ensures that the algorithm considers all possible paths of increasing lengths, gradually refining its estimates until the shortest paths are determined. Moreover, the Bellman-Ford Algorithm is versatile and applicable to various scenarios, including those where negative edge weights are present or where the graph is not fully connected. Despite its flexibility and ability to handle a wide range of graph structures, the Bellman-Ford Algorithm may exhibit slower performance compared to Dijkstra's Algorithm, especially in graphs with sparse connectivity or when negative cycles are present. Nevertheless, its reliability and ability to handle negative edge weights make it a valuable tool in diverse optimization and pathfinding contexts.

	\paragraph{Operational Principles}
	The operational essence of the Bellman-Ford Algorithm is encapsulated in its systematic approach, which iterates through all edges of the graph up to $\left|V\right| - 1$ times, where $\left|V\right|$ is the number of vertices. This iterative process ensures that the shortest paths are correctly computed, as it accounts for the fact that the longest path without cycles in any graph can have at most $\left|V\right| - 1$ edges. Furthermore, this algorithm uniquely incorporates a check for negative cycles, offering the capability to report their presence, as such cycles imply the non-existence of a global shortest path. This iterative nature of the Bellman-Ford Algorithm sets it apart from other algorithms like Dijkstra's, which operate based on a single-source shortest path principle. And while Dijkstra's algorithm excels in efficiency for graphs with non-negative edge weights, the Bellman-Ford Algorithm remains applicable in scenarios where negative edge weights or cycles exist, making it a versatile tool for various graph-related problems.

	\paragraph{Applicability and Versatility}
	The Bellman-Ford Algorithm's ability to handle graphs with negative edge weights extends its applicability beyond that of many other shortest-path algorithms. This unique feature allows it to efficiently compute shortest paths even in scenarios where costs associated with edges are negative, enabling its utilization in various real-world applications such as financial modeling, where transactions may involve costs or gains. Moreover, in systems analysis, where interactions between components can have negative implications, the Bellman-Ford Algorithm proves to be indispensable, providing insights into the shortest paths while considering negative weights.
	
	Furthermore, the algorithm's capability to detect negative cycles adds another layer of versatility, making it an invaluable tool in applications requiring cycle detection. This feature is particularly crucial for preventing infinite loops or identifying potentially hazardous sequences of operations in various systems, including transportation networks, computer networks, and project scheduling.
	
	In summary, the Bellman-Ford Algorithm's applicability and versatility stem from its ability to handle negative edge weights and detect negative cycles, making it a powerful tool in various domains where traditional shortest-path algorithms may fall short. Its usage extends to financial modeling, systems analysis, transportation networks, and beyond, highlighting its significance in solving real-world problems efficiently and effectively.

	In summary, the Bellman-Ford Algorithm is a versatile and robust tool for shortest-path computations, distinguished by its capacity to manage negative edge weights and identify negative cycles. Its foundational role in the realms of algorithm design and graph analysis underscores the continuing relevance and importance of understanding and applying this algorithm in solving complex problems within and beyond computer science.

	\subsubsection{Mathematical Foundations}
	The Bellman-Ford Algorithm leverages the principle of relaxation to iteratively improve the estimate of the shortest path from a single source to all other vertices in a graph, accommodating edges with negative weights. This subsection delves into the mathematical foundations that underpin the algorithm, providing insight into its operational mechanism and theoretical robustness.
	
	\paragraph{Principle of Relaxation}
	The relaxation process, pivotal in the Bellman-Ford Algorithm, serves to adjust the distance to a vertex $v$ if a shorter path through an adjacent vertex $u$ is discovered. Essentially, it involves comparing the distance to $v$ calculated via the current path with the sum of the distance to $u$ and the weight of the edge $(u, v)$. If the latter sum is less than the current distance to $v$, the algorithm updates $\text{distance}[v]$ to $\text{distance}[u] + w(u, v)$, thereby shortening the known path to $v$. This iterative process continues across all edges, gradually refining the estimations of path lengths. This mechanism ensures that the algorithm converges on the shortest paths from the source vertex to all other vertices in the graph. Moreover, it facilitates adaptability to changes in edge weights, allowing for dynamic updates to the shortest paths as the algorithm progresses. Consequently, the principle of relaxation underpins the efficiency and effectiveness of the Bellman-Ford Algorithm in finding shortest paths in weighted graphs.

	\paragraph{Handling Negative Weights and Cycles}
	The Bellman-Ford Algorithm's capability to handle negative weights derives from its exhaustive edge relaxation process, which occurs $\left|V\right| - 1$ times, where $\left|V\right|$ represents the number of vertices in the graph. This comprehensive iteration ensures the discovery of the shortest path, if one exists, irrespective of negative weights. To identify negative cycles, an extra relaxation iteration is performed. If any distance undergoes an update during this iteration, it indicates the presence of a negative cycle, as the shortest path should be determined after $\left|V\right| - 1$ iterations.

	\paragraph{Algorithmic Complexity}
	The computational complexity of the Bellman-Ford Algorithm is $O\left(\left|V\right| \cdot \left|E\right|\right)$, where $\left|E\right|$ represents the number of edges. This reflects the algorithm's iterative nature, requiring a pass through all edges for each vertex. Additionally, each vertex may relax its adjacent edges multiple times, contributing to the overall time complexity. Despite its polynomial time complexity, the Bellman-Ford Algorithm remains a fundamental tool for pathfinding, particularly in scenarios where negative edge weights are present. Unlike Dijkstra's Algorithm, which requires non-negative edge weights, Bellman-Ford can handle graphs with negative weights, making it suitable for a broader range of applications. However, this flexibility comes at the cost of efficiency, as the algorithm may iterate over the entire edge set multiple times to ensure the correct shortest paths are computed. Nevertheless, its ability to handle negative cycles and its relatively simple implementation make it a valuable asset in graph theory and network optimization.

	The mathematical foundation of the Bellman-Ford Algorithm, characterized by the relaxation principle and its iterative application, equips it to navigate graphs with negative edge weights and identify negative cycles. This foundation not only underscores the algorithm's versatility in addressing complex pathfinding problems but also highlights its significance in applications that span financial modeling, network design, and beyond, where dynamic and potentially adverse conditions influence path selection.
	
	\subsubsection{Handling Negative Edge Weights and Cycles}
	The Bellman-Ford Algorithm stands out for its proficient handling of graphs containing negative edge weights and its capability to detect negative cycles, which are critical in certain applications where costs or distances might decrease along certain paths. This subsection illuminates the algorithm's approach to these challenges, emphasizing its mathematical rationale and operational strategy.
	
	\paragraph{Adaptation to Negative Edge Weights}
	Unlike many pathfinding algorithms that assume non-negative edge weights, the Bellman-Ford Algorithm is designed to accurately compute shortest paths in graphs that may include edges with negative weights. It achieves this through a rigorous relaxation process, iteratively updating the $\text{distance}\left[v\right]$ for each vertex $v$ based on the formula: if $\text{distance}\left[v\right] > \text{distance}\left[u\right] + w\left(u, v\right)$ for an edge $\left(u, v\right)$, then $\text{distance}\left[v\right]$ is set to $\text{distance}\left[u\right] + w\left(u, v\right)$. This procedure, performed $\left|V\right| - 1$ times, where $\left|V\right|$ is the total number of vertices, ensures that the shortest paths are determined even when paths involve traversing through negative-weight edges. Additionally, the algorithm handles negative-weight cycles by detecting them during the $\left|V\right|$th iteration, ensuring that the shortest paths are correctly identified despite the presence of such cycles. Consequently, the Bellman-Ford Algorithm stands out as a versatile solution for pathfinding in graphs with diverse edge weight distributions, offering robustness and accuracy in scenarios where other algorithms may falter.

	\paragraph{Negative Cycle Detection}
	The Bellman-Ford Algorithm possesses a unique capability in detecting negative cycles within a graph. This distinctive feature emerges from its iterative approach to edge relaxation. After completing the standard $\left|V\right| - 1$ iterations, where $\left|V\right|$ represents the number of vertices in the graph, the algorithm conducts an additional check. During this supplementary step, each edge is relaxed once more. Here, the algorithm scrutinizes for any decrease in the $\text{distance}\left[v\right]$ values. A reduction in any of these values signifies the presence of a negative cycle within the graph.
	
	This detection mechanism holds paramount importance, especially in scenarios where path optimality and stability are critical. A negative cycle denotes a cycle in the graph where the total weight is negative. This implies that traversing the cycle repeatedly could perpetually decrease the path length. Consequently, the existence of such cycles renders the notion of a shortest path obsolete, as paths may continuously shorten, never reaching an optimal solution.
	
	In summary, the Bellman-Ford Algorithm's ability to detect negative cycles provides valuable insights into the graph's structure and behavior. It serves as a warning sign for scenarios where path optimality cannot be guaranteed, thus guiding decision-making processes in various applications.

	\paragraph{Operational Implications and Applications}
	The Bellman-Ford Algorithm's ability to handle negative edge weights and detect negative cycles expands its operational applicability, rendering it suitable for various scenarios unaddressed by algorithms like Dijkstra's. In financial modeling, where transactions may yield net losses (negative weights), or in network routing, where adjustments or anomalies might temporarily introduce negative costs, the algorithm offers a robust solution. Furthermore, its cycle detection capability proves invaluable for pinpointing potentially problematic loops in systems or networks, ensuring informed decision-making and preserving system integrity.

	In essence, the Bellman-Ford Algorithm's mathematical and operational foundation equips it to navigate the complexities of graphs with negative edge weights and cycles, affirming its value in scenarios requiring nuanced pathfinding capabilities and cycle analysis.

	\subsubsection{Standard Applications and Limitations}
	The Bellman-Ford Algorithm is integral to various fields due to its ability to find the shortest paths in graphs, even those with negative edge weights, and its unique capability to detect negative cycles. This versatility enables its application in a range of scenarios, from network design and optimization to economic models. However, the algorithm also encounters limitations, primarily related to its computational efficiency and the presence of negative cycles. This subsection outlines both the extensive applications and the limitations of the Bellman-Ford Algorithm.
	
	\paragraph{Applications of the Bellman-Ford Algorithm}
	The Bellman-Ford Algorithm's ability to handle negative edge weights makes it particularly useful in scenarios where costs can decrease along a path, such as financial transactions that might involve debts or losses. It is also applied in network routing protocols, where it can dynamically adjust to changing conditions and costs. Additionally, in urban planning and logistics, the algorithm helps in optimizing routes under complex conditions, including varying traffic patterns and road closures. Moreover, it finds utility in cycle detection within networks, where identifying negative cycles can prevent potential system failures or inefficiencies. The algorithm's versatility extends to various fields, including telecommunications, transportation, and computer networking, where its robustness and adaptability are indispensable. Furthermore, its role in shortest path calculations contributes to efficient resource allocation, cost optimization, and risk management strategies. Therefore, the Bellman-Ford Algorithm stands as a fundamental tool in the arsenal of algorithms, offering solutions to a diverse array of real-world problems.

	\paragraph{Limitations of the Bellman-Ford Algorithm}
	Despite its robustness and versatility, the Bellman-Ford Algorithm's computational complexity of $O(\left|V\right| \cdot \left|E\right|)$, where $\left|V\right|$ and $\left|E\right|$ represent the number of vertices and edges, respectively, poses challenges for large-scale graphs. This makes it less suitable for applications requiring real-time pathfinding solutions. Furthermore, while the algorithm's ability to detect negative cycles is beneficial, it also means that the existence of such cycles can render the task of finding shortest paths undefined, limiting the algorithm's applicability in graphs where negative cycles are a feature rather than an anomaly. Additionally, the Bellman-Ford Algorithm's performance degrades significantly when dealing with graphs that have a high density of edges or when the edges have varying weights, as it explores all possible paths in each iteration. Despite these limitations, the Bellman-Ford Algorithm remains a valuable tool in scenarios where negative cycles need to be identified or when the computational resources allow for its usage without significant performance constraints.

	\paragraph{Navigating the Trade-offs}
	The selection of the Bellman-Ford Algorithm should be informed by a clear understanding of its computational demands and the nature of the graph being analyzed. For graphs with a moderate number of vertices and edges, or where negative edge weights are present, the algorithm offers a comprehensive solution. However, for very large graphs or real-time applications, alternative algorithms or optimizations may be necessary to achieve the desired efficiency and performance. While the Bellman-Ford Algorithm guarantees finding the shortest path even in the presence of negative edge weights, its time complexity of O(V*E) can become prohibitive for graphs with a large number of vertices and edges. Therefore, in such scenarios, algorithms like Dijkstra's or the Floyd-Warshall Algorithm, with better time complexities under certain conditions, may be more suitable. Moreover, parallelizing the computation or applying heuristics to guide the search process can also alleviate the computational burden. Additionally, in real-time applications where responsiveness is crucial, trade-offs between optimality and efficiency may need to be made, favoring faster but less optimal solutions. Ultimately, the decision should be guided by a thorough analysis of the specific requirements and constraints of the problem at hand, weighing the trade-offs between computational complexity, solution optimality, and real-time performance.

	In conclusion, while the Bellman-Ford Algorithm is a powerful tool in the arsenal of graph theory, its effective deployment requires careful consideration of its strengths and limitations. Balancing these factors is key to leveraging the algorithm's capabilities to address complex pathfinding and cycle detection problems in a wide array of applications.

	\subsubsection{Algorithmic Pseudocode for Bellman-Ford}
	The Bellman-Ford Algorithm is a robust methodical approach devised for efficiently determining the shortest paths from a designated source vertex to all other vertices within a weighted graph. It stands out by iteratively relaxing edges, gradually refining its estimates of shortest paths until convergence is reached. This iterative process allows Bellman-Ford to effectively handle graphs with negative edge weights. The core functionality of the Bellman-Ford Algorithm is delineated in pseudocode \ref{fig:bellmanford-pseudocode}, illustrating its systematic procedure for traversing the graph and updating distance estimates.
	
	\begin{algorithm}
		\caption{Bellman-Ford Algorithm}
		\begin{algorithmic}[1]
			\Procedure{BellmanFord}{$G$, $s$}
			\State Initialize $\text{distance}\left[v\right] = \infty$ for all $v \in G.V$ except $\text{distance}\left[s\right] = 0$
			\State Initialize $\text{predecessor}\left[v\right] = \text{NIL}$ for all $v \in G.V$
			\For{$i = 1$ to $\left|G.V\right|-1$}
			\For{each edge $\left(u, v\right) \in G.E$}
			\If{$\text{distance}\left[u\right] + w\left(u, v\right) < \text{distance}\left[v\right]$}
			\State $\text{distance}\left[v\right] = \text{distance}\left[u\right] + w\left(u, v\right)$
			\State $\text{predecessor}\left[v\right] = u$
			\EndIf
			\EndFor
			\EndFor
			\For{each edge $\left(u, v\right) \in G.E$}
			\If{$\text{distance}\left[u\right] + w\left(u, v\right) < \text{distance}\left[v\right]$}
			\State \Return "Graph contains a negative-weight cycle"
			\EndIf
			\EndFor
			\State \Return $\text{distance}$, $\text{predecessor}$
			\EndProcedure
		\end{algorithmic}\label{fig:bellmanford-pseudocode}
	\end{algorithm}

	\subsection{Previous Work on ML and AI Interplay with the Bellman-Ford Algorithm}
	Recent studies have explored the integration of Machine Learning (ML) and Artificial Intelligence techniques with the Bellman-Ford algorithm, commonly utilized for determining shortest paths in graphs. These investigations aim to harness ML and AI capabilities to enhance the efficiency and precision of the algorithm. An example of such endeavors is as follows.
	
	One method introduces Neural Bellman-Ford Networks (NBFNet) \cite{zhu2021neural}, a framework that merges the Bellman-Ford algorithm with neural networks for graph link prediction. NBFNet exploits the path-based representation learning features of the Bellman-Ford algorithm and integrates them with neural network elements to enhance the accuracy of predicting links between nodes in the graph. This approach serves to connect conventional graph algorithms with neural network capabilities, demonstrating encouraging outcomes across various standard datasets.

\subsection{Algogenic Enhancements for Bellman-Ford}
\subsubsection{Graph Integrity Analysis}
\paragraph{Introduction to Graph Integrity Analysis}
The strategic integration of Graph Integrity Analysis within the Algogenic framework tailored for the Bellman-Ford algorithm marks a significant leap towards the refinement and optimization of graph-based computations. This enhancement, meticulously designed to leverage the advanced capabilities of generative AI, embarks on a comprehensive examination of the graph's structural framework prior to the algorithm's deployment. The primary objective of this rigorous analysis is to identify and rectify potential inefficiencies or structural anomalies that might otherwise hamper the algorithm's performance. Such inefficiencies include, but are not limited to, redundant edges that contribute no additional value to the graph's connectivity, superfluous nodes that complicate the graph without enhancing its informational content, or poorly configured connections that could potentially introduce computational complexity or, in worse cases, give rise to negative weight cycles that undermine the algorithm's integrity.

This proactive analysis phase is ingeniously designed as a strategic intervention aimed at streamlining the graph's topology, thereby enhancing its structural coherence and operational efficiency. This foundational optimization sets the stage for the Bellman-Ford algorithm to execute with enhanced precision and computational agility. By systematically purging the graph of extraneous elements and meticulously fine-tuning its structural framework, this preliminary phase effectively primes the algorithm for success. It equips the algorithm with the necessary agility to traverse the graph's landscape with enhanced precision and efficiency, navigating through the complexities of the graph with an unprecedented level of effectiveness.

Furthermore, the integration of generative AI into this preparatory process introduces an intelligent layer of adaptability and foresight. Through sophisticated data analysis techniques and advanced pattern recognition capabilities, the algorithm is endowed with deep insights into the intricacies of the graph's topology. This dynamic analysis capability enables the algorithm to uncover and leverage optimization opportunities that might remain elusive to traditional, static analysis methods. By adopting this flexible and responsive approach to graph analysis, the algorithm is effectively armored to adapt to and thrive in the face of evolving data dynamics and structural variations, significantly bolstering its robustness and resilience across a wide spectrum of real-world applications.

\paragraph{Implementing Graph Integrity Analysis}
The practical implementation of Graph Integrity Analysis as an Algogenic enhancement involves the strategic deployment of state-of-the-art large language models. These advanced models are tasked with conducting a deep and comprehensive analysis of the graph's topology, drawing upon a rich repository of historical data and leveraging the current configuration to unearth patterns, anomalies, or inefficiencies indicative of potential structural issues. This meticulous analysis encompasses a wide array of considerations, including, but not limited to, the assessment of the graph's susceptibility to negative weight cycles - a critical concern for the Bellman-Ford algorithm - and the examination of node connectivity and reachability to ensure a coherent and fully integrated graph structure. Furthermore, the analysis extends to scrutinize the distribution and configuration of edge weights, identifying any outliers or inconsistencies that may detrimentally impact the accuracy and efficiency of pathfinding operations. Additionally, this comprehensive evaluation includes an examination of the graph's density and its implications for computational efficiency, alongside an assessment of node centrality and the strategic importance of various nodes within the graph, thereby guiding optimization strategies to enhance pathfinding efficacy. The culmination of this analysis phase necessitates the development and refinement of sophisticated algorithms and visualization techniques, designed to facilitate a clear and intuitive understanding of the analysis outcomes, thereby enabling informed decision-making in the optimization of the graph's structure for optimal pathfinding performance.

\paragraph{Impact on the Bellman-Ford Algorithm}
The meticulous optimization of the graph's structure through the application of Graph Integrity Analysis profoundly enhances the performance capabilities of the Bellman-Ford algorithm. By proactively identifying and rectifying potential structural inefficiencies and vulnerabilities, the algorithm is empowered to operate with a heightened level of smoothness and efficiency. This optimization significantly reduces the computational burden by minimizing the number of iterations required to ascertain the shortest paths, thereby conserving computational resources and enhancing the algorithm's operational efficiency. Furthermore, by preemptively addressing and mitigating the risks associated with negative weight cycles, the algorithm is fortified to deliver more reliable and accurate pathfinding outcomes, thereby broadening its applicability and enhancing its reliability in navigating through the intricacies of complex networked environments.

Moreover, this enhanced structural foundation facilitates the Bellman-Ford algorithm's adaptability to dynamic network changes, ensuring that pathfinding operations remain both timely and accurate, even as the network topology evolves. Additionally, the strategic integration of Graph Integrity Analysis empowers the algorithm to leverage contextual insights derived from the graph's structure, thereby informing the selection of optimal paths in real-time, based on prevailing network conditions and constraints.

Furthermore, the targeted optimization of the graph's structure not only amplifies the algorithm's scalability and computational efficiency but also significantly reduces the likelihood of algorithmic failures or inconsistencies, thereby bolstering the overall robustness and reliability of the Bellman-Ford algorithm in practical implementations. This comprehensive approach to graph optimization, characterized by its proactive identification and resolution of structural inefficiencies, ensures that the algorithm remains resilient to noisy or incomplete data inputs, thereby maintaining its capacity to produce accurate and reliable pathfinding outcomes even in the most challenging and uncertain environments. Ultimately, the profound impact of Graph Integrity Analysis on the Bellman-Ford algorithm heralds a new era in algorithmic capabilities, significantly expanding the algorithm's utility and effectiveness across a diverse array of domains, ranging from telecommunications and transportation to finance and logistics, thereby revolutionizing its potential applications in the modern world.

\subsubsection{Dynamic Edge Weight Prediction}
\paragraph{Exploring Dynamic Edge Weight Prediction}
The introduction of Dynamic Edge Weight Prediction into the Algogenic framework specifically tailored for enhancing the Bellman-Ford algorithm represents a strategic and forward-thinking adaptation. This enhancement is meticulously designed to harness the predictive power of advanced generative AI technologies, enabling the algorithm to dynamically anticipate and adjust to fluctuations in edge conditions in real time. This capability is particularly crucial in addressing the inherent challenges presented by the dynamic nature of real-world applications, such as rapidly evolving transportation networks and communication systems, where external factors such as fluctuating traffic patterns, variable weather conditions, and network congestion can significantly impact the graph's topology and, consequently, the algorithm's pathfinding efficiency.

At the core of Dynamic Edge Weight Prediction lies its groundbreaking ability to forecast imminent changes within the graph's environment and proactively recalibrate edge weights to reflect these anticipated dynamics. By leveraging a sophisticated blend of historical data analysis and real-time environmental inputs, the algorithm is equipped to optimize its pathfinding strategies based on the most current and relevant information available. This adaptive mechanism ensures that the Bellman-Ford algorithm remains agile and responsive to the ever-changing conditions, delivering path solutions that are not only contextually relevant but also optimized for efficiency and effectiveness.

Moreover, the strategic incorporation of Dynamic Edge Weight Prediction significantly enhances the algorithm's resilience to transient disruptions and environmental fluctuations. Through a continuous process of learning and adaptation, the algorithm is capable of navigating dynamic graphs with an enhanced level of reliability, providing optimized path solutions that are robust against the volatility inherent in real-world conditions.

Furthermore, by synergizing with the Bellman-Ford algorithm's iterative approach to pathfinding, Dynamic Edge Weight Prediction seamlessly integrates predictive analytics into the algorithmic framework. This integration not only augments the algorithm's overall performance and efficacy but does so in a manner that is particularly well-suited to scenarios characterized by frequent and unpredictable changes in edge weights. This enhancement represents a paradigm shift in the field of graph-based pathfinding algorithms, heralding a new era of adaptability and responsiveness that promises to transform the landscape of dynamic environment navigation.

\paragraph{Implementing Predictive Weight Adjustments}
The practical implementation of Dynamic Edge Weight Prediction necessitates the deployment of sophisticated large language models, which are tasked with the complex analysis of both historical and contemporary data relevant to the graph's edges. These cutting-edge models are employed to generate accurate forecasts regarding future changes in edge conditions, thereby enabling the algorithm to preemptively adjust edge weights in anticipation of these changes. For instance, should an LLM predict a significant increase in traffic along a particular route, the corresponding edge in the graph would see an adjustment in its weight to account for the expected delay, effectively recalibrating the algorithm's pathfinding strategy before these conditions manifest in reality.

This intricate process of predictive weight adjustment requires not only a deep understanding of the multifaceted factors influencing edge weights but also a robust framework for accurately forecasting their future trajectories. Additionally, the implementation phase involves rigorous data preprocessing to ensure the accuracy and relevance of the input data, encompassing techniques such as data normalization, feature engineering, and outlier detection. Moreover, the development of efficient model training pipelines is essential for continuously updating the LLMs with fresh data, thereby refining their predictive capabilities over time.

Furthermore, the successful integration of predictive weight adjustments into the Bellman-Ford algorithm's operational framework also brings to the fore considerations regarding computational resource allocation. The real-time or near-real-time adjustment of edge weights, particularly in large-scale graphs or under conditions of frequent environmental change, demands significant computational power. As such, the deployment of scalable infrastructure and the application of parallel processing techniques become imperative to manage the computational load effectively.

Moreover, ethical considerations surrounding data privacy and the mitigation of algorithmic bias assume critical importance in the context of predictive weight adjustments. Ensuring that the data used for training the LLMs is ethically sourced and that the predictive models do not inadvertently perpetuate existing biases or discriminate against specific groups is of paramount importance. Establishing mechanisms for transparency and accountability is crucial in addressing any potential biases or errors inherent in the predictive models.

In conclusion, the implementation of Dynamic Edge Weight Prediction within the Bellman-Ford algorithm represents a significant technological advancement, offering the potential to significantly enhance pathfinding efficiency and accuracy by dynamically adapting to anticipated changes in edge conditions. However, this implementation also presents a series of challenges, including data preprocessing, computational resource management, and ethical considerations, that must be carefully navigated to fully realize the transformative potential of this enhancement.

\paragraph{Benefits to Pathfinding with the Bellman-Ford Algorithm}
The integration of Dynamic Edge Weight Prediction within the Bellman-Ford algorithm framework marks a significant milestone in the evolution of pathfinding algorithms, imbuing the algorithm with a newfound level of adaptability and foresight. This Algogenic enhancement enables the algorithm to not merely react to changes in edge conditions but to anticipate and adapt to these changes proactively. By doing so, the algorithm transcends the limitations of static pathfinding methodologies, offering solutions that are not only optimized for the current state of the graph but also resilient against imminent changes. This predictive capability fundamentally transforms the nature of pathfinding, converting a traditionally reactive process into a dynamic, forward-looking system that continuously aligns its strategies with the evolving landscape of the graph.

Moreover, the ability to dynamically adjust to changing edge conditions significantly enhances the Bellman-Ford algorithm's utility and reliability in dynamic environments. By maintaining an up-to-date understanding of the graph's topology and adapting its pathfinding strategies accordingly, the algorithm ensures that the paths it identifies remain optimal even as actual conditions evolve. This enhancement effectively minimizes the risk of relying on outdated or suboptimal paths, thereby increasing the overall reliability and utility of the Bellman-Ford algorithm in applications where edge conditions are subject to rapid and unpredictable changes.

Furthermore, the synergistic relationship between Dynamic Edge Weight Prediction and the iterative nature of the Bellman-Ford algorithm creates a powerful feedback loop that continuously refines the algorithm's pathfinding outcomes. As the algorithm iterates through the graph, updating distance estimates and identifying the shortest paths, the predictive component of the enhancement continuously fine-tunes its forecasts based on the latest patterns and trends. This iterative refinement process fosters a cycle of continuous improvement, enhancing the accuracy and efficiency of the algorithm's pathfinding decisions.

Additionally, the adaptive capabilities conferred by Dynamic Edge Weight Prediction render the Bellman-Ford algorithm particularly well-suited to a wide range of dynamic routing scenarios. From managing network traffic and optimizing transportation logistics to allocating resources in distributed systems, the ability to anticipate and adapt to fluctuations in edge weights is indispensable for maintaining efficient and reliable pathfinding operations. In these contexts, where conditions can change swiftly and unpredictably, the enhanced Bellman-Ford algorithm, equipped with predictive insights, becomes a critical tool for navigating the complexities of modern networks.

In essence, the incorporation of Dynamic Edge Weight Prediction into the Bellman-Ford algorithm represents a profound enhancement to its capabilities, elevating the algorithm beyond its traditional confines and establishing it as a versatile and powerful instrument for dynamic environment navigation. By augmenting the algorithm's inherent strengths with advanced predictive analytics, this enhancement not only boosts its performance and efficacy but also expands its applicability across a diverse array of domains, heralding a new era of intelligent, adaptive pathfinding solutions capable of meeting the challenges of dynamic and complex environments.

\subsubsection{Predictive Negative Cycle Detection}
\paragraph{Introduction to Predictive Negative Cycle Detection}
Integrating Predictive Negative Cycle Detection into the sophisticated Algogenic framework specifically optimized for the Bellman-Ford algorithm represents a pioneering advancement in algorithmic design. This cutting-edge enhancement leverages the formidable predictive prowess of generative AI to proactively identify and mitigate the formation of negative weight cycles within the graph. Such cycles pose a significant challenge to the integrity and reliability of pathfinding algorithms, potentially distorting the computation of shortest paths and undermining the algorithm's overall performance. The introduction of this predictive analysis capability signifies a strategic shift towards a more anticipatory approach to algorithmic optimization, wherein potential disruptions to the pathfinding process are addressed preemptively, thereby ensuring a higher level of accuracy and reliability in the algorithm's outputs.

The essence of Predictive Negative Cycle Detection lies in its proactive identification and mitigation strategy. By harnessing the advanced analytical capabilities of generative AI, this enhancement empowers the Bellman-Ford algorithm with the ability to forecast the emergence of negative weight cycles based on an extensive analysis of historical data, current graph configurations, and a deep understanding of the underlying patterns and trends that may contribute to such phenomena. This forward-looking perspective enables the algorithm to implement corrective measures before the negative cycles materialize, effectively safeguarding the pathfinding process from potential distortions and ensuring the integrity of the computed paths. Furthermore, the adaptive nature of this enhancement ensures that the algorithm's predictive models are continuously refined based on evolving data and structural variations within the graph, thereby maintaining the algorithm's effectiveness and reliability even in the face of dynamic and unpredictable environmental changes.

Moreover, the integration of Predictive Negative Cycle Detection significantly contributes to the efficiency and scalability of the Bellman-Ford algorithm. By preemptively identifying and addressing the risks associated with negative weight cycles, the algorithm is able to streamline its computational processes, reducing unnecessary iterations and enhancing its overall performance. This efficiency gain is particularly valuable in large-scale networks or in applications where real-time responses are critical, as it enables the algorithm to deliver timely and accurate pathfinding results with a reduced computational footprint. Additionally, the predictive capabilities of this enhancement equip the algorithm with a heightened level of adaptability, allowing it to navigate complex and dynamic graph environments with unprecedented agility and precision.

In summary, Predictive Negative Cycle Detection represents a transformative enhancement to the Bellman-Ford algorithm, introducing a novel dimension of predictive analytics to the realm of algorithmic optimization. By preemptively identifying potential disruptions to the pathfinding process and implementing strategic corrective measures, this enhancement not only enhances the accuracy and reliability of the algorithm's outputs but also significantly improves its computational efficiency and scalability. Furthermore, the introduction of this predictive capability encourages further exploration and innovation in the field of algorithmic design, opening new avenues for research and development in the optimization of pathfinding algorithms and beyond.

\paragraph{Implementing Predictive Cycle Detection}
The implementation of Predictive Negative Cycle Detection within the Bellman-Ford algorithm's operational framework involves the strategic deployment of advanced large language models, which are meticulously engineered to analyze and interpret a wide array of data points and indicators that may signify the potential for negative weight cycle formation. These sophisticated models leverage an extensive dataset comprising historical performance metrics, current graph configurations, and external factors that could influence the graph's topology, employing advanced machine learning algorithms to identify patterns, anomalies, and trends indicative of impending negative cycles. By synthesizing this wealth of information, LLMs are able to generate predictive insights that guide the algorithm in making informed adjustments to the graph or its operational parameters, thereby preventing the formation of negative weight cycles or mitigating their impact on the pathfinding process.

This proactive approach to cycle detection represents a significant departure from traditional, reactive methodologies, wherein negative cycles are addressed post-formation, often resulting in suboptimal pathfinding outcomes and diminished algorithmic reliability. By integrating predictive cycle detection into the Bellman-Ford algorithm, a paradigm shift is achieved, emphasizing preventive measures and strategic foresight over reactive corrections. This shift not only enhances the robustness and reliability of the pathfinding process but also instills a greater level of confidence in the algorithm's performance across a diverse range of applications and conditions.

Furthermore, the predictive nature of this enhancement enables the Bellman-Ford algorithm to dynamically adapt to changing graph dynamics and environmental factors, ensuring that its performance remains optimal even as the underlying graph structure evolves or external influences fluctuate. This dynamic adaptability is achieved through the continuous refinement of the LLMs' predictive models, which are regularly updated with new data and insights, allowing the algorithm to stay ahead of potential disruptions and maintain its pathfinding efficiency and accuracy over time.

Additionally, the incorporation of Predictive Negative Cycle Detection introduces a new layer of sophistication to the pathfinding process, enabling the Bellman-Ford algorithm to detect and respond to subtle patterns and correlations that may not be immediately apparent through conventional analytical techniques. This enhanced analytical capability allows the algorithm to proactively identify and address potential risks, aligning with the principles of preventive maintenance and strategic risk management. By adopting this proactive stance, the likelihood of pathfinding failures or suboptimal outcomes is significantly reduced, thereby enhancing the overall system reliability and user satisfaction.

In summary, the implementation of Predictive Negative Cycle Detection within the Bellman-Ford algorithm represents a significant technological advancement, offering a host of benefits including early risk identification, proactive adaptation to changing conditions, and enhanced algorithmic reliability. By leveraging the advanced predictive analytics capabilities of large language models, the algorithm is equipped to anticipate and mitigate potential negative cycle formations, ensuring smoother and more efficient navigation through complex graph environments. This implementation not only underscores the importance of predictive analytics in enhancing pathfinding algorithms but also highlights the potential for generative AI to revolutionize traditional algorithmic approaches, paving the way for more intelligent and adaptive solutions in the field of computational graph analysis.

\paragraph{Impact on the Bellman-Ford Algorithm}
The incorporation of Predictive Negative Cycle Detection within the Bellman-Ford algorithm framework significantly elevates its capabilities, transforming it into a more robust and resilient computational tool capable of navigating the complexities of modern networked environments. This Algogenic enhancement not only shields the algorithm from the disruptive effects of negative weight cycles but also ensures that the shortest path calculations remain accurate and meaningful, even in dynamically changing landscapes. By preemptively identifying and addressing potential negative cycles, the algorithm is able to offer more stable and reliable pathfinding solutions, thereby increasing its utility and applicability across a wide range of complex network scenarios where conditions and configurations may vary frequently.

Moreover, the integration of this predictive enhancement introduces a proactive element to the algorithm's decision-making process, allowing it to anticipate and mitigate potential negative cycles before they exert their influence on the pathfinding outcomes. This proactive approach aligns with the principles of adaptive and anticipatory computing, where systems are designed to actively foresee and adapt to changing conditions in order to achieve optimal performance. By embracing this forward-looking perspective, the Bellman-Ford algorithm transcends the limitations of reactive computing models, establishing itself as a leader in the domain of adaptive pathfinding solutions.

Furthermore, the introduction of Predictive Negative Cycle Detection into the Bellman-Ford algorithm does not compromise its computational efficiency. While the enhancement adds a layer of complexity to the algorithm's operations, the benefits gained in terms of improved accuracy, stability, and resilience far outweigh the additional computational overhead. This careful balancing act between complexity and performance is critical, especially in applications where the reliability and timeliness of pathfinding results are of paramount importance. By optimizing the algorithm's performance to accommodate the predictive analysis of negative cycles, a new benchmark is set for efficiency and reliability in the field of pathfinding algorithms.

Additionally, the enhanced Bellman-Ford algorithm, equipped with Predictive Negative Cycle Detection, opens up new possibilities for its application in domains where path reliability is a critical concern. In transportation networks, for example, where safety and efficiency are paramount, the ability to predict and preempt negative cycles can significantly enhance route planning and optimization efforts. By providing a more reliable and stable foundation for pathfinding, the algorithm contributes to the development of safer and more efficient transportation systems, ultimately benefiting both operators and users alike.

In essence, the integration of Predictive Negative Cycle Detection within the Bellman-Ford algorithm represents a significant leap forward in the evolution of pathfinding technologies. By incorporating advanced predictive analytics into the algorithm's framework, a new level of adaptability and foresight is achieved, enabling the algorithm to navigate complex and dynamic environments with unparalleled precision and reliability. This enhancement not only showcases the potential of Algogenic approaches to transform traditional algorithmic methodologies but also highlights the critical role of predictive analytics in advancing the field of computational graph analysis. As the Bellman-Ford algorithm continues to evolve and adapt to the challenges of modern networked environments, the integration of Predictive Negative Cycle Detection stands as a testament to the power of innovation and foresight in shaping the future of algorithmic design and optimization.

\subsubsection{Adaptive Relaxation Optimization}
\paragraph{Redefining Relaxation in Bellman-Ford}
Adaptive Relaxation Optimization heralds a transformative enhancement in the Algogenic framework tailored for the Bellman-Ford algorithm, establishing a new paradigm in the algorithm's approach to relaxation. Traditionally, the Bellman-Ford algorithm updates the distances to nodes iteratively, treating all edges with uniform consideration. However, Adaptive Relaxation Optimization introduces a dynamic, informed strategy that leverages the analytical prowess of generative AI to identify and prioritize edges or nodes based on their potential impact on the efficiency and accuracy of finding the shortest paths.

This innovative approach to relaxation in the Bellman-Ford algorithm is not merely an incremental improvement but a strategic overhaul that infuses the algorithm with a level of adaptability and foresight previously unattainable. By integrating generative AI, the algorithm gains the capability to dynamically adjust its focus, channeling computational resources toward parts of the graph that are most likely to influence optimal pathfinding outcomes. This targeted optimization does not only streamline the search for the shortest paths but also significantly enhances the computational efficiency of the algorithm, reducing the time and resources required to achieve convergence.

Furthermore, the introduction of Adaptive Relaxation Optimization imbues the Bellman-Ford algorithm with a remarkable resilience to changes in graph structure or edge weights. Where traditional static approaches may falter in the face of such dynamism, this enhancement ensures that the algorithm remains robust and responsive, continuously fine-tuning its relaxation strategy to align with the evolving topology of the graph.

In essence, Adaptive Relaxation Optimization revolutionizes the Bellman-Ford algorithm, transforming it into a more agile, efficient, and adaptive tool for tackling the challenges of shortest path problems in complex and dynamic graph environments. This groundbreaking approach not only promises significant improvements in algorithmic performance but also opens new avenues for the application of generative AI in enhancing traditional algorithmic methodologies.

\paragraph{Implementing Adaptive Relaxation}
Implementing Adaptive Relaxation Optimization within the Bellman-Ford algorithm involves a sophisticated application of large language models to conduct an in-depth analysis of the graph's structure. This analysis meticulously evaluates historical pathfinding outcomes, current edge weights, and the overall topology of the graph to uncover patterns, trends, and key insights that could inform a more efficient relaxation strategy. Armed with this comprehensive understanding, the algorithm is then able to dynamically adjust its approach, prioritizing edges and nodes that have shown to be critical in influencing pathfinding efficacy.

The deployment of Adaptive Relaxation Optimization signifies a strategic shift towards a more intelligent, data-driven approach to algorithmic problem-solving. By prioritizing critical components of the graph, the algorithm optimizes the allocation of computational resources, enhancing both the speed and accuracy of the pathfinding process. This dynamic adjustment is made possible through the continuous analysis of real-time data, allowing the algorithm to adapt its strategy in response to changes within the graph or in the external environment.

This process of implementing Adaptive Relaxation Optimization not only enhances the performance of the Bellman-Ford algorithm but also significantly increases its adaptability and precision in navigating complex graph structures. Through the integration of advanced generative AI capabilities, the algorithm is equipped to identify and exploit optimization opportunities that traditional methods might overlook, ensuring that it remains at the forefront of algorithmic efficiency and effectiveness.

\paragraph{Enhancing Efficiency and Accuracy}
The incorporation of Adaptive Relaxation Optimization into the Bellman-Ford algorithm marks a pivotal advancement in the pursuit of enhanced computational efficiency and algorithmic accuracy. By intelligently focusing computational efforts on the most influential parts of the graph, this enhancement reduces unnecessary calculations and accelerates the pathfinding process, enabling the algorithm to converge on optimal solutions with unprecedented speed.

Moreover, this targeted approach to relaxation not only optimizes the algorithm's performance but also significantly improves the accuracy of the resulting paths. By dynamically adjusting the relaxation process based on a nuanced understanding of the graph's topology and the relative importance of its components, the algorithm ensures that updates to node distances are prioritized in a manner that directly contributes to the discovery of the shortest and most reliable paths.

Additionally, the adaptability introduced through Adaptive Relaxation Optimization ensures that the Bellman-Ford algorithm remains effective and efficient across a broad spectrum of graph environments. Whether faced with static or dynamically changing graphs, the enhanced algorithm is capable of adjusting its strategy to maintain optimal performance, demonstrating a level of flexibility and resilience that sets a new standard in the field of pathfinding algorithms.

In summary, the integration of Adaptive Relaxation Optimization represents a significant leap forward in the development of more efficient, accurate, and adaptable pathfinding algorithms. By leveraging the capabilities of generative AI to refine and optimize the relaxation process, this enhancement not only improves the operational efficiency of the Bellman-Ford algorithm but also ensures its continued relevance and effectiveness in addressing the complexities of modern graph-based problems.

\subsubsection{Path Contextualization and Enhancement}
\paragraph{Broadening Pathfinding Perspectives}
The strategic incorporation of Path Contextualization and Enhancement within the Algogenic framework tailored for the Bellman-Ford algorithm signifies a monumental shift towards a more nuanced and holistic approach to pathfinding. This innovative enhancement leverages the advanced capabilities of generative AI, specifically large language models, to imbue the algorithm with the ability to consider not only the quantitative metrics traditionally associated with path optimization but also a diverse array of qualitative factors. These factors encompass user preferences, environmental considerations, and other context-specific attributes such as safety, scenic value, or cultural significance, thereby transforming the pathfinding output from a mere numerical solution into a rich, multifaceted decision-making tool that resonates with the complex needs and preferences of users.

By extending the algorithm's focus beyond the simplistic minimization of cost or distance, Path Contextualization and Enhancement introduces a paradigm where paths are evaluated and recommended based on a comprehensive understanding of the broader context in which they exist. This approach is particularly relevant in applications such as urban planning, where the selection of routes is influenced by a multitude of factors beyond physical distance, including the promotion of pedestrian-friendly pathways, the avoidance of areas with high crime rates, or the inclusion of routes that pass through areas of historical or cultural interest. In such scenarios, the enhanced Bellman-Ford algorithm emerges as a powerful tool capable of delivering tailored solutions that cater to the diverse and often complex requirements of modern urban environments.

Moreover, the implications of this enhanced pathfinding approach extend well beyond the realm of navigation and transportation. In sectors such as logistics and supply chain management, where factors like traffic congestion, road conditions, and delivery time windows critically impact operational efficiency, the integration of contextual insights can lead to the identification of routes that optimize not just time and cost but also reliability and sustainability. This broader perspective enables organizations to make more informed decisions that align with their strategic objectives and operational constraints, thereby driving improvements in efficiency, customer satisfaction, and environmental sustainability.

Additionally, the ability to incorporate qualitative attributes into the pathfinding process opens up new avenues for innovation and creativity across a wide range of applications. By considering factors such as environmental impact, cultural heritage, and community engagement, the enhanced algorithm can contribute to the advancement of sustainable development goals, promote cultural appreciation and preservation, and foster a greater sense of community and belonging among users. This holistic approach to pathfinding not only elevates the utility of the Bellman-Ford algorithm but also underscores the role of technology in addressing some of the most pressing challenges facing society today.

In conclusion, the integration of Path Contextualization and Enhancement into the Bellman-Ford algorithm represents a significant advancement in the field of pathfinding, marking a departure from traditional optimization techniques towards a more inclusive and context-aware approach. By leveraging the power of generative AI to integrate a wide range of qualitative factors into the pathfinding process, this enhancement ensures that the solutions provided are not only efficient and effective but also aligned with the broader societal, environmental, and cultural values. As such, the enhanced Bellman-Ford algorithm stands as a testament to the potential of Algogenic enhancements to transform conventional algorithmic paradigms, paving the way for the development of more intelligent, adaptive, and human-centric computational tools.

\paragraph{Implementing Contextual and Qualitative Enhancements}
The implementation of Path Contextualization and Enhancement within the Bellman-Ford algorithm framework entails a sophisticated application of large language models to analyze and synthesize a vast array of contextual information pertinent to the paths identified by the algorithm. This process begins with the meticulous collection and integration of diverse data sources, ranging from geographical and environmental datasets to cultural and socio-economic databases, in order to construct a comprehensive contextual landscape surrounding each potential path. Leveraging state-of-the-art data processing techniques, including natural language processing (NLP) and sentiment analysis, the LLMs are equipped to interpret and evaluate qualitative aspects of the environment, such as the aesthetic appeal of scenic routes, the historical significance of landmarks, or the safety and accessibility of urban pathways.

By integrating this rich contextual information into the pathfinding algorithm, the paths generated are enriched with layers of qualitative insights, transforming them from mere routes into curated experiences that resonate with the users' values and preferences. For instance, in the context of urban navigation, the algorithm can highlight paths that not only optimize travel time but also offer pedestrian-friendly routes, access to green spaces, or proximity to cultural landmarks, thereby enriching the user's journey with meaningful interactions and experiences.

Furthermore, the enhancement process involves a dynamic evaluation of paths against a set of user-defined preferences and broader societal criteria. This allows the algorithm to assign additional value to paths based on factors such as environmental sustainability, safety, and cultural richness, moving beyond the traditional focus on efficiency to embrace a more holistic approach to route optimization. By doing so, the algorithm not only meets the individual needs of users but also contributes to the achievement of collective societal goals, such as reducing carbon emissions, enhancing public safety, and promoting cultural heritage.

The successful implementation of Path Contextualization and Enhancement also requires continuous refinement and adaptation to ensure that the algorithm remains responsive to evolving data landscapes and changing user preferences. As new information becomes available and societal values shift, the algorithm must be capable of integrating these changes into its decision-making process, thereby ensuring that the paths recommended remain relevant and aligned with current needs and expectations. This iterative process of learning and adaptation is facilitated by the advanced capabilities of generative AI, which enable the algorithm to update its contextual understanding in real-time, ensuring that the pathfinding solutions offered are not only accurate and efficient but also reflective of the complex and dynamic world in which we live.

In summary, the implementation of Path Contextualization and Enhancement represents a significant leap forward in the evolution of pathfinding algorithms, offering a more nuanced and user-centric approach to route optimization. By harnessing the power of generative AI to integrate a wide range of contextual information into the pathfinding process, this enhancement not only improves the relevance and utility of the paths recommended but also opens up new possibilities for the application of technology in addressing complex societal challenges. As the Bellman-Ford algorithm continues to evolve and adapt to the needs of modern users, the integration of contextual and qualitative enhancements stands as a powerful example of how Algogenic approaches can transform traditional computational methodologies into more intelligent, adaptive, and human-centered solutions.

\paragraph{Impact on Path Selection and User Experience}
The integration of Path Contextualization and Enhancement into the Bellman-Ford algorithm framework ushers in a new era of pathfinding, one characterized by a profound transformation in the user experience and a significant expansion in the algorithm's applicability and relevance. By incorporating a rich tapestry of contextual and qualitative factors into the pathfinding process, this Algogenic enhancement elevates the algorithm from a tool for identifying the shortest or least costly paths to a sophisticated decision-support system that offers tailored, context-aware route recommendations. This shift not only enhances the practical utility of the algorithm but also imbues the pathfinding experience with a deeper sense of personalization and relevance, fostering a closer alignment between the algorithm's outputs and the users' real-world needs and preferences.

The inclusion of context-awareness within the algorithm's decision-making framework significantly enhances its adaptability to a wide range of user scenarios and objectives. This flexibility allows the algorithm to cater to diverse user preferences, from prioritizing scenic beauty and cultural enrichment to ensuring safety and accessibility, thereby delivering route recommendations that are not only optimized for physical parameters but also enriched with values that matter to the users. As a result, users are empowered with a broader spectrum of choices that reflect their individual priorities and circumstances, fostering a sense of agency and satisfaction with the algorithm's recommendations.

Moreover, the ability to integrate contextual and qualitative insights into the path selection process not only optimizes the efficiency and relevance of the routes identified but also contributes to a more enriching and meaningful user experience. By offering paths that resonate with users on multiple levels, the enhanced algorithm transcends the limitations of traditional pathfinding approaches, opening up new dimensions of engagement and interaction between users and their environment. This holistic approach to navigation not only facilitates more informed and satisfying travel decisions but also encourages a deeper connection with the surrounding world, enhancing the overall quality and enjoyment of the journey.

Furthermore, the adaptive nature of Path Contextualization and Enhancement ensures that the algorithm remains responsive and relevant in the face of changing environmental conditions and evolving user preferences. By continuously updating its contextual understanding and refining its recommendations based on real-time data and feedback, the algorithm demonstrates a remarkable capacity for learning and adaptation, ensuring that its pathfinding solutions remain aligned with the dynamic complexities of modern life. This ongoing process of refinement and improvement not only underscores the algorithm's commitment to delivering high-quality, user-centric solutions but also highlights the potential of Algogenic enhancements to drive continuous innovation and progress in the field of computational algorithms.

In summary, the integration of Path Contextualization and Enhancement within the Bellman-Ford algorithm represents a paradigm shift in the domain of pathfinding, introducing a new level of sophistication and user-centricity to the algorithm's capabilities. By leveraging the power of generative AI to incorporate a wide array of contextual and qualitative factors into the pathfinding process, this enhancement not only improves the relevance and utility of the routes recommended but also significantly enhances the user experience, fostering a more engaging, personalized, and meaningful interaction with the algorithm. As the Bellman-Ford algorithm continues to evolve and adapt to the needs of modern users, the integration of Path Contextualization and Enhancement stands as a testament to the transformative potential of Algogenic approaches in redefining the landscape of algorithmic design and optimization, paving the way for the development of more intelligent, adaptive, and human-centric computational tools.

\subsubsection{Outcome-based Learning for Weight Adjustment}
\paragraph{Enhancing Algorithmic Adaptability}
Outcome-based Learning for Weight Adjustment introduces an evolutionary leap in the Bellman-Ford algorithm's capability, embedding a self-optimizing mechanism that dynamically refines edge weights based on real-world path outcomes. This Algogenic enhancement utilizes the analytical prowess of generative AI to assimilate feedback from actual path traversals, assessing discrepancies between predicted and realized path efficiencies. This continuous learning loop allows for the nuanced adjustment of edge weights, ensuring the algorithm's path predictions evolve in accuracy and reliability. Such a mechanism not only enhances the algorithm's adaptability to changing network conditions but also significantly boosts its performance, making it a paragon of modern, data-driven pathfinding methodologies.

\paragraph{Implementing Feedback Loops}
The practical implementation of this enhancement requires the establishment of sophisticated feedback loops that capture and analyze path outcome data. This data, encompassing aspects like actual travel times versus predicted, cost variances, and user feedback, feeds into generative AI models. These models, through deep learning and pattern recognition, adjust the graph's edge weights, enhancing the fidelity of future path predictions. This process of iterative refinement and adjustment marks a significant departure from static pathfinding approaches, propelling the Bellman-Ford algorithm into a new era of dynamic, responsive pathfinding.

\paragraph{Optimizing Pathfinding Over Time}
Incorporating Outcome-based Learning for Weight Adjustment ensures the Bellman-Ford algorithm not only adapts to present conditions but also anticipates future changes, maintaining its efficiency and accuracy over time. This foresight, grounded in real-world performance data, transforms the algorithm into a continually evolving entity, capable of self-optimization and increasingly accurate path prediction. Such dynamic adaptability underscores the algorithm's enhanced utility across a spectrum of applications, from urban navigation to logistics and beyond, where changing conditions are the norm.

\subsubsection{Path Reliability and Risk Analysis}
\paragraph{Assuring Path Stability and Safety}
Path Reliability and Risk Analysis significantly elevates the Bellman-Ford algorithm's utility by integrating a multifaceted assessment of path stability and safety. Leveraging generative AI, this enhancement evaluates paths not just on efficiency but on their resilience to disruptions and safety metrics. This forward-thinking approach ensures that paths recommended by the algorithm are not only optimal in terms of distance or time but also reliable and secure, meeting the high standards required in critical applications like emergency response and infrastructure management.

\paragraph{Implementing Reliability and Risk Assessments}
Implementing this enhancement entails the use of generative AI to perform comprehensive risk assessments, analyzing data from various sources, including historical performance, current conditions, and predictive insights into future risks. This analysis facilitates the development of a reliability score for each path, enabling users to make informed decisions based on a holistic understanding of potential risks and benefits. This advanced capability transforms the Bellman-Ford algorithm into a more versatile and dependable tool for pathfinding, capable of navigating the complexities of the modern world with enhanced precision and insight.

\paragraph{Enhancing Decision-Making with Comprehensive Insights}
By incorporating Path Reliability and Risk Analysis, the Bellman-Ford algorithm becomes a pivotal tool in decision-making processes, offering insights that extend beyond traditional pathfinding metrics. This comprehensive evaluation framework empowers users to select paths that align with broader objectives, such as minimizing environmental impact, enhancing public safety, or ensuring operational resilience. The integration of this Algogenic enhancement not only broadens the algorithm's applicability but also reinforces its role in supporting sustainable, safe, and reliable navigation and logistical planning.

	\subsubsection{Pseudocode for Algogenic Bellman-Ford}
	The Algogenic Bellman-Ford approach utilizes AI to enhance the traditional Bellman-Ford algorithm by dynamically adjusting algorithm parameters and strategies based on observed behavior and real-time error estimates. This pseudocode, available in \ref{fig:bellman-ford-Algogen-pseudocode}, outlines an advanced framework incorporating AI-driven enhancements for adaptive scheme selection, domain partitioning, error estimation, and real-time parameter optimization.
	
	\begin{algorithm}
		\caption{Algogenic Bellman-Ford Pseudocode}
		\begin{algorithmic}[1]
			\Procedure{AlgogenicBellmanFord}{Graph, Source}
			\State PerformGraphIntegrityAnalysis(Graph) \Comment{Optimize graph structure}
			\State InitializeDistancesAndPredecessors(Graph, Source)
			\For{$i = 1$ to $|V|-1$} \Comment{Iterate |V|-1 times over all edges}
			\For{each edge $(u, v)$ in Graph}
			\State PredictiveWeightAdjustment($(u, v)$) \Comment{Adjust edge weights dynamically}
			\If{distance[u] + weight$(u, v)$ < distance[v]}
			\State distance[v] = distance[u] + weight$(u, v)$
			\State predecessor[v] = u
			\EndIf
			\EndFor
			\State AdaptiveRelaxationOptimization() \Comment{Optimize relaxation steps}
			\EndFor
			\State CheckForNegativeWeightCycles(Graph) \Comment{Use predictive cycle detection}
			\For{each vertex $v$ in Graph}
			\If{Path to $v$ exists}
			\State Path[$v$] = ReconstructPath(Source, $v$, predecessor)
			\State EnhancePath(Path[$v$]) \Comment{Contextualize and enhance path}
			\EndIf
			\EndFor
			\State ApplyOutcomeBasedLearning(Graph) \Comment{Learn from path outcomes}
			\State \Return Path
			\EndProcedure
		\end{algorithmic}\label{fig:bellman-ford-Algogen-pseudocode}
	\end{algorithm}

	\begin{figure}
		\centering
		\includegraphics[width=0.47\textwidth]{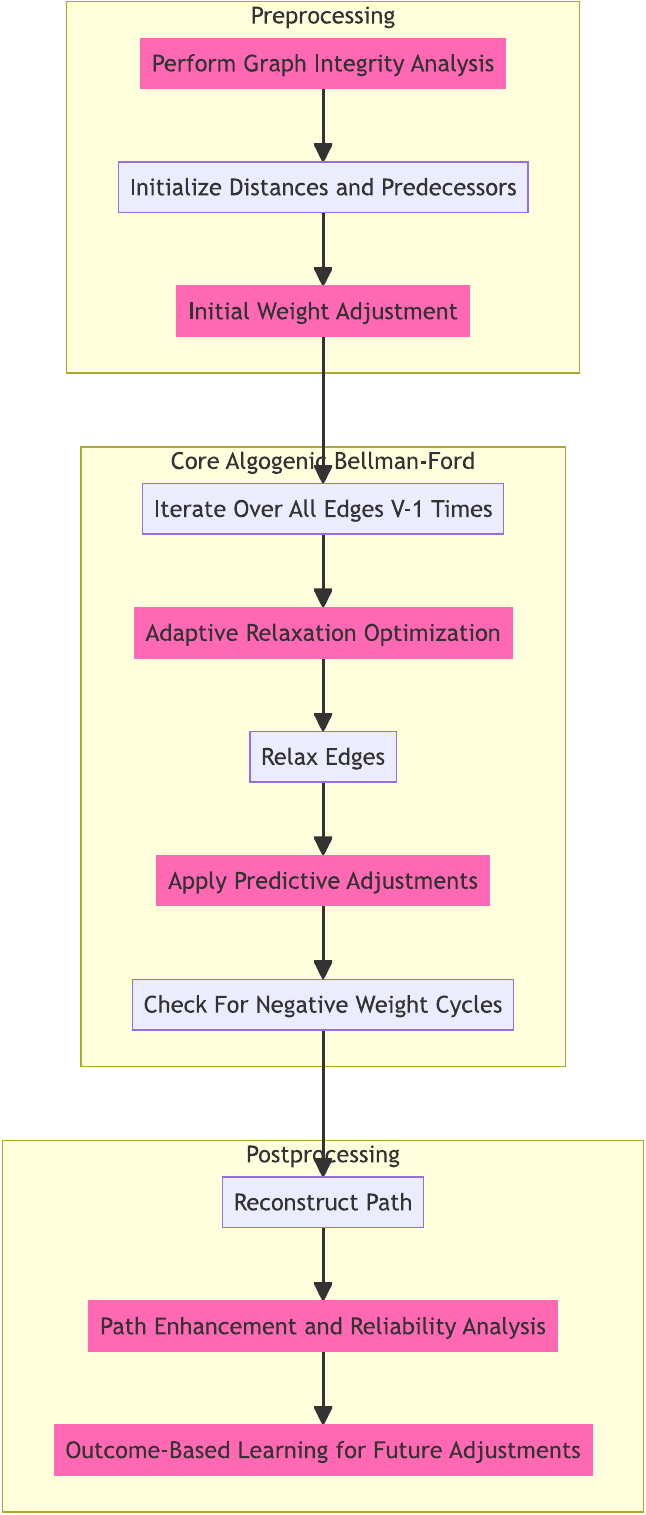} 
		\caption{Integrating Algogenic Enhancements into the Bellman-Ford Algorithm: This diagram presents the structured implementation of Algogenic enhancements within the Bellman-Ford algorithm, highlighting the strategic application of generative AI across various phases of the algorithm. In the preprocessing phase, 'Perform Graph Integrity Analysis' and 'Initial Weight Adjustment' utilize generative AI to analyze and optimize the graph's structure and adjust edge weights based on predictive insights, preparing the graph for more effective pathfinding. The core phase introduces 'Adaptive Relaxation Optimization' before edge relaxation, applying generative AI to optimize the relaxation process, followed by 'Apply Predictive Adjustments' to dynamically adjust edge weights in response to evolving conditions, enhancing the algorithm's adaptability to real-time changes. The postprocessing phase sees 'Path Enhancement and Reliability Analysis' where paths are enriched with contextual information and assessed for reliability through generative AI, ensuring the recommended paths are not only efficient but also aligned with broader qualitative criteria. 'Outcome-Based Learning for Future Adjustments' concludes the process, enabling continuous refinement of the algorithm based on the outcomes of previous executions. This Algogenic approach significantly augments the traditional Bellman-Ford algorithm, offering a more dynamic, responsive, and insightful pathfinding solution suitable for complex and changing environments.}
		\label{fig:bellman_ford}
	\end{figure}

	
	\chapterimage{pngs/optimization.png} 
	
	\chapter{Optimization Algogens}\index{Optimization Algogens}
	
	\section{Genetic Algorithms}\index{Genetic Algorithms}
	
	\subsection{Introduction to Genetic Algorithms}
	\paragraph{Overview of Genetic Algorithms}
	Genetic Algorithms (GAs) stand out as a powerful computational approach rooted in the principles of natural selection and genetics, offering a versatile solution to optimization and search problems across diverse domains. At their core, GAs mimic the process of natural evolution by iteratively evolving a population of candidate solutions to find the most suitable one. This iterative process involves several key components, including selection, crossover, and mutation.
	
	In a typical GA, the initial population comprises a set of potential solutions encoded as strings of parameters, often referred to as chromosomes or individuals. These individuals are then evaluated based on a predefined objective function, which measures their fitness or suitability for the given problem. The selection process, inspired by the concept of survival of the fittest, determines which individuals are allowed to proceed to the next generation based on their fitness scores. 
	
	Next, the selected individuals undergo crossover and mutation operations to produce offspring, thereby introducing diversity and exploration into the population. During crossover, pairs of parent individuals exchange genetic information to generate new solutions, while mutation introduces random changes to further diversify the population and prevent premature convergence. These operations, coupled with the selection mechanism, drive the iterative improvement of solutions over successive generations.
	
	One of the distinguishing features of GAs is their ability to search large solution spaces efficiently and effectively, even in the presence of complex constraints or non-linear relationships. Moreover, GAs offer a high degree of flexibility and scalability, making them suitable for tackling a wide variety of optimization problems, ranging from engineering design and scheduling to financial modeling and machine learning.
	
	In summary, Genetic Algorithms represent a sophisticated yet intuitive approach to optimization and search, leveraging principles from nature to address complex computational challenges. By harnessing the power of evolution, GAs offer a robust framework for finding optimal solutions in diverse problem domains.

	\paragraph{Foundational Principles}
	Genetic Algorithms (GAs) represent a fascinating emulation of nature's evolutionary processes within the realm of computational problem-solving. At the heart of this computational paradigm lies the ingenious concept of mimicking the mechanisms of natural selection and genetic variation to iteratively evolve potential solutions to complex optimization problems. This emulation begins with the creation of a diverse population of candidate solutions, often referred to as individuals or chromosomes, which encode potential solutions to the problem domain. Each individual within this population embodies a unique combination of parameters or characteristics, representing a candidate solution to the given problem.
	
	The evaluation process, facilitated by a fitness function tailored to the specific problem at hand, serves as the cornerstone of GA operations. This fitness function acts as the guiding compass, assessing the quality or effectiveness of each individual within the population relative to the problem's objectives. Through rigorous evaluation against predetermined criteria, individuals are assigned fitness scores that quantitatively capture their performance or suitability as potential solutions. Consequently, this process inherently drives the optimization process, directing the evolutionary trajectory towards increasingly promising regions of the solution space.
	
	Following the evaluation phase, genetic operators such as crossover and mutation come into play, mirroring the mechanisms of genetic recombination and mutation observed in natural evolution. These operators inject diversity and exploration into the population, allowing for the exploration of novel solution combinations and the preservation of beneficial traits across generations. Through crossover, genetic material from two parent individuals is combined to produce offspring with characteristics inherited from both parents, fostering the propagation of favorable traits. Conversely, mutation introduces stochastic perturbations to individual chromosomes, thereby introducing variability and preventing premature convergence towards suboptimal solutions.
	
	In essence, Genetic Algorithms stand as a testament to the power of harnessing nature-inspired principles to tackle complex optimization challenges. By capitalizing on the principles of natural selection, genetic variation, and survival of the fittest, GAs offer a robust and versatile framework for exploring solution spaces, uncovering innovative solutions, and navigating the intricate landscapes of optimization landscapes with efficiency and efficacy.

	\paragraph{Operational Mechanics}
	The operational mechanics of a Genetic Algorithm encompass a series of iterative steps designed to efficiently explore and exploit the solution space. At the outset, the process initiates with the creation of a diverse population comprising potential solutions represented as chromosomes. These chromosomes encode candidate solutions to the optimization problem at hand, with each gene in the chromosome representing a specific attribute or decision variable.
	
	Following the population initialization phase, the algorithm progresses into the selection stage. During selection, individuals within the population are chosen probabilistically based on their fitness scores, which reflect their performance in solving the given problem. This process mimics the natural selection mechanism, favoring individuals with higher fitness values for reproduction while allowing diversity to be maintained through the selection of less fit individuals as well.
	
	Subsequently, the reproduction phase involves the application of genetic operators such as crossover and mutation. Crossover facilitates the exchange of genetic material between selected parent chromosomes, producing offspring with a combination of traits from both parents. Meanwhile, mutation introduces random changes to the genetic makeup of individual chromosomes, promoting exploration of novel regions within the solution space.
	
	The offspring generated through these genetic operations form the basis for the next generation of the population. This iterative cycle of evaluation, selection, and variation continues over multiple generations, with each successive iteration refining the population towards solutions of higher fitness. Through this process, the algorithm progressively converges towards optimal or near-optimal solutions, leveraging the principles of evolution and natural selection to solve complex optimization problems efficiently.

	\paragraph{Genetic Operators}
	Key to the operation of genetic algorithms (GAs) are the genetic operators: selection, crossover, and mutation. Selection operators play a crucial role in determining which individuals from the current population will be chosen as parents for the next generation. These operators employ various strategies, such as roulette wheel selection, tournament selection, or rank-based selection, each with its advantages and trade-offs. For instance, while roulette wheel selection favors individuals with higher fitness values, tournament selection offers a balance between exploration and exploitation by randomly selecting a subset of individuals and choosing the best among them.
	
	Crossover operators, on the other hand, facilitate the exchange of genetic information between selected parents to produce offspring. The process typically involves selecting a crossover point along the chromosome and swapping genetic material between the parents. This recombination of genetic material introduces diversity into the population, enabling the exploration of new solution spaces. Various crossover techniques exist, including one-point crossover, two-point crossover, and uniform crossover, each influencing the exploration-exploitation trade-off differently.
	
	Mutation operators serve to introduce randomness into the genetic algorithm by making small, random modifications to individual solutions. These mutations are essential for maintaining genetic diversity within the population and preventing premature convergence to local optima. Mutation rates are typically kept low to ensure that beneficial genetic material is not lost too quickly, yet high enough to allow for sufficient exploration of the search space. Mutation operators can vary in their intensity and scope, ranging from simple bit flips to more complex structural changes, depending on the problem domain and algorithm requirements.
	
	In essence, genetic operators work together synergistically to drive the evolutionary process within genetic algorithms, balancing exploration and exploitation to search for optimal or near-optimal solutions in complex problem spaces.

	\paragraph{Applications and Limitations}
	Genetic Algorithms (GAs) have garnered widespread recognition for their effectiveness in tackling diverse problem domains, ranging from optimization in engineering design to the realms of machine learning and artificial intelligence. Their versatility and adaptability render them particularly adept at addressing complex problems that conventional optimization techniques often struggle to solve. In engineering, GAs find application in optimizing parameters for intricate systems, such as designing efficient mechanical structures or refining electrical circuit layouts. Moreover, in machine learning and AI, GAs serve as powerful tools for evolving solutions to challenging tasks, such as optimizing neural network architectures or fine-tuning model parameters.
	
	However, despite their efficacy, GAs are not without limitations. One significant concern is their potentially slow convergence rates, especially when applied to high-dimensional or computationally intensive problems. This issue arises due to the stochastic nature of genetic algorithms, which may require a large number of iterations to converge to an optimal solution. Additionally, the effectiveness of GAs heavily relies on the appropriate selection of parameters governing the genetic operators, including selection, crossover, and mutation. Determining these parameters can be a non-trivial task, as suboptimal choices may lead to premature convergence or insufficient exploration of the solution space. Consequently, practitioners often face the challenge of striking a delicate balance between exploration and exploitation during the optimization process, aiming to prevent premature convergence while efficiently traversing the search space.
	
	Despite these limitations, the widespread adoption of GAs across various domains underscores their significance as powerful optimization tools. With ongoing research focused on enhancing their performance and addressing their limitations, genetic algorithms continue to hold promise for tackling complex optimization problems in diverse fields.

	\paragraph{Conclusion}
	Genetic Algorithms (GAs) represent a cornerstone in the realm of computational intelligence, offering a multifaceted approach to tackling intricate problems across various domains. By emulating the fundamental principles of evolution, GAs navigate through solution spaces with a delicate balance between exploration and exploitation. This nuanced strategy enables them to traverse diverse landscapes of potential solutions, leveraging both randomization and structured search methodologies to uncover optimal or near-optimal solutions. Through successive generations, GAs iteratively refine candidate solutions, gradually converging towards increasingly efficient outcomes.
	
	The allure of GAs lies in their inherent adaptability and resilience. Their ability to adapt to evolving problem spaces and dynamic environments renders them invaluable in scenarios where traditional optimization techniques falter. Moreover, the robustness of GAs ensures their efficacy across a spectrum of optimization challenges, ranging from combinatorial problems in logistics and scheduling to parameter optimization in machine learning algorithms.
	
	However, despite their prowess, GAs are not without limitations. Their reliance on randomness and population-based exploration strategies may lead to suboptimal solutions or premature convergence in certain scenarios. Additionally, the computational overhead associated with evaluating fitness functions and maintaining diverse populations can pose challenges in resource-constrained environments.
	
	Nevertheless, the enduring appeal of GAs persists, driven by their versatility and effectiveness in addressing complex optimization tasks. As computational capabilities continue to advance, coupled with ongoing research efforts aimed at enhancing GA methodologies, the future holds promise for further advancements in this field. Through continued innovation and refinement, GAs are poised to remain a cornerstone of computational intelligence, unlocking new frontiers in optimization and problem-solving.

	\subsubsection{Key Concepts and Operating Mechanisms}
	Genetic Algorithms (GAs) are characterized by a set of foundational concepts and mechanisms that underpin their ability to efficiently navigate complex search spaces to find optimal or near-optimal solutions. This section provides an in-depth exploration of these core concepts and the operational dynamics that drive the success of GAs in solving a wide array of computational problems.
	
	\paragraph{Population}
	The population in a Genetic Algorithm constitutes a fundamental component driving the evolutionary process towards finding optimal solutions. It comprises a diverse set of individuals, each embodying a potential solution to the problem at hand. This diversity is paramount as it enables the algorithm to explore a broad spectrum of potential solutions, thereby increasing the chances of discovering an optimal or near-optimal solution. 
	
	In a typical GA setup, individuals within the population are often represented as strings of binary digits, although alternative encodings such as real-valued vectors or permutations are also employed based on the problem's requirements. This representation scheme encapsulates the characteristics or features relevant to the problem domain, facilitating the exploration and manipulation of potential solutions through genetic operations.
	
	Moreover, the population size plays a critical role in the algorithm's performance. A larger population size generally promotes greater exploration of the solution space but may entail increased computational overhead. Conversely, a smaller population size may lead to premature convergence or insufficient diversity, potentially hindering the algorithm's ability to discover optimal solutions.
	
	The selection of an appropriate population size involves a trade-off between exploration and exploitation, where a balance must be struck to ensure effective convergence towards satisfactory solutions within reasonable computational resources. Additionally, mechanisms such as elitism, which preserves the best-performing individuals across generations, contribute to maintaining diversity and preventing premature convergence.
	
	In summary, the population within a GA serves as the breeding ground for potential solutions, embodying the diversity and complexity necessary for effective exploration of the solution space. Through careful management of population size and diversity, coupled with genetic operators and selection mechanisms, the algorithm navigates towards optimal or near-optimal solutions in a manner inspired by natural evolution.

	\paragraph{Fitness Function}
	The fitness function serves as a pivotal element within Genetic Algorithms (GAs), constituting the cornerstone for evaluating the efficacy of potential solutions generated by the algorithm. Essentially, it acts as a metric, quantifying the performance or suitability of each candidate solution within the population relative to the problem domain. By assigning a numerical score or fitness value to each individual, the fitness function provides a basis for discerning the quality of solutions, thereby facilitating the selection of individuals for reproduction and the propagation of genetic material.
	
	In practical terms, the design and formulation of the fitness function profoundly influence the GA's evolutionary process and its ability to converge towards optimal or near-optimal solutions. A well-crafted fitness function encapsulates the objectives and constraints of the optimization problem, effectively guiding the search process towards solutions that align with predefined criteria. Moreover, the fitness function acts as a guiding beacon for the genetic operators employed by the GA, influencing the mechanisms of selection, crossover, and mutation to favor individuals with higher fitness scores.
	
	Mathematically, the fitness function can take various forms depending on the nature of the problem being addressed. In some cases, it may involve simple arithmetic calculations, while in others, it could entail complex computations or even involve machine learning models to assess solution quality. Regardless of its complexity, the fitness function encapsulates the essence of the optimization problem, distilling the objective into a quantifiable measure that drives the evolutionary process of the GA.
	
	In summary, the fitness function serves as a pivotal bridge between the problem space and the optimization process in Genetic Algorithms. Its design and implementation are critical considerations, as they directly influence the algorithm's ability to explore the solution space effectively and converge towards desirable outcomes.

	\paragraph{Selection}
	Selection is a fundamental component of evolutionary algorithms, serving as the mechanism for determining which individuals from the current population will contribute to the next generation. Inspired by the principle of natural selection, where organisms with advantageous traits are more likely to survive and reproduce, selection methods in evolutionary algorithms aim to favor individuals with higher fitness or performance.
	
	One commonly used selection method is \textit{roulette wheel selection}, also known as \textit{fitness proportionate selection}. In this method, each individual's probability of selection is proportional to its fitness relative to the total fitness of the population. Mathematically, the probability \( P(i) \) of selecting an individual \( i \) with fitness \( f(i) \) can be expressed as:
	
	\[ P(i) = \frac{f(i)}{\sum_{j} f(j)} \]
	
	This approach ensures that individuals with higher fitness have a greater chance of being selected, mimicking the natural process of favoring individuals with higher reproductive success.
	
	Another common selection method is \textit{tournament selection}, where a predefined number of individuals are randomly selected from the population, and the one with the highest fitness among them is chosen for reproduction. Tournament selection provides a balance between exploration and exploitation by allowing less fit individuals to occasionally be selected, preventing premature convergence to local optima.
	
	Both selection methods play crucial roles in maintaining diversity within the population and driving the evolutionary process towards optimal solutions. While roulette wheel selection tends to favor individuals with higher fitness more consistently, tournament selection introduces stochasticity by considering a subset of individuals in each selection event, which can be advantageous in certain problem domains.

	\paragraph{Crossover}
	Crossover, also known as recombination, constitutes a fundamental genetic operator within evolutionary algorithms, playing a pivotal role in the generation of new offspring. Essentially, this process involves the fusion of genetic material from two parent individuals to create progeny with a combination of traits inherited from each parent. This mechanism is analogous to biological reproduction, where genetic information is exchanged between organisms during mating to produce offspring with unique genetic compositions. The overarching objective of crossover in evolutionary algorithms is to explore the search space effectively by generating diverse solutions that potentially possess favorable attributes.
	
	In the context of genetic algorithms, crossover serves as a mechanism for promoting exploration and exploitation simultaneously. By blending genetic information from different parents, crossover facilitates the creation of offspring that inherit beneficial characteristics from both parents while introducing variations that could potentially lead to improved solutions. This diversification strategy helps prevent premature convergence and enhances the algorithm's ability to explore promising regions of the search space.
	
	The effectiveness of crossover hinges on several factors, including the choice of crossover points and the method employed to combine parental genetic material. The selection of crossover points determines the segments of genetic material exchanged between parents, influencing the extent of exploration and the diversity of offspring generated. Various crossover techniques, such as single-point crossover, multi-point crossover, and uniform crossover, offer different approaches to combining parental genes, each with its own implications for solution quality and convergence speed.
	
	Moreover, the balance between exploration and exploitation is crucial in determining the algorithm's convergence behavior and solution quality. While crossover promotes exploration by creating diverse offspring, it must be complemented by other genetic operators, such as mutation, to ensure adequate exploration of the search space. Additionally, the interaction between crossover and selection mechanisms, such as tournament selection or roulette wheel selection, plays a critical role in shaping the evolutionary dynamics and convergence properties of genetic algorithms.
	
	In summary, crossover stands as a cornerstone of evolutionary algorithms, driving the exploration of the solution space and facilitating the discovery of high-quality solutions through the combination of genetic information from parent individuals.

	\paragraph{Mutation}
	Mutation serves as a pivotal mechanism in evolutionary algorithms, injecting randomness into the genetic makeup of individuals within a population. This process enables the exploration of novel solutions by introducing variations that deviate from the existing gene pool. Essentially, mutation acts as a driving force for diversification, counteracting the tendency of the population to converge prematurely towards suboptimal solutions. By introducing randomness, mutation ensures that the algorithm does not become trapped in local optima but instead continues to explore the solution space in search of potentially superior solutions. 
	
	Moreover, the mutation rate plays a crucial role in determining the balance between exploration and exploitation within the algorithm. A higher mutation rate increases the likelihood of introducing significant changes into the genetic makeup of individuals, fostering exploration at the expense of potentially disrupting beneficial traits. Conversely, a lower mutation rate may lead to a more conservative approach, where incremental changes dominate, potentially slowing down the exploration process.
	
	However, striking the right balance in setting the mutation rate is not a trivial task and often requires empirical tuning based on the problem domain and algorithmic characteristics. A mutation rate that is too low may result in premature convergence, limiting the algorithm's ability to explore diverse regions of the search space. Conversely, an excessively high mutation rate may lead to excessive exploration, hindering the algorithm's ability to exploit promising solutions efficiently.
	
	Therefore, careful consideration and experimentation are necessary to determine an optimal mutation rate that facilitates effective exploration while ensuring convergence towards high-quality solutions. Additionally, ongoing monitoring and adaptation of the mutation rate throughout the evolutionary process may be necessary to respond to changing dynamics and evolving problem landscapes.

	\paragraph{Convergence}
	Convergence in genetic algorithms (GAs) is a pivotal aspect that underscores the algorithm's efficacy in finding optimal or near-optimal solutions to complex optimization problems. At its essence, convergence represents the gradual refinement and improvement of the population of candidate solutions over successive generations. This iterative refinement process is driven by the genetic operators of selection, crossover, and mutation, which collectively steer the population towards better solutions. 
	
	Throughout the evolutionary process, the population evolves through generations, with each subsequent generation ideally exhibiting improved fitness values compared to its predecessors. This improvement in fitness signifies progress towards the optimization goal and is indicative of convergence towards an optimal solution. However, it's essential to note that convergence is not always guaranteed in every run of a GA, as the algorithm's effectiveness can be influenced by various factors, including the problem's complexity, the choice of genetic operators, and the population size.
	
	Monitoring convergence is a critical task in GA optimization. Various convergence metrics and criteria can be employed to assess the algorithm's progress and determine when to halt the evolutionary process. These metrics may include the average fitness of the population, the best fitness achieved, or the diversity of solutions within the population. By tracking these metrics over successive generations, practitioners can gain insights into the algorithm's behavior and performance, enabling informed decisions regarding termination criteria.
	
	In practice, achieving convergence in a GA involves striking a balance between exploration and exploitation. While exploration involves discovering diverse regions of the solution space to avoid premature convergence to suboptimal solutions, exploitation focuses on intensifying the search around promising regions to refine the solutions further. Finding the optimal balance between exploration and exploitation is crucial for facilitating convergence towards high-quality solutions within a reasonable computational budget.

	\paragraph{Conclusion}
	The success of Genetic Algorithms (GAs) in solving complex optimization problems is undeniable, primarily owing to their ability to emulate sophisticated evolutionary processes. GAs effectively mimic the principles of natural selection, where diverse solutions within a population compete and evolve over successive generations. This emulation starts with the initialization of a diverse population representing potential solutions to the optimization problem at hand. Each solution's fitness is then evaluated based on a predefined objective function, determining its suitability for survival and reproduction in the algorithmic environment.
	
	Throughout the iterative process, GAs employ a strategic combination of selection, crossover, and mutation operators to drive population evolution. Selection mechanisms favor individuals with higher fitness, mimicking the natural process of survival of the fittest. Crossover facilitates the exchange of genetic information between selected individuals, promoting the exploration of novel solution spaces. Mutation introduces random changes to the genetic makeup of offspring, injecting diversity into the population and preventing premature convergence to suboptimal solutions.
	
	Crucially, GAs strike a delicate balance between exploration and exploitation. While exploration allows the algorithm to search the solution space broadly for promising areas, exploitation focuses on refining and intensifying search efforts around high-quality solutions. This balance is achieved through the adaptive adjustment of evolutionary parameters, such as mutation rates and selection pressures, throughout the optimization process.
	
	In essence, the iterative refinement of the population through selection, crossover, and mutation enables GAs to navigate complex solution spaces effectively. By leveraging the principles of evolution, GAs continuously drive towards increasingly better solutions, making them indispensable tools for tackling a wide range of optimization challenges across various domains.

	\subsubsection{Applications and Limitations}
	Genetic Algorithms (GAs) have found widespread applications across various domains, solving complex optimization and search problems with their robust, evolutionary computation techniques. Despite their versatility, GAs also encounter specific limitations that can affect their efficacy under certain conditions. This section explores both the broad applicability of GAs in solving real-world problems and the inherent limitations that researchers and practitioners must navigate.
	
	\paragraph{Wide-Ranging Applications}
	Genetic Algorithms (GAs) have established themselves as versatile problem-solving tools with applications spanning various domains. In the realm of engineering, they have emerged as indispensable aids in design optimization tasks. Consider scenarios where engineers seek to fine-tune the parameters of a complex system while adhering to multiple constraints, such as material limitations, cost considerations, and performance requirements. GAs excel in navigating the vast design space to identify optimal configurations, balancing competing factors to achieve the desired outcome efficiently and effectively.
	
	In the financial sector, where uncertainty and volatility are constants, GAs offer valuable assistance in portfolio optimization. Investing in diverse assets involves weighing the trade-offs between risk and return, a complex optimization challenge. Here, GAs come into play, leveraging their ability to explore diverse investment strategies and adapt to changing market conditions. By iteratively evaluating and adjusting investment portfolios, GAs empower financial analysts to make informed decisions that maximize returns while managing risk exposure.
	
	Meanwhile, in the realm of machine learning and data mining, GAs contribute significantly to feature selection and model optimization tasks. In the quest to extract meaningful insights from vast datasets, identifying the most relevant features and fine-tuning model parameters is paramount. GAs provide a systematic approach to this endeavor, efficiently exploring the high-dimensional space of possible configurations to uncover optimal solutions. By leveraging evolutionary principles, GAs enable data scientists to enhance the performance and generalization capabilities of machine learning models, ultimately leading to more accurate predictions and actionable insights.
	
	Moreover, the utility of GAs extends to addressing complex scheduling and routing problems encountered in logistics and telecommunications. From optimizing transportation routes to scheduling manufacturing processes, GAs offer robust solutions to intricate optimization challenges. Their ability to handle combinatorial optimization tasks and adapt to dynamic environments makes them invaluable tools in real-world scenarios where efficiency and resource utilization are paramount.
	
	In addition, GAs have demonstrated prowess in tackling challenging puzzles and games, where traditional problem-solving approaches often fall short. Whether it's solving Sudoku puzzles, playing chess at a grandmaster level, or optimizing strategies in strategic board games, GAs showcase their versatility and adaptability in diverse gaming environments.
	
	Overall, the wide-ranging applications of GAs underscore their versatility and effectiveness across various domains, making them indispensable tools for solving complex optimization and decision-making problems.

	\paragraph{Inherent Limitations}
	Despite their adaptability, Genetic Algorithms (GAs) come with inherent limitations that can significantly affect their performance and applicability in various problem domains. One of the primary challenges faced by GAs is their computational cost, particularly evident when dealing with large-scale optimization problems. Evaluating the fitness of each individual in a sizable population across numerous generations can impose substantial computational overhead, leading to longer execution times and increased resource utilization. This computational complexity often necessitates the allocation of considerable computing resources, which may not always be feasible or practical, especially in resource-constrained environments.
	
	Moreover, GAs may encounter the issue of premature convergence, wherein the evolutionary process halts prematurely, resulting in the convergence of the population to suboptimal solutions. This phenomenon typically arises due to the lack of genetic diversity within the population or suboptimal design choices in the genetic operators employed. Premature convergence restricts the exploration of the solution space, preventing the algorithm from discovering potentially superior solutions. Overcoming premature convergence requires careful consideration of population initialization strategies, selection mechanisms, and operator design to maintain population diversity throughout the evolutionary process.
	
	Additionally, selecting appropriate parameter values for GAs poses a significant challenge. Parameters such as population size, mutation rate, and crossover rate exert a substantial influence on the algorithm's behavior and performance. However, determining optimal parameter values often involves a considerable amount of trial and error or domain-specific expertise. Inadequate parameter selection can hinder the algorithm's convergence, leading to suboptimal solutions or prolonged convergence times. Thus, practitioners must invest time and effort in fine-tuning these parameters to achieve satisfactory performance across different problem instances.

	\paragraph{Overcoming Limitations}
	Efforts to overcome these limitations have led to the development of advanced GA variants and hybrid algorithms that combine GAs with other optimization techniques, aiming to enhance performance and reduce computational demands. These efforts have been driven by the recognition of the inherent challenges faced by traditional genetic algorithms, such as premature convergence and difficulty in handling high-dimensional search spaces. By integrating GAs with techniques like adaptive parameter tuning, which dynamically adjusts GA parameters based on performance feedback, researchers aim to mitigate these challenges and improve overall efficiency. Additionally, the incorporation of local search methods into GA frameworks has shown promise in refining solutions and overcoming local optima. These methods work by iteratively exploring the solution space around candidate solutions, seeking improvements and potentially identifying more optimal solutions. 
	
	Furthermore, leveraging parallel computing resources can alleviate the computational burden associated with GA-based optimization, enabling GAs to tackle larger and more complex problems more effectively. Parallelization techniques distribute the computational workload across multiple processing units, allowing for concurrent execution of GA operations and speeding up the optimization process. This approach is particularly beneficial for problems with computationally intensive fitness evaluations or large population sizes. By harnessing the power of parallel computing, researchers can unlock new possibilities for applying genetic algorithms to real-world optimization challenges, pushing the boundaries of what is achievable with these versatile optimization techniques.

	In conclusion, Genetic Algorithms stand out for their flexibility and broad applicability across various domains, offering powerful solutions to optimization and search problems that are difficult to solve with traditional methods. While they face limitations related to computational efficiency, convergence, and parameter selection, ongoing research and methodological advancements continue to expand their capabilities and address these challenges, underscoring the enduring value of GAs in computational intelligence and optimization.

	\subsubsection{Pseudocode for Genetic Algorithms}
	The Genetic Algorithm is a versatile optimization technique tailored for solving complex problems by mimicking the process of natural selection and evolution. It operates by maintaining a population of candidate solutions and iteratively evolving them through processes such as selection, crossover, and mutation to generate fitter offspring. This approach allows GAs to efficiently explore the solution space and converge towards optimal or near-optimal solutions. The fundamental principles of GAs are elucidated in the pseudocode depicted in Figure \ref{fig:ga-pseudocode}, demonstrating its systematic methodology for exploring and refining candidate solutions.
	
	\begin{algorithm}
		\caption{Genetic Algorithm}
		\begin{algorithmic}[1]
			\Procedure{GeneticAlgorithm}{}
			\State Initialize population with random individuals
			\State Evaluate the fitness of each individual
			\While{termination condition not met}
			\State Select parents from the population
			\State Perform crossover on parents to create offspring
			\State Apply mutation to offspring
			\State Evaluate the fitness of the offspring
			\State Select individuals for the next generation
			\EndWhile
			\State \Return The best solution found
			\EndProcedure
		\end{algorithmic}\label{fig:ga-pseudocode}
	\end{algorithm}

	\subsection{Previous Work on ML and AI Interplay with Genetic Algorithms}
	\paragraph{Evolving Search Spaces with Variational Autoencoders}
	
	One line of research explores using machine learning to improve the search space itself. \cite{bentley2022evolving} proposes an approach that leverages variational autoencoders (VAEs) to learn improved representations of the search space. The VAE is trained on a set of high-performing solutions, allowing it to capture the essential features that lead to success. This learned representation can then be used to guide the search process towards more promising regions of the search space. This work demonstrates the potential of machine learning for dynamically adapting the search space during the optimization process.
	
	\paragraph{Machine Learning for Phylogenetic Tree Search}
	
	Another direction involves employing machine learning to guide the search for optimal solutions. \cite{azouri2021harnessing} investigates the use of machine learning to guide the search for phylogenetic trees, which are essential for understanding evolutionary relationships. The proposed approach utilizes a reinforcement learning framework, where the agent learns to select search operators based on feedback about the quality of the resulting solutions. This approach demonstrates the ability of machine learning to learn effective search strategies, potentially leading to more efficient exploration of the search space.
	
	\paragraph{Fitness Approximation through Machine Learning}
	
	Recent work explores the use of machine learning to approximate the fitness function, a crucial component of many optimization algorithms. \cite{tzruia2023fitness} proposes a framework that leverages machine learning to learn an accurate surrogate model of the true fitness function. This surrogate model can then be used to evaluate candidate solutions more efficiently, reducing the computational cost of the optimization process. This work highlights the potential of machine learning to alleviate the computational burden associated with evaluating complex fitness functions.

\subsection{Algogenic Enhancements for Genetic Algorithms}
\subsubsection{Problem Formulation Enhancement}
\paragraph{Refining Problem Representation with Generative AI}
Integrating generative AI into the initial stages of Genetic Algorithms represents a thoughtful evolution from traditional problem representation methods. By employing Large Language Models, such as GPT, we can reimagine how problems are structured, moving away from basic numerical or binary encodings towards a representation that captures the complexity and nuance of the domain. This shift allows for a more comprehensive understanding of potential solutions and their interrelations, potentially leading to a more informed exploration of the solution space. It is suggested that this approach could enable Genetic Algorithms to encapsulate complex data types and relationships akin to those found in natural language or conceptual frameworks, offering a pathway to more effective problem analysis and solution identification. However, the practical implementation of this enhancement requires careful consideration of how these richer representations are translated into the genetic algorithm's operations, ensuring that the benefits of enhanced problem understanding directly contribute to improved algorithm performance.

\paragraph{Evolving Fitness Function Dynamics}
The dynamic refinement of fitness functions, guided by insights from generative AI, presents a method to keep the evaluation criteria of Genetic Algorithms in tune with the evolving landscape of the solution space. By adjusting fitness functions based on the analysis of current population performance and potential exploration areas, we propose that Genetic Algorithms could navigate towards underexplored yet promising regions. This approach relies on the nuanced analysis capabilities of LLMs to identify and correct biases in the fitness function, potentially leading to a more balanced and effective exploration of the solution space. However, skepticism arises in the practicality of continuously adapting fitness functions without compromising the stability and convergence properties of the algorithm. The challenge lies in implementing these adjustments in a way that benefits the genetic algorithm's search process, ensuring that changes to the fitness function lead to tangible improvements in solution quality and algorithm efficiency.

\paragraph{Enhancing Efficiency and Effectiveness through Problem Formulation}
The integration of generative AI into problem formulation processes for Genetic Algorithms suggests a route to greater adaptability and contextual awareness. This enhancement is theorized to allow Genetic Algorithms to dynamically adjust their exploration strategies in response to complex problem landscapes, potentially leading to more efficient identification of promising solution areas. By embedding deeper contextual insights and domain knowledge into the algorithm's operations, we might see an improvement in decision-making and solution relevance. Nevertheless, the practical translation of these enhancements into real-world applications requires a careful balance between algorithm complexity and computational efficiency. The value of this approach will be measured by its ability to not just theoretically, but practically improve the Genetic Algorithm's ability to navigate and adapt to evolving problem spaces, demonstrating tangible benefits in diverse application scenarios.

\subsubsection{Dynamic Fitness Function Adjustment}
\paragraph{Real-Time Fitness Function Adaptation}
Introducing the concept of real-time fitness function adjustment into Genetic Algorithms proposes an evolutionary step towards more responsive and adaptable optimization processes. By leveraging generative AI to monitor and modify the fitness function in response to ongoing changes in the solution landscape, it's suggested that Genetic Algorithms could better align with the current optimization context. This real-time adaptation aims to enhance the algorithm's ability to identify and exploit promising areas of the solution space, potentially leading to more effective and efficient problem-solving. However, the implementation of such dynamic adjustments raises questions about the balance between adaptability and algorithm stability. The challenge lies in ensuring that these adaptations contribute positively to the optimization process without introducing volatility that could hinder the Genetic Algorithm's performance.

\paragraph{Incorporating Adaptive Fitness Evaluation}
The adaptation of fitness evaluation strategies, informed by continuous analysis and feedback, represents a nuanced approach to enhancing the performance of Genetic Algorithms. By dynamically adjusting evaluation criteria based on the insights derived from generative AI, we propose a method for Genetic Algorithms to maintain relevance and effectiveness in evolving problem landscapes. This process involves a careful interpretation of population dynamics and algorithmic feedback, aiming to refine the fitness evaluation to better match the optimization objectives and constraints. However, skepticism may arise regarding the feasibility and impact of such continuous adaptations, particularly in terms of computational overhead and the potential for overfitting to transient solution characteristics. The practical value of adaptive fitness evaluation will ultimately depend on its ability to improve the Genetic Algorithm's convergence behavior and solution quality without compromising the efficiency or robustness of the optimization process.

\paragraph{Optimizing Algorithm Outcomes through Fitness Function Dynamics}
The proposition of dynamically refining fitness functions based on generative AI insights seeks to enhance the adaptability and outcome of Genetic Algorithms. This approach suggests that by continuously updating the evaluation criteria to reflect the current state of the solution space and optimization goals, Genetic Algorithms could achieve more nuanced and effective exploration. However, the practical application of this concept requires a careful consideration of how these dynamic adjustments are implemented, ensuring that they contribute to the algorithm's ability to find optimal solutions efficiently. The challenge lies in managing the complexity introduced by frequent fitness function modifications while maintaining the algorithm's focus on relevant solution areas. The effectiveness of dynamic fitness function adjustments will be measured by their ability to enhance the Genetic Algorithm's performance in real-world optimization tasks, demonstrating tangible benefits in terms of solution quality and algorithm efficiency.

\subsubsection{Predictive Crossover and Mutation Strategies}
\paragraph{Enhancing Genetic Operations through Predictive Insights}
Integrating predictive insights into crossover and mutation strategies offers a forward-looking enhancement to Genetic Algorithms. By utilizing generative AI to analyze population dynamics and performance trends, we suggest a methodology for adapting genetic operations to the evolving needs of the optimization process. This approach aims to refine the algorithm's ability to generate potentially successful offspring and introduce beneficial genetic variations. However, the practical implementation of predictive strategies in genetic operations raises considerations about the balance between exploration and exploitation, as well as the computational resources required for continuous analysis. The challenge lies in ensuring that these predictive adjustments effectively contribute to the Genetic Algorithm's ability to navigate complex solution spaces, improving both the efficiency and quality of the optimization process.

\paragraph{Tailoring Genetic Strategies to Enhance Performance}
The proposition of tailoring genetic strategies based on predictive analytics and real-time feedback introduces a nuanced approach to optimizing Genetic Algorithms. By leveraging insights from generative AI, it's suggested that algorithms could dynamically adjust their crossover and mutation strategies to better suit the current optimization landscape. This adaptive mechanism aims to maintain a balance between exploring new solution regions and exploiting known promising areas, potentially leading to more effective problem-solving. However, the practicality of implementing such tailored strategies raises questions about the algorithm's ability to respond to changes without introducing undue complexity or computational overhead. The value of this approach will be evaluated based on its impact on the Genetic Algorithm's performance, particularly in terms of convergence speed and solution quality.

\paragraph{Improving Algorithm Adaptability with Predictive Strategies}
Introducing predictive strategies into the framework of Genetic Algorithms aims to enhance their adaptability and effectiveness in solving complex optimization problems. By analyzing trends and performance data, we propose a method for dynamically adjusting genetic operations to better navigate the solution space. This approach suggests a potential for Genetic Algorithms to more effectively balance exploration and exploitation, adapting to changes in the optimization context. However, skepticism may arise regarding the feasibility of continuously integrating predictive insights into genetic operations, especially considering the computational demands and the potential for algorithmic instability. The practical implementation of predictive strategies will need to carefully manage these concerns, ensuring that the enhancements contribute positively to the Genetic Algorithm's ability to find optimal solutions in diverse and dynamic problem environments.

\subsubsection{Adaptive Selection Pressure}
\paragraph{Refining Selection Mechanisms for Enhanced Evolution}
The concept of adaptive selection pressure introduces a strategic enhancement to Genetic Algorithms, aimed at optimizing the selection process in response to changing problem dynamics. By leveraging generative AI, it's suggested that algorithms could dynamically adjust the intensity of selection pressure to maintain an effective balance between diversity and convergence. This approach seeks to enhance the algorithm's ability to explore the solution space comprehensively while efficiently converging on high-quality solutions. However, the practical application of adaptive selection mechanisms must consider the potential for disruption to the evolutionary balance, ensuring that adjustments to selection pressure contribute positively to the Genetic Algorithm's overall performance. The challenge lies in implementing these adaptive mechanisms in a way that improves the efficiency and effectiveness of the optimization process without compromising the stability or robustness of the algorithm.

\paragraph{Implementing Dynamic Selection to Optimize Performance}
The proposal to implement dynamic selection mechanisms within Genetic Algorithms offers a method for enhancing algorithm performance through adaptive selection pressure. By analyzing population dynamics and evolutionary trends with the aid of generative AI, we suggest a framework for adjusting selection criteria to better align with the optimization goals. This adaptive approach aims to improve the algorithm's ability to navigate the solution space, potentially leading to more effective problem-solving. However, skepticism may arise regarding the complexity of continuously adapting selection mechanisms and their impact on the Genetic Algorithm's convergence behavior. The practical value of implementing dynamic selection will be assessed based on its ability to enhance the Genetic Algorithm's adaptability and effectiveness in solving complex optimization problems, without introducing undue computational overhead or compromising algorithm stability.

\paragraph{Balancing Exploration and Exploitation through Adaptive Selection}
Introducing adaptive selection mechanisms into Genetic Algorithms proposes a nuanced approach to optimizing the balance between exploration and exploitation. By leveraging insights from generative AI, it's suggested that algorithms could dynamically adjust selection pressure based on real-time feedback and analysis. This process aims to ensure that the Genetic Algorithm remains responsive to evolving problem landscapes, potentially leading to more efficient and effective optimization. However, the practical implementation of adaptive selection raises questions about the algorithm's ability to maintain stability and consistency in the face of changing selection dynamics. The challenge lies in implementing these adaptive mechanisms in a way that positively contributes to the Genetic Algorithm's performance, improving its ability to find high-quality solutions while navigating complex solution spaces.

\subsubsection{Semantic Encoding of Solutions}
\paragraph{Advancing Solution Representation through Semantic Encoding}
The integration of semantic encoding into Genetic Algorithms represents an innovative step towards more meaningful and contextually aware solution representations. By employing generative AI to encode solutions with semantic information, we suggest a methodology for capturing the complexities and nuances of the solution space in a more intuitive manner. This approach aims to enhance the algorithm's understanding of potential solutions, potentially leading to more effective genetic operations and improved problem-solving. However, the practical application of semantic encoding introduces considerations about the computational resources required for processing complex encodings and the impact on the algorithm's efficiency. The challenge lies in balancing the benefits of enriched solution representations with the need to maintain computational tractability and algorithm performance.

\paragraph{Implementing Semantically Rich Solution Representations}
The proposition of implementing semantically rich solution representations within Genetic Algorithms offers a pathway to enhancing the algorithm's capacity for understanding and manipulating solutions. By leveraging the capabilities of generative AI, it's suggested that solutions could be encoded in a manner that captures their inherent semantic relationships and context. This approach aims to improve the effectiveness of genetic operations by ensuring that solutions are represented in a way that aligns with their conceptual and functional significance. However, skepticism may arise regarding the complexity of translating these semantically rich representations into practical algorithmic operations and the potential impact on the Genetic Algorithm's computational efficiency. The value of implementing semantic encoding will be measured by its ability to facilitate more informed and effective problem-solving within Genetic Algorithms, without compromising the algorithm's performance or scalability.

\paragraph{Enhancing Genetic Algorithm Capabilities with Semantic Information}
Integrating semantic information into the encoding scheme of Genetic Algorithms proposes an enhancement aimed at improving the algorithm's understanding and manipulation of solutions. By adopting semantically rich representations, we suggest a method for capturing the complexity and nuance of the solution space in a more intuitive and meaningful manner. This approach seeks to enhance the effectiveness of genetic operations, potentially leading to more efficient problem-solving. However, the practicality of incorporating semantic information into Genetic Algorithms raises questions about the balance between enriched solution representation and computational demands. The challenge lies in ensuring that semantic encoding contributes positively to the Genetic Algorithm's ability to navigate complex solution spaces, improving both the quality and relevance of the solutions generated.

\subsubsection{Solution Interpretation and Refinement}
\paragraph{Enhancing Solution Evaluation with Generative AI}
The integration of solution interpretation and refinement mechanisms into Genetic Algorithms, guided by generative AI, represents a strategic enhancement aimed at improving the practicality and feasibility of generated solutions. By employing Large Language Models to analyze and refine solutions in the context of real-world constraints and objectives, we suggest a methodology for ensuring that algorithm outputs are not only optimal within the computational framework but also viable in practical scenarios. This approach seeks to deepen the algorithm's understanding of solution implications, potentially leading to more informed and effective decision-making. However, the implementation of such refinement mechanisms introduces considerations about the computational resources required for in-depth analysis and the impact on the algorithm's efficiency. The challenge lies in balancing the benefits of enhanced solution evaluation with the need to maintain computational tractability and timely problem-solving.

\paragraph{Implementing AI-Driven Solution Refinement Processes}
The proposal to implement AI-driven solution refinement processes within Genetic Algorithms offers a method for enhancing the quality and applicability of algorithm outputs. By leveraging the analytical capabilities of generative AI, it's suggested that solutions could be evaluated and refined based on a comprehensive set of criteria that extend beyond the algorithm's initial fitness function. This adaptive approach aims to improve the relevance and feasibility of solutions, potentially leading to more effective problem-solving. However, skepticism may arise regarding the feasibility of continuously refining solutions based on AI-driven insights, especially considering the computational demands and the potential for delaying the optimization process. The practical value of AI-driven solution refinement will be evaluated based on its ability to enhance the Genetic Algorithm's outputs, ensuring that solutions are not only technically sound but also practically implementable.

\paragraph{Elevating Solution Quality and Applicability through Refinement}
Introducing solution interpretation and refinement mechanisms into Genetic Algorithms, informed by generative AI, proposes an enhancement aimed at elevating the quality and practical applicability of solutions. By employing advanced analytics to evaluate and refine solutions in the context of real-world constraints, we suggest a methodology for ensuring that algorithm outputs align closely with practical requirements and objectives. This approach seeks to enhance the algorithm's ability to generate solutions that are not only optimal from a computational perspective but also viable and effective in practical applications. However, the practical implementation of these refinement mechanisms raises questions about the algorithm's efficiency and the computational resources required for in-depth solution analysis. The challenge lies in implementing refinement processes that contribute positively to the Genetic Algorithm's performance, improving the relevance and feasibility of solutions without compromising the efficiency of the optimization process.

\subsubsection{Continuous Evolution Strategy}
\paragraph{Adapting Genetic Algorithms to Dynamic Environments}
Introducing a Continuous Evolution Strategy into Genetic Algorithms represents a forward-thinking enhancement aimed at enabling dynamic adaptation to evolving problem landscapes. By leveraging generative AI to continually adjust algorithm parameters and objectives, we propose a method for maintaining the relevance and effectiveness of Genetic Algorithms in dynamic and uncertain environments. This approach suggests a potential for algorithms to anticipate changes and refine their strategies autonomously, potentially leading to more robust and adaptable problem-solving. However, the practical application of a Continuous Evolution Strategy introduces considerations about the algorithm's ability to remain stable and efficient in the face of continuous adjustments. The challenge lies in balancing the benefits of dynamic adaptation with the need to ensure that the Genetic Algorithm remains focused and effective in achieving optimization goals.

\paragraph{Implementing Feedback-Driven Evolutionary Processes}
The proposition of implementing feedback-driven evolutionary processes within Genetic Algorithms offers a pathway to enhancing their adaptability and responsiveness to changing problem dynamics. By establishing a continuous feedback loop informed by generative AI insights, it's suggested that algorithms could dynamically adjust their strategies based on evolving conditions and performance data. This adaptive mechanism aims to improve the algorithm's ability to navigate complex and uncertain problem landscapes, potentially leading to more effective and efficient problem-solving. However, skepticism may arise regarding the complexity of continuously integrating feedback into the evolutionary process and its impact on the Genetic Algorithm's convergence behavior. The practical value of feedback-driven evolutionary processes will be assessed based on their ability to enhance the Genetic Algorithm's performance in dynamic environments, without introducing undue computational overhead or compromising algorithm stability.

\paragraph{Ensuring Long-Term Algorithm Effectiveness through Continuous Evolution}
Integrating a Continuous Evolution Strategy into Genetic Algorithms proposes an enhancement aimed at ensuring their long-term effectiveness and relevance. By employing generative AI to continuously adapt algorithm parameters and strategies, we suggest a method for maintaining the algorithm's adaptability in the face of evolving problem landscapes. This approach seeks to enable Genetic Algorithms to anticipate changes and refine their optimization strategies autonomously, potentially leading to more robust and effective problem-solving. However, the practical implementation of a Continuous Evolution Strategy raises questions about the balance between adaptability and algorithmic stability. The challenge lies in implementing continuous evolution mechanisms that positively contribute to the Genetic Algorithm's ability to find optimal solutions, improving its performance and relevance in dynamic and complex problem environments.

\subsubsection{LLM-Guided Problem Decomposition}
\paragraph{Enhancing Problem Solving with AI-Guided Decomposition}
The introduction of LLM-Guided Problem Decomposition into Genetic Algorithms represents a strategic enhancement aimed at improving the algorithm's ability to tackle complex problems. By leveraging the analytical capabilities of generative AI to break down problems into more manageable sub-components, we propose a methodology for enabling more focused and effective solution exploration. This approach suggests a potential for Genetic Algorithms to navigate complex problem spaces more efficiently, potentially leading to faster convergence and improved solution quality. However, the practical application of AI-guided decomposition introduces considerations about the algorithm's ability to integrate and synthesize solutions across decomposed sub-problems. The challenge lies in balancing the benefits of focused problem-solving with the need to maintain a coherent and comprehensive approach to optimization.

\paragraph{Implementing Strategic Problem Decomposition}
The proposal to implement strategic problem decomposition within Genetic Algorithms offers a method for enhancing the algorithm's capacity for solving complex optimization problems. By leveraging generative AI to identify logical sub-components within the problem domain, it's suggested that algorithms could adopt a more structured and efficient approach to problem-solving. This adaptive mechanism aims to simplify the optimization process, potentially leading to more effective exploration of the solution space. However, skepticism may arise regarding the complexity of managing decomposition and reintegration processes and their impact on the Genetic Algorithm's overall performance. The practical value of strategic problem decomposition will be evaluated based on its ability to facilitate more focused and efficient problem-solving within Genetic Algorithms, without compromising the coherence or quality of the solutions generated.

\paragraph{Optimizing Solutions through Decomposed Problem Solving}
Introducing LLM-Guided Problem Decomposition into Genetic Algorithms proposes an enhancement aimed at optimizing the problem-solving process. By employing generative AI to break down complex problems into more manageable sub-problems, we suggest a methodology for enabling more targeted and effective solution exploration. This approach seeks to improve the Genetic Algorithm's ability to navigate intricate problem spaces, potentially leading to more efficient problem-solving and improved solution quality. However, the practical implementation of problem decomposition raises questions about the algorithm's ability to synthesize solutions across decomposed sub-components effectively. The challenge lies in implementing decomposition strategies that contribute positively to the Genetic Algorithm's performance, improving its efficiency and effectiveness in solving complex optimization problems.

\subsubsection{Narrative-Based Evolution}
\paragraph{Innovating Genetic Algorithms with Narrative Integration}
The integration of Narrative-Based Evolution into Genetic Algorithms represents an innovative enhancement aimed at simulating solution development within dynamic scenarios. By employing generative AI to create narratives that reflect potential future states or challenges, we propose a methodology for evolving solutions that are robust, adaptable, and effective across a range of possible futures. This approach suggests a potential for Genetic Algorithms to develop foresight and adaptability, anticipating and preparing for future challenges. However, the practical application of narrative-based evolution introduces considerations about the algorithm's ability to process and adapt to narrative-driven scenarios effectively. The challenge lies in balancing the benefits of narrative integration with the need to ensure that the Genetic Algorithm remains focused and effective in achieving optimization goals.

\paragraph{Crafting Dynamic Scenarios for Solution Evolution}
The proposition of crafting dynamic scenarios for solution evolution within Genetic Algorithms offers a pathway to enhancing the algorithm's adaptability and foresight. By leveraging generative AI to generate scenarios that simulate a range of conditions and challenges, it's suggested that algorithms could better prepare solutions for future uncertainties. This adaptive mechanism aims to improve the robustness and effectiveness of solutions, potentially leading to more resilient problem-solving. However, skepticism may arise regarding the complexity of creating and integrating dynamic scenarios into the evolutionary process and their impact on the Genetic Algorithm's efficiency. The practical value of narrative-based evolution will be assessed based on its ability to enhance the Genetic Algorithm's performance, ensuring that solutions are not only optimal but also adaptable to evolving conditions and future challenges.

\paragraph{Enhancing Solution Robustness through Narrative Contextualization}
Integrating Narrative-Based Evolution into Genetic Algorithms proposes an enhancement aimed at elevating solution robustness and adaptability. By employing generative AI to contextualize the evolutionary process within dynamic scenarios, we suggest a method for preparing solutions to navigate future uncertainties effectively. This approach seeks to imbue solutions with foresight and resilience, potentially leading to more effective problem-solving in dynamic and uncertain environments. However, the practical implementation of narrative contextualization raises questions about the algorithm's ability to interpret and adapt to narrative-driven dynamics effectively. The challenge lies in implementing narrative-based evolution mechanisms that positively contribute to the Genetic Algorithm's ability to generate robust and adaptable solutions, improving its performance and relevance in addressing complex optimization challenges.

	\subsubsection{Pseudocode for Algogenic Genetic Algorithms}
	Genetic Algogens integrate AI techniques to augment traditional genetic algorithms by dynamically adjusting genetic operators and strategies based on the observed behavior of the population and real-time fitness evaluations. This pseudocode, available in \ref{fig:genetic-algorithm-Algogen-pseudocode}, outlines an advanced framework incorporating AI-driven enhancements for adaptive selection of genetic operators, population partitioning, fitness estimation, and real-time parameter optimization.
	
	\begin{algorithm}
		\caption{Algogenic Genetic Algorithm Pseudocode}
		\begin{algorithmic}[1]
			\Procedure{AlgogenicGeneticAlgorithm}{PopulationSize, Generations}
			\State InitializePopulation(PopulationSize) \Comment{Create initial population}
			\State EvaluateFitness() \Comment{Assess initial population}
			\For{$generation = 1$ to $Generations$}
			\State Selection() \Comment{Select individuals for reproduction}
			\State PredictiveCrossover() \Comment{Generative AI guides crossover}
			\State PredictiveMutation() \Comment{Generative AI guides mutation}
			\State AdaptiveSelectionPressure() \Comment{Adjust selection pressure dynamically}
			\State EvaluateFitness() \Comment{Re-evaluate fitness post-evolution}
			\State SemanticEncoding() \Comment{Enhance solution representation}
			\State SolutionInterpretationAndRefinement() \Comment{Refine solutions with AI}
			\State ContinuousEvolutionAdjustment() \Comment{Adjust GA parameters for next gen}
			\EndFor
			\State BestSolution $\gets$ IdentifyBestSolution() \Comment{Select best solution found}
			\State \Return BestSolution
			\EndProcedure
		\end{algorithmic}\label{fig:genetic-algorithm-Algogen-pseudocode}
	\end{algorithm}

	\begin{figure}
		\centering
		\includegraphics[width=0.45\textwidth]{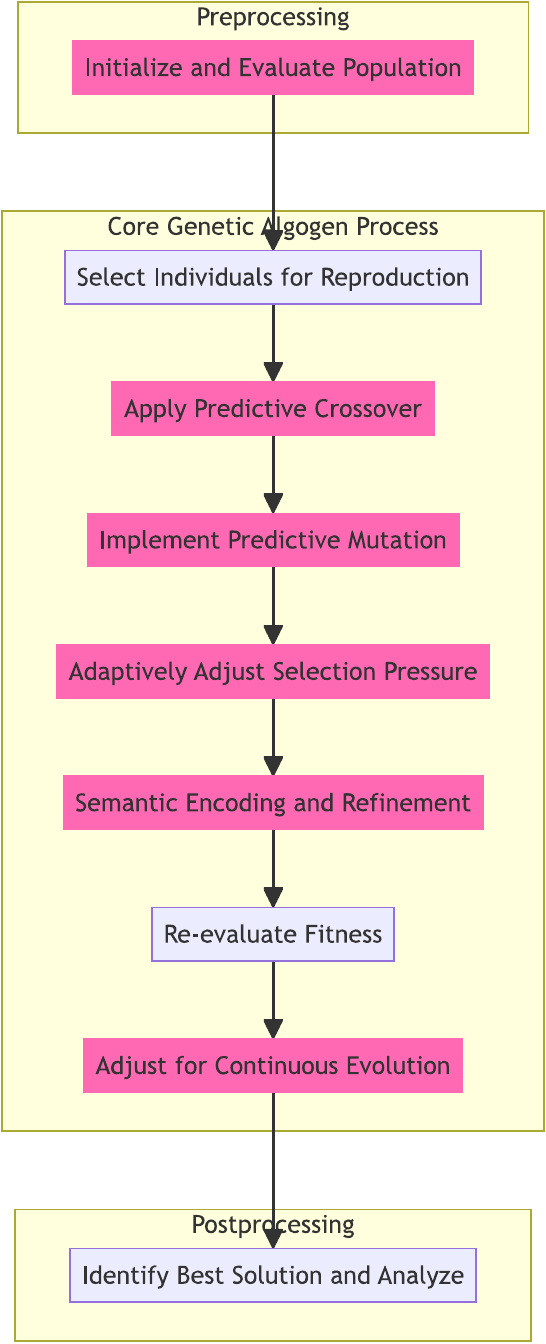} 
		\caption{Integrating Algogenic Enhancements into Genetic Algorithms: This diagram visualizes the comprehensive integration of Algogenic enhancements within the framework of Genetic Algorithms. Beginning with the preprocessing phase, 'Initialize and Evaluate Population' is marked by the application of generative AI to optimize initial population setup and evaluation, setting a strong foundation for the evolutionary process. The core algorithm process unfolds through a series of generative AI-enhanced steps, including 'Select Individuals for Reproduction' for targeted genetic propagation, 'Apply Predictive Crossover' and 'Implement Predictive Mutation' for dynamic genetic operations informed by AI predictions, and 'Adaptively Adjust Selection Pressure' to maintain an optimal balance of diversity and fitness within the population. 'Semantic Encoding and Refinement' leverages AI to imbue solutions with deep semantic value and refine them based on contextual insights, ensuring evolved solutions are not only optimal but meaningful. Continuous AI-driven adjustments, represented by 'Adjust for Continuous Evolution,' ensure the algorithm's parameters evolve in tandem with the solution space, enhancing adaptability and long-term effectiveness. The postprocessing phase, 'Identify Best Solution and Analyze,' underscores the culmination of the Algogenic process, where the best solutions are selected and subjected to further AI-driven analysis for potential improvement or insight, illustrating a holistic approach to problem-solving that leverages the synergy between genetic algorithms and generative AI for enhanced efficiency, adaptability, and solution quality in complex environments.}
		\label{fig:genetic_algorithm}
	\end{figure}

	\section{Gradient Descent}\index{Gradient Descent}
	
	\subsection{Introduction to Gradient Descent}
	\subsubsection{The Concept of Gradient Descent}
	\paragraph{Definition and Purpose} 
	Gradient Descent is an optimization algorithm used for minimizing the cost function in various machine learning algorithms, particularly in linear regression, logistic regression, and neural networks. It serves as the guiding force behind the iterative process of adjusting model parameters to optimize performance. By iteratively updating parameters in the direction of the steepest descent of the cost function, Gradient Descent enables machine learning models to converge towards the optimal solution. This iterative approach allows models to learn from data and make accurate predictions by gradually reducing prediction errors. Additionally, Gradient Descent plays a pivotal role in training deep neural networks, where the optimization of millions of parameters requires efficient and scalable optimization algorithms. Overall, Gradient Descent is instrumental in the success of machine learning algorithms, empowering them to learn from data and improve their performance over time.

	\paragraph{Operational Mechanism} 
	The operational mechanism of the algorithm hinges on iterative parameter adjustment to minimize the cost function. Initially, the algorithm computes the gradient of the cost function, representing the direction of steepest ascent. By moving in the opposite direction of this gradient, the algorithm progresses toward the optimal solution. This iterative process continues until convergence is achieved, signifying that further adjustments do not significantly reduce the cost function. Mathematically, this can be expressed as:
	
	\[
	\theta := \theta - \alpha \nabla J(\theta)
	\]
	
	where \( \theta \) represents the parameters, \( \alpha \) denotes the learning rate, and \( \nabla J(\theta) \) signifies the gradient of the cost function with respect to the parameters. Through this mechanism, the algorithm systematically refines its parameters, ultimately converging to the optimal solution.

	\paragraph{Learning Rate Significance} 
	
	The learning rate in Gradient Descent plays a pivotal role in determining the efficiency and effectiveness of the optimization process. Essentially, it controls the size of the steps taken towards the minimum of the cost function. If the learning rate is set too high, the algorithm may overshoot the minimum point, leading to oscillations or divergence from the optimal solution. Conversely, if the learning rate is too small, the convergence towards the minimum becomes sluggish, necessitating numerous iterations for convergence. Striking the right balance in setting the learning rate is therefore crucial for achieving optimal performance in Gradient Descent. Moreover, the choice of learning rate is intertwined with the nature of the cost function and the dataset characteristics. In practice, selecting an appropriate learning rate often involves experimentation and fine-tuning to find the optimal value that ensures swift convergence without sacrificing accuracy.
	
	\paragraph{Variants of Gradient Descent} 
	There are several variants of Gradient Descent, each tailored to address specific challenges encountered in optimization tasks. Batch Gradient Descent computes the gradient using the entire dataset, making it computationally expensive but offering precise updates. On the other hand, Stochastic Gradient Descent randomly selects single instances from the dataset for gradient computation, resulting in faster convergence but with more noise in the updates. Mini-batch Gradient Descent strikes a balance by using a small random subset of the dataset for each iteration, combining the advantages of both batch and stochastic approaches. These variants provide flexibility in optimizing different objective functions, enabling efficient optimization in various machine learning models.

	\subsubsection{Key Principles and Mechanisms}
	\paragraph{Objective Function Optimization} 
	The primary principle behind Gradient Descent is the optimization of an objective function, commonly known as the cost or loss function, which measures the difference between the predicted output of the model and the actual output. This function serves as a guide for the learning process, indicating the direction in which the model parameters should be adjusted to minimize errors. Through iterative updates, Gradient Descent gradually refines the parameter values to converge towards the optimal solution. By continuously evaluating and adjusting the objective function, the algorithm strives to improve the model's predictive accuracy and overall performance. Consequently, the optimization of the objective function lies at the heart of Gradient Descent and is fundamental to the success of machine learning algorithms.

	\paragraph{Gradient Calculation} 
	Gradient Descent relies on the calculation of gradients, which are partial derivatives of the cost function with respect to each parameter in the model. These gradients provide the direction in which the cost function has the steepest ascent. Consequently, by moving in the opposite direction, Gradient Descent aims to minimize the cost function, achieving convergence to the optimal solution. The calculation of gradients involves computing the change in the cost function corresponding to infinitesimal changes in each parameter. This process requires differentiation techniques, such as the chain rule, to determine how changes in one parameter affect the overall cost function. Moreover, in complex models with numerous parameters, computing gradients efficiently becomes a computational challenge. Therefore, optimization techniques like automatic differentiation and computational graph frameworks are employed to streamline the gradient calculation process and enhance efficiency. Additionally, the accuracy of gradient calculation is crucial for the convergence and stability of Gradient Descent algorithms. Even small errors in gradient computation can lead to suboptimal solutions or divergence from the desired outcome. Hence, meticulous attention to numerical precision and algorithmic implementation is essential to ensure reliable gradient calculation in optimization tasks.

	\paragraph{Update Rule}
	The update rule is fundamental to the iterative process of Gradient Descent, driving the optimization of model parameters towards minimizing the cost function. Essentially, it dictates how the parameters of the model are adjusted based on the calculated gradients. In each iteration, the current parameters \(\theta\) are updated by subtracting the product of the learning rate \(\alpha\) and the gradient of the cost function \(\nabla_\theta J(\theta)\). This adjustment aims to progressively refine the model's predictions by moving towards the direction of steepest descent in the parameter space. Consequently, smaller learning rates result in slower but potentially more stable convergence, while larger rates may lead to faster convergence but risk overshooting the optimal solution. Thus, the update rule plays a crucial role in balancing the trade-off between convergence speed and stability.

	\paragraph{Convergence Criteria} 
	The algorithm iterates this update process until it reaches a point where the cost function no longer decreases significantly with each iteration, indicating convergence to a minimum. This point is considered an optimal solution within the limits of the given model and data. However, it's important to note that this solution may only be a local minimum rather than a global minimum, depending on the nature of the cost function and the initial conditions of the algorithm. Consequently, the algorithm's convergence does not guarantee the discovery of the globally optimal solution. Nevertheless, by carefully selecting convergence criteria and monitoring convergence behavior, practitioners can ensure the algorithm's effectiveness in finding satisfactory solutions to optimization problems.
	
	\paragraph{Impact of Learning Rate}
	The learning rate \(\alpha\) is akin to the speed at which the algorithm learns from the data. It dictates the size of steps taken towards the optimal solution during each iteration of Gradient Descent. An ill-chosen learning rate can lead to suboptimal convergence, where the algorithm may either take excessively large steps, overshooting the minimum, or tiny steps, prolonging convergence unnecessarily. Therefore, striking the right balance with the learning rate is crucial for the efficiency and effectiveness of Gradient Descent. A too-high learning rate might cause the algorithm to oscillate around the minimum or even diverge, while a too-low learning rate might lead to sluggish convergence, prolonging the training process. Consequently, careful experimentation and tuning of the learning rate are essential to ensure the success of the optimization process. Moreover, adjusting the learning rate dynamically based on the feedback from the training process can further enhance the algorithm's adaptability and convergence speed.

	\subsubsection{The Role of Learning Rate}
	\paragraph{Defining Learning Rate} 
	The learning rate, denoted as \(\alpha\), plays a crucial role in guiding the Gradient Descent algorithm towards the minimum of the cost function during each iteration. Essentially, it determines the magnitude of the steps taken in parameter space, balancing the trade-off between convergence speed and stability. A smaller learning rate leads to slower convergence but ensures more precise optimization, while a larger learning rate accelerates convergence but risks overshooting the minimum. Therefore, selecting an appropriate learning rate is a delicate task, requiring careful consideration of the specific optimization problem and the characteristics of the dataset. Moreover, the learning rate interacts intricately with other hyperparameters and regularization techniques, influencing the overall performance and generalization ability of the learning algorithm.

	\paragraph{Impact on Convergence}
	The choice of the learning rate significantly influences the convergence behavior of the algorithm. A high learning rate may lead to rapid changes in the model's parameters, causing it to overshoot the optimal solution or even diverge from it altogether. Conversely, opting for a low learning rate results in slower adjustments to the parameters, potentially prolonging the convergence process and increasing computational overhead. Finding the optimal balance is crucial; a learning rate that is too high risks instability, while one that is too low may impede convergence efficiency. Therefore, careful tuning and experimentation are necessary to determine the most suitable learning rate for a given problem. Furthermore, understanding the underlying dynamics of the learning rate's impact on convergence can provide valuable insights into optimizing algorithm performance and efficiency.

	\paragraph{Adaptive Learning Rates} 
	To address the challenges associated with choosing an optimal learning rate, various strategies for adapting the learning rate during the optimization process have been developed. One common approach is learning rate decay, where the learning rate gradually decreases over time. This technique helps stabilize the training process and prevent overshooting of the optimal solution. Additionally, more sophisticated methods like Adam and RMSprop dynamically adjust the learning rate based on the gradient magnitudes and past gradients. These adaptive algorithms offer advantages in different scenarios, with Adam, for example, incorporating momentum to speed up convergence in the presence of sparse gradients. By dynamically adjusting the learning rate, these methods can effectively navigate complex optimization landscapes, leading to improved convergence and performance in training deep neural networks.

	\paragraph{Balancing Speed and Stability}
	The learning rate plays a pivotal role in optimizing the convergence process of algorithms, especially in iterative optimization methods like gradient descent. It acts as a guiding factor, determining the size of steps taken towards the minimum of the objective function. Too high a learning rate may lead to overshooting, where the algorithm fails to converge, bouncing back and forth around the optimal solution. Conversely, too low a learning rate may result in slow convergence, prolonging the optimization process unnecessarily. Hence, selecting an appropriate learning rate involves a delicate balance between the speed of convergence and the stability of the optimization process. This balance ensures that the algorithm efficiently reaches the minimum point of the objective function without oscillating or diverging.

	\paragraph{Experimentation and Tuning} 
	Experimentation and tuning are integral aspects of determining the optimal learning rate. In practical applications, this process often involves iterative experimentation with different learning rates to identify the one that maximizes the performance of the algorithm. Initially, a range of learning rates is selected, encompassing both high and low values. Through experimentation, the algorithm's behavior is observed, focusing on key performance metrics such as convergence speed and the accuracy of the resulting model. This iterative approach allows for a systematic narrowing down of the learning rate range, leading to the selection of an optimal value that balances the trade-off between convergence speed and model accuracy. Additionally, tuning involves adjusting other hyperparameters, such as regularization strength or network architecture, to further optimize the model's performance.

	\subsubsection{Applications and Limitations}
	\paragraph{Wide Range of Applications} 
	Gradient Descent, with its diverse applications, serves as a cornerstone in the realm of optimization algorithms, particularly within the domain of machine learning. Its utility spans across various models and tasks, ranging from simple linear and logistic regression to more complex neural networks and deep learning architectures. This algorithm's adaptability and effectiveness in minimizing cost functions render it indispensable in numerous fields where optimization plays a pivotal role. Its straightforward implementation and ability to handle large-scale datasets make it an ideal choice for tackling real-world problems across industries such as finance, healthcare, and marketing. Furthermore, its iterative nature allows for continuous refinement, making it suitable for dynamic environments where data distributions may change over time. Thus, Gradient Descent stands as a fundamental tool empowering advancements in machine learning and artificial intelligence.

	\paragraph{Scalability to Large Datasets}
	One of the notable advantages of Gradient Descent, particularly its stochastic and mini-batch variants, lies in its ability to handle large datasets efficiently. By updating model parameters using a subset of the data in each iteration, these variants mitigate the computational burden associated with processing extensive datasets. This approach not only accelerates the optimization process but also allows for the utilization of datasets that may exceed the memory capacity of the computing infrastructure. Additionally, the scalability of Gradient Descent enables the exploration of larger and more diverse datasets, facilitating more comprehensive learning and capturing intricate patterns within the data. Moreover, the parallelizability of stochastic and mini-batch Gradient Descent further enhances scalability by distributing computations across multiple processing units simultaneously, thereby leveraging parallel processing capabilities for expedited model training. Consequently, Gradient Descent's scalability to large datasets makes it a preferred optimization technique in scenarios where computational resources are limited or when dealing with massive data volumes.

	\paragraph{Challenges with Non-Convex Functions}
	Gradient Descent, while effective in optimizing convex functions, encounters significant challenges when dealing with non-convex functions. These functions often possess multiple local minima, complicating the convergence process. Despite efforts to reach the global minimum, Gradient Descent may converge to a local minimum instead, influenced by factors like initial parameter values and learning rates. In the realm of deep neural networks, characterized by complex architectures and high-dimensional parameter spaces, navigating non-convex loss landscapes becomes even more daunting. Consequently, practitioners must grapple with the inherent uncertainty of Gradient Descent's convergence behavior, necessitating careful hyperparameter tuning and exploration strategies to mitigate the risk of suboptimal solutions. While alternative optimization algorithms exist to address these challenges, such as stochastic gradient descent variants and metaheuristic approaches, they also have their own trade-offs and computational demands, underscoring the ongoing quest for efficient optimization techniques in the realm of non-convex functions.

	\paragraph{Sensitivity to Initial Conditions and Hyperparameters}
	Gradient Descent, a fundamental optimization algorithm, exhibits a pronounced sensitivity to its initial conditions and hyperparameters. The algorithm's effectiveness hinges on the proper configuration of initial parameter values and the meticulous selection of hyperparameters, notably the learning rate. Even minor deviations from optimal settings can have profound ramifications on the algorithm's performance. For instance, erroneous initialization or ill-suited learning rates can impede convergence, prolong convergence times, or steer the optimization process toward suboptimal solutions. Consequently, practitioners must exercise caution and judiciously fine-tune these parameters to mitigate the risk of convergence issues and ensure the algorithm's efficacy. The intricacies of calibrating these parameters underscore the nuanced nature of optimization tasks, emphasizing the need for meticulous experimentation and parameter tuning to unlock the algorithm's full potential.

	\paragraph{Computationally Intensive for Large Models} 
	While variants of Gradient Descent are designed to handle large datasets efficiently, the algorithm can still be computationally intensive when applied to models with a large number of parameters, such as deep neural networks. The computational burden is due to the need for repeated evaluations of gradients and updates to a large number of parameters over many iterations. Additionally, the computational complexity escalates with the size of the dataset and the depth of the network architecture. This heightened computational demand can lead to longer training times and increased resource consumption, posing challenges for real-time applications or environments with limited computational resources. Moreover, the computational intensity may necessitate the use of specialized hardware accelerators or distributed computing frameworks to expedite the training process and mitigate resource constraints. Consequently, while Gradient Descent variants offer scalability and efficiency for large-scale datasets, their application to complex models requires careful consideration of computational requirements and optimization strategies to ensure practical feasibility and performance.

	\paragraph{Advances and Innovations} Despite its limitations, ongoing research and development in the field of optimization algorithms have led to numerous enhancements and variations of Gradient Descent. These advancements aim to overcome its limitations, such as introducing momentum to accelerate convergence, employing adaptive learning rates to handle complex loss surfaces, and developing algorithms that are more robust to the choice of hyperparameters. Furthermore, researchers have explored techniques like Nesterov Accelerated Gradient (NAG) and AdaGrad, which offer improvements in terms of convergence speed and adaptability to different optimization landscapes. Moreover, recent innovations include the incorporation of second-order optimization methods like Newton's method and variants like BFGS and L-BFGS, addressing challenges related to saddle points and ill-conditioned matrices. Additionally, advancements in stochastic gradient descent variants like Adam and RMSProp have demonstrated superior performance in various deep learning tasks by effectively adjusting learning rates for individual parameters based on their gradients' historical behavior.

	\subsubsection{Pseudocode for Algorithmic Gradient Descent}
	The Gradient Descent algorithm is a versatile optimization technique used for minimizing a function by iteratively moving in the direction of steepest descent. Unlike A*, which focuses on finding the most cost-effective path within a graph, Gradient Descent is employed in continuous optimization tasks. It operates by iteratively adjusting parameters to minimize a loss function, aiming to reach a local minimum. Pseudocode for Gradient Descent is depicted in Figure \ref{fig:gradient-descent-pseudocode}, demonstrating its iterative approach to optimizing functions.
	
	\begin{algorithm}
		\caption{Gradient Descent Pseudocode}
		\begin{algorithmic}[1]
			\Procedure{GradientDescent}{$J(\theta)$, $\alpha$, $\epsilon$, $\theta_{\text{init}}$}
			\State Initialize $\theta$ to $\theta_{\text{init}}$
			\State Initialize gradient norm, $g_{\text{norm}}$, to a value greater than $\epsilon$
			\While{$g_{\text{norm}} > \epsilon$}
			\State Compute gradient, $g = \nabla_{\theta} J(\theta)$
			\State Update $\theta$: $\theta = \theta - \alpha \cdot g$
			\State Compute $g_{\text{norm}} = \left\| g \right\|$
			\EndWhile
			\State \Return $\theta$
			\EndProcedure
		\end{algorithmic}\label{fig:gradient-descent-pseudocode}
	\end{algorithm}

\subsection{Algogenic Enhancements for Gradient Descent}
\subsubsection{Problem Space Analysis and Optimization}
\paragraph{Enhancing Problem Understanding with AI}
Integrating generative AI into Gradient Descent's preprocessing phase significantly augments the algorithm's ability to understand and navigate the problem space. By analyzing vast datasets, generative AI identifies and addresses potential impediments such as local minima and steep gradients. This enriched understanding suggests strategic interventions like advanced data preprocessing and insightful normalization, thereby reshaping the problem space for more effective gradient descent exploration. Furthermore, this integration facilitates a more informed decision-making process, blending domain expertise with data-driven insights. The iterative nature of generative AI promotes continuous adaptation, ensuring the optimization framework's agility in real-time, fostering innovation, and enhancing the overall problem-solving approach.

\paragraph{Optimizing Initial Conditions}
Optimizing initial conditions is crucial for the gradient descent algorithm's efficiency. By leveraging generative AI, the algorithm benefits from predictive analytics to select optimal starting points and parameter configurations, enhancing the exploration and avoiding potential pitfalls. This strategy not only improves robustness but also facilitates a dynamic adjustment of initial conditions, responding adeptly to evolving problem dynamics. The strategic selection and continuous adaptation of initial conditions play a vital role in navigating the solution space effectively, underscoring the importance of informed initial setups in achieving optimal convergence.

\paragraph{Impact on Gradient Descent}
The application of generative AI to gradient descent optimization significantly improves the algorithm's understanding of the optimization landscape, enabling a more intelligent navigation and faster convergence. By optimizing problem space analysis and initial conditions, gradient descent benefits from a strategic start and an enhanced ability to overcome optimization challenges. This approach not only accelerates the search for the global minimum but also expands the algorithm's applicability to more complex problems, thereby transforming gradient descent into a more versatile and effective optimization tool.

\subsubsection{Initial Parameter Optimization}
\paragraph{Tailoring Starting Points with AI Insight}
Leveraging generative AI for the selection of starting parameters in Gradient Descent, such as the initial point and learning rate, enhances the optimization process. This AI-driven approach intelligently predicts suitable starting conditions, improving convergence speed and avoiding common pitfalls. By analyzing past optimization tasks, generative AI adapts initialization strategies dynamically, enhancing the optimization's adaptability and efficiency. This method not only increases the robustness of the algorithm but also encourages exploration of diverse optimization landscapes, fostering continuous learning and iterative performance improvement.

\paragraph{Strategic Parameter Selection}
Generative AI facilitates strategic parameter selection by incorporating insights from similar optimization tasks and domain knowledge, improving gradient descent's efficiency and effectiveness. This tailored approach, informed by past experiences and specific problem characteristics, optimizes the exploration-exploitation balance, enhancing convergence rates and overall optimization outcomes. Strategic parameter selection, by adapting parameters like learning rates and regularization terms, ensures that the optimization process is both efficient and aligned with the problem's intricacies.

\paragraph{Enhancing Convergence Efficiency}
Generative AI significantly enhances Gradient Descent's convergence efficiency by strategically selecting initial parameters and refining the learning rate. This proactive adjustment reduces computational demands and increases the optimization process's robustness. By dynamically adapting to the problem's specific characteristics, Gradient Descent achieves faster convergence and improved performance, demonstrating the value of integrating generative AI in optimizing initial parameter settings.

\subsubsection{Dynamic Learning Rate Adjustment}
\paragraph{Optimizing Convergence with Real-time AI Analysis}
Dynamic Learning Rate Adjustment, powered by generative AI, enables Gradient Descent to modulate the learning rate dynamically, optimizing the balance between rapid convergence and stability. This real-time adjustment mechanism adapts to changing optimization landscapes, enhancing the algorithm's efficiency and effectiveness. By leveraging AI analysis, Gradient Descent navigates through complex problems more adeptly, demonstrating the significant impact of dynamic learning rate adjustments on optimizing algorithm performance.

\paragraph{Implementing AI-driven Adaptations}
Generative AI's role in dynamically adjusting the learning rate involves real-time analysis of the optimization trajectory, enabling strategic interventions to accelerate or stabilize the descent. This AI-driven approach ensures Gradient Descent's adaptability, allowing for anticipatory adjustments that mitigate risks and exploit favorable conditions. Implementing AI-driven adaptations enhances the algorithm's responsiveness and strategic foresight, underlining the importance of adaptive strategies in optimization processes.

\paragraph{Achieving Enhanced Algorithm Performance}
Dynamic Learning Rate Adjustment significantly improves Gradient Descent's performance by allowing adaptive learning rate modifications based on AI insights. This enhancement ensures optimal convergence speeds, stability, and adaptability to varying problem complexities, showcasing the profound impact of generative AI on the algorithm's efficiency and effectiveness. By dynamically adjusting the learning rate, Gradient Descent navigates optimization challenges more adeptly, highlighting the crucial role of adaptive learning strategies in achieving superior optimization outcomes.

\subsubsection{Pathway Optimization}
\paragraph{Navigating the Cost Function Landscape Intelligently}
Pathway Optimization employs generative AI to enhance Gradient Descent's ability to navigate the cost function landscape efficiently. By predicting potential challenges and adjusting the descent trajectory intelligently, the algorithm avoids inefficiencies and converges faster towards the optimal solution. This approach not only adapts dynamically to changes but also utilizes global context to bypass obstacles, significantly improving the algorithm's efficiency and effectiveness across various optimization tasks.

\paragraph{Strategic Path Selection with AI Insights}
Generative AI enhances Gradient Descent by strategically selecting pathways based on predictive models and real-time data, optimizing the descent trajectory. This approach enables the algorithm to navigate complex landscapes more effectively, avoiding challenges and exploiting favorable conditions. By leveraging reinforcement learning and domain knowledge, generative AI continuously refines its path selection strategy, leading to more efficient and effective optimization.

\paragraph{Enhancing Efficiency and Convergence}
Integrating Pathway Optimization into Gradient Descent significantly improves the algorithm's efficiency and convergence reliability. By optimizing the traversal pathway, the algorithm reduces unnecessary iterations and avoids suboptimal regions, enhancing its effectiveness. This generative AI-driven approach ensures that Gradient Descent navigates the cost function landscape with precision and adaptability, marking a significant advancement in optimization strategies.

\subsubsection{Solution Refinement and Validation}
\paragraph{Ensuring Practicality and Robustness of Solutions}
Solution Refinement and Validation enhance Gradient Descent by ensuring solutions are practical and robust for real-world applications. This phase leverages generative AI to rigorously evaluate solutions against various criteria, including domain-specific requirements and scalability. By simulating real-world conditions and incorporating feedback loops, this process iteratively improves solutions, enhancing their viability and relevance across diverse scenarios.

\paragraph{Applying Generative AI for Comprehensive Analysis}
Generative AI facilitates a comprehensive analysis of Gradient Descent's solutions, simulating implementation scenarios and comparing against benchmarks. This approach uncovers potential improvements and compliance issues, ensuring solutions are effective and relevant. By continuously learning and adapting, generative AI anticipates future challenges, enabling ongoing refinement and optimization, thus driving continuous improvement and innovation.

\paragraph{Achieving Solution Excellence}
Integrating Solution Refinement and Validation into Gradient Descent, powered by generative AI, ensures the delivery of high-quality, practically applicable solutions. This Algogenic enhancement enables iterative refinement and rigorous validation, enhancing the algorithm's ability to produce robust solutions suitable for real-world deployment. This comprehensive approach to optimization not only achieves theoretical excellence but also translates into tangible benefits in practical applications.

\subsubsection{Continuous Learning Loop}
\paragraph{Adapting to Evolving Environments and Data}
The Continuous Learning Loop, powered by generative AI, enables Gradient Descent to adapt to changing environments and data over time. This Algogenic enhancement supports iterative learning, augmenting the algorithm's dataset diversity and resilience against data shifts. By incorporating feedback mechanisms, Gradient Descent continuously refines its strategies and predictions, demonstrating the transformative impact of continuous learning on optimization techniques.

\paragraph{Implementing Feedback Mechanisms}
Sophisticated feedback mechanisms, integrated through generative AI, allow Gradient Descent to analyze and learn from the outcomes of its optimization processes. This cyclical learning process iteratively improves the algorithm's decision-making and adaptability, ensuring relevance and effectiveness in diverse contexts. Implementing feedback mechanisms represents a crucial step towards evolving algorithmic frameworks that respond dynamically to changes and challenges.

\paragraph{Evolving Algorithmic Intelligence}
Incorporating a Continuous Learning Loop significantly enhances Gradient Descent's intelligence, enabling it to learn from experiences and adapt to new challenges. This evolution transforms the algorithm into a dynamic entity capable of offering more accurate and efficient solutions, demonstrating the critical role of continuous learning in advancing algorithmic performance and adaptability in a changing world.

\subsubsection{Predictive Convergence Analysis}
\paragraph{Forecasting Optimization Success}
Predictive Convergence Analysis, utilizing generative AI, forecasts the likelihood of successful convergence in Gradient Descent. This enhancement evaluates the algorithm's trajectory, allowing for real-time adjustments to optimize outcomes. By anticipating convergence challenges and navigating through them, this approach significantly improves the efficiency and effectiveness of the optimization process, showcasing the pivotal role of predictive analysis in optimization algorithms.

\paragraph{Strategic Adjustments Based on Predictions}
Predictive models enable strategic adjustments in Gradient Descent, enhancing decision-making with foresight. Anticipating future challenges, the algorithm optimizes its path, balancing exploration and exploitation. This proactive strategy, informed by predictive insights, guides the algorithm towards desirable outcomes, demonstrating the importance of adaptability and strategic foresight in optimizing performance.

\paragraph{Enhancing Efficiency and Outcome Reliability}
Incorporating Predictive Convergence Analysis into Gradient Descent improves efficiency and reliability, streamlining the optimization process and ensuring confidence in the outcomes. By adapting to predictive insights, the algorithm navigates optimization challenges more adeptly, showcasing the critical role of predictive analysis in achieving superior optimization results and enhancing algorithmic robustness.

	\subsubsection{Pseudocode for Algogenic Gradient Descent}
	The Algogenic gradient descent approach leverages AI to augment traditional gradient descent methods by dynamically adjusting descent parameters and strategies based on observed function behavior and real-time error estimates. This pseudocode, accessible in \ref{fig:gradient-descent-Algogen-pseudocode}, outlines a sophisticated framework integrating AI-driven enhancements for adaptive step size selection, domain partitioning, error estimation, and real-time parameter optimization.
	
	\begin{algorithm}
		\caption{Algogenic Gradient Descent Pseudocode}
		\begin{algorithmic}[1]
			\Procedure{AlgogenicGradientDescent}{$f, \mathbf{x}_0, \epsilon$}
			\State $\mathbf{x} \gets \mathbf{x}_0$ \Comment{Initialize starting point}
			\State $lr \gets \Call{InitialParameterOptimization}{}$ \Comment{Optimize initial learning rate}
			\While{$\lVert \nabla f(\mathbf{x}) \rVert > \epsilon$}
			\State $\mathbf{g} \gets \Call{GradientPredictionAndSmoothing}{\mathbf{x}}$ \Comment{Predict and smooth gradient}
			\State $lr \gets \Call{DynamicLearningRateAdjustment}{lr, \mathbf{g}}$ \Comment{Adjust learning rate dynamically}
			\State $\mathbf{x} \gets \mathbf{x} - lr \times \mathbf{g}$ \Comment{Update position}
			\State $\Call{PathwayOptimization}{\mathbf{x}}$ \Comment{Optimize descent pathway}
			\EndWhile
			\State $\mathbf{x}^* \gets \Call{SolutionRefinementAndValidation}{\mathbf{x}}$ \Comment{Refine and validate solution}
			\State $\Call{ContinuousLearningLoop}{f, \mathbf{x}^*}$ \Comment{Update AI models for future runs}
			\State \Return $\mathbf{x}^*$
			\EndProcedure
		\end{algorithmic}\label{fig:gradient-descent-Algogen-pseudocode}
	\end{algorithm}

	\begin{figure}
		\centering
		\includegraphics[width=0.7\textwidth]{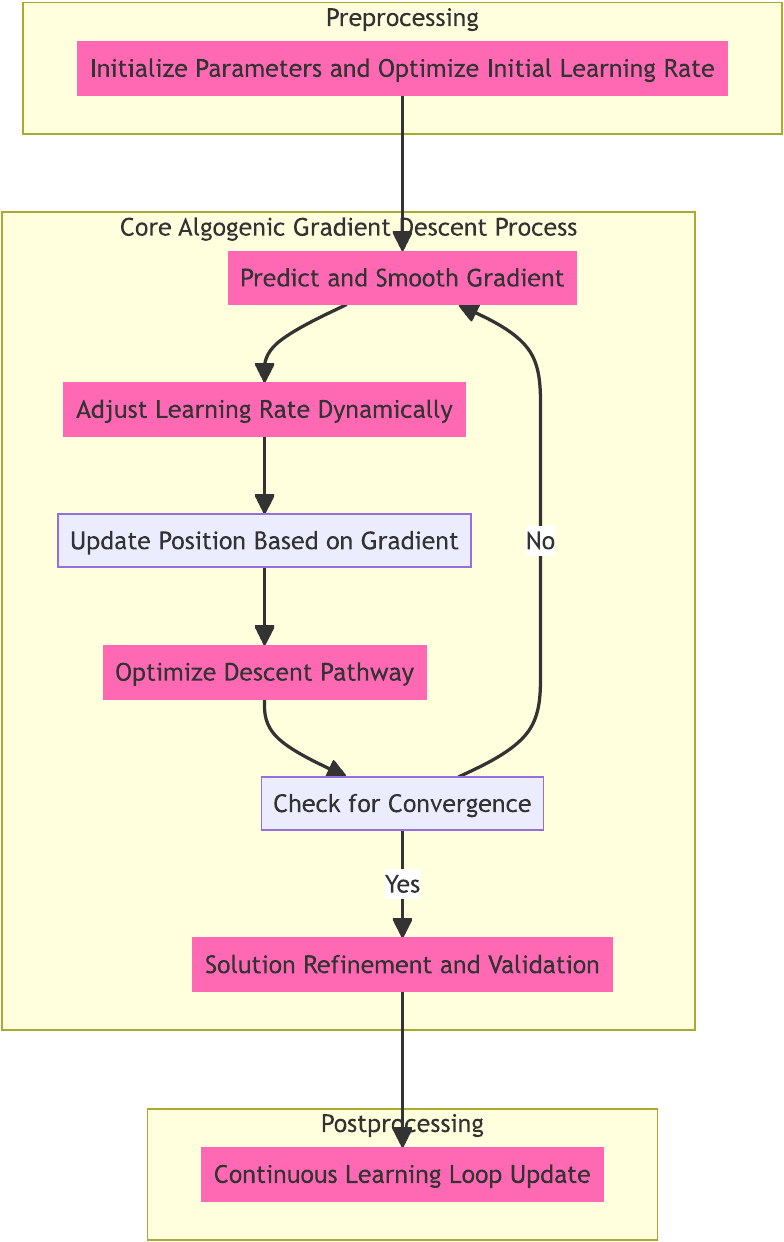} 
		\caption{Integrating Algogenic Enhancements into Gradient Descent: This diagram visualizes the comprehensive integration of generative AI enhancements within the Gradient Descent optimization process. Starting with the preprocessing phase, 'Initialize Parameters and Optimize Initial Learning Rate' utilizes AI to set optimal initial conditions, priming the algorithm for efficient convergence. The core process unfolds through AI-enhanced steps such as 'Predict and Smooth Gradient' and 'Adjust Learning Rate Dynamically,' which ensure the descent is guided intelligently through the cost function landscape. 'Update Position Based on Gradient' represents the algorithmic heart of Gradient Descent, directly influenced by AI optimizations. 'Optimize Descent Pathway' and iterative 'Check for Convergence' further leverage AI to refine the descent trajectory and validate progress. Postprocessing with 'Solution Refinement and Validation' and 'Continuous Learning Loop Update' closes the loop, applying AI to ensure solutions are not only optimal but applicable, and feeding insights back into the system for continuous improvement. This Algogenic approach transforms Gradient Descent into a dynamic, self-improving algorithm capable of tackling complex optimization challenges with enhanced adaptability and insight.}
		\label{fig:gradient_descent}
	\end{figure}

	\section{Simulated Annealing}\index{Simulated Annealing}
	\subsection{Introduction to Simulated Annealing}
	
	\subsubsection{The Concept of Simulated Annealing}
	\paragraph{Origins and Inspiration}
	Simulated Annealing (SA) is an optimization technique inspired by the physical process of annealing in metallurgy. This process involves heating and then slowly cooling a material to decrease defects and increase the size of its crystals, thereby reducing its system energy to a minimum. Analogously, SA seeks to find a minimum (or maximum) of a function that represents some "energy" of a system, often in the context of complex optimization problems. While traditional optimization algorithms may get stuck in local minima, SA employs a probabilistic approach, allowing it to escape these local optima and explore a wider solution space. Additionally, SA's ability to accept worse solutions with a certain probability enables it to navigate through rugged landscapes, making it particularly suitable for problems with multiple local minima. This adaptability and exploration capability are key factors contributing to SA's effectiveness in finding near-optimal solutions for various real-world optimization challenges.

	\paragraph{Probabilistic Approach to Optimization}
	In simulated annealing (SA), the introduction of a probabilistic element revolutionizes the traditional deterministic approach to optimization. Unlike deterministic algorithms, SA injects a dose of randomness into its decision-making process, akin to rolling a dice to explore potential solutions within the solution space. This probabilistic exploration is governed by a metaphorical "temperature" parameter, representing the algorithm's willingness to accept suboptimal solutions. At higher temperatures, the algorithm behaves more erratically, willingly accepting worse solutions with a higher probability. This characteristic allows SA to escape local optima and explore a broader range of potential solutions early in the optimization process. Over time, as the temperature cools according to a predefined schedule, the algorithm's behavior becomes more deterministic, gradually converging towards an optimal solution. This dynamic interplay between randomness and determinism imbues SA with the flexibility to navigate complex solution spaces and converge towards globally optimal solutions, making it a powerful tool in optimization tasks across various domains.

	\paragraph{Cooling Schedule and Convergence}
	The cooling schedule in simulated annealing (SA) plays a pivotal role in guiding the optimization process towards a satisfactory solution. It acts as a bridge between exploration and exploitation, delicately balancing the trade-off between thorough search and convergence. The choice of cooling schedule significantly influences the convergence rate and the quality of the final solution. A gradual decrease in temperature allows the algorithm to explore a wider solution space in the initial stages, preventing premature convergence to local optima. Conversely, a rapid decrease may hinder exploration, causing the algorithm to converge too quickly, possibly to suboptimal solutions. Therefore, designing an effective cooling schedule requires careful consideration of the problem's complexity, computational resources, and desired solution quality. Mathematical formulations, such as exponential or logarithmic functions, are often utilized to regulate the cooling rate, ensuring a smooth transition towards convergence while avoiding stagnation.

	\paragraph{Acceptance Criteria}
	The acceptance of new solutions in simulated annealing (SA) relies on the Metropolis-Hastings algorithm, a pivotal mechanism ensuring the exploration of solution space. When a new solution enhances the objective function, it is promptly accepted, driving iterative improvement. However, the algorithm also allows for the acceptance of solutions that may not directly enhance the objective function. This acceptance is contingent upon a probabilistic assessment, determined by factors including the discrepancy in objective function values between the current and prospective solutions, alongside the prevailing temperature. This probabilistic evaluation, encapsulated in the formula $P(\text{accept}) = \exp\left(-\frac{\Delta E}{kT}\right)$, underscores the nuanced balance between exploitation and exploration within SA. Here, $\Delta E$ signifies the change in objective function value, $T$ denotes the current temperature, and $k$ serves as a scaling constant modulating temperature's influence.

	\paragraph{Application Scope}
	SA is particularly suited for optimization problems where the search space is complex and not well understood, including those with multiple local minima. Its flexibility and general applicability make it a valuable tool across a wide range of disciplines. Furthermore, SA's ability to escape local optima and explore diverse regions of the solution space sets it apart from traditional optimization methods. Moreover, its iterative nature allows for the incorporation of various constraints and objectives, making it adaptable to diverse problem domains. Additionally, SA's effectiveness in handling non-linear and non-convex optimization problems enhances its relevance in fields such as machine learning, where complex objective functions are prevalent. Hence, SA serves as a versatile and powerful optimization technique, offering insights and solutions to challenging problems in operations research, engineering, and beyond.
	
	\subsubsection{Key Principles and Mechanisms}
	\paragraph{Exploration and Exploitation}
	Simulated Annealing operates on the balance between exploration of the search space and exploitation of the best solutions found. At high temperatures, the algorithm encourages exploration by allowing acceptance of solutions that are worse than the current solution, enabling it to escape local minima and explore more of the solution space. Conversely, as the temperature decreases, the algorithm gradually shifts its focus towards exploitation, becoming more selective in accepting new solutions and concentrating its search around the best solutions discovered. This delicate interplay between exploration and exploitation is essential for Simulated Annealing to effectively navigate complex optimization landscapes and converge to high-quality solutions.
	
	\paragraph{Temperature as a Control Parameter}
	In Simulated Annealing, the temperature acts as a pivotal control parameter governing the balance between exploration and exploitation within the search space. As the temperature increases, the algorithm's propensity for accepting worse solutions grows, facilitating a more extensive exploration of the solution landscape. This heightened exploration is akin to a broadened horizon, allowing the algorithm to venture into diverse regions of the search space, regardless of their immediate objective function value. Consequently, in the initial stages of the optimization process, characterized by higher temperatures, the algorithm is bestowed with the flexibility to escape local optima and explore potentially promising regions that might harbor globally optimal solutions. Thus, the temperature parameter serves as a mechanism for injecting an essential dose of randomness into the search process, enabling the algorithm to navigate through the solution space with versatility and efficacy. By embracing a broader perspective facilitated by elevated temperatures, Simulated Annealing can effectively overcome the limitations of local exploration and pave the way towards discovering high-quality solutions.

	\paragraph{Decreasing Temperature Schedule}
	The mechanism by which the temperature decreases over time, known as the cooling schedule, is pivotal to the success of Simulated Annealing. The cooling schedule serves as a guiding principle, orchestrating the gradual reduction of temperature throughout the optimization process. This meticulous orchestration is imperative, as it directly influences the algorithm's ability to navigate the solution space effectively. One commonly employed strategy is the implementation of a geometric decay, where the temperature decreases exponentially over iterations. This approach ensures a systematic exploration of the search space, striking a delicate balance between exploration and exploitation. However, alternative methods exist, each tailored to specific optimization scenarios. For instance, in scenarios where swift convergence is desired, a linear cooling schedule might be preferred, ensuring a steady and predictable decline in temperature. Conversely, in situations where intricate exploration of the solution space is paramount, a logarithmic cooling schedule might be more appropriate. Ultimately, the choice of cooling schedule hinges on the unique characteristics of the optimization problem at hand and the trade-off between exploration and exploitation it entails.
	
	\paragraph{Acceptance Probability}
	The acceptance probability, as defined by the Boltzmann distribution, plays a pivotal role in simulated annealing algorithms. As the temperature decreases, the probability of accepting a worse solution diminishes gradually. This mathematical relationship is crucial for the algorithm's ability to explore the solution space effectively while gradually converging towards optimal or near-optimal solutions. The formula $\exp\left(-\frac{\Delta E}{kT}\right)$ encapsulates this probabilistic decision-making process, where $\Delta E$ represents the difference in energy or cost between the current and proposed solutions, $T$ denotes the temperature, and $k$ serves as a constant factor regulating the effect of temperature on acceptance probability.
	
	In practical terms, as the algorithm progresses through iterations, the temperature decreases, signifying a decrease in the system's "excitability" or willingness to accept suboptimal solutions. This mechanism mirrors the real-world annealing process, where materials cool down to achieve a more stable state. By incorporating the Boltzmann distribution, simulated annealing strikes a balance between exploration and exploitation, allowing it to navigate complex solution landscapes effectively. Consequently, the algorithm can escape local optima and search for globally optimal solutions, making it a versatile and powerful optimization technique in various problem domains.
	
	\paragraph{Convergence to Global Optimum}
	The Simulated Annealing algorithm navigates towards the global optimum of the objective function through a process characterized by the gradual reduction in temperature and the probabilistic acceptance of new solutions. This methodical approach ensures that the algorithm explores the solution space comprehensively, avoiding premature convergence to local optima. While theoretical guarantees of convergence exist under certain conditions, such as an infinitely slow cooling schedule, practical implementations prioritize striking a balance between computational efficiency and the quality of the solution obtained. By iteratively adjusting the temperature parameter and accepting or rejecting candidate solutions based on a stochastic criterion, Simulated Annealing converges towards an optimal solution over time. This convergence process is akin to the annealing of metal, where the material is gradually cooled to reach a stable and desirable state. Through this iterative refinement process, Simulated Annealing effectively explores the solution landscape, ultimately converging to the global optimum with a high probability.

	\subsubsection{The Role of the Cooling Schedule}
	\paragraph{Defining the Cooling Schedule}
	The cooling schedule in Simulated Annealing plays a pivotal role in orchestrating the delicate balance between exploration and exploitation throughout the optimization process. Essentially, it serves as a blueprint guiding the algorithm's progression by determining how rapidly the temperature decreases over time. This schedule is not arbitrary but meticulously designed to ensure efficient exploration of the solution space in the algorithm's initial stages while gradually shifting focus towards exploiting promising solutions as the temperature diminishes. Consequently, the cooling schedule directly influences the algorithm's ability to escape local optima and converge towards globally optimal solutions. Moreover, the selection of an appropriate cooling schedule is heavily reliant on the problem domain and the specific characteristics of the objective function being optimized. Therefore, careful consideration and experimentation are paramount in crafting an effective cooling schedule tailored to the intricacies of each optimization problem.
	
	\paragraph{Importance of Balance Between Exploration and Exploitation}
	A well-designed cooling schedule ensures that there is sufficient exploration at the beginning of the algorithm when the temperature is high. This allows the algorithm to escape local minima and explore a wide range of potential solutions. As the temperature decreases, the algorithm gradually shifts its focus towards exploitation, honing in on areas of the search space that contain promising solutions and refining these to find the optimum. It's crucial to strike a balance between exploration and exploitation to avoid getting stuck in local optima or missing out on potentially better solutions. This delicate equilibrium is akin to navigating a rugged terrain where one must tread carefully to avoid getting trapped in valleys while also climbing peaks to reach the summit. Therefore, the interplay between exploration and exploitation is vital for the success of optimization algorithms, ensuring they efficiently converge to high-quality solutions without prematurely converging to suboptimal ones. Moreover, this balance is not static but dynamically evolves throughout the optimization process, adapting to the changing landscape of the search space. Consequently, algorithms with well-tailored exploration-exploitation strategies exhibit robust performance across a wide range of optimization tasks, making them indispensable tools in various domains.
	
	\paragraph{Common Cooling Schedules}
	Various strategies exist for cooling, each with its own advantages and applications. The geometric cooling schedule, for instance, offers a straightforward approach by decreasing the temperature at each iteration by a fixed ratio. This method is particularly attractive for its simplicity and computational efficiency, making it a popular choice in many optimization problems. However, linear and logarithmic cooling schedules also play crucial roles in certain scenarios. Linear schedules involve decrementing the temperature by a constant amount in each iteration, providing a steady reduction in temperature. On the other hand, logarithmic schedules gradually decrease the temperature, resulting in a slower cooling rate over time. The selection of a suitable cooling schedule depends on the nature of the problem at hand and the desired trade-offs between exploration and exploitation. Factors such as the initial temperature, final temperature, and cooling rate must be carefully tuned to achieve optimal performance. Ultimately, the effectiveness of a cooling schedule hinges on its ability to strike a balance between exploration of the solution space and exploitation of promising regions, ultimately leading to convergence towards an optimal solution.
	
	\paragraph{Impact on Algorithm Convergence}
	The cooling schedule plays a pivotal role in determining the convergence of the Simulated Annealing algorithm towards a global optimum. It acts as a guiding mechanism, orchestrating the balance between exploration and exploitation within the search space. If the temperature decreases too rapidly, the algorithm risks getting trapped in local minima, impeding its ability to explore promising regions. Conversely, a sluggish decrease in temperature prolongs the exploration phase excessively, leading to computational inefficiency. Striking the right balance is imperative, necessitating careful consideration of the cooling rate. Theoretical analyses underscore the importance of a gradual temperature reduction, ensuring that the algorithm can effectively traverse the solution landscape. Empirical studies corroborate these findings, demonstrating that adhering to appropriate cooling schedules significantly enhances the algorithm's likelihood of converging to the global optimum. Thus, the meticulous design and calibration of the cooling schedule emerge as critical factors in shaping the convergence behavior of Simulated Annealing.

	\paragraph{Adaptive Cooling Schedules}
	Recent advancements in Simulated Annealing have explored the use of adaptive cooling schedules, where the rate of cooling is adjusted dynamically based on the algorithm's progress. This approach allows for more flexibility and can lead to improved performance on complex optimization problems by automatically adjusting the exploration-exploitation balance in response to the observed behavior of the search process. Adaptive cooling schedules address the inherent challenge of selecting an appropriate cooling rate, which can significantly impact the convergence speed and final solution quality. By adaptively adjusting the cooling rate during the optimization process, the algorithm can effectively navigate rugged search spaces and escape local optima more efficiently. Furthermore, these schedules enhance the algorithm's ability to explore the solution space comprehensively while exploiting promising regions for potential improvements. This adaptive mechanism ensures that the algorithm maintains a robust exploration strategy throughout the optimization process, effectively balancing exploration and exploitation to achieve optimal solutions. Moreover, adaptive cooling schedules offer a dynamic approach to optimization, allowing the algorithm to respond flexibly to changes in the problem landscape and adapt its search strategy accordingly. Overall, the integration of adaptive cooling schedules into Simulated Annealing represents a significant advancement in optimization algorithms, offering greater adaptability and efficiency in solving complex optimization problems.

	\subsubsection{Applications and Limitations}
	\paragraph{Broad Range of Applications}
	Simulated Annealing has been successfully applied to a broad spectrum of optimization challenges, spanning diverse domains and industries, which underscores its adaptability and efficacy. Notably, it has found utility in addressing scheduling conundrums such as the renowned traveling salesman problem, where it excels in finding near-optimal solutions amid intricate spatial configurations. Moreover, within engineering disciplines, Simulated Annealing emerges as a formidable tool for optimizing design parameters, facilitating the creation of robust and efficient systems. In logistics, it plays a pivotal role in tackling allocation dilemmas, optimizing resource allocation to streamline operations and minimize costs. Additionally, its utility extends to the realm of machine learning and statistics, where it contributes to model fitting endeavors by navigating high-dimensional parameter spaces to uncover optimal configurations. Its distinct advantage lies in its ability to traverse complex search landscapes devoid of gradient information, making it indispensable for scenarios where conventional optimization techniques encounter insurmountable challenges.

	\paragraph{Advantages in Complex Search Spaces}
	Simulated Annealing possesses distinct advantages in navigating complex search spaces, particularly those characterized by multimodal distributions and numerous local optima. While traditional gradient-based optimization methods often struggle in such environments due to their susceptibility to getting stuck in local minima, Simulated Annealing's probabilistic approach to accepting suboptimal moves enables it to explore the search space more extensively. By occasionally accepting worse solutions, the algorithm can effectively escape local traps and continue its pursuit of global optima. This characteristic makes Simulated Annealing well-suited for tackling problems with intricate landscapes, where the presence of multiple peaks and valleys poses significant challenges to traditional optimization techniques. Through its adaptive exploration strategy, Simulated Annealing demonstrates robustness and versatility in optimizing complex objective functions, offering a valuable tool for solving real-world optimization problems in diverse domains.

	\paragraph{Dependence on the Cooling Schedule}
	The effectiveness of Simulated Annealing is heavily influenced by the choice of the cooling schedule, including the initial temperature, the cooling rate, and the termination condition. Finding the right cooling schedule is often problem-specific and can require significant experimentation and tuning. This dependence can be seen as both a strength and a limitation. On one hand, it offers flexibility to tailor the algorithm to the specific problem at hand, allowing for optimization in various scenarios. Conversely, this reliance on the cooling schedule also poses challenges, as identifying the optimal parameters may necessitate extensive empirical testing and computational resources. Moreover, the sensitivity of the algorithm to changes in the cooling schedule underscores the importance of understanding the underlying problem dynamics and selecting appropriate parameters accordingly. Thus, while the dependence on the cooling schedule offers customization potential, it also demands careful consideration and analysis to ensure the algorithm's effectiveness and efficiency.

	\paragraph{Computational Cost and Convergence Time}
	Simulated Annealing (SA) offers a powerful approach for optimization, aiming to reach the global optimum solution. However, despite its theoretical potential for convergence, the practical execution of SA often encounters challenges related to computational cost and convergence time. The algorithm's iterative nature involves exploring the solution space through probabilistic transitions, which can lead to a slow convergence process. Moreover, for large or complex problem instances, the computational requirements for achieving convergence to the global optimum can become prohibitively high. This arises due to the need for a large number of iterations to adequately explore the solution landscape and identify the optimal configuration. As a result, practitioners often face a trade-off between computational resources and the desired solution quality. To address this challenge, heuristic stopping criteria are commonly employed to terminate the SA algorithm based on predefined conditions such as a maximum number of iterations or a threshold on solution quality. These criteria aim to balance computational efficiency with the need to obtain acceptable solutions within practical time frames, making SA applicable to a wide range of optimization problems.

	\paragraph{Limitations in Problem-Specific Performance}
	While Simulated Annealing (SA) stands as a robust and versatile optimization technique, its efficacy may encounter challenges when applied to certain problem domains. Particularly in scenarios where the problem landscape offers clear and consistent gradient information, methods like gradient descent or Newton's method may emerge as more expedient alternatives, capable of swiftly converging to optimal solutions with greater precision. This disparity arises from SA's reliance on stochastic exploration, which may exhibit slower convergence rates compared to deterministic methods in gradient-rich environments. Consequently, the suitability of SA hinges heavily upon the nature of the optimization task at hand, necessitating a judicious selection process that weighs the trade-offs between solution quality, computational resources, and convergence speed. Therefore, while SA presents itself as a versatile asset within the optimization toolkit, prudent consideration of problem-specific characteristics remains imperative to ensure optimal algorithmic selection and performance across diverse problem landscapes.

	\subsubsection{Pseudocode for Algorithmic Simulated Annealing}
	Simulated Annealing is a powerful optimization algorithm utilized for finding near-optimal solutions to combinatorial optimization problems. It operates by iteratively exploring the solution space, allowing for occasional uphill moves (accepting solutions that worsen the objective function) to escape local optima. Simulated Annealing's operational procedure is illustrated in pseudocode \ref{fig:simulated-annealing-pseudocode}, which demonstrates its systematic exploration of the solution landscape.
	
	\begin{algorithm}
		\caption{Algorithmic Simulated Annealing Pseudocode}
		\begin{algorithmic}[1]
			\Procedure{SimulatedAnnealing}{}
			\State Initialize temperature, $T$, to a high value
			\State Select an initial solution, $s$, at random or heuristically
			\State Evaluate the energy, $E(s)$, of the initial solution
			\While{termination conditions not met}
			\State Select a neighboring solution, $s'$, of $s$
			\State Evaluate the energy, $E(s')$, of the new solution
			\State Calculate $\Delta E = E(s') - E(s)$
			\If{$\Delta E < 0$ or $\exp(-\Delta E / T) > \text{random}(0, 1)$}
			\State Accept the new solution: $s = s'$
			\EndIf
			\State Update $T$ according to the cooling schedule
			\EndWhile
			\State \Return The best solution found
			\EndProcedure
		\end{algorithmic}\label{fig:simulated-annealing-pseudocode}
	\end{algorithm}

\subsection{Previous Work on ML and AI Interplay with Simulated Annealing}

The integration of machine learning and artificial intelligence into Simulated Annealing (SA) has resulted in the development of Neural Simulated Annealing (NSA), which represents an improvement over traditional SA methods. In 2023, a paper presented at the International Conference on Artificial Intelligence and Statistics introduced NSA, employing deep learning techniques to optimize the proposal distribution mechanisms inherent in SA \cite{correia2023neural}. This approach utilizes neural networks to dynamically learn optimal proposal distributions, contrasting with the static, manually-tuned distributions used in conventional SA. NSA has shown improvements in efficiency and effectiveness, with enhanced convergence rates and the ability to handle more complex optimization challenges. The research suggests the potential of leveraging unsupervised learning to further refine proposal distributions, opening avenues for exploration in optimization algorithms. This advancement demonstrates the impact of AI and machine learning on enhancing traditional optimization frameworks, paving the way for future innovations that may incorporate more advanced AI techniques, including generative models, to enhance the performance of SA.

	\subsection{Algogenic Enhancements for Simulated Annealing}
	\subsubsection{Problem Structure Analysis}
	\paragraph{Enhancing Initial Problem Understanding}
	We suggest initiating Algogenic enhancements for Simulated Annealing with a comprehensive Problem Structure Analysis. This phase leverages generative AI to deepen our understanding of the problem landscape prior to beginning the annealing process. By analyzing the solution space, such AI models can uncover characteristics like potential barriers or dense solution regions, aiding in setting appropriate initial parameters for the annealing process and designing tailored strategies for more efficient navigation through the solution space. This early insight into the problem structure is critical for dynamically adapting exploration and exploitation mechanisms, ensuring a targeted search for optimal solutions. Furthermore, insights from this analysis guide the selection of cooling schedules and neighborhood structures, enhancing the algorithm's ability to avoid local optima and explore diverse solution areas, thus leading to robust and high-quality outcomes.
	
	\paragraph{Optimizing the Solution Space for Efficient Exploration}
	Optimizing the solution space for Simulated Annealing through generative AI involves identifying areas where the solution space can be transformed or preconditioned to enhance SA's effectiveness. Suggesting re-scaling techniques for variables and proposing modifications to the cost function to smooth out steep gradients or sparse solution regions are examples of such optimizations. These preparatory steps not only enhance the SA algorithm's robustness but also accelerate its convergence towards a global optimum by providing an optimized environment, significantly reducing computational resources and time required for satisfactory solutions.
	
	\paragraph{Tailoring the Annealing Process}
	Tailoring the annealing process based on Problem Structure Analysis enables a more informed initialization of the annealing process, including setting an appropriate initial temperature and designing a cooling schedule. Insights from the problem's intricacies help in discerning patterns and dependencies that inform the initialization phase, crucial for the algorithm's performance. This tailored approach promises a more efficient and effective optimization journey, improving convergence rates and solution quality by aligning the process with the problem's specific characteristics.
	
	\subsubsection{Cooling Schedule Optimization}
	\paragraph{Crafting an AI-Informed Cooling Strategy}
	Integrating generative AI for Cooling Schedule Optimization in Simulated Annealing allows for a novel approach where the algorithm dynamically adjusts its cooling schedule based on predictive insights from historical optimization data. This AI-informed strategy enhances the balance between exploration and exploitation, improving the algorithm's convergence speed and solution quality. The adaptive approach enables the algorithm to respond effectively to changes in the problem landscape, leveraging generative AI's predictive power for crafting an effective cooling strategy that guides the algorithm towards more efficient exploration and convergence.
	
	\paragraph{Dynamic Adjustment for Enhanced Efficiency}
	Incorporating dynamic adjustment strategies based on real-time feedback from the optimization process enables the Simulated Annealing algorithm to modify its cooling schedule in response to observed search behaviors. This flexibility ensures that the algorithm remains efficient throughout the search process, enhancing its adaptability to different optimization landscapes and improving convergence speed and solution quality. Furthermore, this approach enhances the algorithm's robustness against noisy environments and fosters a more interactive optimization process.
	
	\paragraph{Implementing a Tailored Cooling Approach}
	Implementing a tailored cooling approach with generative AI significantly advances the optimization methodologies in Simulated Annealing. This approach allows for dynamic adaptation of the cooling schedule in real-time based on evolving problem landscapes and solution trajectories, streamlining the optimization process and enhancing the algorithm's robustness against local minima traps. Moreover, it ensures a balance between exploration and exploitation, enhancing the algorithm's performance and efficiency across diverse problem domains.
	
	\subsubsection{Adaptive Temperature Adjustment}
	\paragraph{Refining Thermal Dynamics for Optimal Search}
	Adaptive Temperature Adjustment leverages generative AI to dynamically adjust the temperature parameter in Simulated Annealing based on the current state and performance of the algorithm. This enhancement improves the efficiency and effectiveness of the algorithm by intelligently adapting the temperature to the search's current state, allowing for more exploration or exploitation as needed. This adaptive approach enhances the algorithm's robustness and performance across different problem domains and instances.
	
	\paragraph{Real-time Temperature Modulation}
	Real-time temperature modulation enables dynamic adjustment of the cooling process based on the current search state, allowing the algorithm to regulate the cooling rate for a balanced exploration and exploitation. This enhancement improves the optimization process's efficiency and increases the likelihood of finding high-quality solutions within a reasonable timeframe by preventing premature convergence and encouraging thorough exploration of the solution space.
	
	\paragraph{Achieving a Harmonized Search Process}
	Implementing Adaptive Temperature Adjustment in Simulated Annealing ensures a balance between exploration and exploitation throughout the optimization process. This dynamic temperature management enhances the search process's efficiency, improving the likelihood of identifying the global optimum. By leveraging generative AI for real-time, adaptive temperature control, the algorithm becomes more versatile and effective for a wide range of optimization tasks.
	
	\subsubsection{Intelligent Move Selection}
	\paragraph{Elevating Solution Exploration with AI Insights}
	Integrating Intelligent Move Selection into Simulated Annealing revolutionizes the optimization process by leveraging generative AI insights for strategic move selection. This approach enables the algorithm to transcend randomness, guiding the exploration towards promising candidate moves based on a deep understanding of the problem's structure and dynamics. Furthermore, this adaptability ensures that the algorithm remains agile and resilient in the face of complex optimization challenges, continuously refining its strategy for more efficient convergence towards optimal solutions.
	
	\paragraph{Strategizing Move Choices for Enhanced Exploration}
	Employing generative AI in move selection enhances exploration by strategically prioritizing moves towards less explored or promising regions of the solution space. This dynamic and context-aware selection process adapts based on the outcomes of previous moves and integrates domain-specific knowledge, leading to more informed and effective move choices. This paradigm shift in exploration strategies enhances the exploration process's efficiency and effectiveness, leading to more robust and insightful solutions.
	
	\paragraph{Optimizing the Search Trajectory}
	Incorporating Intelligent Move Selection into Simulated Annealing transforms the search process into a guided exploration, significantly enhancing efficiency and effectiveness. By strategically focusing on promising areas and avoiding repetitive moves, the algorithm accelerates convergence towards high-quality solutions and effectively navigates complex solution spaces, opening doors to solving increasingly sophisticated optimization problems.
	
	\subsubsection{Solution Space Pruning}
	\paragraph{Focusing the Search on Promising Regions}
	Solution Space Pruning employs generative AI to refine the search area by identifying and excluding regions unlikely to contain the optimum. This targeted approach reduces the computational burden, enhances the scalability of the optimization process, and ensures a more effective allocation of computational resources towards promising solution regions. By dynamically adapting the search strategy based on real-time feedback, this enhancement improves the efficiency, scalability, and adaptability of optimization processes.
	
	\paragraph{Implementing AI-driven Exclusion Criteria}
	Implementing AI-driven exclusion criteria leverages machine learning techniques to identify patterns associated with suboptimal outcomes, streamlining the search process by excluding less promising regions. This strategic pruning of the search space, based on learned patterns and dynamic adaptation, enhances the efficiency of algorithms like Simulated Annealing, focusing computational resources on exploring regions with the highest potential for optimal solutions.
	
	\paragraph{Enhancing Search Efficiency and Outcomes}
	Solution Space Pruning streamlines the optimization process, ensuring that exploratory efforts are directed towards the most promising parts of the solution landscape. This targeted approach accelerates convergence towards optimal solutions, minimizes computational waste, and enhances the robustness of the Simulated Annealing algorithm by avoiding suboptimal or premature solutions, making it a reliable choice for addressing complex optimization challenges.
	
	\subsubsection{Solution Refinement and Analysis}
	\paragraph{Maximizing Solution Quality with LLM Insights}
	Solution Refinement and Analysis leverages Large Language Models to enhance the quality of solutions identified by Simulated Annealing. This process employs LLMs to scrutinize and refine solutions, evaluating them against a diverse range of metrics and facilitating the identification of subtle patterns and emergent properties. Furthermore, the integration of LLM insights stimulates creativity and innovation in the optimization process, enabling the exploration of alternative solution paths and strategies for achieving superior performance or addressing unforeseen constraints.
	
	\paragraph{Employing LLMs for Comprehensive Solution Evaluation}
	Employing LLMs for comprehensive solution evaluation involves validating proposed solutions against additional datasets, simulating solution performance in various scenarios, and leveraging domain-specific knowledge to evaluate practicality. This multifaceted approach ensures the viability and effectiveness of proposed solutions, leading to more robust and impactful outcomes by considering feasibility, effectiveness, and practicality in the evaluation process.
	
	\paragraph{Delivering Optimized, Real-world-ready Solutions}
	The process of Solution Refinement and Analysis ensures that the solutions produced by Simulated Annealing are not only theoretically robust but also practically viable for real-world implementation. Leveraging LLMs within this refinement process enhances the algorithm's capability to produce solutions that are optimal, adaptable, and responsive to dynamic real-world conditions, making Simulated Annealing a powerful tool for complex optimization problems.
	
	\subsubsection{Performance Feedback Loop}
	\paragraph{Cultivating a Self-Improving Algorithmic Ecosystem}
	The Performance Feedback Loop enhances the Simulated Annealing algorithm through continuous learning and refinement, facilitated by generative AI. This dynamic interplay between algorithmic execution and feedback analysis fosters a self-improving ecosystem, optimizing the algorithmic strategies for solution space exploration and intelligent move selection. By leveraging LLMs to analyze outcomes, the algorithm iteratively refines its strategies, enhancing its ability to navigate complex problem domains with precision and efficacy.
	
	\paragraph{Implementing Continuous Learning and Adaptation}
	Implementing continuous learning and adaptation involves refining the LLMs' underlying models powering the algorithm's improvements based on data from each iteration. This process of iteratively updating models with new insights enhances the LLMs' capability to generate contextually relevant text, optimizing generation processes and exploring alternative model architectures for improved scalability, efficiency, and performance.
	
	\paragraph{Achieving Evolutionary Advances in Optimization}
	The integration of a Performance Feedback Loop facilitates evolutionary advances in Simulated Annealing, enabling the algorithm to dynamically adjust its optimization strategies based on real-world performance data. This continuous feedback mechanism allows for the refinement of solution generation processes, enhancing the algorithm's problem-solving capabilities and leading to sophisticated optimization strategies that adapt to complex challenges.
	
	\subsubsection{Contextual Exploration Guidance}
	\paragraph{Navigating the Solution Landscape with AI-Driven Insights}
	Contextual Exploration Guidance integrates language model-based AI to provide tailored recommendations for exploring the solution space, leveraging LLMs to discern patterns and guide the exploration process. This Algogenic enhancement enables the algorithm to adapt dynamically to evolving scenarios, enhancing its ability to discover novel solution paths and ensuring efficient navigation through complex solution landscapes.
	
	\paragraph{Strategic Exploration Based on Contextual Cues}
	Leveraging LLMs for strategic exploration involves synthesizing contextual information to inform move selection, prioritizing exploration towards promising regions and adapting strategies based on learned patterns. This approach enhances exploration efficiency and outcomes by navigating through the solution space with a nuanced understanding of the problem domain, leading to more efficient and effective optimization.
	
	\paragraph{Optimizing Search Efficiency and Outcomes}
	Integrating Contextual Exploration Guidance into Simulated Annealing enhances the search process's efficiency, adaptability, and intelligence. By leveraging LLM-driven insights, the algorithm can navigate the solution space with unprecedented precision and insight, unlocking new frontiers in optimization theory and practice, and paving the way for groundbreaking advancements.
	
	\subsubsection{Semantic Cooling Schedule Design}
	\paragraph{Tailoring the Annealing Process to Problem Semantics}
	Semantic Cooling Schedule Design leverages LLMs to tailor the cooling schedule of Simulated Annealing to the problem's semantics, enhancing performance and efficiency. This approach adapts to diverse problem domains and promotes a more intuitive and interpretable optimization process, aligning the cooling schedule with domain-specific insights for more effective optimization solutions.
	
	\paragraph{Incorporating Domain Knowledge into Cooling Strategies}
	Semantic Cooling Schedule Design involves using LLMs to identify critical variables and adapt cooling schedules based on the problem domain's unique characteristics. This proactive approach enhances the efficiency and efficacy of optimization algorithms, dynamically adjusting the temperature descent to explore fruitful areas and avoid unproductive regions.
	
	\paragraph{Enhancing Optimization with Intuitive Adjustments}
	Incorporating Semantic Cooling Schedule Design transforms Simulated Annealing into a dynamic tool capable of navigating complex search spaces efficiently. By aligning the cooling strategy with semantic cues, the algorithm intelligently prioritizes exploration, accelerating convergence towards high-quality solutions, and enhancing the adaptability and robustness of the optimization process.

	\subsubsection{Pseudocode for Algogenic Simulated Annealing}
	The Algogenic simulated annealing approach leverages AI to augment traditional simulated annealing methods by dynamically adjusting annealing parameters and strategies based on the observed behavior of the system and real-time error estimates. This pseudocode, available in \ref{fig:simulated-annealing-Algogen-pseudocode}, outlines an advanced framework incorporating AI-driven enhancements for adaptive temperature control, neighbor selection, acceptance criteria, and real-time parameter optimization.
	
	\begin{algorithm}
		\caption{Algogenic Simulated Annealing Pseudocode}
		\begin{algorithmic}[1]
			\Procedure{AlgogenicSimulatedAnnealing}{$f, solution, T_{start}, T_{end}, \alpha$}
			\State $currentSolution \gets solution$
			\State $bestSolution \gets solution$
			\State $T \gets T_{start}$ \Comment{Initialize temperature using AI-optimized starting point}
			\While{$T > T_{end}$}
			\State $newSolution \gets \Call{IntelligentMoveSelection}{currentSolution}$ \Comment{AI guides move selection}
			\State $deltaE \gets f(newSolution) - f(currentSolution)$
			\If{$\Call{AcceptanceCriterion}{deltaE, T}$}
			\State $currentSolution \gets newSolution$
			\If{$f(newSolution) < f(bestSolution)$}
			\State $bestSolution \gets newSolution$
			\EndIf
			\EndIf
			\State $T \gets \Call{AdaptiveTemperatureAdjustment}{T, \alpha}$ \Comment{AI adjusts cooling rate}
			\State $\Call{SolutionSpacePruning}{}$ \Comment{AI prunes non-promising regions}
			\EndWhile
			\State $\Call{SolutionRefinementAndAnalysis}{bestSolution}$ \Comment{Refine and validate solution}
			\State $\Call{PerformanceFeedbackLoop}{}$ \Comment{Update AI models for future runs}
			\State \Return $bestSolution$
			\EndProcedure
		\end{algorithmic}\label{fig:simulated-annealing-Algogen-pseudocode}
	\end{algorithm}

	\begin{figure}
		\centering
		\includegraphics[width=0.7\textwidth]{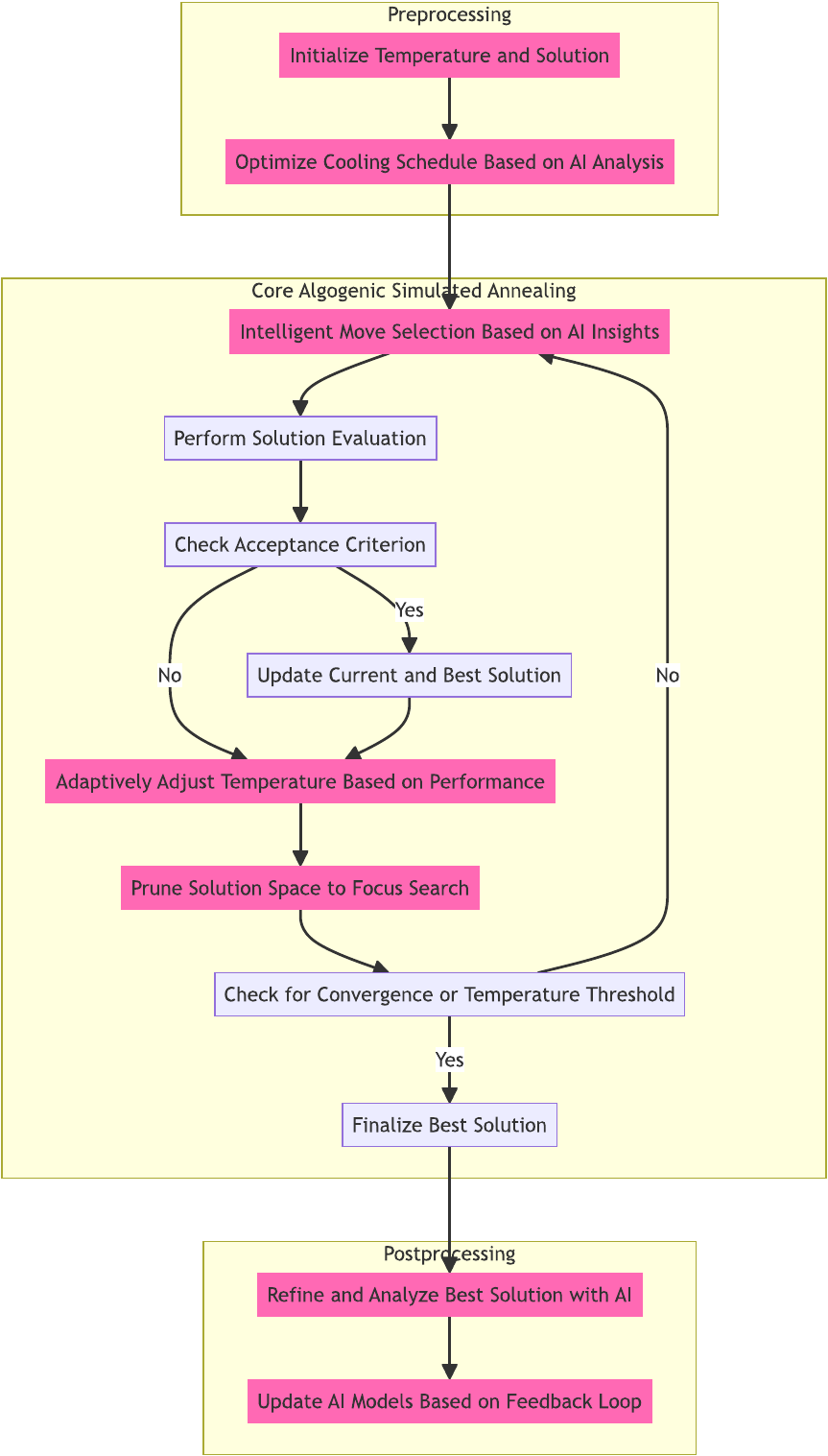} 
		\caption{Integrating Algogenic Enhancements into Simulated Annealing: This diagram visualizes the seamless integration of generative AI enhancements within the Simulated Annealing optimization process. Starting from the preprocessing phase, 'Initialize Temperature and Solution' and 'Optimize Cooling Schedule Based on AI Analysis' set the stage for an AI-optimized exploration of the solution space. The core process unfolds through intelligent move selection, adaptive temperature adjustments, and targeted solution space pruning, all guided by AI insights to navigate the algorithm towards optimal solutions efficiently. 'Adaptive Temperature Adjustment' and 'Intelligent Move Selection Based on AI Insights' ensure the algorithm dynamically responds to the evolving search landscape, enhancing exploration effectiveness. The postprocessing phase, 'Refine and Analyze Best Solution with AI' followed by 'Update AI Models Based on Feedback Loop,' closes the optimization cycle, applying deep learning for final solution refinement and leveraging the performance feedback to inform future runs. This Algogenic approach transforms Simulated Annealing into a dynamic, self-improving algorithm capable of tackling complex optimization challenges with enhanced adaptability, precision, and effectiveness.}
		\label{fig:simulated_annealing}
	\end{figure}

	
	\chapterimage{pngs/statistics.png} 
	\chapter{Statistical Algogens}\index{Statistical Algogens}
	
	\section{Expectation-Maximization (EM)}\index{Expectation-Maximization (EM)}
	\subsection{Introduction to the Expectation-Maximization Algorithm}
	\subsubsection{The Concept of the Expectation-Maximization Algorithm}
	
	\paragraph{Introduction to Expectation-Maximization}
	The Expectation-Maximization (EM) algorithm stands as a pivotal tool within the realm of statistical analysis, revered for its adeptness in handling scenarios characterized by incomplete datasets or the presence of latent variables. At its essence, the EM algorithm endeavors to unveil the most probable parameters of a given probabilistic model, deftly navigating the intricacies posed by missing information. This iterative approach meticulously oscillates between two distinctive phases: the Expectation (E) step and the Maximization (M) step, each assuming a critical role in the estimation of parameters. In the Expectation phase, the algorithm computes the expected value of the latent variables based on current parameter estimates, while in the Maximization phase, it maximizes the likelihood function with respect to the parameters, leveraging the computed expectations from the previous step. This cyclical process continues iteratively until convergence, yielding parameter estimates that optimize the likelihood of the observed data under the model assumptions.

	\paragraph{Mechanics of the E Step}
	During the Expectation step, the algorithm employs the current estimates of the model parameters to infer the missing data's likely values. This step is fundamental for initializing the iterative optimization process, as it sets the foundation for refining the parameter estimates. By leveraging the observed data and the current parameter estimates, the algorithm calculates the conditional expectation of the log-likelihood concerning the latent variables. This expectation, denoted as \(\mathbb{E}\left[\log L(\Theta; X, Z) | X, \Theta^{(t)}\right]\), encapsulates the anticipated log-likelihood of the observed data given the current state of the model. Essentially, the E step acts as a preparatory phase, guiding the subsequent parameter updates in the Maximization step. It establishes a quantitative measure of how well the current parameter estimates align with the observed data, offering insights into the model's fit and informing the optimization trajectory. Consequently, the meticulous execution of the E step is pivotal for the algorithm's convergence and the generation of accurate model predictions.

	\paragraph{Dynamics of the M Step}
	The Maximization (M) step, an integral part of the Expectation-Maximization (EM) algorithm, plays a pivotal role in refining the parameters of the probabilistic model. After the Expectation (E) step, where the algorithm estimates the missing data, the M step takes center stage in iteratively updating the model's parameters (\(\Theta\)). This step is driven by the pursuit of maximizing the Q-function, a critical objective that entails optimizing the parameters to enhance the model's fit to the observed data augmented by the estimated missing values. Symbolically, the update equation in the M step, \(\Theta^{(t+1)} = \arg \max_{\Theta} \mathbb{E}\left[\log L(\Theta; X, Z) | X, \Theta^{(t)}\right]\), encapsulates the essence of parameter refinement. Here, \(\Theta^{(t+1)}\) represents the new estimates of the parameters, derived through the maximization process, based on the current parameter estimates \(\Theta^{(t)}\) and the observed data \(X\) along with the estimated missing data \(Z\). Through this iterative optimization, the M step progressively fine-tunes the model parameters, iteratively improving the model's alignment with the observed data distribution.

	\paragraph{Convergence and Iteration}
	The EM algorithm operates by alternating between the expectation step (E-step) and the maximization step (M-step), continually refining parameter estimates until a convergence criterion is satisfied. This iterative process begins with an initial guess for the parameters and then proceeds to update them iteratively. In the E-step, the algorithm computes the expected values of the unobserved variables, given the current parameter estimates. These expected values are used to estimate the likelihood function. Subsequently, in the M-step, the algorithm maximizes this likelihood function to obtain improved parameter estimates. This iterative cycle continues until the algorithm converges, which is determined by monitoring the change in the log-likelihood function or the parameter values between successive iterations. When the change falls below a predefined threshold, indicating that further iterations are unlikely to significantly improve the parameter estimates, the algorithm terminates, and the final parameter estimates are obtained.
	
	\paragraph{Significance and Application}
	The elegance of the EM algorithm lies in its general applicability and robustness, making it a versatile tool for a wide range of applications in statistical inference, machine learning, and data mining. From clustering and classification to the estimation of complex models like mixture models and hidden Markov models, the EM algorithm facilitates a structured approach to dealing with incomplete data, extracting valuable insights from the shadows of uncertainty and ambiguity that latent variables introduce. Furthermore, its iterative nature allows for refinement and optimization, ensuring convergence to meaningful solutions even in challenging scenarios. Moreover, the EM algorithm's ability to handle missing data gracefully enhances its practical utility in real-world datasets, where data completeness is often compromised. Thus, its widespread adoption in various domains underscores its importance as a fundamental technique for probabilistic modeling and parameter estimation.
	
	The iterative refinement of parameters through the Expectation and Maximization steps embodies a methodical search for clarity and understanding within datasets that conceal their complete stories, showcasing the EM algorithm's pivotal role in the quest for knowledge from incomplete data.

	\subsubsection{Key Principles and Mechanisms}
	
	\paragraph{Foundational Principles of the EM Algorithm}
	At the core of the Expectation-Maximization (EM) algorithm lies the principle of iteratively refining the estimates of a model's parameters, particularly when dealing with hidden or latent variables. This iterative process serves as a powerful tool for statistical inference in scenarios where complete data is not available. The EM algorithm operates under the assumption that a complete understanding of the data generation process requires knowledge of both observed and unobserved variables. In the Expectation (E) step, the algorithm calculates the expected values of the latent variables given the observed data and current parameter estimates. This step essentially computes the missing information, allowing for a more comprehensive assessment of the underlying data structure. Conversely, in the Maximization (M) step, the algorithm updates the model parameters to maximize the likelihood of the observed data given the computed expected values of the latent variables. By iteratively alternating between these two steps, the EM algorithm converges towards a local maximum of the likelihood function, effectively refining the parameter estimates and revealing hidden patterns within the data. Through this iterative refinement process, the EM algorithm facilitates a deeper understanding of complex data distributions and enhances the accuracy of statistical models. Moreover, the EM algorithm's flexibility in handling missing or incomplete data makes it a valuable tool in various fields such as machine learning, signal processing, and bioinformatics.

	\paragraph{The Iterative Process}
	The journey of the EM algorithm commences with initial guesses or estimates of the model parameters, which may be derived from prior knowledge, random initialization, or heuristic methods. These initial estimates serve as the starting point for the iterative process, where the algorithm oscillates between the E and M steps. In the E step, the algorithm computes the expected value of the log-likelihood function, considering the current parameter estimates and incorporating the estimated distribution of the latent variables. This step essentially constructs a bridge over the gaps in the data, enabling the M step to proceed on firmer ground. Furthermore, in the M step, the algorithm updates the model parameters by maximizing the log-likelihood function based on the data and the expected values obtained in the E step. Consequently, this iterative cycle continues until convergence, where the parameter estimates stabilize, indicating that the algorithm has found a local maximum of the likelihood function. Therefore, through this iterative process, the EM algorithm iteratively refines its estimates of the model parameters, gradually converging towards optimal values that best capture the underlying data distribution.

	\paragraph{Optimization in the M Step}
	The M step, following the completion of the estimation phase, delves into an intricate process of optimization. This phase is pivotal as it endeavors to fine-tune the model parameters, steering them towards optimizing the likelihood function established during the preceding E step. This optimization pursuit is akin to a meticulous calibration, where the model parameters undergo adjustments to better align with the observed data. Each iteration of parameter updates is driven by a resolute commitment to enhancing the model's fidelity to the underlying data generation process. Through this iterative refinement, the model progressively converges towards a configuration that maximizes the likelihood of generating the observed data. This iterative process is not merely a mechanical adjustment but rather a strategic maneuver aimed at incrementally refining the model's representation of the latent variables and their interdependencies. Consequently, the M step serves as a crucial bridge between the initial estimation phase and the eventual convergence towards a more accurate and reliable model representation.

	\paragraph{Convergence Criterion}
	A pivotal aspect of the EM algorithm's iterative process is the convergence criterion, which serves as a guiding principle for determining when to terminate the algorithm's iterations. This criterion acts as a safeguard against unnecessary computational overhead by specifying conditions under which further iterations are deemed unnecessary. Typically, convergence is declared when successive iterations produce only marginal improvements in the likelihood function or when the changes in parameter estimates become negligible. This decision ensures that the algorithm halts its iterations when additional adjustments to the parameters are unlikely to yield substantial enhancements in likelihood estimation, thereby conserving computational resources. By adhering to the convergence criterion, the algorithm strikes a balance between computational efficiency and optimization accuracy, ensuring that the final parameter estimates sufficiently represent the underlying data distribution without unnecessary computational burden.

	\paragraph{Mechanisms Ensuring Robust Estimation}
	The EM algorithm's mechanisms ensure robust estimation in dealing with challenges posed by latent variables and incomplete data. Iteratively estimating missing data and optimizing parameter estimates, the EM algorithm navigates uncertainties inherent in such datasets, providing a robust framework for statistical inference. Its application spans various domains, offering a systematic approach to unraveling the complexities of data and extracting meaningful insights from incomplete information. Reliance on the principles of expectation and maximization ensures that each iteration contributes to a more accurate and comprehensive understanding of the data's underlying structure. Furthermore, the EM algorithm's iterative nature allows it to adapt and refine estimates over successive iterations, converging towards optimal solutions even in the presence of complex data distributions. Additionally, the algorithm's versatility enables its integration into various statistical models, enhancing their capability to handle missing data effectively. Through these mechanisms, the EM algorithm not only addresses the challenges of latent variables and incomplete data but also facilitates robust estimation in diverse statistical applications.

	Through its iterative refinement of parameters, the EM algorithm demonstrates a powerful principle of statistical learning: that even in the face of incomplete information, methodical and iterative approaches can lead to the discovery of underlying truths hidden within the data.

	\subsubsection{The Role of Latent Variables}
	
	\paragraph{Defining Latent Variables}
	Latent variables, often referred to as hidden variables, play a crucial role in statistical modeling and data analysis. They represent underlying factors that are not directly observable but have a significant impact on the observed data. These variables are fundamental in various machine learning algorithms, including the Expectation-Maximization (EM) algorithm, where they serve as essential components for uncovering hidden structures within the data. In the realm of EM, latent variables act as bridges between the observed data and the model parameters, encapsulating the unobservable information necessary for accurately modeling the data distribution. For example, in clustering tasks, latent variables can represent the cluster assignments of data points, guiding the algorithm in grouping similar observations together. Moreover, in the context of probabilistic graphical models, latent variables enable the representation of complex dependencies among observed variables, allowing for more nuanced and accurate modeling of real-world phenomena. Thus, understanding and appropriately defining latent variables are critical steps in designing effective machine learning models and algorithms for various applications.

	\paragraph{Latent Variables in the EM Framework}
	Within the EM algorithm, latent variables serve a multifaceted role, extending beyond mere placeholders for missing data. They act as conduits that link the observed data to the underlying mechanisms governing the data generation process. In essence, latent variables represent the hidden dimensions of the data, encapsulating unobservable factors that influence the observed outcomes. During the Expectation (E) step of the EM algorithm, these latent variables are inferred based on the available observed data and the current parameter estimates. This inference process involves estimating the distribution and characteristics of the latent variables, given the observed data and the model parameters. Subsequently, during the Maximization (M) step, the algorithm updates the model parameters based on these inferred latent variables to maximize the likelihood of the observed data. This iterative interplay between estimating latent variables and updating model parameters lies at the heart of the EM algorithm's efficacy in handling incomplete or partially observed datasets. By leveraging latent variables, the EM algorithm can uncover hidden patterns and structures in the data, facilitating robust parameter estimation and model fitting even in the presence of missing or unobservable information.

	\paragraph{Inference and Estimation of Latent Variables}
	The EM algorithm's efficacy in inferring the characteristics of latent variables underscores its versatility and applicability across diverse domains. Through its iterative nature, the EM algorithm navigates through the complex landscape of hidden variables, gradually unraveling their underlying traits. This iterative estimation process operates within the framework of probabilistic modeling, wherein each iteration refines the approximations of latent variables based on the most recent parameter updates. Consequently, the model's fidelity to the observed data progressively improves, elucidating the latent structures inherent in the dataset. This nuanced understanding of latent variables empowers practitioners to glean deeper insights into the intricacies of the phenomena under study. Moreover, the EM algorithm's probabilistic foundations instill confidence in the inferred latent variables, facilitating robust decision-making in the face of uncertainty.

	\paragraph{Applications Leveraging Latent Variables}
	The integration of latent variables into the EM algorithm's framework significantly expands its range of applications, rendering it a versatile tool for statistical analysis and machine learning endeavors. These latent variables play a pivotal role across various domains, facilitating sophisticated modeling techniques and insights extraction from complex datasets. For instance, in unsupervised learning tasks such as clustering, latent variables are employed to represent cluster memberships, allowing the algorithm to identify underlying patterns and groupings within the data autonomously. Additionally, in more intricate models like hidden Markov models (HMMs) utilized in sequence analysis, latent variables serve to encode hidden states, enabling the algorithm to infer underlying temporal structures and dependencies. Such adaptability enables the EM algorithm to address a diverse array of problems, ranging from pattern recognition to anomaly detection, with robustness and efficacy.

	\paragraph{Unveiling Hidden Insights}
	The EM algorithm operates on the principle of uncovering latent variables, thereby delving into the depths of data beyond its surface manifestations. This methodology acknowledges that the observable attributes of data often represent only a fraction of its true complexity, with underlying factors exerting significant influence. By iteratively estimating these latent variables and updating model parameters, EM untangles the intricate web of interactions within the data, revealing patterns and structures that may otherwise remain obscured. This process resembles peering through a multifaceted prism, where each iteration brings a clearer understanding of the underlying phenomena. Consequently, EM transcends conventional statistical techniques by elucidating the hidden dynamics driving observed behaviors, enabling analysts to grasp the underlying mechanisms shaping the data landscape. Moreover, the EM algorithm's ability to seamlessly integrate observable and latent variables facilitates a holistic perspective, empowering researchers to extract nuanced insights and formulate informed hypotheses about the underlying phenomena. Through this lens, EM emerges as a powerful tool not only for parameter estimation but also for illuminating the intricate interplay between observable phenomena and their underlying causes.

	\subsubsection{Applications and Limitations}
	
	\paragraph{Diverse Applications of the EM Algorithm}
	The Expectation-Maximization (EM) algorithm finds extensive application across a broad spectrum of fields, underscoring its versatility and effectiveness in dealing with incomplete data and latent variables. In clustering, particularly with Gaussian Mixture Models (GMMs), the EM algorithm is instrumental in identifying underlying groups in the data, where latent variables represent cluster memberships. Image analysis benefits from the EM algorithm through techniques like image segmentation, where it helps in modeling the distribution of pixels into different segments. Natural language processing (NLP) applications, such as topic modeling, leverage the EM algorithm for discovering latent topics within large collections of text documents, significantly aiding in the organization, understanding, and summarization of vast textual information. Furthermore, the EM algorithm plays a crucial role in parameter estimation for hidden Markov models (HMMs), contributing to advancements in speech recognition, bioinformatics, and finance. Additionally, in the domain of signal processing, the EM algorithm facilitates signal decomposition and denoising, leading to improved signal reconstruction and analysis. Moreover, the EM algorithm is employed in neuroscience for decoding neural signals and uncovering underlying patterns in brain activity, fostering advancements in cognitive science and brain-computer interfaces. Hence, the diverse applications of the EM algorithm underscore its importance and utility across various domains, driving innovation and progress in research and industry.

	\paragraph{Sensitivity to Initial Parameter Estimates}
	One notable limitation of the EM algorithm is its sensitivity to the initial parameter estimates. While EM is powerful in estimating model parameters from incomplete data, its convergence heavily relies on the initial guesses of these parameters. If the initial estimates are far from the true values, the algorithm may converge to local optima instead of the global maximum likelihood solution. This sensitivity underscores the importance of meticulously selecting initial parameter values, which often involves leveraging domain expertise or employing heuristic approaches. Moreover, the algorithm's performance can be enhanced by conducting preliminary analyses to gain insights into the data distribution and refine the initial estimates accordingly. Despite its effectiveness in handling missing data, the EM algorithm's susceptibility to initialization underscores the need for caution and thorough exploration of parameter space to mitigate the risk of convergence to suboptimal solutions.

	\paragraph{Convergence to Local Maxima}
	The EM algorithm's tendency to converge to local maxima rather than global ones presents a significant challenge in optimization. This behavior arises from its iterative nature, wherein it gradually improves parameter estimates based on the current likelihood landscape. Particularly in intricate models housing multiple maxima, the algorithm may become trapped in a local maximum that does not correspond to the optimal solution globally. This inherent limitation underscores the necessity for employing diverse strategies to mitigate it effectively. One approach involves executing the algorithm multiple times with varied initial parameters to explore a broader solution space. Additionally, techniques like simulated annealing can be leveraged to escape local maxima by incorporating probabilistic acceptance of less optimal solutions, thereby facilitating exploration of alternative regions of the parameter space. Through these concerted efforts, practitioners can enhance the algorithm's robustness and increase the likelihood of converging to the global maximum.

	\paragraph{Strategies to Mitigate Limitations}
	To mitigate these limitations, several strategies can be employed. One approach is to utilize multiple initializations of the algorithm from various starting points. By doing so, the algorithm can explore the parameter space more thoroughly, reducing the risk of convergence to suboptimal local maxima. Incorporating domain-specific knowledge into the initialization process can provide a more guided and potentially more accurate starting point for parameter estimation. This integration of domain expertise ensures that the algorithm's optimization trajectory aligns more closely with the underlying structure of the data, thus enhancing its performance. Furthermore, advanced optimization techniques can be leveraged to refine the estimation process. Modifications to the EM algorithm, such as stochastic EM or variational EM methods, offer alternative paths to enhance the algorithm's robustness and convergence properties. These techniques introduce randomness or approximate inference methods, enabling the algorithm to escape local optima and achieve more stable convergence. By combining these strategies, researchers can effectively address the limitations of the algorithm and improve its overall performance in practical applications.

	\paragraph{The Balance of Utility and Challenges}
	Despite its limitations, the EM algorithm remains a cornerstone method in statistical analysis and machine learning, valued for its ability to extract meaningful insights from incomplete data. The algorithm's iterative nature allows it to converge towards maximum likelihood estimates even in scenarios with missing or latent variables. However, navigating its complexities requires a nuanced understanding of its underlying assumptions and constraints. For instance, the EM algorithm's reliance on local maxima can lead to suboptimal solutions, necessitating careful initialization and convergence criteria selection. Moreover, its sensitivity to initial parameter values underscores the importance of robust initialization strategies to avoid convergence to undesirable solutions. Nevertheless, when applied judiciously, the EM algorithm offers unparalleled utility in various domains, from clustering and classification to density estimation and latent variable modeling. By acknowledging its challenges and adopting appropriate mitigation strategies, researchers can effectively leverage the EM algorithm's power while minimizing its limitations.

	\subsubsection{Pseudocode for Algorithmic EM}
	
	The Expectation Maximization (EM) Algorithm is a sophisticated framework designed for efficiently estimating parameters in statistical models, particularly when dealing with latent variables. It distinguishes itself by iteratively maximizing the likelihood function, incorporating both observed data and latent variables to refine parameter estimates. The operational essence of EM is encapsulated in pseudocode \ref{fig:em-pseudocode}, illustrating its iterative approach to parameter estimation.
	
	\begin{algorithm}
		\caption{Pseudocode for the Expectation-Maximization Algorithm}
		\begin{algorithmic}[1]
			\Procedure{ExpectationMaximization}{Data, InitialParameters}
			\State \textbf{initialize} parameter estimates $\Theta^{(0)}$ with InitialParameters
			\State \textbf{set} iteration counter $t \gets 0$
			\While{not converged}
			\State \textbf{// E-Step: Estimate missing data given current parameters}
			\For{each data point $x_i$ in Data}
			\State Estimate $E[Z|x_i, \Theta^{(t)}]$, the expected value of latent variables $Z$
			\EndFor
			\State \textbf{// M-Step: Maximize expected log-likelihood w.r.t. $\Theta$}
			\State $\Theta^{(t+1)} \gets \arg\max_{\Theta} \sum_i \log p(x_i, E[Z|x_i, \Theta^{(t)}]; \Theta)$
			\If{convergence criterion is met}
			\State \textbf{break}
			\EndIf
			\State $t \gets t + 1$
			\EndWhile
			\State \Return $\Theta^{(t+1)}$ \textbf{as} FinalParameterEstimates
			\EndProcedure
		\end{algorithmic}\label{fig:em-pseudocode}
	\end{algorithm}

\subsection{Previous Work on ML and AI Interplay with the Expectation Maximization Algorithm}

\paragraph{Federated Learning Perspective}
The Expectation Maximization (EM) algorithm has been applied within federated learning frameworks, as discussed in \cite{louizos2021expectation}. This work introduces a perspective that integrates the EM algorithm with federated learning, addressing challenges related to data privacy and security in distributed data environments. By incorporating sparsity-inducing priors and variational inference techniques, the study aims to enhance model training efficiency across decentralized datasets, providing insights into scalable and privacy-preserving machine learning models.

\paragraph{Semi-supervised Learning Enhancement}
In the realm of semi-supervised learning, \cite{sula2022semi} presents advancements in semi-supervised Expectation Maximization. The framework combines labeled and unlabeled data to improve learning outcomes, demonstrating enhanced convergence rates and model performance. The study underscores the algorithm's adaptability to semi-supervised settings and offers insights into its behavior across different data regimes.

\paragraph{Big Learning Approach}
The concept of 'Big Learning' in EM, as proposed in \cite{cong2023big}, aims to address limitations of traditional EM algorithms, particularly in escaping local optima. This work employs a multi-stage strategy, including joint, marginal, and conditional matching, to enhance convergence properties and applicability to complex mixture models. Integration of orthogonal transformations and dimensionality reduction techniques further extends the algorithm's utility in large-scale learning tasks.

\paragraph{Unsupervised Clustering Driven by Supervised Learning}
The intersection of unsupervised clustering and supervised learning through the EM algorithm is explored in \cite{louiset2021ucsl}. This framework synergizes unsupervised clustering with supervised learning objectives to enhance the discovery of latent subtypes in datasets, particularly in domains with limited labeled data. The study reflects a trend of blending learning paradigms for improved model performance across diverse contexts.

\paragraph{Deep Learning of Semi-Competing Risk Data}
\cite{salerno2022deep} discusses the application of the EM algorithm in semi-competing risks data. The proposed neural EM algorithm extends traditional EM frameworks to address complexities in semi-competing risks models, offering a methodology for estimating baseline hazards and risk functions. Integration of deep learning enhances predictive capabilities and expands applicability to biomedical and survival analysis tasks, showcasing the symbiotic relationship between machine learning advancements and statistical algorithms.

	\subsection{Algogenic Enhancements for Expectation Maximization}
	\subsubsection{Data Completeness Analysis}
	\paragraph{Enhancing Data Preparation with AI Insights}
	Prior to the application of the Expectation Maximization algorithm, conducting a comprehensive Data Completeness Analysis using generative AI significantly enhances the understanding of dataset completeness. This analysis, crucial for determining the presence and nature of missing data, influences the EM algorithm's efficacy. By leveraging Large Language Models, this step surpasses traditional statistical measures, enabling a deeper exploration of complex data relationships and the prediction of missing data's impact on model performance. Additionally, this AI-driven approach identifies underlying missing data patterns and guides the selection of appropriate handling methods, simulates missing data effects, and uncovers hidden data dependencies, thus optimizing the EM algorithm's performance through informed data preparation strategies.
	
	\paragraph{Strategizing Data Imputation and Algorithm Adaptation}
	Insights from Data Completeness Analysis via LLMs guide the development of tailored data imputation strategies and EM algorithm adaptations to address dataset-specific challenges. This involves recommending imputation methods that maintain the underlying data distribution and modifying the EM algorithm to incorporate missing data uncertainty, thereby enhancing robustness and reliability in the analysis.
	
	\paragraph{Setting the Stage for Optimized EM Processing}
	Effective preparation of the dataset lays the groundwork for optimized Expectation Maximization processing, ensuring data quality and completeness. This phase, crucial for algorithm performance, involves detailed data analysis and preprocessing, guided by Algogenic enhancements, to identify and address anomalies, biases, or missing data, thus setting a solid foundation for the EM algorithm's application.
	
	\subsubsection{Model Structure Optimization}
	\paragraph{Tailoring Model Architecture with AI}
	The optimization of the model structure, using generative AI insights, ensures the EM algorithm is well-suited to the specific data and problem context. This involves analyzing preliminary outcomes and adjusting the model accordingly, leveraging AI's capability for dynamic adaptation and exploration of novel architectures, thereby enhancing computational efficiency and model accuracy.
	
	\paragraph{Optimizing for Computational Efficiency and Accuracy}
	Model Structure Optimization focuses on enhancing computational efficiency and accuracy through model architecture analysis and parameter adjustment. This process, informed by domain expertise, seeks to balance model complexity with performance, ensuring scalable and interpretable models that perform effectively across various datasets.
	
	\paragraph{Enhancing Model Performance and Interpretability}
	Strategic model structuring, incorporating generative AI insights, not only improves the EM algorithm's performance but also its interpretability. This approach ensures the model accurately captures the underlying data distribution, facilitating a deeper understanding of data generation mechanisms and enabling more informed decision-making.
	
	\subsubsection{Dynamic Parameter Initialization}
	\paragraph{Leveraging AI for Strategic Parameter Selection}
	Implementing Dynamic Parameter Initialization with generative AI significantly enhances the EM algorithm by selecting optimal initial parameter values. This approach utilizes AI to analyze data patterns and optimize parameter selection, improving algorithm effectiveness and efficiency.
	
	\paragraph{Adapting to Dataset Specifics and Historical Insights}
	Generative AI facilitates adaptive parameter initialization, considering dataset specifics and historical data insights. This customization ensures a tailored approach to each dataset, enhancing model performance and efficacy through informed parameter selection strategies.
	
	\paragraph{Optimizing the Path to Convergence}
	Dynamic Parameter Initialization accelerates the EM algorithm's convergence and increases its robustness against local optima by strategically selecting initial parameters. This Algogenic enhancement leverages AI insights for a more efficient and effective optimization process.
	
	\subsubsection{Adaptive Step Size Adjustment}
	\paragraph{Fine-Tuning the Maximization Phase with AI}
	Adaptive Step Size Adjustment, guided by generative AI, dynamically adjusts the EM algorithm's step size, enhancing parameter update efficiency and reliability. This Algogenic enhancement improves the optimization process, leading to faster convergence and better model performance.
	
	\paragraph{Balancing Exploration and Convergence}
	This technique optimizes the EM algorithm by balancing exploration and convergence, dynamically adjusting step sizes based on the optimization stage and insights from generative AI, facilitating a more effective search for the optimal solution.
	
	\paragraph{Enhancing Efficiency and Accuracy of Parameter Estimation}
	Incorporating Adaptive Step Size Adjustment into the EM algorithm enhances its efficiency and accuracy, optimizing parameter updates to mitigate common optimization issues and adapt to changing data distributions, thus improving model performance and reliability.
	
	\subsubsection{Intelligent Stopping Criterion}
	\paragraph{Enhancing Termination Decisions with AI}
	The Intelligent Stopping Criterion leverages generative AI to improve the EM algorithm's termination decisions, dynamically evaluating progress and incorporating a broader range of metrics for a more effective convergence assessment and model performance evaluation.
	
	\paragraph{Context-Aware Termination Strategies}
	Generative AI enables context-aware termination strategies in the EM algorithm, adjusting stopping criteria based on iterative progress, historical data insights, and real-time analysis, enhancing the efficiency and accuracy of the optimization process.
	
	\paragraph{Optimizing Model Performance and Computational Efficiency}
	Intelligent Stopping Criterion ensures optimal performance and efficiency in the EM algorithm, dynamically adjusting termination criteria based on AI-driven insights, enhancing model accuracy and reducing computational overhead.
	
	\subsubsection{Solution Validation and Refinement}
	\paragraph{Ensuring Optimal Solutions through AI Analysis}
	Solution Validation and Refinement uses generative AI to assess and enhance EM algorithm solutions, ensuring statistical soundness and practical viability through comprehensive analysis and domain-specific insights, thereby optimizing model performance and applicability.
	
	\paragraph{AI-Driven Refinement for Real-World Applications}
	LLMS-driven refinement enhances model adaptability and performance in real-world scenarios, incorporating external data, simulations, and domain knowledge to improve model accuracy and interpretability, ensuring operational feasibility and effectiveness.
	
	\paragraph{Maximizing Solution Utility and Impact}
	Integrating Solution Validation and Refinement with the EM algorithm ensures solutions are practically applicable and optimized for impact, enhancing the robustness and reliability of the model for real-world decision-making and problem-solving.
	
	\subsubsection{Model Interpretability Enhancement}
	\paragraph{Translating Complex Models into Understandable Insights}
	Model Interpretability Enhancement uses LLMs to make the EM algorithm's outcomes more accessible and comprehensible, translating statistical complexities into clear insights, thereby democratizing access to advanced analytical techniques and empowering informed decision-making.
	
	\paragraph{Facilitating Broader Understanding and Application}
	LLMS enhances the understanding and application of model outputs, translating complex findings into accessible insights, fostering trust and facilitating informed decision-making, thereby enabling a broader application of data-driven insights in decision-making processes.
	
	\paragraph{Enhancing Decision-Making with AI-Augmented Explanations}
	Integrating Model Interpretability Enhancement with the EM algorithm enhances decision-making by making model outputs and their implications more understandable, fostering transparency and collaboration among stakeholders, and driving informed decisions across various domains.
	
	\subsubsection{Semantic Analysis of Convergence}
	\paragraph{Deepening Understanding of Optimization Dynamics}
	Semantic Analysis of Convergence leverages LLMS to provide insights into the EM algorithm's convergence process, offering a deeper understanding of optimization dynamics and facilitating the development of strategies to enhance algorithm performance and efficiency.
	
	\paragraph{Identifying Patterns and Providing Actionable Insights}
	LLMS identifies optimization patterns and provides actionable insights for refining the EM algorithm, enhancing its efficiency and effectiveness by guiding model adjustments and optimization strategies based on comprehensive data analysis.
	
	\paragraph{Enhancing Algorithmic Efficiency and Effectiveness}
	Incorporating Semantic Analysis of Convergence into the EM algorithm improves its self-awareness and adaptability, streamlining the optimization process and leading to more accurate and reliable model estimates, enhancing statistical optimization techniques.
	
	\subsubsection{Predictive Model Selection}
	\paragraph{Optimizing Model Choice Pre-Optimization}
	Predictive Model Selection utilizes LLMs to optimize model selection before the EM algorithm's optimization process begins, analyzing the dataset to predict the most suitable model structures, saving computational resources and improving efficiency.
	
	\paragraph{Incorporating Domain Knowledge and Data Insights}
	This process integrates domain-specific knowledge and data insights, leveraging LLMS to inform model selection, ensuring the chosen model aligns with the data characteristics and enhances model optimization and performance.
	
	\paragraph{Streamlining the Path to Optimal Solutions}
	Employing Predictive Model Selection positions the EM algorithm for success, focusing on efficient and theoretically sound model structures, enhancing the optimization process's targeted effectiveness and leading to optimized solutions with greater precision.

	\subsubsection{Pseudocode for Algogenic Expectation Maximization}
	The Algogenic expectation maximization approach integrates AI to enhance conventional expectation maximization techniques by dynamically adjusting optimization parameters and strategies according to the observed behavior of the system and real-time error estimates. This pseudocode, available in \ref{fig:expectation-maximization-Algogen-pseudocode}, outlines a sophisticated framework incorporating AI-driven improvements for adaptive parameter adjustment, expectation maximization steps, convergence criteria, and real-time optimization of parameters.
	
	\begin{algorithm}
		\caption{Algogenic Expectation Maximization Pseudocode}
		\begin{algorithmic}[1]
			\Procedure{AlgogenicEM}{$data, model$}
			\State $parameters \gets \Call{DynamicParameterInitialization}{data}$ \Comment{AI optimizes starting parameters}
			\State $model \gets \Call{PredictiveModelSelection}{data}$ \Comment{AI selects optimal model structure}
			\While{not \Call{IntelligentStoppingCriterion}{parameters}}
			\State $EStep \gets \Call{Expectation}{data, parameters}$ \Comment{E-step with current parameters}
			\State $MStep \gets \Call{Maximization}{data, EStep}$ \Comment{M-step, AI adjusts step size}
			\State $parameters \gets \Call{AdaptiveStepSizeAdjustment}{MStep}$ \Comment{Dynamically adjust parameters}
			\State $\Call{SemanticAnalysisOfConvergence}{parameters}$ \Comment{AI analyzes convergence behavior}
			\EndWhile
			\State $validatedParameters \gets \Call{SolutionValidationAndRefinement}{parameters}$ \Comment{Refine solution}
			\State $\Call{ModelInterpretabilityEnhancement}{validatedParameters}$ \Comment{Enhance model interpretability}
			\State \Return $validatedParameters$
			\EndProcedure
		\end{algorithmic}\label{fig:expectation-maximization-Algogen-pseudocode}
	\end{algorithm}
	
	\begin{figure}
		\centering
		\includegraphics[width=0.7\textwidth]{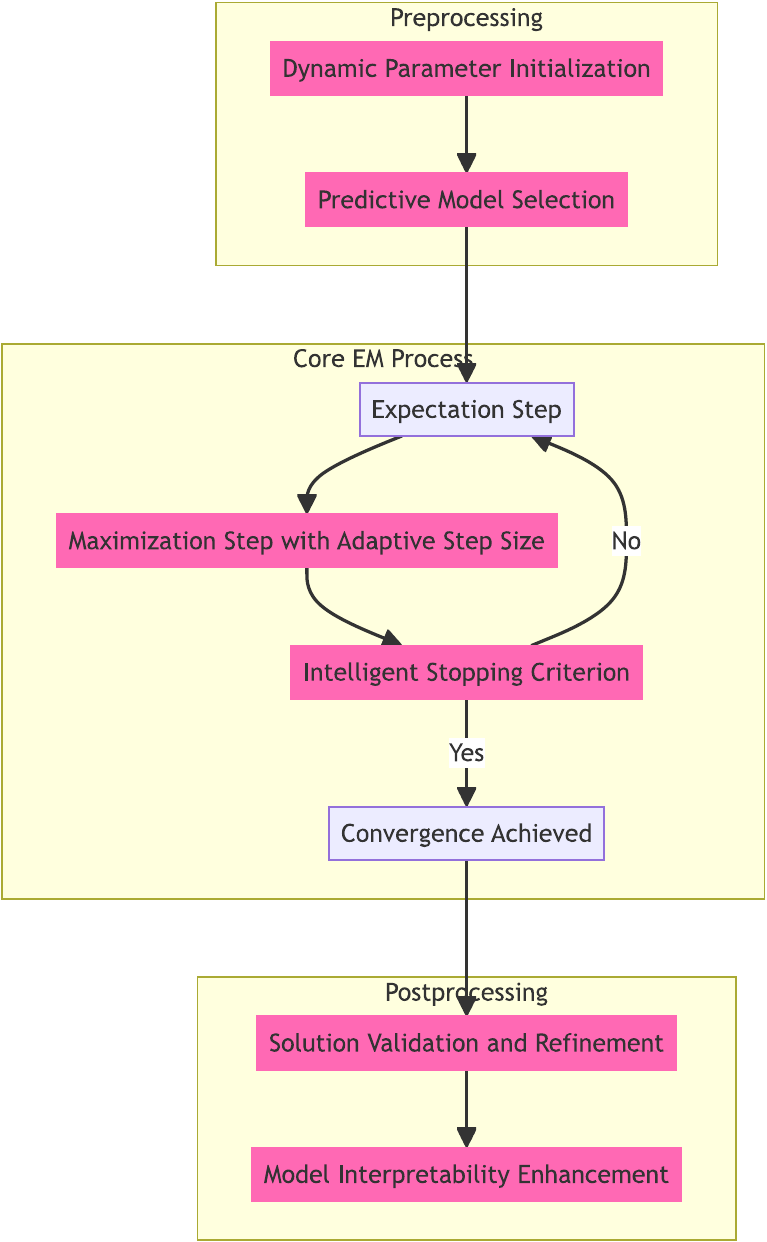} 
		\caption{Integrating Algogenic Enhancements into Expectation Maximization: This diagram visualizes the seamless integration of generative AI enhancements within the Expectation Maximization (EM) algorithm process. Starting with the preprocessing phase, 'Dynamic Parameter Initialization' and 'Predictive Model Selection' utilize AI to set the foundation for a more informed and optimized EM process. The core process illustrates the iterative nature of EM, enhanced with 'Adaptive Step Size' adjustments and an 'Intelligent Stopping Criterion'—both AI-driven—to ensure efficient and effective convergence. The postprocessing phase, including 'Solution Validation and Refinement' and 'Model Interpretability Enhancement,' highlights the role of AI in ensuring that the solutions are not only statistically robust but also practically applicable and easily interpretable. This Algogenic approach transforms the EM algorithm into a dynamic, self-improving framework capable of tackling complex statistical challenges with enhanced adaptability and precision.}
		\label{fig:expectation_maximization}
	\end{figure}

	\section{Bayesian Inference}\index{Bayesian Inference}
	\subsection{Introduction to Bayesian Inference}
	\subsubsection{The Concept of Bayesian Inference}
	
	\paragraph{Introduction to Bayesian Inference}
	Bayesian Inference stands as a cornerstone within the realm of statistical analysis, distinguished by its foundational approach to integrating probability with statistical problems. At its core, Bayesian Inference harnesses the power of Bayes' theorem, a seminal principle that guides the updating of probabilities for hypotheses in light of new evidence or information. This methodological framework is not merely a statistical technique; it is a philosophical approach to understanding uncertainty, predicated on the concept of belief updating. As new data becomes available, Bayesian Inference provides a structured mechanism for revising and refining these beliefs, encapsulating the dynamic and iterative nature of learning from data. Moreover, it allows for a coherent incorporation of prior knowledge or assumptions into the analysis, enhancing the interpretability and robustness of the results. Additionally, Bayesian methods offer a flexible framework for modeling complex phenomena, accommodating various types of data and facilitating the integration of multiple sources of evidence. Consequently, Bayesian Inference serves as a versatile and powerful tool for addressing a wide range of analytical challenges, from parameter estimation to hypothesis testing, decision making, and predictive modeling.

	\paragraph{Bayes' Theorem at the Heart}
	Central to Bayesian Inference is Bayes' theorem, elegantly capturing the essence of probabilistic reasoning. Mathematically expressed as
	\[
	P(H|E) = \frac{P(E|H)P(H)}{P(E)},
	\]
	where $P(H|E)$ represents the posterior probability of the hypothesis $H$ given evidence $E$, $P(E|H)$ denotes the likelihood of observing $E$ if $H$ is true, $P(H)$ is the prior probability of $H$, and $P(E)$ is the probability of observing the evidence. This theorem lays the groundwork for updating our confidence in hypotheses as we encounter new data, embodying the iterative process of refining our understanding of the world.
	
	Moreover, Bayes' theorem serves as a fundamental bridge between prior knowledge and new evidence, allowing us to systematically incorporate data to revise our beliefs. It enables a principled approach to decision-making under uncertainty, providing a framework for rational inference and updating of probabilities. Additionally, Bayes' theorem finds wide application across various fields, including machine learning, statistics, and artificial intelligence, where uncertainty quantification and inference are paramount. Furthermore, its elegant formulation underscores the power of probabilistic thinking in modeling real-world phenomena and making informed decisions. Hence, Bayes' theorem stands as a cornerstone of Bayesian methodology, facilitating a coherent and flexible framework for reasoning in the face of uncertainty.
	
	\paragraph{Dynamic Approach to Statistical Inference}
	Bayesian Inference distinguishes itself through its dynamic approach to statistical inference. Unlike traditional methods that might provide static analyses, Bayesian Inference thrives on adaptability, accommodating new information as it becomes available. This adaptability ensures that conclusions drawn from Bayesian analyses are not fixed; they evolve as more data is gathered, offering a more nuanced and comprehensive understanding of the phenomena under study. Furthermore, this flexibility allows Bayesian models to incorporate prior knowledge effectively, enhancing the robustness and accuracy of the inference process. Moreover, the iterative nature of Bayesian updating fosters a continuous refinement of estimates, enabling researchers to refine hypotheses and predictions over time. Consequently, Bayesian Inference stands as a powerful framework for addressing complex and evolving phenomena, providing insights that can inform decision-making processes across various fields, from healthcare to finance. Hence, its dynamic nature positions Bayesian Inference as a cornerstone in modern statistical practice, offering unparalleled flexibility and insight into the ever-changing landscape of data analysis.
	
	\paragraph{Philosophical Underpinnings}
	Beyond its mathematical formulation, Bayesian Inference is deeply philosophical, rooted in the concept of subjective probability. This perspective views probabilities as expressions of personal belief about the likelihood of events, rather than frequencies or propensities. Such a viewpoint allows for the incorporation of prior knowledge and expert judgment into the analysis, making Bayesian methods particularly valuable in fields where data is scarce or uncertain, and prior expertise is rich. Moreover, this philosophical foundation highlights the importance of acknowledging and quantifying uncertainty in decision-making processes. Furthermore, by embracing subjectivity, Bayesian Inference fosters a more flexible and nuanced approach to statistical analysis, enabling practitioners to account for diverse sources of information and to update beliefs in light of new evidence. Consequently, Bayesian methods offer a comprehensive framework for inference that goes beyond traditional frequentist approaches, empowering researchers to make more informed and contextually relevant conclusions.
	
	\paragraph{Applications Across Disciplines}
	The versatility of Bayesian Inference sees its application across a myriad of disciplines. In genetics, it aids in understanding the distribution of genetic traits, elucidating complex inheritance patterns, and inferring ancestral relationships. Additionally, Bayesian methods are instrumental in machine learning, where they enrich predictive modeling by incorporating prior knowledge or assumptions, thereby enhancing model robustness and generalization. Furthermore, in epidemiology, Bayesian frameworks are indispensable for disease spread modeling, allowing for real-time updates of parameters and predictions based on evolving data, which is critical for effective public health interventions. Similarly, in finance, Bayesian analysis facilitates risk assessment by integrating prior beliefs with observed data to make informed decisions under uncertainty, optimizing portfolio management strategies, and enhancing market forecasting accuracy. Thus, Bayesian Inference serves as a unifying framework across diverse fields, enabling researchers and practitioners to leverage existing knowledge and empirical evidence for more accurate and reliable inference and decision-making.

	Bayesian Inference, with its foundation in Bayes' theorem and its philosophical approach to probability, offers a robust framework for dealing with uncertainty in statistical analysis. By enabling the dynamic updating of beliefs in light of new data, it provides a powerful tool for learning from data, applicable across diverse scientific and practical domains.

	\subsubsection{Key Principles and Mechanisms}
	
	\paragraph{Foundation in Bayes' Theorem}
	The bedrock of Bayesian Inference is Bayes' theorem, a mathematical axiom that elegantly delineates the process of updating the probability of a hypothesis in light of new evidence. This theorem is not just a formula but a fundamental principle of conditional probability, offering a systematic method for integrating prior knowledge with observed data to derive posterior knowledge. Bayes' theorem encapsulates the essence of Bayesian Inference, providing the mathematical mechanism to transition from prior beliefs to updated beliefs, thus embodying the core principle of learning from data.
	
	Furthermore, Bayes' theorem serves as a bridge between subjective prior beliefs and objective evidence, allowing practitioners to incorporate both subjective judgment and empirical data in a coherent framework. Moreover, the flexibility of Bayesian methods enables the incorporation of prior knowledge, which is particularly valuable in situations with limited data availability. Additionally, the iterative nature of Bayesian updating facilitates continuous learning and refinement of hypotheses, leading to more robust and accurate inference outcomes. Hence, Bayes' theorem stands as a cornerstone in the edifice of Bayesian Inference, guiding the rational assimilation of information and fostering a principled approach to decision-making.
	
	\paragraph{Prior Knowledge and Its Role}
	In Bayesian Inference, prior knowledge about the parameters of interest is encapsulated in what is known as the prior distribution. This prior distribution represents our beliefs about the parameters before any new data is observed, reflecting either subjective judgments or objective information derived from previous studies. The choice of prior can significantly influence the outcome of the inference process, underscoring the importance of carefully considering how much weight to give to prior knowledge relative to new evidence.
	
	Moreover, the incorporation of prior knowledge through the prior distribution allows for a principled integration of existing information with new data, thereby enabling a more robust and efficient inference process. Furthermore, the selection of an appropriate prior involves a delicate balance between incorporating relevant information and avoiding undue influence from prior beliefs, highlighting the need for thoughtful consideration and sensitivity analysis. Additionally, the Bayesian framework offers a coherent mechanism for updating beliefs in light of new evidence, facilitating iterative refinement of our understanding of the parameters of interest. Consequently, the judicious incorporation of prior knowledge is essential for producing meaningful and reliable inference results in Bayesian analysis.

	\paragraph{Observing New Evidence}
	The observational phase in Bayesian Inference involves collecting new data that will inform the updating process. This evidence is considered through the likelihood function, which assesses the probability of observing the data given specific values of the model parameters. The likelihood function plays a critical role in linking the observed data with the model's parameters, serving as a bridge between the prior beliefs and the posterior conclusions. \textbf{Moreover}, it encapsulates the essence of how the data influences the updating of beliefs, as it quantifies the plausibility of different parameter values \textbf{and} guides the direction of inference. \textbf{Additionally}, by incorporating the observed data, the likelihood function acts as a lens through which the model's parameters are scrutinized, allowing for the assessment of their compatibility with the data. \textbf{Furthermore}, the process of observing new evidence underscores the dynamic nature of Bayesian reasoning, as beliefs are continuously refined based on incoming data, leading to more informed and precise conclusions.

	\paragraph{Deriving Posterior Knowledge}
	The culmination of the Bayesian Inference process is the calculation of the posterior distribution, which mathematically combines the prior distribution and the likelihood of the observed data. This process integrates both prior knowledge and new evidence to update our understanding. The posterior distribution, represented by $P(\Theta|Data)$, encapsulates our refined beliefs about the parameters $\Theta$ after analyzing the available data. By leveraging Bayes' theorem, we can weigh the strength of our prior beliefs against the evidence provided by the data. This synthesis, denoted by $\frac{P(Data|\Theta)P(\Theta)}{P(Data)}$, highlights the interplay between prior assumptions and observed information, offering a comprehensive view of the underlying parameters. Notably, the normalization factor $P(Data)$ ensures that our posterior distribution is properly scaled, enabling meaningful comparisons and interpretations. Thus, the posterior distribution serves as a powerful tool for decision-making and inference in Bayesian analysis, guiding us towards more informed conclusions.

	\paragraph{Iterative Learning Process}
	Bayesian Inference is inherently iterative, with the posterior distribution from one analysis becoming the prior distribution for the next as more data becomes available. This iterative cycle facilitates a continuous learning process, allowing for the sequential updating of beliefs in light of accumulating evidence. Such an approach is particularly powerful in dynamic environments where data is collected in stages or where models need to adapt over time. Furthermore, this iterative nature enables the incorporation of new information seamlessly into the analysis, ensuring that the model remains relevant and accurate. Moreover, by leveraging past knowledge through the posterior distribution, Bayesian Inference provides a coherent framework for decision-making, where each iteration refines our understanding of the underlying phenomena. Consequently, this iterative learning process not only enhances model performance but also fosters a deeper comprehension of the system under study, leading to more informed and robust conclusions.

	\paragraph{Implications for Statistical Analysis}
	The principles and mechanisms of Bayesian Inference have profound implications for statistical analysis, offering a flexible and coherent framework for dealing with uncertainty and incorporating prior knowledge. \textbf{Moreover}, by mathematically formalizing the process of learning from data, Bayesian Inference provides a robust foundation for a wide range of statistical modeling and decision-making processes, adaptable to the complexities and nuances of real-world data. This framework allows analysts to not only quantify uncertainty but also to update beliefs as new data becomes available. \textbf{Additionally}, Bayesian methods enable the integration of prior information seamlessly into the analysis, enhancing the accuracy and reliability of results. \textbf{Furthermore}, the Bayesian approach facilitates the exploration of model parameters and assumptions, allowing for a more transparent and interpretable analysis. \textbf{On the other hand}, traditional frequentist methods often struggle with incorporating prior knowledge and tend to provide narrower interpretations of uncertainty. Thus, the Bayesian paradigm offers a comprehensive and versatile toolkit for statisticians and data analysts to tackle diverse analytical challenges.

	\subsubsection{The Role of Prior and Posterior Distributions}
	
	\paragraph{Understanding Prior Distributions}
	The concept of prior distributions is fundamental to Bayesian analysis, serving as the mathematical representation of initial beliefs or knowledge about the parameters of interest before any new data is observed. These priors can be based on historical data, expert opinion, or even subjective judgment, provided they are coherently formulated within a probabilistic framework. The selection of a prior distribution is a critical step in Bayesian inference, as it sets the stage for how new evidence will be incorporated into the analysis. Priors can range from non-informative, expressing a state of relative ignorance about the parameters, to highly informative, reflecting strong convictions or well-established knowledge.
	
	Prior distributions play a crucial role in Bayesian analysis, offering a means to incorporate domain expertise or existing information into statistical models. They enable practitioners to encode prior knowledge about the parameters being estimated, thus allowing for more informed inference, especially in cases where data may be limited or uncertain. Moreover, by accounting for uncertainty in prior beliefs, Bayesian methods provide a principled framework for decision-making under uncertainty. The flexibility of priors allows analysts to tailor their models to specific contexts, balancing between incorporating prior information and letting the data speak for itself. Consequently, understanding the implications of different prior choices is essential for conducting reliable Bayesian inference and deriving meaningful conclusions from data.

	\paragraph{Incorporating New Evidence}
	The essence of Bayesian learning lies in the incorporation of new evidence into existing beliefs. This process is facilitated through the observation of data and the application of Bayes' theorem, which mathematically combines the prior distribution with the likelihood of observing the new data under various parameter values. The likelihood function plays a pivotal role in this process, quantifying the probability of the new data given different parameter values and serving as the bridge between the prior knowledge and the observed evidence. Furthermore, the incorporation of new evidence is not a one-time event but rather an iterative process, as each new piece of data updates the posterior distribution, refining our understanding and beliefs about the underlying parameters. Therefore, Bayesian learning allows for a dynamic and adaptive approach to modeling, where the accumulation of evidence continuously shapes and updates our understanding of the underlying phenomena.

	\paragraph{Derivation of Posterior Distributions}
	The posterior distribution, a cornerstone of Bayesian analysis, encapsulates the essence of probabilistic inference, offering a nuanced portrayal of updated beliefs regarding model parameters in light of new evidence. It serves as a refined synthesis of prior knowledge and observed data, embodying the Bayesian update mechanism. Mathematically, the posterior distribution is elegantly expressed as:
	\[
	P(\Theta|Data) = \frac{P(Data|\Theta) \cdot P(\Theta)}{P(Data)},
	\]
	where $P(\Theta|Data)$ denotes the posterior probability of parameters $\Theta$ given the observed data, $P(Data|\Theta)$ represents the likelihood of the data given the parameters, and $P(\Theta)$ signifies the prior probability distribution over the parameters. The denominator, $P(Data)$, acts as a normalization factor, ensuring that the posterior distribution integrates to unity. This formula underscores the dynamic interplay between prior beliefs and incoming data, elucidating how evidence refines our understanding of model parameters. Furthermore, it underscores the iterative nature of Bayesian inference, wherein the posterior of one analysis can serve as the prior for subsequent analyses, fostering a continuous refinement of knowledge.

	\paragraph{From Prior to Posterior: The Bayesian Update}
	The transition from prior to posterior distributions is at the heart of Bayesian learning, illustrating the dynamic nature of the inference process. As new data is observed and analyzed, the posterior distribution evolves, reflecting an updated synthesis of prior beliefs and empirical evidence. This continuous update mechanism allows Bayesian inference to remain responsive to new information, facilitating an ongoing refinement of our understanding of the underlying parameters.
	
	Moreover, the Bayesian update encapsulates a profound philosophical stance towards uncertainty. While classical statistical methods often treat parameters as fixed but unknown quantities, Bayesian inference treats them as random variables with probability distributions. Thus, the posterior distribution represents the quantification of uncertainty after considering both prior beliefs and observed data, providing a comprehensive framework for decision-making under uncertainty.
	
	Furthermore, the Bayesian update process is recursive, enabling iterative refinement of beliefs with each new piece of evidence. This recursive nature allows for adaptive modeling, where the model becomes increasingly tailored to the available data over time. Consequently, Bayesian inference offers a powerful tool for updating and revising hypotheses in light of accumulating evidence, fostering a flexible and nuanced approach to statistical inference.
	
	\paragraph{The Iterative Nature of Bayesian Learning}
	Bayesian analysis inherently supports an iterative learning process, where the posterior distribution from one stage of analysis becomes the prior for the next as more data becomes available. This feature makes Bayesian methods particularly suited to applications where data accumulates over time or where decisions must be updated with the arrival of new information. \textbf{Moreover}, the ability to seamlessly integrate new evidence into existing models underscores the flexibility and adaptability of the Bayesian approach. This iterative framework allows for continual refinement and improvement, ensuring that decisions are not solely based on initial assumptions but are continuously updated to reflect the most current understanding of the data. \textbf{Additionally}, by incorporating prior knowledge with new data, Bayesian analysis can provide more stable estimates, especially in cases where the sample size is small or where data is noisy. \textbf{Furthermore}, this iterative process promotes a deeper understanding of the underlying phenomena being studied, as each iteration allows for the refinement of hypotheses in light of new evidence. In practical terms, this means that Bayesian models can evolve over time, becoming increasingly accurate and reliable as more data is gathered and analyzed.
	
	\paragraph{Impact on Statistical Inference}
	The interplay between prior and posterior distributions in Bayesian analysis not only provides a robust framework for incorporating prior knowledge and new evidence but also highlights the philosophical underpinnings of Bayesian thought. By quantifying uncertainty and belief updating in probabilistic terms, Bayesian inference offers a powerful tool for understanding and navigating the complexities of the world, grounded in the principles of rationality and evidence-based decision-making. Additionally, it fosters a nuanced comprehension of uncertainty, allowing for more informed and nuanced decision-making processes. Moreover, it facilitates the integration of various sources of information, enabling analysts to combine diverse data streams effectively. Furthermore, Bayesian inference promotes a coherent approach to inference by aligning prior beliefs with observed data, thereby reducing biases inherent in traditional statistical methodologies. Consequently, it enhances the interpretability and robustness of statistical analyses, leading to more reliable conclusions and actionable insights. Hence, the adoption of Bayesian principles in statistical inference not only enhances analytical rigor but also promotes a deeper understanding of uncertainty's role in decision-making processes.

	\subsubsection{Applications and Limitations}
	
	\paragraph{Broad Spectrum of Applications}
	Bayesian Inference has cemented its role across a diverse array of fields, showcasing its versatility and the depth of its applicability. In epidemiology, it aids in modeling the spread of diseases and evaluating the effectiveness of interventions, allowing for the incorporation of prior knowledge and real-time data updates. \textbf{Moreover}, the finance sector leverages Bayesian methods to assess risk, forecast market trends, and make informed investment decisions, benefiting from the probabilistic nature of Bayesian models to handle uncertainty. \textbf{Furthermore}, machine learning and artificial intelligence extensively utilize Bayesian approaches for predictive modeling, reinforcement learning, and unsupervised learning tasks, where the ability to update predictions or models in light of new data is crucial.

	\paragraph{Handling Uncertainty with Elegance}
	One of the most compelling aspects of Bayesian Inference is its principled approach to handling uncertainty. By framing both prior knowledge and new evidence in probabilistic terms, Bayesian methods provide a coherent and consistent framework for making inferences and decisions under uncertainty. This approach not only facilitates the integration of diverse sources of information but also allows for the explicit quantification of uncertainty in the conclusions drawn.
	
	Moreover, Bayesian inference accommodates the dynamic nature of knowledge acquisition by updating beliefs in light of new evidence through Bayes' theorem. This iterative process enhances decision-making by continuously refining estimates and predictions, thereby adapting to evolving scenarios. Furthermore, the incorporation of uncertainty in Bayesian models enables decision-makers to assess the robustness of conclusions and consider alternative hypotheses. Additionally, Bayesian frameworks offer a principled approach to handling complex scenarios where data are sparse or noisy, allowing for informed decisions even in challenging contexts. Hence, Bayesian inference stands as a versatile and powerful tool for addressing uncertainty in diverse domains, ranging from scientific research to real-world applications.

	\paragraph{Computational Complexity Challenges}
	Despite its conceptual elegance and broad applicability, Bayesian Inference is not without its limitations. One of the most significant challenges is the computational complexity associated with calculating posterior distributions, especially for high-dimensional models or complex likelihood functions. Advanced computational techniques, such as Markov Chain Monte Carlo (MCMC) methods, have been developed to address these challenges, but they often require substantial computational resources and expertise to implement effectively. Furthermore, the efficiency of these methods can vary depending on the specific characteristics of the model and the data being analyzed. Moreover, as models become increasingly complex or data sets grow larger, the computational burden can become prohibitive, hindering the widespread adoption of Bayesian approaches in certain domains. Consequently, researchers are constantly exploring new algorithms and computational strategies to improve the scalability and efficiency of Bayesian inference methods, aiming to make them more accessible and practical for a wider range of applications.

	\paragraph{Subjectivity in Prior Selection}
	Another limitation stems from the subjective nature of selecting prior distributions. The choice of priors can significantly influence the outcomes of Bayesian analysis, raising concerns about objectivity and reproducibility. While the use of non-informative or weakly informative priors offers a potential solution, the art of choosing appropriate priors remains a critical and sometimes contentious aspect of Bayesian practice, necessitating careful consideration and justification. Furthermore, the dependence on prior selection introduces a layer of subjectivity that may undermine the credibility of Bayesian results. Moreover, the lack of standardized guidelines for selecting priors can lead to inconsistency across analyses, hindering comparability and reproducibility. Nonetheless, despite these challenges, Bayesian practitioners must navigate the complexities of prior selection with diligence and transparency to ensure the robustness and reliability of their analyses. Hence, methodologies for validating prior choices and assessing sensitivity to prior specifications are crucial for enhancing the rigor and trustworthiness of Bayesian inference.

	\paragraph{Navigating the Limitations}
	To navigate these limitations, researchers and practitioners in Bayesian Inference continually develop new methodologies, computational tools, and best practices. Efforts to standardize prior selection, improve computational efficiency, and enhance the transparency of Bayesian analyses are ongoing, reflecting the vibrant and evolving nature of the field. Moreover, as Bayesian methods become increasingly prevalent across diverse disciplines, interdisciplinary collaboration has become imperative. Furthermore, in addressing computational challenges, parallel computing architectures, such as Graphics Processing Units (GPUs) and High-Performance Computing (HPC) clusters, offer promising solutions. Additionally, embracing open-source software and reproducible research practices fosters a culture of transparency and facilitates the exchange of ideas within the Bayesian community. Hence, while navigating these challenges may be daunting, the collective efforts of researchers, coupled with technological advancements, propel Bayesian Inference towards greater accessibility and reliability. Consequently, the landscape of Bayesian analysis continues to evolve, driven by a commitment to innovation and collaboration.

	\paragraph{Conclusion}
	The widespread applications of Bayesian Inference underscore its fundamental importance in scientific research, decision-making, and data analysis. Despite the challenges associated with computational complexity and prior selection, the Bayesian framework's ability to coherently update beliefs in light of new evidence remains an invaluable tool in the quest for understanding complex phenomena. Moreover, as computational capabilities advance and methodologies refine, the potential for Bayesian Inference to provide deep insights and robust solutions to a wide range of problems continues to grow. Furthermore, its flexibility allows for the integration of various data sources and the incorporation of prior knowledge, enhancing the robustness of analyses. Additionally, Bayesian methods offer a principled approach to uncertainty quantification, crucial in fields where accurate risk assessment is paramount. Hence, embracing Bayesian principles not only enriches scientific inquiry but also empowers decision-makers with more informed choices. In the same way, Bayesian Inference fosters interdisciplinary collaborations, bridging gaps between disparate fields through a unified probabilistic framework. Thus, its adoption promises to catalyze innovation and drive progress across diverse domains.

	\subsubsection{Pseudocode for Bayesian Inference}
	The Bayesian Inference Algorithm is a powerful framework tailored for estimating parameters in statistical models, especially when handling latent variables. It operates by iteratively updating the posterior distribution of parameters based on observed data and latent variables, incorporating prior knowledge to refine parameter estimates. The operational procedure of Bayesian inference can be depicted in pseudocode \ref{fig:bayesian-pseudocode}, demonstrating its iterative nature in parameter estimation.
	
	\begin{algorithm}
		\caption{Bayesian Inference Pseudocode}
		\begin{algorithmic}[1]
			\Procedure{BayesianInference}{Data, PriorDistributions}
			\State \textbf{initialize} posterior distributions as PriorDistributions
			\For{each piece of new evidence in Data}
			\State \textbf{calculate} likelihood of new evidence given model parameters
			\State \textbf{update} posterior distributions using Bayes' theorem:
			\State $Posterior = \frac{Likelihood \times Prior}{Evidence}$
			\State \textbf{set} PriorDistributions to updated posterior distributions
			\EndFor
			\State \Return updated posterior distributions as new beliefs
			\EndProcedure
		\end{algorithmic}\label{fig:bayesian-pseudocode}
	\end{algorithm}
	
\subsection{Previous Work on ML and AI Interplay with the Bayesian Inference Algorithm}

\paragraph{Machine Learning in Microseismic Event Analysis}
The integration of machine learning with Bayesian inference for microseismic event analysis has been explored in recent research. In a study from 2023, an approach was proposed to expedite Bayesian posterior inference for localizing microseismic events and identifying their mechanisms. This method utilizes machine learning algorithms to construct a surrogate model trained on seismic data, enabling faster Bayesian inference. By replacing computationally intensive forward modeling, the analysis time has been notably reduced to less than an hour on standard computing hardware without compromising accuracy. This approach signifies the potential of machine learning to streamline traditional Bayesian methods for understanding seismic activities \cite{piras2023towards}.

\paragraph{Bayesian Parameter Estimation in Quantum Systems}
In the field of quantum information science, machine learning has also been applied to improve Bayesian parameter estimation. A study in 2021 introduced a machine learning approach for Bayesian parameter estimation in quantum systems. By combining neural networks with Bayesian estimation techniques, this research demonstrates the accurate estimation of quantum system parameters, such as qubit state rotation angles. A neural network trained on quantum measurement data predicts system parameters, which are then refined through Bayesian inference. This approach not only enhances parameter estimation in quantum systems but also showcases the versatility of machine learning in extending Bayesian inference capabilities across various scientific disciplines. Its success in quantum systems highlights the broader applicability of machine learning-enhanced Bayesian methods in complex parameter estimation tasks \cite{nolan2021machine}.

\subsection{Algogenic Enhancements for Bayesian Inference}
\subsubsection{Prior Knowledge Synthesis}
\paragraph{Integrating Comprehensive Domain Insights}
The enhancement of Bayesian Inference through the synthesis of prior knowledge from domain-specific insights represents a strategic application of Algogens, specifically focusing on the absorption and integration of vast quantities of domain-relevant information into actionable prior distributions. This process meticulously employs Large Language Models to navigate and distill critical insights from extensive databases, scientific literature, and domain-specific repositories, aiming to craft prior distributions that are both statistically robust and deeply reflective of the current domain understanding. The inclusion of such comprehensive domain insights enhances the formation of priors, ensuring they are not only grounded in statistical rigor but also enriched with the nuanced understanding of the domain, thereby improving the interpretability and relevance of Bayesian analyses in addressing complex industry problems.

\paragraph{Enhancing Priors with Contextual Relevance}
The enhancement of priors in Bayesian Inference through the integration of contextual relevance emphasizes the application of Algogens in refining the base knowledge with which Bayesian models operate. This entails a detailed examination and integration of domain-specific variables, empirical findings, and expert knowledge into the formulation of prior distributions, thereby ensuring that the Bayesian models commence with a foundation that is statistically sound and closely aligned with real-world phenomena. The process involves a careful analysis of patterns and dependencies within the domain, translating this complex domain knowledge into quantifiable and applicable prior information, which in turn improves the models' accuracy and responsiveness to the underlying complexities of the data.

\paragraph{Setting the Stage for Informed Inference}
The application of Algogens in setting the stage for informed Bayesian inference through Prior Knowledge Synthesis revolves around equipping the Bayesian models with a level of insight and specificity that significantly enhances their subsequent analysis phases. By initiating the Bayesian inference process with a foundation of enhanced knowledge, these models are better positioned to uncover meaningful, accurate, and actionable insights, thereby advancing the understanding and decision-making capabilities across diverse applications. This strategic enrichment of Bayesian Inference with domain-specific insights underscores the practicality of Algogenic enhancements in leveraging extensive information sources for statistical modeling, ensuring that the models remain relevant, adaptable, and informative in dynamic research environments.

\subsubsection{Data Quality Analysis}
\paragraph{Elevating Data Integrity for Bayesian Computation}
In the context of Bayesian Inference, elevating data integrity through Data Quality Analysis specifically involves the application of Algogens to meticulously evaluate and ensure the dataset's appropriateness for sophisticated statistical analysis. This Algogenic enhancement deploys LLMs to conduct a comprehensive assessment of the data, pinpointing anomalies, inconsistencies, and gaps that could potentially compromise the inference process. By enhancing data quality, the Bayesian computation is grounded on a dataset that is both accurate and reflective of the study's domain, thereby streamlining the preprocessing stage and enabling a more efficient and effective analysis.

\paragraph{Automating Preprocessing Recommendations}
The automation of preprocessing recommendations in Bayesian Inference through Data Quality Analysis represents an Algogenic application aimed at refining data quality. This involves leveraging LLMs to suggest targeted preprocessing strategies based on the comprehensive analysis of the dataset, including handling missing data, correcting outliers, and transforming variables to align with Bayesian analysis requirements. Such enhancements ensure that the dataset is optimized for analysis, thereby facilitating a more accurate and efficient Bayesian modeling process. The strategic application of Algogens in this context underscores the practicality of leveraging AI to automate and refine the preprocessing steps, enhancing the overall quality and reliability of the data used in Bayesian analyses.

\paragraph{Setting the Foundation for Accurate Inference}
In Bayesian Inference, setting the foundation for accurate inference through Data Quality Analysis emphasizes the critical role of Algogens in optimizing the dataset's quality. This entails a thorough enhancement of the data's integrity, ensuring that it is suitably prepared for the intricate analysis processes ahead. By embedding such Algogenic enhancements, the inference process is fortified with a solid foundation of high-quality data, significantly reducing the likelihood of inaccuracies and enabling the derivation of reliable, actionable insights. The application of Algogens in this phase of Bayesian Inference illustrates the shift towards more data-driven, precise, and informed statistical modeling, highlighting the practicality and value of such enhancements in advancing the accuracy and reliability of the inference process.

\subsubsection{Dynamic Prior Updating}
\paragraph{Adapting Priors to New Evidence in Real-time}
The adaptation of priors to new evidence in real-time, within the framework of Bayesian Inference, showcases the specific application of Algogens in maintaining the relevance and accuracy of the inference process. This Algogenic enhancement leverages LLMs to dynamically adjust prior distributions as new data becomes available, ensuring that the Bayesian models continuously reflect the most current knowledge and evidence. Such dynamic updating of priors enhances the model's responsiveness to changing information landscapes, thereby improving the robustness and reliability of the inference outcomes. The strategic application of Algogens here not only highlights the practicality of incorporating real-time data analysis but also illustrates the value of such enhancements in ensuring the Bayesian models remain adaptable and accurate in rapidly evolving domains.

\paragraph{Facilitating a More Responsive Inference Framework}
Facilitating a more responsive inference framework through Dynamic Prior Updating in Bayesian Inference underscores the strategic application of Algogens in enhancing the model's adaptability to new findings. This involves employing LLMs to ensure that the Bayesian models are continually updated with the latest data, thereby maintaining their relevance and accuracy over time. Such an Algogenic enhancement allows for the Bayesian inference process to be more adaptable and responsive, capable of efficiently incorporating new evidence and adjusting priors accordingly. The specific application of Algogens in this context demonstrates the practicality and value of leveraging AI to create more dynamic and flexible models, enhancing their predictive accuracy and relevance in various domains.

\paragraph{Bridging Data and Knowledge with AI Insight}
In Bayesian Inference, bridging data and knowledge with AI insight through Dynamic Prior Updating highlights the application of Algogens in seamlessly integrating new evidence into the model's framework. This Algogenic enhancement utilizes LLMs to analyze incoming data in real-time, adjusting the priors to ensure that the Bayesian models remain aligned with the latest developments and insights. Such enhancements not only improve the models' robustness against data shifts but also maximize the utility of accumulated knowledge, thereby driving more accurate and timely decision-making. The strategic application of Algogens in this phase of Bayesian Inference showcases the practicality and value of AI in enhancing the adaptability and accuracy of statistical models, ensuring they remain relevant and effective in dynamic environments.

\subsubsection{Intelligent Hypothesis Generation}
\paragraph{Expanding the Horizon of Statistical Exploration}
Expanding the horizon of statistical exploration through Intelligent Hypothesis Generation in Bayesian Inference exemplifies the specific application of Algogens in transcending traditional hypothesis-driven analyses. This Algogenic enhancement leverages LLMs to generate novel hypotheses or models, enabling a broader and more exploratory approach to analyzing complex datasets. Such an approach not only accelerates the discovery process but also ensures that the exploration is grounded in statistical rigor and relevance to the data. The strategic use of Algogens here underscores the practicality and value of AI-driven exploration in uncovering hidden patterns and relationships, thereby pushing the boundaries of knowledge discovery in data science.

\paragraph{AI as a Catalyst for Discovery}
The role of AI as a catalyst for discovery in Bayesian Inference, facilitated by Intelligent Hypothesis Generation, highlights the strategic application of Algogens in broadening the scope of statistical exploration. This involves employing LLMs to suggest innovative hypotheses based on an in-depth analysis of the dataset and domain knowledge, thereby enabling a more systematic and efficient method for exploring a wider range of possibilities. The incorporation of AI in this context not only accelerates the discovery process but also enhances the statistical analysis's relevance and rigor. The application of Algogens in Intelligent Hypothesis Generation demonstrates the practicality and value of leveraging AI capabilities to drive scientific discovery and innovation across various domains.

\paragraph{Revolutionizing Model Development and Validation}
In Bayesian Inference, revolutionizing model development and validation through Intelligent Hypothesis Generation underscores the application of Algogens in fostering a more exploratory and innovative approach to statistical analysis. This Algogenic enhancement utilizes LLMs to actively generate and validate new hypotheses, thereby democratizing the discovery process and enabling a broader exploration of data-driven questions. Such enhancements not only enhance the efficiency and efficacy of hypothesis generation but also encourage cross-disciplinary collaboration, enriching the exploration process and fostering innovation. The strategic application of Algogens in this phase of Bayesian Inference highlights the practicality and value of AI-driven methodologies in advancing model development and validation, pushing the boundaries of knowledge discovery and innovation.

\subsubsection{Adaptive Sampling Strategies}
\paragraph{Optimizing the Efficiency of Evidence Gathering}
Optimizing the efficiency of evidence gathering in Bayesian Inference through Adaptive Sampling Strategies exemplifies the specific application of Algogens in refining how data samples are selected and analyzed. This Algogenic enhancement leverages LLMs to dynamically adjust sampling methods based on the evolving state of the inference process, ensuring each sample maximally contributes to refining the model's accuracy. Such strategic application of Algogens not only reduces computational demands but also enhances the quality of insights derived from the data, showcasing the practicality and value of AI in making Bayesian inference more efficient and effective.

\paragraph{Tailoring Sampling to the Needs of the Model}
Tailoring sampling to the needs of the model in Bayesian Inference through Adaptive Sampling Strategies highlights the application of Algogens in optimizing the data selection process for analysis. This involves employing LLMs to assess current model parameters and data distributions, thereby determining the most informative regions for subsequent sampling. Such enhancements ensure that sampling is always aligned with achieving a comprehensive understanding of the underlying phenomena, thereby enhancing the model's capability to capture nuanced patterns and outliers. The strategic use of Algogens in this context underscores the practicality and value of leveraging AI to refine sampling methodologies, facilitating more agile and effective model training and deployment across various applications.

\paragraph{Achieving Deeper Insights with Fewer Resources}
Achieving deeper insights with fewer resources in Bayesian Inference through Adaptive Sampling Strategies underscores the strategic application of Algogens in making the evidence-gathering process more resource-efficient. This Algogenic enhancement utilizes LLMs to intelligently select data points for analysis, focusing computational efforts on areas where information gain is maximized. Such enhancements not only ensure the efficient use of computational resources but also enable the exploration of high-dimensional parameter spaces with limited budgets. The application of Algogens in this phase of Bayesian Inference demonstrates the practicality and value of AI-driven methodologies in enhancing statistical inference processes, enabling deeper insights into complex datasets with fewer resources.

\subsubsection{Result Interpretation and Explanation}
\paragraph{Bridging the Gap Between Complex Models and Actionable Insights}
Bridging the gap between complex models and actionable insights in Bayesian Inference through Result Interpretation and Explanation highlights the application of Algogens in making statistical results accessible and actionable for decision-makers. This involves leveraging LLMs to interpret statistical outcomes, elucidate model parameters' implications, and provide a narrative that contextualizes the findings within the domain of application. Such enhancements not only address the challenge of making complex models understandable for non-experts but also enhance the interpretability and applicability of Bayesian analyses, demonstrating the practicality and value of Algogenic enhancements in translating statistical insights into informed decisions.

\paragraph{Enhancing Understanding Through AI-Driven Narratives}
Enhancing understanding through AI-driven narratives in Bayesian Inference, facilitated by Result Interpretation and Explanation, underscores the strategic application of Algogens in extracting and articulating significance from statistical results. This involves employing LLMs to generate narratives that provide a deeper comprehension of the Bayesian framework, including the influence of prior assumptions and the coherence of predictions within the broader domain context. Such enhancements not only enrich model evaluation with qualitative insights but also foster informed decision-making by bridging the gap between abstract mathematical formalism and tangible practical relevance. The use of Algogens in this context highlights the practicality and value of AI in enhancing model development and stakeholder communication, promoting a more nuanced and comprehensive approach to statistical analysis.

\paragraph{Facilitating Data-Driven Decision Making}
Facilitating data-driven decision-making in Bayesian Inference through Result Interpretation and Explanation emphasizes the application of Algogens in improving the usability of statistical models for real-world applications. This Algogenic enhancement leverages LLMs to provide clear, contextualized explanations of the results, ensuring that the insights generated are not only scientifically robust but also practically relevant and readily applicable. Such enhancements democratize access to advanced statistical findings, enabling a broader range of stakeholders to benefit from data-driven insights and fostering an informed approach to complex challenges. The strategic use of Algogens in this phase of Bayesian Inference demonstrates the practicality and value of AI-driven methodologies in advancing the field of statistical inference and empowering decision-makers with the tools necessary to navigate complex data-rich environments effectively.

\subsubsection{Predictive Validity Checks}
\paragraph{Ensuring Model Robustness through Forward-Looking Analysis}
Ensuring model robustness through forward-looking analysis in Bayesian Inference, facilitated by Predictive Validity Checks, underscores the application of Algogens in assessing the inferred models' applicability and reliability by evaluating their predictive performance on new, unseen data. This Algogenic enhancement leverages LLMs to simulate future data scenarios or utilize split-sample validation techniques, providing a rigorous test of the model's generalizability. Such enhancements not only ensure the statistical soundness of the Bayesian analysis but also its practical reliability, demonstrating the practicality and value of Algogenic enhancements in ensuring the robustness and reliability of Bayesian models for real-world applications.

\paragraph{Tailoring Models to Real-World Applications}
Tailoring models to real-world applications through Predictive Validity Checks in Bayesian Inference highlights the strategic application of Algogens in bridging the gap between theoretical model development and practical application. This involves employing LLMs to identify the most relevant and challenging scenarios for model validation, ensuring that the tests reflect realistic complexities and variabilities. Such enhancements guarantee that the model remains robust and accurate even when confronted with unpredictable real-world data, underscoring the practicality and value of Algogenic enhancements in developing models that are not only theoretically sound but also practically applicable and resilient.

\paragraph{Reinforcing Confidence in Model Predictions}
Reinforcing confidence in model predictions through Predictive Validity Checks in Bayesian Inference emphasizes the application of Algogens in enhancing the credibility of statistical models and boosting stakeholders' confidence in the model's predictions. This Algogenic enhancement ensures that models are thoroughly vetted for predictive accuracy and reliability, making Bayesian inference a more powerful tool for informed decision-making. Such enhancements address concerns regarding the reproducibility and generalizability of model outcomes, promoting transparency in the modeling process and fostering trust and confidence among users and stakeholders. The strategic use of Algogens in this phase of Bayesian Inference showcases the practicality and value of AI-driven methodologies in reinforcing confidence in model predictions, ensuring that Bayesian models deliver actionable insights that drive informed decision-making in high-stakes domains.

\subsubsection{Semantic Analysis of Model Fit}
\paragraph{Deepening Insight with Contextual Model Evaluation}
Deepening insight with contextual model evaluation in Bayesian Inference through Semantic Analysis of Model Fit exemplifies the specific application of Algogens in evaluating the suitability and effectiveness of models beyond traditional statistical metrics. This Algogenic enhancement leverages LLMs to perform a deep, contextual analysis of the model's fit, considering both quantitative measures and qualitative alignment with known domain knowledge. Such enhancements offer a more holistic assessment of model fit, identifying strengths and areas for refinement, and demonstrate the practicality and value of Algogenic enhancements in ensuring that Bayesian models capture the underlying dynamics of the data comprehensively and accurately.

\paragraph{Bridging Quantitative Analysis and Qualitative Insights}
Bridging quantitative analysis and qualitative insights in Bayesian Inference through Semantic Analysis of Model Fit underscores the strategic application of Algogens in amalgamating numerical metrics with subjective interpretation into a cohesive narrative that enriches model evaluation. This involves leveraging LLMs to compute performance indicators and contextualize these metrics within the broader domain knowledge, thereby enhancing model evaluation with deeper insights into the model's capabilities and limitations. Such enhancements not only validate the model against domain-specific benchmarks but also empower stakeholders with actionable insights, underscoring the practicality and value of Algogenic enhancements in promoting a more nuanced and comprehensive approach to model evaluation.

\paragraph{Enhancing Model Development and Stakeholder Communication}
Enhancing model development and stakeholder communication in Bayesian Inference through Semantic Analysis of Model Fit highlights the application of Algogens in improving the interpretability and applicability of Bayesian analyses. This Algogenic enhancement leverages LLMs to articulate the strengths and limitations of the model in a language that resonates with stakeholders, promoting transparent and coherent communication. Such enhancements not only facilitate a more informed decision-making process but also guide researchers in refining the model for greater accuracy and utility, demonstrating the practicality and value of Algogenic enhancements in bridging the gap between statistical results and practical implications, thereby fostering a collaborative approach to model development and refinement.

\subsubsection{Evidence Synthesis from Unstructured Data}
\paragraph{Harnessing Rich Data Sources for Bayesian Enrichment}
Harnessing rich data sources for Bayesian enrichment through Evidence Synthesis from Unstructured Data exemplifies the specific application of Algogens in extending the Bayesian inference framework to incorporate insights from diverse, unstructured data sources. This Algogenic enhancement leverages LLMs to extract and synthesize evidence from text, images, videos, and other non-traditional data formats, significantly broadening the evidential base for Bayesian analysis. Such strategic application of Algogens not only enriches the Bayesian model with a richer, more nuanced understanding of the phenomena under study but also demonstrates the practicality and value of leveraging AI to advance scientific understanding and decision-making across diverse domains.

\paragraph{Transforming Data into Quantifiable Insights}
Transforming data into quantifiable insights through Evidence Synthesis from Unstructured Data in Bayesian Inference highlights the strategic application of Algogens in integrating insights from diverse, unstructured data sources into the Bayesian framework. This involves employing LLMs to interpret unstructured data, identify relevant patterns, and translate these findings into quantifiable inputs for Bayesian analysis. Such enhancements enrich the Bayesian model with insights previously inaccessible, showcasing the practicality and value of Algogenic enhancements in leveraging the wealth of information contained in unstructured data sources to make informed decisions and draw more accurate conclusions about complex phenomena.

\paragraph{Expanding the Frontiers of Statistical Analysis}
Expanding the frontiers of statistical analysis through Evidence Synthesis from Unstructured Data in Bayesian Inference underscores the application of Algogens in embracing a wider spectrum of information sources for Bayesian models. This Algogenic enhancement not only enhances the depth and accuracy of the analysis but also opens up new possibilities for interdisciplinary research and applications. By incorporating qualitative insights alongside quantitative measures, Bayesian analysis transcends traditional boundaries, paving the way for a more nuanced, multidimensional approach to statistical inference and knowledge discovery. The strategic use of Algogens in this phase of Bayesian Inference showcases the practicality and value of AI-driven methodologies in advancing the field of statistical analysis, enabling researchers to capture the complexities and nuances of complex systems comprehensively.

	\subsubsection{Pseudocode for Algogenic Bayesian Inference}
	The Algogenic Bayesian inference approach enhances traditional Bayesian inference methods by dynamically adjusting inference parameters and strategies based on observed data and real-time error estimates. This pseudocode, available in \ref{fig:bayesian-inference-Algogen-pseudocode}, outlines an advanced framework incorporating AI-driven enhancements for adaptive model updating, hypothesis selection, likelihood evaluation, and real-time parameter optimization.
	
	\begin{algorithm}
		\caption{Algogenic Bayesian Inference Pseudocode}
		\begin{algorithmic}[1]
			\Procedure{AlgogenicBayesianInference}{$data$}
			\State $priors \gets \Call{PriorKnowledgeSynthesis}{data}$ \Comment{Use AI to formulate priors}
			\State $data \gets \Call{DataQualityAnalysis}{data}$ \Comment{AI-enhanced data cleaning}
			\While{not \Call{Convergence}{model}}
			\State $hypotheses \gets \Call{IntelligentHypothesisGeneration}{data, priors}$
			\For{each $hypothesis$ in $hypotheses$}
			\State $model \gets \Call{BayesianUpdate}{hypothesis, data, priors}$
			\State $priors \gets \Call{DynamicPriorUpdating}{model, data}$
			\State $samples \gets \Call{AdaptiveSamplingStrategies}{model, data}$
			\EndFor
			\State $validity \gets \Call{PredictiveValidityChecks}{model}$
			\If{$validity$ is satisfactory}
			\State \textbf{break}
			\EndIf
			\EndWhile
			\State $interpretation \gets \Call{ResultInterpretationAndExplanation}{model}$
			\State \Return $interpretation$
			\EndProcedure
		\end{algorithmic}\label{fig:bayesian-inference-Algogen-pseudocode}
	\end{algorithm}

	\begin{figure}
		\centering
		\includegraphics[width=0.6\textwidth]{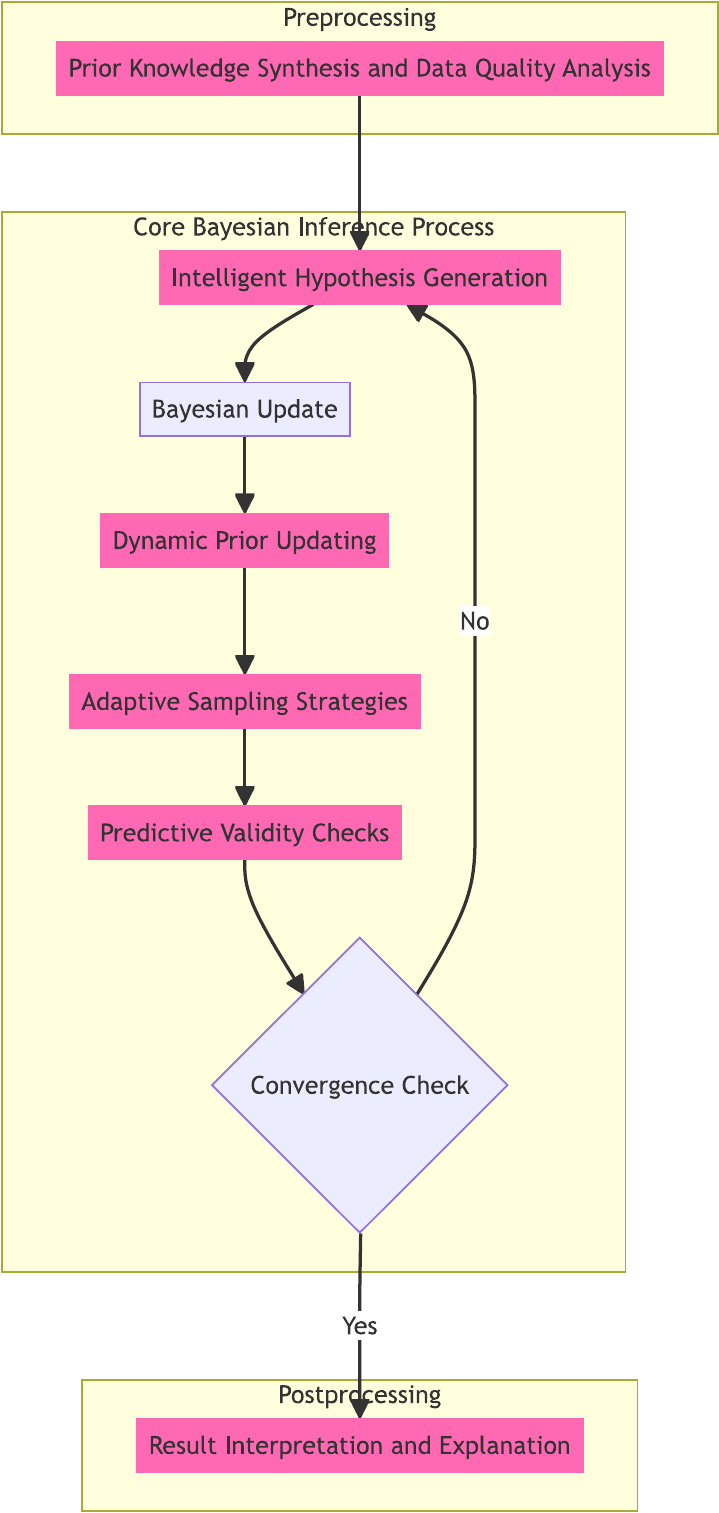} 
		\caption{Integrating Algogenic Enhancements into Bayesian Inference: This diagram visualizes the comprehensive incorporation of generative AI into the Bayesian inference process. It highlights the transformative Algogenic enhancements at each stage, from preprocessing with Prior Knowledge Synthesis and Data Quality Analysis through the core inference process, featuring Intelligent Hypothesis Generation, Dynamic Prior Updating, and Adaptive Sampling Strategies, to the postprocessing phase of Result Interpretation and Explanation. Each step is designed to leverage generative AI for optimizing the inference process, enhancing the adaptability and depth of analysis, and ensuring the resulting insights are robust, relevant, and readily interpretable. This integration exemplifies the synergy between traditional Bayesian methods and modern AI capabilities, setting a new standard for precision and insight in statistical analysis.}
		\label{fig:bayesian_inference}
	\end{figure}

	\section{Principal Component Analysis}\index{Principal Component Analysis}
	
	\subsection{Introduction to PCA}
	\subsubsection{Mathematical Framework of PCA}
	Principal Component Analysis is grounded in a mathematical framework that seeks to transform the original data space into a new space of reduced dimensionality while retaining as much of the variance present in the original dataset as possible. This process involves identifying the principal components (PCs) that capture the most variance within the data.
	
	\paragraph{Covariance Matrix Computation}
	The first step in PCA involves computing the covariance matrix of the original dataset. The covariance matrix, denoted as $\Sigma$, captures the pairwise covariances between variables in the dataset. This matrix encapsulates the relationships between different dimensions, revealing how changes in one variable correspond to changes in another. This foundational computation forms the basis for understanding the underlying structure of the dataset, enabling PCA to identify patterns and extract meaningful features. Furthermore, $\Sigma$ provides insights into the spread and orientation of the data cloud in the multidimensional space, aiding in dimensionality reduction. Consequently, by analyzing the eigenvalues and eigenvectors of $\Sigma$, PCA identifies the principal components that best represent the variability in the data. Through this process, PCA seeks to maximize the variance along the principal components, facilitating dimensionality reduction while preserving as much information as possible from the original dataset. Therefore, the computation of $\Sigma$ is pivotal in PCA, serving as a fundamental building block for subsequent analyses.

	\paragraph{Eigenvalue Decomposition}
	The core of PCA's mathematical framework lies in the eigenvalue decomposition of the covariance matrix $\Sigma$. This decomposition identifies eigenvectors and eigenvalues of $\Sigma$, where each eigenvector represents a principal component direction in the dataset, and the corresponding eigenvalue indicates the amount of variance captured by that direction. Mathematically, this involves solving the equation $\Sigma \mathbf{v} = \lambda \mathbf{v}$, where $\mathbf{v}$ is an eigenvector of $\Sigma$, and $\lambda$ is the corresponding eigenvalue. Furthermore, the eigenvectors are orthogonal to each other, meaning they represent uncorrelated directions in the original feature space. Consequently, the largest eigenvalues correspond to the directions with the most variance, allowing PCA to effectively reduce the dimensionality of the dataset by selecting the top $k$ eigenvectors with the highest eigenvalues. This reduction retains the most significant information while discarding the least informative dimensions. Thus, PCA facilitates the transformation of the original high-dimensional dataset into a lower-dimensional subspace without losing crucial information about its structure and relationships.
	
	\paragraph{Selection of Principal Components}
	After computing the eigenvectors and eigenvalues, the next step is to select the top $k$ eigenvectors that correspond to the largest eigenvalues, as these principal components capture the most variance. The number of components $k$ is chosen based on the amount of total variance one wishes to retain in the reduced data representation. However, determining the optimal $k$ can be challenging, as selecting too few components may lead to loss of important information, while selecting too many may introduce noise or overfitting. Therefore, a balance must be struck to effectively reduce the dimensionality of the data while preserving its essential characteristics. Furthermore, it's crucial to consider the computational cost associated with higher values of $k$, especially for large datasets. Hence, careful consideration should be given to the trade-offs between dimensionality reduction, information retention, and computational efficiency. Additionally, techniques such as scree plots or cumulative explained variance can aid in visually assessing the impact of different $k$ values on retained variance, aiding in informed decision-making.
	
	\paragraph{Projection onto New Feature Space}
	The final step involves projecting the original data onto the new feature space spanned by the selected principal components. This is achieved by multiplying the original dataset by a transformation matrix composed of the top $k$ eigenvectors. \textbf{Consequently}, the result is a new dataset of reduced dimensionality that maximizes the retained variance from the original data. \textbf{Moreover}, this dimensionality reduction aids in simplifying the subsequent analysis, making it computationally more efficient. \textbf{In addition}, by focusing on the principal components with the highest variance, the transformed dataset maintains the essential information necessary for effective modeling while discarding redundant features. \textbf{Furthermore}, this process enhances interpretability by highlighting the most significant patterns in the data, facilitating a deeper understanding of underlying structures.

	This mathematical framework allows PCA to efficiently reduce the dimensionality of a dataset, simplifying the complexity of data analysis while preserving essential variance and structure. The effectiveness of PCA in revealing the underlying patterns in data makes it a fundamental tool in exploratory data analysis, preprocessing for machine learning models, and any domain requiring dimensionality reduction.

	\subsubsection{Significance of Dimensionality Reduction}
	Dimensionality reduction is a critical process in data analysis and machine learning, addressing challenges associated with high-dimensional data, often referred to as the "curse of dimensionality." Principal Component Analysis stands out as a fundamental technique for dimensionality reduction, offering significant benefits for data exploration, visualization, and subsequent analytical tasks. This section highlights the importance of dimensionality reduction and the role of PCA in this context.
	
	\paragraph{Mitigating the Curse of Dimensionality}
	High-dimensional datasets can be difficult to analyze and visualize due to the exponential increase in volume as the number of dimensions grows. This phenomenon not only complicates data exploration but also poses challenges for machine learning models, including overfitting and increased computational complexity. However, dimensionality reduction through PCA helps mitigate these issues by transforming the original data into a lower-dimensional space that captures the most critical variance and patterns, simplifying the dataset while preserving essential information. By reducing the number of dimensions, PCA can alleviate the burden of computational resources required for processing high-dimensional data, making it more feasible to apply sophisticated machine learning algorithms. Furthermore, PCA aids in interpreting and understanding the underlying structure of the data, enabling researchers to make informed decisions in feature selection and model building. Consequently, PCA serves as a valuable tool in various domains, facilitating more efficient and effective data analysis and machine learning tasks.

	\paragraph{Enhancing Data Visualization}
	One of the most immediate benefits of dimensionality reduction is the improved feasibility of data visualization. High-dimensional data cannot be directly visualized, but by reducing the dimensionality to two or three principal components, PCA enables the creation of comprehensible visual representations. These visualizations can reveal underlying data structures, clusters, and relationships that may not be apparent in the high-dimensional space. Furthermore, they provide a means to effectively communicate complex information to stakeholders, decision-makers, and other non-technical audiences. Moreover, the insights gained from these visualizations can inform subsequent data analysis strategies and model development. Additionally, the ability to visually inspect data in lower dimensions allows for the identification of outliers, anomalies, and patterns that might otherwise go unnoticed. Consequently, PCA plays a crucial role in exploratory data analysis, hypothesis generation, and feature selection, contributing to more informed decision-making processes across various domains.

	\paragraph{Improving Model Performance and Efficiency}
	Dimensionality reduction through PCA can significantly enhance the performance and efficiency of machine learning models. By focusing on the principal components that account for the majority of the variance in the dataset, PCA reduces the risk of overfitting by eliminating noise and redundant features. This streamlined dataset can lead to faster training times, lower computational resource requirements, and potentially better model generalization to unseen data. Moreover, PCA facilitates better visualization of high-dimensional data, allowing for easier interpretation of underlying patterns. Additionally, since PCA transforms the original features into a new orthogonal basis, it can help mitigate multicollinearity issues in regression tasks, thus improving the stability and interpretability of the resulting models. Furthermore, PCA serves as a preprocessing step that simplifies subsequent analyses, enabling more efficient exploration of complex datasets. Hence, incorporating PCA into machine learning pipelines is a prudent strategy to enhance both model performance and computational efficiency.

	\paragraph{Facilitating Feature Engineering and Data Preprocessing}
	PCA also plays a vital role in feature engineering and data preprocessing. By identifying the principal components, PCA effectively uncovers the most informative combinations of the original features, providing a transformed dataset that can serve as a more predictive input for machine learning models. This process can uncover hidden patterns and relationships, enhancing the dataset's suitability for complex analytical tasks. Furthermore, PCA aids in dimensionality reduction, mitigating the curse of dimensionality by selecting only the most relevant features while retaining as much variance as possible. Moreover, PCA can assist in outlier detection and removal by highlighting the dominant directions of variability in the data, making it easier to identify observations that deviate significantly from the norm. Additionally, PCA can be coupled with other preprocessing techniques such as scaling and normalization, ensuring that the data is appropriately prepared for subsequent analysis. Hence, integrating PCA into the feature engineering and data preprocessing pipeline can greatly enhance the efficiency and effectiveness of machine learning workflows.

	In summary, the significance of dimensionality reduction in data analysis cannot be overstated, with PCA serving as a key technique in simplifying high-dimensional data into a more manageable and informative form. Whether for visualization, improving model performance, or aiding in feature engineering, PCA's ability to distill essential information from complex datasets makes it an indispensable tool in the data scientist's arsenal.

	\subsubsection{Standard Applications and Limitations}
	Principal Component Analysis has found extensive applications across diverse fields such as finance, biology, social sciences, and machine learning, showcasing its versatility as a tool for data reduction and analysis. Its primary uses include simplifying the complexity of large datasets for visualization, enhancing feature reduction and noise filtration in machine learning models, uncovering patterns in genetic data within bioinformatics, and optimizing portfolios in finance by analyzing asset return correlations. These applications leverage PCA's ability to reduce dimensionality, thereby revealing underlying data structures, simplifying datasets, and improving computational efficiency.
	
	Despite its broad utility, PCA encounters specific limitations that may affect its applicability to certain tasks. One of the primary challenges is its assumption of linearity, implying that principal components are linear combinations of original features, which may not adequately capture complex, nonlinear relationships in the data. Additionally, PCA's outcomes are sensitive to the scale of features, with variables on larger scales potentially dominating the principal components unless the data undergo normalization to ensure equitable scaling. Another concern is the interpretability of the principal components themselves; these components, being linear combinations of original variables, can sometimes be difficult to interpret, particularly in domains requiring explicit explanatory factors. Lastly, PCA may underperform on sparse datasets, such as those common in text analysis, where the data representation leads to a matrix with a majority of zero elements.
	
	Understanding these limitations is crucial when applying PCA in data science projects. While PCA offers a powerful means of data analysis and preprocessing, acknowledging its constraints is essential for its effective and appropriate use. This nuanced perspective ensures that data scientists can leverage PCA to its fullest potential while being mindful of situations where alternative methods or additional preprocessing steps may be required.
	
	\subsubsection{Algorithmic Pseudocode for PCA}
	The Principal Component Analysis technique offers a refined methodology for parameter estimation in statistical models, particularly when latent variables are involved. It operates by iteratively maximizing variance, utilizing both observed data and latent variables to update parameter estimates. The core mechanics of PCA are elucidated through pseudocode \ref{fig:pca-pseudocode}, illustrating its iterative nature in parameter estimation.
	
	\begin{algorithm}
		\caption{Algorithmic Principal Component Analysis}
		\begin{algorithmic}[1]
			\Procedure{PCA}{$X$}
			\State Standardize the dataset $X$
			\State Compute the covariance matrix $\Sigma$ from the standardized data
			\State Perform eigenvalue decomposition on $\Sigma$ to find eigenvectors and eigenvalues
			\State Sort the eigenvectors by decreasing eigenvalues and choose $k$ eigenvectors with the largest eigenvalues to form a matrix $W$
			\State Project the standardized data onto the space spanned by the top $k$ eigenvectors by computing $X_{\text{pca}} = XW$
			\State \Return $X_{\text{pca}}$
			\EndProcedure
		\end{algorithmic}\label{fig:pca-pseudocode}
	\end{algorithm}

\subsection{Algogenic Enhancements for PCA}
\subsubsection{Dynamic Component Selection Based on Content Analysis}
\paragraph{Rethinking Component Selection in PCA}
The proposed enhancement, Dynamic Component Selection, aims to refine Principal Component Analysis by integrating semantic analysis capabilities of Language Models. This approach moves beyond traditional variance-based selection, introducing a layer of semantic evaluation to identify principal components with significant contextual relevance. 

While PCA traditionally prioritizes variance, this enhancement suggests a nuanced selection process. By incorporating semantic understanding from LLMs, the process is enriched, aiming to retain components that are not just statistically significant but also contextually meaningful. This could potentially offer a deeper insight into the data by identifying components that encapsulate crucial information relevant to specific domains or objectives.

The practical implementation involves LLMs assessing each component for its semantic content, considering the relevance to the predefined analytical goals. This could lead to a dynamic adjustment in the number of components retained or a reevaluation of components based on their contextual alignment. Such a methodology suggests a more informed approach to component selection, potentially enhancing the interpretability and applicability of PCA results.

\paragraph{The Process of Semantic Evaluation}
This enhancement entails a detailed examination of principal components through the lens of LLMs. By analyzing loading patterns and correlating them with domain-specific concepts, a more sophisticated understanding of each component's relevance is achieved. This process may uncover latent factors that traditional variance-focused methods might overlook, offering a pathway to more nuanced and informative PCA outcomes.

The LLM's role is to sift through the components, evaluating their significance based on a broader contextual understanding. This might involve identifying components that resonate with key themes or exhibit patterns critical to the analytical aims. The operational aspect requires careful integration of LLM insights into the PCA process, ensuring that the semantic evaluation informs the final component selection in a meaningful way.

\paragraph{Implementing Semantic-Driven Component Selection}
The practicality of incorporating semantic-driven component selection lies in establishing a feedback loop where LLM analysis directly influences the PCA component selection. This approach requires adjustments in the PCA algorithm to accommodate semantic significance alongside statistical variance, potentially redefining how components are chosen and interpreted.

Such integration highlights a shift towards a more goal-oriented PCA, where components are selected not just for their statistical contribution but for their contextual relevance. This could entail redefining the criteria for component retention or adjusting the PCA process to prioritize dimensions identified as semantically pertinent by the LLMs. The value lies in aligning the PCA results more closely with the analytical objectives, enhancing both the utility and the interpretability of the outcomes.

\paragraph{Challenges in Semantic Component Analysis}
Integrating LLM insights poses several challenges, including computational demands and ensuring the semantic adjustments do not compromise the statistical integrity of PCA. The balance between semantic relevance and statistical significance is crucial, necessitating methodologies that can integrate LLM insights without undermining PCA's fundamental principles.

\subsubsection{Intelligent Eigenvector Adjustment}
\paragraph{Refinement of Eigenvectors with LLM Insights}
Intelligent Eigenvector Adjustment leverages LLMs to refine eigenvectors in PCA, aiming to align them more closely with underlying semantic patterns. This enhancement seeks to address the limitations of traditional PCA by ensuring eigenvectors not only maximize variance but also encapsulate meaningful insights relevant to the data's context.

The implementation involves LLMs analyzing the dataset and initial eigenvectors to recommend adjustments that better reflect critical themes or concepts. This process could lead to a reorientation or recalibration of eigenvectors, making them more representative of semantically significant dimensions. Such adjustments promise a PCA outcome where the derived components offer a deeper understanding of the data, potentially revealing insights that conventional approaches might miss.

\paragraph{Challenges in Implementing Eigenvector Adjustments}
The integration of semantic insights into eigenvector adjustment faces challenges, notably in ensuring the adjustments enhance rather than detract from the PCA's objective to uncover variance-driven patterns. Balancing the introduction of semantic relevance with the preservation of statistical principles is key. Additionally, computational considerations must be managed, ensuring that the process remains feasible even with the added complexity of LLM analysis.

\subsubsection{Adaptive Variance Thresholding for Component Retention}
\paragraph{Optimizing Component Selection via Adaptive Thresholding}
Adaptive Variance Thresholding introduces an LLM-driven approach to dynamically set the variance threshold for component retention in PCA. This method aims to tailor the thresholding process, ensuring that the number of components retained is optimized for both information preservation and relevance to the analytical goals.

The core mechanism involves LLMs evaluating the dataset and PCA outcomes to recommend an adaptive threshold that reflects the data's complexity and the specific analysis objectives. This could result in a more nuanced approach to component retention, potentially enhancing PCA's effectiveness by ensuring that retained components are both statistically significant and contextually informative.

\paragraph{Navigating Thresholding Challenges}
The adaptive thresholding approach must contend with the challenge of accurately determining the optimal threshold that balances information retention with computational efficiency. Ensuring the LLM's recommendations are data-driven and relevant requires sophisticated analysis capabilities and a deep understanding of the dataset's semantics. Additionally, the integration of these recommendations into the PCA workflow must be handled with care to maintain the analytical integrity of the process.

\subsubsection{Semantic Noise Filtering in Data Preprocessing}
\paragraph{Refining PCA Through Enhanced Preprocessing}
Semantic Noise Filtering targets the preprocessing phase of PCA, employing LLMs to identify and eliminate semantic noise. This approach aims to cleanse the data of irrelevant or misleading information, ensuring that PCA focuses on data elements that are truly significant to the analysis.

The operational framework involves LLMs analyzing the data to distinguish between valuable information and semantic noise. Recommendations for data cleansing could involve excluding irrelevant features or adjusting data representations to better reflect the semantic content. This preprocessing enhancement promises a PCA outcome that is more accurate and reflective of the data's true structure, potentially leading to insights that traditional preprocessing methods might overlook.

\paragraph{Addressing Preprocessing Challenges}
The challenge in implementing Semantic Noise Filtering lies in accurately identifying semantic noise without discarding potentially valuable data. Balancing the LLM's computational demands with the preprocessing objectives is also crucial, as is ensuring that the filtering actions are transparent and understandable to practitioners. Developing robust methodologies to leverage LLM insights effectively in preprocessing will be key to overcoming these challenges.

\subsubsection{Context-Aware Scaling and Transformation}
\paragraph{Enhancing Data Preparation for PCA}
Context-Aware Scaling and Transformation refines PCA's data preparation phase by incorporating semantic insights from LLMs. This approach aims to tailor scaling and transformation processes to the dataset's context and the specific goals of PCA, ensuring that the variance captured reflects meaningful patterns relevant to the analysis.

The practical implementation involves using LLMs to identify appropriate scaling and transformation techniques that align with the data's semantic characteristics. This could lead to a more targeted preprocessing approach, where adjustments are made to optimize the representation of data for PCA. Such a strategy promises to enhance the interpretability and relevance of PCA results, offering a more nuanced understanding of the data's underlying structure.

\paragraph{Navigating Data Preparation Challenges}
Integrating context-aware methodologies into PCA's data preparation faces challenges, particularly in accurately implementing LLM recommendations and managing the computational complexity of the process. Balancing semantic relevance with statistical considerations is crucial to ensuring that the preprocessing enhancements contribute positively to the PCA outcomes.

	\subsubsection{Pseudocode for Algogenic PCA}
	The Algogenic PCA approach employs AI to enhance conventional PCA techniques by dynamically adjusting principal component analysis parameters and strategies in response to observed system behavior and real-time error estimates. This pseudocode, provided in \ref{fig:pca-Algogen-pseudocode}, delineates a sophisticated framework integrating AI-driven improvements for adaptive component selection, dimensionality reduction, criteria for component acceptance, and real-time parameter optimization.
	
	\begin{algorithm}
		\caption{Algogenic Principal Component Analysis with Enhancements}
		\begin{algorithmic}[1]
			\Procedure{EnhancedPCA}{data}
			\State Analyze data and objectives using LLM to provide Contextual Preprocessing Recommendations
			\State Apply recommended preprocessing steps to data
			\State Perform standard PCA on preprocessed data
			\State Utilize LLM for Adaptive Variance Thresholding to determine optimal components
			\State Adjust eigenvectors with Intelligent Eigenvector Adjustment based on LLM insights
			\State Filter semantic noise from data using LLM recommendations
			\State Apply Context-Aware Scaling and Transformation as suggested by LLM
			\For{each principal component}
			\State Perform Semantic Feature Interpretation for enhanced understanding
			\State Conduct Automated Feature Correlation Analysis to elucidate feature contributions
			\EndFor
			\State Generate Intelligent Post-PCA Application Guidance for effective use of PCA results
			\State \Return Enhanced PCA components and guidance for application
			\EndProcedure
		\end{algorithmic}\label{fig:pca-Algogen-pseudocode}
	\end{algorithm}

	\begin{figure}
		\centering
		\includegraphics[width=0.8\textwidth]{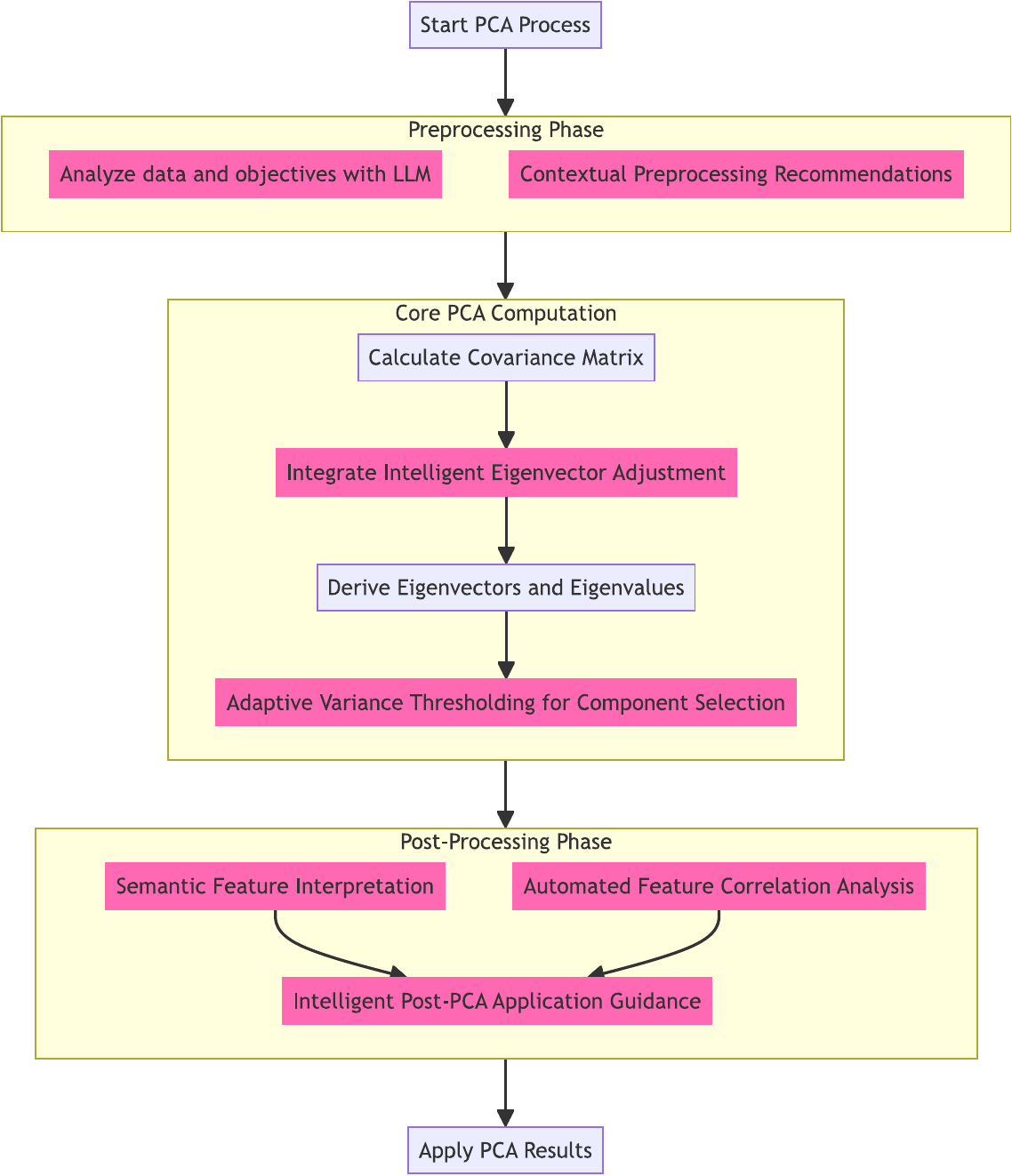} 
		\caption{Enhancing PCA with Algogenic Insights: This diagram visualizes the integration of generative AI (Algogens) within the Principal Component Analysis workflow, segmented into preprocessing, core computation, and post-processing phases. The preprocessing phase utilizes Language Models for data analysis and contextual recommendations, ensuring data is optimally prepared. The core PCA computation phase is highlighted by the calculation of covariance matrices, intelligent eigenvector adjustments, and adaptive variance thresholding, directly embedding Algogenic enhancements into the PCA algorithm to capture semantically significant patterns. The post-processing phase leverages semantic feature interpretation and automated feature correlation analysis, culminating in intelligent guidance for applying PCA results. This framework illustrates a holistic approach to PCA, significantly enriching the algorithm's capability to provide deep, actionable insights across various data analysis contexts.}
		\label{fig:pca}
	\end{figure}

	
	\chapterimage{pngs/machine_learning.png} 
	\chapter{Machine Learning Algogens}\index{Machine Learning Algogens}
	
	\section{$K$-Means Clustering}\index{K-Means Clustering}
	\subsection{Introduction to $K$-Means}
	\subsubsection{The Concept of $K$-Means Clustering}
	
	\paragraph{Foundational Overview}
	$K$-Means Clustering represents a cornerstone in unsupervised machine learning, where the primary objective is to partition $n$ observations into $K$ clusters based on the nearest mean. Each observation is assigned to the cluster with the closest centroid, which is the cluster's mean. The process iteratively refines the positions of centroids to minimize the within-cluster variances, also known as the sum of squared distances between each point and the centroid of its cluster. This iterative refinement continues until the centroids stabilize, indicating that the clusters are as compact and distinct as possible given the initial conditions. Furthermore, this method is computationally efficient, making it suitable for large datasets. Moreover, its simplicity and intuitive nature make it a popular choice for various applications, ranging from customer segmentation to image compression. Additionally, the algorithm's deterministic nature ensures reproducibility, a crucial aspect in scientific research. Thus, $K$-Means Clustering stands as a versatile and robust tool in exploratory data analysis and pattern recognition tasks.

	\paragraph{Mathematical Formulation}
	The mathematical essence of $K$-Means is captured by the objective to minimize the within-cluster sum of squares $\left(WCSS\right)$, which is formulated as:
	\[
	WCSS = \sum_{i=1}^{K} \sum_{x \in S_i} ||x - \mu_i||^2
	\]
	
	where $K$ is the number of clusters, $S_i$ is the set of observations in the $i$th cluster, $x$ represents an observation, and $\mu_i$ is the centroid of $S_i$. The algorithm starts with an initial guess for the centroids, which can significantly influence the outcome. Subsequent iterations adjust the centroids to reduce the WCSS, with each observation reassigned to the cluster whose centroid is nearest. The process converges when assignments no longer change, indicating the algorithm has found a locally optimal partition of the data.
	
	\paragraph{Algorithmic Steps}
	The operational steps of $K$-Means clustering are straightforward yet powerful. Initially, $K$ centroids are either chosen randomly from the dataset or placed using more sophisticated heuristic methods to ensure a diverse starting point. This initial centroid placement is crucial as it can significantly impact the convergence and final clustering result. The algorithm then alternates between two main steps: assignment and update. In the assignment step, each observation is assigned to the nearest centroid's cluster based on a distance metric such as Euclidean distance. This step aims to minimize the intra-cluster variance, ensuring that observations are grouped with similar centroids. Following the assignment step, the update phase recalculates each centroid's position as the mean of all observations assigned to its cluster. This centroid update ensures that the centroids better represent the center of their respective clusters, iterating towards convergence. This alternation between assignment and update phases constitutes the core iterative process of $K$-Means, converging towards a stable clustering solution.

	\paragraph{Applications and Versatility}
	$K$-Means Clustering is widely applied across a range of disciplines, from market segmentation and image compression to document clustering and anomaly detection. Its popularity stems from its simplicity, efficiency, and the intuitive appeal of its results. Additionally, the algorithm's scalability makes it suitable for large datasets, while its straightforward implementation allows for easy integration into various systems. Moreover, $K$-Means is robust to noise and can handle high-dimensional data, making it versatile in real-world scenarios. However, the choice of $K$ and the algorithm's sensitivity to initial centroid placement are critical factors that can affect the quality of the clustering outcome. Consequently, careful consideration and domain expertise are required to determine the appropriate value of $K$ and to mitigate the impact of initialization on the final clusters. Thus, despite its widespread usage and advantages, practitioners must be cautious and methodical in their approach to ensure the effectiveness of $K$-Means Clustering in diverse applications.
	
	\paragraph{Challenges and Considerations}
	While $K$-Means is a powerful tool for pattern discovery, it is not without its challenges. The algorithm assumes clusters are convex and isotropic, which may not hold for all datasets, leading to less meaningful clusters. Additionally, the need to specify $K$ a priori and the algorithm's sensitivity to outliers are considerations that practitioners must navigate. However, despite these challenges, $K$-Means remains a fundamental technique in the data scientist's toolkit. Its simplicity allows for straightforward implementation and interpretation, making it accessible even to those new to clustering algorithms. Furthermore, its computational efficiency enables the analysis of large datasets in a reasonable amount of time, making it suitable for various applications across different domains. Moreover, with proper preprocessing techniques and careful consideration of the dataset's characteristics, many of the algorithm's limitations can be mitigated, enhancing its effectiveness in uncovering hidden structures in data.

	\subsubsection{Key Principles and Mechanisms}
	
	\paragraph{Core Principles}
	At the heart of $K$-Means Clustering lie several key principles that dictate its operation and effectiveness. The algorithm operates on the premise of minimizing the variance within each cluster, a measure that is intrinsically tied to the concept of Euclidean distance in the feature space. This minimization is achieved through an iterative process of assignment and optimization, where data points are grouped based on their proximity to the nearest centroid, and centroids are recalculated to best represent the mean of the assigned points. The simplicity of this approach, relying on distance as the primary metric for cluster formation, underpins the algorithm's widespread applicability and robustness. Furthermore, the iterative nature of the algorithm allows for adaptability to different datasets and convergence towards stable cluster configurations. Moreover, despite its simplicity, $K$-Means can effectively handle large datasets efficiently due to its computational efficiency, making it a popular choice for various clustering tasks in data mining and machine learning applications.
	
	\paragraph{Mechanisms of Clustering}
	The clustering mechanism commences by selecting $K$ initial centroids, a pivotal step in initializing the clustering process. Following this, each data point is meticulously assigned to the closest centroid, thereby forming preliminary clusters based on proximity. This assignment step lays the groundwork for subsequent iterations. After the initial assignment, the centroids undergo a recalibration process to better represent the current cluster compositions. Typically, this involves computing the mean of all points within each cluster. These updated centroids then guide the next round of assignments, perpetuating a cyclical refinement process. The algorithm iterates until convergence, where subsequent iterations yield minimal changes to centroid positions, signifying the attainment of a locally optimal clustering solution. This iterative refinement process encapsulates the essence of clustering algorithms, emphasizing both the dynamic evolution of cluster compositions and the gradual convergence towards an optimal configuration.
	
	\paragraph{Distance Metrics and Variance Reduction}
	The choice of distance metric, typically the Euclidean distance, plays a crucial role in how clusters are formed. The Euclidean distance between a point and a centroid is a direct measure of their dissimilarity, and minimizing this distance across all points in a cluster effectively minimizes the cluster's variance. \textbf{Moreover}, the sum of these minimized variances across all clusters constitutes the objective function that $K$-Means seeks to minimize. This focus on variance reduction is both a strength and a limitation, as it assumes that clusters are spherical and evenly sized, which may not accurately reflect the underlying structure of all datasets. \textbf{However}, despite this limitation, $K$-Means remains widely used due to its simplicity and computational efficiency. \textbf{Furthermore}, exploring alternative distance metrics such as Mahalanobis distance or cosine similarity can address some of these limitations by accommodating different data distributions and cluster shapes, \textbf{thus} enhancing the robustness and accuracy of clustering algorithms.

	\paragraph{Iterative Optimization}
	The iterative optimization process of $K$-Means is designed to gradually improve the clustering outcome with each cycle of assignments and updates. This process is emblematic of the Expectation-Maximization (EM) approach. \textbf{Moreover}, each iteration consists of an expectation step (assigning points to the nearest centroid) followed by a maximization step (recalculating centroids to minimize variance). The algorithm's efficiency stems from its ability to make substantial improvements to the clustering quality in the initial iterations, \textbf{while} experiencing diminishing returns as it approaches convergence. \textbf{In addition}, the iterative nature of the optimization \textbf{furthermore} allows for the exploration of various cluster configurations, thus enabling the algorithm to potentially escape local optima and find better solutions. \textbf{Furthermore}, the iterative refinement process \textbf{also} facilitates adaptability to datasets with varying complexities, as it can dynamically adjust cluster centroids to better fit the data distribution. Overall, the iterative optimization mechanism of $K$-Means plays a crucial role in achieving robust and accurate clustering results.
	
	\paragraph{Convergence Criteria and Algorithmic Complexity}
	$K$-Means converges to a solution when the centroids stabilize, meaning their positions do not change significantly between iterations, or when the decrease in the objective function falls below a predefined threshold. The speed of convergence and the algorithm's overall computational complexity are influenced by the choice of initial centroids, the dataset's characteristics, and the value of $K$. While $K$-Means is generally considered efficient for a wide range of applications, its performance can be affected by the curse of dimensionality and the need to run multiple initializations to escape local minima and find a more globally optimal clustering solution. Furthermore, the efficiency of $K$-Means may vary depending on the distribution of the data and the geometry of the clusters. Moreover, the algorithm's scalability can become an issue for very large datasets, as the computational cost grows linearly with the number of data points and the number of clusters. Therefore, careful consideration of these factors is crucial when applying $K$-Means in practice.

	\subsubsection{Choosing the Number of Clusters}
	
	\paragraph{The Significance of $K$}
	The selection of the optimal number of clusters, denoted by $K$, is a pivotal decision in the application of the $K$-Means Clustering algorithm. This choice directly influences the granularity of the clustering outcome, impacting both the interpretability and utility of the results. An appropriately chosen $K$ can reveal meaningful patterns and distinctions within the data. Moreover, it enhances the ability to extract actionable insights and make informed decisions based on the clustered data. Conversely, an ill-suited $K$ may lead to detrimental outcomes. For instance, if $K$ is too small, it may oversimplify the complexity of the data, blurring the boundaries between distinct groups and failing to capture subtle variations. On the other hand, if $K$ is excessively large, it might result in overfitting, dividing the data into too many clusters and thereby obscuring underlying trends. Thus, the careful selection of $K$ is crucial for obtaining reliable and informative clustering results.
	
	\paragraph{Methodologies for Determining $K$}
	Several methodologies have been developed to assist practitioners in selecting an optimal $K$. One widely recognized approach is the Elbow Method, which involves plotting the within-cluster sum of squares (WCSS) against the number of clusters and identifying the point where the rate of decrease sharply changes, resembling an "elbow". This point is considered to be indicative of the optimal $K$. Another approach is the Silhouette Method, which measures how similar an object is to its own cluster compared to other clusters. The Silhouette Score provides insight into the cohesion and separation of the formed clusters, with higher scores indicating a more appropriate $K$. Additionally, the Gap Statistic compares the total within intra-cluster variation for different values of $K$ with their expected values under null reference distribution of the data.
	
	\paragraph{Challenges in $K$ Selection}
	Despite these methodologies, determining the optimal number of clusters remains a non-trivial challenge, often requiring domain knowledge and iterative exploration. The inherent subjectivity in interpreting the Elbow or Silhouette plots can lead to different conclusions about the best $K$. Moreover, the assumption of spherical clusters in $K$-Means further complicates the selection process, as real-world data may not conform to this geometric arrangement, leading to potential misrepresentation of the true data structure. Additionally, the utilization of the Silhouette coefficient, though informative, can be limited in cases where clusters exhibit irregular shapes or varying densities. Furthermore, the curse of dimensionality exacerbates the challenge, as higher-dimensional data spaces can obscure meaningful cluster boundaries, necessitating dimensionality reduction techniques prior to $K$ selection. Hence, despite the availability of various methodologies, the selection of an appropriate $K$ value remains an intricate task fraught with challenges and nuances.
	
	\paragraph{Impact of $K$ on Clustering Quality}
	The choice of $K$ significantly affects the clustering quality. Too few clusters can result in overly broad groupings that fail to capture important distinctions within the data. Conversely, too many clusters may lead to overfitting, where noise in the data is mistaken for genuine cluster structure. This balance between underfitting and overfitting is critical to achieving meaningful, actionable clustering outcomes. Furthermore, selecting an appropriate value for $K$ requires careful consideration of the specific characteristics of the dataset. Moreover, the determination of $K$ often involves iterative processes, such as cross-validation or silhouette analysis, to identify the optimal number of clusters. Additionally, the impact of $K$ extends beyond the clustering algorithm itself, influencing downstream tasks such as classification or anomaly detection. Hence, thorough experimentation and validation are necessary to ensure the chosen $K$ yields robust and interpretable clusters. 
	
	\paragraph{Adaptive and Heuristic Approaches}
	In response to these challenges, adaptive and heuristic approaches have been proposed to automate or assist in the selection of $K$. These methods aim to balance statistical criteria with computational efficiency, often incorporating machine learning techniques to evaluate potential cluster configurations dynamically. \textbf{However}, despite advancements in these areas, the selection of $K$ remains a fundamental step that requires careful consideration, experimentation, and validation to ensure the clustering results are both meaningful and aligned with the analytical objectives. \textbf{Moreover}, the complexity of the data and the specific characteristics of the problem domain \textbf{furthermore} underline the necessity of adopting flexible strategies that can adapt to diverse scenarios. \textbf{Consequently}, researchers continue to explore innovative algorithms and methodologies to enhance the robustness and reliability of the clustering process. \textbf{Furthermore}, ongoing efforts focus on developing scalable solutions that can handle large-scale datasets efficiently while maintaining the quality of the clustering outcomes. \textbf{In addition}, the integration of domain knowledge \textbf{is crucial} in guiding the selection process, ensuring that the chosen $K$ reflects meaningful patterns inherent in the data. Overall, the synergy between adaptive techniques, heuristic approaches, and domain expertise \textbf{is pivotal} in advancing the state-of-the-art in cluster analysis.

	\subsubsection{Applications and Limitations}
	
	\paragraph{Diverse Applications Across Fields}
	The $K$-Means Clustering algorithm finds its utility in a broad array of applications across various fields due to its simplicity and efficiency. In marketing, it aids in customer segmentation by identifying groups with similar preferences or behaviors, enabling targeted marketing strategies. In bioinformatics, $K$-Means is used to classify genes with similar expression patterns, providing insights into gene function and regulation. Urban planning benefits from $K$-Means by clustering areas with similar land use, facilitating urban development and zoning decisions. Additionally, it plays a crucial role in image segmentation, dividing digital images into distinct segments to simplify their analysis, and in document clustering, grouping documents with similar topics for more efficient information retrieval.
	
	\paragraph{Limitations and Considerations}
	Despite its widespread application, $K$-Means faces several limitations that affect its performance and applicability. The requirement to specify the number of clusters, $K$, a priori, poses a significant challenge, as the optimal number is often not known in advance and can significantly impact the results. $K$-Means assumes that clusters are spherical and of similar size, which may not hold true for all datasets, leading to poor performance with elongated or irregularly shaped clusters. The algorithm's sensitivity to the initial placement of centroids can result in convergence to local minima, necessitating multiple runs with different initializations to achieve a satisfactory outcome. Additionally, $K$-Means is sensitive to outliers, as they can disproportionately influence the calculation of centroids, skewing the results.
	
	\paragraph{Overcoming Limitations}
	Various strategies have been employed to address the limitations of $K$-Means. Techniques such as $K$-Means++ offer an improved method for initial centroid placement, reducing the likelihood of poor convergence. The use of pre-processing steps to remove outliers and normalize data can mitigate the algorithm's sensitivity to noise and scale. Incorporating domain knowledge or employing more sophisticated methods to determine the optimal $K$ can alleviate the challenge of selecting the number of clusters. Furthermore, integrating $K$-Means with other clustering techniques or adopting a hybrid approach can enhance its ability to uncover complex patterns in data.
	
	\paragraph{Future Directions and Enhancements}
	The ongoing evolution of $K$-Means includes research into making the algorithm more robust to outliers, more flexible in identifying clusters of varying shapes and sizes, and more autonomous in determining the optimal number of clusters. Developments in machine learning and data mining continue to expand the applications of $K$-Means, pushing the boundaries of its capabilities. The exploration of Algogenic enhancements, incorporating generative AI to dynamically adjust clustering parameters and interpret complex data structures, represents a promising frontier for extending the utility and applicability of $K$-Means in the era of big data.
	
	\paragraph{Concluding Remarks}
	$K$-Means Clustering remains a fundamental tool in unsupervised learning, valued for its simplicity, efficiency, and versatility. Moreover, its intuitive approach makes it accessible to a wide range of users, from novice practitioners to seasoned data scientists. Furthermore, its ability to handle large datasets efficiently makes it particularly attractive in today's era of big data. However, despite its strengths, $K$-Means is not without limitations. One notable challenge is its sensitivity to initial centroid selection, which can lead to suboptimal clustering results. Nevertheless, ongoing research efforts aim to address these challenges, with promising developments in initialization strategies and optimization techniques. Additionally, the interpretability of $K$-Means clusters enhances its utility in exploratory data analysis and pattern recognition tasks. In contrast, more complex clustering algorithms may offer higher accuracy but often at the expense of interpretability and computational overhead. Therefore, while $K$-Means may not always yield the most precise clustering solution, its balance of simplicity, efficiency, and interpretability ensures its enduring relevance across diverse domains.

	\subsubsection{Algorithmic Pseudocode for $K$-Means Clustering}
	The $K$-Means Algorithm is a powerful method used for clustering data into $K$ distinct groups based on similarities between data points. Unlike Expectation Maximization (EM), which deals with latent variables, $K$-Means focuses solely on observed data. It iteratively assigns data points to the nearest cluster centroid and updates the centroids to minimize the within-cluster variance. This process continues until convergence, resulting in well-defined clusters. For a detailed algorithmic representation, refer to the pseudocode in Figure \ref{fig:kmeans-pseudocode}.

	\begin{algorithm}
		\caption{$K$-Means Clustering Pseudocode}
		\begin{algorithmic}[1]
			\Procedure{$K$-Means}{Data, $K$}
			\State Choose $K$ initial centroids from the dataset (randomly or by a heuristic)
			\State Initialize cluster assignments for each data point to null
			\While{centroids do not converge}
			\For{each data point in the dataset}
			\State Assign the data point to the nearest centroid
			\EndFor
			\For{each centroid}
			\State Recalculate the position of the centroid as the mean of all data points assigned to it
			\EndFor
			\State Check for convergence (no change in centroid positions or minimal change within a threshold)
			\EndWhile
			\State \Return Clusters and their centroids
			\EndProcedure
		\end{algorithmic}\label{fig:kmeans-pseudocode}
	\end{algorithm}

\paragraph{Enhanced Initial Cluster Center Selection}
The refinement of the initialization process of the K-Means algorithm, as discussed in a 2021 study \cite{rahman2021enhanced}, represents progress in clustering techniques. The study proposes a method to address limitations associated with random and heuristic-based initial center selections, aiming to mitigate inefficiencies in clustering performance such as increased convergence iterations and suboptimal cluster formations. By leveraging geometric insights into the dataset, this method enhances the algorithm's efficiency by strategically placing initial cluster centers. This approach accelerates the convergence process and enhances clustering quality, contributing to advancements in data clustering methodologies.

\paragraph{Hybridization with Improved Firefly Algorithm}
In 2023, an approach to automatic cluster determination in high-dimensional datasets was introduced through the hybridization of the K-Means algorithm with an improved version of the firefly algorithm \cite{alam2023hybridization}. This method addresses the challenge of determining the number of clusters a priori, which is crucial in analyzing high-dimensional data. Inspired by the communication behavior of fireflies, the firefly algorithm is adapted to optimize the selection of cluster centers and dynamically determine the number of clusters. This hybrid approach enhances the adaptability and accuracy of the K-Means algorithm in handling complex datasets, demonstrating the potential of integrating traditional clustering methods with nature-inspired algorithms.

	\subsection{Algogenic Enhancements for K-Means}
	\subsubsection{Semantic Feature Engineering for Preprocessing}
	\paragraph{Introduction to Semantic Feature Engineering}
	Enhancing the K-Means clustering algorithm through Semantic Feature Engineering involves the strategic application of large language models for advanced data preprocessing. This technique capitalizes on LLMs' profound linguistic understanding to generate semantic-rich features from the dataset. Unlike traditional numerical features, these semantic features encapsulate the deeper contextual meanings inherent in the data, offering a more meaningful basis for clustering. This approach not only augments the traditional feature extraction methods by identifying complex relationships within the data but also significantly improves clustering interpretability and outcomes.
	
	\paragraph{The Process of Semantic Feature Engineering}
	The process commences with LLMs analyzing the dataset to uncover latent semantic structures, which are not readily apparent through conventional preprocessing methods. LLMs adeptly transform these insights into a semantically enriched vector space, facilitating a nuanced representation of the dataset. This approach ensures the initial grouping in K-Means is informed by the inherent semantics of the data, leading to more meaningful clusters that reflect the dataset's true underlying structure.
	
	\paragraph{Impact on K-Means Clustering}
	Integrating semantic features into the K-Means algorithm enhances its ability to form clusters that are not only numerically cohesive but also semantically meaningful. This approach significantly refines the clustering process, enabling the identification of patterns and relationships that transcend mere numerical similarities. By incorporating semantic understanding into the clustering, K-Means becomes more adept at interpreting complex datasets, thereby expanding its applicability and effectiveness in data-driven decision-making across various domains.
	
	\subsubsection{Dynamic Cluster Initialization}
	\paragraph{The Need for Dynamic Cluster Initialization}
	Dynamic Cluster Initialization enhances the K-Means algorithm by utilizing LLMs to determine optimal initial centroids based on the dataset's semantic structure. This method addresses the traditional shortcomings of random or heuristic-based initializations, enabling a more informed and effective starting point for clustering. By grounding the initialization process in the dataset's intrinsic semantic richness, this Algogenic enhancement promises to improve clustering accuracy and efficiency.
	
	\paragraph{Implementing Dynamic Cluster Initialization}
	Through LLM analysis, this method dynamically identifies initial centroids that are representative of the dataset's varied semantic landscapes. This nuanced approach allows K-Means to better navigate the clustering process, optimizing for both cohesion and separation from the outset. The dynamic initialization not only streamlines the algorithm's convergence but also enhances its robustness, offering a solid foundation for subsequent clustering iterations.
	
	\paragraph{Benefits to the K-Means Clustering Process}
	The incorporation of dynamic cluster initialization provides a significant boost to the K-Means algorithm by facilitating more precise and semantically coherent clusters. This method enhances the algorithm's scalability and adaptability, particularly in handling complex datasets. By offering a more strategic approach to centroid selection, K-Means is positioned to deliver more accurate and interpretable clustering results, thereby enriching its utility in practical applications.
	
	\subsubsection{Adaptive Clustering Criteria}
	\paragraph{Rationale Behind Adaptive Clustering Criteria}
	Adaptive Clustering Criteria introduce a flexible, data-driven approach to refining the K-Means algorithm, enabling the adjustment of clustering criteria based on the evolving characteristics of the dataset. Leveraging LLMs, this enhancement allows the algorithm to dynamically tailor its distance metrics, accommodating the complex and nuanced relationships among data points. This adaptability overcomes the limitations of static metrics, ensuring that clustering decisions are informed by the most relevant features of the data.
	
	\paragraph{Implementation of Adaptive Clustering Criteria}
	The implementation involves LLMs continuously evaluating and adjusting the clustering criteria to better capture the dataset's semantic and structural nuances. This process ensures that the K-Means algorithm remains aligned with the data's inherent complexities, leading to more accurate and meaningful clusters. The adaptive criteria adapt to changes in the dataset, maintaining the algorithm's efficacy across diverse and dynamic data landscapes.
	
	\paragraph{Impact on the K-Means Algorithm}
	Integrating adaptive clustering criteria significantly enhances the flexibility and precision of the K-Means algorithm. This Algogenic enhancement ensures that the clustering process is not only responsive to the data's immediate features but also adaptable to its evolving dynamics. By fostering a more nuanced and context-aware clustering approach, K-Means becomes more capable of uncovering intricate patterns, offering a robust tool for complex data analysis tasks.
	
	\subsubsection{Cluster Refinement with Generative Modeling}
	\paragraph{Introduction to Cluster Refinement through Generative Modeling}
	Cluster Refinement with Generative Modeling represents a sophisticated Algogenic enhancement that employs LLMs to refine K-Means clustering results. By generating synthetic data points, this method improves cluster cohesion and boundary definition, particularly beneficial for datasets with ambiguous or poorly defined clusters. This innovative approach leverages generative capabilities to address gaps in clusters, enhancing both the precision and interpretability of clustering outcomes.
	
	\paragraph{How Generative Modeling Enhances Cluster Refinement}
	Through LLM-generated synthetic data, this process strategically enhances cluster characteristics, addressing issues of sparsity and disjointedness. The careful calibration of synthetic data ensures clusters are more defined and coherent, facilitating a more granified understanding of the dataset's structure. This method not only enriches the clustering process but also maintains the dataset's integrity, avoiding skewing or bias introduction.
	
	\paragraph{Unique Advantages for K-Means Clustering}
	The integration of Cluster Refinement with Generative Modeling provides distinct benefits to K-Means, notably in improving cluster accuracy and robustness to outliers. This enhancement is particularly advantageous for complex or imbalanced datasets, where traditional clustering may falter. Although introducing additional computational complexity, the resulting gains in cluster quality and interpretability justify this approach, marking a significant advancement in the application of K-Means across diverse data scenarios.
	
	\subsubsection{Semantic Cluster Interpretation}
	\paragraph{The Importance of Semantic Cluster Interpretation}
	Semantic Cluster Interpretation enhances the K-Means algorithm by providing a deeper, semantic-based understanding of clustering results. This Algogenic enhancement leverages LLMs to translate numerical clusters into comprehensible, meaningful groupings. By elucidating the semantic relationships within clusters, this approach significantly aids in the interpretation and application of clustering results, bridging the gap between data analysis and actionable insights.
	
	\paragraph{Implementing Semantic Cluster Interpretation}
	The implementation leverages LLMs to analyze clusters, extracting themes, patterns, and outliers based on semantic similarities. This process not only enhances the coherence and relevance of clusters but also provides a framework for identifying and addressing anomalies. By incorporating semantic insights, K-Means transcends numerical analysis, offering a more nuanced and actionable understanding of clustered data.
	
	\paragraph{Enhancing K-Means Clustering with Semantic Insight}
	Semantic Cluster Interpretation significantly augments the K-Means algorithm, facilitating the extraction of actionable insights from clustered data. This enhancement is invaluable in applications requiring a deep understanding of data groupings, such as market segmentation and text analysis. By integrating semantic interpretation, K-Means is transformed into a more comprehensive tool for uncovering and leveraging hidden data patterns.
	
	\subsubsection{Cluster Optimization Feedback Loop}
	\paragraph{The Concept of Cluster Optimization Feedback Loop}
	The Cluster Optimization Feedback Loop is a sophisticated Algogenic enhancement that iteratively refines K-Means clustering through LLM-evaluated feedback. This process dynamically adjusts clustering parameters, aligning the algorithm more closely with specific application goals and data characteristics. By integrating external knowledge and domain-specific insights, this feedback loop ensures continuous improvement and relevance of clustering outcomes, addressing traditional K-Means limitations and fostering adaptability.
	
	\paragraph{Implementing the Feedback Loop}
	This implementation involves LLMs providing targeted feedback on clustering outcomes, guiding adjustments to parameters such as cluster count and initialization. By evaluating cluster cohesion and separation against domain-specific benchmarks, LLMs facilitate a recursive refinement process. This approach not only enhances the algorithm's accuracy but also its responsiveness to evolving data and application needs.
	
	\paragraph{Unique Benefits to the K-Means Algorithm}
	Integrating a Cluster Optimization Feedback Loop offers unique advantages to the K-Means algorithm, transforming it into a dynamic, continuously improving system. This enhancement ensures that clustering outcomes remain relevant and effective, adapting to both data nuances and application requirements. By leveraging LLMs for informed feedback, K-Means is rendered more adaptable and effective, marking a significant advancement in the application of clustering techniques across various domains.

	\subsubsection{Pseudocode for Algogenic K-Means}
	The Algogenic $K$-Means approach harnesses AI to enhance conventional $K$-Means clustering methods by dynamically adjusting clustering parameters and strategies according to the observed behavior of the system and real-time error estimates. This pseudocode, available in \ref{fig:kmeans-Algogen-pseudocode}, delineates a sophisticated framework that integrates AI-driven improvements for adaptive cluster center initialization, point assignment to clusters, convergence criteria, and real-time parameter optimization.
	\begin{algorithm}
		\caption{Algogenic K-Means Pseudocode}
		\begin{algorithmic}[1]
			\Procedure{AlgogenicKMeans}{Data, K}
			\State SemanticFeatures $\gets$ GenerateSemanticFeatures(Data) \Comment{Semantic feature engineering}
			\State InitialCentroids $\gets$ DynamicClusterInitialization(SemanticFeatures, K) \Comment{Dynamic initialization}
			\State Clusters $\gets$ AssignPointsToClusters(SemanticFeatures, InitialCentroids)
			\While{not Converged}
			\State Centroids $\gets$ UpdateCentroids(Clusters)
			\State Clusters $\gets$ AssignPointsToClusters(SemanticFeatures, Centroids)
			\State Clusters, Centroids $\gets$ RefineClustersWithGenerativeModeling(Clusters, Centroids) \Comment{Cluster refinement}
			\State UpdateCriteria $\gets$ AdaptiveClusteringCriteria(Clusters) \Comment{Adapt criteria}
			\If{UpdateCriteria}
			\State Clusters $\gets$ ReassignPoints(Clusters, NewCriteria)
			\EndIf
			\EndWhile
			\State SemanticInterpretation(Clusters) \Comment{Semantic interpretation}
			\State Clusters $\gets$ ClusterOptimizationFeedbackLoop(Clusters) \Comment{Optimization loop}
			\State \Return Clusters
			\EndProcedure
		\end{algorithmic}\label{fig:kmeans-Algogen-pseudocode}
	\end{algorithm}

	\begin{figure}
		\centering
		\includegraphics[width=0.6\textwidth]{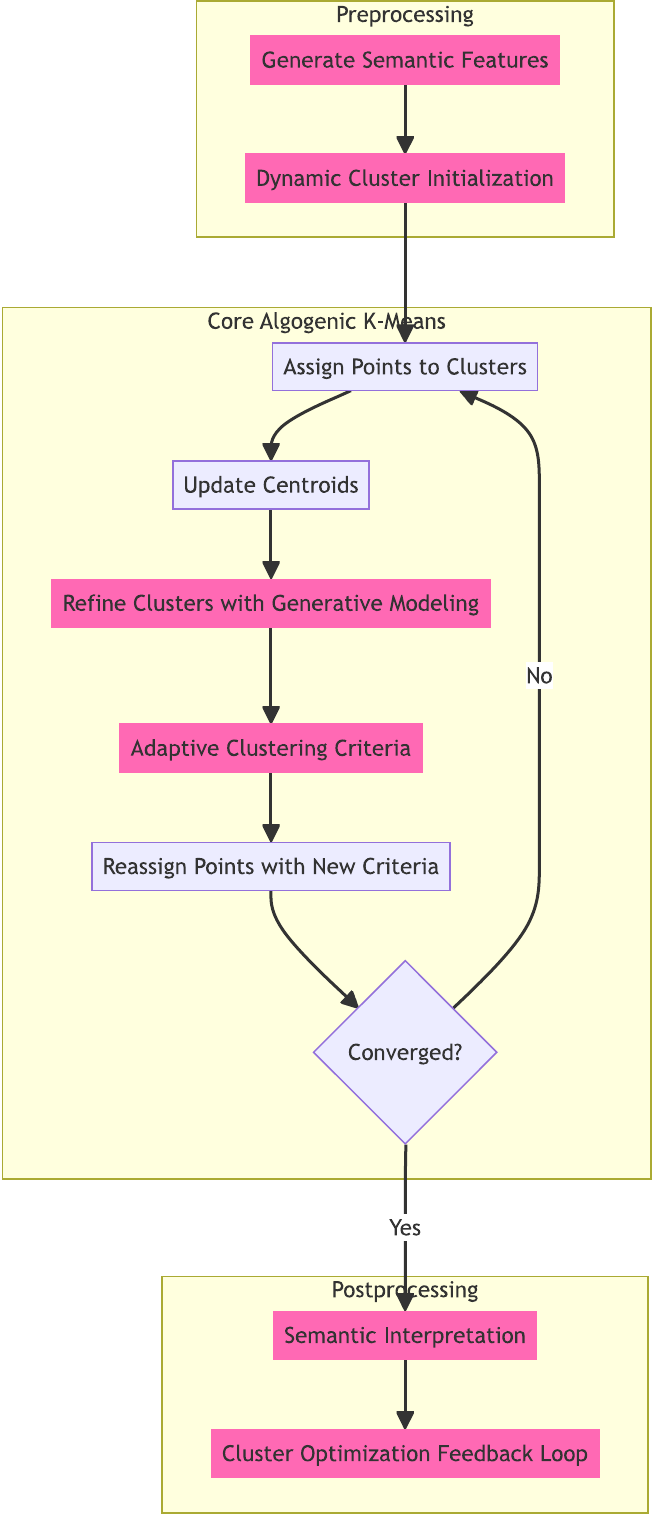} 
		\caption{Algogenic Enhancements for K-Means Clustering: This diagram outlines the Algogenic framework for the K-Means algorithm, divided into preprocessing, core, and postprocessing phases. The preprocessing phase introduces generative AI for generating semantic features and dynamically initializing clusters. In the core phase, an iterative process of cluster assignment and refinement is depicted, where both algorithmic steps and generative AI enhancements, such as adaptive clustering criteria and cluster refinement with generative modeling, are integrated. The postprocessing phase emphasizes the semantic interpretation of clusters and includes a feedback loop for optimization, highlighting how Algogenic K-Means enhances clustering through accuracy, efficiency, and making the clustering results interpretable and actionable.}
		\label{fig:kmeans}
	\end{figure}

	\section{Support Vector Machines}\index{Support Vector Machines}
	
	\subsection{Introduction to SVMs}
	\paragraph{The Concept of Support Vector Machines}
	Support Vector Machines (SVMs) represent a powerful and versatile class of supervised learning algorithms used for classification, regression, and outlier detection tasks. At their core, SVMs are based on the principle of finding the optimal hyperplane that separates different classes in the feature space with the maximum margin. This section delves into the foundational aspects of SVMs, elucidating their operational principles, mathematical formulation, and the unique characteristics that distinguish them from other machine learning algorithms.
	
	Moreover, SVMs offer robustness against overfitting, particularly in high-dimensional spaces, making them suitable for handling complex data sets with relatively small sample sizes. Additionally, SVMs can efficiently handle nonlinear classification tasks through the use of kernel functions, allowing them to capture intricate decision boundaries that may not be linearly separable in the original feature space. Furthermore, the margin-based optimization objective of SVMs promotes generalization performance, leading to models that exhibit strong predictive capabilities on unseen data. Consequently, SVMs have found widespread applications across various domains, including computer vision, bioinformatics, finance, and text classification.
	
	\paragraph{Operational Principles}
	SVMs operate by constructing a hyperplane or set of hyperplanes in a high-dimensional space, which can be used for classification, regression, or other tasks. Furthermore, the algorithm aims to create a hyperplane that effectively separates the classes in the feature space with as wide a margin as possible, a principle known as margin maximization. Additionally, the selection of the optimal hyperplane is crucial, as it directly impacts the generalization capability of the SVM model. On the contrary, failing to find an appropriate hyperplane may result in poor performance and overfitting to the training data. Moreover, the SVM algorithm's effectiveness lies in its ability to handle both linearly separable and non-linearly separable datasets through the use of kernel functions, thereby enhancing its versatility and applicability across various domains. Consequently, SVMs are renowned for their robustness, particularly in scenarios where the data is complex or exhibits nonlinear relationships. Hence, understanding the operational principles of SVMs is fundamental for harnessing their full potential in real-world applications.

	\paragraph{Mathematical Formulation}
	The decision function for a binary classification SVM is given by:
	\[ f(x) = \text{sign}(\sum_{i=1}^{n} \alpha_i y_i \langle x, x_i \rangle + b) \]
	where \(x\) represents the input features, \(x_i\) are the support vectors, \(y_i\) are the labels of the support vectors, \(\alpha_i\) are the Lagrange multipliers obtained from solving the dual optimization problem, \(\langle x, x_i \rangle\) denotes the inner product, and \(b\) is the bias term. The support vectors are the data points that lie closest to the decision boundary and are pivotal in defining the hyperplane.
	
	\paragraph{Kernel Trick}
	A key feature of SVMs is their use of the kernel trick, a method that allows them to operate in a high-dimensional, implicitly transformed feature space without ever computing the coordinates of the data in that space explicitly. This enables the handling of nonlinear relationships between classes by applying linear classification techniques to the transformed data. \textbf{Moreover}, the kernel trick is particularly useful when dealing with complex data distributions where direct linear separation is not feasible. \textbf{Furthermore}, by selecting appropriate kernel functions such as the radial basis function (RBF) or polynomial kernels, SVMs can effectively capture intricate decision boundaries, leading to improved generalization performance. \textbf{In addition}, the kernel trick contributes to computational efficiency since it avoids the need to explicitly transform the data into a higher-dimensional space, thus reducing the computational burden associated with processing large datasets. Overall, the kernel trick significantly enhances the flexibility and performance of SVMs, making them a powerful tool for various classification tasks.

	\paragraph{Versatility and Applications}
	SVMs are known for their versatility, being applicable to a wide range of domains such as image recognition, bioinformatics, and text classification. Their ability to manage both linear and nonlinear data, along with the robust theoretical foundation, makes SVMs a popular choice among machine learning practitioners. Furthermore, SVMs offer robust performance even in high-dimensional spaces, which is particularly useful in fields like bioinformatics where data often has a large number of features. Moreover, SVMs can handle both classification and regression tasks efficiently, providing a unified framework for various machine learning problems. Additionally, SVMs have been successfully applied in domains where interpretability is crucial, such as medical diagnosis systems, due to their ability to provide clear decision boundaries. Therefore, SVMs stand out as a versatile and powerful tool in the machine learning toolkit, offering solutions to diverse problems across multiple domains.

	In summary, the concept of Support Vector Machines encompasses a sophisticated blend of linear algebra, optimization, and geometry, aimed at producing models that not only perform well on the training data but also generalize effectively to unseen data, embodying a balance between complexity and performance.

	\subsubsection{Key Principles and Mechanisms}
	Support Vector Machines (SVMs) are grounded in a set of key principles and mechanisms that enable their high performance across classification, regression, and outlier detection tasks. This subsubsection explores these foundational elements, including the concepts of margin maximization, support vectors, the dual problem, and the kernel trick, which together form the cornerstone of SVM methodology.
	
	\paragraph{Margin Maximization}
	The fundamental goal of an SVM is to find the hyperplane that separates the classes in the feature space with the maximum margin. The margin, representing the distance between the hyperplane and the nearest points from each class, commonly known as support vectors, is crucial for the SVM's performance. Maximizing this margin ensures robustness against outliers and noise in the data, as it provides a larger region of separation between classes. This larger margin not only aids in classifying the training data correctly but also enhances the model's ability to generalize well to unseen data. By maximizing the margin, SVMs aim to strike a balance between fitting the training data closely and maintaining a suitable level of complexity to avoid overfitting. Consequently, the hyperplane derived through margin maximization becomes a robust decision boundary that optimally separates different classes, thereby improving the model's predictive accuracy and reliability.

	\paragraph{Support Vectors}
	Support vectors are pivotal entities within the framework of Support Vector Machines (SVMs). These data points, strategically positioned closest to the decision boundary, wield considerable influence over the SVM's efficacy. As determinants of the hyperplane's orientation and placement, they dictate the margin's width, a fundamental aspect of SVM classification. Any perturbation in the configuration or characteristics of these support vectors can induce significant alterations in the hyperplane, thereby impacting the model's classification performance. Their role transcends mere representation; they serve as anchors in the feature space, guiding the SVM in delineating complex decision boundaries. Consequently, their identification and utilization are central to SVM optimization strategies, ensuring robust and accurate classification outcomes. Moreover, understanding the dynamics of support vectors elucidates the interpretability of SVM models, shedding light on the underlying rationale governing classification decisions. Thus, the strategic positioning and characteristics of support vectors underscore their indispensable role in the SVM paradigm.

	\paragraph{The Dual Problem}
	SVMs operate by solving an optimization problem that seeks to maximize the margin while minimizing classification errors. The solution to this problem can be approached through its dual formulation, which allows for the incorporation of the kernel trick and simplifies the optimization process, especially when dealing with nonlinear data. The dual problem involves finding a set of Lagrange multipliers that maximize the margin, subject to certain constraints. However, solving the dual problem introduces additional computational complexity compared to the primal problem. Moreover, the dual formulation provides insights into the relationships between data points through the kernel function, enabling SVMs to capture complex decision boundaries more effectively. Furthermore, the dual problem offers flexibility in choosing appropriate kernel functions tailored to specific data characteristics. Additionally, by solving the dual problem, SVMs can handle datasets with high dimensionality efficiently. Thus, despite its computational challenges, the dual formulation significantly enhances the versatility and performance of SVMs in various classification tasks.

	\paragraph{The Kernel Trick}
	The kernel trick stands as a cornerstone in Support Vector Machines (SVMs), revolutionizing their capacity for linear classification within an implicitly transformed feature space, obviating the need for explicit transformation computations. By leveraging diverse kernel functions, SVMs adeptly navigate nonlinear relationships by projecting input features into higher-dimensional spaces conducive to linear separations. The deployment of polynomial, radial basis function (RBF), and sigmoid kernels exemplifies this versatility. Polynomial kernels raise input features to various powers, amplifying the feature space's dimensionality and enabling intricate decision boundaries. RBF kernels, with their Gaussian nature, map inputs into infinite-dimensional spaces, where intricate nonlinear relationships unravel into discernible patterns. Sigmoid kernels, inspired by neural networks, simulate neural activity, accommodating complex decision boundaries. This profound methodology transcends conventional linear SVM limitations, offering a sophisticated framework for discerning intricate data patterns without succumbing to the curse of dimensionality.

	\paragraph{Regularization and Soft Margin}
	Regularization in SVMs addresses the trade-off between maximizing the margin and minimizing classification errors. The introduction of the soft margin concept allows for some misclassifications, providing the model with the flexibility to handle noisy and overlapping data distributions. This is achieved by introducing slack variables that permit data points to be on the wrong side of the margin, controlled by a regularization parameter. Furthermore, by incorporating the soft margin, SVMs become more robust to outliers and noisy data, as it reduces the impact of individual data points on the overall decision boundary. Moreover, the regularization parameter allows fine-tuning the balance between margin maximization and error minimization, providing a mechanism to control model complexity and prevent overfitting. Additionally, the soft margin formulation enables SVMs to generalize better to unseen data, as it prioritizes a broader margin over perfect separation. Hence, the integration of regularization and soft margin enhances the adaptability and generalization capability of SVMs, making them suitable for a wide range of real-world classification tasks.

	Together, these principles and mechanisms underpin the operation of SVMs, enabling them to deliver robust and versatile models capable of tackling a wide range of machine learning challenges with high accuracy and generalization capabilities.

	\subsubsection{The Role of the Kernel Trick}
	The kernel trick is a fundamental concept in the operation of Support Vector Machines (SVMs), allowing these models to efficiently handle nonlinear data. This technique is pivotal in extending the applicability of SVMs beyond linear classification problems, enabling the construction of highly accurate and complex decision boundaries. This subsubsection delves into the essence of the kernel trick, its mathematical underpinnings, and the impact it has on the functionality and versatility of SVMs.
	
	\paragraph{Transforming Feature Spaces}
	At its core, the kernel trick involves mapping input features into a higher-dimensional space without explicitly performing the transformation. This is achieved through kernel functions, which compute the inner products of data points in the transformed feature space, effectively simulating the process of transformation and allowing SVMs to operate as if the data were linearly separable in this new space. Furthermore, by employing various kernel functions such as polynomial, radial basis function (RBF), or sigmoid kernels, SVMs can capture complex relationships between data points that may not be linearly separable in the original feature space. Consequently, this flexibility enables SVMs to handle nonlinear decision boundaries with ease, making them powerful tools for classification and regression tasks. Moreover, the kernel trick enhances computational efficiency by avoiding the explicit calculation of the transformed feature vectors, thus saving computational resources, especially in scenarios with high-dimensional data or large datasets. Additionally, the ability to operate in a higher-dimensional space allows SVMs to effectively deal with overlapping classes or classes that cannot be separated by a simple linear decision boundary.

	\paragraph{Mathematical Foundation}
	The mathematical foundation of the kernel trick lies in its ability to replace the standard dot product used in the SVM decision function with a kernel function. For two input vectors \(x\) and \(x'\), a kernel function \(k(x, x')\) returns the dot product of the vectors in the transformed space. Commonly used kernel functions include:
	\begin{itemize}
		\item Linear: \(k(x, x') = x^\top x'\)
		\item Polynomial: \(k(x, x') = (\gamma x^\top x' + r)^d\)
		\item Radial Basis Function (RBF): \(k(x, x') = \exp(-\gamma \|x - x'\|^2)\)
		\item Sigmoid: \(k(x, x') = \tanh(\gamma x^\top x' + r)\)
	\end{itemize}
	where \(\gamma\), \(r\), and \(d\) are parameters that control the shape of the kernel function.
	
	\paragraph{Enabling Nonlinear Classification}
	By employing the kernel trick, Support Vector Machines (SVMs) can construct nonlinear decision boundaries in the original input space that correspond to linear hyperplanes in the transformed feature space. This capability significantly enhances the model's ability to capture complex patterns and relationships in the data. Moreover, it facilitates superior performance on a wide range of nonlinear classification tasks. The kernel trick allows SVMs to implicitly map the input data into a higher-dimensional space where it becomes easier to separate classes with a linear boundary. Consequently, even if the data is not linearly separable in the original space, SVMs can effectively classify it by exploiting the higher-dimensional feature space. Furthermore, this approach not only handles nonlinearities but also mitigates the risk of overfitting, as it focuses on maximizing the margin between classes. Thus, SVMs equipped with the kernel trick offer a versatile and powerful framework for tackling diverse classification problems, ranging from image recognition to financial forecasting.

	\paragraph{Considerations and Challenges}
	While the kernel trick greatly expands the versatility of SVMs, selecting the appropriate kernel function and tuning its parameters ($\gamma$, $r$, and $d$) are critical for achieving optimal model performance. Additionally, the choice of kernel and its parameters can significantly influence the model's susceptibility to overfitting, computational efficiency, and generalization ability. The selection of a kernel function depends on the dataset's characteristics and the problem at hand; for instance, a polynomial kernel may be suitable for capturing complex relationships, while a radial basis function (RBF) kernel might perform better with non-linearly separable data. Moreover, the parameter $\gamma$ in the RBF kernel controls the influence of individual training samples, affecting the smoothness of the decision boundary and the model's capacity to generalize. Tuning these parameters involves a trade-off between bias and variance, as increasing model complexity might lead to overfitting, while overly simplistic models may fail to capture the underlying patterns effectively. Thus, careful consideration and experimentation are necessary to strike the right balance and ensure optimal SVM performance.

	The kernel trick represents a powerful mechanism that underlies the success of SVMs in nonlinear classification problems, enabling these models to effectively navigate the complexities of varied datasets. Its judicious application is key to unlocking the full potential of SVMs in machine learning applications.

	\subsubsection{Applications and Limitations}
	Support Vector Machines (SVMs) are celebrated for their robustness and efficacy across a broad spectrum of applications, ranging from image classification in computer vision to sentiment analysis in natural language processing, bioinformatics for gene, patient, and disease classification, and even market prediction and financial analyses in economics. These diverse applications underscore the versatility of SVMs, capitalizing on their ability to manage high-dimensional data, perform well in sparse data settings, and provide accurate predictions in pattern recognition tasks.
	
	Despite their widespread use and advantages, SVMs encounter limitations that may affect their performance or applicability in certain situations. One notable challenge is their scalability to large datasets; the computational complexity, particularly with non-linear kernels, can become prohibitive, limiting their use in big data scenarios. Additionally, the selection of an appropriate kernel function and the tuning of hyperparameters such as the regularization parameter and kernel parameters can be intricate processes requiring substantial expertise and experimentation to optimize model performance. This complexity also impacts the interpretability of SVM models, especially those employing complex kernels, making them less transparent and harder to explain than simpler models. Furthermore, while SVMs are generally robust, their performance can degrade in the presence of highly noisy datasets or when classes overlap significantly, posing challenges for applications in environments with substantial data noise or ambiguity.
	
	Understanding the strengths and limitations of SVMs is essential for leveraging their capabilities effectively. While they offer powerful solutions for a variety of machine learning challenges, recognizing the contexts in which they excel and those where alternative approaches might be more appropriate is crucial for achieving optimal outcomes in machine learning projects and research.
	
	\subsubsection{Pseudocode for the Algorithmic SVM}
	The Support Vector Machine (SVM) Algorithm is a sophisticated framework designed for efficiently estimating parameters in statistical models, particularly when dealing with optimizing hyperplanes for classification tasks. It distinguishes itself by iteratively adjusting Lagrange multipliers ($\alpha_i$) to satisfy the Karush-Kuhn-Tucker (KKT) conditions, thereby maximizing the margin between the support vectors of the two classes. This operational essence of SVM is encapsulated in pseudocode \ref{fig:svm-pseudocode}, illustrating its iterative approach to parameter estimation.
	
	This pseudocode encapsulates the essence of SVM training, focusing on the iterative adjustment of Lagrange multipliers ($\alpha_i$) to satisfy the Karush-Kuhn-Tucker (KKT) conditions, thereby ensuring the maximization of the margin between the support vectors of the two classes. The procedure \textsc{TrainSVM} iterates over the training set, adjusting $\alpha_i$ and $b$ to find the optimal hyperplane. The \textsc{Predict} procedure then utilizes the trained model to classify new examples based on the sign of the decision function. This algorithmic representation lays the groundwork for understanding how SVMs function and sets the stage for exploring their Algogenic enhancements.
	
	\begin{algorithm}
		\caption{Standard Support Vector Machine Training}
		\begin{algorithmic}[1]
			\Procedure{TrainSVM}{$X, Y, C$}
			\State $n \gets$ length($X$) \Comment{Number of training examples}
			\State Initialize $\alpha_i \gets 0$ for all $i$
			\State Initialize $b \gets 0$
			\While{optimization objective improves}
			\For{each $i$ in $1$ to $n$}
			\If{KKT conditions are violated for $\alpha_i$}
			\State Select $j \neq i$ randomly
			\State Compute $L, H$ based on $C, \alpha_i, \alpha_j, y_i, y_j$
			\State Optimize $\alpha_i, \alpha_j$ using $L, H$ and the objective function
			\State Update $b$ based on optimized $\alpha_i, \alpha_j$
			\EndIf
			\EndFor
			\EndWhile
			\State \textbf{return} $\alpha, b$
			\EndProcedure
			
			\Procedure{Predict}{$x, \alpha, b, X, Y$}
			\State $f(x) \gets \sum_{i=1}^{n} \alpha_i y_i K(x_i, x) + b$ \Comment{Decision function}
			\State \textbf{return} sign($f(x)$)
			\EndProcedure
		\end{algorithmic}\label{fig:svm-pseudocode}
	\end{algorithm}

\subsection{Previous Work on ML and AI Interplay with Support Vector Machines}

\paragraph{Deep support vector machine for hyperspectral image classification}
In 2020, a study presented a method integrating deep learning with support vector machines (SVM) for hyperspectral image classification \cite{okwuashi2020deep}. This approach aimed to utilize deep learning models to extract features from high-dimensional data, improving classification accuracy for hyperspectral images. By employing a deep SVM framework, the research demonstrated enhancements in classification accuracy, suggesting potential applications in remote sensing and environmental monitoring.

\paragraph{Deep support vector neural networks}
In 2020, researchers proposed a hybrid architecture termed deep support vector neural networks \cite{diaz2020deep}. This model combines neural networks and SVMs, leveraging deep learning for feature extraction and representation learning, while utilizing SVMs for final classification. The integration of these components enhances classification performance and generalization to unseen data. This work represents a significant advancement in hybrid machine learning models, offering insights into combining different learning paradigms effectively.

\subsection{Algogenic Enhancements for SVMs}
\subsubsection{Feature Conceptualization and Optimization}

\paragraph{Revisiting Feature Conceptualization with Algogenic Input}
In the context of Support Vector Machines (SVM), the integration of generative AI, specifically LLMs, is proposed to offer a nuanced method for identifying and refining features. This method leverages the capacity of LLMs to analyze complex datasets, potentially unveiling unique feature transformations or kernel functions that conventional methodologies might not identify. We suggest this integration could yield insights that improve SVM's predictive accuracy. However, we acknowledge the exploratory nature of this proposal and the necessity for empirical validation to substantiate these potential enhancements.

\paragraph{Tempering Expectations in SVM Feature Selection Optimization}
We advocate for a cautious approach to enhancing SVM feature selection through insights provided by generative AI. This approach entails a deliberate examination of how various feature combinations might influence SVM's performance, with a particular focus on kernel function selection. Although the prospect of improved model accuracy and adaptability is promising, we recognize the intricate dynamics at play and underscore the importance of empirical evidence to validate these potential advancements.

\paragraph{Generative AI's Tentative Role in Refining SVM Features}
The potential for generative AI to assist in the iterative refinement of SVM features and kernel parameters is approached with cautious optimism. Envisioning a collaborative process, LLMs could guide the development of feature sets and kernel configurations to bolster model performance. Nevertheless, this proposal is advanced with a recognition of its speculative nature, emphasizing the critical role of thorough experimentation and validation in establishing its validity.

\subsubsection{Dynamic Kernel Adjustment}

\paragraph{Adaptive Kernel Function Refinement via Algogenic Insights}
The proposal for dynamically adjusting SVM kernel functions, with guidance from LLM analysis, seeks to refine SVM performance. This approach posits that LLMs might offer valuable recommendations for modifying kernel parameters or exploring new kernel functions, based on a deep analysis of data characteristics. While this hypothesis holds promise, it is advanced with caution, stressing the experimental status of these enhancements and the necessity for rigorous validation.

\paragraph{Kernel Parameter Optimization: A Cautious Exploration}
The potential for leveraging LLM insights for the optimization of SVM kernel parameters is explored as a means to improve model precision and generalization. This exploration includes fine-tuning parameters, such as the gamma value in RBF kernels, guided by data-driven insights from LLMs. Acknowledging the preliminary nature of this exploration, we highlight the importance of continuous evaluation and adaptation based on empirical findings.

\paragraph{Exploring the Synergy between Generative AI and SVM Kernel Strategies}
We propose a prudent examination of how generative AI might interact with SVM kernel strategies to introduce dynamic adaptability into SVM models. This proposal is based on the premise that LLMs could provide real-time, data-informed recommendations for adjusting kernel functions, potentially improving SVM's responsiveness to diverse data characteristics. However, this exploration is undertaken with a skeptical and modest outlook, recognizing the necessity for extensive empirical testing to ascertain the effectiveness of this strategy.

\subsubsection{Geometric Hyperplane Optimization}

\paragraph{Leveraging Algogenic Insights for Hyperplane Fine-tuning}
The optimization of the SVM decision hyperplane, informed by insights from generative AI, is proposed as a means to potentially elevate classification accuracy. This strategy involves analyzing support vectors and their positioning relative to the hyperplane, with the aim of utilizing LLMs to suggest geometric modifications. Although the theoretical basis for this approach is promising, its implementation is approached with caution, underlining the indispensable role of empirical validation.

\paragraph{Predictive Intelligence in Decision Boundary Optimization}
The application of predictive intelligence, gleaned from LLMs, to the refinement of SVM decision boundaries is considered as a potential avenue for enhancing model accuracy and robustness. This strategy relies on the predictive capabilities of LLMs to anticipate the effects of hyperplane adjustments, with the goal of optimizing SVM's classification performance. While the potential for improvement is acknowledged, we emphasize a commitment to meticulous testing and validation.

\paragraph{Investigating the Intersection of Geometric Understanding and Algogenic Enhancement}
The potential integration of geometric insights from LLMs into the optimization of SVM hyperplanes is investigated as a means to enhance model performance. This investigation is approached with an inquisitive mindset, acknowledging the complexities of merging generative AI insights with the geometric underpinnings of SVM. The emphasis is placed on the necessity for empirical research to validate the prospective benefits of this integrative approach.

\subsubsection{Selective Sample Re-weighting}

\paragraph{Strategic Sample Influence via Algogenic Methods}
An exploration into the selective re-weighting of samples in SVM training, informed by LLM insights, is proposed. This strategy aims to dynamically adjust the impact of key data points, potentially enhancing SVM's generalization capability. While the concept is theoretically grounded, its practical application is approached with caution, highlighting the imperative for comprehensive experimentation.

\paragraph{Analyzing and Adjusting Data Points with LLM Insights}
The potential role of LLMs in the strategic adjustment of sample weights within SVM training is considered as a means to improve model performance. This strategy, which employs LLMs for advanced data analysis, is presented as exploratory, emphasizing the importance of empirical validation for these adjustments.

\paragraph{Algogenic Re-weighting: A Path to Enhanced SVM Adaptability}
The concept of Algogenic re-weighting is introduced as a strategy for potentially improving the adaptability and performance of SVM models. This approach involves using LLMs to inform the dynamic adjustment of sample weights, based on their perceived impact on model performance. This idea is approached with cautious optimism, highlighting the exploratory nature of these enhancements and the critical need for empirical validation.

\subsubsection{Interpretation and Explanation Enhancement}

\paragraph{Enhancing SVM Interpretability through Algogenic Insights}
The exploration of enhancing SVM interpretability and explanation capabilities through Algogenic means seeks to leverage the analytical capabilities of LLMs to generate understandable explanations of SVM decisions. While the approach holds promise, it is pursued with caution, recognizing the challenges inherent in providing clear and impactful explanations.

\paragraph{Generative AI's Potential in Facilitating SVM Interpretability}
The potential of generative AI, specifically LLMs, to enhance the interpretability of SVM models by providing narratives that clarify model predictions is explored. This exploration is undertaken with humility, acknowledging the difficulty of translating SVM operations into accessible explanations and the necessity for empirical validation.

\paragraph{Advancing SVM Applications through Enhanced Explanations}
The proposal to advance SVM applications through the integration of enhanced explanation capabilities, facilitated by Algogenic methods, aims to provide stakeholders with deeper insights into the model's decision-making process. Recognizing the exploratory nature of this proposition, the importance of validation and ethical considerations in developing explanation mechanisms is emphasized.

\subsubsection{Prediction Confidence Analysis}

\paragraph{Incorporating Algogenic Confidence Measures into SVM Predictions}
The incorporation of prediction confidence analysis into SVM models, informed by Algogenic insights, is suggested as a method to provide a nuanced understanding of the model's certainty in its predictions. This suggestion is advanced with caution, emphasizing the need for empirical testing to determine the impact of these measures on model utility and interpretability.

\paragraph{Exploring the Role of LLMs in Confidence Estimation}
The exploration of LLMs' role in enhancing SVM prediction confidence is proposed as a means to offer richer model insights. This role involves translating the geometric properties of SVM predictions into meaningful confidence indicators, approached with a sense of exploration and the acknowledgement of the necessity for careful validation.

\paragraph{Enhancing SVM Utility with Detailed Confidence Analysis}
The potential for enhancing SVM utility through detailed confidence analysis, informed by Algogenic insights, is explored. This enhancement aims to provide stakeholders with greater clarity on the reliability of SVM predictions, potentially improving decision-making processes. This exploration is undertaken with a commitment to rigorous testing and ethical implementation.

\subsubsection{Adaptive Post-processing Adjustment}

\paragraph{Exploratory Adaptive Post-processing for SVM Optimization}
The concept of adaptive post-processing adjustment, leveraging LLM insights to fine-tune SVM outputs, is proposed as a means to enhance model applicability and performance. This concept is pursued with an exploratory mindset, emphasizing the necessity for empirical validation and the careful consideration of application-specific requirements.

\paragraph{Synergizing LLMs and SVMs for Enhanced Decision-making}
The potential synergy between LLMs and SVMs in the post-processing phase is explored as a means to refine and enhance decision-making capabilities. This exploration is based on cautious optimism, acknowledging the complexity of effectively integrating these technologies and the importance of empirical testing.

\paragraph{Advancing SVM Applications through Algogenic Post-processing}
The proposal to advance SVM applications through adaptive post-processing adjustments, informed by LLM insights, is approached with an exploratory attitude. This proposal highlights the need for validation and the exploration of novel post-processing strategies to cater to diverse application needs.

	\subsubsection{Pseudocode for Algogenic SVMs}
	The SVM-based approach leverages AI to augment traditional SVM methods by dynamically adjusting parameters and strategies based on the observed behavior of the system and real-time error estimates. This pseudocode, available in \ref{fig:svm-Algogen-pseudocode}, outlines an advanced framework incorporating AI-driven enhancements for adaptive parameter tuning, kernel selection, regularization strategies, and real-time optimization.
	
	\begin{algorithm}
		\caption{Algogenic SVM Framework Pseudocode}
		\begin{algorithmic}[1]
			\Procedure{AlgogenicSVM}{Dataset}
			\State \textbf{Preprocessing:}
			\State Feature Conceptualization with LLM(Dataset)
			
			\State \textbf{Core Training:}
			\State Initialize SVM with Default Parameters
			\While{not Converged}
			\State Train SVM with Current Parameters
			\State Dynamic Kernel Function Optimization with LLM Insights
			\State Selective Sample Re-weighting with LLM Analysis
			\State Adjust Hyperplane with LLM Geometric Optimization
			\If{Performance Improved}
			\State Update Parameters \& Continue Training
			\Else
			\State Reassess \& Adjust Strategy
			\EndIf
			\EndWhile
			
			\State \textbf{Postprocessing:}
			\State Adjust Predictions \& Confidence Levels based on LLM Analysis
			\EndProcedure
		\end{algorithmic}\label{fig:svm-Algogen-pseudocode}
	\end{algorithm}
	
	\begin{figure}
		\centering
		\includegraphics[width=0.8\textwidth]{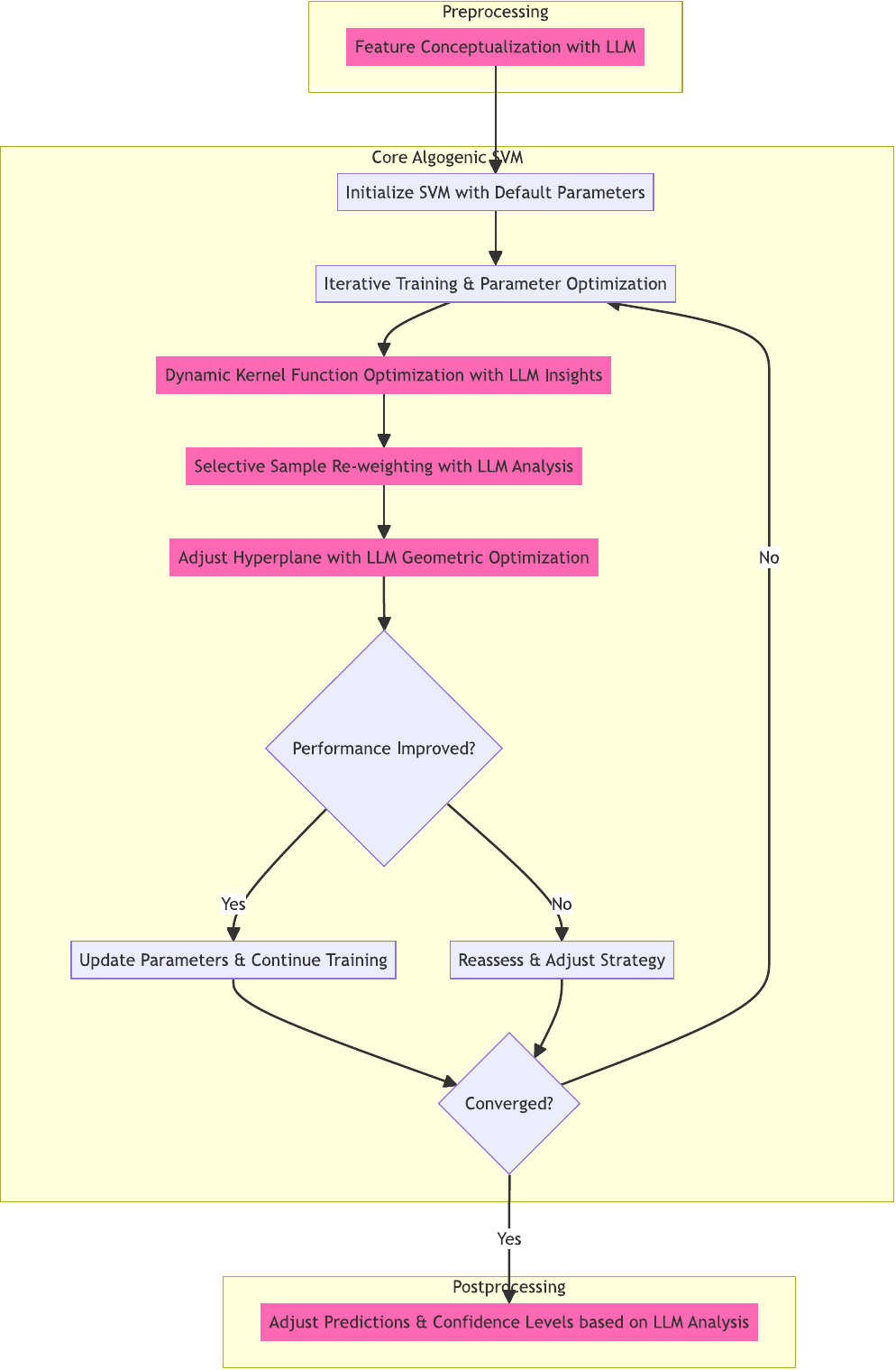} 
		\caption{Integration of Algogenic Enhancements in SVM: This figure delineates the comprehensive framework of Algogenic SVM, illustrating the critical phases of preprocessing, core training, and post-processing. In preprocessing, LLMs are leveraged for advanced feature conceptualization. The core training phase depicts an iterative process involving dynamic kernel optimization, selective sample re-weighting, and hyperplane adjustments, all informed by LLM insights. The post-processing phase concludes the workflow with adjustments to predictions and confidence levels based on LLM analysis, showcasing the holistic application of generative AI to enhance SVM's predictive performance and interpretability.}
		\label{fig:svm}
	\end{figure}

	\section{Gradient Boosting Machines}\index{Gradient Boosting Machines}
	
	\subsection{Introduction to Gradient Boosting Machines}
	\paragraph{Concept and Evolution of Gradient Boosting Machines}
	Gradient Boosting Machines (GBMs) represent a pinnacle in the evolution of ensemble learning techniques, finding extensive applications in both regression and classification tasks within machine learning. GBMs extend the boosting paradigm, whereby weak learners are sequentially added to an ensemble, with each subsequent learner focusing on the residual errors of its predecessors. This iterative process enables GBMs to continually refine predictions, leading to enhanced model performance. Central to the effectiveness of GBMs is the utilization of gradient descent optimization, wherein the algorithm iteratively adjusts model parameters to minimize the loss function. This technique enables GBMs to navigate complex, high-dimensional feature spaces efficiently, making them particularly adept at handling diverse and challenging datasets. Moreover, the inherent flexibility of GBMs allows them to accommodate various loss functions, further enhancing their versatility and applicability across different problem domains. Overall, the concept and evolution of GBMs signify a significant advancement in machine learning, providing practitioners with a robust tool for tackling a wide array of predictive modeling tasks.

	\paragraph{Foundational Ideas}
	The foundational idea behind GBMs is to combine multiple weak learning models, typically decision trees, to create a strong predictive model. Each tree in the sequence is trained to correct the errors made by the previous one, \textbf{and} the learning process is guided by the gradient of the loss function. This iterative correction of errors enables GBMs to gradually improve model accuracy with each addition to the ensemble. \textbf{Furthermore}, the use of decision trees allows for capturing complex nonlinear relationships between features, making GBMs effective in a wide range of predictive modeling tasks. \textbf{Moreover}, the ensemble nature of GBMs helps in reducing overfitting by aggregating the predictions of multiple models. \textbf{Additionally}, the ability to handle both numerical and categorical data makes GBMs versatile for various types of datasets. \textbf{On the other hand}, the sequential nature of training can make GBMs computationally expensive, especially with large datasets, \textbf{but} advancements in parallel processing techniques have mitigated this issue to some extent.

	\paragraph{Historical Progression}
	The evolution of GBMs can be traced back to the work of Yoav Freund and Robert Schapire on the AdaBoost algorithm in the 1990s, which laid the groundwork for boosting techniques. Later, Jerome H. Friedman introduced the concept of gradient boosting in 2001, which formalized the use of gradient descent to minimize error across sequential models. \textbf{Furthermore}, since then, GBMs have seen numerous enhancements, including the introduction of regularization techniques to prevent overfitting, and the development of efficient implementations like XGBoost, LightGBM, and CatBoost. These advancements have \textbf{Moreover} propelled GBMs into the forefront of machine learning algorithms, making them widely used in various fields such as finance, healthcare, and e-commerce. The popularity of GBMs can be attributed to their \textbf{ability to handle diverse data types, scalability to large datasets, and robust performance} across different predictive modeling tasks. \textbf{Additionally}, the continuous research and development in this field have led to the emergence of novel techniques, such as tree boosting and ensemble methods, further \textbf{enhancing} the capabilities and applications of GBMs in real-world scenarios.

	\paragraph{Advancements and Innovations}
	Significant advancements in GBMs have focused on improving their speed, scalability, and accuracy. XGBoost, introduced by Tianqi Chen, became popular for its efficiency, scalability, and performance in machine learning competitions. The framework not only enhanced predictive accuracy but also offered impressive computational efficiency, making it a preferred choice for large-scale datasets. Moreover, its ability to handle sparse data and missing values efficiently contributed to its widespread adoption in various domains. LightGBM, developed by Microsoft, further optimized the training process by utilizing a histogram-based method for faster learning and reduced memory usage. By discretizing continuous features into bins and constructing histograms, LightGBM accelerates the training process by reducing the number of data points to be evaluated, thus enhancing computational efficiency. Additionally, CatBoost, developed by Yandex, introduced innovations in handling categorical variables and reducing model training time. Through advanced techniques like ordered boosting, CatBoost effectively handles categorical features without requiring preprocessing, thereby simplifying the workflow and accelerating model development.

	\paragraph{Impact on Machine Learning}
	The evolution of GBMs has had a profound impact on machine learning, enabling the development of highly accurate models for a wide range of applications, from risk assessment in finance to disease detection in healthcare. GBMs, with their iterative nature, not only handle large and complex datasets effectively but also excel in capturing intricate patterns within the data. Moreover, their adaptability to various types of data distributions and the ability to handle mixed data types make them versatile tools for diverse domains. Furthermore, GBMs have been instrumental in addressing challenges such as overfitting through techniques like regularization and ensemble learning. Consequently, they have become indispensable in the data science toolkit, offering robust solutions that consistently outperform traditional machine learning algorithms. Thus, GBMs play a pivotal role in advancing the capabilities of machine learning systems, driving innovations in predictive analytics and decision-making processes across industries.

	In summary, the concept and evolution of Gradient Boosting Machines reflect a journey of continuous improvement and innovation. From their theoretical underpinnings in boosting and gradient descent to the state-of-the-art implementations that push the boundaries of efficiency and performance, GBMs remain at the forefront of machine learning methodologies, driving forward the capabilities of predictive modeling.
	
	\subsubsection{Core Principles and Mechanisms}
	Gradient Boosting Machines (GBMs) operate on the principles of boosting and gradient descent, employing an ensemble of weak predictors, typically decision trees, to construct a robust predictive model. This subsubsection delves into the foundational principles and operational mechanisms that define GBMs, highlighting how they leverage these concepts to achieve high predictive accuracy.
	
	\paragraph{Ensemble Learning}
	At the heart of GBMs is the concept of ensemble learning, where multiple models (weak learners) are combined to form a stronger predictive model. The key insight is that by sequentially adding models to correct the errors of the ensemble so far, the combined model can achieve accuracy rates that are unattainable by individual learners. This approach is rooted in the idea that a group of weak models can, together, form a highly accurate prediction mechanism.
	
	Furthermore, ensemble learning offers robustness against overfitting, a common pitfall in machine learning, as the diversity among weak learners helps prevent the model from memorizing noise in the data. Moreover, the combination of diverse models through ensemble methods allows for capturing complex relationships within the data, enhancing the model's ability to generalize well to unseen instances. Additionally, ensemble techniques such as Gradient Boosting Machines (GBMs) leverage boosting algorithms to assign higher weights to misclassified instances, effectively focusing subsequent models on the hardest-to-predict cases. Consequently, this iterative refinement process leads to continuous improvement in predictive performance.

	\paragraph{Boosting and Weak Learners}
	Boosting, a form of ensemble learning, is central to Gradient Boosting Machines (GBMs). It involves training weak learners sequentially, with each learner focusing on the mistakes made by the previous ones. In GBMs, these learners are usually decision trees. A weak learner is defined as a model that performs slightly better than random guessing. By focusing on correcting errors, boosting methods, including GBMs, ensure that each successive learner adds value to the ensemble. Moreover, the iterative nature of boosting allows for the creation of a strong learner from a collection of weak ones. Additionally, the adaptive nature of boosting enables the model to handle complex relationships within the data. Furthermore, GBMs have gained popularity due to their ability to effectively handle large datasets with high dimensionality, making them suitable for a wide range of tasks in both classification and regression problems.

	\paragraph{Gradient Descent on Loss Function}
	Gradient descent, a fundamental optimization technique, is indispensable in the context of Gradient Boosting Machines (GBMs) for minimizing prediction error. At each iteration, wherein a new tree is added to the ensemble, GBMs leverage gradient descent to iteratively refine model parameters and reduce the disparity between actual and predicted values. The crux of this approach lies in the manipulation of a designated loss function, which quantifies the deviation between predicted and true values. Within the GBM framework, the term 'gradient' pertains to the gradient of this loss function, serving as a compass for the algorithm to navigate towards optimal parameter values. Through successive iterations, guided by the gradients of the loss function, GBMs adjust their model parameters in a manner that gradually diminishes prediction errors. Consequently, gradient descent emerges as a linchpin mechanism underpinning the iterative enhancement of ensemble performance within GBMs.

	\paragraph{Sequential Model Building}
	Sequential model building distinguishes Gradient Boosting Machines (GBMs) from some other ensemble methods by its sequential nature. Unlike parallel methods, GBMs construct models one after the other. Each subsequent tree in the ensemble is trained on the residual errors left by its predecessors. This sequential approach enables GBMs to systematically refine the ensemble's predictive power. By focusing on reducing the errors step by step, GBMs can effectively adapt to the intricacies of the data. This adaptive refinement mechanism is particularly advantageous as it targets specific areas where earlier models may have faltered, progressively enhancing the overall performance. Through this iterative process, GBMs iteratively correct for errors, iteratively adjusting the model to minimize the loss function. Consequently, GBMs tend to yield models that are well-suited to the underlying data distribution, effectively capturing complex patterns and relationships.

	\paragraph{Regularization Techniques}
	GBMs incorporate regularization techniques to control the model's complexity and prevent overfitting. \textbf{Moreover}, these techniques play a crucial role in fine-tuning the model's performance. By \textit{limiting the number of trees}, practitioners effectively manage computational resources and prevent the model from becoming too complex, which could lead to overfitting. \textit{Controlling the depth of each tree} is another pivotal aspect, as it prevents the trees from growing excessively deep and memorizing the training data, thus enhancing the model's ability to generalize. Additionally, \textit{applying shrinkage or learning rate} serves to slow down the learning process, allowing the model to make more gradual adjustments and avoid drastic changes that might lead to overfitting. These measures collectively ensure that the model not only fits the training data well but also generalizes effectively to unseen data, striking a balance between complexity and performance.

	In essence, the core principles and mechanisms of Gradient Boosting Machines — ensemble learning, boosting with weak learners, gradient descent on the loss function, sequential model building, and regularization — collectively contribute to their effectiveness in predictive modeling. These foundational elements enable GBMs to tackle a wide array of machine learning tasks, providing accurate and robust solutions across various domains.

	\subsubsection{The Role of Loss Functions}
	The loss function is a pivotal element in the training of Gradient Boosting Machines (GBMs), dictating the optimization direction and serving as a measure of the model's prediction error. This subsubsection explores the role of loss functions in GBMs, detailing their impact on model learning and the adaptability of GBMs to different loss functions for various tasks.
	
	\paragraph{Defining Model Optimization Objectives}
	The choice of loss function in GBMs defines the objective of model optimization. It quantitatively expresses how well the model's predictions align with the actual target values, with the aim of minimizing this discrepancy during training. Common loss functions include mean squared error (MSE) for regression tasks and log loss for classification tasks. The optimization process involves adjusting the model parameters to minimize the selected loss function, guiding the sequential addition of weak learners to improve model accuracy. Moreover, the selection of an appropriate loss function is crucial as it directly impacts the learning behavior of the model. Consequently, a carefully chosen loss function ensures that the model learns to prioritize certain aspects of the data, leading to better generalization and performance on unseen data. Additionally, different loss functions may be suitable for different types of problems, and understanding their properties is essential for effective model optimization.

	\paragraph{Gradient Descent and Loss Minimization}
	GBMs employ gradient descent to minimize the loss function iteratively. By calculating the gradient of the loss function with respect to the model parameters, GBMs can update these parameters in the direction that reduces prediction error. This iterative process is crucial because it allows GBMs to continuously refine their predictions, gradually improving model performance. Moreover, GBMs utilize a variety of loss functions tailored to specific tasks, such as mean squared error or cross-entropy loss. This flexibility enables GBMs to address diverse prediction problems effectively. Additionally, gradient descent ensures that GBMs can handle large-scale datasets efficiently, as it updates model parameters incrementally based on local information from each data point. Consequently, GBMs can navigate complex parameter spaces effectively, converging towards optimal parameter values that minimize the loss function. As a result, GBMs achieve superior predictive accuracy and robustness across various domains and applications.

	\paragraph{Customization for Task-Specific Performance}
	One of the strengths of GBMs is their flexibility in accommodating different loss functions, allowing for customization based on the specific requirements of the task at hand. This adaptability ensures that GBMs can be tailored to achieve optimal performance across a wide range of applications, from binary classification to multi-class classification and regression. Moreover, the ability to define custom loss functions enables the handling of complex scenarios and objectives not covered by standard loss functions, further enhancing the versatility of GBMs.
	
	Additionally, the incorporation of task-specific loss functions empowers practitioners to address nuanced challenges effectively. Furthermore, this capability broadens the applicability of GBMs in domains with unique optimization goals. Consequently, practitioners can exploit GBMs' adaptiveness to devise innovative solutions for diverse problem domains. Moreover, the inclusion of custom loss functions facilitates the alignment of model objectives with real-world requirements, fostering more meaningful insights and actionable results. Thus, GBMs stand out as a powerful framework for addressing a myriad of tasks with tailored precision and efficacy.

	\paragraph{Impact on Model Complexity and Generalization}
	The choice of loss function greatly influences the complexity of a model and its capacity for generalization. Incorporating regularization terms into the loss function allows for the discouragement of overly complex models, thereby favoring simpler models that are less prone to overfitting on the training data. By striking a balance between fitting the training data and controlling model complexity, robust Gradient Boosting Machines (GBMs) can be developed, ensuring optimal performance not only on the training set but also on unseen data. This balance is essential in mitigating the risk of overfitting while ensuring that the model captures the underlying patterns in the data effectively. Additionally, by adjusting the loss function, practitioners can tailor the learning process to emphasize certain aspects of the data, such as reducing sensitivity to outliers or prioritizing certain types of errors over others. Consequently, the choice of loss function serves as a critical aspect in the design and fine-tuning of GBMs, directly impacting their effectiveness in real-world applications.

	\paragraph{Challenges in Loss Function Selection}
	Selecting an appropriate loss function and incorporating regularization are pivotal stages in the Gradient Boosting Machine (GBM) training process. The choice of loss function not only influences the model's predictive performance but also shapes its behavior during training and inference. Moreover, the selection process is deeply intertwined with the intrinsic characteristics of the dataset and the overarching objectives of the machine learning task at hand. 
	
	Furthermore, the incorporation of regularization techniques introduces another layer of complexity. Regularization methods such as L1 or L2 regularization impose penalties on the model's parameters to prevent overfitting and enhance generalization. Balancing the trade-offs between optimizing the loss function and regularization parameters involves a delicate interplay between model accuracy, complexity, and computational efficiency.
	
	Hence, the challenge lies in navigating this intricate landscape to strike a harmonious balance that yields a well-performing model while ensuring it remains robust and scalable across different scenarios and datasets.

	In summary, the role of loss functions in Gradient Boosting Machines is multifaceted, influencing not only the direction of model optimization but also the adaptability, complexity, and generalization of the model. The ability to work with diverse loss functions underscores the flexibility and power of GBMs in addressing a broad spectrum of predictive modeling challenges.

	\subsubsection{The Role of Loss Functions}
	The loss function is a pivotal element in the training of Gradient Boosting Machines (GBMs), dictating the optimization direction and serving as a measure of the model's prediction error. This subsubsection explores the role of loss functions in GBMs, detailing their impact on model learning and the adaptability of GBMs to different loss functions for various tasks.
	
	\paragraph{Defining Model Optimization Objectives}
	The choice of loss function in GBMs defines the objective of model optimization. It quantitatively expresses how well the model's predictions align with the actual target values, with the aim of minimizing this discrepancy during training. Common loss functions include mean squared error (MSE) for regression tasks and log loss for classification tasks. The optimization process involves adjusting the model parameters to minimize the selected loss function, guiding the sequential addition of weak learners to improve model accuracy. Furthermore, the optimization objective directly influences the training dynamics, determining the direction and magnitude of parameter updates at each iteration. Consequently, the choice of loss function impacts not only the final model performance but also the computational efficiency of the optimization process. Moreover, the selection of appropriate evaluation metrics is crucial to assess model performance effectively and guide the optimization process towards achieving the desired objectives.

	\paragraph{Gradient Descent and Loss Minimization}
	GBMs employ gradient descent to minimize the loss function iteratively. By calculating the gradient of the loss function with respect to the model parameters, GBMs can update these parameters in the direction that reduces prediction error. This approach enables the model to navigate the parameter space efficiently, converging towards a set of parameters that minimize the loss, thereby optimizing the model's performance.
	
	Furthermore, this iterative process allows GBMs to handle complex, high-dimensional data effectively. Moreover, the flexibility of gradient descent allows GBMs to be adaptable to various types of loss functions, making them versatile for different learning tasks. Additionally, by continuously updating the parameters based on the gradient, GBMs can avoid getting stuck in local minima, ensuring that the model converges towards the global minimum of the loss function. Plus, the ability of GBMs to iteratively refine the model parameters makes them suitable for large-scale datasets, as they can incrementally learn from data in a computationally efficient manner.

	\paragraph{Customization for Task-Specific Performance}
	One of the strengths of GBMs is their flexibility in accommodating different loss functions, allowing for customization based on the specific requirements of the task at hand. This adaptability ensures that GBMs can be tailored to achieve optimal performance across a wide range of applications, from binary classification to multi-class classification and regression. Furthermore, the ability to define custom loss functions enables the handling of complex scenarios and objectives not covered by standard loss functions, thus further enhancing the versatility of GBMs. Additionally, this adaptability empowers practitioners to address nuanced nuances in data patterns, leading to more accurate predictions and model outputs. Hence, by incorporating task-specific considerations into the modeling process, GBMs can effectively capture the intricacies of diverse datasets and deliver superior performance, regardless of the domain or application.

	\paragraph{Impact on Model Complexity and Generalization}
	The choice of loss function also influences model complexity and its ability to generalize to unseen data. Incorporating regularization terms into the loss function, such as L1 or L2 penalties, enables the penalization of overly complex models, thus favoring simpler ones that are less prone to overfitting the training data. Striking a balance between optimizing the fit on the training data and controlling the model's complexity is pivotal in the development of robust Gradient Boosting Machines (GBMs). By regulating the complexity, GBMs can effectively navigate the bias-variance trade-off, enhancing their capability to generalize well not only within the training set but also across new, unseen data. Consequently, the choice of an appropriate loss function plays a crucial role in shaping the model's capacity to generalize and its overall performance.

	\paragraph{Challenges in Loss Function Selection}
	Selecting an appropriate loss function and incorporating regularization are critical steps in the GBM training process. The choice of loss function must reflect the goals of the specific machine learning task and the nature of the data. Moreover, tuning the model to optimize a particular loss function requires careful consideration of the trade-offs between model accuracy, complexity, and computational efficiency. \textbf{Furthermore}, the impact of the chosen loss function on the model's generalization ability should be evaluated, as some loss functions may lead to overfitting or underfitting. \textbf{Additionally}, the interplay between the loss function and the algorithm's optimization process should be understood to ensure convergence and stability. \textbf{Moreover}, the incorporation of regularization techniques such as L1 or L2 regularization \textbf{alongside} the choice of loss function can help prevent overfitting by penalizing overly complex models. \textbf{On the other hand}, too much regularization can lead to underfitting, emphasizing the need for a balanced approach. \textbf{In contrast}, neglecting regularization entirely can result in models that are overly sensitive to noise in the data, compromising their predictive performance.

	In summary, the role of loss functions in Gradient Boosting Machines is multifaceted, influencing not only the direction of model optimization but also the adaptability, complexity, and generalization of the model. The ability to work with diverse loss functions underscores the flexibility and power of GBMs in addressing a broad spectrum of predictive modeling challenges.
	
	\subsubsection{Pseudocode for Algorithmic GBMs}
	The Gradient Boosting Machines (GBMs) framework represents a sophisticated approach for parameter estimation in statistical models, particularly those involving latent variables. It operates iteratively to maximize the likelihood function, leveraging both observed data and latent variables for refining parameter estimates. This iterative process is outlined in the pseudocode \ref{fig:gbm-pseudocode}, illustrating the algorithm's approach to parameter estimation.
	
	This pseudocode initializes the model using a base decision tree, which predicts a value minimizing the loss function $L$ across all training data $Y$. Following initialization, GBMs proceed by iteratively fitting additional trees to the residuals—the disparities between observed and predicted values—from the preceding model iteration. Each subsequent tree aims to rectify the errors of the ensemble up to that point, with the learning rate parameter regulating the impact of each new tree to prevent overfitting. As the algorithm progresses, a specified number of trees are added, culminating in a composite model comprising all sequentially constructed trees, fine-tuned to predict the target variable with reduced error. This iterative process underscores the fundamental algorithmic structure of GBMs, demonstrating their ability to incrementally enhance predictive performance and adapt to intricate data relationships.

	\begin{algorithm}
		\caption{Gradient Boosting Machine Algorithm}
		\begin{algorithmic}[1]
			\Procedure{TrainGBM}{$X, Y, \textit{num\_trees}, \textit{learning\_rate}$}
			\State Initialize model with a single tree: $F_0(x) = \arg\min_{\gamma} \sum_{i=1}^{N} L(y_i, \gamma)$
			\For{$t = 1$ to $\textit{num\_trees}$}
			\State Compute residuals $r_{it} = -\left[ \frac{\partial L(y_i, F(x_i))}{\partial F(x_i)} \right]_{F(x) = F_{t-1}(x)}$ for all $i$
			\State Fit a decision tree $h_t(x)$ to residuals $r_{it}$, producing leaf regions $R_{jm}, j = 1,2,\ldots,J$
			\State Compute output values for each leaf region: $\gamma_{jm} = \arg\min_{\gamma} \sum_{x_i \in R_{jm}} L(y_i, F_{t-1}(x_i) + \gamma)$
			\State Update model: $F_t(x) = F_{t-1}(x) + \textit{learning\_rate} \cdot \sum_{j=1}^{J} \gamma_{jm} \mathbf{1}(x \in R_{jm})$
			\EndFor
			\State \textbf{return} Final model $F_t(x)$
			\EndProcedure
		\end{algorithmic}\label{fig:gbm-pseudocode}
	\end{algorithm}

\subsection{Previous Work on ML and AI Interplay with Gradient Boosting Machines}

\paragraph{Hybrid Approaches in Protein-Protein Interaction Prediction}
In 2021, a study presented a hybrid classifier combining deep neural networks (DNNs) with extreme gradient boosting (XGBoost) for predicting Protein-Protein interactions \cite{mahapatra2021deep}. This approach aimed to leverage the strengths of both methodologies to improve prediction accuracy. DNNs, known for their capacity to model complex non-linear relationships in large datasets, were combined with XGBoost, an ensemble learning method that excels in handling sparse data and preventing overfitting. The hybrid model demonstrated superior performance compared to traditional methods, indicating the potential of combining machine learning techniques to solve complex biological problems.

\paragraph{Enhancing Rainfall-Runoff Simulation Models}
In 2023, a study focused on improving the performance of deep learning techniques for rainfall-runoff simulation by integrating them with gradient boosting methods \cite{abdulaleva2023enhancing}. This integration aimed to address the limitations of standalone deep learning models in capturing the complex dynamics of hydrological processes. By combining deep learning with gradient boosting, the proposed approach sought to enhance model accuracy and reliability in predicting runoff. This work not only contributed to the field of hydrology by providing a more accurate and robust tool for runoff simulation but also illustrated the potential of blending advanced machine learning techniques to improve environmental modeling.

\paragraph{A Gradient Boosting Framework for Neural Network Optimization}
In 2023, a study introduced a gradient boosting framework for optimizing the training of convolutional and deep neural networks \cite{emami2023gradient}. This framework aimed to enhance the performance and efficiency of neural networks by systematically improving the learning process. By applying gradient boosting techniques, the framework iteratively refined the network's parameters, leading to more accurate and generalizable models. This research contributed to the broader understanding of how ensemble learning methods can be effectively applied to neural network training, offering valuable insights into potential algorithmic improvements in machine learning.	
	
	\subsection{Algogenic Enhancements for GBMs}
	\subsubsection{Semantic Feature Engineering}
	
	\paragraph{The Essence of Semantic Feature Engineering in GBMs}
	Semantic feature engineering, tailored specifically for Gradient Boosting Machines (GBMs), leverages the unique strengths of Large Language Models to enrich the preprocessing phase of GBMs with deep semantic insights. This approach delves into both structured and unstructured data to distill complex, contextual relationships into a refined feature set, offering GBMs a broader understanding of the data. The process not only captures explicit information but also implicit nuances, thereby equipping GBMs with the means to discern intricate patterns and relationships that conventional methods might miss. This fusion of LLM-derived features into GBMs aims to bolster model generalization and adaptability, especially beneficial in dynamic environments where data characteristics evolve over time. The implementation of semantic feature engineering could, for example, involve transforming natural language text into contextual embeddings that enrich the GBM's feature space, providing a tangible pathway for enhancing model performance and interpretability.
	
	\paragraph{Operationalizing LLMs for Feature Transformation}
	Incorporating LLMs into GBMs' feature engineering process involves a detailed analysis of the dataset to identify latent semantic patterns and relationships pertinent to the predictive task. LLMs can augment GBMs by transforming existing features or generating new ones that encapsulate a wealth of semantic information, potentially overlooked by traditional feature engineering. This might involve leveraging word embeddings or contextual representations derived from LLMs, which could then be systematically integrated into the GBM's learning process. Such integration not only enriches the GBM's input feature set but also enhances its ability to model complex dependencies, offering a clear, implementable strategy for boosting model accuracy and robustness.
	
	\paragraph{Impact of Semantic Feature Engineering on GBM Performance}
	Integrating semantic feature engineering into GBMs signifies a strategic departure from traditional feature engineering, enriching GBMs with a more nuanced understanding of their operational domain. This enhancement is particularly vital for tasks involving substantial unstructured data. By incorporating semantic insights directly into the GBM framework, models gain the ability to uncover and leverage latent patterns, thereby improving accuracy and generalizability. However, it's essential to navigate the computational demands and ensure the interpretability of the enhanced model, presenting a balanced approach to incorporating semantic features. This careful integration underscores the potential of semantic feature engineering to refine GBMs, making them more adept at tackling complex predictive tasks across diverse domains.
	
	\subsubsection{Dynamic Tree Complexity Adjustment}
	
	\paragraph{Adapting Tree Complexity in Real-Time}
	Dynamic tree complexity adjustment in GBMs, guided by insights from LLMs, offers a nuanced approach to optimizing decision tree architectures within the ensemble. This method dynamically tailors tree depth and complexity based on evolving data characteristics and model performance, aiming to balance the trade-off between model complexity and its ability to capture intricate data patterns. Operationalizing this involves analyzing residual errors and the semantic complexity of the dataset, thus predicting optimal tree configurations that enhance learning efficiency and model accuracy. This adaptive strategy ensures GBMs remain responsive to data dynamics, potentially mitigating overfitting and improving model generalization in changing environments.
	
	\paragraph{LLM-Guided Optimization of Tree Structures}
	LLMs facilitate the optimization of GBM tree structures by analyzing model performance and identifying patterns in the data that suggest adjustments to tree complexity. This could involve recommending deeper trees to better capture data intricacies or simpler trees to avoid overfitting, based on the LLM's analysis of residuals and data characteristics. Implementing this approach allows GBMs to adaptively refine their structure, enhancing predictive power and efficiency. The practical application of LLM-guided tree optimization could be seen in dynamically adjusting tree parameters in response to detected changes in data distribution, illustrating a concrete method for improving GBM performance.
	
	\paragraph{Enhancing GBM Performance Through Intelligent Tree Adjustment}
	Integrating dynamic tree complexity adjustment into GBMs, with insights from LLMs, marks a significant advancement in enhancing model adaptability and performance. This approach allows for real-time adjustments to tree architecture, facilitating a more efficient learning process and improved model accuracy. By automating complexity adjustment, GBMs can better navigate the complexities of diverse datasets, reducing the need for manual parameter tuning and making the model more accessible. The collaboration between GBMs and LLMs in this context exemplifies the potential for Algogenic systems to refine traditional machine learning techniques, leading to more effective, intelligent, and adaptive modeling approaches.
	
	\subsubsection{Adaptive Learning Rate Optimization}
	
	\paragraph{Optimizing Learning Rate with Generative Insights}
	Adaptive learning rate optimization in GBMs, informed by LLMs, dynamically adjusts the learning rate based on real-time analysis of model performance and data complexity. This approach aims to enhance the GBM's convergence efficiency by modulating the learning rate to either expedite or decelerate the learning process as appropriate. Operationalizing this concept involves LLMs assessing the model's progress and recommending learning rate adjustments to optimize performance. This method ensures a more nuanced, data-driven learning strategy, potentially accelerating convergence and improving model stability, illustrating a practical application of LLM insights in refining GBM training processes.
	
	\paragraph{LLM-Guided Learning Rate Decisions}
	LLM-guided learning rate decisions in GBMs enable a more adaptive and efficient training process by analyzing the model's learning trajectory and recommending adjustments to the learning rate. This process involves monitoring training performance and employing LLMs to identify optimal learning rate adjustments, thereby enhancing the model's ability to navigate the learning landscape. By implementing LLM recommendations, GBMs can more effectively adjust their learning rate in real-time, improving convergence rates and model accuracy. This approach highlights a tangible strategy for leveraging LLM capabilities to optimize GBM training, enhancing the algorithm's adaptability and performance.
	
	\paragraph{Impact on GBM Efficiency and Efficacy}
	Integrating adaptive learning rate optimization, guided by LLM insights, into GBMs significantly enhances their training efficiency and predictive efficacy. This Algogenic enhancement enables GBMs to dynamically adjust their learning rate, optimizing the training process and improving model performance. By reducing the need for manual tuning and enhancing model adaptability, this approach streamlines GBM deployment and utilization across various applications, demonstrating a practical application of combining generative AI with algorithmic decision-making to refine machine learning methodologies.
	
	\subsubsection{Predictive Feature Interaction Discovery}
	
	\paragraph{Unveiling Feature Interactions with LLM Insights}
	Predictive feature interaction discovery in GBMs, enhanced by LLM insights, focuses on identifying and leveraging complex interactions among features that traditional methods might overlook. This process employs LLMs to analyze datasets and reveal hidden synergies and dependencies, enriching GBMs with a deeper understanding of the data. Operationalizing this involves integrating LLM-identified interactions into the GBM's feature set, allowing the model to capture more nuanced patterns and relationships. This approach enhances the GBM's predictive accuracy and interpretability, offering a concrete example of how LLM insights can be applied to improve machine learning models.
	
	\paragraph{Operationalizing LLMs for Enhanced Feature Synthesis}
	Enhanced feature synthesis in GBMs, facilitated by LLMs, involves analyzing datasets to identify and incorporate complex feature interactions into the model. LLMs provide recommendations for new or modified features that better capture identified interactions, which are then integrated into the GBM's training process. This approach enriches the model's feature space and aligns it more closely with the underlying data structure, improving predictive performance and adaptability. The practical application of this method could include generating new composite features based on LLM analysis, illustrating a direct way to leverage LLM capabilities for GBM enhancement.
	
	\paragraph{Boosting GBM Performance Through Deep Feature Insights}
	The integration of predictive feature interaction discovery into GBMs, informed by LLM insights, significantly enhances the model's capacity for data representation and understanding. This approach ensures a more informative feature set, including complex interactions identified by LLMs, leading to improved model accuracy and generalizability. By addressing the challenges of traditional feature engineering in high-dimensional datasets, this Algogenic enhancement demonstrates a practical strategy for incorporating LLM-derived insights into GBMs, thereby advancing their performance and applicability across various domains.
	
	\subsubsection{Model Explanation Enhancement}
	
	\paragraph{Enhancing GBM Interpretability with LLMs}
	Enhancing GBM model explanations through LLMs introduces a method for generating detailed, accessible explanations of the model's predictions. This approach leverages LLMs to analyze the GBM's structure and output, providing insights into feature contributions and decision-making logic. Operationalizing this involves translating LLM analyses into coherent narratives or visualizations that explain predictions in understandable terms, offering a direct way to improve GBM interpretability and foster trust among users.
	
	\paragraph{Operational Framework for Algogenic Explanation Generation}
	The operational framework for generating Algogenic explanations in GBMs involves utilizing LLMs to dissect the model's predictions and elucidate the rationale behind them. This process includes analyzing decision trees and feature interactions, with LLMs synthesizing this information into understandable explanations. By implementing this framework, GBMs can offer users clear, insightful explanations of their predictions, enhancing transparency and trustworthiness. This method illustrates a tangible application of LLMs in making complex machine learning models more interpretable and accessible.
	
	\paragraph{The Impact of Enhanced Explanations on GBM Applications}
	Integrating enhanced explanations into GBM applications significantly improves the model's utility and acceptability across critical domains. By providing clear, LLM-generated explanations of predictions, GBMs become more transparent and trustworthy, enabling users to make informed decisions based on the model's outputs. This approach facilitates regulatory compliance, mitigates risks, and promotes user acceptance, illustrating a practical benefit of combining GBMs with LLMs to enhance machine learning interpretability and reliability.
	
	\subsubsection{Prediction Confidence Estimation}
	
	\paragraph{Integrating Confidence Measures into GBM Predictions}
	Integrating prediction confidence estimation into GBMs, with LLM support, introduces a method for assessing the certainty associated with each prediction. This process involves analyzing the model's output and residuals to generate confidence scores, offering a quantifiable measure of prediction reliability. By implementing this enhancement, GBMs can provide users with not only predictions but also insights into their confidence levels, improving decision-making and risk management. This approach highlights a practical application of LLMs in augmenting GBMs with valuable predictive insights.
	
	\paragraph{Operationalizing LLMs for Confidence Score Generation}
	Generating confidence scores in GBMs, facilitated by LLMs, involves a detailed evaluation of the model's predictions and the identification of factors influencing confidence levels. LLMs analyze prediction outcomes and compare them to known data patterns, generating confidence scores that reflect the model's reliability. By operationalizing LLMs for this purpose, GBMs can offer more nuanced predictions, enhancing their applicability and trustworthiness in various domains. This method provides a concrete example of leveraging LLM capabilities to improve the interpretability and utility of GBM predictions.
	
	\paragraph{The Role of Confidence Scores in Enhancing GBM Utility}
	The incorporation of confidence scores into GBM predictions enhances the model's applicability and decision-making capabilities. Confidence scores inform users about the reliability of predictions, enabling more informed and risk-aware decision-making. This Algogenic enhancement not only improves the model's interpretability but also promotes a more responsible and informed use of machine learning in critical applications, demonstrating a practical benefit of integrating confidence estimation into GBMs.
	
	\subsubsection{Adaptive Anomaly Detection in Residuals}
	
	\paragraph{Elevating GBM Robustness with Residual Analysis}
	Adaptive anomaly detection in GBM residuals, powered by LLMs, introduces a proactive approach to enhancing model accuracy and reliability. This process involves continuous monitoring of residuals to identify anomalies that may indicate shifts in data or emerging patterns. By operationalizing LLMs for this task, GBMs can adapt to changes and maintain performance over time, offering a direct way to leverage advanced AI for improving model robustness and adaptability.
	
	\paragraph{Operationalizing LLMs for Enhanced Residual Monitoring}
	Enhanced residual monitoring in GBMs, facilitated by LLMs, involves analyzing residuals for patterns or anomalies indicative of model or data issues. LLMs evaluate residuals, identifying deviations and informing users of potential model adjustments. This approach enables GBMs to respond adaptively to changes, ensuring sustained performance and relevance. The practical application of this method demonstrates how LLMs can be leveraged to maintain and enhance the robustness of GBM models in dynamic environments.
	
	\paragraph{Strengthening GBM Applications Through Intelligent Residual Management}
	Integrating intelligent residual management techniques into GBMs enhances their resilience and long-term viability. By incorporating LLMs for adaptive anomaly detection, GBMs can proactively address deviations in model performance, ensuring their applicability across various domains. This approach not only improves model accuracy but also fosters a continuous improvement cycle, demonstrating a practical strategy for leveraging LLM insights to enhance GBM robustness and reliability in real-world applications.

	\subsubsection{Pseudocode for Algogenic GBMs}
	The Algogenic Gradient Boosting Machines approach utilizes AI to enhance traditional Gradient Boosting methods by dynamically adjusting boosting parameters and strategies based on the observed behavior of the system and real-time error estimates. This pseudocode, available in \ref{fig:gradient-boosting-Algogen-pseudocode}, outlines an advanced framework incorporating AI-driven enhancements for adaptive learning rate control, feature selection, tree construction criteria, and real-time parameter optimization.
	
	\begin{algorithm}
		\caption{Algogenic GBM Framework Pseudocode}
		\begin{algorithmic}[1]
			\Procedure{AlgogenicGBM}{Dataset}
			\State \textbf{Preprocessing:}
			\State Semantic Feature Engineering with LLM(Dataset)
			
			\State \textbf{Core Training:}
			\State Initialize GBM with Default Parameters
			\While{not Converged}
			\State Train GBM with Current Parameters
			\State Dynamic Tree Complexity Adjustment with LLM Insights
			\State Adaptive Learning Rate Optimization with LLM Insights
			\State Predictive Feature Interaction Discovery with LLM Analysis
			\EndWhile
			
			\State \textbf{Postprocessing:}
			\State Model Explanation Enhancement with LLM
			\State Prediction Confidence Estimation with LLM
			\State Adaptive Anomaly Detection in Residuals with LLM
			\EndProcedure
		\end{algorithmic}\label{fig:gradient-boosting-Algogen-pseudocode}
	\end{algorithm}

	\begin{figure}
		\centering
		\includegraphics[width=0.7\textwidth]{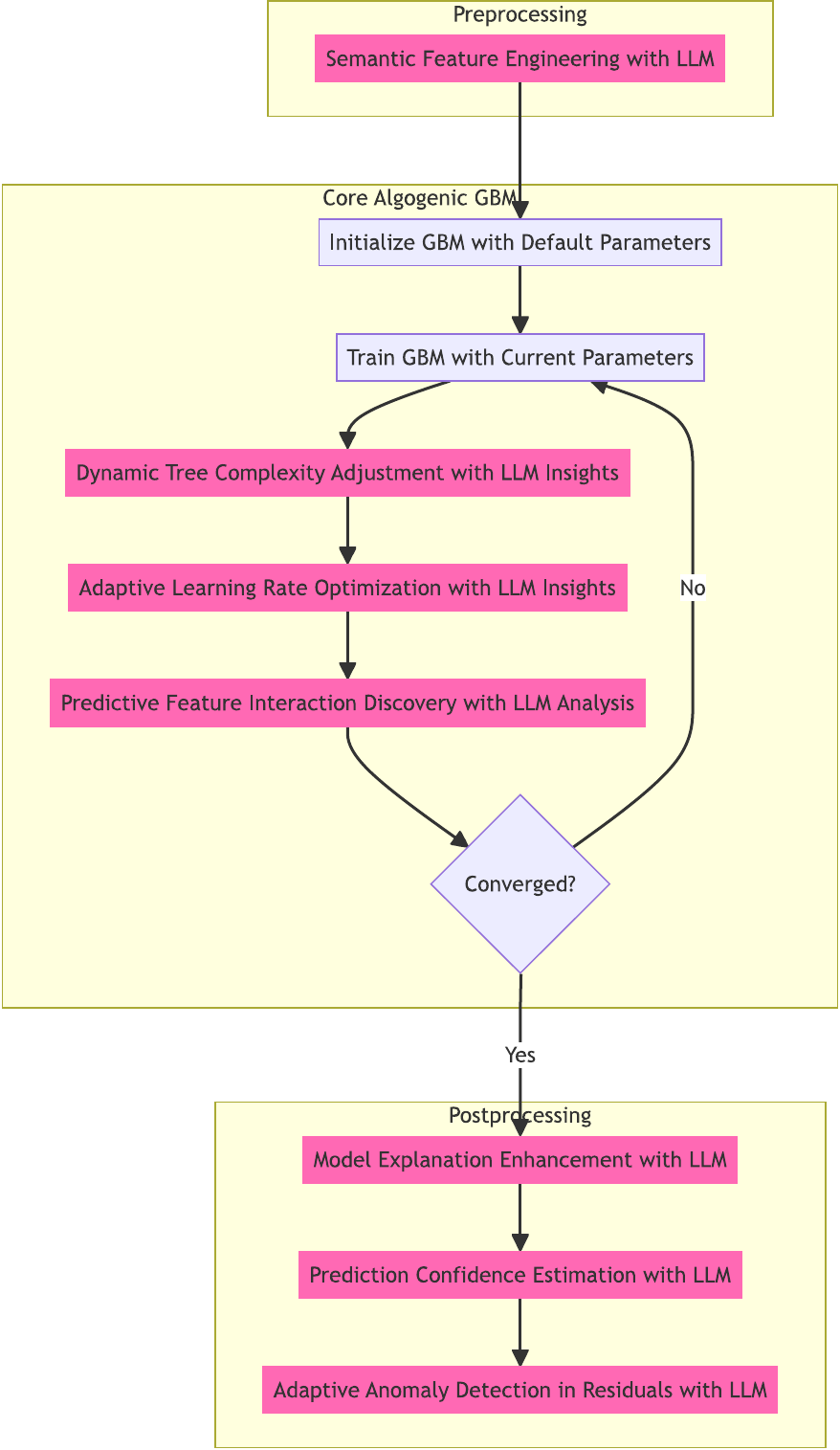} 
		\caption{Innovating GBMs with Algogenic Enhancements: This figure delineates the advanced integration of Algogenic enhancements with Gradient Boosting Machines (GBMs). It highlights the pivotal role of Large Language Models across three main phases: preprocessing, core training, and post-processing. In preprocessing, LLMs enrich feature engineering by unveiling deep semantic relationships. During core training, dynamic adjustments—such as tree complexity, learning rate optimization, and feature interaction discovery—are informed by LLM insights to optimize model performance. The post-processing phase employs LLMs to enhance model explanations, estimate prediction confidence, and detect anomalies in residuals, thereby elevating the GBM's predictive power, interpretability, and adaptability. This comprehensive integration exemplifies the synergistic potential of combining generative AI with traditional machine learning frameworks to address complex predictive tasks.}
		\label{fig:gbm}
	\end{figure}

	
	\chapterimage{pngs/deep_learning.png} 
	
	\chapter{Deep Learning Algogens}\index{Deep Learning Algogens}

	\section{Backpropagation}\index{Backpropagation}
	
	\subsection{Introduction to Backpropagation}
	\paragraph{Understanding Backpropagation}
	Backpropagation, often abbreviated as "backward propagation of errors," stands as a cornerstone algorithm in the realm of artificial neural network training. Its significance lies in its ability to facilitate the minimization of the discrepancy between predicted and actual outputs within supervised learning frameworks. This iterative process involves adjusting the weights of connections in the network to optimize performance. As data flows forward through the network during the training phase, backpropagation enables the computation of gradients with respect to the loss function, which are then propagated backward through the network. This backward propagation allows for the adjustment of weights using gradient descent or its variants, such as stochastic gradient descent or Adam optimization. Through this mechanism, neural networks can iteratively refine their parameters to improve predictive accuracy. Backpropagation's operational mechanism underscores its pivotal role in enhancing the learning capabilities of neural networks, making it a fundamental concept for practitioners and researchers alike.

	\paragraph{Operational Mechanism}
	Backpropagation operates through two main phases in the training process of a neural network: the forward pass and the backward pass. In the forward pass, input data is passed through the network, layer by layer, until the output layer produces a prediction. The prediction's accuracy is then assessed using a loss function, which quantifies the difference between the predicted output and the true output.
	
	The backward pass is where backpropagation truly comes into play. During this phase, the gradient of the loss function is calculated with respect to each weight in the network, effectively determining how much each weight contributes to the error. This gradient information is then propagated back through the network, from the output layer to the input layer, guiding how the weights should be adjusted to reduce the error. This process involves the use of the chain rule from calculus to efficiently compute gradients for each layer.
	
	\paragraph{Significance in Neural Network Training}
	The significance of backpropagation lies in its ability to systematically and efficiently optimize the weights of a neural network, with the ultimate goal of minimizing the loss function. This optimization is typically performed using gradient descent or variants thereof, where small, iterative adjustments to the weights are made in the direction that most reduces the error.
	
	Backpropagation is critical not only for its role in weight optimization but also for its general applicability across various types of neural networks, including feedforward neural networks, convolutional neural networks (CNNs), and recurrent neural networks (RNNs). Its ability to adapt the model to handle complex patterns and relationships within the data makes it indispensable in the field of deep learning.
	
	\paragraph{Applications and Impact}
	The backpropagation algorithm has been instrumental in numerous breakthroughs in machine learning and artificial intelligence, enabling advancements in image and speech recognition, natural language processing, and beyond. Its widespread adoption and continued relevance underscore its foundational impact on the development and success of neural networks.
	
	In summary, understanding backpropagation is essential for anyone involved in the design, implementation, and training of neural networks. Its mechanism for error correction and weight optimization is central to the learning process, allowing neural networks to learn from data and improve over time.
	
	\subsubsection{Mathematical Foundations}
	The mathematical underpinnings of backpropagation are rooted in calculus and linear algebra, providing a systematic approach to updating the weights in a neural network to minimize the loss function. This subsubsection explores the core mathematical concepts that enable backpropagation, including the derivation of gradients, the chain rule of calculus, and the iterative optimization process.
	
	\paragraph{Derivation of Gradients}
	The process of backpropagation initiates with the derivation of gradients, a fundamental step in training neural networks. These gradients represent the sensitivity of the loss function concerning each weight parameter within the network architecture. Essentially, they signify the rate of change of the loss function concerning alterations in individual weights. For a weight \(w\) in the network, the gradient is symbolized as \(\frac{\partial L}{\partial w}\), where \(L\) denotes the loss function. This mathematical expression elucidates the magnitude and direction of adjustments required in the weights to minimize the loss function. Precisely, the gradient vector points towards the direction of the steepest ascent of the loss function, guiding the optimization process towards minimizing the error. This step is pivotal for updating the weights iteratively during the training phase, as it provides insight into how modifications in each weight influence the overall network performance. Furthermore, it serves as the foundation for subsequent adjustments in the network's parameters, facilitating the convergence towards an optimal solution.

	\paragraph{Chain Rule of Calculus}
	The chain rule, a fundamental concept in calculus, serves as the cornerstone of the backpropagation algorithm, particularly crucial for efficiently computing gradients in deep neural networks. It articulates that the derivative of a composite function is the product of the derivatives of its individual components. In the context of neural networks, this rule enables the computation of gradients with respect to weights by sequentially multiplying gradients along the network layers. This recursive application of the chain rule, from the output layer back to the input layer, facilitates the efficient adjustment of weights during the training process, thereby optimizing the network's performance. By leveraging the chain rule, backpropagation effectively traces the impact of each weight on the overall loss, enabling the network to learn and adapt its parameters iteratively. This iterative adjustment, guided by gradients computed via the chain rule, embodies the essence of backpropagation, empowering neural networks to learn complex mappings and perform sophisticated tasks with remarkable accuracy and efficiency.

	\paragraph{Iterative Optimization Process}
	Once the gradients have been computed, the weights are updated in the opposite direction of the gradient to minimize the loss. This is typically done using gradient descent or variations thereof, such as stochastic gradient descent (SGD), Adam, or RMSprop. The update rule is generally of the form:
	\[w_{new} = w_{old} - \eta \frac{\partial L}{\partial w}\]
	where \(\eta\) is the learning rate, a small positive scalar determining the size of the step taken in the direction opposite to the gradient. Moreover, it's noteworthy that the choice of learning rate \(\eta\) significantly influences the convergence speed and stability of the optimization process. While a larger learning rate may lead to faster convergence, it also risks overshooting the minimum and causing oscillations. Conversely, a smaller learning rate might ensure more stable convergence but at the expense of slower progress. Therefore, selecting an appropriate learning rate is crucial for the effectiveness of the optimization algorithm. Additionally, techniques like adaptive learning rates, as employed in Adam and RMSprop, dynamically adjust the learning rate based on past gradients, allowing for a more efficient optimization process in non-convex and high-dimensional spaces.

	\paragraph{Impact on Neural Network Training}
	The mathematical principles guiding backpropagation ensure that each iteration of weight updates brings the neural network closer to the optimal set of weights that minimize the loss function. This iterative process of gradient computation and weight adjustment continues until the network converges to a state where the loss is minimized, or a predefined number of iterations is reached. Consequently, the efficiency of the training process heavily depends on the choice of optimization algorithm and hyperparameters. Furthermore, the complexity of the neural network architecture and the size of the training dataset also play significant roles in determining the training time and convergence behavior. Moreover, the presence of noisy or irrelevant data points may hinder convergence and lead to overfitting. Therefore, meticulous preprocessing and data cleaning are crucial steps in ensuring successful neural network training. Additionally, the availability of computational resources and hardware accelerators like GPUs can expedite the training process, allowing for faster experimentation and model iteration. Conversely, inadequate computational resources may result in prolonged training times or even prevent the training of large-scale models altogether.

	Understanding the mathematical foundations of backpropagation is crucial for designing effective neural networks. It not only informs the choice of network architecture and learning rate but also underlies advanced optimization strategies that can significantly enhance model performance.

	\subsubsection{Role in Neural Network Training}
	Backpropagation stands as the cornerstone of neural network training, enabling these models to learn from data and improve their performance on given tasks. This subsubsection illuminates the pivotal role of backpropagation in the neural network training process, detailing how it facilitates the effective adjustment of model weights and biases to minimize error.
	
	\paragraph{Facilitating Learning}
	At the heart of neural network training lies the objective to minimize the difference between the actual output and the predicted output by the model, often quantified using a loss function. Backpropagation is the mechanism through which information about the error is propagated back through the network, enabling the model to learn. By calculating the gradient of the loss function with respect to each parameter in the network, backpropagation provides a direction for how the weights should be adjusted to reduce error. Furthermore, this iterative process allows the network to refine its predictions iteratively, incrementally improving its performance over time. Moreover, backpropagation not only updates the weights but also adjusts the biases, ensuring that the model becomes more adept at capturing the underlying patterns in the data. Additionally, backpropagation is a fundamental concept in deep learning, forming the backbone of training algorithms in various neural network architectures. Hence, mastering backpropagation is crucial for understanding and effectively training neural networks, paving the way for advancements in artificial intelligence research and applications.

	\paragraph{Gradient Descent Optimization}
	Backpropagation is intrinsically linked with gradient descent optimization, a method that iteratively adjusts parameters to find the minimum of the loss function. Backpropagation computes the gradients necessary for gradient descent, guiding the optimization process by indicating the direction in which the parameters should be updated to decrease the loss.
	
	Furthermore, it's important to note that gradient descent optimization comes in various flavors, including stochastic gradient descent (SGD), mini-batch gradient descent, and batch gradient descent. Each variant has its own characteristics and trade-offs. Moreover, the choice of learning rate, which determines the size of the step taken in the parameter space during each iteration, significantly impacts the convergence and performance of the optimization algorithm. Thus, selecting an appropriate learning rate is crucial for the success of gradient descent optimization.
	
	Additionally, techniques such as momentum, which accelerates convergence by accumulating past gradients, and adaptive learning rate methods like AdaGrad, RMSprop, and Adam further enhance the efficiency and robustness of gradient descent optimization algorithms. These techniques address challenges such as oscillations, slow convergence, and getting stuck in local minima, making gradient descent optimization more effective in training deep neural networks.
	
	In summary, gradient descent optimization, enabled by backpropagation, plays a pivotal role in training neural networks by efficiently adjusting parameters to minimize the loss function. Moreover, the continual advancements in optimization techniques contribute to the effectiveness and scalability of deep learning models.

	\paragraph{Adapting to Diverse Architectures}
	The versatility of backpropagation extends beyond simple feedforward networks to more complex architectures such as convolutional neural networks (CNNs) and recurrent neural networks (RNNs). Despite the structural differences, backpropagation remains applicable, adjusting its calculations to account for the specific dynamics of each architecture. This adaptability underscores the algorithm's fundamental role in the broad spectrum of neural network applications.
	
	Furthermore, the integration of backpropagation into CNNs enables the extraction of hierarchical features from images, leveraging the local connectivity pattern and parameter sharing to efficiently learn spatial hierarchies. Moreover, in RNNs, backpropagation through time (BPTT) facilitates the training of networks capable of processing sequential data by unfolding them in time and propagating errors backward through the unfolded network. Additionally, the flexibility of backpropagation allows for the incorporation of various activation functions and regularization techniques tailored to the requirements of different architectures, ensuring robust learning and generalization. Consequently, whether dealing with image classification, natural language processing, or time-series prediction, the adaptive nature of backpropagation makes it a cornerstone in the advancement of diverse neural network paradigms.

	\paragraph{Improving Model Performance}
	Through the iterative application of backpropagation, neural networks gradually refine their weights and biases to better represent the mapping from inputs to outputs. This process of continuous improvement not only enhances the model's accuracy on the training data, but also, with proper regularization, helps in generalizing well to unseen data. Furthermore, regularization techniques such as L1 and L2 regularization aid in preventing overfitting by penalizing large weights, thereby promoting simpler models that capture essential patterns in the data. Moreover, techniques like dropout introduce randomness during training, forcing the network to learn more robust and diverse features, which contributes to better generalization. Additionally, employing techniques like batch normalization helps in stabilizing and accelerating the training process by normalizing the activations within each mini-batch. Consequently, through a combination of these methods, neural networks achieve improved performance not only on the training set but also on unseen data, ensuring better adaptability and reliability in real-world applications.

	\paragraph{Enabling Advanced Developments}
	Backpropagation's role extends beyond basic training; it is foundational to advancements in deep learning, including transfer learning, deep reinforcement learning, and unsupervised learning models. By enabling efficient error correction and model adjustment, backpropagation facilitates the exploration of complex neural network models and architectures, driving innovation and expanding the capabilities of artificial intelligence systems.
	
	In essence, backpropagation is the engine of learning in neural networks, enabling these models to adapt and evolve based on empirical data. Its central role in neural network training underscores its importance in the field of machine learning and artificial intelligence, serving as a key enabler of the remarkable advancements witnessed in recent years.
	
	\subsubsection{Applications and Limitations}
	Backpropagation has been pivotal in advancing the field of neural network training, enabling progress in diverse areas such as computer vision, natural language processing (NLP), speech recognition, and reinforcement learning. In computer vision, it trains convolutional neural networks for image classification and object detection. For NLP, it enhances recurrent neural networks and transformers for tasks like machine translation and text generation. Speech recognition technologies rely on backpropagation to convert spoken language into text accurately. Moreover, in reinforcement learning, it optimizes the neural networks that underpin decision-making algorithms.
	
	Despite its wide-ranging applications, backpropagation is subject to several limitations. Deep neural networks often encounter the vanishing or exploding gradients problem, which complicates the training process and may hinder convergence. The computational intensity of backpropagation, especially for large networks, necessitates substantial computational resources and time, posing a challenge for training complex models. The algorithm's reliance on differentiable functions limits the types of architectures and functions that can be incorporated into the network. Additionally, there is a risk of overfitting the training data if appropriate regularization techniques are not employed, affecting the model's ability to generalize well to unseen data.
	
	The field continues to evolve with advances in optimization algorithms, network architectures, and regularization techniques aimed at mitigating these limitations. Innovations such as dropout, batch normalization, and alternative activation functions have been developed to address the issues of overfitting and the vanishing/exploding gradient problem. Parallel computing frameworks and hardware advancements have also helped alleviate the computational demands associated with training neural networks using backpropagation.
	
	In essence, backpropagation's role in the development of neural networks is undeniable, driving significant technological advancements across multiple domains. However, understanding and overcoming its limitations remains a critical focus for researchers and practitioners aiming to harness the full potential of neural network technologies.
	
	\subsubsection{Pseudocode for Algorithmic Backpropagation}
	The Backpropagation Algorithm is a powerful technique used for training neural networks, particularly in the context of optimizing parameters. It operates through iterative refinement of these parameters by computing gradients of the loss function with respect to each weight in the network. This process is depicted in the pseudocode \ref{fig:bp-pseudocode}, where it is structured into two primary phases: the forward pass and the backward pass. During the forward pass, input data traverses through the network layers to generate an output. Subsequently, the loss between this output and the target value is computed. Moving to the backward pass, gradients of the loss function are calculated with respect to each weight in the network. The chain rule is employed here to efficiently propagate these gradient updates backward through the network layers, from the output to the input. Finally, the weights are adjusted in a direction that minimizes the error, with the learning rate $\eta$ governing the magnitude of each update step. This iterative process is repeated across multiple epochs, facilitating the gradual reduction of loss and enhancement of the model's predictive capabilities.
	
	\begin{algorithm}
		\caption{Algorithmic Backpropagation in Neural Networks}
		\begin{algorithmic}[1]
			\Procedure{Backpropagation}{$X, Y, \eta, epochs$}
			\State Initialize network weights (often small random values)
			\For{each epoch in epochs}
			\For{each $(x, y)$ in $(X, Y)$}
			\State Forward pass to compute output
			\State Compute loss between predicted and true $y$
			\State Backward pass to compute gradients
			\For{each weight $w$ in network}
			\State $w_{gradient} \gets$ Compute gradient of loss w.r.t. $w$
			\State $w \gets w - \eta \cdot w_{gradient}$ \Comment{Update weight}
			\EndFor
			\EndFor
			\EndFor
			\State \textbf{return} Updated network weights
			\EndProcedure
		\end{algorithmic}\label{fig:bp-pseudocode}
	\end{algorithm}

\subsection{Algogenic Enhancements for Backpropagation}

\subsubsection{Semantic Initialization of Weights}

\paragraph{Redefining Initial Weight Settings}
Initiating neural network weights with semantic relevance, rather than through traditional random or heuristic methods, offers a promising direction for enhancing the efficiency of the backpropagation algorithm. This method involves leveraging the contextual insights from Large Language Models to inform the initial weight settings, aligning them more closely with the semantic structure of the dataset. By doing so, it potentially reduces the number of epochs required to reach convergence, thus streamlining the training process.

\paragraph{Practical Implementation Considerations}
Implementing semantic weight initialization involves translating LLM-derived insights into actionable weight settings. This translation necessitates developing a methodology to quantitatively adjust weights based on qualitative data insights, a challenging task that requires innovative approaches to bridge this gap. The practical value of this enhancement lies in its potential to expedite the training process and improve the model's performance from the outset by providing a more informed starting point.

\paragraph{Skeptical Evaluation of Efficiency Gains}
While the theoretical underpinnings suggest efficiency gains, it is prudent to approach the implementation of semantic weight initialization with caution. The process of translating LLM insights into weight adjustments is complex and may not always yield the expected improvements in convergence speed or final model accuracy. Rigorous testing and validation across various datasets and model architectures are necessary to empirically establish the effectiveness of this approach.

\subsubsection{Dynamic Learning Rate Adjustment}

\paragraph{Adaptive Learning Rate Strategies}
Dynamic adjustment of the learning rate, based on ongoing training performance and semantic understanding of the task, presents a nuanced method to optimize the backpropagation process. By adjusting the learning rate in real-time, this strategy aims to strike a balance between exploration and exploitation in the model's learning trajectory, potentially reducing training time and improving model accuracy.

\paragraph{Operationalizing Adaptive Adjustments}
The operational challenge lies in accurately determining the optimal moments and magnitudes of learning rate adjustments. This requires a complex interplay between the LLM's semantic analysis of training progress and the algorithm's current state, necessitating advanced algorithms that can interpret and act on these insights in real-time.

\paragraph{Critical Perspective on Dynamic Adjustments}
The benefits of dynamic learning rate adjustment, while conceptually appealing, must be critically evaluated in practice. The effectiveness of this approach is highly dependent on the ability to accurately interpret and respond to the nuanced signals regarding training progress and semantic understanding, a non-trivial challenge that may limit its applicability and effectiveness in certain contexts.

\subsubsection{Contextual Error Analysis and Weight Adjustment}

\paragraph{Enhancing Model Training with Contextual Insights}
Incorporating contextual insights into error analysis and weight adjustment represents a sophisticated enhancement to the backpropagation algorithm. This approach leverages LLMs to provide a deeper understanding of errors, enabling targeted adjustments that go beyond simple gradient-based updates. Such contextual adjustments could lead to more nuanced model improvements, enhancing learning efficiency and model performance.

\paragraph{Implementing Contextual Adjustments}
The practical implementation of this enhancement involves developing methodologies to apply LLM-derived insights to the weight adjustment process in a meaningful way. This could include identifying specific patterns of errors that are indicative of underlying issues and adjusting weights in a way that directly addresses these issues, a process that requires a delicate balance between generalization and overfitting.

\paragraph{Evaluating the Impact of Contextual Insights}
The potential of contextual insights to improve training processes must be evaluated with a degree of skepticism. While the approach offers a promising avenue for enhancing model training, the actual impact on model performance and training efficiency can vary significantly across different datasets and model architectures. Rigorous empirical evaluation is essential to validate the effectiveness of incorporating contextual error analysis into the backpropagation process.

\subsubsection{Enhanced Gradient Optimization with Semantic Insights}

\paragraph{Integrating Semantic Insights into Optimization}
Leveraging semantic insights from LLMs to enhance the gradient optimization process represents an innovative approach to improving backpropagation efficiency. By aligning gradient adjustments with semantic structures within the data, this method aims to optimize the learning process, potentially leading to faster convergence and improved model performance.

\paragraph{Operational Challenges and Solutions}
Operationalizing this enhancement involves developing mechanisms to integrate semantic insights into the gradient optimization process. This requires not only the ability to derive meaningful insights from LLMs but also the development of algorithms capable of applying these insights to adjust gradients in a way that positively influences the learning trajectory.

\paragraph{Critical Appraisal of Semantic Gradient Optimization}
While the integration of semantic insights into gradient optimization holds promise, it is important to critically assess its practical effectiveness. The complexity of accurately incorporating semantic insights into the optimization process presents a significant challenge, and the actual benefits in terms of training efficiency and model performance may vary. Empirical validation across a wide range of tasks and datasets is crucial to establishing the value of this approach.

\subsubsection{Predictive Layer Adjustment}

\paragraph{Dynamic Architectural Adjustments}
Predictive Layer Adjustment, which dynamically modifies neural network architecture based on LLM insights, introduces a flexible approach to optimizing model structure during training. This method aims to adapt the network's architecture to better match the complexity of the task at hand, potentially enhancing model efficiency and performance.

\paragraph{Implementing Architectural Adjustments}
The implementation of predictive layer adjustment requires a robust framework for integrating LLM-derived predictions into the model training process. This involves developing algorithms capable of making informed decisions about architectural adjustments and applying these changes in a way that supports continuous learning and improvement.

\paragraph{Skeptical Review of Predictive Adjustments}
The efficacy of predictive layer adjustment, while theoretically appealing, necessitates a careful and critical evaluation. The actual impact on model performance and training efficiency can be highly context-dependent, and the challenges associated with dynamically altering model architecture during training must not be underestimated. Rigorous testing and validation are essential to assess the practical benefits of this approach.

\subsubsection{Model Debugging and Interpretability Enhancement}

\paragraph{Improving Model Transparency}
Enhancing model debugging and interpretability through Algogenic means involves leveraging LLMs to provide insights into the decision-making processes of neural networks. This approach aims to make models more transparent and understandable, facilitating easier debugging and increasing trust in model decisions.

\paragraph{Operational Aspects of Enhancing Interpretability}
Operationalizing this enhancement requires the development of methodologies for extracting and presenting LLM-derived insights in a way that is meaningful and accessible to practitioners. This involves not only identifying the relevant insights but also translating these into actionable information that can improve model debugging and interpretability.

\paragraph{Critical Considerations for Interpretability Enhancements}
The potential of LLMs to enhance model interpretability must be critically examined. While offering a promising route to making neural networks more transparent, the actual effectiveness of these enhancements in improving understanding and trust in model decisions can vary. A rigorous evaluation of the impact on debugging efficiency and model transparency is essential to validate the approach.

\subsubsection{Adaptive Feedback Loop for Continuous Learning}

\paragraph{Continuous Learning through Feedback}
Establishing an adaptive feedback loop for continuous learning entails using LLMs to monitor and provide feedback on model performance over time. This approach aims to ensure that neural networks remain adaptive and relevant, continually refining their performance in response to new data and evolving challenges.

\paragraph{Implementing Continuous Learning Mechanisms}
Operationalizing continuous learning through an adaptive feedback loop requires mechanisms for integrating real-time performance feedback into the training and adaptation process. This involves developing algorithms that can interpret feedback from LLMs and apply it to continuously refine the model, a process that must be carefully managed to avoid overfitting or destabilizing the model.

\paragraph{Evaluating the Effectiveness of Continuous Learning}
The benefits of an adaptive feedback loop for continuous learning, while conceptually appealing, require careful and critical evaluation. The effectiveness of this approach in maintaining model relevance and performance over time is highly dependent on the quality of the feedback and the ability of the model to adapt without compromising stability. Rigorous empirical validation is necessary to establish the practical value of continuous learning mechanisms in neural network training.

	\subsubsection{Pseudocode for Algogenic Backpropagation}
	The Algogenic backpropagation approach harnesses AI to enhance traditional backpropagation methods by dynamically adjusting learning parameters and strategies based on the observed behavior of the system and real-time error estimates. This pseudocode, available in \ref{fig:backpropagation-Algogen-pseudocode}, outlines an advanced framework incorporating AI-driven enhancements for adaptive learning rates, weight initialization, activation functions, and real-time parameter optimization.
	
	\begin{algorithm}
		\caption{Algogenic Backpropagation Framework Pseudocode}
		\begin{algorithmic}[1]
			\Procedure{AlgogenicBackpropagation}{Dataset}
			\State \textbf{Preprocessing:}
			\State Semantic Initialization of Weights using LLM insights
			
			\State \textbf{Core Training:}
			\State Initialize neural network with semantically informed weights
			\While{not Converged}
			\State Perform backpropagation to update weights
			\State Apply Dynamic Learning Rate Adjustment with LLM insights
			\State Conduct Contextual Error Analysis for weight adjustment
			\State Enhance Gradient Optimization with Semantic Insights
			\State Adjust network architecture predictively if needed
			\State \textit{Check for convergence}
			\EndWhile
			
			\State \textbf{Postprocessing:}
			\State Utilize LLM for Model Debugging and Interpretability Enhancement
			\State Implement Adaptive Feedback Loop for continuous learning
			\EndProcedure
		\end{algorithmic}\label{fig:backpropagation-Algogen-pseudocode}
	\end{algorithm}
	
	\begin{figure}
		\centering
		\includegraphics[width=0.75\textwidth]{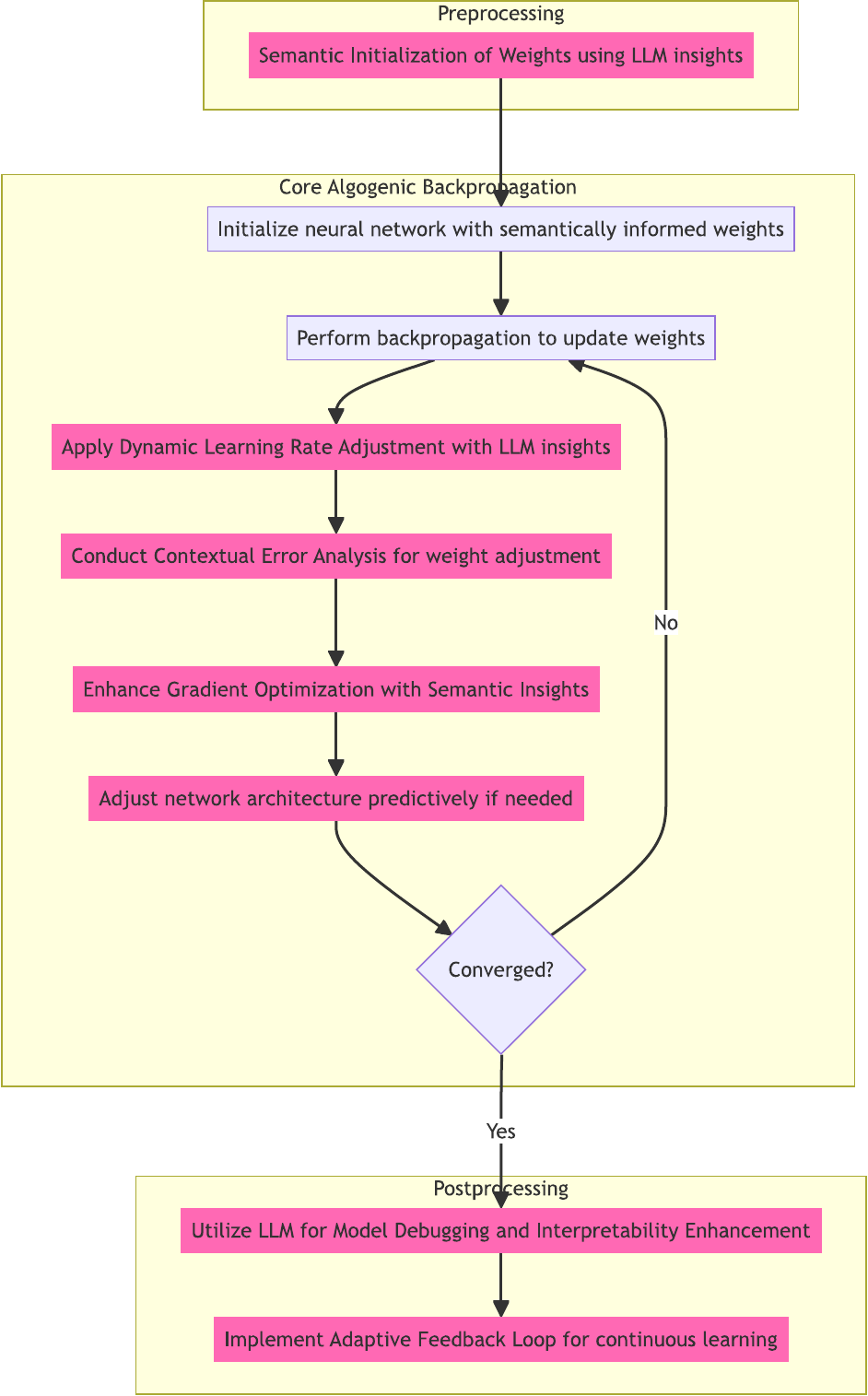} 
		\caption{Integrating Algogenic Enhancements in Backpropagation: This diagram visualizes the innovative application of Algogenic enhancements to the backpropagation algorithm, leveraging the capabilities of Large Language Models. It outlines a comprehensive framework that spans preprocessing, core training, and post-processing phases. In preprocessing, semantic insights from LLMs inform weight initialization, setting a foundation for more effective learning. The core training phase is enriched with dynamic learning rate adjustments, contextual error analysis, and architectural adjustments, all driven by LLM analysis to optimize neural network training. Post-processing utilizes LLMs for model debugging, interpretability enhancement, and establishing an adaptive feedback loop for continuous learning. This holistic integration underscores the transformative potential of combining generative AI with traditional neural network training to enhance efficiency, adaptability, and understanding.}
		\label{fig:backpropagation}
	\end{figure}

	\section{Convolutional Neural Networks (CNNs)}\index{Convolutional Neural Networks (CNNs)}
	\subsection{Introduction to CNNs}
	\subsubsection{The Concept of Convolutional Neural Networks}
	
	\paragraph{Foundational Overview}
	Convolutional Neural Networks (CNNs) stand as a cornerstone in the field of deep learning, particularly renowned for their prowess in processing data with a grid-like topology, such as images. At their core, CNNs employ a mathematical operation known as convolution, which systematically applies filters to the input data to extract meaningful features such as edges, textures, and patterns. This process allows CNNs to capture local spatial dependencies effectively, enhancing their ability to discern intricate patterns within images. Moreover, the hierarchical nature of feature extraction in CNNs enables them to progressively learn abstract representations of the input data, starting from simple features like edges and gradually discerning more complex structures as the layers deepen. This hierarchical representation learning is vital for tasks like image recognition, where objects can be defined by a combination of simpler visual elements. Consequently, CNNs have found widespread applications in various domains beyond computer vision, including natural language processing and speech recognition, highlighting their versatility and efficacy in capturing complex patterns in diverse data modalities.

	\paragraph{Mathematical Underpinnings}
	The convolution operation, which is central to CNNs, can be mathematically represented as the element-wise multiplication of a filter matrix, or kernel, with portions of the input data, followed by a summation. For an input matrix $X$ and a filter $F$, the convolution operation at a position \((i, j)\) is given by:
	\[
	(C * F)(i, j) = \sum_m \sum_n X_{i+m, j+n}F_{m, n}
	\]
	where the sums run over the dimensions of the filter. This operation is applied across the entire input, producing a feature map that highlights the presence of specific features encoded by the filter. Pooling layers follow convolutional layers to reduce the dimensionality of these feature maps, summarizing their most important information and making the network more robust to variations in the input.
	
	\paragraph{Architectural Composition}
	A typical CNN architecture is composed of several layers, each serving a distinct purpose in the feature extraction and learning process. The initial layers are usually convolutional layers paired with activation functions like ReLU to introduce non-linearities, enabling the network to learn complex patterns. And pooling layers intersperse convolutional layers to reduce spatial dimensions and computational load. Towards the network's end, fully connected layers integrate the high-level features extracted by previous layers to make predictions or classifications. This hierarchical structure allows CNNs to effectively learn from a wide range of input data, adapting to the specifics of the task at hand through training. Additionally, the cascading arrangement of layers ensures that each layer can build upon the representations learned by the preceding layers, gradually capturing intricate features and relationships in the input data. Furthermore, the utilization of convolutional and pooling layers in tandem facilitates the extraction of spatial hierarchies and local patterns, which are essential for tasks such as image recognition and object detection. Moreover, the integration of fully connected layers at the network's top enables comprehensive feature aggregation and decision-making, culminating in accurate predictions or classifications. Thus, the architectural design of CNNs embodies a synergistic blend of convolutional, pooling, and fully connected layers, orchestrated to efficiently process diverse input data and yield meaningful insights.

	\paragraph{Training and Learning Process}
	The training process of Convolutional Neural Networks (CNNs) is a fundamental aspect of their development. It involves a meticulous adjustment of the weights associated with the filters spread across all layers. These adjustments are made iteratively to minimize a predefined loss function, which effectively quantifies the disparity between the network's predictions and the actual labels of the training dataset. Backpropagation, in conjunction with optimization algorithms such as stochastic gradient descent, plays a pivotal role in this weight adaptation process. By leveraging backpropagation, CNNs efficiently compute the gradients of the loss function with respect to each weight, thereby enabling precise adjustments that steer the network towards improved performance.
	
	Moreover, the iterative nature of the training process allows CNNs to gradually refine their filters, emphasizing features that are most pertinent to the task at hand. This iterative refinement is crucial for the network to discern and extract the salient characteristics embedded within the input data. Consequently, CNNs exhibit a remarkable capability to learn and adapt to the intricacies of diverse datasets, enabling them to generalize well to unseen examples.
	
	Furthermore, as CNNs progress through the training epochs, they progressively fine-tune their filters to capture the essence of the input data. This fine-tuning process entails adjusting the filter parameters to enhance their sensitivity to relevant features while suppressing noise and irrelevant information. Consequently, CNNs become increasingly adept at discerning complex patterns and structures within the data, ultimately enhancing their predictive accuracy and robustness.
	
	In summary, the training process of CNNs is a multifaceted journey characterized by meticulous weight adjustments, iterative refinement of filters, and progressive fine-tuning, all orchestrated to optimize the network's predictive capabilities and adaptability to diverse datasets.

	\paragraph{Applications and Evolution}
	Convolutional Neural Networks (CNNs) have undeniably catalyzed a paradigm shift in computer vision, redefining the boundaries of what machines can discern from visual data. Through their intricate architectures and hierarchical feature extraction mechanisms, CNNs have not only achieved remarkable feats in tasks like image classification, where they discern objects and scenes with unprecedented accuracy, but also in object detection, enabling real-time identification and localization of multiple objects within images. Moreover, in semantic segmentation, CNNs excel in understanding the pixel-level semantics of images, crucial for applications like autonomous driving and medical imaging.
	
	Beyond their dominance in computer vision, CNNs have transcended disciplinary boundaries, finding utility in diverse domains such as natural language processing (NLP) and time series analysis. In NLP, CNNs are adept at learning hierarchical representations of text, capturing syntactic and semantic features that facilitate tasks like sentiment analysis, document classification, and machine translation. Similarly, in time series analysis, CNNs demonstrate prowess in recognizing temporal patterns and extracting meaningful features from sequential data, empowering applications in financial forecasting, signal processing, and environmental monitoring.
	
	The relentless evolution of CNNs is propelled by a confluence of factors: the relentless pursuit of deeper architectures, novel regularization techniques, and the abundance of computational resources. As deep learning research continues to push the boundaries of model complexity and scalability, CNNs evolve to tackle increasingly intricate tasks and datasets. Moreover, the democratization of computational resources through cloud computing and specialized hardware accelerators fosters widespread adoption and experimentation with CNNs, fueling innovation across industries and research domains.
	
	In essence, the trajectory of CNNs epitomizes the symbiotic relationship between theoretical advancement and practical application in artificial intelligence. As researchers delve deeper into the intricacies of convolutional architectures and feature hierarchies, the applications of CNNs proliferate, permeating every facet of modern technological landscape, from autonomous systems to personalized medicine, underscoring their indelible imprint on the fabric of AI-driven innovation.

	\subsubsection{Key Principles and Mechanisms}
	
	\paragraph{Convolutional Layers: The Core}
	The core principle behind Convolutional Neural Networks (CNNs) lies in their unique structure, particularly the convolutional layers that perform the bulk of feature extraction. These layers use filters or kernels that slide across the input image, systematically applying the convolution operation to capture spatial hierarchies of features. Each filter is designed to detect specific types of features at various levels of abstraction, from simple edges and textures in early layers to complex objects and patterns in deeper layers. \textbf{Furthermore}, the convolution operation's efficiency stems from its ability to preserve the spatial relationship between pixels, making CNNs highly effective for tasks involving image data. \textbf{Moreover}, the hierarchical nature of feature extraction in CNNs allows them to learn increasingly complex representations of the input data, enabling them to discern intricate patterns and structures. \textbf{In addition}, the use of pooling layers after convolutional layers helps in reducing the spatial dimensions of the feature maps while retaining the most relevant information, contributing to the network's efficiency and generalization capabilities. Thus, CNNs leverage the power of convolutional layers to efficiently extract meaningful features from raw input data, making them indispensable in various computer vision tasks.

	\paragraph{Activation Functions: Introducing Non-linearity}
	Following the convolution operation, an activation function is applied to introduce non-linearities into the model, enabling it to learn and represent more complex patterns. The Rectified Linear Unit (ReLU) is commonly used for its simplicity and effectiveness in facilitating faster convergence during training. By applying a non-linear transformation, activation functions allow CNNs to compile and interpret the linear combinations of features extracted by the convolutional layers, contributing to the network's overall ability to discern and classify varied and complex inputs.
	
	\textbf{Moreover}, activation functions play a crucial role in mitigating the vanishing gradient problem often encountered in deep neural networks. This problem arises due to the repeated application of linear transformations, which can cause gradients to shrink exponentially during backpropagation, hindering the learning process. However, with the introduction of non-linearities through activation functions like ReLU, the gradients are kept within a reasonable range, ensuring more stable and efficient training.
	
	\textbf{Furthermore}, the choice of activation function can significantly impact the network's performance and convergence speed. While ReLU is widely favored for its simplicity and effectiveness, other activation functions such as Sigmoid and Tanh are also employed in specific scenarios. Sigmoid functions are commonly used in binary classification tasks, where the output needs to be within the range [0, 1], representing probabilities. On the other hand, Tanh functions, which squash the output to the range [-1, 1], are suitable for tasks where the input data is standardized or centered around zero, preventing saturation of gradients.
	
	\textbf{Additionally}, the introduction of non-linearities enables CNNs to capture complex relationships between features, facilitating better generalization to unseen data. This capability is essential in tasks such as image classification, where the model needs to recognize objects under various conditions such as different viewpoints, lighting conditions, and occlusions.

	\paragraph{Pooling Layers: Reducing Dimensionality}
	Pooling layers serve as crucial components in the convolutional neural network (CNN) architecture, facilitating the reduction of spatial dimensions within the feature maps produced by preceding convolutional layers. This reduction in dimensionality is pivotal for effectively summarizing the essential information while concurrently diminishing the computational burden imposed on subsequent layers. Among the various pooling techniques available, max pooling stands out prominently owing to its innate capability to extract the most significant features. By selecting the maximum value from a predetermined window of pixels, max pooling ensures that the salient characteristics of the feature maps are preserved, thereby enhancing the network's ability to discern important patterns amidst input variations and translations.
	
	Moreover, pooling operations contribute significantly to the overall robustness of the CNN architecture. By abstracting the spatial information and emphasizing the most relevant features, pooling layers play a pivotal role in ensuring that the network remains invariant to slight variations in the input data. This robustness is essential for tasks such as image classification, where the network must accurately identify objects despite differences in their positions or orientations within the input images.
	
	In essence, pooling layers act as effective mechanisms for dimensionality reduction and feature abstraction, enabling CNNs to focus on the most discriminative aspects of the data while alleviating computational overhead. Through techniques like max pooling, these layers serve as critical components in the success of convolutional neural networks, enhancing their ability to extract meaningful representations from complex datasets.

	\paragraph{Fully Connected Layers: Integration for Decision Making}
	Towards the end of a CNN architecture, fully connected layers play a pivotal role in synthesizing the abstract representations of features extracted by preceding layers. These layers serve as the nexus where high-level features are consolidated and processed to facilitate final predictions or classifications. Unlike convolutional layers that focus on local patterns, fully connected layers embrace a global perspective by connecting each neuron to every neuron in the preceding layer. This extensive interconnection imbues the network with a holistic understanding of the input data, enabling it to leverage the entirety of extracted features in decision-making.
	
	The comprehensive integration afforded by fully connected layers ensures that no relevant information is overlooked during the decision-making process. By aggregating inputs from all neurons in the preceding layer, these layers consider the full spectrum of features, encompassing both subtle nuances and prominent characteristics present in the input data. Consequently, the network's ability to discern intricate patterns and nuances is greatly enhanced, leading to more accurate and robust predictions.
	
	Moreover, the utilization of fully connected layers facilitates the incorporation of contextually rich information into the decision-making process. Each neuron in these layers processes a weighted combination of features, allowing for nuanced feature interactions and hierarchies to be captured. This holistic approach to feature integration fosters a deeper understanding of the input data, enabling the network to discern complex relationships and make informed decisions.
	
	In essence, fully connected layers serve as the linchpin of the CNN architecture, orchestrating the integration of abstract features to drive the decision-making process. Their capacity to consolidate and process information from preceding layers empowers the network to extract meaningful insights from the input data, ultimately culminating in accurate predictions or classifications.

	\paragraph{Backpropagation and Optimization: Learning from Errors}
	The training of Convolutional Neural Networks (CNNs) relies heavily on backpropagation, a fundamental technique for efficiently computing gradients of the loss function concerning each weight in the network. This method enables the network to learn from errors by iteratively adjusting its parameters based on the computed gradients. Backpropagation facilitates the propagation of error signals backward through the network, allowing adjustments to be made in the direction that reduces the overall loss, thereby refining the network's predictions.
	
	Coupled with backpropagation are optimization algorithms such as stochastic gradient descent (SGD) or Adam, which play a crucial role in updating the network's parameters. These algorithms leverage the gradients computed via backpropagation to iteratively minimize the loss function. SGD updates the weights by taking small steps in the direction opposite to the gradient, gradually converging towards a local minimum. Adam, on the other hand, adapts the learning rates for each parameter based on their past gradients, enhancing the convergence speed and stability of the training process.
	
	Through the synergy of backpropagation and optimization algorithms, CNNs adjust their filter weights to minimize the discrepancy between predicted and actual labels. This process is essential for the network to learn meaningful representations of the input data. By iteratively fine-tuning the weights, CNNs become adept at capturing intricate patterns and features present in the data, thereby improving their ability to make accurate predictions. Consequently, CNNs exhibit an adaptive and dynamic learning mechanism, continually refining their internal representations to better align with the complexities of the input data.

	\subsubsection{The Role of Convolution and Pooling Layers}
	
	\paragraph{Convolution Layers: Feature Detectors}
	Convolution layers serve as the foundation of Convolutional Neural Networks (CNNs), functioning as advanced feature detectors crucial for image recognition tasks. These layers employ a collection of learnable filters or kernels, each with its unique weights, which systematically convolve across the input image. By sliding across the image, these filters apply a mathematical operation that captures local patterns, enabling the network to discern intricate details within the data.
	
	The essence of convolution layers lies in their ability to extract spatial and temporal hierarchies inherent in the input data. This hierarchical representation facilitates the recognition of fundamental features such as edges, colors, and textures, as well as more complex structures as the network deepens. Through a process of feature abstraction, each filter specializes in detecting particular patterns, enhancing the network's capability to discern meaningful information from raw pixel values.
	
	Moreover, the utilization of convolution layers introduces parameter sharing, a key concept that promotes efficiency and robustness within CNNs. By sharing weights across different spatial locations, these layers drastically reduce the number of trainable parameters, thereby mitigating the risk of overfitting while enhancing generalization performance.
	
	In essence, convolution layers act as feature extractors by systematically analyzing the input data, enabling CNNs to discern intricate patterns and hierarchies crucial for accurate classification and recognition tasks.

	\paragraph{Pooling Layers: Spatial Hierarchy Simplification}
	Pooling layers, positioned strategically after convolution layers, serve a pivotal role in streamlining the spatial dimensions of the convolved features. Through operations like max pooling or average pooling, these layers extract essential information by summarizing the presence of features across non-overlapping subregions of the input. This strategic downsizing of spatial dimensions serves multiple purposes within the neural network architecture.
	
	Firstly, by reducing the spatial size, pooling layers effectively curtail the computational burden for subsequent layers. This downsampling operation contributes significantly to computational efficiency, making the network more manageable and less prone to overfitting, especially in scenarios where computational resources are limited.
	
	Secondly, the condensation of feature representation achieved by pooling layers is crucial for extracting the most salient features from the input data. By focusing on the most prominent features while discarding less relevant information, these layers enhance the network's ability to discern meaningful patterns within the data. This process of feature abstraction aids in generalization, enabling the network to perform effectively across a diverse range of inputs.
	
	Furthermore, pooling layers play a pivotal role in establishing scale and orientation invariance within the network. By summarizing features across local regions, these layers ensure that the network's perception remains consistent regardless of variations in scale or orientation of the input. This property is particularly beneficial in tasks such as object recognition, where objects may appear in different sizes or orientations within an image.
	
	In essence, pooling layers contribute significantly to the spatial hierarchy of features established by convolution layers. They streamline the representation of features, reduce computational complexity, enhance generalization capabilities, and promote scale and orientation invariance, thereby facilitating effective feature extraction and pattern recognition within neural networks.

	\paragraph{Synergy Between Convolution and Pooling}
	The interplay between convolution and pooling layers defines the operational essence of CNNs. Convolution layers extract a rich set of features from the input data, producing feature maps that encapsulate the detected patterns. Pooling layers then simplify these feature maps, reducing their dimensions while preserving the most critical information. Moreover, this synergy not only enhances the computational efficiency of CNNs but also contributes significantly to their robustness against variations in input data. By strategically combining convolutional and pooling operations, CNNs can effectively mitigate overfitting by reducing the spatial dimensions of feature maps, thereby promoting generalization to unseen data. Furthermore, pooling layers facilitate translation invariance, allowing CNNs to recognize patterns regardless of their spatial location within the input data. Consequently, CNNs equipped with convolution and pooling layers can efficiently process large volumes of data, extracting and condensing information layer by layer. This systematic approach not only enhances the interpretability of learned features but also enables CNNs to capture hierarchical representations of input data, ranging from simple edges to complex objects. Thus, the symbiotic relationship between convolution and pooling layers empowers CNNs to achieve superior performance across a wide range of tasks, including image classification, object detection, and semantic segmentation.

	\paragraph{Impact on Network Performance and Generalization}
	The combination of convolution and pooling layers significantly impacts the network's performance and its ability to generalize across different inputs. Convolutional Neural Networks (CNNs) leverage the convolution operation to extract spatial hierarchies of features from input data. This process enables the network to capture local patterns, such as edges and textures, and gradually combine them to form higher-level representations of objects or concepts. \textbf{Additionally}, pooling layers play a crucial role in downsampling the feature maps generated by convolutional layers, reducing the spatial dimensions while preserving the most salient information. \textbf{Furthermore}, the hierarchical structure fostered by convolution and pooling layers allows CNNs to abstract and learn increasingly complex representations of visual data. This hierarchical learning is instrumental in achieving robustness to variations in input, such as changes in scale, rotation, or illumination, enhancing the network's generalization capabilities across diverse datasets and scenarios.
	
	The synergy between convolution and pooling operations empowers CNNs to excel in various computer vision tasks, including image classification, object detection, and semantic segmentation. \textbf{Moreover}, the ability of CNNs to automatically learn relevant features from raw data reduces the need for handcrafted feature engineering, making them adaptable to different domains and applications. This adaptability, coupled with their effectiveness in processing large-scale datasets, has propelled CNNs to the forefront of modern machine learning and artificial intelligence research.
	
	In conclusion, the integration of convolution and pooling layers forms the backbone of CNN architecture, enabling efficient feature extraction, hierarchical learning, and robust generalization across diverse visual inputs. The continuous advancement of CNNs and their applications underscores their significance in shaping the landscape of computer vision and machine learning.

	\subsubsection{Applications and Limitations}
	
	\paragraph{Wide-ranging Applications}
	Convolutional Neural Networks (CNNs) have revolutionized the field of computer vision, boasting a wide array of applications that span beyond image recognition to include video analysis, natural language processing, and medical image diagnosis. CNNs have become indispensable tools in image and video recognition tasks, leveraging their robust architectures to excel at identifying objects, classifying images into categories, and detecting anomalies or specific events. \textbf{Moreover}, their capability to process spatial hierarchies efficiently makes them \textbf{ideal} for applications such as facial recognition systems, autonomous vehicle navigation, and surveillance, where real-time and accurate identification are crucial. 
	
	\textbf{In addition}, CNNs have demonstrated remarkable success in medical imaging by facilitating the detection and diagnosis of diseases from MRI and CT scans. This contribution significantly enhances the accuracy and efficiency of medical evaluations, aiding healthcare professionals in making informed decisions \textbf{furthermore}. The adaptability of CNNs to process sequential data \textbf{also} extends their applicability to natural language tasks. In this domain, they prove useful for sentence classification, topic categorization, and even generating text descriptions for images, thereby \textbf{further} expanding their utility across various domains.
	
	\paragraph{Limitations and Challenges}
	Despite their versatility and power, CNNs are not without limitations. One of the primary challenges is the requirement for large labeled datasets to train the models effectively. This dependency on extensive data can be a significant barrier in domains where data is scarce, sensitive, or expensive to annotate. \textbf{Moreover}, CNNs, like many deep learning models, suffer from a lack of interpretability; their decision-making process is often described as a "black box," making it difficult to understand or explain the rationale behind their predictions. This issue is particularly problematic in critical applications such as healthcare, where explainability is crucial for trust and adoption. Another limitation is the computational cost associated with training and deploying CNNs, requiring significant resources in terms of memory and processing power, which can be a constraint for real-time applications or devices with limited capabilities. \textbf{Additionally}, CNNs are inherently susceptible to adversarial attacks, where slight, often imperceptible, alterations to the input data can lead to incorrect predictions, raising concerns about their robustness and security in sensitive applications.

	\paragraph{Overcoming Limitations}
	Efforts to overcome the limitations of CNNs have led to several advancements, including the development of transfer learning and data augmentation techniques to address the challenge of limited training data. Transfer learning allows CNNs to leverage knowledge gained from one task to perform another related task, reducing the need for large labeled datasets. Data augmentation artificially expands the training dataset by applying various transformations to the input images, enhancing the model's generalization capabilities. To tackle the issue of interpretability, researchers are exploring methods such as attention mechanisms and layer-wise relevance propagation to provide insights into the model's focus and decision-making process. Optimizations in model architecture and deployment strategies, including network pruning and quantization, aim to reduce the computational demands of CNNs, making them more accessible and efficient. Additionally, ongoing research into adversarial training seeks to fortify CNNs against malicious attacks, enhancing their robustness and reliability.
	
	\paragraph{Future Directions}
	The continuous evolution of CNNs, driven by both the challenges they face and their potential for innovation, promises to expand their applicability and effectiveness across a broader spectrum of tasks and domains. As advancements in hardware, algorithms, and understanding of deep learning progress, CNNs are set to play an even more significant role in shaping the future of technology and society.
	
	Furthermore, as computing hardware continues to advance, with the emergence of specialized accelerators such as TPUs and GPUs tailored for deep learning tasks, CNNs will benefit from increased computational power and efficiency. Moreover, ongoing research in algorithmic optimization, including advancements in model compression, pruning, and quantization techniques, will make CNNs more accessible for deployment on resource-constrained devices, extending their reach to edge computing and IoT applications.
	
	Additionally, the interdisciplinary nature of CNN research, drawing from fields such as computer vision, natural language processing, and reinforcement learning, will foster cross-pollination of ideas and methodologies, leading to hybrid models that leverage the strengths of different domains. Likewise, the exploration of novel architectures, such as attention mechanisms and capsule networks, will drive innovation in CNN design, enabling them to tackle complex tasks with greater precision and robustness.
	
	Consequently, the integration of CNNs into various sectors, including healthcare, finance, and autonomous systems, will revolutionize industries by enabling automation, improving decision-making processes, and unlocking new possibilities for innovation. Thus, the future trajectory of CNNs holds immense potential for societal impact, driving forward the frontiers of artificial intelligence and shaping the technological landscape for years to come.

	\subsubsection{Algorithmic Pseudocode for Convolutional Neural Networks}
	The Convolutional Neural Network (CNN) architecture serves as a powerful tool for image processing and classification tasks, operating through a series of convolutional and pooling layers followed by fully connected layers for classification. This structured approach, depicted in pseudocode \ref{fig:cnn-pseudocode}, begins with convolutions applied to the input image using predetermined filters, extracting diverse features and patterns from the image. After each convolutional operation, a Rectified Linear Unit (ReLU) activation function is applied to introduce non-linearity, facilitating the network's ability to capture intricate patterns within the data. Subsequently, pooling layers are employed to downsample the feature maps, reducing computational complexity while enhancing feature detection capabilities. The resulting output from the convolutional and pooling layers is then flattened and fed into fully connected layers, which leverage the learned features for classification purposes. Finally, the softmax function is applied in the output layer to generate a probability distribution across possible classes, enabling the network to make predictions based on the highest probability output.
	
	\begin{algorithm}
		\caption{Convolutional Neural Networks Pseudocode}
		\begin{algorithmic}[1]
			\Procedure{CNN}{InputImage, Filters, PoolSize, FullyConnectedLayers}
			\State Initialize weights for all filters randomly
			\For{each layer in CNN}
			\If{Convolutional Layer}
			\State Apply convolution operation using Filters
			\State Apply ReLU activation function
			\ElsIf{Pooling Layer}
			\State Apply Max Pooling with PoolSize
			\EndIf
			\EndFor
			\State Flatten the output for Fully Connected Layer input
			\For{each layer in FullyConnectedLayers}
			\State Apply weight and bias
			\State Apply ReLU activation function
			\EndFor
			\State Apply Softmax function for classification
			\State \Return Output of the network
			\EndProcedure
		\end{algorithmic}\label{fig:cnn-pseudocode}
	\end{algorithm}

\subsection{Previous Work on ML and AI Interplay with Convolutional Neural Networks}

\paragraph{DualConv: Enhancing Lightweight Deep Neural Networks}
The method known as DualConv was introduced in 2022 as an approach to enhance lightweight deep neural networks \cite{zhong2022dualconv}. By integrating dual convolutional kernels into the network architecture, DualConv aims to improve computational efficiency without sacrificing model accuracy. This method capitalizes on the convolutional layer's potential for optimization by simultaneously processing features with different receptive fields. The introduction of DualConv offers a scalable solution to balancing performance and computational demands, making advanced neural networks more accessible in resource-constrained environments.

\paragraph{Learning Strides in Convolutional Neural Networks}
In 2022, a study proposed an adaptive learning strategy for strides in convolutional neural networks (CNNs) \cite{riad2022learning}. This strategy dynamically adjusts stride values during the training phase to optimize the convolutional process. By allowing for flexible stride adjustments, the model demonstrates improved efficiency in processing spatial hierarchies of features. This research challenges traditional neural network configurations, opening new possibilities for algorithmic innovation and demonstrating the continuous interplay between machine learning and artificial intelligence.

\paragraph{Switchable Self-Attention Module}
The Switchable Self-Attention Module (SSAM) was introduced in 2022 as a mechanism to dynamically switch between self-attention and convolutional operations within neural networks \cite{zhong2022switchable}. This innovation allows the model to allocate computational resources based on task requirements, optimizing for either performance or efficiency. SSAM highlights the versatility of attention mechanisms in enhancing model performance and represents a milestone in the evolution of neural network architectures, offering insights into harmonizing efficiency with effectiveness in artificial intelligence.

\subsection{Algogenic Enhancements for CNNs}

\subsubsection{Semantic Data Augmentation}

\paragraph{Expanding Training Data through Semantic Understanding} To refine Convolutional Neural Networks (CNNs), we suggest augmenting training data by exploiting Large Language Models for a deeper semantic grasp of image data. LLMs, with their nuanced understanding of context, can pinpoint and fill the gaps in training datasets, ensuring a broader representation of scenarios. This semantic data augmentation, unlike traditional methods, leverages LLMs to generate semantically rich and diverse data, thereby enhancing the dataset's coverage without compromising label accuracy. This method promises a nuanced improvement in the CNN's ability to generalize from seen to unseen data, potentially bolstering its performance across varied applications.

\paragraph{Operationalizing Semantic Augmentation} Operationalizing semantic augmentation involves using LLMs to identify dataset deficiencies and generate descriptions for missing scenarios. These descriptions guide the creation or acquisition of new images, enriching the dataset. Incorporating LLM-generated insights enables a targeted approach to data augmentation, ensuring the new data is both relevant and varied. This process, while enhancing the dataset's diversity, maintains the integrity of labels, crucial for preserving the CNN's learning accuracy.

\paragraph{Enhancing Model Performance and Generalization} Augmenting training datasets semantically is anticipated to not only address data scarcity and bias but also enhance the CNN's generalization capabilities. By training on a dataset enriched with semantically diverse data, CNNs are expected to develop a more refined understanding of the visual world, leading to improved performance on various tasks. While promising, this approach requires careful implementation to ensure the augmented data accurately reflects real-world scenarios.

\subsubsection{Dynamic Filter Optimization}

\paragraph{Tailoring Convolutional Filters to Data Complexity} We propose dynamically optimizing CNN filters, informed by LLM analysis, to adaptively refine the network's feature extraction capabilities. By adjusting filters based on the semantic complexity identified by LLMs, this approach aims at a more efficient processing and improved accuracy. Dynamic filter optimization, though computationally demanding, is posited to allow CNNs to better capture nuanced features in complex datasets.

\paragraph{Implementing LLM-Guided Filter Adjustments} This entails LLMs analyzing the data and suggesting adjustments to the CNN's convolutional layers for enhanced feature extraction. Implementing these adjustments requires a flexible CNN architecture but promises to improve the network's responsiveness to the data's semantic intricacies, potentially leading to higher accuracy in tasks like image classification and object detection.

\paragraph{Enhancing CNN Adaptability and Performance} The dynamic optimization of filters, guided by LLM insights, is expected to enhance CNNs' adaptability and performance. This approach allows for a more nuanced and efficient processing of diverse datasets, potentially improving task-specific accuracy. However, its practical implementation involves balancing computational efficiency with the benefits of adaptability.

\subsubsection{Adaptive Activation Function Selection}

\paragraph{Customizing Non-linearity for Enhanced Learning} We suggest the adaptive selection of activation functions, guided by LLM insights, to optimize non-linearity in CNNs based on the data's characteristics. This approach aims to improve learning efficiency by dynamically choosing activation functions that best match the data's semantic content, potentially leading to improved network performance.

\paragraph{Operationalizing Activation Function Adaptation} Implementing this strategy involves LLMs evaluating the network's performance with different activation functions, recommending adjustments to optimize learning. While promising for enhancing model adaptability and performance, this strategy's success hinges on accurately matching activation functions with the data's semantic nuances.

\paragraph{Boosting CNN Performance Through Intelligent Non-linearity} Adaptive activation function selection, informed by LLM analysis, is anticipated to enhance CNNs' ability to learn complex patterns, thereby improving their performance and generalization capabilities. However, the practical benefits of this approach depend on the effective implementation of LLM recommendations, balancing the complexity of adaptive functions with the goal of improved performance.

\subsubsection{Contextual Regularization Adjustment}

\paragraph{Optimizing Model Complexity with Semantic Insights} We propose dynamically adjusting regularization techniques in CNNs, guided by LLM-derived insights into the data's semantic complexity. This approach aims to fine-tune model complexity for optimal performance, addressing overfitting or underfitting as indicated by LLM analysis. While promising for enhancing model generalization, the implementation must carefully balance regularization adjustments to avoid compromising model performance.

\paragraph{Implementing LLM-Guided Regularization Strategies} This involves LLMs monitoring training progress and recommending regularization adjustments to optimize model complexity. The approach promises improved model performance by dynamically tailoring regularization to the training data's needs. However, its effectiveness will depend on the precision of LLM recommendations and the ability to implement adjustments without introducing new biases or performance issues.

\paragraph{Advancing CNN Training Through Intelligent Regularization} Contextual regularization adjustment, informed by LLM analysis, is expected to improve CNNs' training efficiency and generalization. This approach, by dynamically adjusting regularization based on semantic insights, promises to enhance model robustness. However, the challenge lies in accurately interpreting LLM insights and effectively implementing suggested adjustments.

\subsubsection{Semantic Interpretation of Feature Maps}

\paragraph{Deciphering CNN's Visual Cognition} We suggest leveraging LLMs for semantic interpretation of CNN feature maps to bridge the gap between complex data representations and human-understandable concepts. This approach aims to enhance model transparency and trust by providing insights into the CNN's decision-making process. While promising for improving interpretability, this strategy requires careful implementation to ensure accurate and meaningful semantic interpretations.

\paragraph{Operationalizing Feature Map Interpretation} Implementing semantic interpretation involves using LLMs to analyze CNN activations and generate descriptive annotations. This approach promises to enhance understanding of the model's focus and decision-making process. However, its success hinges on the accuracy of LLM-generated interpretations and their relevance to the model's learning objectives.

\paragraph{Enhancing Model Transparency and Trust} Semantic interpretation of feature maps, facilitated by LLMs, is anticipated to improve CNN transparency and trustworthiness. By providing interpretable insights into the model's internal workings, this approach aims to make CNNs more accessible and understandable. However, the practical implementation must ensure that interpretations are both accurate and helpful for users, without oversimplifying the model's complexity.

\subsubsection{LLM-Enhanced Model Debugging}

\paragraph{Elevating Debugging Processes with Generative AI Insights} We propose integrating LLM capabilities into CNN debugging processes to provide deeper, contextual insights into model performance issues. This approach aims to enhance the efficiency and effectiveness of debugging by leveraging LLMs to identify and suggest solutions to complex problems. While promising for improving model reliability, this strategy requires a sophisticated integration of LLM insights to effectively identify and address issues without introducing new complications.

\paragraph{Operationalizing Deep Learning Model Debugging} Implementing LLM-enhanced debugging involves using LLMs to analyze model outputs and suggest adjustments. This approach promises to streamline the debugging process, making it more efficient and effective. However, its success will depend on the accuracy of LLM analyses and the feasibility of implementing suggested adjustments.

\paragraph{Advancing CNN Development through Intelligent Debugging} Integrating LLM insights into CNN debugging is expected to transform model development, making it more informed and efficient. This approach aims to leverage the analytical capabilities of LLMs to identify and resolve issues proactively. However, the challenge lies in effectively integrating LLM insights into the debugging process, ensuring that recommendations are actionable and lead to tangible improvements.

\subsubsection{Predictive Performance Optimization}

\paragraph{Proactive Enhancement of Model Efficacy} We suggest using LLMs to proactively identify and address potential performance bottlenecks in CNNs, optimizing models before issues arise. This approach aims to enhance model efficiency and effectiveness by leveraging predictive insights from LLMs. While promising for improving model performance, the practical implementation of this strategy requires careful analysis and timely adjustments to preemptively optimize the model.

\paragraph{Operationalizing Predictive Adjustments} Implementing predictive performance optimization involves continuous LLM analysis of the CNN's training progress, identifying potential issues, and suggesting preemptive adjustments. This approach promises to improve model performance by proactively addressing challenges. However, its effectiveness will depend on the accuracy of LLM predictions and the ability to implement adjustments efficiently.

\paragraph{Elevating CNN Training Through Forward-Looking Insights} Predictive performance optimization, informed by LLM analysis, is anticipated to enhance the training and development of CNNs by providing a proactive framework for model optimization. This approach aims to improve model performance and adaptability by leveraging LLM insights. However, the practical benefits of this strategy will depend on the effectiveness of predictive adjustments and their impact on model performance.

	\subsubsection{Pseudocode for Algogenic CNNs}
	The Algogenic convolutional neural network (CNN) approach harnesses AI to enhance traditional CNN methods by dynamically adjusting network parameters and strategies based on observed system behavior and real-time error estimates. This pseudocode, available in \ref{fig:CNN-Algogen-pseudocode}, outlines an advanced framework incorporating AI-driven enhancements for adaptive filter tuning, feature selection, activation function optimization, and real-time parameter adjustment.
	
	\begin{algorithm}
		\caption{Algogenic CNN Framework Pseudocode}
		\begin{algorithmic}[1]
			\Procedure{AlgogenicCNN}{Dataset}
			\State \textbf{Preprocessing:}
			\State Semantic Data Augmentation using LLM insights
			
			\State \textbf{Core Training:}
			\State Initialize CNN with semantically informed architecture
			\While{not Converged}
			\State Perform forward propagation with dynamic filters
			\State Apply Adaptive Activation Function Selection with LLM insights
			\State Conduct backpropagation to update weights
			\State Contextual Regularization Adjustment with LLM insights
			\State Predictive Performance Optimization for architecture tuning with LLM insights
			\State \textit{Check for convergence}
			\EndWhile
			
			\State \textbf{Postprocessing:}
			\State Semantic Interpretation of Feature Maps with LLM
			\State LLM-Enhanced Model Debugging
			\If{Adjustments Required}
			\State Return to Core Training with adjustments
			\Else
			\State Finalize Model
			\EndIf
			\EndProcedure
		\end{algorithmic}\label{fig:CNN-Algogen-pseudocode}
	\end{algorithm}
	
	\begin{figure}
		\centering
		\includegraphics[width=1.0\textwidth]{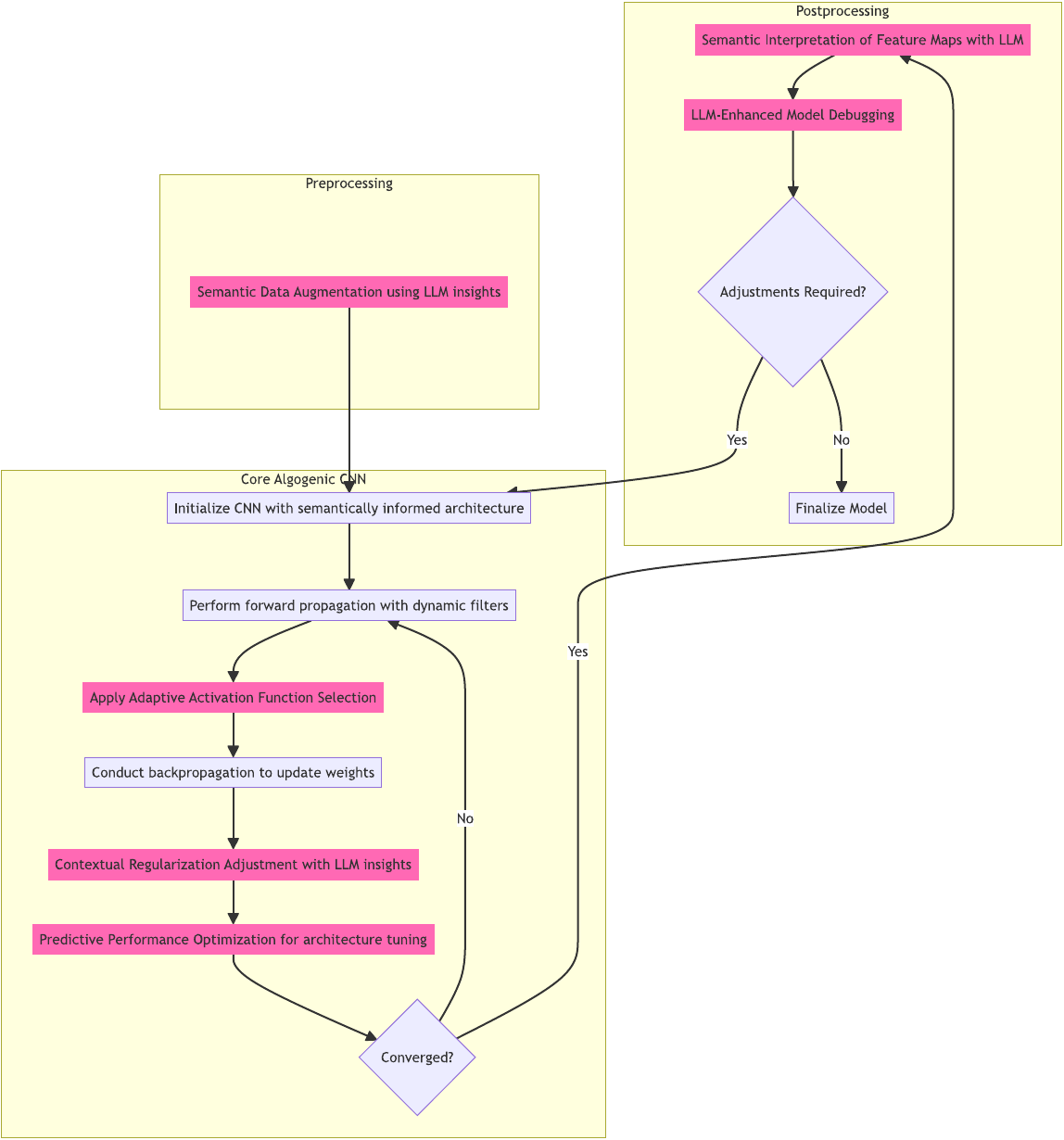} 
		\caption{Integrating Algogenic Enhancements in CNNs: This diagram visualizes the sophisticated integration of Algogenic enhancements with Convolutional Neural Networks (CNNs), leveraging the insights and capabilities of Large Language Models. It outlines a comprehensive framework that spans preprocessing with semantic data augmentation, core training adjustments including dynamic filter optimization and adaptive activation function selection, and post-processing with semantic interpretation of feature maps and LLM-enhanced model debugging. The diagram also illustrates the iterative refinement loop enabled by model debugging insights, showcasing how CNNs can evolve in response to the analysis of their performance and the data they process. This holistic approach underscores the transformative potential of combining generative AI with CNNs to enhance model adaptability, performance, and interpretability, setting a new standard for the development of visual recognition systems.}
		\label{fig:cnn}
	\end{figure}

	\section{Transformer}\index{Transformer}
	\subsection{Introduction to Transformers}
	\subsubsection{The Concept of Transformers}
	
	\paragraph{Introduction to the Transformative Model}
	The concept of Transformers revolutionized the field of deep learning by introducing a model architecture exclusively based on attention mechanisms, devoid of the recurrent layers traditionally used in sequence-to-sequence processing. Introduced by Vaswani et al. in the landmark paper "Attention is All You Need," Transformers have set new standards for a wide array of natural language processing (NLP) tasks, including but not limited to translation, text summarization, and sentiment analysis. The core idea behind Transformers is to process input data, typically textual, in parallel rather than sequentially, enabling significantly more efficient training and the ability to capture long-range dependencies in data.
	Furthermore, Transformers have shown remarkable adaptability across various domains, from language modeling to image captioning. Moreover, they have demonstrated exceptional performance even with limited labeled data, owing to their ability to learn intricate patterns from large-scale unlabeled corpora. Additionally, Transformers have facilitated transfer learning, allowing pre-trained models to be fine-tuned for specific downstream tasks with minimal supervision, thus reducing the need for extensive task-specific labeled data. Consequently, this versatility has led to widespread adoption of Transformer-based architectures in both academia and industry, propelling advancements in NLP and beyond.

	\paragraph{Architectural Foundations}
	The Transformer architecture revolutionized natural language processing by introducing the self-attention mechanism, a groundbreaking concept that underpins its unparalleled performance. This mechanism, reminiscent of the human cognitive process, empowers each position within the input sequence to dynamically weigh its relevance to every other position, fostering a holistic comprehension of the data. Through the orchestration of multiple attention heads, the Transformer extracts a multifaceted representation of the input, synthesizing diverse contextual perspectives. These attention heads, acting as specialized lenses, scrutinize the input from various angles, enriching the model's understanding with nuanced insights. As a result, the Transformer transcends traditional sequential architectures, embracing a parallelized approach that efficiently captures long-range dependencies. The structural framework of the Transformer further underscores its versatility and power. Comprising an encoder and a decoder, each composed of multiple layers, it facilitates intricate transformations through a hierarchical series of operations. Within these layers, self-attention mechanisms collaborate with position-wise feed-forward networks to iteratively refine the representation, iteratively enhancing its discriminative capacity. The elegance of this design lies in its modular nature, enabling seamless scalability and adaptability to diverse tasks and datasets. Thus, the Transformer architecture not only redefines the landscape of natural language processing but also serves as a cornerstone for advancing the frontiers of artificial intelligence.

	\paragraph{Mathematical Underpinnings of Attention}
	The self-attention mechanism can be mathematically represented by the equation:
	\[
	\text{Attention}(Q, K, V) = \text{softmax}\left(\frac{QK^T}{\sqrt{d_k}}\right)V
	\]
	where $Q$, $K$, and $V$ represent the queries, keys, and values matrices, respectively, derived from the input embeddings, and $d_k$ denotes the dimensionality of the keys. This formulation allows the model to dynamically allocate attention across different positions of the input sequence, based on the computed similarity between queries and keys, thereby enabling the selective aggregation of information in the values matrix $V$.
	
	\paragraph{Innovations and Advancements}
	Transformers have introduced several innovations, including positional encoding, which imbues the model with a sense of word order, and layer normalization, which stabilizes the learning process. \textbf{Moreover}, the architecture's ability to handle inputs in parallel, its scalability, and its capacity to model complex dependencies without the constraints of recurrent processing have made Transformers the architecture of choice for a broad spectrum of NLP tasks and beyond. \textbf{Additionally}, Transformers have proven to be highly adaptable, capable of accommodating various input modalities, such as text, images, and graphs, making them versatile tools across domains. \textbf{Furthermore}, the self-attention mechanism in Transformers enables them to capture long-range dependencies efficiently, surpassing the limitations of traditional sequential models. \textbf{On the contrary}, while recurrent neural networks (RNNs) and convolutional neural networks (CNNs) have been widely used for NLP tasks in the past, they often struggle with capturing long-range dependencies and suffer from vanishing gradient problems. Thus, \textbf{in contrast}, Transformers offer a more effective solution by allowing information flow across the entire sequence simultaneously, mitigating the issues faced by RNNs and CNNs. Therefore, the adoption of Transformers represents a significant breakthrough in natural language processing, paving the way for advancements in various AI applications.

	\paragraph{Beyond Natural Language Processing}
	Despite being initially conceived for NLP, the Transformer architecture has transcended its original purpose, finding applications in diverse fields such as computer vision and audio processing. This expansion is attributed to the intrinsic adaptability of Transformers, which capitalize on self-attention mechanisms and parallel processing paradigms. While traditionally applied in linguistic contexts, the deployment of Transformers in computer vision exploits their ability to discern complex patterns across spatial data, enabling tasks like image classification and object detection to achieve remarkable accuracy. Similarly, in audio processing, Transformers excel in capturing temporal dependencies and extracting meaningful features from spectrograms or waveforms, revolutionizing tasks like speech recognition and sound synthesis. The emergence of models like GPT and BERT signifies a paradigm shift in machine learning, where pre-training on vast corpora followed by fine-tuning for specific tasks has become the norm. These models leverage the immense scale of data available on the internet, enabling them to learn rich representations that generalize well across domains. Consequently, they have become benchmarks for performance, surpassing traditional approaches in tasks ranging from language understanding to image generation. Furthermore, the modularity of Transformer-based architectures facilitates transfer learning, allowing knowledge gained from one domain to enhance performance in another. This versatility underscores the transformative impact of Transformers, propelling advancements not only in NLP but across the spectrum of AI applications.

	\subsubsection{Key Principles and Mechanisms}
	
	\paragraph{Self-Attention: The Core Mechanism}
	The foundational principle of the Transformer architecture is the self-attention mechanism, enabling each element in the input sequence to interact with every other element, regardless of their positions. This mechanism calculates the attention scores based on the similarity between elements, allowing the model to dynamically focus on different parts of the input for each output element. Self-attention provides the flexibility to capture complex dependencies and relationships within the data, including those that span long distances in the input sequence, a task that traditional recurrent neural networks (RNNs) find challenging.
	
	Furthermore, the self-attention mechanism empowers the Transformer model to process inputs efficiently by parallelizing computation. Unlike sequential models such as RNNs, where processing steps are inherently sequential, self-attention allows for simultaneous consideration of all input elements. This parallelism leads to significant speed-ups in training and inference, making Transformers highly scalable and applicable to large datasets.
	
	Moreover, self-attention promotes interpretability by assigning importance weights to each input element based on its relevance to the output. These attention scores provide insights into which parts of the input sequence contribute most to the model's predictions, aiding in model debugging and analysis. This interpretability aspect is particularly crucial in applications where understanding the model's decision-making process is essential, such as natural language processing tasks like text generation or sentiment analysis.
	
	Additionally, self-attention facilitates long-range dependencies modeling by allowing direct connections between distant elements in the input sequence. Traditional sequential models like RNNs struggle with capturing such dependencies due to vanishing or exploding gradient problems over long sequences. With self-attention, the Transformer can attend to relevant information regardless of its distance, enabling more effective learning of contextual relationships across the entire input sequence.
	
	Therefore, the self-attention mechanism stands as a pivotal innovation in the Transformer architecture, revolutionizing natural language processing and advancing the state-of-the-art in various other domains, including computer vision and speech recognition.

	\paragraph{Multi-Head Attention: Enhancing Representation}
	Transformers leverage multi-head attention mechanisms to enrich the model's ability to capture intricate patterns and relationships within the input sequence. By employing multiple attention heads in parallel, each with its distinct set of learnable parameters, the model gains the flexibility to focus on diverse aspects of the input simultaneously. This parallelization enables the transformer to attend to different positions with varied attention weights, facilitating the extraction of rich contextual information from different parts of the sequence.
	
	Moreover, the utilization of multiple attention heads allows the transformer to capture a multitude of dependencies at different levels of granularity. For instance, while some attention heads may focus on local dependencies within a short range, others may attend to long-range dependencies spanning the entire sequence. This hierarchical attention mechanism enables the model to discern both fine-grained details and broader contextual relationships within the input data.
	
	Furthermore, by concatenating the outputs of the individual attention heads and applying a linear transformation, the transformer integrates the diverse insights obtained from each head into a comprehensive representation of the input sequence. This integrated representation encapsulates a holistic understanding of the input, incorporating information from multiple perspectives and levels of abstraction.
	
	In essence, multi-head attention empowers transformers to capture complex dependencies and encode rich contextual information in an efficient and scalable manner. By leveraging parallelized attention mechanisms, transformers enhance their representational capacity, enabling them to effectively model the intricate structure of natural language and other sequential data.

	\paragraph{Positional Encoding: Incorporating Sequence Order}
	Given the Transformer's reliance on self-attention, which inherently lacks a mechanism to recognize sequence order, positional encoding is introduced to retain positional information. This encoding adds a unique vector to each input token's embedding, signifying its position in the sequence. \textbf{Moreover}, positional encoding serves as a crucial element in enhancing the Transformer's ability to understand and process sequential data effectively. \textbf{Additionally}, the choice between sinusoidal functions or learned embeddings for implementing positional encoding offers flexibility and allows for adaptation to different tasks and datasets. \textbf{Furthermore}, by incorporating positional encoding, the Transformer model can capture not only the content-based relationships between tokens through self-attention but also their sequential dependencies, \textbf{thus} enabling a more comprehensive understanding of the input sequence structure. \textbf{On the other hand}, neglecting positional encoding could lead to the model overlooking important sequential patterns, potentially resulting in suboptimal performance, especially in tasks where sequence order is essential for accurate predictions. \textbf{In contrast}, when positional encoding is effectively utilized, it facilitates the Transformer in distinguishing between tokens with similar content but different positions, \textbf{thereby} improving its ability to generate coherent and contextually relevant outputs. \textbf{Consequently}, the incorporation of positional encoding aligns with the overarching goal of the Transformer architecture, which aims to leverage the power of self-attention while preserving the sequential structure inherent in many real-world datasets.

	\paragraph{Layered Architecture: Encoder and Decoder}
	The Transformer model, with its distinctive layered architecture comprising an encoder and a decoder, epitomizes a paradigm shift in sequence processing tasks. This architectural design, characterized by multiple identical layers within each component, facilitates intricate information processing across sequential data. Within the encoder, the input sequence undergoes a transformative journey, where each layer meticulously refines its representation through self-attention mechanisms and position-wise feed-forward networks. This sequential refinement not only captures intricate dependencies within the input but also ensures robustness in handling diverse data modalities. Conversely, the decoder module, tailored for sequence generation endeavors, embarks on a nuanced process. While attending to the encoder's output, the decoder leverages self-attention mechanisms to reconcile its input, orchestrating a coherent flow for generating target sequences. This bifurcated structure empowers the Transformer with unparalleled versatility, enabling it to excel in an array of tasks spanning from encoding endeavors, such as sentence embedding generation, to decoding challenges like language translation. This duality of function, underscored by the distinct roles of the encoder and decoder, underscores the transformative potential of the Transformer architecture, heralding a new era in sequence processing methodologies.

	\paragraph{Optimization and Training Techniques}
	Training Transformers involves optimizing a vast number of parameters, facilitated by techniques such as layer normalization and dropout, which help stabilize and regularize the learning process. Furthermore, the use of scaled dot-product attention in computing attention scores contributes to the model's efficiency by enabling a softmax operation over scaled scores, reducing the impact of large values and improving gradient flow. Moreover, the Transformer architecture benefits significantly from attention-based optimization strategies and advanced training algorithms, like the Adam optimizer, which together enhance the model's learning dynamics and overall performance. Additionally, techniques like layer normalization and dropout not only stabilize and regularize the learning process but also mitigate overfitting by introducing noise during training. Likewise, the use of the Adam optimizer facilitates faster convergence by adapting learning rates for each parameter individually based on the past gradients and exponentially decaying averages of past squared gradients. Moreover, scaled dot-product attention allows the model to attend to relevant parts of the input sequence efficiently, improving its ability to capture long-range dependencies. Furthermore, the incorporation of advanced training algorithms enhances the model's ability to generalize to unseen data and handle complex tasks. Thus, the combination of these optimization and training techniques plays a crucial role in the success of Transformer models in various natural language processing tasks.

	\subsubsection{The Role of Attention Mechanisms}
	
	\paragraph{Foundation of Transformers' Functionality}
	The attention mechanism serves as the cornerstone of the Transformer architecture, fundamentally altering the approach to sequence modeling in deep learning. Unlike previous models that processed data sequentially, Transformers leverage attention to weigh the importance of different input elements relative to each other for a given task. This mechanism allows the model to focus on relevant parts of the input data. Additionally, it facilitates capturing long-distance dependencies and nuanced relationships within the data more effectively than traditional recurrent or convolutional models. 
	
	Moreover, by employing attention, Transformers can handle variable-length sequences with ease, a significant advantage over fixed-length approaches. This flexibility is particularly crucial in natural language processing tasks where input lengths can vary drastically. Furthermore, the attention mechanism enhances parallelism in computation, enabling faster training times compared to sequential models.
	
	On the other hand, despite their numerous advantages, Transformers can be computationally intensive, especially when dealing with large datasets. Nonetheless, advancements in hardware and optimization techniques have mitigated this issue to some extent, making Transformers increasingly practical for real-world applications. 
	
	Consequently, the Transformer architecture has become the go-to choice for various sequence-based tasks, including machine translation, text generation, and language understanding, demonstrating its versatility and effectiveness in the field of deep learning.

	\paragraph{Mechanics of Attention}
	At its core, the attention mechanism operates by computing a set of attention scores, which serve to gauge the significance or contribution of each input element towards generating the output. This pivotal process involves the fundamental step of calculating the dot product between the query vector and the key vectors associated with every element in the sequence. Subsequently, these dot products are subjected to a softmax operation, thereby normalizing the scores to attain a probability distribution. These attention weights, thus obtained, play a crucial role in generating a weighted sum of the corresponding value vectors. By doing so, the mechanism effectively synthesizes an output that strategically amalgamates information gleaned from various parts of the input sequence. This deliberate and selective focus on pertinent elements is precisely what empowers Transformers to adeptly process and comprehend intricate patterns and dependencies present within the data. Therefore, the attention mechanism emerges as a cornerstone of Transformer architectures, facilitating their remarkable ability to capture long-range dependencies and contextual information across sequences. Furthermore, this mechanism not only enhances the model's capacity for information integration but also endows it with the capability to dynamically adapt its focus based on the input, thus facilitating robust and contextually sensitive processing.
	
	\paragraph{Scaling with Multi-Head Attention}
	Transformers revolutionize natural language processing (NLP) tasks through the innovative integration of multi-head attention mechanisms. Unlike traditional attention mechanisms, which rely on a single set of learned query, key, and value vectors, multi-head attention splits the attention process into several heads, each equipped with its own set of parameters. This subdivision enables the model to simultaneously process and fuse information from multiple perspectives, akin to having multiple experts analyzing different aspects of the input data. Consequently, the Transformer architecture excels at capturing intricate relationships within the input sequences, leading to enhanced performance across various NLP tasks.
	
	The multi-head attention mechanism empowers Transformers to perform intricate computations in parallel, thereby significantly improving computational efficiency. Rather than relying on a single attention mechanism to process the entire input sequence, the model delegates this task to multiple heads, each focusing on a different aspect of the data. This parallelization not only accelerates the training and inference processes but also facilitates the model's scalability to longer sequences, a crucial advantage in handling real-world applications where inputs vary in length and complexity.
	
	Furthermore, the multi-head attention mechanism enables Transformers to learn diverse representations of the input data. By attending to different subsets of the input features simultaneously, the model can capture nuanced patterns and dependencies that might be overlooked by a single attention mechanism. This comprehensive analysis of the input data fosters a deeper understanding of the underlying semantics, allowing the model to make more informed predictions and produce more coherent outputs.
	
	In summary, multi-head attention serves as a cornerstone of Transformer architecture, enabling the model to scale effectively and perform complex computations with remarkable efficiency. By leveraging multiple heads to analyze input sequences from various perspectives, Transformers achieve superior performance across a wide range of NLP tasks, underscoring the transformative impact of multi-head attention on modern machine learning paradigms.

	\paragraph{Impact on Sequential Data Processing}
	The attention mechanisms in Transformers revolutionize the processing of sequential data by making every element of the sequence directly accessible to every other element, eliminating the limitations associated with processing data in strict sequential order. This approach not only increases the efficiency of the model, allowing for parallel processing of sequence elements, but also enhances the model's ability to understand and generate natural language, handle time-series data, and perform on tasks requiring an intricate understanding of sequential relationships.
	
	Moreover, by enabling parallel processing, Transformers significantly reduce the computational burden traditionally associated with sequential data analysis. This reduction in computational complexity allows for faster training and inference times, making Transformers particularly advantageous for real-time applications where speed is crucial.
	
	Additionally, the bidirectional nature of attention mechanisms in Transformers enables the model to capture long-range dependencies within sequences more effectively. Unlike traditional sequential models that process data in a unidirectional manner, Transformers can incorporate information from both past and future elements of a sequence simultaneously, leading to richer representations and more accurate predictions.
	
	Furthermore, the self-attention mechanism in Transformers allows the model to assign varying degrees of importance to different parts of the input sequence dynamically. This adaptive attention mechanism enables Transformers to focus on relevant elements while ignoring irrelevant ones, enhancing the model's robustness and interpretability.
	
	Consequently, the adoption of Transformers has led to significant advancements in various fields, including natural language processing, speech recognition, machine translation, and time-series forecasting. The versatility and effectiveness of Transformers in handling sequential data have made them a cornerstone of modern deep learning architectures, paving the way for further innovations in sequential data processing.

	\paragraph{Broadening the Scope of Applications}
	The versatility and effectiveness of attention mechanisms have broadened the scope of applications for Transformers. \textbf{Moreover}, the extension of their use \textbf{beyond} natural language processing to fields such as computer vision and audio signal processing reflects their adaptability and robustness. \textbf{Furthermore}, the incorporation of attention mechanisms enables Transformers to delve into the intricacies of complex, multimodal datasets. This capacity to dynamically focus on different aspects of the input data is pivotal \textbf{since} it facilitates detailed analysis and interpretation, making Transformers particularly well-suited for tasks requiring comprehensive understanding. \textbf{Consequently}, as researchers delve deeper into attention-based models, the role of attention mechanisms in deep learning is poised to expand. This expansion holds \textbf{promise for} further innovations and advancements in artificial intelligence, ushering in a new era of sophisticated applications and capabilities across various domains.

	\subsubsection{Applications and Limitations}
	
	\paragraph{Versatile Applications Across Domains}
	Transformers have found widespread applications across numerous domains, primarily revolutionizing the field of natural language processing (NLP). They are the backbone of models like BERT (Bidirectional Encoder Representations from Transformers) for understanding contextual relationships in text. Additionally, GPT (Generative Pre-trained Transformer) leverages Transformers for generating coherent and diverse text, enabling tasks such as text completion, translation, and summarization with unprecedented fluency and accuracy. Moreover, T5 (Text-to-Text Transfer Transformer) represents a significant advancement by framing NLP tasks as a unified text-to-text problem, allowing for seamless adaptation and transfer learning across various tasks.
	
	Beyond NLP, Transformers have been adapted for use in computer vision with models like Vision Transformer (ViT). ViT's innovative approach treats image patches as sequence elements, allowing it to achieve state-of-the-art results on image classification tasks. This paradigm shift in computer vision demonstrates the adaptability and versatility of Transformer architectures across different modalities.
	
	In the realm of audio processing, Transformers facilitate tasks such as speech recognition and music generation by effectively modeling temporal dependencies. Their attention mechanisms enable capturing long-range dependencies in audio signals, leading to enhanced performance in tasks requiring understanding of sequential audio data.
	
	Furthermore, the inherent ability of Transformers to handle sequential data makes them suitable for time-series forecasting. By capturing patterns and dependencies in historical data, Transformers can predict future values with remarkable accuracy, offering valuable insights for decision-making in various domains such as finance, weather forecasting, and resource management.

	\paragraph{Limitations and Challenges}
	Despite their versatility, Transformers are not without limitations. One of the most significant challenges is their computational and memory intensity, especially for large-scale models and datasets, necessitating substantial hardware resources for training and inference. This computational demand limits their accessibility and can hinder rapid experimentation and deployment. However, despite these challenges, the widespread adoption of Transformers underscores their remarkable capabilities in various domains. Another challenge is the potential for overfitting, particularly in smaller datasets, due to the model's large number of parameters. Transformers also struggle with efficiently processing very long sequences due to quadratic complexity in the self-attention mechanism, posing challenges for tasks requiring the analysis of extensive context. This issue may be mitigated through techniques such as hierarchical or sparse attention mechanisms, but these approaches introduce additional complexity and computational overhead. Additionally, while Transformers offer improved performance on a variety of tasks, their "black box" nature, common to many deep learning models, can obscure the understanding of how decisions are made, complicating efforts to diagnose errors or biases in the model. Nonetheless, ongoing research efforts aim to enhance interpretability and transparency in Transformer models, with approaches such as attention visualization and attribution methods. Despite these challenges, Transformers continue to revolutionize natural language processing and other domains, driving advancements in AI research and applications.

	\paragraph{Overcoming Limitations}
	Efforts to overcome these limitations include the development of more efficient Transformer architectures, such as Linformer, Performer, and Reformer, which aim to reduce the computational complexity of attention mechanisms. These architectures introduce innovative strategies to address the challenges posed by long sequences, enabling the application of Transformers in diverse real-world scenarios.
	
	Techniques like knowledge distillation are employed to compress large models into smaller, more manageable versions without significant loss of performance, enhancing their usability on resource-constrained devices. By distilling the knowledge learned by a complex model into a simpler one, researchers can retain essential information while reducing computational demands, facilitating deployment in settings with limited computational resources.
	
	Researchers are also exploring methods to improve the interpretability of Transformers, recognizing the importance of understanding model decisions for trust and usability. Attention visualization techniques provide insights into which parts of the input receive more focus during processing, shedding light on the model's decision-making process. Probing tasks delve deeper into the internal representations of the model, uncovering patterns and biases that may influence its behavior.
	
	Furthermore, advancements in training methodologies, such as adaptive and sparse attention mechanisms, offer promising avenues for mitigating issues related to sequence length and computational efficiency. These approaches dynamically adjust the attention mechanism based on the input, allocating computational resources more efficiently and effectively. Sparse attention mechanisms focus on processing only relevant parts of the input, reducing redundant computations and improving scalability for longer sequences.

	\paragraph{Future Directions}
	As the field continues to evolve, the applications of Transformers are expanding into hybrid models that combine the strengths of Transformers with other architectures, and into novel domains beyond traditional NLP and computer vision tasks. Ongoing research is focused not only on addressing the current limitations but also on harnessing the potential of Transformers to unlock new capabilities in artificial intelligence, signaling a future where they play a central role in advancing machine learning and its applications across a broad spectrum of industries and tasks.
	
	Moreover, the integration of Transformers with other architectures such as recurrent neural networks (RNNs) and convolutional neural networks (CNNs) opens up possibilities for even more powerful models. By leveraging the sequential processing capabilities of RNNs and the spatial hierarchies learned by CNNs, hybrid models can tackle complex tasks with improved efficiency and effectiveness. This integration paves the way for advancements in various domains, including natural language understanding, image recognition, and beyond.
	
	Furthermore, as the demand for AI solutions grows across industries, there is a pressing need to deploy models that can adapt to diverse data modalities and domains. Transformers, with their inherent flexibility and scalability, offer a promising avenue for addressing these challenges. Whether deployed in healthcare for medical diagnosis, in finance for predictive analytics, or in robotics for autonomous decision-making, the versatility of Transformers enables their widespread adoption in real-world applications.
	
	Additionally, ongoing efforts in research and development are exploring ways to enhance the interpretability and explainability of Transformer models. As these models become increasingly complex, understanding their decision-making processes becomes crucial for building trust and facilitating their deployment in sensitive domains. Techniques such as attention visualization and saliency mapping are being actively investigated to shed light on how Transformers process information, enabling stakeholders to interpret model predictions and ensure alignment with ethical and regulatory standards.
	
	In summary, the future of Transformers lies not only in overcoming current limitations but also in pushing the boundaries of AI innovation across diverse domains and applications. Through continuous research, integration with complementary architectures, and efforts to enhance interpretability, Transformers are poised to drive transformative advancements in machine learning, shaping the future landscape of artificial intelligence.

	\subsubsection{Algorithmic Pseudocode for Transformers}
	The Transformer Model represents a sophisticated framework tailored for effective parameter estimation in sequence-to-sequence tasks, particularly in natural language processing and other sequential data domains. It distinguishes itself by its unique architecture, which employs multiple layers of self-attention and feed-forward networks. This architecture enables the model to encode input sequences while incorporating positional information, and subsequently decode them to generate predictions. The operational essence of the Transformer is encapsulated in pseudocode \ref{fig:transformer-pseudocode}, which illustrates its iterative approach to sequence-to-sequence tasks. This concise pseudocode encapsulates the core functionality of a Transformer model, emphasizing the essential processes of positional encoding, encoding the input sequence, and decoding to generate predictions. The encoding phase applies multiple layers of self-attention and feed-forward networks to the input, enriched with positional information. The decoding phase iteratively generates the output sequence, leveraging both self-attention mechanisms to maintain coherence and attention over the encoder's output to integrate context. The process iterates until the model predicts an end-of-sequence token, producing the final sequence of predictions.
	
	\begin{algorithm}
		\caption{Pseudocode for a Transformer Model}
		\begin{algorithmic}[1]
			\Procedure{Transformer}{InputSequence}
			\State $PositionEncodedInput \gets$ AddPositionalEncoding(InputSequence)
			\State $EncoderOutput \gets$ Encode($PositionEncodedInput$)
			\State $Predictions \gets$ Decode($EncoderOutput$)
			\State \Return $Predictions$
			\EndProcedure
			
			\Function{Encode}{Input}
			\For{each encoder layer}
			\State Apply multi-head self-attention on Input
			\State Apply position-wise feed-forward network
			\EndFor
			\State \Return Output of last encoder layer
			\EndFunction
			
			\Function{Decode}{EncoderOutput}
			\State Initialize output sequence with [START] token
			\While{not [END] token predicted}
			\State Apply masked multi-head self-attention on output sequence
			\State Apply multi-head attention over EncoderOutput
			\State Predict next token using linear layer and softmax
			\State Append predicted token to output sequence
			\EndWhile
			\State \Return output sequence without [START] token
			\EndFunction
		\end{algorithmic}\label{fig:transformer-pseudocode}
	\end{algorithm}

	\subsection{Algogenic Enhancements for Transformers}
	\subsubsection{Semantic Embedding Initialization}
	
	\paragraph{Enhancing Initial Representations}
	We suggest enhancing the initial embeddings of Transformer models through the incorporation of semantic insights derived from Large Language Models. This process, termed Semantic Embedding Initialization, aims to provide the Transformer with an advanced starting point for understanding linguistic nuances. By embedding a rich semantic understanding into the initial embeddings, the model may better grasp complex semantic relationships early in its training phase. This approach could potentially reduce the necessity for extensive domain-specific fine-tuning, as the model begins with a more nuanced understanding of language. However, it's important to acknowledge that while promising, the effectiveness of this enhancement depends on the quality and breadth of the semantic analysis performed by the LLMs.
	
	\paragraph{Operationalizing Semantic Insights}
	To operationalize this enhancement, LLMs are tasked with analyzing extensive textual data to distill semantic relationships and nuances. These insights are then encoded into the initial embeddings of the Transformer model. Such an approach involves careful calibration of embeddings to ensure they accurately reflect semantic nuances such as synonymy and antonymy. Despite the potential benefits, the practicality of implementing Semantic Embedding Initialization hinges on the ability to effectively capture and encode a comprehensive semantic understanding, which remains a significant challenge.
	
	\paragraph{Implications for Model Performance}
	Initiating Transformer models with semantically rich embeddings is theorized to enhance their performance, especially in tasks requiring nuanced language understanding. This could lead to more efficient training processes and improved model generalization. However, the actual impact on model performance will vary based on the depth of semantic analysis and the relevance of pre-encoded knowledge to specific tasks. The integration of this enhancement underscores the potential for a more nuanced understanding of language within Transformer models, although its practical effectiveness needs careful evaluation.
	
	\subsubsection{Dynamic Attention Mechanism Adjustment}
	
	\paragraph{Optimizing Attention for Enhanced Focus}
	We propose dynamically adjusting the attention mechanism of Transformer models based on insights from LLMs, allowing for a more focused analysis of semantically significant parts of the input. This Dynamic Attention Mechanism Adjustment could potentially improve the model's ability to prioritize crucial information. However, implementing such a dynamic system requires a nuanced approach to ensure that adjustments enhance rather than detract from the model's ability to understand and generate text.
	
	\paragraph{Implementing Adaptive Attention Strategies}
	Implementing this enhancement involves utilizing LLMs to analyze input data in real-time, identifying key elements that should receive more attention. Adjustments to the attention mechanism are made based on this analysis. The feasibility and success of this approach depend on the model's capacity to accurately identify and adjust to the most relevant aspects of the data, which may introduce additional computational complexity and require sophisticated optimization strategies.
	
	\paragraph{Implications for Transformer Model Efficacy}
	Adapting the attention mechanism dynamically is expected to improve the Transformer model's efficacy by enabling it to respond more adeptly to the nuances of the input data. This could enhance performance across a range of tasks, though the extent of improvement will likely depend on the precision of the dynamic adjustments and the model's initial sensitivity to semantic importance.
	
	\subsubsection{Adaptive Positional Encoding}
	
	\paragraph{Refining Temporal Context Understanding}
	Enhancing Transformer models with Adaptive Positional Encoding aims to allow for a more nuanced understanding of the position and order within input sequences, based on contextual importance derived from LLM analyses. This could improve the model's handling of temporal and sequential data by adjusting its sensitivity to positional information. The practical application of this enhancement involves sophisticated mechanisms to dynamically adjust positional encodings, which may present challenges in ensuring the model's adaptability without compromising its understanding of sequence order.
	
	\paragraph{Implementing Context-Aware Positional Signals}
	To implement Adaptive Positional Encoding, insights from LLMs regarding the contextual significance of sequence positions are used to adjust the model's positional encodings. This requires a deep semantic analysis of the input data and a flexible encoding scheme capable of reflecting the varying importance of positions within different contexts. The complexity of dynamically adjusting positional encoding based on context underscores the technical challenges in enhancing model flexibility while maintaining accurate sequence understanding.
	
	\paragraph{Enhancing Model Flexibility and Semantic Precision}
	The introduction of Adaptive Positional Encoding is anticipated to enhance the flexibility and semantic precision of Transformer models by enabling them to adjust their sensitivity to positional information. This could lead to improved performance in tasks requiring a nuanced understanding of sequence order. However, the practical benefits of this approach will depend on the model's ability to effectively integrate dynamic positional adjustments without compromising the overall coherence and accuracy of its outputs.
	
	\subsubsection{Contextual Layer Weighting}
	
	\paragraph{Tailoring Layer Contributions for Enhanced Understanding}
	We suggest optimizing the contribution of different layers within Transformer models based on the context of the input, leveraging insights from LLMs. This Contextual Layer Weighting aims to enhance the model's processing by dynamically adjusting the influence of each layer's output. While promising, the implementation of this enhancement must be approached with caution to ensure that adjustments improve, rather than inadvertently impact, the model's ability to process and understand input data effectively.
	
	\paragraph{Operationalizing Dynamic Layer Integration}
	The operationalization of Contextual Layer Weighting involves analyzing the input and its context to determine the optimal layer contributions. This dynamic adjustment process requires a sophisticated understanding of the model's architecture and the specific functions of its layers, presenting challenges in accurately modulating layer weights without disrupting the model's overall performance.
	
	\paragraph{Elevating Model Performance Through Intelligent Processing}
	Contextual Layer Weighting is expected to elevate the Transformer model's performance by allowing for a more intelligent processing strategy that adapts to the input context. However, the effectiveness of this approach will depend on the precision of the layer adjustments and their alignment with the model's learning objectives.
	
	\subsubsection{Semantic Output Analysis}
	
	\paragraph{Deepening Output Comprehension with Generative Insights}
	We propose employing LLMs to conduct a Semantic Output Analysis of Transformer-generated text, aiming to ensure semantic coherence and contextual appropriateness. This process involves a detailed examination of the outputs to identify and address any discrepancies or areas for enhancement. While this approach holds promise for improving output quality, it also introduces challenges in terms of computational overhead and the complexity of accurately assessing and refining semantic content.
	
	\paragraph{Implementing Output Evaluation and Enhancement}
	To implement Semantic Output Analysis, a detailed post-generation evaluation of the Transformer's outputs is conducted using LLMs. This involves identifying areas where adjustments are needed to improve semantic coherence and accuracy. The practical application of this enhancement will require sophisticated evaluation mechanisms and could significantly impact the model's efficiency and throughput.
	
	\paragraph{Advancing Transformers Toward Semantic Precision}
	Semantic Output Analysis aims to advance Transformer models toward greater semantic precision in their outputs. The success of this enhancement will depend on the depth and accuracy of the semantic evaluation process, highlighting the challenges in achieving a balance between output quality and computational efficiency.
	
	\subsubsection{Enhanced Explanation Generation}
	
	\paragraph{Clarifying Model Decisions through Rich Narratives}
	Enhanced Explanation Generation involves leveraging LLMs to generate detailed explanations for the decisions and outputs of Transformer models. This aims to improve transparency and user understanding of the model's processes. Implementing this enhancement requires careful consideration of how best to generate explanations that are both informative and accessible to users, without overwhelming them with technical details.
	
	\paragraph{Operationalizing Insightful Explanations}
	To operationalize Enhanced Explanation Generation, LLMs are used to analyze the model's outputs and the factors influencing its decisions, generating accessible explanations. The effectiveness of this approach will depend on the LLMs' ability to accurately interpret and convey the model's decision-making processes in a way that enhances user understanding and trust.
	
	\paragraph{Empowering Users with Deep Model Insights}
	By providing enhanced explanations for model decisions, this approach seeks to empower users with deeper insights into the workings of Transformer models. The potential benefits of this enhancement include increased transparency and trust in AI systems. However, achieving these benefits in practice will require a nuanced approach to explanation generation that balances detail with accessibility.
	
	\subsubsection{Predictive Performance Enhancement}
	
	\paragraph{Proactive Optimization for Future Challenges}
	Predictive Performance Enhancement focuses on using LLMs to anticipate and address future performance challenges of Transformer models. This forward-looking approach aims to ensure the model's continued efficacy and adaptability. Implementing this enhancement involves continuous analysis and adjustment, presenting challenges in accurately predicting future challenges and effectively implementing preemptive optimizations.
	
	\paragraph{Implementing Anticipatory Model Adjustments}
	The implementation of Predictive Performance Enhancement involves continuous monitoring and analysis to identify and preemptively address potential performance issues. The success of this approach will depend on the accuracy of the predictions and the effectiveness of the implemented adjustments in maintaining or enhancing model performance.
	
	\paragraph{Elevating Transformer Capabilities Through Strategic Foresight}
	Predictive Performance Enhancement aims to elevate the capabilities of Transformer models through strategic foresight and proactive optimization. While the concept is promising, its practical implementation raises challenges in terms of predictive accuracy and the potential for unintended consequences from preemptive adjustments.
	
	\subsubsection{Challenges and Opportunities in Algogenic Transformers}
	
	\paragraph{Navigating the Complexities of Integration}
	Integrating Algogenic enhancements into Transformer models presents both challenges and opportunities. Technical complexities, computational resource considerations, and the potential exacerbation of biases are among the key challenges. Addressing these challenges requires a multifaceted approach that balances innovation with practicality and ethical considerations.
	
	\paragraph{Exploiting the Synergy for Advanced Model Performance}
	The integration of Algogenic enhancements offers significant opportunities for advancing Transformer model performance. However, realizing these benefits in practice involves overcoming the associated technical and computational challenges. The synergy between LLMs and Transformers holds the potential for transformative improvements in natural language processing and beyond, provided that the complexities of integration can be effectively managed.
	
	\paragraph{Shaping the Future of AI with Algogenic Transformers}
	The exploration of Algogenic Transformers represents an exciting direction for the future of AI, promising more intelligent, adaptable, and effective systems. The successful integration of these enhancements requires collaboration across disciplines to address the technical, computational, and ethical challenges involved. By navigating these challenges, the AI community can unlock new possibilities for AI systems that better understand and interact with the world.

	\subsubsection{Pseudocode for Algogenic Transformers}
	The Algogenic Transformer approach harnesses generative AI advancements to enhance traditional Transformer methods by dynamically adjusting its parameters and strategies based on the observed behavior of the system and real-time error estimates. This pseudocode, available in \ref{fig:transformer-Algogen-pseudocode}, outlines an advanced framework incorporating AI-driven enhancements for adaptive attention mechanisms, layer-wise optimization, input encoding, and real-time parameter tuning.
	
	\begin{algorithm}
		\caption{Algogenic Transformer Framework Pseudocode}
		\begin{algorithmic}[1]
			\Procedure{AlgogenicTransformer}{InputData}
			\State \textbf{Preprocessing:}
			\State Semantic Embedding Initialization using LLM insights
			
			\State \textbf{Core Processing:}
			\State Initialize Transformer with enhanced embeddings
			\For{each input sequence in InputData}
			\State Apply Dynamic Attention Mechanism Adjustment with LLM insights
			\State Use Adaptive Positional Encoding with LLM insights
			\State Perform forward and backward pass with Contextual Layer Weighting with LLM insights
			\State \textit{Check for convergence or end of epoch}
			\If{not Converged and not end of epoch}
			\State Continue to next epoch or input sequence
			\Else
			\State Proceed to Postprocessing
			\EndIf
			\EndFor
			
			\State \textbf{Postprocessing:}
			\State Semantic Output Analysis with LLM
			\State Enhanced Explanation Generation using LLM
			\State Predictive Performance Enhancement based on LLM insights
			\EndProcedure
		\end{algorithmic}\label{fig:transformer-Algogen-pseudocode}
	\end{algorithm}

	\begin{figure}
		\centering
		\includegraphics[width=0.73\textwidth]{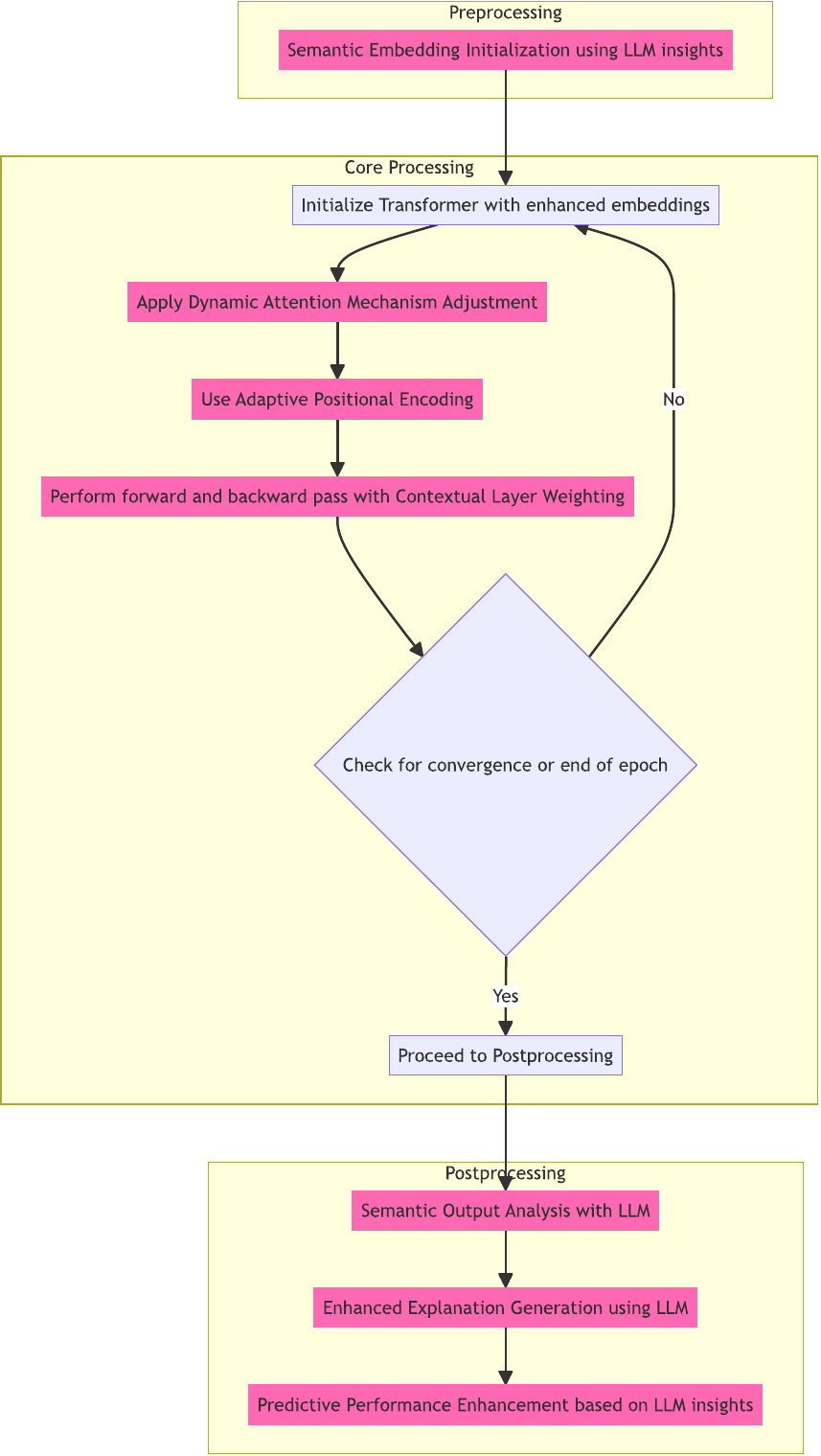} 
		\caption{Integrating Algogenic Enhancements in Transformers: This diagram visualizes the groundbreaking integration of Algogenic enhancements with Transformer models, leveraging the insights and analytical prowess of Large Language Models. It outlines a comprehensive framework that spans preprocessing with semantic embedding initialization, core processing adjustments including dynamic attention mechanism adjustment, adaptive positional encoding, and contextual layer weighting, all informed by LLM insights. The post-processing phase includes semantic output analysis, enhanced explanation generation, and predictive performance enhancement, illustrating how Transformers can evolve in response to complex linguistic tasks. This holistic integration underscores the transformative potential of combining generative AI with Transformer models to enhance natural language processing capabilities, setting a new benchmark for AI-driven linguistic analysis and generation.}
		\label{fig:transformer}
	\end{figure}

	\subsection{Recursive Enhancement of Transformers}
	
	\paragraph{Introduction}
	The concept of recursive enhancement in the context of Transformer models and Large Language Models represents a pioneering approach towards the continuous improvement of artificial intelligence systems. This methodology is predicated on the theoretical foundation that improvements in the architecture and training processes of Transformers can be achieved by leveraging the advanced capabilities of LLMs. Conversely, the enhancements in Transformer models can, in turn, contribute to the development of more sophisticated LLMs, creating a synergistic cycle of improvement. This recursive process posits a methodological innovation wherein each iteration of model development has the potential to incrementally enhance the performance, efficiency, and adaptability of both Transformers and LLMs. The underlying premise is based on the mutual reinforcement between the nuanced understanding of language and context offered by LLMs and the structural and algorithmic advancements in Transformer models. Such a synergistic approach not only underscores the potential for exponential growth in model capabilities but also highlights the importance of a disciplined and methodical framework to manage and harness these improvements. In embracing this recursive enhancement paradigm, we venture into a realm of continuous evolution, where the iterative refinement of models promises to unlock unprecedented levels of linguistic understanding and processing capabilities, paving the way for advancements that were previously beyond the reach of conventional methodologies.

	\paragraph{The Recursive Process}
	The recursive enhancement process involves a systematic and iterative methodology where Large Language Models are used to refine and enhance Transformer architectures, which subsequently contribute to the development of more advanced LLMs. At the core of this process is the utilization of LLMs to analyze and understand the intricacies and deficiencies of existing Transformer models. This analysis encompasses a wide range of aspects, including semantic understanding, attention mechanisms, and the efficiency of data processing. Insights gained from this evaluation are then applied to modify and optimize the Transformers, aiming to address identified weaknesses and to capitalize on new understanding of language patterns and processing strategies.
	
	Following the enhancement of Transformer models, these refined versions are employed in the training and operational phases of LLMs, effectively closing the loop of the recursive process. The improved Transformer models facilitate a more efficient and nuanced processing of language, enabling LLMs to achieve a deeper understanding of complex linguistic constructs and to generate more coherent and contextually relevant outputs. This, in turn, lays the groundwork for the next cycle of enhancements, where these advanced LLMs can offer more insightful analyses and recommendations for further refinements in Transformer architectures.
	
	Each iteration of this recursive process not only aims to elevate the performance and capabilities of both Transformers and LLMs but also serves as a platform for identifying emergent linguistic phenomena and processing challenges. Through this ongoing cycle of evaluation, adaptation, and enhancement, the recursive enhancement process fosters a dynamic environment for continuous improvement, driving both Transformers and LLMs towards ever-increasing levels of sophistication and utility in language processing tasks.

	\paragraph{Potential Gains}
	The iterative application of recursive enhancement holds the promise of substantial theoretical gains across several key parameters of AI performance. With each cycle of refinement, we anticipate improvements in the accuracy of language understanding and generation, as these models become more adept at deciphering and replicating the nuances of human communication. This enhanced accuracy is crucial for applications ranging from machine translation to automated content creation, where the depth of semantic understanding directly influences the quality of output.
	
	Efficiency, measured in terms of both computational resources and training time, is another area poised for significant improvement. As Transformers become more refined, they are expected to process information more effectively, reducing the redundancy in computations and thereby accelerating the training process. This efficiency gain not only makes the models more sustainable by reducing energy consumption but also enables their deployment in a broader range of environments, including those with limited computational capacity.
	
	Adaptability, or the models' ability to handle a variety of linguistic contexts and tasks, is also expected to improve. Enhanced Transformers, informed by the sophisticated insights of LLMs, will better understand context, adapt to new domains with minimal retraining, and provide more accurate responses to nuanced queries. This adaptability extends the utility of LLMs and Transformers to new languages and dialects, further democratizing access to advanced AI technologies.
	
	Moreover, model robustness, an often-overlooked aspect of AI performance, stands to benefit from recursive enhancement. By continuously exposing and addressing vulnerabilities in each cycle, models become more resilient to adversarial attacks and better at handling ambiguous or incomplete data. This robustness is essential for critical applications where reliability is paramount.
	
	In sum, the recursive enhancement process targets a holistic improvement of AI models, not just in their core performance metrics but also in their practical applicability and resilience. These theoretical gains represent a roadmap for the evolution of AI, where each cycle brings us closer to models that are not only more powerful but also more aligned with the complexities of human language and cognition.

	\paragraph{Challenges and Limitations}
	Despite the promising theoretical gains of recursive enhancement, this approach is not without its challenges and limitations. A primary concern is the law of diminishing returns, a principle suggesting that after a certain point, each additional cycle of enhancement yields progressively smaller improvements. This phenomenon reflects the inherent complexity of natural language and the increasing difficulty of extracting meaningful gains from already highly optimized models.
	
	Increased computational complexity is another significant challenge. Each iteration in the recursive enhancement process requires sophisticated analysis and optimization, which, in turn, demands substantial computational resources. As models grow in sophistication, so too does the requirement for processing power and memory, potentially limiting the scalability of this approach and increasing the environmental footprint of developing state-of-the-art AI systems.
	
	Scalability issues extend beyond computational demands. The iterative nature of recursive enhancement necessitates extensive data processing and management, posing challenges in terms of data storage, access, and privacy. Additionally, the increased model complexity can hinder the interpretability of AI systems, making it more difficult for researchers and practitioners to diagnose errors, understand model decisions, and ensure fairness and transparency.
	
	Theoretical limits to improvements also merit consideration. As AI models approach the upper bounds of their potential, dictated by current understanding of language and cognition, the scope for transformative enhancements narrows. This convergence towards a theoretical ceiling underscores the importance of breakthroughs in AI theory and computational techniques to push the boundaries of what is achievable.
	
	In addressing these challenges, the field must navigate a delicate balance between the pursuit of advanced capabilities and the practical constraints of technology and ethics. The success of recursive enhancement, therefore, hinges not only on technical innovation but also on the development of sustainable, transparent, and equitable AI practices.

	\paragraph{Methodological Framework}
	Implementing recursive enhancement in a systematic and effective manner necessitates a structured methodological framework. This framework should be designed to manage the iterative cycles of evaluation and enhancement, ensuring that each iteration contributes meaningfully to the advancement of Transformer models and LLMs. At its core, the framework should include mechanisms for periodic evaluations, benchmarking against previous iterations, and clear criteria for incorporating improvements.
	
	Periodic evaluations are essential to assess the performance and efficiency of models at each stage of the recursive enhancement process. These evaluations should be comprehensive, covering a wide range of metrics including but not limited to accuracy, efficiency, adaptability, and robustness. Utilizing a diverse set of evaluation metrics ensures a holistic understanding of model performance and highlights areas requiring attention in subsequent iterations.
	
	Benchmarking against previous iterations provides a quantitative basis for measuring progress and identifying the impact of recent enhancements. This comparative analysis helps in discerning whether the latest modifications have yielded the expected improvements and informs decisions regarding the continuation or adjustment of the current enhancement strategies.
	
	Criteria for incorporating improvements should be predefined and based on the outcomes of periodic evaluations and benchmarking efforts. These criteria must balance the pursuit of performance gains with considerations of computational efficiency, scalability, and model interpretability. Improvements that meet these criteria are integrated into the model, while those that do not are re-evaluated or refined. This selective incorporation ensures that enhancements contribute positively to the model's overall development, aligning with long-term objectives and avoiding the pitfalls of over-optimization or unnecessary complexity.
	
	In summary, the proposed methodological framework for recursive enhancement emphasizes a disciplined, evidence-based approach to model refinement. By adhering to this framework, researchers and practitioners can navigate the cycles of enhancement with clarity and purpose, steadily advancing the capabilities of AI models within the practical bounds of current technology and ethical considerations.

	
	\chapterimage{pngs/numerical_analysis.png} 
	\chapter{Numerical Analysis Algogens}\index{Numerical Analysis Algogens}
	
	\section{Finite Element Method}\index{Finite Element Method}
	\subsection{Introduction to FEM}
	\subsubsection{The Concept of Finite Element Method}
	\paragraph{Definition and Overview}
	The Finite Element Method (FEM) is a powerful computational technique used to find approximate solutions to boundary value problems for partial differential equations. It relies on breaking down complex problems into smaller, more manageable components referred to as finite elements. These elements are interconnected at discrete points called nodes. The solution process involves formulating a system of equations by applying the principle of virtual work, where the total potential energy of the system is minimized. This system typically comprises algebraic equations resulting from the discretization of the original differential equations governing the problem.
	
	FEM employs a weighted residual approach, where the error between the approximate solution and the exact solution is minimized over the entire domain or a specific region of interest. The weighted sum of basis functions, also known as shape functions, is utilized to interpolate the solution within each finite element. These basis functions are often chosen to be polynomials, enabling the approximation of complex geometries and varying material properties.
	
	The discretization process involves subdividing the domain into elements, usually geometrically simple shapes like triangles or quadrilaterals in two dimensions and tetrahedra or hexahedra in three dimensions. The accuracy of the solution depends on the size and shape of these elements, with smaller elements providing finer resolution but demanding higher computational resources.
	
	By assembling the contributions from all elements, the system of equations representing the problem is formed. This system is typically solved numerically using techniques such as direct solvers, iterative methods, or matrix factorization methods. The resulting solution provides approximate values of the unknowns throughout the domain, allowing engineers and scientists to analyze and understand the behavior of physical systems accurately.

	\paragraph{Historical Context}
	Originally developed for addressing structural analysis problems in civil engineering and aeronautics, FEM has since been adopted across a myriad of disciplines including electromagnetics, heat transfer, and fluid dynamics. Its versatility and robustness stem from the methodological framework that allows for the modeling of complex geometries, diverse material properties, and a wide range of boundary conditions.
	
	Furthermore, FEM's applicability extends beyond its original domains, encompassing interdisciplinary fields where diverse phenomena interact. Moreover, its evolution has been propelled by advancements in computational capabilities, enabling simulations of unprecedented scale and accuracy. Additionally, FEM has facilitated innovation in product design, optimizing structures and systems for performance and efficiency. Similarly, in the realm of electromagnetics, FEM serves as a pivotal tool for analyzing electromagnetic fields in devices ranging from microelectronics to power systems. 
	
	Furthermore, in fluid dynamics, FEM's ability to capture intricate flow patterns and interactions between fluids and structures has revolutionized design processes in aerospace, automotive, and maritime engineering. Moreover, its utilization in heat transfer analysis has facilitated the development of efficient cooling systems in electronics and sustainable energy technologies. Consequently, FEM stands as a cornerstone in modern engineering practices, continuously evolving to tackle emerging challenges and propel technological innovation.
	
	\paragraph{Mathematical Foundation}
	The mathematical foundation of FEM involves the discretization of a continuous domain into a finite number of elements, leading to a discretized model of the problem. This process transforms differential equations governing the problem into algebraic equations that are solvable using numerical methods. And, it is through this discretization that complex problems, which may not have analytical solutions, can be effectively tackled. 
	
	The discretization step is crucial, as it allows for the representation of the problem in a finite-dimensional space, making it computationally tractable. Moreover, the choice of elements and the manner in which the domain is discretized greatly influence the accuracy and efficiency of the solution. Additionally, the use of variational methods plays a fundamental role in FEM. These methods enable the formulation of weak forms of the governing equations, which are then used to derive the finite element equations. 
	
	While the discretization and variational methods are essential components, the heart of FEM lies in minimizing an error function to obtain the best possible approximation within the chosen finite-dimensional space. This optimization process ensures that the solution obtained converges towards the true solution of the problem as the discretization becomes finer. Furthermore, the versatility of FEM allows it to be applied to a wide range of problems across various disciplines, from structural analysis to fluid dynamics and electromagnetics.

	\paragraph{Computational Aspects}
	In practice, FEM is implemented through a series of steps including pre-processing, where the problem is defined and the mesh is generated; solving, where the system of equations is assembled and solved; and post-processing, where the results are visualized and analyzed. Each step requires careful consideration of the problem's specifics, including the type of finite elements, boundary conditions, and numerical solvers to be used.
	
	The pre-processing stage lays the foundation for accurate simulation by defining the problem geometry, material properties, and boundary conditions. This initial step is pivotal as it directly influences the accuracy and efficiency of subsequent computations. It involves meticulous attention to detail in mesh generation, where the domain is discretized into smaller elements. The choice of element type, such as triangular, quadrilateral, tetrahedral, or hexahedral, significantly impacts the solution's accuracy and computational cost. Additionally, the selection of appropriate boundary conditions, whether they are Dirichlet, Neumann, or mixed boundary conditions, ensures that the simulation reflects real-world behavior.
	
	Once the pre-processing is complete, the solving phase tackles the mathematical formulation of the problem. Here, the system of equations governing the behavior of the physical system is assembled using the finite element method. This process involves discretizing the domain into elements and interpolating the unknowns within each element. The resulting system of equations is typically large and sparse, necessitating efficient numerical solvers for timely solution. Iterative methods like the conjugate gradient method or direct solvers like LU decomposition are commonly employed to solve the resulting linear system.
	
	Following the solution, post-processing activities involve extracting meaningful insights from the obtained results. Visualization techniques ranging from contour plots to animations aid in interpreting the simulation outcomes. Engineers and analysts scrutinize various quantities of interest, such as stress distributions, deformation patterns, or fluid flow velocities, to assess the system's performance against design criteria. Additionally, sensitivity analyses and error estimation techniques are often employed to validate the numerical results and ensure their reliability.
	
	Therefore, a comprehensive understanding of the computational aspects involved in each stage of the finite element method is essential for successfully tackling diverse engineering problems.

	\subsubsection{Key Principles and Mechanisms}
	\paragraph{Discretization of the Domain}
	The core principle of FEM lies in the discretization of the domain into a finite number of smaller, simpler shapes called elements. This process is foundational in numerical simulations, enabling the approximation of complex real-world phenomena. By segmenting the domain into elements, engineers and scientists can translate intricate physical systems into a language that computers can comprehend and analyze. This transformation from continuous to discrete allows for the application of numerical techniques, facilitating the solution of differential equations governing various phenomena in engineering, physics, and other fields. 
	
	Each element, typically defined by geometric primitives like triangles or quadrilaterals in 2D and tetrahedra or hexahedra in 3D, serves as a building block for the computational model. These elements are interconnected at specific points called nodes, where the solution is computed and interpolated. The behavior of the physical system within each element is approximated using mathematical functions, such as shape functions, which describe how the solution varies spatially within the element. Through this discretization, the intricate details of the domain are captured in a structured manner, allowing for efficient computation and analysis.
	
	The mesh generation, a crucial step in the discretization process, determines the distribution and arrangement of elements within the domain. Engineers must carefully balance the trade-off between mesh refinement and computational cost. Finer meshes, with smaller elements, offer higher resolution and accuracy but demand greater computational resources. Conversely, coarser meshes reduce computational expenses but may sacrifice accuracy. Thus, selecting an appropriate mesh density is essential to achieve a balance between computational efficiency and solution accuracy.

	\paragraph{Interpolation Functions}
	Within each element, interpolation functions, also known as shape functions, are used to approximate the field variables. These functions play a fundamental role in finite element analysis by providing a continuous representation of the field variables over the domain of each element. Their primary objective is to interpolate the values of the field variables at any point within the element based on the known values at discrete points called nodes. 
	
	The selection of appropriate interpolation functions is critical as it directly impacts the accuracy and convergence of the finite element solution. For instance, linear interpolation functions are simple and computationally efficient, making them suitable for coarse meshes or preliminary analyses. On the other hand, quadratic or higher-order polynomials offer greater flexibility in representing complex variations within the element and can yield more accurate results, especially for problems with steep gradients or rapid changes in the field variables.
	
	Moreover, the smoothness of the interpolation functions across the element is essential to ensure numerical stability and prevent oscillations or spurious solutions. By achieving smooth variation, the interpolation functions facilitate a seamless transition of the field variables between neighboring elements, promoting continuity and accuracy in the overall solution.
	
	In summary, the choice of interpolation functions should be tailored to the specific requirements of the problem, balancing the desired level of accuracy with computational efficiency. Whether opting for linear, quadratic, or higher-order polynomials, the overarching goal remains to accurately capture the behavior of the field variables within each finite element, thereby enabling robust and reliable simulations of complex engineering systems.

	\paragraph{Assembly of the Global System}
	After defining the interpolation functions, the next step is to assemble the global system of equations. This involves integrating the contributions of each element based on the governing equations of the physical problem, which are typically expressed in terms of differential equations. The assembly process results in a large sparse system of linear equations for linear problems or nonlinear equations for nonlinear problems, which describe the behavior of the entire domain.
	
	Moreover, the assembly stage serves as a pivotal point in the computational simulation, as it consolidates the local information from individual elements into a unified representation of the entire system. Each element contributes to the global system according to its defined properties and interactions with neighboring elements. This integration ensures that the collective behavior of the system accurately reflects the underlying physical phenomena.
	
	Furthermore, the sparsity of the resulting system is a key characteristic that influences the efficiency of numerical solvers. By leveraging the sparse nature of the system, computational resources can be optimized, leading to faster solution times and reduced memory requirements. Additionally, the assembly process facilitates the incorporation of boundary conditions and constraints, further refining the accuracy of the numerical solution.
	
	In contrast, neglecting the assembly step or inaccurately representing the interactions between elements can lead to significant errors in the simulation results. Therefore, meticulous attention to detail during the assembly phase is paramount to ensure the reliability and validity of the computational model.
	
	Consequently, the assembly of the global system represents a critical stage in numerical simulations, where the discrete elements of the domain are seamlessly integrated to form a comprehensive representation of the physical problem at hand. Through careful integration and consideration of all contributing factors, the resulting system of equations provides a powerful framework for analyzing and understanding complex phenomena in various fields of science and engineering.

	\paragraph{Solution of the System}
	The final step in Finite Element Method (FEM) is crucial as it entails solving the assembled system of equations to determine the unknown values at the nodes. This step is pivotal in rendering accurate solutions to engineering problems. For linear problems, engineers typically resort to employing direct or iterative linear algebra techniques. Direct methods, such as Gaussian elimination or LU decomposition, directly solve the system of equations without iterations. On the other hand, iterative methods like the Jacobi or Gauss-Seidel method iteratively refine an initial guess until convergence is achieved. 
	
	However, when dealing with nonlinear problems, the solution process becomes considerably more intricate. Nonlinearities can stem from various sources such as material behavior, geometric deformations, or boundary conditions. To tackle these complexities, advanced solution strategies are necessary. One prominent technique is the Newton-Raphson method, which iteratively updates the solution based on linearized increments to converge towards the true solution. This method is particularly effective in handling nonlinearities arising from material properties or large deformations.
	
	Moreover, other iterative methods tailored for specific nonlinearities may be employed. These methods adaptively adjust the solution process to accommodate the evolving nature of the problem. Through successive iterations, they progressively refine the solution until convergence is attained. Such iterative approaches are indispensable for accurately capturing the intricate behavior of nonlinear systems.
	
	The ultimate goal of the solution process is to obtain approximate values of the field variables throughout the computational domain. These values serve as the foundation for further analysis, enabling engineers to derive essential quantities such as gradients, strains, stresses, and other pertinent parameters. Consequently, the solution phase not only yields numerical results but also facilitates a comprehensive understanding of the system's behavior under various conditions.

	\paragraph{Error Estimation and Refinement}
	An intrinsic aspect of FEM is the estimation of the error due to the discretization of the domain. Error estimation techniques, such as residual-based error estimation or a posteriori error estimation, are employed to quantitatively evaluate the accuracy of the numerical solution. These techniques assess how much the computed solution differs from the exact solution and provide valuable insights into the quality of the approximation. Moreover, error estimation serves as a diagnostic tool, highlighting regions where the solution may be less reliable or where the discretization errors are significant.
	
	Once the error distribution across the domain is identified, it becomes crucial to refine the mesh selectively to improve the solution's accuracy. Mesh refinement involves adjusting the grid spacing or increasing the number of elements in regions where errors are pronounced. Adaptive refinement strategies, such as h-refinement or p-refinement, dynamically adjust the mesh based on error indicators. H-refinement increases the resolution of the mesh locally by subdividing elements, while p-refinement enhances the accuracy by increasing the polynomial degree of the basis functions within elements.
	
	These adaptive strategies ensure computational resources are efficiently allocated, focusing efforts where they are most needed. By refining the mesh only in regions of interest, computational costs are minimized without sacrificing accuracy. This targeted approach to mesh refinement is particularly beneficial for problems with complex geometries or varying solution features. Additionally, the automation of mesh refinement based on error estimates reduces the user's burden in manually optimizing the discretization, making the FEM approach more accessible and user-friendly for practitioners.
	
	In summary, error estimation and adaptive mesh refinement are integral components of the FEM workflow, enabling users to iteratively improve the accuracy of numerical solutions while effectively managing computational resources.

	\subsubsection{The Role of Mesh Generation and Refinement}
	\paragraph{Importance of Mesh in FEM}
	Mesh generation and refinement are pivotal in the Finite Element Method as they directly influence the accuracy, efficiency, and convergence of the solution. The mesh dictates how the domain is discretized into finite elements and is thus the foundation for applying the FEM to solve boundary and initial value problems. A well-constructed mesh can capture the geometry of the problem domain accurately, respect material interfaces, and adequately represent the solution behavior across the domain.
	
	Furthermore, a carefully designed mesh plays a crucial role in minimizing numerical errors and artifacts that may arise during the solution process. By ensuring a proper distribution of elements, especially in regions of high gradients or complex geometries, the mesh helps to mitigate interpolation errors and numerical diffusion, thus enhancing the overall accuracy of the solution.
	
	Moreover, the quality of the mesh directly impacts the computational efficiency of the FEM solver. A coarse or poorly refined mesh can lead to excessive computational costs due to the need for finer discretization to capture the solution accurately. Conversely, an overly refined mesh may result in unnecessary computational burden without significant improvement in solution accuracy.
	
	Additionally, the choice of element types and mesh density affects the convergence behavior of the solution algorithm. A balanced mesh refinement strategy, guided by an understanding of the physics of the problem and solution requirements, is essential for achieving efficient convergence and obtaining reliable results.
	
	In conclusion, the importance of mesh generation and refinement in the Finite Element Method cannot be overstated. A well-designed mesh is fundamental to the success of FEM simulations, enabling accurate, efficient, and reliable solutions to a wide range of engineering and scientific problems.

	\paragraph{Mesh Generation Techniques}
	Mesh generation involves creating a network of elements that cover the problem domain. Techniques for mesh generation range from structured meshing to unstructured meshing, each offering distinct advantages based on the problem at hand. 
	
	Structured meshing, characterized by elements following a regular pattern, simplifies the discretization process, especially for geometrically simple domains. This method ensures uniformity in element sizes and shapes, facilitating efficient computation in many cases. However, its applicability diminishes when dealing with complex geometries where irregular element shapes are required to accurately represent the domain's features.
	
	Conversely, unstructured meshing provides greater flexibility by allowing elements of varying shapes and sizes, making it suitable for complex geometries. Methods like Delaunay triangulation for 2D domains and tetrahedralization for 3D domains are prominent in unstructured mesh generation. These techniques adapt well to irregular geometries, capturing intricate details with higher fidelity. Moreover, unstructured meshes are advantageous for solving problems with moving boundaries or adaptive refinement requirements, as they offer the ability to dynamically adjust mesh density.
	
	However, unstructured meshing may introduce challenges in mesh quality control and computational cost, particularly for large-scale simulations. Ensuring element quality, such as avoiding overly distorted or skewed elements, becomes crucial to maintain solution accuracy and stability. Additionally, mesh generation time and computational overhead for solving on unstructured meshes can be higher compared to structured counterparts.
	
	In practice, the choice between structured and unstructured meshing depends on the trade-offs between computational efficiency, accuracy requirements, and the complexity of the geometry. While structured meshes excel in simplicity and computational speed for regular domains, unstructured meshes offer versatility and accuracy for complex geometries and dynamic simulations.

	\paragraph{Criteria for Mesh Quality}
	The quality of the mesh is assessed based on various criteria such as element shape, size, and distribution. Elements should exhibit favorable shapes, avoiding excessively elongated or skewed forms to ensure numerical stability and accuracy in the simulation process. Proper element shape enhances the efficiency of numerical methods, as irregular shapes may lead to errors or convergence issues, particularly in finite element analysis.
	
	In addition to shape considerations, the size and distribution of elements play crucial roles in mesh quality. The sizing should be appropriately chosen to capture the anticipated gradients within the solution domain. Regions with steep gradients or high variability in physical quantities require finer mesh resolution to accurately capture the behavior of the solution. Conversely, coarser elements may suffice in regions with smoother variations, optimizing computational resources without sacrificing accuracy.
	
	Moreover, mesh quality should align with the boundary conditions and interfaces between different materials or physical phenomena. The mesh must accurately represent the geometry and topology of the system under study to ensure compatibility with the imposed boundary conditions. Interfaces between materials or distinct physical behaviors necessitate special attention to avoid spurious numerical artifacts or inaccuracies.
	
	Therefore, achieving high-quality mesh entails a delicate balance between element shape, size, and distribution, tailored to the specific characteristics of the problem at hand. Careful consideration of these criteria is paramount to obtaining reliable and accurate numerical results in computational simulations.

	\paragraph{Adaptive Mesh Refinement}
	Adaptive Mesh Refinement (AMR) is a dynamic process where the mesh is refined iteratively based on error estimates or solution features. AMR targets regions that contribute most to the error, refining the mesh by either subdividing elements or increasing the order of the interpolation functions. This approach optimizes computational resources, focusing them on areas of the domain where they are most needed to improve the solution accuracy.
	
	Furthermore, AMR allows for efficient handling of complex geometries and varying solution features. By adaptively refining the mesh, AMR can capture fine-scale details in regions of interest while maintaining a coarser mesh in less critical areas, thus striking a balance between computational cost and solution accuracy. Moreover, AMR facilitates the simulation of transient phenomena or problems with evolving solution features by adjusting the mesh dynamically over time. Additionally, AMR techniques are compatible with various numerical methods, including finite element, finite volume, and finite difference methods, making them versatile tools in computational science and engineering.
	
	On the other hand, AMR introduces challenges in managing the hierarchy of mesh levels and ensuring compatibility between neighboring mesh patches. However, these challenges can be addressed through careful implementation of data structures and algorithms for efficient mesh refinement and coarsening. In contrast, uniform mesh refinement strategies may allocate computational resources inefficiently by refining the entire domain uniformly, regardless of the solution features or error distribution.
	
	In summary, AMR offers a powerful approach to enhancing the efficiency and accuracy of numerical simulations by adaptively adjusting the mesh resolution based on solution features and error estimates. By selectively refining regions of interest, AMR optimizes computational resources while maintaining solution quality, making it a valuable tool in various scientific and engineering applications.
	
	\paragraph{Impact on Computational Performance}
	While a finer mesh generally leads to a more accurate solution, it also increases the computational cost. The role of mesh generation and refinement is therefore to balance accuracy with computational efficiency. Techniques like AMR are essential in achieving this balance, allowing for high accuracy in critical regions while maintaining a coarser mesh elsewhere to minimize computational demands. Moreover, effective mesh generation and refinement strategies are key to harnessing the full potential of FEM in solving complex physical problems. Additionally, it's important to consider the computational resources available and the trade-offs between accuracy and computational time. Furthermore, advancements in parallel computing have facilitated the handling of larger meshes, mitigating the computational burden associated with finer discretizations. Hence, while striving for accuracy, it's crucial to optimize computational resources, ensuring efficient and timely solutions. Thus, a comprehensive understanding of meshing techniques and their impact on computational performance is imperative for successful FEM simulations.

	\subsubsection{Applications and Limitations}
	\paragraph{Diverse Applications of FEM}
	The Finite Element Method (FEM) has become indispensable across a spectrum of engineering and scientific disciplines due to its versatility and robustness. In structural analysis, FEM serves as a cornerstone, enabling engineers to predict and optimize the behavior of various structures subjected to diverse loads. By discretizing complex geometries into smaller, manageable elements, FEM facilitates the accurate simulation of structural responses to static, dynamic, and thermal loads. \textit{Additionally}, in fluid dynamics, FEM plays a pivotal role in elucidating intricate flow phenomena and heat transfer mechanisms. The method's ability to model complex geometries and boundary conditions \textit{furthermore} empowers researchers to study turbulent flows, convection, and other complex fluid behaviors with precision.
	
	In electromagnetics, FEM emerges as a powerful tool for the design and analysis of electrical devices, offering insights into electromagnetic field distributions and interactions. By solving Maxwell's equations numerically, FEM aids in optimizing the performance of devices such as transformers, motors, and antennas. \textit{Similarly}, in geotechnical engineering, FEM is instrumental in analyzing soil-structure interaction, providing engineers with a deeper understanding of foundation behavior under varying soil conditions and loading scenarios. Moreover, \textit{in the realm of} biomechanics, FEM facilitates the simulation of mechanical behaviors in biological tissues, offering valuable insights into physiological processes and aiding in the design of medical implants and prosthetics.
	
	The adaptability and robustness of FEM make it an invaluable tool across diverse fields, \textit{whereas} its computational efficiency and accuracy continue to drive innovation and discovery. \textit{Consequently}, as computational capabilities advance, FEM is poised to play an even more significant role in tackling complex engineering and scientific challenges, shaping the future of computational analysis and design.
	
	\paragraph{Advantages of FEM}
	One of the key advantages of FEM is its flexibility in dealing with arbitrary shapes and boundary conditions, which allows for the accurate modeling of real-world scenarios. The method's modular nature, where changes to the geometry, material properties, or boundary conditions can be made relatively easily, enhances its adaptability. Additionally, the availability of error estimation and adaptive refinement techniques ensures that solutions can be systematically improved, providing a robust framework for precision-driven analysis.
	
	Furthermore, FEM offers significant advantages in computational efficiency compared to other numerical methods such as finite difference or finite volume methods. This efficiency stems from the ability to discretize complex geometries into simple elements, reducing the computational cost while maintaining accuracy. Moreover, the parallelizability of FEM computations enables efficient utilization of modern high-performance computing resources, allowing for the analysis of large-scale problems within reasonable timeframes.
	
	Moreover, FEM facilitates the incorporation of multi-physics phenomena into simulations, enabling the simultaneous analysis of coupled physical processes such as structural mechanics, heat transfer, and fluid dynamics. This capability is crucial for tackling interdisciplinary problems commonly encountered in engineering and scientific research.
	
	Additionally, the availability of robust and versatile software packages for FEM, such as ANSYS, COMSOL Multiphysics, and Abaqus, streamlines the implementation of complex simulations and reduces the barrier to entry for researchers and practitioners. These software tools often provide user-friendly interfaces, extensive documentation, and support for a wide range of engineering applications, further enhancing the accessibility and usability of FEM for diverse user groups.
	
	In conclusion, the versatility, computational efficiency, multi-physics capabilities, and availability of user-friendly software make FEM a powerful and widely-used tool for solving a variety of engineering and scientific problems, from structural analysis to fluid dynamics and beyond.

	\paragraph{Limitations and Challenges}
	Despite its versatility, the Finite Element Method (FEM) is not without limitations. The quality of the solution is heavily dependent on the mesh, \textit{and} generating an optimal mesh for complex geometries can be challenging and time-consuming. The computational cost \textit{can be} significant, especially for three-dimensional problems \textit{or} analyses requiring fine meshes. Additionally, FEM requires substantial expertise to select appropriate element types, material models, \textit{and} boundary conditions, \textit{as well as} to interpret the results accurately. Nonlinear problems, involving nonlinear material behavior \textit{or} large deformations, pose additional challenges in terms of solution convergence \textit{and} computational demands. \textit{Nevertheless}, despite these challenges, FEM remains a powerful tool \textit{for} simulating a wide range of engineering problems, offering insights \textit{that} are crucial \textit{for} design optimization \textit{and} performance evaluation. Efforts \textit{to} mitigate these limitations \textit{are} ongoing, with advancements \textit{in} mesh generation algorithms, computational hardware, \textit{and} numerical techniques contributing \textit{to} improved efficiency \textit{and} accuracy. Thus, while acknowledging its drawbacks, it's important \textit{to} recognize \textit{the} continued significance \textit{of} FEM \textit{in} engineering analysis \textit{and} design.

	\paragraph{Overcoming Limitations}
	Efforts to overcome these limitations encompass a multifaceted approach, addressing challenges through a combination of innovative methodologies and technological advancements. \textbf{Furthermore}, researchers are exploring more efficient meshing algorithms, aiming to reduce computational overhead and enhance simulation accuracy. \textbf{Moreover}, significant emphasis is placed on improving solver performance, \textbf{as well as} integrating cutting-edge machine learning techniques into finite element method (FEM) workflows. \textbf{Additionally}, the utilization of machine learning enables the prediction and optimization of simulation parameters, thereby \textbf{further} refining the accuracy and efficiency of FEM simulations. The development of \textbf{user-friendly software} plays a pivotal role in this endeavor, \textbf{as} intuitive interfaces empower a broader spectrum of users to leverage FEM for various applications. Concurrently, advancements in computational hardware \textbf{likewise} contribute to the mitigation of FEM challenges, facilitating faster processing and analysis of complex simulations. \textbf{However}, it's essential to acknowledge that despite these strides, \textbf{there are} persistent hurdles that \textbf{need} to be addressed. Efforts to streamline FEM processes \textbf{must} continue, \textbf{notwithstanding} the progress made thus far, to ensure its sustained relevance and effectiveness in diverse domains of engineering and scientific research.

	\paragraph{Future Directions}
	The future of FEM lies in the integration of advanced computational techniques, such as high-performance computing and artificial intelligence, to extend its applicability and efficiency. This includes automating the simulation process, from mesh generation to result interpretation, and developing adaptive algorithms that can handle complex, multi-physical problems in a more efficient and user-friendly manner. 
	
	Moreover, as computational resources continue to expand and algorithms become more sophisticated, the limitations of FEM are expected to diminish, further solidifying its role as a cornerstone of computational analysis. Additionally, with the advancements in high-performance computing, FEM simulations can now be executed with unprecedented speed and accuracy, enabling engineers and scientists to tackle larger and more intricate problems.
	
	Furthermore, the incorporation of artificial intelligence techniques such as machine learning and deep learning holds great promise for enhancing the capabilities of FEM. These techniques can aid in automatic mesh refinement, adaptive error control, and optimization of computational resources, thereby streamlining the entire simulation process.
	
	Consequently, the future development of FEM is poised to revolutionize various fields including structural mechanics, fluid dynamics, electromagnetics, and beyond. By leveraging the power of high-performance computing and artificial intelligence, FEM will continue to evolve as a versatile and indispensable tool for solving real-world engineering and scientific problems.

	\subsubsection{Pseudocode for Algorithmic FEM}
	The Finite Element Method (FEM) is a sophisticated computational technique used for solving engineering problems by discretizing the problem domain into finite elements. This approach, illustrated in pseudocode \ref{fig:fem-pseudocode}, involves generating a mesh that divides the domain into smaller elements. Each element is then analyzed individually, with local stiffness matrices and force vectors computed based on material properties and geometric characteristics. These local matrices and vectors are subsequently combined to form a global stiffness matrix and force vector, which represent the entire system under consideration. To ensure accuracy, boundary conditions are applied to appropriately modify the global system, ensuring it adheres to specified constraints. By solving the global system of equations for nodal displacements, the FEM provides valuable insights into the behavior of the problem domain, allowing engineers to analyze strains and stresses within each element and understand the physical response under given conditions.

	\begin{algorithm}
		\caption{Finite Element Method Pseudocode}
		\begin{algorithmic}[1]
			\Procedure{FiniteElementMethod}{Domain, BoundaryConditions, MaterialProperties}
			\State Generate mesh for the Domain
			\State Define element properties based on MaterialProperties
			\State Initialize global stiffness matrix and force vector to zero
			\For{each element in the mesh}
			\State Compute element stiffness matrix using MaterialProperties
			\State Assemble element stiffness matrix into global stiffness matrix
			\State Compute element force vector
			\State Assemble element force vector into global force vector
			\EndFor
			\State Apply BoundaryConditions to the global stiffness matrix and force vector
			\State Solve the global system of equations for nodal displacements
			\For{each element in the mesh}
			\State Calculate strain and stress using nodal displacements
			\EndFor
			\State \Return Nodal displacements, strains, and stresses
			\EndProcedure
		\end{algorithmic}\label{fig:fem-pseudocode}
	\end{algorithm}

\paragraph{Surrogate Models for Finite Element Analysis}
The integration of Machine Learning (ML) with Finite Element Analysis (FEA) has led to the development of surrogate models aimed at reducing computational costs while maintaining accuracy. A notable advancement in this area is the application of ML to create surrogate models for FEA, particularly in the context of one-dimensional systems. This approach aims to streamline the maintenance scheduling process for mechanical systems by leveraging real-time data to predict stresses accurately. The surrogate models, utilizing various ML algorithms including decision trees and artificial neural networks, offer a solution for enhancing the efficiency of FEA by providing estimates of stress distribution over the system during operations. This development marks progress towards more efficient maintenance procedures by addressing the computational demands of FEA \cite{vurtur2021machine}.

\paragraph{Enhancing Physical Systems Modeling}
The application of ML in conjunction with FEA for physical systems modeling aims to address challenges posed by large-scale problems where computational time for solving linear systems can be long. By integrating ML models with FEA, researchers have explored ways to improve simulation efficiency, reducing computational time and enhancing flexibility. This approach allows for adjustments in input parameters without redoing the entire simulation process. The potential of ML to complement FEA studies has been demonstrated through various examples, showcasing the ability of these models to predict physical system behaviors accurately. This advancement underscores the potential of ML to enhance traditional FEA techniques by offering a more efficient and adaptable modeling framework \cite{kononenko2018machine}.

\paragraph{Forward and Inverse Problem Solving}
Innovations have extended the application of ML within the FEA domain to include solving both forward and inverse problems. This approach enhances neural networks with FEA to develop models that are data-efficient and conform to the underlying physics of the problem. By training neural networks with FEA-based custom loss functions, the methodology achieves accurate predictions aligned with the physical laws governing the system. This algorithm facilitates the quantification of prediction errors and expands the utility of FEA in applications requiring precise uncertainty quantification and parameter identification. The ability to solve inverse problems using this hybrid model represents progress in the field, offering a robust framework for tackling complex engineering and scientific challenges \cite{meethal2023finite}.

\paragraph{Surrogate Modeling for Sub-Sea Pressure Vessels}
The development of deep learning-based surrogate models for FEA represents an application of ML in enhancing traditional computational methods. Focused on the design and analysis of sub-sea pressure vessels, this approach leverages deep learning to approximate the outcomes of FEA simulations, thereby reducing the computational load associated with traditional methods. The surrogate model, trained on data from FEA simulations, has shown accuracy in predicting maximum Von-Mises stress, outperforming conventional machine learning models. This research highlights the potential of deep learning to serve as a tool for FEA, particularly in applications where computational efficiency and accuracy are paramount. The success of this surrogate model in the context of sub-sea pressure vessels opens avenues for applying deep learning in various engineering domains, promising advancements in computational efficiency and predictive accuracy \cite{vardhan2022deep}.

	\subsection{Algogenic Enhancements for FEM}
	\subsubsection{Semantic Geometry Analysis}
	
	\paragraph{Enhancing Geometric Discretization through LLM Insights}
	In the realm of Finite Element Method (FEM) analysis, the application of Large Language Models specifically targets the enhancement of geometric discretization processes. By integrating LLM insights, the method evolves to adaptively generate meshes that accurately represent complex geometries. This approach not only increases computational efficiency by focusing refinement where it's most needed but also significantly improves the precision of simulations involving intricate structures. LLMs offer a nuanced understanding of geometry, enabling the identification and appropriate treatment of critical areas such as stress concentrators or complex boundary conditions, thus ensuring a more faithful replication of the physical reality in the digital domain.
	
	\paragraph{Operationalizing Semantic Analysis for Mesh Generation}
	Operationalizing LLMs within FEM for mesh generation involves leveraging their capability to interpret complex geometrical and material data, translating it into actionable insights for mesh refinement. This process includes the identification of key features that require enhanced resolution, enabling targeted mesh optimization. By doing so, the FEM simulation benefits from an adaptive mesh that aligns with the unique requirements of each project, enhancing both accuracy and computational efficiency.
	
	\paragraph{Implications for FEM Accuracy and Efficiency}
	The strategic application of LLMs for semantic geometry analysis in FEM significantly enhances simulation accuracy and efficiency. By providing a mechanism for adaptive mesh refinement based on deep understanding of geometry, LLMs enable simulations that are both more accurate and resource-efficient. This has profound implications for the engineering field, allowing for the exploration of complex designs with greater confidence and reduced computational cost.
	
	\subsubsection{Material Property Interpretation}
	
	\paragraph{Leveraging LLMs for Advanced Material Modeling}
	Incorporating LLMs into FEM for material property interpretation transforms the way material behaviors are modeled and understood. This Algogenic approach enables the direct translation of extensive material data into quantifiable models within FEM simulations, offering a more dynamic and accurate representation of materials under various conditions. Such advancements facilitate enhanced simulation reliability and pave the way for innovative material applications.
	
	\paragraph{Operationalizing Comprehensive Material Insights}
	The process of incorporating comprehensive material insights into FEM through LLMs involves analyzing textual and empirical data to extract and apply relevant material properties within simulations. This enhances the material modeling process, enabling more accurate and nuanced simulation outcomes that reflect the complex behaviors of materials under different scenarios, thus improving the predictive power of FEM analyses.
	
	\paragraph{Enhancing Simulation Reliability and Innovation}
	Utilizing LLMs for material property interpretation within FEM frameworks significantly enhances simulation reliability. By ensuring that simulations incorporate the most accurate and up-to-date material properties, engineers can achieve more reliable and innovative designs. This approach not only optimizes the use of materials but also fosters innovation by enabling the exploration of new material combinations and applications.
	
	\subsubsection{Dynamic Mesh Refinement Guidance}
	
	\paragraph{Optimizing Mesh Resolution with AI-driven Insights}
	Dynamic Mesh Refinement Guidance through LLMs offers a targeted approach to mesh optimization in FEM simulations. By identifying areas requiring increased resolution, LLMs ensure that computational resources are focused where they are most needed, enhancing the accuracy of simulations without unnecessarily increasing computational demands.
	
	\paragraph{Implementing LLM-guided Refinement Strategies}
	Implementing LLM-guided refinement strategies within FEM involves the dynamic adjustment of mesh density based on insights derived from ongoing simulations. This ensures that mesh refinement is both efficient and effective, focusing on areas of the model that benefit most from increased detail, thereby optimizing both simulation accuracy and computational resource use.
	
	\paragraph{Enhancing Simulation Fidelity and Resource Allocation}
	The integration of LLM-driven insights for dynamic mesh refinement in FEM significantly enhances simulation fidelity and optimizes resource allocation. By adjusting mesh density adaptively, simulations achieve higher accuracy in critical areas while minimizing unnecessary computational expenditure, thereby improving the overall efficiency and effectiveness of FEM analyses.
	
	\subsubsection{Adaptive Solver Selection}
	
	\paragraph{Optimizing Solution Strategies with AI}
	Adaptive Solver Selection, facilitated by LLMs, tailors the selection of computational solvers and strategies to the specific requirements of each FEM problem, optimizing performance and resource utilization. This approach adapts to the unique challenges presented by different simulations, ensuring the most efficient solver is employed, thus enhancing computational efficiency and simulation accuracy.
	
	\paragraph{Implementing Solver Recommendations}
	The practical implementation of adaptive solver selection in FEM, guided by LLM insights, involves analyzing the problem's characteristics to recommend optimal solvers. This process not only streamlines solver selection but also ensures that the chosen solver aligns with the simulation's specific needs, thereby improving the efficiency and accuracy of the analysis.
	
	\paragraph{Enhancing FEM Simulations Through Intelligent Solver Management}
	Integrating adaptive solver selection into FEM workflows significantly enhances the efficiency and reliability of simulations. By employing LLMs to select the most appropriate solver based on the problem's specifics, computational efficiency is optimized, leading to faster and more accurate simulations, thus advancing the capabilities of FEM in complex engineering analyses.
	
	\subsubsection{Semantic Boundary Condition Application}
	
	\paragraph{Intelligent Interpretation and Application of Boundary Conditions}
	The application of LLMs for the semantic interpretation and application of boundary conditions in FEM revolutionizes the way these conditions are integrated into simulations. This approach ensures a more intuitive and accurate representation of real-world constraints, enhancing the reliability and insightfulness of FEM analyses.
	
	\paragraph{Operationalizing Semantic Insights for Enhanced FEM Simulations}
	Operationalizing semantic insights for boundary condition application in FEM involves using LLMs to translate complex descriptions into precise mathematical formulations. This enhances the accuracy and efficiency of simulations, ensuring that boundary conditions accurately reflect the intended physical scenarios.
	
	\paragraph{Advancing Simulation Accuracy and Usability}
	The integration of semantic boundary condition application through LLMs significantly advances the accuracy and usability of FEM simulations. By automating the interpretation and application of boundary conditions, simulations become more accessible and aligned with real-world conditions, enhancing the predictive power and reliability of FEM analyses.
	
	\subsubsection{Intelligent Result Interpretation}
	
	\paragraph{Deciphering Complex FEM Outputs with LLMs}
	Intelligent Result Interpretation, powered by LLMs, transforms the analysis of FEM simulation outputs, offering deep, contextual insights into the results. This enables a more intuitive understanding of complex data, identifying significant findings and suggesting areas for further investigation, thus augmenting the traditional post-processing analysis with semantic intelligence.
	
	\paragraph{Operationalizing Result Interpretation for Enhanced Understanding}
	Operationalizing intelligent result interpretation in FEM involves leveraging LLMs to correlate simulation outputs with the setup and conditions, providing a coherent narrative of the findings. This facilitates a comprehensive understanding of the implications, aiding in decision-making and further refining of simulation parameters.
	
	\paragraph{Advancing FEM Analysis Through Semantic Insights}
	Incorporating intelligent result interpretation into FEM significantly advances analysis capabilities. By providing semantic insights into simulation outcomes, engineers gain a deeper understanding of the phenomena under study, enabling more accurate and insightful analyses that drive innovation and improve design processes.
	
	\subsubsection{Automated Report Generation}
	
	\paragraph{Streamlining Documentation through Generative AI}
	Automated Report Generation, leveraging LLMs, automates the documentation of FEM simulations, transforming raw data into structured, comprehensible reports. This not only enhances communication among stakeholders but also streamlines the reporting process, allowing engineers to focus on analysis rather than documentation.
	
	\paragraph{Operational Framework for Generating Insights-Driven Reports}
	The operational framework for generating insights-driven reports through LLMs involves analyzing simulation data to extract key insights and trends. These are then structured into comprehensive reports that communicate findings effectively, enhancing the decision-making process and facilitating continuous improvement in FEM applications.
	
	\paragraph{Enhancing FEM Applications with Comprehensive Reporting}
	The integration of automated report generation into FEM applications enhances the utility and applicability of simulations. By providing a streamlined pathway for reporting and communication, stakeholders can quickly grasp the implications of results, leading to rapid iteration and refinement of designs and analyses.
	
	\subsubsection{Predictive Maintenance Recommendations}
	
	\paragraph{Forecasting Maintenance Needs with AI Insights}
	Predictive Maintenance Recommendations, utilizing LLM insights in conjunction with FEM, enable the forecasting of maintenance needs, allowing for proactive interventions. This reduces downtime and extends system lifespan, showcasing the potential of integrating AI with traditional engineering simulations to enhance operational reliability and efficiency.
	
	\paragraph{Operationalizing Predictive Analytics in Maintenance Planning}
	Operationalizing predictive analytics for maintenance planning involves using LLMs to analyze FEM outputs and operational data, identifying patterns indicative of potential failures. This enables the scheduling of targeted maintenance activities, optimizing resource allocation and minimizing downtime.
	
	\paragraph{Enhancing System Reliability and Performance}
	Incorporating Predictive Maintenance Recommendations into FEM workflows enhances system reliability and performance. By enabling proactive maintenance planning based on AI-driven insights, the longevity and efficiency of systems are improved, demonstrating the transformative impact of Algogenic enhancements on engineering practices.

	\subsubsection{Pseudocode for Algogenic FEM}
	The Algogenic Finite Element Method (FEM) approach integrates AI to enhance conventional FEM techniques by dynamically adjusting parameters and strategies based on the system's behavior and real-time error estimates. This pseudocode, depicted in \ref{fig:Algogenic-fem-pseudocode}, illustrates a sophisticated framework that incorporates AI-driven enhancements for adaptive element sizing, node selection, convergence criteria, and real-time parameter optimization.
	
	\begin{algorithm}
		\caption{Algogenic FEM Framework Pseudocode}
		\begin{algorithmic}[1]
			\Procedure{AlgogenicFEM}{DesignDomain, MaterialProperties}
			\State \textbf{Preprocessing:}
			\State Analyze DesignDomain with LLM for Semantic Geometry Analysis
			\State Interpret MaterialProperties using LLM for enhanced understanding
			
			\State \textbf{Mesh Generation:}
			\State Generate initial mesh based on LLM insights
			\While{Mesh not optimized}
			\State Apply Dynamic Mesh Refinement Guidance using LLM recommendations
			\State Adjust mesh based on Semantic Boundary Condition Application
			\State \textit{Check if mesh is optimized}
			\EndWhile
			
			\State \textbf{Simulation Execution:}
			\State Select solver with Adaptive Solver Selection using LLM
			\State Execute FEM analysis
			
			\State \textbf{Postprocessing:}
			\State Perform Intelligent Result Interpretation with LLM
			\State Generate Automated Report and Predictive Maintenance Recommendations using LLM
			\EndProcedure
		\end{algorithmic}\label{fig:Algogenic-fem-pseudocode}
	\end{algorithm}

	\begin{figure}
		\centering
		\includegraphics[width=0.8\textwidth]{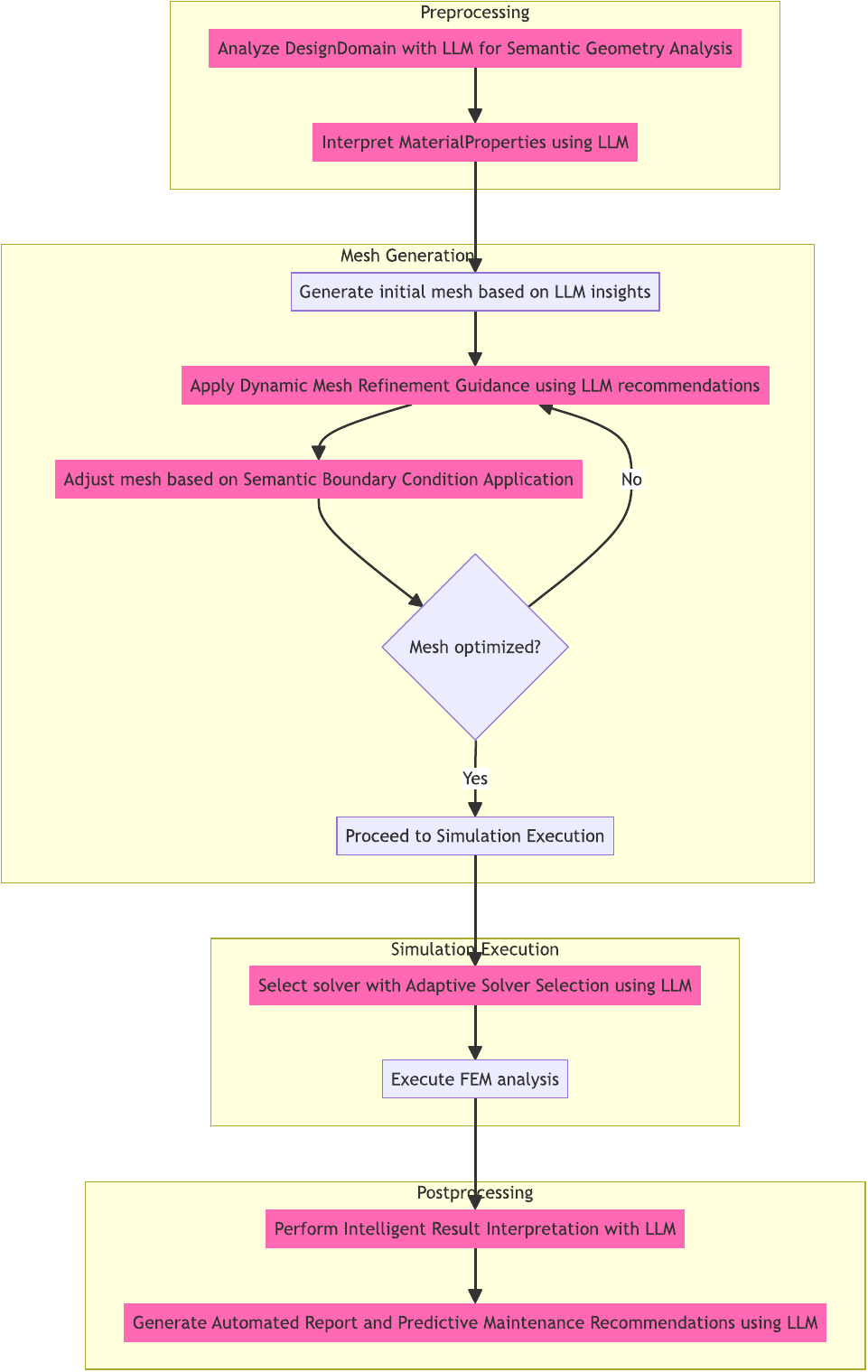} 
		\caption{Integrating Algogenic Enhancements in Finite Element Method (FEM): This diagram showcases the innovative integration of Algogenic enhancements with the Finite Element Method, leveraging Large Language Models for a comprehensive improvement of the FEM process. From preprocessing with semantic geometry analysis and material property interpretation, through dynamic mesh generation with LLM-guided refinement and boundary condition application, to simulation execution optimized by adaptive solver selection, and concluding with intelligent post-processing for result interpretation and maintenance recommendations. This holistic integration exemplifies the transformative potential of combining generative AI with traditional computational engineering methods, enhancing FEM's precision, efficiency, and application scope in tackling complex physical problems.}
		\label{fig:finite_elements}
	\end{figure}

	\section{Fast Fourier Transform}\index{Fast Fourier Transform}
	\subsection{Introduction to FFT}
	\subsubsection{The Concept of Fast Fourier Transform}
	\paragraph{Historical Context and Fundamental Idea}
	The Fast Fourier Transform stands as a hallmark in the realm of computational mathematics, offering a groundbreaking solution to the arduous computation of the Discrete Fourier Transform (DFT). Originally conceptualized by Gauss in the early 19th century, the FFT algorithm gained prominence in the latter half of the 20th century owing to its remarkable efficiency. By breaking down the DFT into smaller subproblems through a process of iterative decomposition, the FFT achieves a complexity of \(\mathcal{O}\left(N\log N\right)\), a monumental advancement from the traditional \(\mathcal{O}\left(N^2\right)\) complexity of direct DFT computation. This efficiency revolutionized numerous fields, particularly digital signal processing, where the ability to rapidly analyze frequency content became indispensable.
	
	Unlike the traditional DFT computation, which involves exhaustive pairwise multiplications and additions, the FFT leverages symmetries and periodicities inherent in sinusoidal functions to expedite the computation. Through techniques such as decimation in time or frequency, the FFT algorithm dramatically reduces the number of arithmetic operations required, making it suitable for real-time applications and large-scale datasets.
	
	Additionally, the FFT facilitates a deeper understanding of signals by providing a succinct representation of their frequency components. This decomposition into constituent frequencies enables various analyses, including filtering, modulation, and spectral estimation, thereby empowering researchers and engineers to extract valuable insights from complex data.
	
	The impact of the FFT extends beyond signal processing into diverse domains such as telecommunications, medical imaging, and finance, where rapid Fourier analysis plays a pivotal role in data interpretation and system optimization. Its ubiquity in modern technology underscores its enduring significance as a cornerstone algorithm in computational mathematics.

	\paragraph{The Core Principle of the FFT}
	At the heart of the FFT lies the fundamental understanding that the Discrete Fourier Transform (DFT) can be decomposed recursively into smaller DFTs. This concept leverages the inherent symmetries and periodicities embedded within the mathematical framework of the Fourier transform. By breaking down the DFT in this manner, the FFT employs a divide-and-conquer strategy, which dramatically diminishes the number of arithmetic operations necessary to compute the DFT.
	
	The Cooley-Tukey algorithm, the quintessential FFT method, exemplifies this principle. It initiates by partitioning the DFT of a sequence into two distinct sets: one comprising the even-indexed points and the other comprising the odd-indexed points. Subsequently, this partitioning process iterates recursively until the size of the DFTs becomes sufficiently small to be directly solvable. Once these smaller DFTs are computed, their outcomes are amalgamated to yield the final Fourier transform of the original sequence.
	
	This recursive subdivision approach capitalizes on the fact that smaller DFTs possess fewer computation requirements compared to larger ones. Consequently, through the systematic reduction of the DFT into progressively smaller components, the FFT attains a remarkable efficiency in computational complexity, rendering it indispensable in various applications requiring rapid Fourier transform calculations.
	
	Moreover, this recursive strategy not only accelerates the computation but also exploits the inherent structure of the DFT, thereby enhancing computational efficiency. This strategic decomposition, coupled with efficient merging of results, constitutes the cornerstone of the Fast Fourier Transform's prowess, making it an indispensable tool in diverse fields ranging from signal processing to scientific computing.

	\paragraph{Implications and Importance}
	The Fast Fourier Transform stands as a testament to the power of computational algorithms in revolutionizing various fields of science and engineering. Its efficiency in computing the Fourier transform, a fundamental operation in signal processing, has unleashed a cascade of advancements across diverse domains. In digital signal processing, the FFT serves as a linchpin, facilitating tasks ranging from denoising signals to precisely analyzing frequency components. By swiftly decomposing signals into their constituent frequencies, it enables engineers to extract meaningful information and make informed decisions.
	
	Moreover, the impact of FFT extends prominently into image processing. Here, its role in image compression is pivotal, enabling the storage and transmission of visual data with remarkable efficiency. Through techniques like JPEG compression, where the image is transformed into the frequency domain using FFT, redundant information is discarded, resulting in significant reductions in file size without perceptible loss in quality. Furthermore, in feature extraction, the FFT aids in identifying critical characteristics of images, essential in tasks such as pattern recognition and computer vision.
	
	The influence of FFT transcends traditional boundaries, finding application in numerical methods for solving partial differential equations. In engineering and physics, where phenomena are often modeled mathematically, the ability to numerically solve these equations is paramount. By efficiently computing Fourier transforms, the FFT plays a central role in iterative methods like spectral techniques, providing accurate solutions to complex problems in fluid dynamics, electromagnetism, and quantum mechanics.
	
	Beyond its foundational roles, the FFT serves as a cornerstone of modern computational science and engineering. Its versatility finds expression in telecommunications, where it underpins the modulation and demodulation of signals, ensuring reliable communication over vast distances. Additionally, in fields such as radar and financial analysis, the FFT enables rapid data processing, facilitating timely decision-making and risk management.
	
	In essence, the FFT's prowess in efficiently computing Fourier transforms has propelled advancements across a spectrum of applications, shaping the landscape of contemporary technology and scientific inquiry. Its ubiquity underscores its indispensability, rendering it not merely a tool but a catalyst for innovation and discovery.

	\subsubsection{Key Principles and Mechanisms}
	\paragraph{Divide and Conquer Strategy}
	The Fast Fourier Transform algorithm employs a divide and conquer strategy, a fundamental technique in algorithm design, to efficiently compute the Discrete Fourier Transform (DFT) of a sequence. By decomposing the original DFT problem into smaller subproblems, the FFT significantly reduces the computational complexity involved in computing the DFT. This strategy operates on the principle that a DFT of a sequence can be expressed as the sum of two smaller DFTs: one for the even-indexed elements and one for the odd-indexed elements of the sequence. Thus, the sequence is recursively divided into halves, with each half representing a smaller DFT problem. This recursive division continues until each subproblem becomes trivial, consisting of DFTs of sequences of length 1, which are themselves. 
	
	This recursive decomposition allows the FFT to exploit the inherent symmetry and periodicity properties of the DFT computation. By breaking down the problem into smaller, more manageable components, the FFT algorithm effectively reduces the number of arithmetic operations required to compute the DFT, leading to a substantial improvement in computational efficiency. Furthermore, the divide and conquer approach facilitates parallelization and efficient implementation on modern computing architectures, making the FFT a widely used and versatile algorithm in various signal processing and scientific computing applications.

	\paragraph{Exploitation of Symmetry and Periodicity}
	The FFT algorithm capitalizes on the inherent properties of symmetry and periodicity within the Discrete Fourier Transform (DFT), strategically reducing computational complexity. Through careful examination of the DFT, it becomes evident that certain symmetries exist, enabling the FFT to recycle previously computed results. This recycling process eliminates redundant computations, significantly enhancing efficiency. Moreover, the periodicity inherent in the sine and cosine functions utilized in the Fourier transform plays a pivotal role in the optimization achieved by the FFT. By leveraging the periodicity of these functions, the FFT algorithm orchestrates computations in a manner that efficiently manages the requisite multiplications and additions. It recognizes that many of the factors involved in these operations exhibit repetition, thus mitigating the overall number of unique multiplications required. Consequently, the FFT's ability to exploit symmetry and periodicity not only streamlines computational efforts but also optimizes resource utilization. This characteristic is particularly advantageous in scenarios where computational resources are constrained or where real-time processing is imperative. Therefore, the FFT algorithm's proficiency in leveraging these fundamental properties underscores its widespread adoption and enduring relevance in diverse applications across various domains.

	\paragraph{Complexity Reduction}
	The primary mechanism by which the FFT reduces the computational complexity of the DFT is through its algorithmic structure. The Cooley-Tukey algorithm, the most well-known FFT algorithm, effectively reduces the number of arithmetic operations from $\mathcal{O}(N^2)$ to $\mathcal{O}(N\log N)$, where $N$ is the number of points in the input signal.
	
	\textbf{Efficient Splitting:} 
	One key aspect of the Cooley-Tukey algorithm is its efficient splitting of the DFT into smaller DFTs. By recursively breaking down the original problem into smaller subproblems, each of which can be solved independently, the algorithm achieves significant computational savings. This recursive approach ensures that as the problem size increases, the increase in computational cost is only logarithmic rather than polynomial.
	
	\textbf{Doubling Efficiency:}
	Each split of the DFT into smaller DFTs effectively doubles the efficiency of the computation. This doubling arises from the fact that the algorithm exploits the inherent symmetries and periodicities present in the input signal. By reusing computations and leveraging symmetries, the algorithm avoids redundant operations, leading to a logarithmic reduction in the number of arithmetic operations required.
	
	\textbf{Logarithmic Reduction:}
	As a result of the efficient splitting and doubling of efficiency, the overall computational complexity of the FFT is drastically reduced to $\mathcal{O}(N\log N)$. This logarithmic reduction is a fundamental property of the Cooley-Tukey algorithm and is crucial for enabling fast computation of the DFT for large input signals.
	
	Therefore, through its algorithmic structure, the FFT achieves a remarkable reduction in computational complexity, making it a powerful tool for various signal processing applications.

	\paragraph{Algorithmic Variants and Optimizations}
	Several variants of the Fast Fourier Transform exist, each tailored to specific conditions to enhance its efficiency and applicability. The Cooley-Tukey algorithm stands out as particularly efficient for sequences with lengths that are powers of two. This method leverages the divide-and-conquer strategy, recursively breaking down the DFT computation into smaller sub-problems until reaching base cases, which are then efficiently computed. However, for sequences of arbitrary length, alternative approaches like the Prime Factor Algorithm (PFA) and Bluestein's FFT algorithm come into play. 
	
	The Prime Factor Algorithm capitalizes on the unique factorization of the sequence length to optimize the FFT computation. By decomposing the length into its prime factors, PFA strategically combines smaller DFTs to construct the final transform, yielding significant efficiency gains, especially for non-power-of-two lengths. Conversely, Bluestein's FFT algorithm tackles the challenge of arbitrary length sequences through a technique known as convolution via chirp-z transform. This method involves zero-padding the input sequence to a suitable length, enabling efficient convolution using FFTs.
	
	These optimizations ensure that the FFT remains versatile and robust, capable of handling diverse signal processing tasks with optimal efficiency. Moreover, advancements in hardware architectures have led to specialized FFT implementations, further enhancing its performance. The ubiquity of FFT-based algorithms across various domains underscores its indispensability in modern digital signal processing, scientific computing, and beyond, solidifying its status as a cornerstone algorithm in computational mathematics.

	\subsubsection{The Role of FFT in Signal Processing}
	\paragraph{Frequency Domain Analysis}
	The FFT (Fast Fourier Transform) is a cornerstone in signal processing, serving as a pivotal tool for converting signals from the time domain to the frequency domain. This transformation facilitates a profound understanding of signals by unveiling their frequency components. Unlike the time domain representation, where signals are depicted as amplitude versus time, the frequency domain showcases the signal's spectral content, revealing the relative strengths of different frequency components.
	
	Through FFT analysis, engineers and scientists can discern the dominant frequencies within a signal, providing crucial insights into its underlying characteristics. This capability proves invaluable in numerous applications, such as audio processing, telecommunications, and biomedical signal analysis. For instance, in audio processing, FFT aids in identifying specific frequency bands associated with various sounds or disturbances, enabling efficient noise reduction or equalization techniques.
	
	Moreover, FFT enables the removal of unwanted noise from signals through filtering operations in the frequency domain. By selectively attenuating certain frequency components, engineers can enhance signal clarity and fidelity. Additionally, FFT facilitates signal compression by representing signals in a more compact form based on their frequency components, enabling efficient storage and transmission.
	
	Analyzing signals in the frequency domain not only enhances signal processing capabilities but also enables the discovery of hidden insights. Complex phenomena, such as resonance or harmonic content, are often more discernible in the frequency domain, allowing for targeted analysis and optimization. Overall, the utilization of FFT and frequency domain analysis empowers engineers and scientists to extract meaningful information from signals, driving advancements across various fields.

	\paragraph{Signal Filtering and Noise Reduction}
	One of the primary applications of the FFT in signal processing is in the area of signal filtering and noise reduction. By transforming a signal into the frequency domain, it becomes possible to isolate and remove noise components or to enhance desired signals. This is achieved by applying various filters, such as low-pass, high-pass, band-pass, or band-stop filters, to the frequency domain representation of the signal. The filtered signal is then transformed back into the time domain using the inverse FFT (IFFT), resulting in a signal with improved clarity and reduced noise.
	
	The FFT provides a powerful tool for discerning the frequency components present in a signal. With this information, engineers and researchers can effectively design filters tailored to the specific characteristics of the signal and the noise to be removed. For instance, a low-pass filter allows only frequencies below a certain cutoff to pass through, effectively eliminating high-frequency noise. Conversely, a high-pass filter attenuates low-frequency signals while preserving higher frequencies, useful for extracting signals from a noisy background.
	
	Moreover, the flexibility of FFT-based filtering extends beyond simple frequency separation. Advanced techniques such as adaptive filtering can dynamically adjust filter parameters based on the signal's characteristics, offering improved noise reduction without distorting the desired signal. Additionally, the FFT facilitates real-time processing, allowing for rapid analysis and filtering of streaming data in applications ranging from audio processing to telecommunications.
	
	Furthermore, the inverse FFT plays a crucial role in reconstructing the filtered signal in the time domain. By converting the filtered signal from the frequency domain back to its original time-domain representation, the IFFT ensures that the processed signal retains its temporal integrity while benefiting from noise reduction. This seamless transition between domains enables engineers to apply sophisticated filtering algorithms without sacrificing temporal accuracy, ensuring optimal signal fidelity.
	
	In conclusion, the FFT's application in signal filtering and noise reduction is indispensable in modern signal processing. Its ability to transform signals between the time and frequency domains, coupled with the versatility of filter designs and real-time processing capabilities, empowers engineers to extract meaningful information from noisy data and enhance the quality of signals across various domains.

	\paragraph{Data Compression and Efficient Storage}
	The FFT (Fast Fourier Transform) is pivotal in the realms of data compression and efficient storage of signals. By leveraging the FFT to convert signals into the frequency domain, redundant or non-essential components can be readily identified and subsequently eliminated. This process is fundamental in optimizing the storage of signals by minimizing the space required for their representation. For instance, in multimedia applications like image compression (e.g., JPEG) and audio/video compression (e.g., MPEG), the FFT serves as a cornerstone technology. Through its application, these compression algorithms can significantly reduce the volume of data needed to faithfully reconstruct high-fidelity signals.
	
	Moreover, the utilization of FFT facilitates the extraction of essential signal components while discarding unnecessary information. This selective removal of redundancies not only conserves storage space but also enhances transmission efficiency. The frequency domain representation obtained through FFT enables the identification and quantification of signal characteristics, allowing for targeted compression strategies. Consequently, data can be compactly stored and transmitted without compromising perceptual quality.
	
	Furthermore, the efficiency gains achieved through FFT-based compression extend beyond mere storage considerations. In scenarios where bandwidth is limited, such as streaming services or wireless communication, the reduced data size resulting from FFT-based compression translates directly to improved transmission rates and lower resource utilization. Thus, FFT-based compression techniques not only optimize storage but also enhance data transfer capabilities, making them indispensable in modern multimedia systems.
	
	In summary, the integration of FFT in data compression and storage workflows revolutionizes signal processing by enabling the extraction of pertinent information while discarding redundancies. This streamlined approach not only conserves storage resources but also enhances transmission efficiency, making FFT a cornerstone technology in multimedia applications.

	\paragraph{Spectral Analysis and Characterization}
	Another significant application of the FFT is in spectral analysis, where it is used to characterize the spectral content of signals. This includes identifying the frequency components present in a signal, measuring the amplitude or phase of these components, and understanding the signal's behavior over time. Spectral analysis is essential in various fields, including telecommunications, where it aids in the design and optimization of communication systems, and in astronomy, where it helps in the analysis of light from celestial objects to determine their composition and motion.
	
	Furthermore, spectral analysis plays a crucial role in fields such as audio signal processing, where it enables the extraction of meaningful information from audio signals. By decomposing a complex audio signal into its frequency components, engineers can analyze and manipulate different aspects of the sound, such as pitch, timbre, and harmonics. This capability is exploited in applications ranging from music production and sound synthesis to speech recognition and noise cancellation.
	
	Moreover, in the field of medical imaging, spectral analysis techniques are utilized to extract diagnostic information from various types of scans, such as MRI and CT scans. By analyzing the frequency spectrum of the acquired signals, medical professionals can identify abnormalities, localize lesions, and assess tissue properties. This aids in the early detection and accurate diagnosis of diseases, leading to more effective treatment strategies and improved patient outcomes.
	
	Additionally, spectral analysis is employed in environmental monitoring and geophysics to study natural phenomena such as seismic waves, ocean currents, and atmospheric disturbances. By analyzing the frequency content of signals recorded by sensors and instruments, researchers can infer valuable information about the underlying processes and dynamics of the Earth's systems. This knowledge is instrumental in understanding and predicting natural hazards, mitigating environmental risks, and informing policy decisions related to climate change and resource management.

	\paragraph{Enhancement of Modern Technologies}
	The Fast Fourier Transform stands as a pivotal tool driving advancements across various fields due to its unparalleled efficiency and adaptability. Its significance reverberates through domains as diverse as medical imaging and seismology, where its transformative capabilities have revolutionized technological landscapes.
	
	In medical imaging, particularly MRI (Magnetic Resonance Imaging) and CT (Computed Tomography) scans, the FFT plays an integral role in extracting precise information from raw data. By swiftly converting signals into the frequency domain, it facilitates the creation of detailed anatomical images crucial for diagnosis and treatment planning. Moreover, the FFT's speed is paramount in these applications, as it enables real-time processing essential for rapid medical interventions.
	
	The realm of seismology benefits immensely from the FFT's prowess in signal analysis. By analyzing seismic data, scientists can discern patterns and detect subtle variations in ground vibrations, aiding in the prediction and mitigation of seismic events. This predictive capability holds immense societal value, potentially saving lives and minimizing the impact of earthquakes on infrastructure and communities.
	
	Beyond these specific applications, the FFT's broader utility permeates modern technologies, underpinning advancements in telecommunications, audio processing, and radar systems, among others. Its ability to rapidly decompose signals into their constituent frequencies facilitates efficient data transmission, high-fidelity audio reproduction, and precise target detection.
	
	In essence, the FFT serves as a cornerstone of modern technological innovation, empowering scientists, engineers, and innovators to push the boundaries of possibility. Its versatility and efficiency continue to catalyze progress across a spectrum of disciplines, driving forward the digital age and shaping the future of technology.

	\subsubsection{Applications and Limitations}
	\paragraph{Wide-ranging Applications}
	The FFT's applications are vast and span across many fields, demonstrating its versatility and importance. In engineering, it is used for signal analysis, filtering, and system design. These applications are crucial in fields such as telecommunications, where efficient signal processing is essential for transmitting and receiving data accurately and rapidly. For instance, in wireless communications, FFT plays a pivotal role in OFDM (Orthogonal Frequency Division Multiplexing) systems, enabling simultaneous transmission of multiple data streams over a single channel with minimal interference. Moreover, in digital audio processing, FFT algorithms are employed for tasks like spectral analysis and equalization, ensuring high-quality sound reproduction in various audio devices. 
	
	In physics, FFT algorithms are indispensable for solving partial differential equations (PDEs) that describe complex physical phenomena. Whether it's simulating fluid dynamics, modeling electromagnetic wave propagation, or understanding quantum mechanics, FFT techniques facilitate numerical solutions to these intricate equations, allowing scientists and engineers to predict and analyze the behavior of physical systems accurately. 
	
	In finance, particularly in the realm of option pricing models, FFT plays a crucial role in evaluating the characteristic functions of stochastic processes. These models heavily rely on Fourier analysis to estimate future asset prices, calculate option values, and manage risk effectively. By applying FFT algorithms, financial analysts can derive insights into market trends, volatility, and pricing dynamics, aiding in informed decision-making and portfolio management.
	
	Furthermore, in the medical field, FFT finds extensive use in image processing techniques applied to MRI and CT scans. These imaging modalities generate vast amounts of data that require efficient processing to produce clear and accurate diagnostic images. FFT algorithms enable tasks such as image reconstruction, noise reduction, and feature extraction, enhancing the quality and reliability of medical imaging diagnostics. Overall, the widespread adoption of FFT across diverse disciplines underscores its significance in advancing technology, science, and healthcare.

	\paragraph{Limitations and Challenges}
	Despite its widespread use and efficiency, the FFT also has limitations. The need for the length of the input data to be a power of two for optimal efficiency is a significant constraint, although algorithmic variations have been developed to address this. \textbf{Additionally}, the FFT can also be sensitive to numerical errors, especially in floating-point computations, which can accumulate and lead to inaccuracies in the final output. Another challenge is the FFT's susceptibility to "leakage" when analyzing finite signals that are not perfectly periodic within the sample window, leading to distortions in the frequency spectrum. Moreover, the FFT assumes a uniform sampling of the signal, which can be a limitation in applications where the signal is irregularly sampled or contains gaps. \textbf{Furthermore}, in scenarios where computational resources are limited, the computational complexity of the FFT algorithm may pose a challenge due to its $\mathcal{O}(n \log n)$ complexity, although this is generally efficient compared to alternative methods. \textbf{However}, it's important to note that despite these limitations, the FFT remains a versatile and powerful tool in various fields ranging from signal processing to scientific computing.

	\paragraph{Addressing the Limitations}
	Various strategies have been developed to mitigate the limitations of the Fast Fourier Transform. \textbf{Windowing techniques}, such as the Hamming or Blackman-Harris window, are frequently employed to reduce spectral leakage by tapering the signal in the time domain before computing the FFT. This helps alleviate the problem of spectral leakage, particularly when analyzing signals with discontinuities or sharp transitions. \textbf{Zero-padding} is another effective method utilized to enhance frequency resolution, especially when dealing with signals whose lengths are not powers of two. By appending zeros to the input signal, the FFT effectively interpolates between existing samples, resulting in a higher resolution frequency spectrum. Moreover, employing \textbf{advanced numerical algorithms} like the Cooley-Tukey algorithm or the FFTW library contributes to minimizing numerical errors inherent in FFT computations, thereby improving the accuracy of the frequency analysis. Additionally, utilizing \textbf{precision data types}, such as double-precision floating-point numbers, can further reduce the impact of round-off errors and enhance the fidelity of FFT results. Furthermore, \textbf{Non-uniform Fast Fourier Transform (NFFT)} techniques have emerged to address the challenge of analyzing signals sampled at non-uniform intervals. By adaptively adjusting the FFT algorithm to account for irregularly spaced samples, the NFFT offers a more accurate frequency analysis compared to traditional FFT methods when dealing with non-uniformly sampled signals.

	\paragraph{Future Directions}
	The ongoing research and development in the field of FFT and digital signal processing continue to push the boundaries of its applications and address its limitations. The integration of AI and machine learning techniques with FFT, for example, opens up new possibilities for adaptive signal processing, noise reduction, and feature extraction. This integration allows for the creation of intelligent systems capable of learning and adapting to different signal characteristics, thereby enhancing the efficiency and accuracy of signal processing tasks. Moreover, by leveraging machine learning algorithms, FFT-based systems can autonomously identify patterns in signals, leading to improved performance in tasks such as classification, prediction, and anomaly detection.
	
	As computational hardware evolves, so too does the potential for more complex and computationally intensive FFT applications. The advent of specialized hardware accelerators, such as GPUs and FPGAs, enables the implementation of FFT algorithms with higher throughput and lower latency, facilitating real-time processing of large-scale data in various scientific and industrial domains. Furthermore, advancements in parallel computing architectures allow for the efficient distribution of FFT computations across multiple processing units, leading to scalability and improved performance for handling massive datasets.
	
	In addition to enhancing traditional signal processing tasks, the integration of AI with FFT opens doors to novel applications and interdisciplinary research areas. For instance, in biomedical engineering, AI-powered FFT algorithms can aid in the analysis of medical signals, such as electrocardiograms (ECG) and electroencephalograms (EEG), facilitating early diagnosis of diseases and monitoring patient health. Similarly, in environmental monitoring, FFT-based systems augmented with AI can process sensor data to detect environmental changes, predict natural disasters, and optimize resource management strategies.
	
	The synergy between FFT, AI, and evolving computational hardware not only expands the scope of signal processing applications but also fosters innovation in diverse fields, driving advancements in technology and addressing societal challenges.

	\subsubsection{Pseudocode for Algorithmic FFT}
	The Fast Fourier Transform Algorithm is a powerful technique used for efficiently computing the Discrete Fourier Transform (DFT) of a sequence. Unlike other methods, the FFT dramatically reduces the computational complexity from \(O(N^2)\) to \(O(N\log N)\), where \(N\) is the length of the sequence. The algorithm's operation is detailed in pseudocode \ref{fig:fft-pseudocode}, which begins by checking if the sequence length is 1, in which case it simply returns the sequence since the DFT of a single element is itself. For longer sequences, the FFT recursively applies itself to the even and odd indexed elements, effectively breaking down the problem into smaller parts. It then combines these results using complex exponentials, representing rotations of the odd elements in the complex plane before merging them with the even elements. This step is crucial for capturing the essence of the Fourier transform, which involves summing sinusoids of different frequencies. Through iterative application and efficient combination, the FFT achieves significant computational savings, making it indispensable in various signal processing and data analysis tasks.
	
	\begin{algorithm}
		\caption{Cooley-Tukey FFT Algorithm Pseudocode}
		\begin{algorithmic}[1]
			\Procedure{FFT}{sequence}
			\State $N \gets \text{length of } sequence$
			\If{$N = 1$}
			\State \Return sequence
			\EndIf
			\State Even $\gets$ FFT(even-indexed elements of sequence)
			\State Odd $\gets$ FFT(odd-indexed elements of sequence)
			\State Create a complex array $T$ of size $N/2$ for temporary storage
			\For{$k = 0$ to $N/2 - 1$}
			\State $T[k] \gets \exp\left(-2\pi i \cdot \frac{k}{N}\right) \cdot \text{Odd}[k]$
			\State $\text{sequence}[k] \gets \text{Even}[k] + T[k]$
			\State $\text{sequence}[k + N/2] \gets \text{Even}[k] - T[k]$
			\EndFor
			\State \Return sequence
			\EndProcedure
		\end{algorithmic}\label{fig:fft-pseudocode}
	\end{algorithm}

\subsection{Previous Work on ML and AI Interplay with the Fourier Transform}

\paragraph{Fourier Transform Approach to Machine Learning I: Fourier Regression}
The exploration of Fourier Transform in the context of Machine Learning, specifically through Fourier Regression, provides insights into integrating the Fourier Transform into machine learning algorithms \cite{mehrabkhani2019fourier}. This study investigates how the Fourier Transform can be incorporated into regression problems to potentially enhance performance. By leveraging the Fourier Transform, the paper presents a method for regression tasks. The approach utilizes the Fourier Transform's capability to decompose functions into frequency components, offering an alternative approach to data analysis. This method introduces a new tool for machine learning practitioners and suggests avenues for further research on Fourier Transform applications in machine learning. The paper's contributions include presenting a technique for handling regression problems using the Fourier Transform and laying the groundwork for future exploration of their synergy.

\paragraph{Approximate Fast Fourier Transform-based Preprocessing for Edge AI}
The integration of Approximate Fast Fourier Transform in preprocessing for Edge AI addresses optimization challenges in machine learning algorithms for edge computing \cite{krupp2022approximate}. This paper introduces a preprocessing methodology that utilizes Approximate FFT to reduce computational load on edge devices. The strategy involves approximating FFT computations to balance accuracy and computational efficiency, enabling faster data processing on edge devices. This innovation enhances the performance of edge AI applications and extends their applicability in real-world scenarios with limited computational resources. By addressing challenges in edge device processing, this approach facilitates the deployment of AI applications in resource-constrained environments.

\subsection{Algogenic Enhancements for FFT}

\subsubsection{Semantic Signal Classification}

\paragraph{Enhancing FFT Through Intelligent Signal Categorization}
Applying Algogenic enhancements specifically to FFT, the process of Semantic Signal Classification can significantly refine the algorithm's capacity to analyze and interpret signals with precision. By integrating LLMs to meticulously categorize signals into semantic clusters such as "audio," "vibrational," or "electromagnetic," the FFT can be tailored to address each signal type with enhanced precision and efficiency. This customization facilitates the selection of optimal preprocessing techniques, windowing functions, and resolution parameters specific to the signal's nature, thereby optimizing the FFT's performance for specific applications. This approach not only boosts the accuracy and relevance of FFT outcomes but also elevates the adaptability of the algorithm to diverse signal types, thus broadening its applicability across various real-world scenarios.

\paragraph{Operationalizing Semantic Insights for Preprocessing}
Incorporating Semantic Signal Classification within the FFT workflow fundamentally transforms its preprocessing phase. By leveraging LLMs to extract semantic tags and discern pivotal signal attributes, the FFT setup becomes dynamically adaptive, significantly streamlining the preprocessing stage. This integration reduces manual intervention and enhances objectivity, ensuring that FFT parameters are finely tuned according to the semantic nuances of the signal. The automation of semantic classification and the subsequent adaptive adjustment of FFT parameters exemplify a significant leap in signal processing methodologies, enhancing both the efficiency and the accuracy of the FFT analysis across diverse applications.

\paragraph{Implications for FFT Analysis and Application}
Embedding Semantic Signal Classification into FFT analysis innovatively enhances the algorithm's efficiency and applicability across a spectrum of domains. By enabling FFT to analyze signals in a contextually relevant manner, this enhancement not only improves the quality of analysis in conventional applications but also facilitates its extension to new areas requiring nuanced analytical approaches. This Algogenic enhancement signifies a strategic evolution in FFT methodologies, fostering a more intelligent, adaptable, and application-specific analysis process.

\subsubsection{Automated Preprocessing Recommendations}

\paragraph{Optimizing Signal Preparation with LLM Insights}
The integration of Automated Preprocessing Recommendations into the FFT workflow revolutionizes signal preparation through LLM-driven insights. This Algogenic enhancement automates the selection of preprocessing methods, ensuring the signal is optimally conditioned for FFT analysis. By identifying effective preprocessing techniques tailored to each signal's semantic characteristics, LLMs facilitate a significant improvement in FFT's efficiency and accuracy, paving the way for enhanced frequency domain analysis in telecommunications, audio processing, and beyond.

\paragraph{Implementing Intelligent Preprocessing Pathways}
Intelligent Preprocessing Pathways, informed by LLM analysis, optimize FFT by tailoring the preprocessing steps to the signal's unique attributes. This process involves a detailed evaluation of the signal, leading to a precise application of the recommended window function and ensuring an effective mitigation of potential distortions during FFT computation. The adaptability of this approach ensures that each analysis benefits from preprocessing that is closely aligned with the signal's specific characteristics, enhancing the FFT's accuracy and relevance.

\paragraph{Enhancing FFT Through Precision Preprocessing}
Adopting Automated Preprocessing Recommendations within the FFT framework significantly broadens its practical utility. By ensuring optimal signal conditioning, this Algogenic enhancement not only elevates the quality and relevance of FFT analyses but also extends its applicability to a diverse range of fields. Automated Preprocessing Recommendations represent a pivotal advancement in signal processing, offering a path toward more automated, efficient, and insightful analyses in the digital era.

\subsubsection{Adaptive Window Function Selection}

\paragraph{Optimizing FFT with Context-Sensitive Windowing}
Utilizing Adaptive Window Function Selection, FFT's ability to mitigate spectral leakage and enhance frequency resolution is significantly improved through the use of LLM insights. This process dynamically selects the most suitable window function for each signal, optimizing the balance between resolution and leakage based on the signal's semantic categorization. This innovation not only refines FFT's spectral analysis capabilities but also enhances its adaptability to diverse signal types and conditions, promoting a more efficient and accurate analysis process.

\paragraph{Implementing LLM Recommendations in FFT Processing}
The incorporation of LLM recommendations for Adaptive Window Function Selection into FFT processing customizes the algorithm's approach to each signal. By meticulously choosing an optimal window function that aligns with the signal's characteristics, FFT preprocessing becomes highly targeted, enhancing the signal's representation and the accuracy of the subsequent analysis. This tailored approach exemplifies a significant enhancement in FFT processing, fostering a more intelligent and responsive signal analysis methodology.

\paragraph{Enhancing FFT Flexibility and Accuracy Across Applications}
Integrating Adaptive Window Function Selection into FFT workflows significantly enhances the algorithm's analytical capabilities. This Algogenic enhancement not only improves the precision of spectral analysis but also expands FFT's utility across various domains. By enabling FFT to adaptively select window functions based on semantic insights, this approach marks a substantial advancement in signal processing technologies, driving forward the capabilities of FFT in the data-driven analysis era.

\subsubsection{Dynamic Resolution Adjustment}

\paragraph{Optimizing Spectral Resolution for Enhanced Analysis}
Dynamic Resolution Adjustment in FFT, powered by LLM insights, allows for the adaptive tuning of frequency resolution to meet the analysis's specific needs. This enhancement enables FFT to dynamically adjust its parameters, optimizing the resolution to capture essential frequency details accurately or to provide a broader overview as required. This flexibility significantly enhances FFT's applicability and efficiency across various fields, enabling a more meaningful interpretation of signals.

\paragraph{Implementing Adaptive FFT Parameter Selection}
The process of Dynamic Resolution Adjustment entails LLM-driven analysis to dynamically alter FFT parameters, enhancing spectral analysis's precision and efficiency. By tailoring the FFT setup to the signal's requirements, this approach ensures that the analysis remains optimized, facilitating a more accurate and responsive examination of spectral data. This adaptive strategy underscores the importance of flexibility in spectral analysis, leveraging LLM insights to maximize FFT's utility across diverse applications.

\paragraph{Enhancing FFT Flexibility and Precision}
Incorporating Dynamic Resolution Adjustment into FFT processes significantly enhances the algorithm's adaptability and precision. This Algogenic enhancement ensures that FFT analyses are always optimized for the task at hand, improving the quality of outcomes across diverse fields. By making FFT more responsive to the nuances of each signal, Dynamic Resolution Adjustment exemplifies the integration of LLM insights into classical algorithms, leading to significant advancements in signal processing.

\subsubsection{Intelligent Zero-padding Guidance}

\paragraph{Optimizing Frequency Resolution with AI-Driven Strategies}
Intelligent Zero-padding Guidance, utilizing LLM insights, optimizes FFT's application of zero-padding to enhance frequency resolution and spectral analysis clarity. By intelligently guiding the zero-padding process, this enhancement ensures optimal resolution across different frequency ranges, facilitating a more accurate interpretation of spectral components. This AI-driven approach marks a significant advancement in spectral analysis techniques, offering unprecedented insights into signals' underlying characteristics.

\paragraph{Implementing Zero-padding Based on Semantic Analysis}
Implementing Intelligent Zero-padding Guidance involves LLM-driven semantic analysis to recommend an optimal zero-padding strategy, enhancing FFT's spectral resolution and precision. By tailoring zero-padding to the signal's specific characteristics and analysis goals, this approach optimizes FFT's performance, ensuring a more detailed and accurate frequency domain representation. This integration exemplifies the value of semantic analysis in refining FFT methodologies, enabling a more nuanced and effective optimization of spectral analysis.

\paragraph{Enhancing FFT Flexibility and Precision}
The integration of Intelligent Zero-padding Guidance into FFT workflows significantly enhances the algorithm's flexibility and precision. By enabling dynamic adjustment of zero-padding based on LLM insights, this enhancement facilitates a more refined spectral analysis, improving the resolution and mitigating leakage and aliasing effects. This advancement not only improves FFT's analytical capabilities but also expands its applicability, fostering a more intelligent and adaptable approach to signal processing.

\subsubsection{Adaptive Algorithmic Pathways with LLM Insights}

\paragraph{Enhancing FFT Through Intelligent Pathway Selection}
Incorporating LLMs into FFT analysis facilitates intelligent selection of algorithmic pathways, optimizing performance based on the signal's characteristics and computational environment. By analyzing the signal and leveraging LLM insights, FFT can adaptively choose the most suitable algorithm, enhancing computational efficiency and accuracy. This approach not only signifies a leap in FFT methodologies but also broadens its application across diverse fields, enabling a more versatile and effective analysis.

\paragraph{Customizing FFT Execution Based on Signal and System Analysis}
Customizing FFT execution through LLM insights involves a nuanced examination of signal properties and computational resources, enabling dynamic adaptation of the FFT's strategy. This adaptive approach optimizes FFT's performance, ensuring the algorithm is tailored to the specific demands of each application. By enabling real-time adjustments and leveraging computational resources efficiently, this methodology enhances FFT's precision and responsiveness, fostering advancements in signal processing.

\paragraph{Implementing Real-Time Algogenic Adjustments}
Integrating real-time Algogenic adjustments into FFT, informed by LLM analysis, enables the algorithm to adapt its strategy dynamically, ensuring optimal performance. This approach not only enhances FFT's applicability in dynamic environments but also optimizes computational efficiency. By incorporating continuous LLM feedback, FFT can effectively handle diverse signal inputs, demonstrating the value of adaptability in modern signal processing applications.

\subsubsection{Intelligent Error Correction and Precision Enhancement with LLM}

\paragraph{Predicting and Mitigating Computational Errors}
LLMs can enhance FFT by predicting and mitigating computational errors, ensuring more reliable and accurate results. By analyzing known error patterns, LLMs can adjust computation parameters preemptively, addressing potential errors before they impact FFT's output. This proactive approach to error mitigation exemplifies a significant advancement in FFT methodologies, improving the reliability and efficiency of FFT-based computations across various applications.

\paragraph{Enhancing Precision Through Adaptive Techniques}
Incorporating adaptive precision techniques into FFT, guided by LLM insights, enables dynamic adjustment of computational precision, improving the accuracy of spectral analysis. This approach optimizes resource allocation, ensuring each signal segment receives appropriate precision levels based on its characteristics. By enhancing precision adaptively, FFT analyses become more accurate and efficient, demonstrating the potential of LLM-guided techniques in advancing signal processing methodologies.

\paragraph{Utilizing Error Correction Codes}
Integrating error correction codes (ECC) into FFT, based on LLM recommendations, enhances the algorithm's accuracy by correcting computational errors. This approach provides a robust solution for mitigating errors introduced during computation, improving FFT's reliability and precision. By dynamically adjusting ECC levels, FFT maintains computational efficiency while ensuring high accuracy, showcasing the value of LLMs in refining FFT methodologies for diverse applications.

\subsubsection{Semantic-Driven Dynamic Resolution Adjustment with LLM}

\paragraph{Tailoring Frequency Resolution to Signal Content}
Dynamic resolution adjustment, informed by LLM-driven semantic analysis, enables FFT to tailor its frequency resolution based on the signal's content, optimizing the analysis for specific application requirements. By focusing computational resources on relevant frequency components, FFT achieves enhanced precision and efficiency in spectral analysis. This approach demonstrates the integration of semantic insights into FFT methodologies, fostering a more intelligent and adaptable analysis process.

\paragraph{Adapting Window Functions and FFT Length}
LLM insights guide the selection of window functions and adjustments to FFT length, optimizing spectral analysis based on the signal's characteristics. By tailoring window functions and FFT length, FFT achieves a balance between resolution and leakage, enhancing the accuracy and relevance of spectral analysis. This methodology exemplifies the potential of LLMs in refining FFT processes, advancing signal processing techniques for diverse applications.

\paragraph{Semantic Prioritization in Spectral Analysis}
Integrating semantic priorities into FFT analysis enables a more focused examination of relevant spectral features, enhancing the interpretability and applicability of FFT results. By leveraging LLM insights, FFT can adapt its analysis to emphasize critical frequency components, facilitating a deeper understanding of the underlying signal dynamics. This approach underscores the value of semantic analysis in enhancing FFT methodologies, fostering more meaningful and actionable insights.

\subsubsection{Optimizing FFT Algorithms with LLM Insights}

\paragraph{Algorithm Selection Based on Signal Characteristics}
Utilizing LLM insights for FFT algorithm selection optimizes the analysis based on signal characteristics such as sparsity and noise level. By tailoring the choice of algorithm, FFT achieves enhanced computational efficiency and accuracy, demonstrating the potential of integrating LLM insights into signal processing methodologies. This approach not only improves FFT performance but also extends its applicability across various signal types and computational environments.

\paragraph{Adaptive Algorithmic Complexity}
Incorporating LLM insights into FFT algorithms enables adaptive management of algorithmic complexity, optimizing computational resources based on the signal's characteristics and the analysis requirements. By dynamically adjusting algorithmic parameters, FFT maintains optimal performance, enhancing the efficiency and accuracy of spectral analysis. This methodology exemplifies the integration of LLM-driven insights into FFT processes, advancing signal processing techniques for diverse applications.

\paragraph{LLM-Guided Parameter Tuning}
LLM-guided parameter tuning enhances FFT by optimizing parameters such as twiddle factors based on signal characteristics and computational constraints. This approach ensures FFT's computational efficiency and accuracy, demonstrating the potential of LLM insights in refining FFT methodologies. By dynamically adjusting parameters, FFT achieves enhanced performance, fostering advancements in signal processing across various domains.

\subsubsection{Dynamic Parameter Adjustment within FFT}

\paragraph{FFT Length and Resolution}
Dynamic adjustment of FFT length, informed by LLM insights, optimizes frequency resolution based on the analysis goals, enhancing the precision and relevance of spectral analysis. By tailoring FFT length, the analysis achieves a balance between computational efficiency and resolution requirements, demonstrating the value of integrating LLM-driven insights into FFT methodologies. This approach advances signal processing techniques, enabling more accurate and efficient spectral analysis across diverse applications.

\paragraph{Window Function Application}
Adapting window functions based on LLM insights enhances FFT's ability to mitigate spectral leakage, improving the accuracy of spectral analysis. By selecting window functions tailored to the signal's characteristics, FFT achieves optimal balance between resolution and leakage, demonstrating the potential of LLM-driven insights in refining FFT methodologies. This approach fosters advancements in spectral analysis, enhancing FFT's applicability across various signal processing tasks.

\paragraph{Adaptive Zero-padding for Frequency Bin Precision}
Implementing adaptive zero-padding strategies, guided by LLM insights, enhances FFT's frequency bin precision, improving the interpretability of spectral analysis. By optimizing zero-padding based on the analysis goals, FFT achieves enhanced precision in frequency domain representation, demonstrating the potential of LLM-driven insights in advancing FFT methodologies. This approach fosters more accurate and efficient spectral analysis, enhancing FFT's utility across diverse applications.

\subsubsection{Enhanced Spectral Analysis using LLMs}

\paragraph{Semantic Interpretation of Spectral Features}
Integrating LLM-driven semantic interpretation into FFT analysis enhances the understanding of spectral features, linking them to real-world phenomena and facilitating actionable insights. By providing a semantic context for spectral data, FFT analysis becomes more accessible and relevant, demonstrating the potential of combining LLM insights with signal processing techniques. This approach fosters advancements in various domains, enhancing the interpretability and applicability of FFT results.

\paragraph{Real-time FFT Adjustment Based on Intermediate Results}
Incorporating real-time FFT adjustments based on intermediate results, informed by LLM analysis, enhances the adaptability and precision of spectral analysis. By dynamically adjusting FFT parameters in response to emerging trends, FFT achieves enhanced accuracy and efficiency, demonstrating the potential of integrating LLM insights into FFT methodologies. This approach fosters a more responsive and intelligent signal analysis process, advancing FFT's utility across diverse applications.

\paragraph{Linking Spectral Analysis to Semantic Content}
The integration of LLM insights into FFT analysis bridges the gap between spectral data and semantic understanding, enhancing the interpretability and relevance of FFT results. By linking spectral analysis to real-world phenomena, FFT becomes a more powerful tool for extracting actionable insights from complex signals, demonstrating the potential of combining LLM-driven insights with signal processing techniques. This approach fosters advancements in various domains, enhancing FFT's applicability and impact.

\subsubsection{Semantic Interpretation of FFT Results}

\paragraph{Translating Frequency Domain Insights into Understandable Concepts}
The Semantic Interpretation of FFT Results, leveraging LLM insights, transforms complex frequency domain data into accessible, actionable insights. By providing a clear interpretation of FFT outcomes, this enhancement enables a broader understanding of spectral analysis, fostering informed decision-making across diverse fields. This approach demonstrates the potential of integrating LLM insights into FFT methodologies, enhancing the accessibility and utility of FFT results.

\paragraph{Operationalizing LLMs for Enhanced Result Interpretation}
Implementing Semantic Interpretation of FFT Results through LLM-driven analysis enhances the comprehensibility and relevance of spectral analysis. By generating detailed narratives contextualizing FFT findings, LLMs facilitate a deeper understanding of signal dynamics, fostering informed decision-making. This approach exemplifies the integration of LLM insights into FFT methodologies, advancing signal processing techniques and enhancing FFT's utility across various domains.

\paragraph{Advancing FFT Applications through Intelligent Analysis}
Incorporating Semantic Interpretation of FFT Results into FFT applications enhances the algorithm's analytical capabilities, broadening its applicability and impact. By providing clear, contextualized insights into spectral data, this enhancement fosters innovation and optimization across diverse fields, demonstrating the potential of integrating LLM insights into FFT methodologies. This approach advances signal processing techniques, enabling more informed and strategic decision-making based on FFT analysis.

\subsubsection{Automated Report Generation}

\paragraph{Streamlining FFT Analysis Documentation}
Automated Report Generation, utilizing LLM insights, automates the documentation of FFT analysis, enhancing the efficiency and comprehensibility of reports. By generating detailed, insightful reports based on FFT results, LLMs facilitate a quicker understanding of spectral analysis, fostering informed decision-making across diverse applications. This approach demonstrates the potential of integrating LLM insights into FFT methodologies, streamlining the analysis process and enhancing the accessibility of FFT results.

\paragraph{Operationalizing Insightful Communication}
Implementing Automated Report Generation within the FFT workflow, leveraging LLM-driven insights, enhances the documentation and communication of FFT analysis. By providing detailed narratives and contextual explanations of FFT results, LLMs foster a deeper understanding of spectral data, facilitating informed decision-making. This approach exemplifies the integration of LLM insights into FFT methodologies, advancing signal processing techniques and enhancing FFT's utility across various domains.

\paragraph{Enhancing Accessibility and Impact of FFT Analyses}
The integration of Automated Report Generation into FFT workflows significantly enhances the accessibility and impact of FFT analyses. By providing clear, concise summaries of complex data, this enhancement enables a broader range of stakeholders to engage with and benefit from FFT results. This approach fosters innovation and collaboration across diverse fields, demonstrating the potential of integrating LLM insights into FFT methodologies and advancing the utility of FFT in various applications.

\subsubsection{Predictive Analysis for Signal Evolution}

\paragraph{Forecasting Signal Changes with LLM-Driven Insights}
Predictive Analysis for Signal Evolution, utilizing LLM insights, enhances FFT by forecasting future signal changes, providing a forward-looking perspective on signal dynamics. This approach enables proactive decision-making based on anticipated trends, demonstrating the potential of integrating LLM-driven insights into FFT methodologies. By enhancing FFT's predictive capabilities, this enhancement fosters advancements in various domains, enabling more strategic responses to evolving signals.

\paragraph{Operationalizing Predictive Signals Analysis}
Implementing Predictive Analysis for Signal Evolution within the FFT framework, leveraging LLM insights, enhances signal processing by identifying patterns indicative of future changes. This approach enables a proactive analysis of signal evolution, fostering informed decision-making and strategic planning. By integrating LLM-driven insights into FFT methodologies, this enhancement advances signal processing techniques, fostering a more adaptive and forward-looking approach to analyzing complex signals.

\paragraph{Enhancing Signal Processing Applications}
Incorporating Predictive Analysis for Signal Evolution into FFT applications significantly enhances the utility and effectiveness of signal processing efforts. By enabling the anticipation and response to signal changes, this enhancement fosters innovation and optimization across diverse fields. This approach demonstrates the potential of combining LLM insights with FFT methodologies, advancing signal processing techniques and enabling more proactive and strategic decision-making based on FFT analysis.

	\subsubsection{Pseudocode for Algogenic FFT}
	The Algogenic Fast Fourier Transform approach integrates AI to enhance conventional FFT methods by adaptively modifying algorithmic parameters and strategies according to the system's behavior and real-time error estimates. This pseudocode, provided in \ref{fig:fft-Algogen-pseudocode}, illustrates a sophisticated framework integrating AI-driven improvements for dynamic frequency domain analysis, windowing techniques, data manipulation, and real-time parameter optimization.
	
	\begin{algorithm}
		\caption{Enhanced FFT with Algogenic Enhancements}
		\begin{algorithmic}[1]
			\Procedure{EnhancedFFT}{Signal}
			\State \textbf{Preprocessing:}
			\State Analyze signal characteristics and context with LLM for semantic classification
			\State Apply automated preprocessing recommendations based on LLM insights
			
			\State \textbf{Core FFT Execution:}
			\State Determine adaptive algorithmic pathways with LLM insights
			\State Implement intelligent error correction and precision enhancement using LLM
			\State Select an appropriate window function for the signal
			\State Apply zero-padding to improve FFT resolution and computational efficiency
			\State Adjust FFT resolution dynamically with semantic-driven insights from LLM
			\State Optimize computational resources based on the current environment with LLM guidance
			\State Perform the FFT, incorporating real-time feedback loops for iterative analysis with LLM
			\State Detect enhanced spectral features using LLM for deeper analysis
			
			\State \textbf{Postprocessing:}
			\State Interpret FFT results with advanced semantic analysis using LLM
			\State Generate automated reports, including predictive analysis for signal evolution, with LLM insights
			\EndProcedure
		\end{algorithmic}\label{fig:fft-Algogen-pseudocode}
	\end{algorithm}

	\section{Sparse Matrix Computations}\index{Sparse Matrix Computations}
	\subsection{Introduction to Sparse Matrix Computations}
	\subsubsection{The Concept of Sparse Matrices}
	\paragraph{Definition and Characteristics}
	Sparse matrices are distinguished by their significant number of zero-valued elements, setting them apart from dense matrices where most elements are non-zero. The primary characteristic of a sparse matrix is that the non-zero elements are dispersed throughout the matrix, which can be leveraged to optimize both storage and computational operations. This optimization is crucial in large-scale computations where the matrix dimensions can be very large, making the storage and manipulation of dense matrices impractical.
	
	Furthermore, the sparsity pattern of a matrix plays a vital role in determining the efficiency of various algorithms used for operations like multiplication, inversion, and decomposition. In contrast to dense matrices, where every element must be stored and processed, sparse matrices allow algorithms to exploit the presence of zeros, reducing memory requirements and computational complexity. 
	
	Moreover, sparse matrices often arise in practical applications such as finite element analysis, network analysis, and computational fluid dynamics, where the underlying systems exhibit inherent sparsity. In such domains, utilizing sparse matrix representations not only conserves memory but also accelerates computations, leading to significant performance gains.
	
	Therefore, understanding the properties and handling of sparse matrices is essential for efficiently solving large-scale computational problems. By exploiting sparsity, practitioners can develop algorithms and techniques tailored to the specific structure of sparse matrices, enabling faster and more memory-efficient computations across various domains.

	\paragraph{Storage Efficiency}
	The efficiency in storing sparse matrices comes from the fact that only the non-zero elements and their positions need to be stored. This selective storage approach significantly reduces memory requirements compared to dense matrix storage, where every element occupies space regardless of its value. \textit{Moreover}, various storage schemes, such as Compressed Sparse Row (CSR) or Compressed Sparse Column (CSC), are designed to minimize the memory footprint by efficiently encoding the position and value of these non-zero elements. \textit{Additionally}, these schemes often employ techniques like pointer compression and data structure optimizations to further enhance storage efficiency. This efficient storage is particularly beneficial \textit{for} applications like finite element analysis, where the matrices involved are large but contain very few non-zero elements relative to their size. \textit{Furthermore}, the efficiency gained in storage directly translates into computational benefits, as operations on sparse matrices can be performed more efficiently due to reduced memory access overhead. Thus, sparse matrix storage techniques not only conserve memory but also contribute to faster computations and overall improved performance in a wide range of scientific and engineering applications.

	\paragraph{Computational Advantages}
	From a computational standpoint, sparse matrices offer significant advantages in terms of efficiency and speed. By storing and manipulating only the non-zero elements, sparse matrix operations can be performed much more quickly compared to dense matrices. This optimization is particularly beneficial for algorithms heavily reliant on matrix operations, such as those involved in numerical methods and simulations.
	
	Sparse matrices excel in scenarios where computational resources are limited or where large-scale simulations are conducted. For instance, in finite element analysis or computational fluid dynamics, where matrices representing physical systems can be extremely large but predominantly sparse, the use of sparse matrix techniques drastically reduces memory consumption and computational time.
	
	Moreover, the efficiency gains extend beyond basic operations like matrix multiplication to more complex tasks such as matrix inversion and solving systems of linear equations. In these cases, the computational complexity is significantly reduced due to the sparsity of the matrices involved, leading to faster solution times and overall improved performance.
	
	Furthermore, the advantages of sparse matrices become even more pronounced when considering parallel and distributed computing environments. Sparse matrix algorithms can be parallelized effectively, allowing for efficient utilization of multi-core processors and distributed computing clusters. This scalability makes sparse matrix techniques indispensable in modern computational science and engineering applications.
	
	In conclusion, the computational benefits of sparse matrices, including faster execution times, reduced memory overhead, and improved scalability, make them a crucial tool in various scientific and engineering domains, where computational efficiency is paramount.

	\paragraph{Implications for Software and Hardware}
	The characteristics of sparse matrices have significant implications for both software and hardware design. On the software side, algorithms and data structures must be specifically designed to take advantage of sparsity. Utilizing efficient sparse matrix representations such as Compressed Sparse Row (CSR) or Compressed Sparse Column (CSC) becomes imperative for reducing memory overhead and computational complexity. \textit{Moreover}, specialized algorithms like iterative solvers (e.g., Conjugate Gradient, GMRES) or direct solvers (e.g., LU decomposition with sparse pivoting) are \textit{likewise} tailored to exploit the sparsity pattern efficiently, minimizing unnecessary operations on zero elements. \textit{Furthermore}, parallelization techniques, \textit{such as} task parallelism or data parallelism, can be employed to distribute computations effectively across multiple processors, enhancing overall performance. 
	
	\textit{On the other hand}, from a hardware perspective, memory access patterns play a crucial role in exploiting sparsity. Caches and memory hierarchies need to be optimized to minimize cache misses when accessing non-contiguous memory locations in sparse matrices. \textit{Consequently}, specialized memory architectures, like hybrid memory systems combining fast, low-capacity memory with slower, high-capacity memory, \textit{are} often \textit{employed} to accommodate the irregular access patterns inherent in sparse computations. 
	
	\textit{Additionally}, processing units need to be equipped with efficient instruction sets tailored for sparse matrix operations, allowing for streamlined execution of common sparse matrix algorithms. \textit{Furthermore}, hardware accelerators, such as Graphics Processing Units (GPUs) or Field-Programmable Gate Arrays (FPGAs), can be leveraged to offload sparse matrix computations from the CPU, exploiting their parallel processing capabilities to achieve significant speedups. \textit{In contrast}, traditional dense matrix operations, which benefit from regular memory access patterns and data locality, may not fully utilize the potential of such hardware accelerators.

	\subsubsection{Key Principles and Mechanisms}
	\paragraph{Fundamental Concepts of Sparsity}
	The core principle behind sparse matrix computations is the focus on the non-zero elements, which are the essence of the matrix's informational content. By concentrating computational and storage efforts on these elements, sparse matrix methodologies significantly reduce the resources required for matrix operations. This principle is fundamental in computational mathematics and computer science, especially in applications dealing with large datasets or matrices where the majority of elements are zero.
	
	Moreover, sparse matrix computations enable efficient handling of massive datasets by exploiting the inherent sparsity present in many real-world problems. In addition to reducing storage requirements, this approach accelerates computational processes, leading to faster algorithms for tasks such as solving linear systems, eigenvalue computations, and optimization problems. Furthermore, the sparsity-driven paradigm facilitates the development of specialized algorithms tailored to exploit specific structural properties of sparse matrices, enhancing performance even further.
	
	On the other hand, while sparse matrix techniques offer significant advantages in terms of resource efficiency and computational speed, they also introduce challenges in algorithm design and implementation. Unlike dense matrices, sparse matrices require specialized data structures and algorithms to handle efficiently. Consequently, developing robust and scalable sparse matrix libraries becomes crucial for practitioners working with sparse data representations. Nonetheless, the benefits of sparsity in terms of resource savings and computational efficiency make these challenges worthwhile to address.
	
	Therefore, understanding the fundamental concepts of sparsity and mastering the techniques for exploiting it in computational tasks are essential skills for researchers and practitioners in fields ranging from scientific computing to machine learning and data analysis.

	\paragraph{Storage Schemes and Their Impact}
	Various storage schemes have been developed to efficiently manage sparse matrices, each with its unique advantages and suited to particular types of sparsity patterns or computational requirements. The Compressed Sparse Row (CSR) format, for instance, is optimized for row-wise traversal and operations, making it ideal for certain linear algebra computations. The Coordinate List (COO) format, on the contrary, stores each non-zero element by its row and column indices, offering a straightforward representation that is particularly useful during the construction phase of a matrix. These storage schemes are critical in achieving the computational and storage efficiencies that sparse matrices can offer.
	
	Moreover, the CSR format facilitates efficient matrix-vector multiplication by storing the data in a way that aligns with row-wise traversal, reducing memory access overhead and improving cache utilization. Similarly, the COO format, while not as efficient for direct operations, provides a simple and intuitive representation during matrix assembly, making it favorable for dynamic or irregular matrices where structure may change frequently.
	
	Furthermore, the choice of storage scheme can significantly impact the performance of algorithms operating on sparse matrices. For example, algorithms relying heavily on matrix-vector products may benefit greatly from using CSR format due to its optimized structure for such operations. Conversely, algorithms involving frequent modifications to the matrix structure may find COO format more suitable due to its ease of manipulation during construction.
	
	In addition to CSR and COO, other storage formats like the Compressed Sparse Column (CSC) or the Ellpack-Itpack (ELL) offer alternative trade-offs between memory usage, computational efficiency, and ease of manipulation. Understanding the characteristics of each storage scheme and matching them to the specific requirements of the problem at hand is crucial for optimizing the performance of sparse matrix computations.

	\paragraph{Reducing Computational Overhead}
	The selective processing of non-zero elements inherent in sparse matrix operations leads to a substantial reduction in computational overhead. This reduction is not just in terms of the number of arithmetic operations but also in the improved cache utilization and reduced memory bandwidth requirements, which are crucial performance factors in modern computing architectures. 
	
	Sparse matrix operations, by virtue of focusing solely on non-zero elements, alleviate the burden on computational resources. Traditional dense matrix operations necessitate processing all elements, regardless of their value, resulting in redundant calculations and increased computational complexity. However, in sparse matrices, only non-zero elements are considered, significantly reducing the number of arithmetic operations required. This targeted approach not only speeds up computations but also optimizes resource utilization, leading to enhanced efficiency.
	
	Moreover, the efficiency gains extend beyond computational aspects. Sparse matrix operations tend to exhibit improved cache utilization due to their data sparsity. Since only non-zero elements are accessed and manipulated, there is a higher likelihood of data residing in the cache, thereby minimizing memory access latency. This optimized cache usage translates into faster data retrieval and manipulation, contributing to overall performance enhancement.
	
	Additionally, sparse matrix computations impose lighter demands on memory bandwidth. With fewer non-zero elements to process, there is a reduced need for data movement between the processor and memory. Consequently, the strain on memory bandwidth is alleviated, allowing for smoother data transfer and reducing potential bottlenecks in system performance.
	
	Efficient sparse matrix operations play a pivotal role in mitigating both time and energy consumption in computational tasks, particularly in high-performance computing environments. By focusing computational efforts solely on relevant data points, sparse matrices enable significant improvements in processing speed and resource utilization, ultimately leading to more efficient and sustainable computing practices.

	\paragraph{Optimization Techniques}
	Beyond storage and basic operations, optimization techniques play a crucial role in maximizing the efficiency of sparse matrix computations. These techniques encompass a spectrum of strategies aimed at mitigating computational bottlenecks and enhancing overall performance. Reordering algorithms, for instance, strategically rearrange the structure of sparse matrices to minimize fill-in during matrix factorizations. By intelligently reorganizing the matrix elements, these algorithms reduce the number of non-zero entries in the factors, thereby optimizing memory utilization and computational efficiency. Furthermore, partitioning methods are instrumental in facilitating parallel computations by decomposing the sparse matrix into smaller, manageable subproblems that can be solved concurrently. This parallelization significantly accelerates computation, especially for large-scale sparse systems where traditional sequential methods may become prohibitively time-consuming. Additionally, iterative solvers tailored for sparse systems offer efficient solutions by exploiting the specific characteristics of sparse matrices. Unlike direct solvers, which require dense factorizations and entail high memory overhead, iterative methods iteratively refine an initial guess towards the solution, making them particularly well-suited for large, sparse systems. These optimization techniques are underpinned by sophisticated theoretical frameworks that carefully balance the trade-offs between preprocessing time, memory usage, and computational speed, ensuring robust and efficient sparse matrix computations in diverse applications.

	\paragraph{Interplay with High-Performance Computing}
	The principles and mechanisms of sparse matrix computations are closely aligned with the goals of high-performance computing (HPC). Sparse matrices, characterized by their large proportion of zero elements, offer significant advantages in terms of memory utilization and computational efficiency. This sparsity property enables optimizations tailored to exploit the structure of sparse matrices, resulting in reduced storage requirements and accelerated computations. Moreover, the parallelism inherent in sparse matrix operations resonates with the parallel processing capabilities of modern HPC architectures.
	
	\textbf{Furthermore}, the optimization of storage and computations for sparse matrices directly contributes to the scalability and efficiency of HPC applications. \textbf{Moreover}, advancements in specialized hardware, such as GPUs and FPGAs, have been driven by the demand for accelerating sparse matrix operations within HPC environments. These hardware accelerators leverage the inherent parallelism of sparse matrix computations, enabling significant performance gains compared to traditional CPU-based approaches. 
	
	\textbf{On the other hand}, the development of efficient software libraries and algorithms tailored for sparse matrices is paramount in maximizing the benefits of sparse matrix computations in HPC. \textbf{In contrast}, dense matrix computations, which require storage and computation proportional to the square of the matrix dimension, pose significant challenges for memory bandwidth and computational resources in HPC systems. Thus, the adoption of sparse matrix techniques \textbf{instead} of dense representations is crucial for achieving optimal performance in HPC workloads.

	\subsubsection{The Role of Sparsity in Computational Efficiency}
	\paragraph{Accelerating Computational Tasks}
	Sparsity significantly accelerates computational tasks by reducing the number of operations required to perform matrix manipulations. In dense matrices, operations such as matrix multiplication, inversion, and determinant calculation involve every element of the matrices, leading to computational complexity that scales quadratically or even cubically with the size of the matrix. \textbf{Moreover}, sparse matrix operations focus on the non-zero elements, dramatically reducing the computational workload. This efficiency is particularly beneficial \textbf{for} iterative algorithms common in scientific computing and machine learning, where the same operations are repeated many times. \textbf{Additionally}, sparsity allows for more efficient storage and memory usage, as only the non-zero elements need to be stored, resulting in significant savings in memory footprint, especially for large-scale problems. \textbf{Furthermore}, the reduced computational burden enables the application of more sophisticated algorithms and models, contributing to improved accuracy and scalability in various domains. \textbf{In contrast}, dense matrix operations may become infeasible or impractical for large-scale problems due to memory constraints and computational overhead. Thus, embracing sparsity is \textbf{essential} for achieving efficient and scalable computational solutions in modern data-intensive applications.

	\paragraph{Minimizing Memory Requirements}
	The sparse representation of matrices minimizes memory requirements by storing only the non-zero elements and their indices. This approach contrasts with the storage of dense matrices, where space is allocated for every element regardless of its value. The reduced memory footprint of sparse matrices not only conserves valuable system memory but also enhances cache efficiency, leading to faster access times and overall computational performance. 
	
	Moreover, in large-scale data analysis and simulations, where data can easily grow to gigabytes or terabytes, the ability to compactly store and efficiently process data is invaluable. Additionally, the utilization of sparse matrices facilitates more efficient memory management, enabling systems to handle larger datasets without encountering memory overflow issues. 
	
	Furthermore, the optimization of memory usage through sparse matrix representation contributes to improved scalability, allowing algorithms and computations to be applied to increasingly larger datasets without necessitating a proportional increase in memory resources. This scalability is particularly crucial in fields such as machine learning, scientific computing, and big data analytics, where the size and complexity of datasets continue to expand rapidly.
	
	Thus, by employing sparse matrix representations, organizations and researchers can effectively manage memory resources, optimize computational performance, and scale their data processing capabilities to meet the demands of modern data-intensive applications.

	\paragraph{Facilitating Large-Scale Computations}
	Sparsity, a fundamental concept in various domains, plays a pivotal role in enabling the practical execution of large-scale computations, particularly those involving matrices. Sparse matrices, characterized by having a vast majority of zero elements, offer significant advantages over dense matrices in terms of computational efficiency and memory utilization. 
	
	In fields like graph theory, sparse matrix techniques are indispensable for handling massive networks comprising millions of nodes. By efficiently representing the sparse connectivity structure of such networks, computations like shortest path algorithms and network analysis become computationally tractable. This capability is crucial in diverse applications ranging from social network analysis to infrastructure optimization.
	
	Moreover, in the realm of machine learning, sparse feature matrices revolutionize the training process, especially when dealing with massive datasets. In this context, each feature's relevance to only a subset of the data naturally leads to sparse representations. Leveraging sparse matrices not only reduces computational overhead but also facilitates scalability, allowing algorithms to handle large-scale datasets with ease. This efficiency is particularly evident in tasks like text classification, where feature matrices can be extremely high-dimensional, yet sparse due to the nature of language.
	
	Furthermore, the sparsity-induced computational benefits extend beyond graph theory and machine learning. In computational physics, for instance, sparse matrix methods are instrumental in solving systems of partial differential equations arising from complex physical phenomena. The ability to efficiently manipulate sparse matrices significantly accelerates simulations and optimizations, enabling researchers to tackle problems of unprecedented scale and complexity.
	
	In conclusion, the exploitation of sparsity in computations unlocks new frontiers for tackling large-scale problems across various domains. Its impact spans from enhancing algorithmic efficiency in machine learning to revolutionizing computational approaches in fields like graph theory and physics, ultimately paving the way for groundbreaking advancements in science and technology.

	\paragraph{Enhancing Parallel Computing}
	The role of sparsity extends to enhancing parallel computing strategies. Sparse matrix storage formats and computational techniques are often designed with parallelism in mind, enabling the distribution of computations across multiple processors or computing nodes efficiently. This parallelism is key to scaling up scientific simulations, data analysis tasks, and machine learning algorithms to leverage the full capabilities of modern multi-core and distributed computing environments.
	
	Moreover, sparse matrix computations facilitate efficient communication and synchronization among parallel computing nodes. By reducing the amount of data that needs to be exchanged between nodes, sparse matrices minimize communication overhead, allowing parallel systems to operate more efficiently. This streamlined communication is crucial for achieving high-performance computing (HPC) objectives in various domains, including weather forecasting, fluid dynamics simulations, and large-scale optimization problems.
	
	Furthermore, sparsity-driven parallel computing strategies can lead to significant savings in computational resources and energy consumption. Since sparse matrices inherently contain fewer non-zero elements, parallel computations involving sparse data structures require fewer arithmetic operations and memory accesses compared to dense counterparts. As a result, parallel computing platforms can achieve better resource utilization and energy efficiency, making them more environmentally sustainable and cost-effective.
	
	Additionally, the utilization of sparse matrices in parallel computing environments promotes fault tolerance and resilience. By distributing computations across multiple nodes, parallel systems can continue functioning even if individual nodes fail or experience errors. The inherent redundancy in parallel computing architectures, coupled with sparsity-based techniques, enhances system reliability and availability, ensuring uninterrupted operation in mission-critical applications.
	
	In summary, the integration of sparsity principles into parallel computing paradigms offers a multitude of benefits, ranging from improved performance and scalability to enhanced resource efficiency and fault tolerance. Leveraging sparse matrix representations and algorithms enables parallel systems to tackle increasingly complex computational tasks while maximizing the utilization of available hardware resources.

	\paragraph{Challenges and Solutions in Sparsity}
	While sparsity brings numerous computational advantages, it also presents challenges, such as the irregularity of non-zero elements which can complicate the parallelization and optimization of computations. Advanced algorithmic strategies and data structures have been developed to address these challenges, ensuring that the potential of sparsity in enhancing computational efficiency is fully realized. 
	
	Furthermore, the irregular distribution of non-zero elements in sparse matrices often leads to inefficient memory usage and cache utilization. This inefficiency can hamper the performance gains expected from sparsity. Consequently, innovative memory management techniques, such as compressed sparse row (CSR) and compressed sparse column (CSC) formats, have been devised to mitigate these issues. These formats optimize memory allocation and access patterns, allowing for more efficient storage and retrieval of sparse matrix data.
	
	Moreover, while traditional dense matrix operations can be readily parallelized using standard techniques, the inherent irregularity of sparse matrices poses unique challenges for parallel computing. However, parallel algorithms tailored specifically for sparse matrices, including parallel sparse matrix-vector multiplication (SpMV) and parallel sparse LU decomposition, have been developed to harness the computational power of modern parallel architectures effectively. These algorithms exploit the structure of sparse matrices to distribute computations efficiently across multiple processing units, enabling scalable performance improvements.
	
	Additionally, the choice of appropriate data structures and algorithms is crucial for achieving optimal performance with sparse matrices in diverse computational tasks. Techniques such as hierarchical matrix representations and multigrid solvers have been employed to handle large-scale sparse systems encountered in scientific simulations and numerical modeling. By leveraging these specialized tools, researchers can address the computational challenges posed by sparsity and unlock new possibilities for solving complex problems in science and engineering.

	\subsubsection{Applications and Limitations}
	\paragraph{Diverse Applications across Disciplines}
	Sparse matrix computations find extensive applications across a wide range of disciplines, showcasing their versatility and critical importance. In engineering and physics, they are pivotal in solving systems of linear equations that arise in the analysis of structures, fluid dynamics, and electrical circuits, where the underlying matrices representing physical relationships are typically sparse. \textbf{Moreover}, graph theory and network analysis, fundamental to understanding social networks, biological systems, and the structure of the internet, rely heavily on sparse matrices to represent and analyze the connections within large, complex networks efficiently. In the realm of machine learning and data mining, sparse matrices are instrumental in handling high-dimensional data, where they enable the efficient storage and processing of datasets with a large number of features but relatively few nonzero feature values per instance. \textbf{Additionally}, sparse matrices facilitate parallel computations, a crucial aspect in modern computing environments where processing large datasets efficiently is essential. \textbf{Furthermore}, their usage extends to optimization problems, where techniques like compressed sensing exploit sparsity to reconstruct signals accurately from limited measurements, finding applications in medical imaging, signal processing, and many other domains.

	\paragraph{Optimization Challenges}
	Despite their widespread application, optimizing sparse matrix computations presents significant challenges. The effectiveness of optimizations often depends on the specific sparsity pattern of the matrix, which can vary widely across different applications and even within different datasets in the same application. For instance, diagonal sparsity patterns, block sparsity, and random sparsity each require different storage formats and computational strategies to achieve optimal efficiency. And, the scale of the problem also introduces complexity, as very large matrices may necessitate distributed computing solutions that introduce additional layers of optimization challenges, such as data distribution, load balancing, and communication overhead. Moreover, achieving high performance in sparse matrix computations requires careful consideration of memory access patterns and cache utilization, especially when dealing with irregular data structures inherent to sparse matrices. Furthermore, while hardware advancements such as multi-core processors and specialized accelerators offer opportunities for performance improvements, exploiting these resources efficiently often requires sophisticated parallelization techniques tailored to the characteristics of sparse matrix algorithms. Additionally, the dynamic nature of many sparse matrix computations, where the sparsity pattern evolves over time or varies between iterations, poses challenges for achieving sustained performance across different phases of the computation. Therefore, addressing these optimization challenges demands a holistic approach that integrates algorithmic innovations, architectural optimizations, and parallelization strategies tailored to the specific characteristics of sparse matrices and their associated computations.

	\paragraph{Algorithmic and Storage Solutions}
	To address these challenges, a variety of algorithmic and storage solutions have been developed. \textbf{Furthermore}, algorithmic solutions encompass iterative solvers and preconditioners customized for sparse systems, thereby significantly accelerating convergence for specific sparsity patterns. These solvers exploit the inherent structure of sparse matrices, effectively reducing the computational complexity associated with solving large linear systems arising from discretization processes in various fields such as computational fluid dynamics, structural mechanics, and electromagnetics. 
	
	\textbf{Moreover}, storage solutions play a pivotal role in managing memory resources efficiently. Formats like CSR (Compressed Sparse Row) and COO (Coordinate List) are \textbf{designed to minimize memory usage and access time for sparse matrices}. By storing only the non-zero elements along with their corresponding row and column indices, these formats alleviate the burden of storing unnecessary zero entries present in dense matrices, particularly beneficial for large-scale simulations where memory overhead can be a significant concern. 
	
	\textbf{In addition}, recent advancements in software and hardware have spurred the development of specialized libraries and computing architectures tailored explicitly for sparse matrix operations. These innovations encompass highly parallelized algorithms and hardware accelerators like GPUs (Graphics Processing Units) and TPUs (Tensor Processing Units), \textbf{which are optimized for efficiently handling sparse data structures}. Such dedicated solutions help to mitigate some of the inherent computational bottlenecks associated with sparse matrix computations, enabling faster execution and scalability across diverse application domains.

	\paragraph{Future Directions and Research}
	The limitations of sparse matrix computations are an active area of research, with ongoing efforts aimed at developing more adaptive and intelligent algorithms that can automatically adjust to the sparsity pattern and scale of the problem. Furthermore, the integration of machine learning techniques into sparse matrix optimization processes represents a promising avenue for automatically identifying optimal computational strategies. Moreover, as computational capabilities continue to advance, it is likely that many of the current limitations will be overcome, further expanding the applicability and efficiency of sparse matrix computations in solving complex problems across various domains.
	
	One promising direction is the exploration of hybrid approaches that combine traditional numerical methods with machine learning-based strategies. This fusion of techniques could lead to the development of algorithms capable of dynamically adapting to changing problem characteristics, thus improving both accuracy and efficiency. Additionally, research focusing on the development of specialized hardware architectures optimized for sparse matrix computations is gaining traction. These tailored hardware solutions have the potential to significantly accelerate computations by exploiting the inherent sparsity present in many real-world datasets.
	
	Moreover, investigating the application of sparse matrix computations in emerging fields such as quantum computing and computational biology holds great promise. The unique characteristics of these domains present new challenges and opportunities for advancing sparse matrix algorithms and techniques. For instance, in quantum computing, where qubits can represent sparse matrices, developing efficient algorithms tailored to quantum hardware could revolutionize quantum simulations of complex systems.
	
	Furthermore, interdisciplinary collaborations between mathematicians, computer scientists, and domain experts are essential for pushing the boundaries of sparse matrix computations. By leveraging insights from diverse fields, researchers can develop more holistic approaches that address the multifaceted challenges associated with sparse matrix optimization.
	
	In conclusion, the future of sparse matrix computations is characterized by a rich landscape of research avenues spanning algorithmic innovations, hardware advancements, interdisciplinary collaborations, and novel application domains. By embracing these opportunities and addressing the current limitations, the field is poised to make significant strides towards unlocking the full potential of sparse matrix techniques in tackling complex problems.

	\subsubsection{Algorithmic Pseudocode for Sparse Matrix Computations}
	Sparse Matrix Computation is a highly efficient technique utilized in various computational tasks, particularly those involving large matrices with many zero elements. It operates by iteratively processing non-zero elements of the matrix to perform computations, significantly reducing computational effort and memory usage compared to traditional dense matrix operations. The computational procedure begins with the initialization of a result vector $R$ to store the outcome of the multiplication. Then, it proceeds to iterate through each non-zero element $e$ in the sparse matrix. During each iteration, the algorithm retrieves the row index $i$, column index $j$, and value of the non-zero element. It then multiplies this value by the corresponding entry in the input vector $Vector[j]$ and adds the result to the appropriate entry in the result vector $R[i]$. This approach effectively leverages the sparsity of the matrix, ensuring that computations are only performed on non-zero elements, thereby optimizing both computational efficiency and memory utilization. For a visual representation of the operational essence of Sparse Matrix Computation, refer to pseudocode \ref{fig:sparse-matrix-pseudocode}.
	
	\begin{algorithm}
		\caption{Algorithmic Sparse Matrix-Vector Multiplication Pseudocode}
		\begin{algorithmic}[1]
			\Procedure{SparseMatrixVectorMultiply}{SparseMatrix, Vector}
			\State Initialize a result vector $R$ of appropriate size with all zeros
			\For{each non-zero element $e$ in SparseMatrix}
			\State $i \gets$ row index of $e$
			\State $j \gets$ column index of $e$
			\State $value \gets$ value of $e$
			\State $R[i] \gets R[i] + value \times Vector[j]$
			\EndFor
			\State \Return $R$
			\EndProcedure
		\end{algorithmic}\label{fig:sparse-matrix-pseudocode}
	\end{algorithm}

\subsection{Previous Work on ML and AI Interplay with Sparse Matrix Computations}

\paragraph{Optimization for Graph Neural Networks}
A study from 2021 focuses on optimizing sparse matrix multiplications for graph neural networks \cite{qiu2021optimizing}. This research addresses challenges posed by the sparse nature of matrices encountered in graph neural networks by applying algorithmic optimizations and machine learning insights. The resulting improvements enhance the efficiency, scalability, and performance of graph neural networks, offering potential directions for future research in optimizing sparse matrix multiplications through AI integration.

\paragraph{Deep Learning for Sparse Matrix Classification}
In 2018, an innovative approach to sparse matrix classification using deep learning techniques was introduced \cite{pichel2018new}. This approach leverages deep learning to handle the complexities associated with sparse matrices, enhancing accuracy and speed in classification tasks. This advancement provides a foundation for more intelligent sparse matrix handling strategies, offering new possibilities for efficient management in scientific computing and engineering applications.

\paragraph{DL and Sparse Matrix Format Selection}
A 2018 study focuses on automating sparse matrix format selection using deep learning algorithms \cite{zhao2018bridging}. By predicting the most efficient formats for storing and processing sparse matrices, this research reduces the overhead associated with sparse matrix operations. The integration of deep learning techniques demonstrates potential in optimizing computational challenges and improving the performance of high-dimensional data processing tasks.

	\subsection{Algogenic Enhancements for Sparse Matrix Computations}
	
	\subsubsection{Semantic Analysis of Matrix Structure}
	
	\paragraph{Identifying Optimal Storage and Preprocessing Strategies}
	Semantic analysis applied directly to sparse matrix structure identifies the most suitable storage formats and preprocessing strategies, enhancing computational efficiency. This process, tailored to the specifics of sparse matrices, uses language models to analyze the matrix's origin and intended application, thereby predicting which storage formats (CSR, COO, CSC) and preprocessing methods (e.g., bandwidth reduction) are most beneficial. The approach considers not just the sparsity pattern but also the semantic context, ensuring that the chosen strategies are aligned with the matrix's characteristics and use case. Dynamic adaptation to changing data or computational requirements further refines this optimization, making it a practical and valuable tool in sparse matrix computations.
	
	\paragraph{Operationalizing Structural Insights}
	Operationalizing structural insights involves translating the outcomes of semantic analysis into concrete actions, such as selecting the most appropriate storage format or recommending specific algorithms to improve sparse matrix computation efficiency. This process is informed by the matrix's structure, enabling automated and dynamic optimization. It introduces a system that evolves over time, adapting to new findings and ensuring that computational strategies remain optimal, providing a clear path toward the practical application of these insights in real-world scenarios.
	
	\paragraph{Enhancing Efficiency and Accuracy}
	Applying semantic analysis to sparse matrix computations aims to significantly boost both efficiency and accuracy. By aligning storage and preprocessing methods with the matrix's inherent characteristics, computations are executed more swiftly and with enhanced precision. This method not only accelerates processing but also ensures that outcomes are reliable and meaningful, showcasing its practicality across various application areas.
	
	\subsubsection{Predictive Preconditioning Guidance}
	
	\paragraph{Tailoring Preconditioners to Matrix Characteristics}
	Predictive preconditioning guidance utilizes semantic context to recommend preconditioners that improve iterative solver performance, demonstrating a practical application of LLMs in enhancing solver efficiency. By analyzing the matrix and computational goals, this guidance adapts preconditioning techniques dynamically, ensuring optimal solver performance through a context-aware and adaptable approach.
	
	\paragraph{Automating the Selection of Preconditioning Techniques}
	Automating preconditioner selection through LLM insights streamlines the optimization process in sparse matrix computations, ensuring that the most appropriate techniques are employed based on matrix characteristics. This automation not only enhances computational efficiency but also introduces adaptability, allowing for real-time adjustments to preconditioning strategies as computational or matrix properties evolve.
	
	\paragraph{Impact on Solver Efficiency and Solution Quality}
	Predictive preconditioning significantly influences solver efficiency and the quality of solutions by ensuring the use of the most effective preconditioners. This approach enhances solver performance, reduces computation time, and improves result accuracy, demonstrating its practical value in advancing sparse matrix computations.
	
	\subsubsection{Dynamic Storage Format Selection}
	
	\paragraph{Optimizing Memory Usage and Access Patterns}
	Dynamic storage format selection, informed by LLMs, optimizes sparse matrix storage to improve memory efficiency and computational speed. By choosing the most suitable storage format based on the matrix's sparsity pattern and computational requirements, this strategy ensures that memory and access efficiency are maximized, showcasing a practical application of LLM insights in enhancing sparse matrix computations.
	
	\paragraph{Adapting to Computational Contexts}
	Adapting storage formats to computational contexts involves LLMs considering available resources, computational load, and specific operations to adjust the storage format dynamically. This adaptability optimizes computations by selecting formats that balance compression efficiency and access complexity, demonstrating the practical application of dynamic storage format selection in various computational scenarios.
	
	\paragraph{Enhancing Computational Performance and Scalability}
	Dynamic selection of storage formats, guided by LLMs, significantly enhances the computational performance and scalability of sparse matrix operations. By optimizing storage formats based on the matrix's characteristics and the computational task, this approach facilitates faster and more efficient computations, highlighting its practical benefits in handling large-scale sparse matrices.
	
	\subsubsection{Adaptive Algorithm Pathways}
	
	\paragraph{Customizing Computational Strategies for Efficiency}
	Adaptive algorithm pathways, informed by LLM analysis, optimize sparse matrix computations by selecting the most efficient computational strategies. This customization considers the matrix's specific characteristics and the computational task, enhancing efficiency and performance. This approach demonstrates the practical implementation of adaptive strategies, significantly improving computational outcomes.
	
	\paragraph{Incorporating Contextual and Semantic Insights}
	Incorporating contextual and semantic insights into algorithm selection enriches sparse matrix computations by aligning computational strategies with the semantic significance of the tasks. This integration facilitates a deeper understanding and optimization of computational processes, showcasing the practical benefits of a context-aware approach in enhancing algorithm selection and application.
	
	\paragraph{Enhancing Solver Performance and Accuracy}
	Adaptive algorithm pathways improve solver performance and accuracy in sparse matrix computations by tailoring computational strategies to each matrix's needs. This approach leads to faster convergence rates and more efficient operations, underscoring the practical advantages of adaptively optimizing computational strategies.
	
	\subsubsection{Intelligent Sparsity Pattern Recognition}
	
	\paragraph{Optimizing Computations Through Pattern Analysis}
	Intelligent sparsity pattern recognition employs LLMs to identify unique sparsity patterns, enabling the customization of computational strategies. This optimization leads to reduced complexity and improved efficiency, highlighting the practical application of pattern recognition in enhancing sparse matrix computations.
	
	\paragraph{Dynamic Adaptation to Matrix Changes}
	Dynamic adaptation to matrix changes, facilitated by LLMs, ensures that computational strategies remain optimized as sparsity patterns evolve. This proactive adjustment maintains computational efficiency and robustness, demonstrating the practical value of intelligent sparsity pattern recognition in dynamic computational environments.
	
	\paragraph{Facilitating Parallel Processing and Scalability}
	Intelligent sparsity pattern recognition enhances parallel processing and scalability in sparse matrix computations. By identifying segments suitable for parallelization, this approach improves processing speed and scalability, showcasing the practical benefits of leveraging sparsity patterns for optimized computation.
	
	\subsubsection{Adaptive Precision Management}
	
	\paragraph{Balancing Computational Efficiency with Numerical Accuracy}
	Adaptive Precision Management optimizes numerical accuracy in sparse matrix operations by dynamically adjusting precision levels. This balance enhances computational efficiency without compromising accuracy, demonstrating a practical approach to managing precision in complex computational tasks.
	
	\paragraph{Implementation of Precision Adjustment Strategies}
	The implementation of precision adjustment strategies involves LLMs dynamically optimizing precision levels based on computational needs. This adaptive approach maintains computational efficiency and accuracy, showcasing the practical application of Adaptive Precision Management in enhancing sparse matrix computations.
	
	\paragraph{Enhancing Sparse Matrix Computations Across Domains}
	Adaptive Precision Management improves sparse matrix computations across various domains by ensuring efficient resource allocation and maintaining numerical accuracy. This approach enables the tackling of larger and more complex problems, demonstrating its practical value in advancing computational methodologies.
	
	\subsubsection{Predictive Load Balancing for Parallel Operations}
	
	\paragraph{Optimizing Parallel Processing in Sparse Matrix Computations}
	Predictive Load Balancing optimizes task distribution in parallel sparse matrix operations, enhancing computational efficiency. By predicting efficient task allocation, this approach ensures optimal processor utilization, showcasing the practical benefits of LLM-driven strategies in improving parallel processing performance.
	
	\paragraph{Dynamic Adaptation to Computational Workloads}
	Dynamic adaptation to computational workloads ensures optimal task distribution in sparse matrix operations, enhancing computational efficiency. This adaptability allows for real-time adjustments, maintaining high performance and resource utilization, highlighting the practical application of predictive load balancing in managing dynamic workloads.
	
	\paragraph{Facilitating Scalable and Efficient Sparse Matrix Operations}
	Predictive Load Balancing facilitates scalable and efficient sparse matrix operations, enabling effective handling of large-scale problems. By optimizing task distribution, this approach improves performance and scalability, demonstrating its practical value in enhancing high-performance computing environments.
	
	\subsubsection{Semantic Result Interpretation}
	
	\paragraph{Contextualizing Computational Outcomes}
	Semantic Result Interpretation contextualizes the outcomes of sparse matrix computations, transforming numerical results into meaningful insights. This process enhances the understanding and applicability of computational findings, showcasing the practical benefits of LLMs in bridging the gap between raw data and actionable information.
	
	\paragraph{Bridging Numerical Analysis and Domain Knowledge}
	Bridging numerical analysis with domain knowledge enhances the interpretation of sparse matrix results, providing insights that extend beyond raw data. This integration facilitates a deeper understanding of computational findings, demonstrating the practical application of LLMs in enriching result interpretation with domain-specific context.
	
	\paragraph{Enhancing Reporting and Communication}
	Semantic interpretation enhances the reporting and communication of computational results, enabling the generation of accessible and informative reports. This capability improves stakeholder understanding and decision-making, showcasing the practical benefits of LLMs in translating complex computations into actionable insights.
	
	\subsubsection{Automated Documentation and Reporting}
	
	\paragraph{Streamlining Report Generation with LLM Insights}
	Automated Documentation and Reporting streamlines the generation of reports on sparse matrix computations, leveraging LLM insights for efficiency and clarity. This process ensures comprehensive and accessible documentation, demonstrating the practical application of LLMs in enhancing the communication of computational outcomes.
	
	\paragraph{Enhancing Communication of Complex Computational Results}
	Enhancing communication through automated documentation and reporting facilitates the understanding of complex computational results, making them accessible to a broader audience. This approach underscores the practical benefits of LLMs in bridging the gap between technical computations and stakeholder engagement.
	
	\paragraph{Customizing Reports to Audience Needs}
	Customizing reports to audience needs ensures that documentation is relevant and engaging, highlighting the practical application of LLMs in tailoring communication strategies. This adaptability enhances the impact of computational findings, demonstrating the value of automated documentation in addressing diverse informational needs.
	
	\subsubsection{Predictive Maintenance and Update Suggestions}
	
	\paragraph{Proactive Optimization of Sparse Matrix Computations}
	Predictive Maintenance and Update Suggestions enable proactive optimization of sparse matrix computations, anticipating and addressing future computational needs. This approach ensures sustained efficiency and performance, showcasing the practical benefits of LLM-driven strategies in maintaining computational integrity over time.
	
	\paragraph{Adapting to Evolving Computational Requirements}
	Adapting to evolving computational requirements through predictive maintenance ensures that sparse matrix computations remain efficient and relevant. This adaptability demonstrates the practical application of LLM insights in optimizing computational strategies to meet changing needs and challenges.
	
	\paragraph{Ensuring Long-term Computational Integrity and Performance}
	Ensuring long-term computational integrity and performance through predictive maintenance and update suggestions highlights the practical benefits of LLMs in maintaining the efficiency and relevance of sparse matrix computations. This approach fosters continuous improvement and innovation, ensuring that computational frameworks can adapt and thrive in dynamic environments.

	\subsubsection{Pseudocode for Algogenic Sparse Matrix Computations}
	The Algogenic sparse matrix computation approach utilizes AI to enhance conventional methods by dynamically adjusting computation parameters and strategies in response to system behavior and real-time error estimates. This pseudocode, available in \ref{fig:sparse-matrix-computation-Algogen-pseudocode}, outlines a sophisticated framework integrating AI-driven enhancements for adaptive computation control, element selection, optimization criteria, and real-time parameter adjustments.
	
	\begin{algorithm}
		\caption{Algogenic Sparse Matrix Computations Pseudocode}
		\begin{algorithmic}[1]
			\Procedure{AlgogenicSparseMatrixComputation}{SparseMatrix}
			
			\Comment{Preprocessing Phase}
			\State Analyze the matrix structure with LLM for semantic insights.
			\State Apply predictive preconditioning based on LLM insights.
			\State Select the optimal storage format dynamically with LLM recommendations.
			
			\Comment{Core Computation Phase}
			\State Choose the most suitable computational pathways with insights from LLM.
			\State Perform intelligent sparsity pattern recognition using LLM.
			\State Manage numerical precision adaptively throughout the operations.
			\State Balance computational load predictively across parallel processes.
			\State Execute the core sparse matrix operations.
			
			\Comment{Postprocessing Phase}
			\State Interpret computational results semantically with LLM for enhanced understanding.
			\State Generate automated documentation and reporting using LLM insights.
			\State Recommend predictive maintenance and updates based on LLM analysis.
			
			\EndProcedure
		\end{algorithmic}\label{fig:sparse-matrix-computation-Algogen-pseudocode}
	\end{algorithm}

	\begin{figure}
		\centering
		\includegraphics[width=0.83\textwidth]{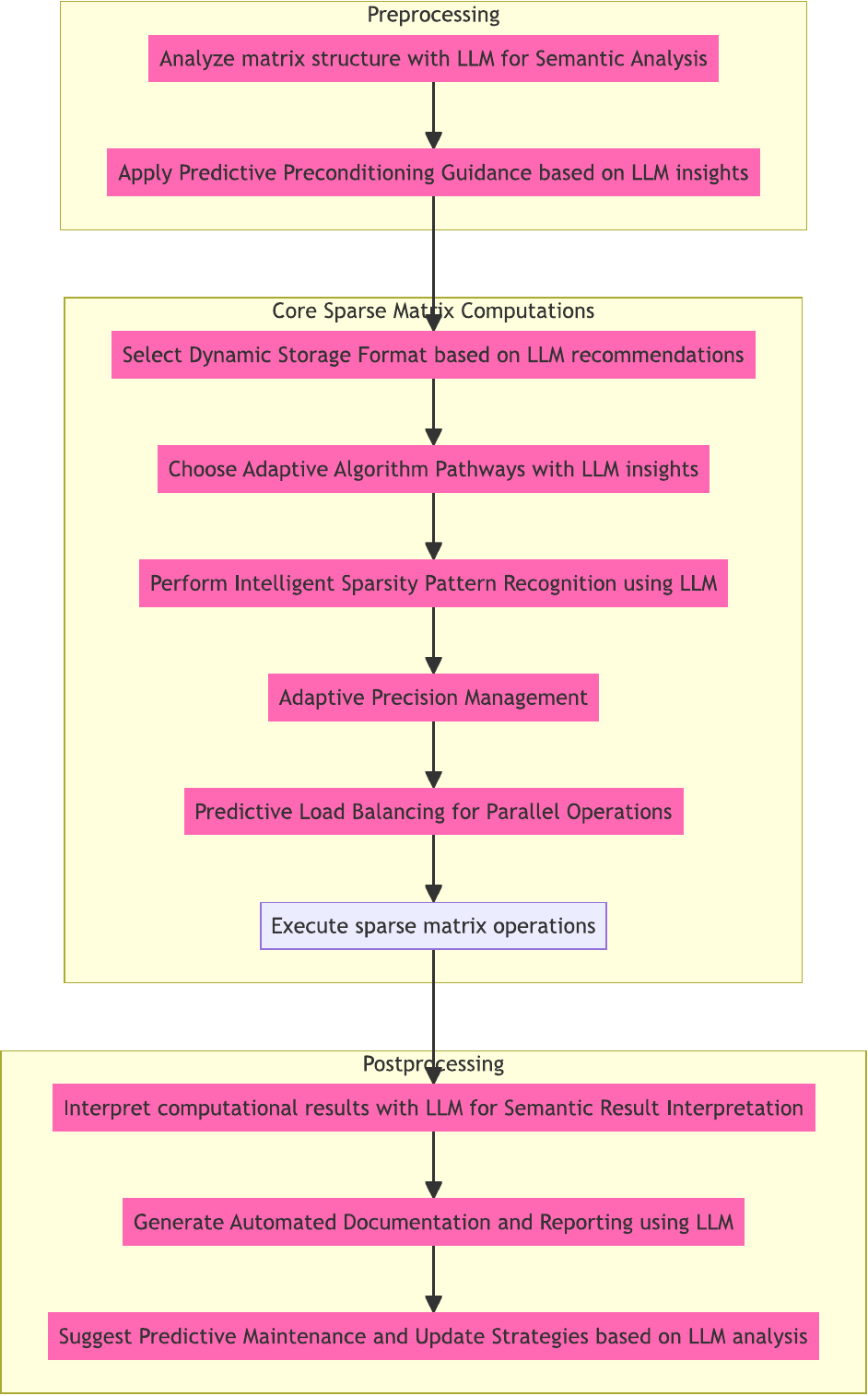}
		\caption{Integrating Algogens with Sparse Matrix Computations: This diagram visualizes the comprehensive Algogenic framework applied to sparse matrix computations, highlighting the seamless integration of generative AI across preprocessing, core computation, and postprocessing phases. The framework begins with LLM-driven semantic analysis and predictive preconditioning in the preprocessing phase, ensuring matrices are optimally prepared. In the core phase, AI dynamically selects computational pathways and manages precision, enhancing efficiency and accuracy. Postprocessing leverages LLMs for semantic interpretation of results and automated reporting, culminating in a sophisticated, AI-enhanced computational process that significantly improves the performance and applicability of sparse matrix operations in various domains.}
		\label{fig:sparse_matrices}
	\end{figure}

	\section{Numerical Integration}\index{Numerical Integration}
	\subsection{Introduction to Numerical Integration}
	\subsubsection{The Concept of Numerical Integration}
	\paragraph{Overview of Numerical Integration}
	Numerical integration is a fundamental technique in computational mathematics that aims to approximate the integral of a function over a specified interval. This method is particularly valuable when the integral cannot be determined analytically due to the complexity of the function or when dealing with integrals over irregular domains. The core idea is to estimate the area under the curve represented by the function over the interval of interest.
	
	Moreover, numerical integration plays a crucial role in various scientific and engineering applications where precise calculations are required. It enables researchers and practitioners to solve a wide range of problems, including those in physics, engineering, economics, and more. By employing numerical integration techniques, scientists can analyze complex systems, model physical phenomena, and predict outcomes with high accuracy.
	
	Furthermore, numerical integration methods come in various forms, each with its advantages and limitations. Some common techniques include the trapezoidal rule, Simpson's rule, and Gaussian quadrature. These methods differ in their approaches to approximating the integral and their computational complexity. While simpler techniques like the trapezoidal rule are easy to implement and understand, more advanced methods like Gaussian quadrature offer higher accuracy but may require more computational resources.
	
	Additionally, the choice of numerical integration method depends on factors such as the smoothness of the function, the desired level of accuracy, and computational efficiency. Practitioners often need to balance these considerations when selecting an appropriate technique for a particular problem.
	
	In summary, numerical integration is a versatile tool that allows for the approximation of integrals in situations where analytical solutions are impractical or unavailable. Its widespread use across various disciplines underscores its importance in modern computational mathematics and scientific research.

	\paragraph{Mathematical Formulation}
	The process of numerical integration serves as a pivotal technique in approximating definite integrals, especially when exact solutions are unattainable or impractical. It hinges on dissecting the interval of interest, \([a, b]\), into smaller, manageable subintervals. This division is crucial as it allows us to encapsulate the behavior of the function \(f(x)\) within these localized regions, thereby facilitating a more accurate estimation of the overall integral. Each subinterval is assigned specific points where the function's value is evaluated. These points play a pivotal role in the accuracy of the approximation; their selection is governed by various numerical integration methods, each with its own set of advantages and limitations.
	
	The heart of numerical integration lies in the choice of an appropriate formula for estimating the integral over each subinterval. These formulas, often termed quadrature rules, encapsulate diverse strategies for capturing the integral's essence. They typically involve evaluating \(f(x)\) at designated points within the subinterval and then employing a weighted sum to approximate the integral. The weights are intricately tied to the width of the subinterval, reflecting the intuition that larger intervals contribute more significantly to the overall integral.
	
	Among the plethora of numerical integration methods, diverse approaches emerge. Some methods, like the trapezoidal rule, approximate the function within each subinterval using linear interpolation, providing a straightforward yet effective estimation. Others, such as Simpson's rule, leverage higher-order polynomial approximations, offering enhanced accuracy at the expense of increased computational complexity. The choice of method hinges on a delicate balance between accuracy requirements and computational efficiency, with each method offering its own trade-offs.
	
	Additionally, the partitioning of the interval \([a, b]\) into smaller segments underscores the discretization inherent in numerical integration. This discretization introduces approximation errors, especially when dealing with functions exhibiting rapid fluctuations or irregular behavior. Consequently, practitioners must strike a careful balance between partitioning the interval finely enough to capture the function's nuances and avoiding excessive computational overhead.
	
	Ultimately, numerical integration stands as a cornerstone in computational mathematics, offering a diverse array of tools for approximating integrals with varying degrees of accuracy and efficiency. By understanding the underlying principles governing these methods, practitioners can navigate the trade-offs inherent in numerical integration and select the most suitable approach for a given problem.

	\paragraph{Significance in Various Fields}
	Numerical integration is indispensable across various scientific and engineering disciplines. It plays a crucial role in solving differential equations, optimizing engineering designs, calculating areas and volumes, and performing financial analyses where precise integral values are required but are difficult to obtain through analytical methods. The ability to accurately approximate integrals enables researchers and professionals to model and solve real-world problems that involve complex systems and relationships.
	
	Furthermore, in the field of physics, numerical integration is extensively utilized to simulate dynamic systems governed by differential equations. For instance, in celestial mechanics, numerical integration techniques are employed to predict the motion of planets and satellites, enabling astronomers to study celestial phenomena such as eclipses and planetary orbits with high precision.
	
	Moreover, in computational biology, numerical integration is essential for modeling biological processes such as population dynamics, biochemical reactions, and neural networks. These models help biologists understand complex biological systems, predict the behavior of organisms under different conditions, and design targeted interventions for diseases.
	
	In addition, numerical integration finds wide application in financial mathematics for pricing complex derivative securities and managing risk. By numerically integrating stochastic differential equations, financial analysts can simulate the behavior of asset prices and assess the performance of investment portfolios under uncertain market conditions, facilitating informed decision-making in the financial industry.
	
	Therefore, the versatility and efficacy of numerical integration techniques make them indispensable tools across diverse fields, empowering researchers and practitioners to tackle complex problems and drive innovation.

	\subsubsection{Key Principles and Mechanisms}
	\paragraph{Simplification Through Approximation}
	The essence of numerical integration lies in the art of simplification, where the daunting task of integrating complex functions is tamed through strategic approximations. This pivotal process involves transforming intricate functions into more tractable forms, facilitating their seamless integration. A common strategy entails replacing the original function with a polynomial counterpart, chosen judiciously to closely mirror the behavior of the original function within the designated interval. Through this substitution, the integration process transcends the intricacies of the original function, yielding a manageable expression amenable to computational methods. The fidelity of this approximation profoundly impacts the accuracy of the resultant integral, as the efficacy of numerical integration hinges upon the ability of the surrogate function to faithfully emulate the nuances of the original across the specified domain.

	\paragraph{Discretization of the Integration Domain}
	The process of numerical integration involves discretizing the integration domain into smaller segments or subintervals. This discretization is crucial because it enables the application of approximation methods over manageable pieces of the domain, thus making the computation feasible. \textbf{Moreover}, by breaking down the integration domain into smaller parts, numerical integration methods can better capture the intricacies of the function being integrated, especially when dealing with functions that exhibit varying behavior across different regions. 
	
	For instance, consider the Trapezoidal Rule. \textbf{While} it divides the domain into trapezoids, \textbf{the Simpson's Rule}, on the other hand, \textbf{divides} the domain into panels that are approximated by parabolas. This difference in approach not only affects the accuracy of the approximation but also the computational effort required. \textbf{Additionally}, by choosing smaller subintervals, the accuracy of the approximation can be increased, at the expense of computational resources. 
	
	However, it is important to note that the choice of discretization method should strike a balance between accuracy and computational efficiency. \textbf{Furthermore}, the selection of the appropriate number of subintervals is crucial. Too few subintervals may result in an inaccurate approximation, \textbf{while} too many may lead to unnecessary computational overhead. \textbf{Hence}, the discretization process plays a pivotal role in the success of numerical integration, as it directly impacts the quality of the approximation obtained.

	\paragraph{Advanced Techniques for Enhanced Accuracy}
	For functions with complex behaviors or for high-accuracy requirements, more sophisticated numerical integration techniques are employed. \textbf{Gaussian Quadrature}, for example, optimizes both the placement of sample points and the weighting given to each point, providing high accuracy with fewer evaluations of the function. This method strategically selects the integration points and weights to match the integrand's behavior, thereby minimizing error. Unlike simple methods like the trapezoidal rule or Simpson's rule, Gaussian Quadrature achieves high precision even with relatively few points, making it efficient for functions with rapid oscillations or steep gradients.
	
	\textbf{Monte Carlo Integration}, on the other hand, employs random sampling within the integration domain, making it particularly useful for high-dimensional integrals where traditional methods become computationally prohibitive. By randomly selecting points, Monte Carlo Integration bypasses the need for a grid-like structure, allowing it to handle irregularly shaped regions and functions with discontinuities. While it typically requires more function evaluations to achieve a similar level of accuracy compared to deterministic methods, its strength lies in its ability to provide approximate solutions for complex integrals that defy conventional techniques. Moreover, Monte Carlo Integration naturally lends itself to parallelization, enabling efficient use of computational resources for large-scale integration tasks.

	\paragraph{Error Estimation and Adaptivity}
	A crucial aspect of numerical integration is the estimation of error and adaptivity in the integration process. Many numerical integration techniques include methods for estimating the error of the approximation, allowing the user to adjust the parameters of the method (such as the number of subdivisions or sample points) to achieve a desired level of accuracy. Adaptive integration methods go a step further by automatically adjusting these parameters in regions where the function is difficult to approximate, thereby optimizing both the accuracy and efficiency of the integration.
	
	Furthermore, in adaptive integration, the adjustment of parameters is typically based on local error estimates. This means that instead of uniformly dividing the interval into smaller subintervals, the method intelligently allocates more computational resources to regions where the function exhibits significant variations or sharp changes. Consequently, the computational effort is focused where it is most needed, leading to a more efficient use of computational resources.
	
	Moreover, adaptive integration techniques often employ sophisticated strategies for error estimation, such as Richardson extrapolation or hierarchical error estimation. These techniques allow for a more nuanced assessment of the accuracy of the approximation, taking into account factors such as the smoothness of the function and the behavior of its derivatives. As a result, the adaptive integration process can provide accurate solutions even for functions with complex behavior or discontinuities.
	
	Additionally, the adaptivity in numerical integration methods is essential for handling problems with highly oscillatory or singular integrands. Traditional fixed-step integration methods may struggle to accurately capture the behavior of such functions, leading to significant errors in the approximation. However, adaptive techniques can dynamically adjust the step size or sampling density to effectively capture the underlying structure of the integrand, thereby improving the accuracy of the numerical solution.
	
	In conclusion, error estimation and adaptivity play crucial roles in numerical integration, allowing for the accurate approximation of integrals across a wide range of functions and domains. By intelligently adjusting parameters and allocating computational resources based on local error estimates, adaptive integration methods can provide accurate solutions efficiently, even for challenging problems with complex or singular behavior.

	\paragraph{Mathematical Representation}
	Mathematically, the choice of approximation and discretization strategies leads to different formulations. For a given function \(f(x)\) over an interval \([a, b]\), the integral \(\int_a^b f(x) \, dx\) is approximated by a sum \(\sum_{i} w_i f(x_i)\), where \(x_i\) are the chosen points within the interval and \(w_i\) are the weights associated with each point. However, the specific choice of \(x_i\) and \(w_i\) depends on the numerical integration method being used. 
	
	Different methods employ diverse sets of \(x_i\) and \(w_i\) to achieve accurate approximations efficiently. For instance, the Gaussian quadrature method selects \(x_i\) as roots of orthogonal polynomials and \(w_i\) as corresponding quadrature weights, ensuring high precision for polynomials of certain degrees. Conversely, the trapezoidal rule partitions the interval into equal subintervals, employing equally spaced \(x_i\) and \(w_i\), leading to straightforward implementation but sacrificing accuracy, especially for functions with rapid variations. 
	
	Furthermore, the choice of \(x_i\) and \(w_i\) directly impacts computational resources. Techniques like adaptive quadrature dynamically adjust \(x_i\) to concentrate computation where the function varies most, potentially reducing the number of evaluations and improving efficiency. Conversely, fixed methods like Simpson's rule predefine \(x_i\), simplifying implementation but possibly requiring more points to achieve desired accuracy.
	
	In summary, while various numerical integration methods aim to maximize accuracy while minimizing computational resources, the specific choice of approximation points and weights profoundly influences the effectiveness and efficiency of the integration process.

	\subsubsection{The Role of Error Estimation}
	\paragraph{Importance of Error Estimation}
	Error estimation in numerical integration serves as a critical tool for assessing the reliability of the integral approximation. It quantifies the difference between the true value of the integral and its numerical estimate, offering a measure of the accuracy achieved by the computational method. This insight is invaluable, as it guides practitioners in making informed decisions about whether the approximation is sufficiently accurate for their purposes or whether further refinement is necessary.
	
	Moreover, error estimation aids in understanding the limitations of the chosen numerical integration technique. It provides a clear indication of where the method might falter, allowing for adjustments or alternative approaches to be considered. Additionally, it enhances the overall confidence in the results obtained through numerical integration, especially in scientific and engineering applications where precision is paramount.
	
	Furthermore, error estimation facilitates the comparison of different numerical integration methods. By quantifying the accuracy of each approach, researchers can determine which method is most suitable for a particular problem or dataset. This comparative analysis ensures that the chosen method not only provides accurate results but also does so efficiently, saving computational resources and time.
	
	Consequently, integrating error estimation into the numerical integration process is essential for ensuring the validity and reliability of the results obtained. It empowers practitioners to make informed decisions, understand the limitations of their methods, and select the most appropriate approach for their specific needs. In essence, error estimation serves as a cornerstone in the practice of numerical integration, guiding the way towards more accurate and dependable computational solutions.

	\paragraph{Mechanisms of Error Estimation}
	Error estimation mechanisms play a pivotal role in numerical integration methods, ensuring the reliability and accuracy of computed results. These mechanisms, while diverse in their approaches, primarily operate by gauging the behavior of the integrated function or by comparing different approximations.
	
	One common approach involves assessing the rate of change of the function being integrated. By scrutinizing how rapidly the function varies within the integration domain, analysts can infer the potential errors in the computed results. This strategy is particularly evident in techniques like Richardson extrapolation. Here, successive refinements of the integration domain are employed to observe the convergence behavior of the numerical solution. By extrapolating the results obtained with finer subdivisions, analysts can approximate the error associated with the integration process.
	
	Moreover, error estimation in numerical integration often involves comparing approximations of varying orders. Techniques such as Richardson extrapolation explicitly rely on this principle, where solutions obtained with different levels of accuracy are juxtaposed to infer the error characteristics. Similarly, adaptive integration methods operate on the premise of estimating local errors within each subinterval. By dynamically adjusting the subdivision strategy based on these localized error assessments, these methods ensure that the overall error across the entire integration domain remains within acceptable bounds.
	
	In essence, the mechanisms of error estimation in numerical integration methods serve to guide the refinement and adaptation of computational strategies, ultimately enhancing the accuracy and efficiency of numerical computations.

	\paragraph{Dynamic Parameter Adjustment}
	The ability to estimate error dynamically is a cornerstone of adaptive numerical integration techniques. These methods automatically adjust the parameters of the integration process, such as the number and placement of sample points or the granularity of the subdivision, based on the estimated error. By focusing computational efforts on regions where the function exhibits complex behavior and the approximation error is likely to be higher, adaptive methods achieve a more efficient allocation of resources, optimizing the balance between computational effort and accuracy.
	
	Moreover, adaptive integration techniques offer versatility in handling functions with varying degrees of complexity. Whereas traditional numerical integration methods employ fixed parameters throughout the computation, adaptive approaches dynamically refine the approximation, allowing for finer discretization in regions of rapid function variation. This adaptability is particularly advantageous when dealing with functions characterized by localized spikes, oscillations, or discontinuities. Consequently, the integration process can effectively capture intricate features of the function's behavior, ensuring enhanced accuracy without unnecessarily dense sampling in regions of smooth variation.
	
	Furthermore, adaptive methods inherently promote computational efficiency by reducing the total number of sampling points required to achieve a desired level of accuracy. By selectively refining the approximation only where necessary, computational resources are utilized judiciously, mitigating the computational burden associated with exhaustive sampling strategies. Additionally, the adaptive nature of these techniques empowers them to dynamically respond to changes in the function's behavior during the integration process. Thus, they can efficiently adapt to evolving requirements and effectively accommodate functions with evolving characteristics.
	
	In summary, adaptive numerical integration techniques provide a robust framework for efficiently and accurately approximating the integral of complex functions. Through dynamic parameter adjustment and targeted refinement of the approximation, these methods offer superior performance compared to traditional approaches, making them indispensable tools for a wide range of scientific and engineering applications.

	\paragraph{Mathematical Formulation of Error Estimation}
	Mathematically, error estimation often involves calculating an upper bound on the error based on known properties of the function being integrated. This process is crucial in numerical methods for approximating definite integrals as it provides insights into the reliability of the obtained results. For instance, if \(E\) represents the error of an approximation, it might be bounded by an expression involving the maximum value of the function’s derivative within the integration domain. This bound serves as a crucial metric in assessing the quality of the numerical integration.
	
	Moreover, this mathematical framework enables practitioners to make informed decisions regarding the granularity of the approximation. For example, in the case of the Trapezoidal Rule, \(E \leq \frac{M(b-a)^3}{12n^2}\), where \(M\) denotes the maximum value of the second derivative of the function, \(b-a\) represents the width of the integration interval, and \(n\) signifies the number of subdivisions. This formula not only provides a theoretical estimate of the error but also guides the selection of \(n\) to achieve a desired level of accuracy.
	
	Furthermore, understanding the relationship between the error and the parameters involved in the numerical method is essential for optimizing computational resources. By leveraging mathematical analysis, practitioners can efficiently allocate computing power while ensuring that the error remains within acceptable bounds.
	
	Hence, error estimation serves as a cornerstone in the development and application of numerical integration techniques, enabling the attainment of accurate results in various scientific and engineering domains.

	\paragraph{Adapting to Function Characteristics}
	Effective error estimation requires consideration of the function's characteristics, such as smoothness, presence of singularities, or rapid oscillations. Techniques must be selected and adapted based on these characteristics to provide accurate error estimates. 
	
	\textbf{Smoothness}: For smooth functions, traditional error estimation methods like Richardson extrapolation or Romberg integration can be highly effective. These techniques leverage the continuity and differentiability of smooth functions to extrapolate accurate estimates from a sequence of successively refined approximations.
	
	\textbf{Presence of Singularities}: Functions with singularities pose challenges to conventional error estimation approaches due to their discontinuities or infinite behavior at certain points. Specialized methods such as adaptive quadrature or singularity detection algorithms become crucial in such cases. These techniques dynamically adjust the integration step size to concentrate computational effort around singularities, ensuring accurate estimates despite the challenging function behavior.
	
	\textbf{Rapid Oscillations}: Functions exhibiting rapid oscillations require careful handling to prevent significant error accumulation. Approaches like adaptive mesh refinement or Filon quadrature excel in capturing the fine details of oscillatory functions by strategically placing integration points in regions of high oscillation frequency. This adaptability ensures that the integration accurately captures the oscillatory behavior without sacrificing computational efficiency.
	
	For functions with known problematic features, specialized error estimation techniques that take these features into account can offer more reliable accuracy assessments, ensuring that numerical integration results are both precise and trustworthy.

	Error estimation plays a pivotal role in numerical integration, enabling the dynamic adjustment of integration parameters to achieve an optimal trade-off between computational effort and the accuracy of results. Through careful application and understanding of error estimation techniques, practitioners can significantly enhance the quality and reliability of numerical integration outcomes.

	\subsubsection{Applications and Limitations}
	\paragraph{Broad Spectrum of Applications}
	Numerical integration is a versatile tool with applications spanning across multiple fields, illustrating its fundamental importance in both theoretical and applied sciences. In physics, it is essential for solving integrals that arise in the study of motion, electromagnetism, and quantum mechanics, where exact solutions are often unattainable. \textbf{Moreover}, numerical integration plays a crucial role in computational simulations, allowing physicists to model complex systems and phenomena, such as fluid dynamics or celestial mechanics, where analytical solutions are impractical or non-existent. 
	
	Engineering applications \textbf{likewise} benefit extensively from numerical integration techniques. Structural analysis, for instance, heavily relies on integration for calculating stress distributions, deformations, and stability conditions in diverse engineering structures like bridges, buildings, and aircraft components. \textbf{Furthermore}, in materials science, numerical integration aids in predicting material behavior under various conditions, aiding in the design of innovative materials with tailored properties, from composite materials to alloys used in aerospace applications.
	
	In the realm of finance, numerical integration methods are indispensable \textbf{for} pricing exotic financial derivatives and managing portfolio risks. These methods enable financial analysts to compute complex integrals arising in option pricing models, stochastic calculus, and risk assessment frameworks. \textbf{Additionally}, numerical integration facilitates the calibration of financial models to market data, allowing for accurate pricing and risk management strategies in volatile markets.
	
	The fields of statistics and data science heavily rely on numerical integration for \textbf{handling} probability distributions, cumulative distribution functions, and statistical inference. \textbf{Moreover}, in Bayesian statistics, numerical integration plays a central role in estimating posterior distributions and conducting Bayesian model comparison, facilitating robust decision-making based on complex probabilistic models and large-scale datasets.

	\paragraph{Inherent Limitations and Challenges}
	Despite its wide applicability, numerical integration is subject to several limitations and challenges that stem from its approximate nature. One of the primary limitations is the potential for approximation errors, which can vary significantly depending on the method used, the characteristics of the function being integrated, and the specific requirements of the application. These errors necessitate careful error estimation and management strategies to ensure that the results are within acceptable bounds for the task at hand.
	
	Computational complexity presents another challenge, particularly for high-dimensional integrals or functions that require a fine resolution to accurately approximate. As the dimensionality or resolution increases, the computational resources needed can grow exponentially, making some problems intractable with direct numerical integration methods.
	
	Additionally, functions with singularities, sharp discontinuities, or highly oscillatory behavior pose specific difficulties for numerical integration. Traditional methods may struggle to accurately capture the behavior of such functions, requiring specialized techniques or adaptive methods that can more effectively manage these features.
	
	\paragraph{Overcoming Limitations with Advanced Techniques}
	The limitations of numerical integration have spurred the development of advanced techniques designed to address these challenges. Adaptive integration methods, such as those incorporating recursive partitioning or adaptive step-size control, offer a way to manage complex functions more effectively by dynamically adjusting their parameters based on the function's behavior. This dynamic adjustment allows the integration algorithm to concentrate computational effort where it is most needed, thereby improving efficiency and accuracy, particularly in regions of rapid function variation or high curvature.
	
	Multidimensional integration techniques, including Monte Carlo and quasi-Monte Carlo methods, present alternative strategies for handling high-dimensional problems more efficiently than traditional numerical methods. By randomly or quasi-randomly sampling points from the integration domain, these methods effectively explore the function space, yielding estimates of the integral with reduced computational effort compared to grid-based approaches. This makes them particularly well-suited for problems with a large number of dimensions where grid-based methods become computationally prohibitive.
	
	For functions characterized by singularities or discontinuities, specialized integration methods tailored to exploit knowledge of the function's properties can significantly enhance accuracy and reliability. Techniques such as adaptive mesh refinement, where the integration domain is subdivided into smaller regions around singularities or points of discontinuity, allow for more precise handling of these challenging features. Additionally, techniques like singular value decomposition or spline interpolation can be employed to model and approximate the behavior of the function in problematic regions, further improving the accuracy of the integration process.

	\paragraph{The Role of Algogenic Enhancements}
	Algogenic enhancements, which integrate generative AI into numerical integration processes, present a promising avenue for addressing both the limitations and challenges inherent in traditional numerical integration methods. By leveraging AI to adaptively select methods, manage errors, and optimize computational strategies, these enhancements have the potential to significantly improve the efficiency, accuracy, and applicability of numerical integration across a wide range of problems and disciplines.
	
	Moreover, these enhancements offer a holistic approach to numerical integration, integrating advanced AI algorithms seamlessly into the computational workflow. Furthermore, the incorporation of AI allows for real-time adaptation and refinement of integration techniques, ensuring that computational resources are utilized optimally and effectively. Additionally, the ability to dynamically manage errors through AI-driven error analysis and correction mechanisms enhances the robustness and reliability of numerical integration processes.
	
	Furthermore, the versatility of Algogenic enhancements enables their application across diverse domains, ranging from scientific research and engineering to finance and economics. Consequently, these enhancements pave the way for more accurate simulations, predictions, and analyses, facilitating informed decision-making and problem-solving in various fields.
	
	As a result, the integration of generative AI into numerical integration processes represents a paradigm shift in computational mathematics, offering unprecedented opportunities for innovation and advancement. Therefore, the adoption of Algogenic enhancements is poised to revolutionize numerical integration methodologies and redefine the landscape of computational science and engineering.

	Numerical integration's applications and limitations highlight its critical role in scientific computation and the ongoing need for innovation to extend its capabilities. The development of advanced techniques and Algogenic enhancements continues to expand the boundaries of what can be achieved, enabling more accurate and efficient solutions to the complex integration problems faced in research and industry.

	\subsubsection{Algorithmic Pseudocode for Numerical Integration}
	The Numerical Integration Algorithm is a powerful computational method utilized for approximating definite integrals. This algorithm partitions the integration interval \([a, b]\) into \(n\) equal subintervals, each with a width of \(h\). It then evaluates the function at the endpoints and midpoints of these subintervals. By summing the areas of the trapezoids formed by these points, with adjustments made for the overestimation at the boundaries by treating the first and last points as half-width trapezoids, the algorithm provides an approximation of the integral of the function over the interval \([a, b]\). This process balances computational simplicity with accuracy, making it an efficient tool for numerical integration. For further details on the operational steps of this algorithm, refer to the provided pseudocode \ref{fig:numerical-integration-pseudocode}.

	\begin{algorithm}
		\caption{Numerical Integration Using the Trapezoidal Rule}
		\begin{algorithmic}[1]
			\Procedure{NumericalIntegrationTrapezoidal}{Function, a, b, n}
			\State $h \gets (b - a) / n$
			\State $sum \gets 0.5 \times (Function(a) + Function(b))$
			\For{$i \gets 1$ to $n-1$}
			\State $x_i \gets a + i \times h$
			\State $sum \gets sum + Function(x_i)$
			\EndFor
			\State $approximation \gets h \times sum$
			\State \Return $approximation$
			\EndProcedure
		\end{algorithmic}\label{fig:numerical-integration-pseudocode}
	\end{algorithm}
	
\paragraph{Efficient Multi-dimensional Integration with Machine Learning}
The paper by Yoon (2021) \cite{yoon2021machine} discusses the application of machine learning techniques to enhance the efficiency of multi-dimensional integration. The method proposed in this study aims to reduce computational complexity and improve accuracy by integrating machine learning models into the computational framework. By adopting a data-driven approach, the study demonstrates how machine learning algorithms can be trained to predict the outcomes of integrations with precision, thereby reducing the need for extensive numerical computations. The utilization of machine learning streamlines the integration process and offers potential for handling complex multi-dimensional datasets, which can be challenging for traditional numerical methods.

\paragraph{Tensor Neural Network for Numerical Integration}
Wang et al. (2022) \cite{wang2022tensor} introduce a Tensor Neural Network (TNN) framework for numerical integration. This research explores the intersection of deep learning and numerical analysis by developing a neural network architecture tailored for numerical integration tasks. The TNN approach utilizes tensor operations and deep learning to approximate integrals accurately, especially in scenarios with high dimensionality or complex domain geometries. The proposed framework enhances the precision of numerical integration and reduces computational costs, making it applicable to various fields such as physics and engineering.

\paragraph{Bayesian Numerical Integration with Neural Networks}
Ott et al. present a methodology for Bayesian numerical integration utilizing neural networks \cite{ottbayesian}. This approach incorporates Bayesian inference principles into the neural network framework to achieve numerical integration. By integrating Bayesian techniques with neural networks, the method estimates the uncertainty of integration outcomes, providing precise numerical results along with confidence measures. This methodology adapts its performance as more data becomes available and is beneficial for dealing with complex, high-dimensional integrals in scientific and engineering applications. The integration of Bayesian statistics and neural networks offers potential for developing uncertainty-aware numerical integration techniques.

	\subsection{Algogenic Enhancements for Numerical Integration}
	
	\subsubsection{Adaptive Integration Scheme Selection}
	\paragraph{Utilizing AI for Scheme Selection}
	The integration of AI into selecting numerical integration methods for specific functions enhances accuracy and efficiency. By analyzing characteristics such as smoothness and periodicity, AI can recommend optimal methods like the Trapezoidal Rule or Gaussian Quadrature. This AI-driven approach ensures that the method aligns with the function's requirements, considering factors like desired accuracy and computational resources. It dynamically adjusts to the function's behavior, optimizing the integration process by matching it with the most suitable technique.
	
	\paragraph{Dynamic Adaptation Based on Function Analysis}
	AI-driven dynamic adaptation in numerical integration involves partitioning the domain based on the function's varying characteristics. This method tailors the integration approach within each segment, employing suitable techniques for different behaviors, such as oscillations or discontinuities. This strategy ensures optimal accuracy and efficiency by adapting the integration method to the local behavior of the function, offering a nuanced solution to complex integration tasks.
	
	\paragraph{Mathematical Basis for Scheme Selection}
	AI's scheme selection is grounded in mathematical analysis, considering factors like derivative magnitudes and inflection points. This analysis enables AI to pair the function with an integration method minimizing expected error. By leveraging insights into the function's behavior, AI optimizes the selection process, enhancing the accuracy of numerical integration across various applications.
	
	\paragraph{Enhancing Computational Efficiency}
	Adaptive scheme selection by AI not only improves accuracy but also boosts computational efficiency by choosing the most appropriate method for each task. This approach minimizes unnecessary computations, particularly in resource-constrained scenarios, and allows for tailored computation strategies that align with the function's complexity and integration requirements.
	
	\paragraph{Implementation Considerations}
	Implementing adaptive integration schemes entails seamless AI integration with numerical software, allowing real-time method selection and adjustment. This process requires efficient communication protocols and computational resource management to ensure the AI's recommendations enhance the integration process without imposing excessive computational costs.
	
	\subsubsection{Dynamic Subdivision of Integration Domains}
	\paragraph{Adaptive Domain Partitioning via Generative Models}
	LLMs facilitate adaptive domain partitioning in numerical integration, enhancing accuracy and computational efficiency. By analyzing the function's behavior, AI dynamically partitions the domain, focusing on complex regions for improved precision. This approach efficiently allocates computational resources, optimizing the integration process across varying function behaviors.
	
	\paragraph{Analyzing Function Characteristics for Subdivision}
	AI meticulously examines the function's characteristics to guide domain subdivision. By assessing derivatives and critical points, AI identifies regions needing finer subdivision, optimizing accuracy while maintaining computational efficiency. This adaptive strategy ensures precise integration across diverse function behaviors and domains.
	
	\paragraph{Mathematical Framework for Subdivision}
	The mathematical framework for dynamic subdivision relies on local error estimations, guiding AI in optimizing the partitioning process. By calculating error based on function derivatives and segment size, AI minimizes overall integration error, enhancing both accuracy and efficiency in numerical integration algorithms.
	
	\paragraph{Enhancing Accuracy Through Intelligent Partitioning}
	Intelligent partitioning, guided by AI, significantly improves numerical integration accuracy. By allocating computational resources based on the function's local behavior, this strategy captures complex features effectively, optimizing integration outcomes and enabling precise solutions to challenging problems.
	
	\paragraph{Implementation and Computational Considerations}
	Implementing dynamic subdivision requires integrating AI analysis capabilities into the integration process, balancing adaptive partitioning benefits with computational overhead. This approach enhances the accuracy and efficiency of numerical integration, addressing complex integration challenges through sophisticated AI-driven strategies.
	
	\subsubsection{Error Estimation and Correction}
	\paragraph{Predictive Error Analysis with AI}
	AI-driven predictive error analysis revolutionizes error estimation in numerical integration, enabling real-time adjustments for accuracy enhancement. By learning from extensive datasets, AI identifies error patterns, facilitating proactive correction strategies. This approach improves efficiency by reducing trial-and-error iterations, advancing numerical integration techniques.
	
	\paragraph{Real-time Parameter Adjustment for Error Minimization}
	AI dynamically adjusts integration parameters to minimize errors, enhancing algorithm accuracy and responsiveness. By tailoring step size and subdivisions based on the function's behavior, AI optimizes integration performance, ensuring precise and efficient computation across various applications.
	
	\paragraph{Mathematical Underpinnings of AI-driven Error Estimation}
	AI-driven error estimation is rooted in mathematical principles, leveraging function derivatives to predict errors accurately. This mathematical approach enables AI to refine integration algorithms, optimizing accuracy and efficiency in computational tasks across scientific and engineering domains.
	
	\paragraph{Enhancing Integration Outcomes Through Error Correction}
	LLMs play a crucial role in error correction within numerical integration, recommending adjustments to improve accuracy. This iterative refinement process, informed by AI analysis, enhances the reliability of numerical integration outcomes, offering precise solutions to complex computational problems.
	
	\paragraph{Challenges and Implementation Strategies}
	Implementing AI-driven error estimation and correction involves addressing challenges such as model training and computational efficiency. By developing modular frameworks and optimizing AI models, researchers can integrate AI recommendations effectively, enhancing numerical integration processes across various applications.
	
	\subsubsection{Real-time Parameter Optimization}
	\paragraph{Adaptive Adjustment of Integration Parameters}
	AI-driven real-time parameter optimization in numerical integration enhances efficiency and accuracy by dynamically adjusting critical parameters based on real-time feedback. This approach allows for adaptive computation, optimizing performance and enabling exploration of complex integration problems.
	
	\paragraph{AI-driven Strategies for Parameter Selection}
	AI develops sophisticated strategies for parameter selection, tailoring integration approaches to function characteristics. This adaptive selection process improves numerical integration accuracy and efficiency, demonstrating AI's potential to enhance traditional computational techniques.
	
	\paragraph{Mathematical Basis for Optimization Decisions}
	AI's optimization decisions are based on mathematical analyses aimed at balancing accuracy with computational efficiency. By managing numerical errors and exploring optimization strategies, AI enhances numerical integration algorithms, improving their performance in various computational tasks.
	
	\paragraph{Enhancing Convergence and Computational Efficiency}
	AI optimizes integration parameters in real-time, improving convergence and computational efficiency. This dynamic adjustment strategy, supported by parallel processing capabilities, accelerates integration processes and enables efficient exploration of complex mathematical problems.
	
	\paragraph{Implementation Considerations and Challenges}
	Implementing AI-driven real-time parameter optimization requires careful consideration of model complexity and computational efficiency. By addressing these challenges with a multidisciplinary approach, researchers can leverage AI to refine numerical integration techniques, enhancing their applicability in scientific computing.

	\subsubsection{Pseudocode for Algogenic Numerical Integration}
	The Algogenic numerical integration approach harnesses AI to enhance conventional numerical integration methods by dynamically adjusting integration parameters and strategies based on the observed behavior of the system and real-time error estimates. This pseudocode, available in \ref{fig:numerical-integration-Algogen-pseudocode}, outlines an advanced framework incorporating AI-driven enhancements for adaptive step size control, function evaluation, convergence criteria, and real-time parameter optimization.
	
	\begin{algorithm}
		\caption{Algogenic Numerical Integration Pseudocode}
		\begin{algorithmic}[1]
			\Procedure{AlgogenicNumericalIntegration}{Function, Domain, ErrorThreshold}
			
			\Comment{Preprocessing Phase}
			\State DecomposeProblem(Function, Domain) \Comment{AI-driven decomposition}
			\State SchemeSelection $\gets$ SelectIntegrationScheme(Function, Domain)
			
			\Comment{Core Computation Phase}
			\While{not Converged and Error > ErrorThreshold}
			\State IntegrationMethod $\gets$ EvaluateScheme(SchemeSelection)
			\State Precision $\gets$ AdjustPrecision(Function, Domain, SchemeSelection)
			\State Result $\gets$ Integrate(Function, Domain, IntegrationMethod, Precision)
			\State Error $\gets$ EstimateError(Result, Function, Domain)
			\If{Error > ErrorThreshold}
			\State SchemeSelection $\gets$ AdaptScheme(Function, Domain, Error)
			\State Domain $\gets$ SubdivideDomain(Function, Domain, Error)
			\EndIf
			\EndWhile
			
			\Comment{Postprocessing Phase}
			\State InterpretResults(Result, Function, Domain) \Comment{AI-driven interpretation}
			\State UpdateLearningLoop(Function, Domain, Result, Error) \Comment{Feedback to AI}
			
			\EndProcedure
		\end{algorithmic}\label{fig:numerical-integration-Algogen-pseudocode}
	\end{algorithm}

	\begin{figure}
		\centering
		\includegraphics[width=0.9\textwidth]{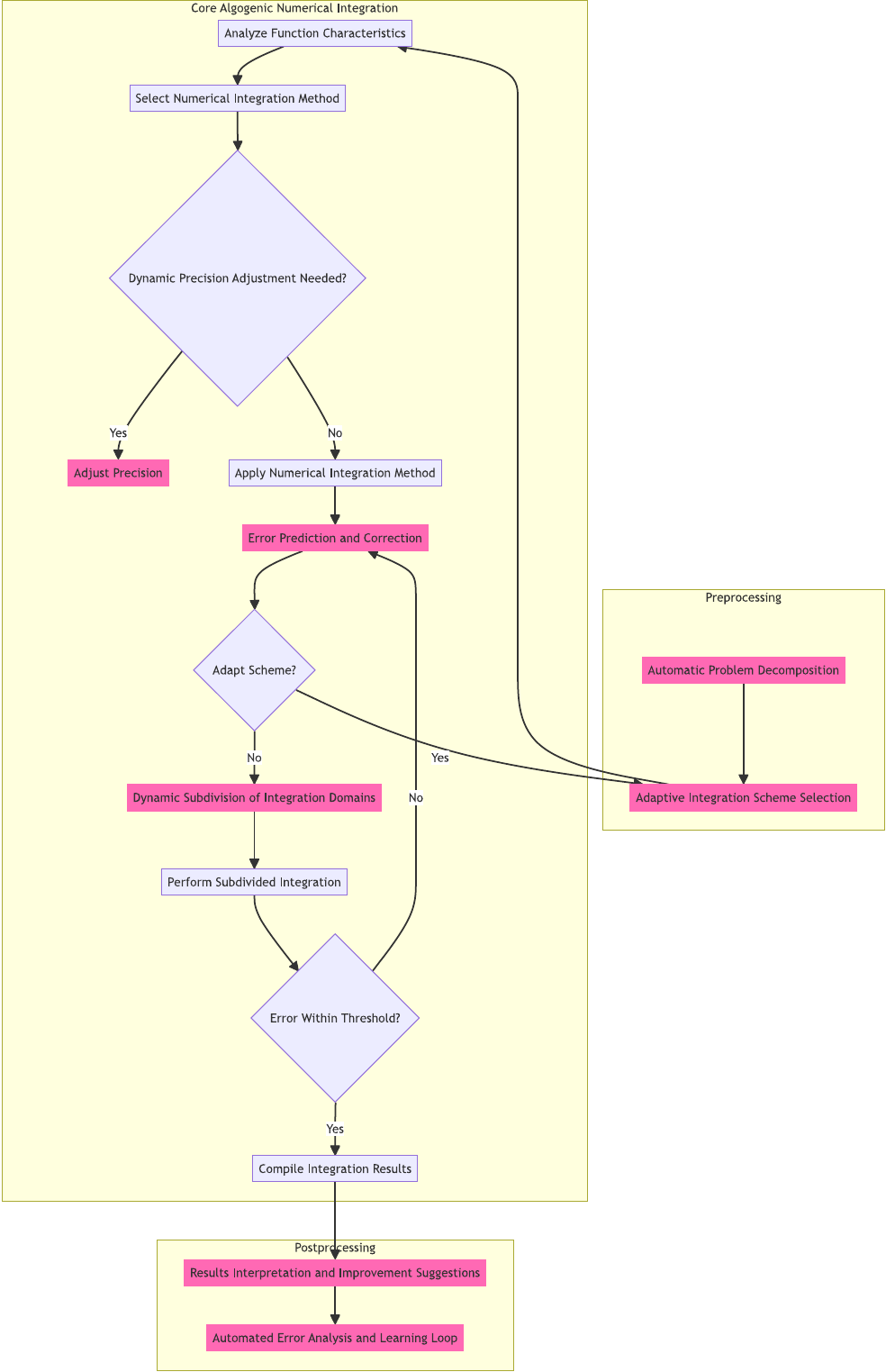}
		\caption{Integration of Algogenic Enhancements with Numerical Integration: This figure conceptualizes the Algogenic Numerical Integration algorithm, highlighting the seamless interplay between AI-driven enhancements and traditional numerical methods. Key phases include AI-driven problem decomposition in preprocessing, adaptive integration scheme selection, dynamic precision adjustment, and error prediction and correction during the core computation phase, followed by AI-enhanced results interpretation and automated error analysis in postprocessing. Each step underscores the use of generative AI to optimize the integration process, improve accuracy, and enhance computational efficiency, showcasing a sophisticated framework for tackling numerical integration challenges.}
		\label{fig:numerical_integration}
	\end{figure}

	\section{Solving Differential Equations}\index{Solving Differential Equations}
	\subsection{Introduction to Solving Differential Equations}
	\subsubsection{The Concept of Differential Equations}
	\paragraph{Fundamental Overview}
	Differential equations constitute a major area of mathematical modeling, providing a framework for describing the change in physical quantities over time or space. They are equations that relate a function with one or more of its derivatives, capturing the rates at which these quantities change. These equations serve as fundamental tools across various fields such as physics, engineering, economics, and biology, enabling the analysis and prediction of dynamic systems' behavior. 
	
	Ordinary differential equations (ODEs) form a cornerstone in modeling systems with a single independent variable. These equations commonly arise when dealing with phenomena evolving in time, such as population dynamics, chemical reactions, and electrical circuits. By expressing how a function and its derivatives interrelate, ODEs provide insights into the evolution of dynamic systems over time intervals. In contrast, partial differential equations (PDEs) extend this framework to systems with multiple independent variables, often representing phenomena evolving in both space and time. 
	
	PDEs find widespread application in fields like fluid dynamics, heat transfer, and quantum mechanics, where understanding how quantities vary across different dimensions is crucial. Unlike ODEs, which focus on the time evolution of a system, PDEs capture spatial variations alongside temporal changes, offering a comprehensive description of diverse physical processes. The solutions to these equations, whether numerical or analytical, provide valuable insights into the behavior of complex systems, aiding in decision-making, design, and optimization processes.
	
	In summary, differential equations serve as a fundamental language for describing the dynamics of natural and engineered systems, bridging theoretical concepts with real-world observations. Whether through ODEs or PDEs, these equations offer powerful tools for modeling, analyzing, and understanding a wide range of phenomena across various scientific and engineering disciplines.
	
	\paragraph{Representation and Solution}
	Mathematically, an ODE is typically represented as $\frac{dy}{dx} = f(x, y)$, where $y$ is the dependent variable, $x$ is the independent variable, and $f(x, y)$ is a function describing the rate of change of $y$ with respect to $x$. The goal is to find a function $y(x)$ that satisfies this equation for given initial conditions or boundary values. Solutions to differential equations can be explicit functions, implicit functions, or series expansions, and they provide crucial insights into the behavior and characteristics of the system being modeled.
	
	In addition, understanding the behavior of solutions often involves analyzing the stability, existence, and uniqueness of solutions. Stability analysis determines whether small perturbations in the initial conditions lead to bounded or unbounded changes in the solution over time. Moreover, the existence and uniqueness theorem guarantees that under certain conditions, there exists a unique solution that satisfies the given initial or boundary conditions. These theorems are fundamental in ensuring the reliability and predictability of the solutions obtained from differential equations.
	
	Furthermore, differential equations play a pivotal role in various fields such as physics, engineering, biology, and economics. They are used to model a wide range of phenomena including population dynamics, heat transfer, fluid flow, electrical circuits, and quantum mechanics. By solving differential equations, researchers and engineers gain valuable insights into the underlying mechanisms governing these phenomena, enabling them to make informed decisions and design efficient systems.
	
	Moreover, the study of differential equations often involves numerical methods for approximating solutions when analytical solutions are not feasible. Numerical techniques such as Euler's method, Runge-Kutta methods, and finite element methods allow for the approximation of solutions with a desired level of accuracy. These numerical approaches are indispensable in practical applications where exact solutions are elusive or computationally expensive to obtain.
	
	Thus, the representation and solution of ordinary differential equations are essential tools in scientific and engineering disciplines, facilitating the understanding, analysis, and prediction of various natural and artificial systems.

	\paragraph{Role in Modeling Natural Phenomena}
	Differential equations serve as the cornerstone in elucidating and predicting the intricate behaviors exhibited by a diverse range of natural occurrences. Through their application, scientists and engineers delve into understanding celestial mechanics, where they decipher the orbital paths of planets and moons by formulating equations that govern their motion under the influence of gravitational forces. These equations not only allow us to trace the trajectories of celestial bodies but also enable precise predictions of astronomical events, such as eclipses and planetary conjunctions, essential for space exploration and celestial navigation.
	
	Moreover, in the realm of fluid dynamics, differential equations provide a framework for analyzing the flow of liquids and gases. They help in designing efficient transportation systems, optimizing the performance of engines, and predicting weather patterns by modeling atmospheric phenomena like wind currents and oceanic tides. Additionally, in the study of population dynamics, these equations facilitate the formulation of models that elucidate the growth and interactions of various species within ecosystems. By simulating scenarios of population growth, researchers gain insights into biodiversity, resource management, and the impacts of human activities on the environment.
	
	Furthermore, in epidemiology, differential equations play a crucial role in understanding and mitigating the spread of infectious diseases. By modeling the dynamics of transmission within populations, scientists can assess the effectiveness of different intervention strategies, such as vaccination campaigns and social distancing measures. This enables policymakers to make informed decisions to curb outbreaks and protect public health.
	
	In essence, the versatility and efficacy of differential equations in capturing the underlying mechanisms governing natural phenomena empower scientists and engineers to advance knowledge, innovate technologies, and address pressing challenges across various disciplines.

	\subsubsection{Key Principles and Mechanisms}
	\paragraph{Analytical vs. Numerical Solutions}
	While some differential equations can be solved analytically, providing exact solutions, many practical problems involve equations that are too complex for analytical solutions. \textbf{However}, in such cases, numerical methods offer a powerful alternative. These methods, including Euler's method, Runge-Kutta methods, and finite difference methods for PDEs, \textbf{rely on discretizing the equations and iteratively solving them over small increments}. This iterative nature enables numerical methods to handle complex equations that defy analytical solution. \textbf{Moreover}, numerical solutions are often more flexible in handling boundary conditions and irregular geometries compared to analytical solutions. 
	
	\textbf{Additionally}, numerical methods allow for the incorporation of various physical phenomena into the model, such as nonlinearities or variable coefficients, which may not be feasible with analytical methods. \textbf{Furthermore}, these methods provide a systematic approach to deal with large-scale systems, where analytical solutions may become computationally prohibitive. \textbf{On the other hand}, while numerical solutions offer versatility and efficiency, they come with their own set of challenges, such as numerical stability, convergence issues, and discretization errors, which must be carefully addressed to ensure accuracy. 
	
	In summary, \textbf{although} analytical solutions remain invaluable for understanding fundamental properties of differential equations, numerical methods \textbf{offer a practical means} to tackle real-world problems, \textbf{bridging} the gap between theory and application.

	\paragraph{Discretization and Integration}
	The process of solving differential equations numerically often involves discretizing the time or space over which the equation is defined. This allows us to transform continuous problems into discrete ones that can be handled computationally. For time-dependent problems, discretization entails breaking the time interval into small steps, often of equal length, using techniques like the Euler method or higher-order methods such as the Runge-Kutta methods. Each time step represents a snapshot of the system's state, and by iteratively applying the numerical scheme, we can approximate how the solution evolves over time.
	
	In spatial problems, especially those described by partial differential equations (PDEs), discretization typically involves partitioning the domain into a grid or mesh. This grid may be structured, like a uniform Cartesian grid, or unstructured, such as a triangular or tetrahedral mesh. At each grid point or element of the mesh, the differential equation is approximated using finite difference, finite volume, or finite element methods. This approximation allows us to represent the continuous PDE as a system of algebraic equations, which can be solved using iterative techniques like Gauss-Seidel or conjugate gradient methods.
	
	Moreover, the choice of discretization scheme impacts the accuracy, stability, and computational efficiency of the numerical solution. For instance, finer discretizations in time or space generally lead to more accurate results but may require more computational resources. Conversely, coarser discretizations can reduce computational costs but may sacrifice accuracy. Therefore, there's often a trade-off between accuracy and computational efficiency, and the selection of an appropriate discretization strategy depends on the specific problem and computational resources available.

	\paragraph{Convergence and Stability}
	Key considerations in the numerical solution of differential equations include convergence — ensuring that the solution approximates the true solution as the step size or mesh is refined — and stability, particularly for stiff equations where certain numerical methods can produce erroneous results. The choice of numerical method and its parameters must be carefully managed to balance accuracy, computational efficiency, and stability.
	
	Convergence, \textbf{however}, poses a significant challenge in numerical methods. While reducing the step size or refining the mesh often improves accuracy, it also increases computational costs. \textbf{Furthermore}, for certain differential equations, such as those with rapidly oscillating solutions or steep gradients, achieving convergence can be particularly demanding. In such cases, specialized methods may be required, \textbf{such as} adaptive step-size control or high-order numerical schemes.
	
	Stability, on the other hand, is crucial for the reliability of numerical solutions. \textbf{Moreover}, stiff equations present a formidable obstacle as they can lead to numerical instability with certain methods. For instance, explicit methods may exhibit stability issues when applied to stiff problems, necessitating the use of implicit schemes which \textbf{additionally} demand higher computational costs per time step. \textbf{Furthermore}, the stiffness of the problem often dictates the choice of time-stepping methods; for mildly stiff equations, implicit methods might suffice, whereas highly stiff problems may require implicit methods with stiffness detection and adaptation mechanisms.
	
	In practice, achieving a balance between accuracy, efficiency, and stability is a multifaceted task. \textbf{Therefore}, researchers and practitioners frequently resort to a combination of strategies, such as utilizing adaptive algorithms that dynamically adjust the step size based on local error estimates and employing hybrid methods that switch between different numerical schemes depending on the characteristics of the problem. \textbf{Thus}, while ensuring convergence and stability remains paramount, the overarching goal is to develop robust numerical techniques capable of efficiently solving a wide range of differential equations.

	\subsubsection{The Role of Numerical Methods}
	\paragraph{Bridging Theory with Computation}
	Numerical methods for differential equations bridge the gap between theoretical models and practical computational solutions. They provide algorithms for approximating the solutions of differential equations to any desired level of accuracy, subject to computational constraints. These methods transform the continuous problem of solving a differential equation into a discrete problem that can be handled by digital computers, enabling the analysis and simulation of complex systems that are intractable analytically.
	
	Furthermore, these numerical techniques play a pivotal role in various scientific and engineering disciplines. They allow researchers and engineers to tackle real-world problems where analytical solutions are elusive or impractical. Moreover, numerical methods facilitate sensitivity analysis and parameter estimation, essential tasks in understanding the behavior of dynamical systems. Additionally, they empower the exploration of multi-dimensional spaces and the study of phenomena that involve intricate interactions between numerous variables.
	
	On the other hand, it's crucial to acknowledge the inherent limitations and challenges associated with numerical methods. While they offer practical solutions, numerical approximations can introduce errors, especially when dealing with stiff equations or discontinuous solutions. Nonetheless, through careful consideration of numerical stability, convergence, and discretization strategies, these challenges can be mitigated to a large extent.
	
	In summary, the symbiotic relationship between theoretical models and computational methods is central to advancing scientific understanding and technological innovation. By leveraging numerical techniques, researchers can unlock new insights into complex phenomena, driving progress across a spectrum of fields, from physics and chemistry to biology and engineering.

	\paragraph{Diversity of Numerical Techniques}
	The diversity of numerical techniques reflects the wide array of differential equations encountered in various fields and their distinct applications. When tackling ordinary differential equations (ODEs), practitioners confront a spectrum of methods tailored to suit different complexities and computational requirements. From the elementary yet robust Euler's method to the adaptive precision of methods like the Runge-Kutta-Fehlberg scheme, the toolbox is rich with options. These methods exhibit varying behaviors in terms of stability, accuracy, and computational cost, providing analysts with flexibility in choosing the most suitable approach depending on the problem at hand.
	
	In the realm of partial differential equations (PDEs), the landscape expands further, encompassing finite difference, finite element, and spectral methods. Each method offers a unique perspective on discretizing and solving PDEs, catering to diverse problem domains and computational resources. Finite difference methods, for instance, discretize the spatial domain into a grid, transforming PDEs into a system of algebraic equations. Conversely, finite element methods decompose the domain into smaller, more manageable elements, offering flexibility in handling irregular geometries and boundary conditions. Spectral methods, on the other hand, exploit the inherent periodicity of functions, representing solutions as a sum of basis functions with coefficients determined through spectral techniques.
	
	The choice among these techniques hinges on several factors, including the nature of the equation (e.g., linearity, order), boundary conditions, and the desired level of accuracy. Furthermore, considerations such as computational efficiency and ease of implementation play crucial roles, especially in large-scale simulations where resource optimization is paramount. Thus, while Euler's method may suffice for simple ODEs with low accuracy requirements, more sophisticated methods like the Runge-Kutta-Fehlberg scheme or spectral methods become indispensable for tackling complex problems demanding high precision and computational fidelity.

	\paragraph{Implementation and Software}
	Numerous software libraries and packages have been developed to implement these numerical methods, providing researchers and practitioners with powerful tools for solving differential equations. These tools range from general-purpose mathematical software to specialized libraries focused on particular types of equations or numerical methods, facilitating the modeling, simulation, and analysis of complex systems across scientific and engineering disciplines.
	
	Furthermore, advancements in computational techniques have led to the creation of user-friendly interfaces and integrated development environments (IDEs), making it easier for both experts and novices to utilize these tools effectively. Moreover, collaborative platforms and open-source communities play a significant role in the continuous improvement and dissemination of these software solutions, fostering innovation and knowledge-sharing within the scientific community. Additionally, the integration of parallel computing and distributed systems has enabled the efficient execution of computationally intensive simulations, thereby accelerating the pace of research and development in various fields.
	
	On the other hand, despite the availability of sophisticated software, challenges such as numerical instability, convergence issues, and computational overhead persist, requiring careful consideration and expertise during the implementation and utilization of these numerical methods. Nevertheless, ongoing research efforts aim to address these challenges through algorithmic improvements, optimization techniques, and error analysis, thereby enhancing the reliability and efficiency of numerical simulations.
	
	In contrast, traditional analytical methods may face limitations in handling complex systems with nonlinear dynamics or irregular geometries, highlighting the importance of numerical techniques in tackling real-world problems. Moreover, the flexibility and scalability offered by numerical approaches allow for the exploration of diverse phenomena and scenarios, enabling researchers to gain insights into the behavior of complex systems under varying conditions.
	
	Overall, the availability and advancement of software tools for implementing numerical methods have revolutionized the field of computational science and engineering, empowering researchers to tackle increasingly complex problems and push the boundaries of knowledge and innovation.

	\subsubsection{Key Principles and Mechanisms}
	\paragraph{Solving Strategies and Their Foundations}
	The core objective in solving differential equations, whether ordinary (ODEs) or partial (PDEs), is to find a function or a set of functions that satisfy the given equations under specific initial or boundary conditions. The complexity of these equations and their conditions dictates the choice of solving strategy. Analytical methods aim to find exact solutions and are suitable for simpler equations where such solutions exist. \textbf{Moreover}, these methods leverage a deep understanding of mathematical functions and their properties, applying techniques like separation of variables, integrating factors, and characteristic equations.
	
	Analytical methods provide a rigorous framework for solving differential equations. \textbf{Furthermore}, they often rely on well-established mathematical principles, such as the existence and uniqueness theorem, which guarantees the existence of solutions under certain conditions. \textbf{In addition}, these methods offer insights into the behavior of solutions, particularly through the study of special functions like Bessel functions, Legendre polynomials, and hypergeometric functions.
	
	However, analytical methods have limitations. \textbf{On the other hand}, for nonlinear or complex equations, exact analytical solutions may be elusive or non-existent. \textbf{Instead}, numerical methods become indispensable in such cases. Numerical techniques, such as finite difference methods, finite element methods, and numerical integration, approximate solutions by discretizing the problem domain and employing iterative algorithms. These methods excel in handling nonlinearities, discontinuities, and high-dimensional systems.
	
	In conclusion, while analytical methods provide elegant solutions for certain classes of differential equations, numerical methods offer versatility and efficiency in tackling more challenging problems. A comprehensive understanding of both analytical and numerical techniques equips mathematicians and scientists with powerful tools to address a wide range of differential equations encountered in various fields of study.

	\paragraph{Numerical Approaches for Complex Equations}
	When analytical solutions are not feasible due to the complexity of the equations or the conditions imposed, numerical methods provide an alternative by approximating the solution at discrete points. These methods include finite difference methods, which approximate derivatives using differences between function values at adjacent points; finite element methods, which divide the domain into smaller, simpler regions (elements) and approximate the solution piecewise; and spectral methods, which approximate the solution as a sum of basis functions, typically chosen for their advantageous properties in the Fourier or polynomial domains.
	
	Additionally, when dealing with highly nonlinear systems or systems with irregular geometries, numerical approaches offer a practical solution. While analytical methods often struggle with such complexities, numerical techniques excel in handling them by discretizing the problem domain and iteratively solving the resulting system of equations.
	
	Moreover, numerical methods allow for the incorporation of boundary conditions and other constraints seamlessly, enabling the investigation of a wide range of real-world problems in various fields such as fluid dynamics, structural mechanics, electromagnetics, and finance.
	
	Furthermore, these methods facilitate the implementation of algorithms that can efficiently handle large-scale computations, leveraging advancements in computational hardware and parallel processing architectures. Consequently, numerical approaches have become indispensable tools for researchers and engineers seeking to tackle intricate mathematical models and simulate complex phenomena accurately.

	\paragraph{Initial and Boundary Conditions}
	The solution to a differential equation is significantly influenced by initial and boundary conditions, which ground the solution in physical or practical reality. Initial conditions specify the state of the system at the beginning of the observation period, primarily used in ODEs. Boundary conditions define the behavior of the solution at the boundaries of the domain of interest, critical for PDEs. These conditions ensure that the solution not only satisfies the differential equation but also aligns with the specific scenario being modeled.
	
	Moreover, initial conditions serve as starting points for the solution trajectory, offering a snapshot of the system's state at a particular moment. They provide essential information for solving the differential equation by determining the arbitrary constants present in the general solution. Likewise, boundary conditions set constraints on the solution's behavior, ensuring its compatibility with the physical environment or the problem's setup.
	
	Furthermore, initial conditions often reflect the system's history or prior knowledge, capturing the dynamics from which the system evolves. Conversely, boundary conditions encapsulate external influences or constraints imposed on the system's behavior at its spatial or temporal limits. In contrast to initial conditions, which typically involve specifying values for dependent variables at a single point or within a small interval, boundary conditions extend across the entire boundary of the domain, influencing the solution's behavior across its entirety.
	
	Therefore, both initial and boundary conditions play indispensable roles in shaping the solution to a differential equation, anchoring it within the context of the problem being analyzed and ensuring its validity and relevance.

	\paragraph{Finite Difference and Finite Element Methods}
	The finite difference method (FDM) and the finite element method (FEM) are two widely used numerical techniques for solving partial differential equations (PDEs) in various scientific and engineering applications.
	
	FDM discretizes the computational domain into a grid, where each grid point represents a discrete location in space or time. Differential operators in the PDE are approximated as finite differences between neighboring grid points. This approach simplifies the PDE into a system of algebraic equations, making it well-suited for problems with structured domains or uniform geometries. And, despite its simplicity, FDM can accurately capture the behavior of the solution in many cases, especially when the problem exhibits smooth variations.
	
	On the contrary, FEM employs a different discretization strategy by dividing the domain into smaller, geometrically simple elements, such as triangles or quadrilaterals in 2D or tetrahedra or hexahedra in 3D. Within each element, the solution is approximated using interpolation functions, also known as basis functions, which are typically piecewise defined over the element. This allows FEM to handle complex geometries and irregular domains more effectively. Furthermore, FEM provides a flexible framework for refining the mesh in regions where the solution varies rapidly or requires higher accuracy.
	
	Moreover, FEM naturally accommodates different boundary conditions by incorporating them directly into the variational formulation of the problem. This property makes FEM particularly versatile for problems with diverse boundary conditions or where boundary effects play a crucial role in the solution behavior.
	
	In conclusion, while FDM is advantageous for problems with regular geometries and uniform discretizations, FEM excels in handling complex geometries and irregular domains, offering greater flexibility and accuracy in approximating solutions to PDEs.

	\paragraph{Spectral Methods for High Accuracy}
	Spectral methods stand out in the realm of numerical analysis for their remarkable capacity to provide high accuracy in solving problems characterized by a series representation in orthogonal basis functions. This representation often manifests in the form of sine and cosine functions within Fourier series or as polynomials within Chebyshev series. The essence of spectral methods lies in their adeptness at handling smooth problems involving high-order derivatives.
	
	By leveraging orthogonal basis functions, spectral methods excel in capturing intricate details of the solution, especially in scenarios where traditional numerical techniques struggle. This proficiency becomes particularly pronounced in problems necessitating the calculation of high-order derivatives, where spectral methods showcase their prowess. Through the judicious selection of appropriate basis functions, spectral methods can achieve exponential convergence rates, a feat that significantly surpasses the capabilities of conventional numerical approaches.
	
	The efficacy of spectral methods becomes apparent when considering the accuracy-computational effort trade-off. While some numerical methods may require substantial computational resources to achieve a desired level of accuracy, spectral methods often outshine them by offering superior accuracy per unit of computational effort. This superiority makes spectral methods an attractive choice for engineers and scientists grappling with computationally demanding problems where precision is paramount.
	
	In essence, spectral methods represent a cornerstone in numerical analysis, particularly for problems characterized by smoothness and where high-order derivatives play a pivotal role. Their ability to harness the power of orthogonal basis functions and achieve exponential convergence rates underscores their importance in various fields of science and engineering.

	\paragraph{Selection Based on Problem Characteristics}
	The choice among these techniques is dictated by the specific characteristics of the differential equation and the problem domain, including the equation's linearity, the nature of the domain, and the desired accuracy and computational resources available. Numerical methods offer a spectrum of approaches tailored to different scenarios. \textit{For} linear differential equations, methods like finite difference and finite element are often preferred due to their ability to efficiently handle linearity while preserving accuracy. \textit{However}, for nonlinear equations, implicit methods such as the backward Euler method may be more suitable, \textit{since} they can better capture complex dynamics and avoid instability issues associated with explicit schemes. \textit{Moreover}, in domains with irregular boundaries or complex geometries, finite element methods shine, \textit{since} they allow for flexible meshing strategies and accurate representation of the domain. \textit{Additionally}, when high accuracy is paramount, spectral methods \textit{alongside} finite difference or finite element techniques can be employed, as they excel in capturing rapid variations in the solution. \textit{On the other hand}, in situations where computational resources are limited, simpler methods like finite difference \textit{or} finite element may be favored \textit{since} they offer a balance between accuracy and computational cost. \textit{In contrast}, when computational resources are abundant, more computationally intensive methods like spectral methods \textit{or} high-order finite element methods \textit{may} be \textit{chosen} \textit{as} they promise greater accuracy. \textit{Nevertheless}, regardless of the method chosen, careful consideration of the problem characteristics \textit{is} necessary \textit{in order to} ensure the accuracy and efficiency of the numerical solution.

	\subsubsection{The Role of Numerical Methods}
	\paragraph{Bridging the Gap to Practical Solutions}
	Numerical methods play a pivotal role in translating the abstract constructs of differential equations into tangible solutions essential for real-world applications. While the elegance of theoretical models lies in their ability to encapsulate physical phenomena within mathematical frameworks, the complexities inherent in many systems often render direct analytical solutions unattainable. Here, numerical methods step in as indispensable tools, offering a pragmatic pathway to address the inherent challenges.
	
	In instances where differential equations defy straightforward analytical resolution—be it owing to the intricate interplay of nonlinear terms, the multidimensionality of the system, or the imposition of intricate boundary and initial conditions—numerical methods shine. By discretizing the solution space and approximating values at discrete points, these methods empower researchers and engineers to traverse the intricate landscapes of complex systems. Whether navigating the turbulent dynamics of fluid flow, elucidating the intricate behaviors of quantum mechanical systems, or modeling the complex interactions within biological systems, numerical methods furnish a versatile toolkit for scientific inquiry and engineering innovation.
	
	Moreover, numerical methods afford flexibility in exploring diverse scenarios and parameter spaces, enabling a comprehensive understanding of system behavior under varied conditions. Through iterative refinement and convergence, numerical techniques facilitate the refinement of models, fostering deeper insights and more accurate predictions. Consequently, they serve as indispensable enablers, facilitating the translation of abstract theoretical frameworks into actionable insights and practical solutions.
	
	In essence, numerical methods stand as the linchpin bridging the chasm between theoretical abstraction and practical exigency. By harnessing the power of computation to navigate the complexities of real-world phenomena, these methods pave the way for transformative advancements across a myriad of disciplines, driving innovation and progress in science and engineering.

	\paragraph{Discretization of Continuous Problems}
	The essence of numerical methods lies in the discretization of the continuous problem posed by differential equations. This process involves transforming the continuous domain of the equation into a set of discrete points or elements and then approximating the solution's behavior at these points. \textbf{Moreover}, the approach taken for discretization—whether dividing the domain into a grid for finite difference methods, breaking it into elements for finite element methods, or expressing the solution in terms of a finite series of basis functions for spectral methods—depends on the specific nature of the differential equation and the domain over which it is defined.
	
	Finite difference methods, \textbf{for example}, divide the domain into a grid, where the differential equation is approximated by a set of finite difference equations relating values of the function at neighboring grid points. This approach is particularly suited for problems with regular geometries and well-behaved derivatives.
	
	On the other hand, finite element methods \textbf{involve} breaking the domain into smaller elements, where the differential equation is approximately solved over each element using interpolation functions. This method offers flexibility in handling irregular geometries and complex boundary conditions.
	
	Similarly, spectral methods \textbf{rely on} expressing the solution as a weighted sum of basis functions, such as Fourier or Chebyshev polynomials. These methods excel in problems with smooth solutions and periodic boundary conditions, providing highly accurate approximations.
	
	Regardless of the specific method used, discretization enables the transformation of continuous problems into manageable discrete systems, facilitating numerical analysis and computation. \textbf{Furthermore}, it allows for the application of iterative techniques and computational algorithms to approximate solutions efficiently.
	
	In summary, the choice of discretization method depends on various factors, including the problem's geometry, boundary conditions, and the desired accuracy of the solution. Each method offers unique advantages and limitations, \textbf{and} selecting the most suitable approach is essential for obtaining accurate and efficient numerical solutions to differential equations.

	\paragraph{Adaptation to Equation Characteristics}
	The selection of a numerical method is guided by the characteristics of the differential equation in question. For instance, the finite difference method may be preferred for problems with simple geometries and where high precision is not paramount. In contrast, the finite element method offers superior flexibility for complex geometries and varying material properties, making it better suited for engineering applications involving structural analysis or fluid dynamics. Spectral methods, offering high accuracy for smooth problems, are ideal for applications where the solution can be accurately captured with global basis functions.
	
	Furthermore, it's crucial to consider the computational cost associated with each method. Finite difference methods often involve discretizing the domain into a grid, which can lead to high memory consumption and computational expense, especially for problems in higher dimensions. On the other hand, finite element and spectral methods typically require solving systems of linear equations derived from the discretization process. While this can also be computationally intensive, advancements in numerical algorithms and computational hardware have made these methods more feasible for a wide range of applications.
	
	Moreover, the choice of numerical method may depend on the specific requirements of the problem at hand. For example, in time-dependent problems where stability and conservation properties are essential, implicit finite difference schemes or spectral methods with appropriate time-stepping techniques may be preferred. Additionally, the presence of boundary conditions and constraints can influence the selection of the most suitable numerical approach.
	
	In conclusion, the adaptability of numerical methods to the characteristics of the differential equation, including geometry, material properties, and solution smoothness, plays a crucial role in ensuring accurate and efficient simulations in various engineering and scientific domains.

	\paragraph{Accuracy and Computational Considerations}
	Achieving the desired level of accuracy in numerical methods is paramount, as it directly impacts the reliability and usefulness of the solutions obtained. This pursuit of accuracy, however, must often be balanced against computational considerations. While it's tempting to increase accuracy by refining the discretization, doing so incurs a proportional increase in computational resources. This trade-off underscores the importance of judiciously selecting the level of accuracy necessary for the specific problem at hand, considering the available computational capabilities.
	
	Numerical methods offer various strategies to manage this balance effectively. Error estimation techniques provide insights into the accuracy of the solution, enabling practitioners to gauge whether further refinement is warranted. By assessing the magnitude of errors introduced at different levels of discretization, one can make informed decisions about allocating computational resources optimally. Furthermore, methods like adaptive refinement empower algorithms to dynamically adjust the discretization based on solution characteristics, directing computational effort where it promises the greatest improvement in accuracy.
	
	Moreover, it's essential to recognize that achieving higher accuracy doesn't always translate to better solutions, especially if the problem inherently contains uncertainties or noise. In such cases, overly precise solutions may obscure the underlying trends or introduce spurious artifacts. Therefore, practitioners must exercise discretion in pursuing accuracy, considering the broader context of the problem domain and the intended application of the numerical results.
	
	In summary, while accuracy remains a primary goal in numerical computations, it must be pursued judiciously, mindful of the computational resources available and the specific requirements of the problem. By leveraging error estimation techniques and adaptive methods, practitioners can navigate the accuracy-computational trade-off effectively, ensuring that computational efforts are focused where they yield the most meaningful improvements in solution quality.

	\paragraph{Enabling Complex Systems Analysis}
	Numerical methods have enabled the analysis of complex systems that would be intractable with analytical solutions alone. They facilitate the simulation of phenomena ranging from weather patterns and fluid flows to economic models and population dynamics. By providing a means to approximate solutions to differential equations, numerical methods have expanded the scope of problems that can be addressed scientifically, contributing profoundly to advances in technology, science, and mathematics.
	
	Moreover, the versatility of numerical methods allows researchers to tackle problems across various disciplines. For example, in meteorology, numerical weather prediction models utilize sophisticated algorithms to simulate atmospheric processes, enabling forecasters to predict weather patterns with increasing accuracy. Likewise, in fluid dynamics, computational fluid dynamics (CFD) methods employ numerical techniques to model fluid behavior in diverse scenarios, aiding in the design of aerodynamic structures for aircraft or optimizing the efficiency of industrial processes.
	
	Additionally, numerical methods have revolutionized economic analysis by enabling the development of complex macroeconomic models. These models incorporate numerous variables and interactions, providing insights into the behavior of economies under different conditions and informing policy decisions. Similarly, in ecology and epidemiology, numerical simulations of population dynamics allow scientists to study the spread of diseases, assess environmental impacts, and design conservation strategies.
	
	Furthermore, the advancements in numerical techniques have led to the emergence of high-performance computing (HPC) systems, capable of handling massive datasets and executing complex simulations in parallel. This synergy between numerical methods and HPC has propelled research in fields such as computational finance, where intricate models require extensive computational resources for accurate predictions.
	
	In conclusion, numerical methods play a pivotal role in modern scientific inquiry, offering a powerful toolkit for understanding and analyzing complex systems across a wide range of disciplines. Their continued development and application promise to further deepen our understanding of the natural world, drive technological innovation, and address pressing societal challenges.

	\paragraph{The Continual Evolution of Numerical Techniques}
	The field of numerical methods for differential equations is one of continual innovation, with ongoing research focused on developing more efficient, accurate, and robust methods. This research is driven by the increasing complexity of the models used to describe real-world phenomena and the growing computational power available to tackle these models. As computational resources advance, \textbf{new challenges arise,} demanding novel approaches to numerical computation. \textbf{Moreover,} the interdisciplinary nature of modern scientific problems often requires methods that can seamlessly integrate with other disciplines, \textbf{thus} fostering the development of hybrid numerical techniques. \textbf{Additionally,} the advent of high-performance computing architectures \textbf{further accelerates} the pace of innovation in numerical methods, enabling the simulation of previously inaccessible phenomena and the refinement of existing models. \textbf{Furthermore,} the pursuit of more accurate results drives researchers to explore \textbf{alternative numerical formulations,} such as spectral methods or meshless techniques, which may offer advantages over traditional finite difference or finite element methods. \textbf{On the other hand,} the quest for efficiency motivates the exploration of \textbf{adaptive algorithms} that dynamically adjust the computational grid or time step to optimize accuracy while minimizing computational cost. \textbf{In contrast,} the need for robustness in the face of uncertain inputs or model parameters \textbf{prompts} the development of \textbf{stability analysis} techniques and \textbf{uncertainty quantification} methods. In summary, the continual evolution of numerical techniques \textbf{not only} addresses the increasing complexity of models and the escalating demand for computational power but also opens up \textbf{new avenues} for tackling challenging scientific problems across various disciplines.

	\subsubsection{Applications and Limitations}
	\paragraph{Wide-ranging Applications in Science and Engineering}
	Differential equations are indispensable tools in a vast array of scientific and engineering disciplines, serving as the backbone for modeling an extensive variety of dynamic systems and phenomena. In physics, they are used to describe the laws of motion, electromagnetism, and thermodynamics, providing insights into the fundamental workings of the universe. \textbf{Moreover}, engineering applications are equally diverse, ranging from the design and analysis of mechanical structures to the simulation of electrical circuits and the optimization of chemical processes. In the realm of economics and finance, differential equations model market dynamics, interest rates, and other factors critical to understanding economic behavior and forecasting. Environmental science, biology, and medicine also rely heavily on differential equations to model ecological systems, disease spread, and physiological processes. \textbf{Furthermore}, in environmental science, differential equations are instrumental in understanding the interactions between various components of ecosystems, such as predator-prey dynamics and nutrient cycling. \textbf{In addition}, in biology and medicine, these equations play a crucial role in modeling the spread of infectious diseases, predicting population dynamics, and optimizing drug dosage regimens. Therefore, the versatility and applicability of differential equations across disciplines underscore their significance in advancing scientific knowledge and technological innovation.

	\paragraph{Modeling Complex Phenomena}
	The strength of differential equations lies in their ability to model complex phenomena with remarkable precision. By incorporating time, space, and other variables into their formulations, these equations can capture the nuances of change and interaction within systems, allowing for predictions and analyses that are foundational to advances in technology, policy making, and scientific understanding.
	
	Furthermore, the versatility of differential equations enables researchers to address a wide array of phenomena across various disciplines. From describing the behavior of physical systems like fluid flow, electromagnetic fields, and mechanical vibrations to elucidating the dynamics of biological processes such as population growth, enzyme kinetics, and neural networks, these equations serve as indispensable tools for exploring the intricate workings of the natural world.
	
	Moreover, differential equations play a pivotal role in engineering applications, facilitating the design and optimization of structures, circuits, and control systems. Engineers rely on differential equation models to simulate and analyze complex systems, ensuring the safety, efficiency, and reliability of technological advancements in fields ranging from aerospace and automotive engineering to telecommunications and renewable energy.
	
	Additionally, the integration of differential equations with computational techniques has revolutionized scientific inquiry, enabling the simulation of complex phenomena that defy analytical solutions. Through numerical methods like finite element analysis, finite difference methods, and computational fluid dynamics, researchers can tackle intricate problems with unprecedented accuracy and efficiency, paving the way for groundbreaking discoveries and innovations.
	
	Hence, the interdisciplinary nature of differential equations, coupled with their computational prowess, underscores their significance in advancing knowledge and driving progress across diverse domains.

	\paragraph{Limitations in Solving Differential Equations}
	Despite their broad utility, solving differential equations, particularly complex or high-dimensional PDEs, presents significant challenges. One major limitation is the inherent complexity of the equations themselves, which may not have analytical solutions or may require assumptions that simplify the real-world systems they aim to represent. This complexity often necessitates the use of numerical methods, which, while powerful, introduce their own set of challenges. Moreover, numerical methods can be computationally intensive, requiring significant computational resources and time. Additionally, the accuracy of numerical solutions depends heavily on the choice of discretization methods, grid sizes, and other parameters, which can be non-trivial to optimize, especially for highly nonlinear or ill-conditioned problems. Furthermore, the stability and convergence of numerical schemes can be problematic, particularly when dealing with stiff systems or long-time simulations. Consequently, practitioners often face a trade-off between computational efficiency and solution accuracy, making the process of solving complex differential equations a delicate balancing act. Nevertheless, ongoing advancements in computational techniques, along with improvements in hardware capabilities, continue to push the boundaries of what is feasible in this field, offering hope for more effective solutions to these challenging problems in the future.

	\paragraph{Computational Cost and Accuracy Concerns}
	The computational cost of numerical methods can be substantial, especially for equations describing phenomena in three dimensions or involving time evolution over extended periods. High-resolution simulations require significant computational resources, which can limit the feasibility of certain studies. \textbf{Moreover}, accurately capturing boundary and initial conditions is crucial for the reliability of the solutions but can be difficult in practice, particularly for complex geometries or conditions that vary over time. \textbf{Furthermore}, the trade-off between computational cost and accuracy is often a significant consideration in numerical simulations. Researchers must carefully balance the need for higher resolution and accuracy against the available computational resources and time constraints. \textbf{Additionally}, parallel computing techniques can alleviate some of the computational burden by distributing the workload across multiple processors or nodes. However, this introduces its own challenges, such as communication overhead and load balancing, which must be carefully managed to ensure efficient utilization of resources. \textbf{On the other hand}, simplifying assumptions or coarse discretizations may reduce computational cost but can lead to loss of accuracy, especially in capturing fine-scale features or transient phenomena. Thus, there exists a complex interplay between computational cost, accuracy, and model complexity that researchers must navigate when designing numerical simulations for scientific or engineering applications.

	\paragraph{Navigating Limitations with Advanced Techniques and Technologies}
	The limitations of traditional approaches in solving differential equations have spurred the development of advanced numerical techniques and the adoption of high-performance computing technologies. \textbf{Furthermore}, methods such as adaptive mesh refinement, parallel computing, and machine learning algorithms for predicting solution behaviors offer ways to overcome computational challenges, improving the efficiency and accuracy of simulations. \textbf{Moreover,} the integration of physical insights into computational models, \textbf{such as} through techniques like physics-informed neural networks, helps ensure that solutions remain realistic and grounded in the underlying science, even as they become increasingly complex. \textbf{Additionally}, by leveraging parallel computing, simulations can be distributed across multiple processors or computing nodes, \textbf{thus} reducing the computational time required for large-scale simulations. \textbf{On the other hand}, while these advanced techniques show promise in addressing computational limitations, they also introduce new challenges such as algorithmic complexity and the need for specialized expertise in implementing and optimizing these methods. \textbf{However,} with proper training and investment in computational resources, researchers can harness the power of these advanced techniques to tackle previously intractable problems in science and engineering. \textbf{Consequently}, the intersection of advanced numerical methods and high-performance computing technologies opens up new avenues for exploration and discovery, pushing the boundaries of what is possible in computational science and engineering.

	\paragraph{The Evolving Landscape of Differential Equation Solving}
	As computational capabilities continue to grow and numerical methods advance, the landscape of differential equation solving is evolving rapidly. These developments promise to extend the reach of differential equation models, enabling more accurate, comprehensive, and computationally feasible simulations of the complex systems that characterize the natural and human-made world.
	
	Moreover, the integration of advanced algorithms with high-performance computing resources opens up new avenues for tackling previously intractable differential equations. This synergy empowers researchers and engineers to explore phenomena with unprecedented detail and fidelity, from the intricate dynamics of biological systems to the behavior of complex fluid flows in engineering applications.
	
	Furthermore, alongside the refinement of numerical methods, the parallel advancement of hardware architectures ensures that computational tasks can be executed more efficiently than ever before. Consequently, simulations that once demanded prohibitive amounts of time and computational resources can now be completed in a fraction of the time, democratizing access to sophisticated modeling and analysis tools.
	
	Additionally, as the understanding of differential equations deepens and interdisciplinary collaborations flourish, innovative approaches emerge for solving challenges across diverse domains. This interdisciplinary synergy fosters the development of hybrid models that combine differential equations with techniques from machine learning, optimization, and statistical inference, further enriching the toolbox available to researchers and practitioners.
	
	In summary, the evolution of differential equation solving is driven by a convergence of computational, algorithmic, and interdisciplinary advancements. This evolution not only enhances our ability to accurately represent and analyze complex systems but also paves the way for groundbreaking discoveries and technological innovations with far-reaching implications.

	The applications and limitations of differential equations highlight both their fundamental importance in modeling dynamic systems and the challenges inherent in their solution. Addressing these challenges through ongoing research and technological innovation remains a critical focus in the application of differential equations to real-world problems.

	\subsubsection{Algorithmic Pseudocode for Solving Ordinary Differential Equations (ODEs) Using Euler's Method}
	Euler's Method exemplifies the basic principle behind numerical ODE solving: discretizing the continuous problem and iteratively approximating the solution. This method begins with the initial conditions $(x_0, y_0)$, where $x_0$ represents the starting point, and $y_0$ is the corresponding value of the solution $y$. With a specified step size $h$ and the number of steps $n$, Euler's Method operates by iteratively updating the value of $y$. At each step, it calculates the derivative of $y$ with respect to $x$, denoted as $f(x, y)$, and approximates the value of $y$ at the next point by adding the product of the step size $h$ and the derivative $f(x, y)$. This process continues for $n$ steps or until the desired $x$ value is reached. Despite its simplicity and known limitations, such as susceptibility to significant errors with large step sizes or highly nonlinear functions, Euler's Method lays down the foundational concept of numerical integration for differential equations. It provides a piecewise linear approximation of the solution curve, serving as a fundamental building block for more advanced numerical techniques that aim for improved accuracy and stability in solving differential equations. (See pseudocode \ref{fig:euler-pseudocode} for a visual representation of the algorithm's steps.)

	\begin{algorithm}
		\caption{Euler's Method for Solving ODEs}
		\begin{algorithmic}[1]
			\Procedure{EulersMethod}{f, x0, y0, h, n}
			\State $x \gets x0$
			\State $y \gets y0$
			\For{$i \gets 1$ to $n$}
			\State $y \gets y + h \times f(x, y)$ \Comment{Update $y$ using the slope $f(x, y)$}
			\State $x \gets x + h$ \Comment{Move to the next $x$ value}
			\EndFor
			\State \Return $y$ \Comment{Approximation of $y(x)$ at $x = x0 + n \times h$}
			\EndProcedure
		\end{algorithmic}\label{fig:euler-pseudocode}
	\end{algorithm}
	
\subsection{Previous Work on ML and AI Interplay with Differential Equation Solvers}

\paragraph{Neural Ordinary Differential Equations}
The integration of neural networks with differential equations was introduced in \cite{chen2018neural}, proposing a framework for modeling dynamic systems. This framework utilizes neural networks to represent the dynamics of ordinary differential equations (ODEs), allowing for end-to-end training of both the neural network parameters and the differential equation model through backpropagation. This approach facilitates the application of deep learning in scientific computing, enabling efficient and accurate modeling of complex systems.

\paragraph{On Neural Differential Equations}
Expanding upon the foundational work on neural ODEs, \cite{kidger2022neural} critically examines theoretical and practical aspects. The paper discusses mechanisms, challenges, and potential improvements, offering insights into integrating machine learning with differential equation solving. It emphasizes adaptivity in solver selection, efficient training techniques, and the broader applicability of neural differential equations, providing directions for future research.

\paragraph{High Precision PDE Solving with Neural Networks}
In \cite{jiang2023neural}, an algorithm for solving partial differential equations (PDEs) using neural networks is presented. This algorithm achieves high precision by leveraging deep learning to approximate solutions accurately. Through a specialized network architecture, the authors significantly improve solution precision compared to traditional numerical methods, showcasing the potential of neural networks in numerical simulation.

\paragraph{Machine Learning-based Spectral Methods}
\cite{meuris2023machine} explores the application of machine learning to enhance spectral methods for PDE solving. Integrating machine learning with spectral methods improves solution efficiency and accuracy. The hybrid method combines the high-resolution and convergence properties of spectral methods with the adaptability and predictive power of machine learning, offering a comprehensive framework for tackling complex PDEs.

\paragraph{AI Poincaré 2.0: Learning Conservation Laws}
\cite{liu2022ai} introduces an approach to discovering conservation laws in differential equations using machine learning. Termed "AI Poincaré 2.0," this method identifies and learns conservation laws in dynamical systems, offering a tool for analyzing complex systems. It demonstrates the potential of AI in uncovering fundamental laws of nature and integrating machine learning with classical physics theories.

\paragraph{Deep Learning for PDEs and Parameter Identification}
Research in \cite{tanyu2023deep} focuses on applying deep learning methods to solve PDEs and parameter identification problems. The study demonstrates the effectiveness of deep learning in addressing challenging aspects of PDEs, including identifying unknown parameters. Novel architectures and training strategies outperform traditional methods in accuracy and efficiency, highlighting the versatility of deep learning in computational mathematics.

\paragraph{Autonomous ODEs with Machine Learning}
In \cite{bouchereau2023machine}, machine learning methods for autonomously solving ordinary differential equations are introduced. This work combines autonomous systems and machine learning to create models capable of self-improvement and adaptation. By embedding machine learning algorithms within ODE frameworks, the authors achieve accurate solutions efficiently, paving the way for autonomous systems in scientific computing.

	\subsection{Algogenic Enhancements for Solving Differential Equations}
	\subsubsection{Adaptive Solution Strategies}
	\paragraph{Integrating AI for Strategy Selection}
	The implementation of adaptive solution strategies in solving differential equations through the application of large language models presents a nuanced approach to selecting the most suitable numerical method. This methodology is predicated on an in-depth analysis of the equation's characteristics, including its classification as an ordinary or partial differential equation, linearity, stiffness, and the presence of discontinuities. By harnessing the analytical capabilities of LLMs, a tailored selection process is facilitated, wherein the methodological choice—ranging from explicit to implicit approaches for ODEs, or from finite difference to finite element methods for PDEs—is dynamically aligned with the equation's specific attributes. This adaptive selection is further refined through ongoing analysis, allowing for real-time adjustments that enhance the solver's robustness and efficiency. Moreover, this integration democratizes the solving process, enabling broader accessibility without requiring deep expertise in numerical methods, thus revolutionizing the approach to differential equations across various fields.
	
	\paragraph{AI-Driven Method Adaptation}
	The paradigm shift introduced by LLMs in solving differential equations through AI-driven method adaptation reflects a deep understanding of the equation's nuances and the computational strategies' strengths and weaknesses. By analyzing historical data and outcomes, LLMs can recommend optimal numerical methods tailored to the problem's specific requirements. This approach not only improves efficiency and accuracy by adapting recommendations to novel scenarios but also provides insights into the problem's dynamics, enhancing the solving process's transparency and reproducibility. The practical implication is a streamlined computational process, facilitating the exploration of complex phenomena and fostering innovation in diverse scientific domains.
	
	\paragraph{Dynamic Adjustment During Computation}
	The dynamic adjustment capability of LLM-driven algorithms in solving differential equations, particularly in fields like computational fluid dynamics, introduces an adaptive mechanism that enhances solution accuracy and efficiency. By autonomously adjusting parameters like step size or mesh density in response to changing solution dynamics, these algorithms optimize computational resource allocation. This adaptability allows for real-time refinement, ensuring precision in simulations with varying complexities, thereby transcending traditional static methods and unlocking new potential in scientific exploration and problem-solving.
	
	\paragraph{Mathematical Foundations and Implementation}
	The foundation of adaptive solution strategies in LLMs is deeply rooted in mathematical analysis and machine learning algorithms, utilizing techniques like regression analysis, pattern recognition, and deep learning to understand the complex relationships between differential equation characteristics and numerical method efficacy. This comprehensive analysis facilitates informed decision-making in method selection and parameter optimization, enhancing differential equation solvers' efficiency and accuracy through a rigorous mathematical framework.
	
	\paragraph{Enhancing Efficiency and Accuracy}
	The integration of adaptive solution strategies via LLMs marks a significant advance in solving differential equations, optimizing computational efforts and ensuring precision across various problem types. This approach not only allocates computational resources more judiciously but also broadens the scope of solvable problems, enabling the tackling of complex, nonlinear equations that previously posed significant challenges. This Algogenic enhancement underscores a transformative step in numerical solutions, combining AI's adaptability with traditional methodologies for unparalleled efficiency and accuracy.
	
	\subsubsection{Dynamic Step Size Adjustment}
	\paragraph{Leveraging AI for Step Size Optimization}
	The application of AI in dynamically adjusting step sizes during differential equation solutions significantly enhances both the accuracy and efficiency of numerical methods. By continuously evaluating solution behavior and adapting step sizes accordingly, LLM algorithms optimize computational focus, applying finer or coarser steps as dictated by the solution's changing dynamics. This intelligent optimization addresses the inefficiencies of fixed step sizes, especially in equations with variable behaviors, ensuring precise and efficient solutions across a range of applications.
	
	\paragraph{Real-time Analysis and Adaptation}
	The real-time analysis and adaptive step sizing facilitated by LLMs in solving differential equations embody a significant leap in computational science. This approach dynamically adjusts step sizes based on the solution's evolving characteristics, enhancing efficiency and precision. By focusing computational efforts where they are most needed and allowing for real-time methodological adjustments, LLMs drive a paradigm shift towards more adaptable and intelligent numerical solutions.
	
	\paragraph{Mathematical Criteria for Step Size Adjustment}
	The AI-driven dynamic step size adjustment is underpinned by mathematical criteria that balance error estimation and stability requirements, ensuring that step sizes are optimized for accuracy without compromising computational efficiency. This approach adapts to the equation's specific challenges, such as stiffness or nonlinearity, and leverages mathematical formulations to guide the adjustment process, thereby achieving a nuanced balance between precision and computational demand.
	
	\paragraph{Implementation Challenges and Solutions}
	Integrating dynamic step size adjustment with LLMs into numerical solvers presents challenges such as computational efficiency and model robustness. By employing machine learning and algorithmic optimizations, these challenges can be addressed, enabling efficient and reliable adjustments that enhance the solver's performance. This integration represents a nuanced approach to overcoming traditional limitations, paving the way for more advanced numerical simulations.
	
	\paragraph{Impact on Numerical Solution Processes}
	The introduction of AI-driven dynamic step size adjustment transforms the numerical solution of differential equations by enhancing adaptability and precision. This innovation not only improves the accuracy of solutions but also optimizes computational efficiency, enabling the exploration of complex systems with greater detail and within computational constraints, thus marking a significant advancement in numerical methods.
	
	\subsubsection{Intelligent Parameter Tuning for PDEs}
	\paragraph{Optimization of Discretization Parameters}
	The optimization of discretization parameters through AI in solving PDEs represents a targeted approach to balancing computational load and solution precision. By adaptively tuning parameters such as mesh density, LLMs ensure efficient resource allocation, enhancing the accuracy of solutions to complex PDEs across various domains. This intelligent tuning adapts to the problem's evolving requirements, providing a dynamic, efficient solution process that transcends traditional discretization strategies.
	
	\paragraph{AI-driven Discretization Strategy}
	LLMs employ advanced machine learning algorithms to optimize discretization strategies for PDEs, identifying areas requiring refined computational focus. This adaptive approach ensures precise solution representations, maximizing computational efficiency and enabling the detailed exploration of complex phenomena, thus significantly enhancing the performance of numerical methods in solving multidimensional PDEs.
	
	\paragraph{Real-time Adaptation and Mesh Refinement}
	The real-time adaptation and mesh refinement facilitated by LLMs in solving PDEs underscore a dynamic approach to numerical solutions. By continuously assessing solution accuracy and adapting discretization parameters accordingly, LLMs optimize computational resource allocation, enhancing solution fidelity while maintaining computational efficiency, thus representing a significant advancement in the numerical analysis of differential equations.
	
	\paragraph{Mathematical Foundations of Parameter Optimization}
	The mathematical foundation of intelligent parameter tuning in LLMs incorporates error estimation and optimization algorithms to navigate the complex solution spaces of differential equations. This rigorous approach enables effective parameter adjustments, balancing accuracy with computational efficiency, and overcoming traditional challenges in numerical methods through a data-driven, adaptive strategy.
	
	\paragraph{Enhancing Multidimensional PDE Solutions}
	Intelligent parameter tuning significantly impacts the solution of multidimensional PDEs by enabling adaptive discretization strategies that cater to the problem's specific complexities. This approach not only ensures computational efficiency but also enhances solution accuracy, facilitating the exploration of complex phenomena with greater precision and contributing to advances in various scientific and engineering domains.
	
	\subsubsection{Error Prediction and Correction}
	\paragraph{Forecasting Solution Errors with AI}
	The incorporation of AI in forecasting solution errors represents a proactive approach to enhancing the accuracy and reliability of numerical solvers. By predicting potential inaccuracies and enabling real-time methodological adjustments, AI-driven error forecasting optimizes the solving process, introducing a dynamic, adaptive framework that improves efficiency and expands the solver's capabilities in handling complex differential equations.
	
	\paragraph{Real-time Method Adjustments for Enhanced Accuracy}
	AI-driven real-time method adjustments, informed by predictive error analysis, ensure targeted computational efforts for maintaining solution accuracy. This dynamic approach allows for the optimization of numerical methods during computation, adapting to predicted error landscapes and enhancing both solution precision and computational efficiency, thereby marking a significant evolution in numerical analysis techniques.
	
	\paragraph{Leveraging Machine Learning for Error Estimation}
	The integration of machine learning in error estimation leverages extensive datasets to predict and correct errors in differential equation solutions. This approach not only enhances the accuracy and reliability of numerical simulations but also contributes to the ongoing improvement of error prediction models, showcasing the potential of AI in advancing computational mathematics and solving complex problems.
	
	\paragraph{Optimizing Computational Workflows}
	Incorporating AI-driven error prediction and correction mechanisms optimizes computational workflows by balancing accuracy with computational efficiency. This approach minimizes the propagation of errors, streamlines resource allocation, and fosters continuous improvement in the computational process, enabling more effective and reliable solutions to differential equations across diverse applications.
	
	\paragraph{Impact on Complex Differential Equations}
	Predictive error correction through AI significantly impacts the solution of complex differential equations, enabling a nuanced, adaptive strategy that navigates potential challenges with enhanced accuracy and efficiency. This approach not only broadens the solver's applicability to complex problems but also fosters innovation and progress in computational mathematics, demonstrating the transformative potential of Algogenic enhancements in numerical analysis.

	\subsubsection{Pseudocode for Algogenic Differential Equation Solving}
	The Algogenic differential equations solving approach integrates AI to enhance traditional methods by dynamically adjusting solving parameters and strategies according to the observed behavior of the system and real-time error estimates. This pseudocode, accessible in \ref{fig:differential-equations-Algogen-pseudocode}, delineates a sophisticated framework integrating AI-driven enhancements for adaptive solving parameter control, equation manipulation, acceptance criteria, and real-time parameter optimization.
	
	\begin{algorithm}
		\caption{Algogenic Differential Equation Solving Pseudocode}
		\begin{algorithmic}[1]
			\Procedure{AlgogenicDEsSolving}{Equation, InitialConditions, Domain}
			
			\Comment{Preprocessing Phase}
			\State strategy $\gets$ AnalyzeAndSelectStrategy(Equation, InitialConditions)
			\State detailedAnalysis $\gets$ AnalyzeEquationCharacteristics(Equation, Domain)
			
			\Comment{Core Computation Phase}
			\State method $\gets$ SelectNumericalMethod(detailedAnalysis)
			\While{!Converged \&\& Error > AcceptableThreshold}
			\If{NeedDynamicAdjustment(detailedAnalysis)}
			\State strategy $\gets$ AdaptSolutionStrategy(detailedAnalysis)
			\EndIf
			\State result $\gets$ ApplyNumericalMethod(Equation, method, Domain)
			\State result, stepSize $\gets$ DynamicStepSizeAdjustment(result, Equation)
			\State result $\gets$ PerformIntegrationOrDiscretization(result, stepSize)
			\If{ErrorNotWithinThreshold(result)}
			\State method, Domain $\gets$ IntelligentParameterTuning(Equation, Domain, result)
			\Else
			\State result $\gets$ ErrorPredictionAndCorrection(result, Equation)
			\EndIf
			\EndWhile
			
			\Comment{Postprocessing Phase}
			\State finalResult $\gets$ InterpretAndEnhanceResults(result, Equation, Domain)
			
			\EndProcedure
		\end{algorithmic}\label{fig:differential-equations-Algogen-pseudocode}
	\end{algorithm}

	\begin{figure}
		\centering
		\includegraphics[width=0.7\textwidth]{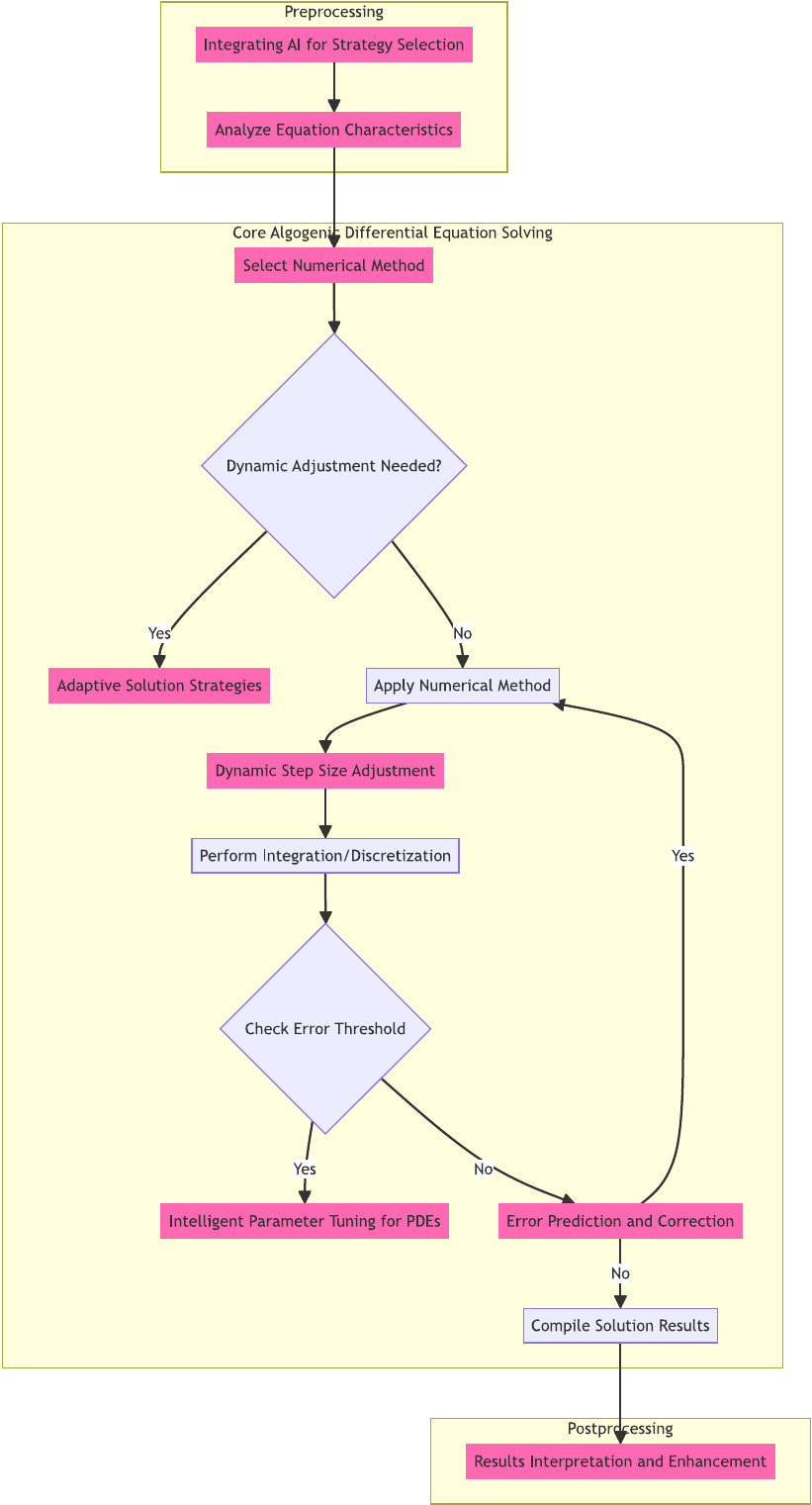}
		\caption{Integration of Algogenic Enhancements with Differential Equation Solving: This figure would illustrate the advanced framework of Algogenic Differential Equation Solving, highlighting the strategic integration of AI-driven enhancements throughout the solving process. It showcases the initial AI-driven strategy selection and detailed equation analysis in the preprocessing phase, followed by dynamic method selection, step size adjustment, and intelligent parameter tuning in the core computation phase. The diagram would further depict real-time error prediction and correction, ensuring solution precision and efficiency. This integration exemplifies how Algogenic enhancements optimize the solving of differential equations, making the process more adaptive, accurate, and computationally efficient, especially in handling complex equations with varying characteristics.}
		\label{fig:differential_equations}
	\end{figure}

	
	\chapterimage{pngs/image_processing.png} 
	\chapter{Image Processing Algogens}\index{Image Processing Algogens}
	
	\section{Image Denoising}\index{Image Denoising}
	\subsection{Introduction to Image Denoising}
	\subsubsection{The Concept of Image Denoising}
	\paragraph{Understanding Image Denoising}
	Image denoising is a critical preprocessing step in image processing that focuses on the reduction or removal of noise from digital images. Noise, characterized as unwanted interference, manifests in various forms such as random variations in brightness or color, sensor imperfections, or transmission errors. These disturbances obscure or degrade the quality of images, posing significant challenges to visual perception and analysis. The presence of noise can obscure fine details, reduce contrast, and introduce artifacts, making it essential to mitigate its effects.
	
	The primary aim of image denoising techniques is to recover the original, undistorted image as accurately as possible, thus enhancing both its aesthetic appeal and functional utility for subsequent processing or analysis tasks. Various approaches exist to tackle image denoising, ranging from simple filtering methods like median filtering and Gaussian smoothing to more sophisticated algorithms such as wavelet-based denoising, sparse representation, and deep learning-based methods.
	
	In recent years, deep learning techniques, particularly convolutional neural networks (CNNs), have gained prominence in image denoising tasks due to their ability to learn complex mappings from noisy to clean images directly from data. These models leverage large datasets to automatically learn the underlying structure of noise and effectively remove it, producing visually pleasing and semantically meaningful results.
	
	Despite advancements, image denoising remains a challenging problem, especially in scenarios with high levels of noise or complex noise patterns. Researchers continue to explore innovative strategies, incorporating domain knowledge, advanced mathematical models, and computational techniques to improve denoising performance across diverse applications, including medical imaging, surveillance, remote sensing, and photography.

	\paragraph{Sources of Noise in Digital Imaging}
	Noise in digital images can arise from a multitude of factors, each exerting its influence on image fidelity in distinctive manners. Sensor imperfections represent a primary contributor, where inherent flaws within image sensors introduce stochastic deviations in pixel values, manifesting as variations in brightness or color across the image. These imperfections stem from manufacturing inconsistencies or limitations in sensor technology, such as electronic noise and dark current. Environmental factors during image acquisition also play a pivotal role in noise generation. Conditions like diminished ambient light or elevated temperatures exacerbate noise levels, compromising image clarity and detail resolution. Furthermore, the transmission and storage stages of digital imaging workflows serve as potential breeding grounds for noise. Errors in data encoding, transmission, and subsequent decoding can introduce artifacts, leading to signal distortion and information loss. Additionally, compression algorithms employed during storage can inadvertently amplify existing noise, particularly in regions of low signal intensity. Hence, while digital imaging technologies continue to advance, mitigating noise remains a persistent challenge, necessitating the development of sophisticated noise reduction techniques and robust image processing algorithms to preserve image quality amidst diverse noise sources.

	\paragraph{Characteristics of Noise}
	Noise can manifest in digital images in several forms, including but not limited to Gaussian noise, salt-and-pepper noise, and speckle noise. \textbf{Gaussian noise}, characterized by a normal distribution, is commonly encountered in digital imaging and represents random variations in intensity across the image. It arises due to the combined effect of many random processes such as thermal noise in electronic circuits or quantization error in analog-to-digital conversion. This type of noise is often modeled as additive, where pixel values are perturbed by random values drawn from a Gaussian distribution with zero mean and a certain standard deviation, influencing the overall smoothness of the image. \textbf{Salt-and-pepper noise}, on the other hand, introduces sharp, sudden disturbances in the image, presenting as randomly scattered white or black (or both) pixels. It can result from errors during image acquisition or transmission, where certain pixels take extreme values compared to their surroundings, causing a disruptive effect on image quality. \textbf{Speckle noise}, often found in radar and ultrasound images, is caused by the interference of coherent signals. It manifests as granular patterns that degrade image clarity and can obscure important features. Speckle noise is inherent in coherent imaging systems and arises from the random phase differences of the scattered signals, leading to constructive or destructive interference. Moreover, it complicates image analysis and interpretation, requiring specialized denoising techniques tailored to its unique characteristics.

	\paragraph{The Challenge of Denoising}
	The challenge in image denoising lies in effectively removing noise while preserving the essential attributes of the image. This delicate balance is crucial as overly aggressive denoising can obliterate important details, while insufficient denoising leaves distracting noise artifacts. And, the complexity of this challenge is compounded by the need to adapt denoising techniques to the specific type and level of noise present in the image.
	
	Furthermore, adapting denoising methods requires a deep understanding of both the noise characteristics and the underlying image content. For, without this understanding, denoising algorithms may fail to differentiate between noise and actual image features. Moreover, denoising techniques must also consider the trade-off between noise reduction and preservation of image details such as edges, textures, and fine features. 
	
	Additionally, denoising algorithms often involve sophisticated mathematical models and computational processes. Likewise, they frequently rely on statistical methods to distinguish between signal and noise components in the image data. Furthermore, the effectiveness of denoising algorithms may vary depending on factors such as image resolution, color depth, and the presence of structured or random noise patterns.
	
	Consequently, researchers continually strive to develop novel denoising approaches that strike an optimal balance between noise reduction and preservation of image quality. Thus, advancements in denoising technology play a crucial role in various fields, including medical imaging, surveillance, satellite imaging, and digital photography.

	\paragraph{Mathematical Formulation}
	Mathematically, the image denoising problem can be formulated as an equation where the observed noisy image $I_{noisy}$ is the sum of the original clean image $I_{clean}$ and the noise $N$, i.e., $I_{noisy} = I_{clean} + N$. This straightforward relationship encapsulates the fundamental challenge in denoising: separating the underlying signal from the unwanted noise. However, the task is far from trivial due to the intricate nature of real-world noise, which can manifest in various forms such as Gaussian noise, salt-and-pepper noise, or even more complex patterns in certain applications like medical imaging or surveillance. Consequently, devising effective denoising algorithms necessitates a deep understanding of both the characteristics of the noise and the structure of the underlying signal. Moreover, the formulation implies a crucial assumption that the noise is additive, which may not always hold true in practical scenarios. Despite this limitation, denoising algorithms leverage this assumption as a starting point, exploiting statistical properties of the noise and employing sophisticated models to approximate the clean image. Techniques ranging from classical methods like median filtering to state-of-the-art deep learning approaches are employed to tackle this problem. Furthermore, the success of these algorithms heavily relies on the quality and richness of the available data. Thus, a comprehensive understanding of the interplay between noise and signal, coupled with innovative algorithmic designs and abundant data, is paramount for achieving robust and accurate image denoising.

	\subsubsection{Key Principles and Mechanisms}
	\paragraph{Balancing Noise Reduction and Detail Preservation}
	The cornerstone of image denoising is the delicate balance between the removal of noise and the preservation of critical image features. This balance is crucial because the primary objective of denoising is not just to reduce noise but to do so in a way that maintains the integrity of the original image. Essential details, such as edges, textures, and fine structures, carry significant information about the image content and are vital for both human perception and automated image analysis tasks.
	
	Furthermore, achieving optimal noise reduction while preserving essential image features presents a challenging optimization problem. It requires sophisticated algorithms that can distinguish between noise and signal reliably. Moreover, the effectiveness of denoising techniques often depends on the specific characteristics of the noise present in the image, such as its distribution and intensity.
	
	However, blindly removing noise without considering its impact on image details can lead to undesirable consequences, such as loss of sharpness or blurring of important structures. Therefore, denoising algorithms must strike a careful balance, employing strategies like adaptive filtering or incorporating prior knowledge about the image content to ensure that crucial details are preserved while noise is effectively suppressed.
	
	Moreover, in practical applications, the trade-off between noise reduction and detail preservation may vary depending on the specific requirements of the task at hand. For instance, in medical imaging, preserving fine anatomical structures is critical for accurate diagnosis, whereas in surveillance or satellite imagery, reducing noise to enhance object detection may take precedence.
	
	In summary, successful image denoising hinges on striking the right balance between reducing noise and preserving essential image features. Achieving this balance requires sophisticated algorithms, careful consideration of noise characteristics, and adaptation to the specific demands of the application.

	\paragraph{Spatial Filtering Techniques}
	Spatial filtering represents one of the most basic approaches to image denoising, involving the direct manipulation of pixel values based on their spatial location. Techniques such as Gaussian blur are foundational. A Gaussian filter is applied to average pixel values in a local neighborhood, smoothing out variations. This filter, characterized by its bell-shaped curve, weights the contributions of neighboring pixels based on their distance from the central pixel, effectively reducing high-frequency noise while somewhat blurring edges and details.
	
	Moreover, alongside Gaussian blur, other spatial filtering techniques like median filtering exist. Median filtering, unlike Gaussian blur, replaces each pixel's value with the median value of its neighborhood, effectively eliminating outliers and impulse noise without significantly blurring edges or details. Additionally, mean filtering, similar to Gaussian blur, replaces each pixel's value with the average of its neighboring pixels, providing a smoother image at the expense of potentially blurring fine details more than median filtering.
	
	Furthermore, spatial filtering techniques can be adaptive, where the filter parameters vary depending on the local image characteristics. Adaptive filters adjust their parameters based on the local image statistics, allowing for better preservation of edges and details in areas with significant variations while effectively denoising smoother regions. However, the complexity of adaptive filtering may lead to higher computational costs compared to non-adaptive methods.
	
	In contrast to spatial filtering, frequency domain techniques such as Fourier transform-based denoising exploit the frequency components of an image. These techniques, while powerful, may require more computational resources and are often used in conjunction with spatial filtering for optimal denoising results.
	
	Therefore, understanding the principles and trade-offs of different spatial filtering techniques is essential for effective image denoising in various applications.

	\paragraph{Frequency Domain Approaches}
	Beyond spatial filtering, frequency domain approaches offer a different perspective on image denoising. Techniques like wavelet transforms decompose the image into components at various scales and frequencies, allowing for selective noise reduction in the frequency domain. By identifying and attenuating the components predominantly associated with noise, while preserving those corresponding to the actual image signal, wavelet-based methods can achieve more nuanced denoising, particularly for images where noise and signal characteristics vary across different scales.
	
	Furthermore, wavelet-based denoising techniques are advantageous because they can handle non-stationary signals effectively. While spatial filtering methods assume that the statistical properties of the image remain constant throughout, wavelet transforms adaptively adjust to the varying characteristics of the image content, making them suitable for denoising tasks where noise characteristics change across different regions of the image. Moreover, in addition to noise reduction, wavelet-based approaches often yield enhanced feature preservation compared to spatial filtering alone. This is because they operate on multi-scale representations of the image, allowing for more precise manipulation of noise components while minimizing the impact on essential image features.
	
	Additionally, wavelet-based denoising methods offer computational efficiency by exploiting the sparsity of wavelet representations. Since many natural images exhibit sparsity in the wavelet domain, where only a small fraction of coefficients contain significant signal information, wavelet-based denoising algorithms can achieve substantial noise reduction with relatively low computational cost. This efficiency is particularly beneficial for real-time applications or scenarios where computational resources are limited.
	
	Nevertheless, wavelet-based denoising is not without its challenges. One limitation is the selection of an appropriate wavelet basis and decomposition level, which can significantly impact the denoising performance. Moreover, the trade-off between noise reduction and preservation of fine image details requires careful tuning of denoising parameters, which may necessitate extensive experimentation or domain-specific knowledge. Despite these challenges, the versatility and effectiveness of wavelet-based denoising make it a valuable tool in the image processing toolbox.

	\paragraph{Non-Local Means and Advanced Algorithms}
	The non-local means algorithm represents a significant advancement in denoising techniques, moving beyond the local consideration of pixels to explore redundancy within the entire image. By averaging pixels based on the similarity of their local neighborhoods across the whole image, non-local means can preserve detailed structures while effectively reducing noise. This method highlights the principle that pixels with similar patterns, regardless of their spatial proximity, can contribute to a more accurate estimation of the noise-free image.
	
	Furthermore, the utilization of non-local means underscores the importance of considering global information in image processing tasks. While traditional denoising methods rely solely on local information, non-local means incorporate knowledge from distant parts of the image, leading to more robust denoising performance. Moreover, the algorithm's ability to adaptively weigh the contributions of different pixels based on their similarity enhances its versatility across various types of images and noise characteristics.
	
	Additionally, non-local means offer computational advantages by exploiting redundancies in the image data. Despite the increased computational complexity compared to local methods, the efficiency gains from leveraging redundant information often result in competitive performance in terms of both accuracy and speed. Furthermore, the algorithm's simplicity in implementation and parameter tuning contributes to its widespread adoption in practical applications.
	
	Consequently, the integration of non-local means into advanced algorithms has propelled the field of image denoising towards more sophisticated and effective solutions. By embracing the concept of non-local similarity, these algorithms can achieve superior denoising results across a diverse range of imaging scenarios, from medical imaging to digital photography. Thus, non-local means stand as a testament to the power of incorporating global context in image processing algorithms.

	\paragraph{Adaptive and Model-Based Techniques}
	Modern denoising methods increasingly rely on adaptive and model-based techniques, where the denoising process is informed by models of image formation and noise characteristics. These methods, including sophisticated algorithms like BM3D (Block-Matching and 3D filtering), adapt the denoising strategy based on the estimated local properties of noise and signal. \textbf{Moreover}, by modeling the noise and leveraging the inherent redundancy in natural images, these techniques can achieve high-quality denoising across a wide range of noise levels and types.
	
	The utilization of adaptive techniques allows denoising algorithms to dynamically adjust their parameters and processing steps according to the local characteristics of the image. \textbf{Furthermore}, model-based approaches enable the incorporation of prior knowledge about the image structure and noise statistics, leading to more effective noise reduction. This integration of models into the denoising process enhances the ability of algorithms like BM3D to accurately distinguish between signal and noise components, \textbf{thereby} preserving important image details while suppressing unwanted artifacts.
	
	In particular, BM3D exploits similarities between image patches to efficiently estimate and remove noise. \textbf{Additionally}, the 3D collaborative filtering employed in BM3D takes advantage of the redundancy present in image volumes, resulting in enhanced denoising performance. \textbf{Consequently}, these adaptive and model-based techniques offer a robust solution for denoising tasks in various applications, including medical imaging, surveillance, and photography.
	
	The effectiveness of these methods is further demonstrated by their ability to handle different types of noise, such as Gaussian, Poisson, or impulse noise. \textbf{Furthermore}, their adaptability ensures reliable performance even in scenarios with non-uniform noise characteristics or complex image structures. \textbf{Hence}, the combination of adaptive strategies and model-based principles represents a significant advancement in the field of image denoising, providing versatile and efficient solutions for real-world applications.

	\paragraph{Mathematical Representation}
	The mathematical underpinnings of image denoising involve formulations that explicitly or implicitly model the noise and the image. Denoising methodologies often adopt an optimization framework to address this challenge. One common approach formulates denoising as an optimization problem, where the primary objective is to minimize a cost function. This cost function typically comprises two components: a fidelity term and a regularization term. The fidelity term measures the agreement between the denoised image and the observed noisy image. It ensures that the denoised image retains essential features present in the noisy input. Conversely, the regularization term incorporates prior knowledge about the structure of natural images. This knowledge is often enforced through constraints that promote certain properties, such as smoothness or sparsity, in the denoised image.
	
	And, this dual-component formulation allows denoising algorithms to strike a balance between fidelity to the observed data and adherence to prior assumptions about the underlying image structure. But, achieving this balance is not trivial, as it involves navigating trade-offs between different sources of information. Nevertheless, by carefully designing the cost function and selecting appropriate regularization techniques, denoising algorithms can effectively suppress noise while preserving important image features. Moreover, the flexibility of this framework enables the incorporation of various image priors, making it adaptable to different denoising scenarios and imaging modalities. Furthermore, the optimization process itself can be tailored to exploit specific properties of the noise model and the image structure, leading to efficient and robust denoising solutions.

	The exploration of key principles and mechanisms in image denoising reveals a field driven by the goal of intelligently reducing noise while safeguarding the richness and authenticity of the original image. Through a combination of spatial, frequency, and model-based approaches, image denoising continues to evolve, offering increasingly sophisticated tools for enhancing image quality.

	\subsubsection{The Role of Advanced Filtering Techniques}
	\paragraph{Evolution Beyond Simple Averaging}
	Advanced filtering techniques represent a significant evolution in image denoising, moving beyond the simple averaging approaches of early spatial filters. These methods incorporate more sophisticated strategies that account for the inherent structure and information within the image. By considering both the spatial proximity and the intensity similarity of pixels, advanced filters can more effectively discriminate between noise and important image features, leading to denoising that preserves essential details such as edges and textures.
	
	Furthermore, these advanced techniques not only surpass the limitations of basic averaging methods but also address the challenges posed by complex noise patterns and high-frequency components present in modern digital images. Unlike traditional spatial filters, which indiscriminately blur the entire image to reduce noise, advanced filters utilize complex algorithms to selectively attenuate noise while preserving the sharpness of image edges and fine textures. Additionally, they adaptively adjust their filtering parameters based on local image characteristics, ensuring optimal denoising performance across diverse image content.
	
	Moreover, the incorporation of advanced filtering techniques opens up new possibilities for enhancing image quality beyond mere noise reduction. By leveraging the rich information embedded in the image structure, these methods enable more sophisticated image restoration tasks such as super-resolution and inpainting. Through the synergistic combination of advanced denoising algorithms with other image processing modules, researchers can achieve unprecedented levels of fidelity and visual clarity in digital imagery, revolutionizing applications ranging from medical imaging to satellite photography.

	\paragraph{Non-Local Means Denoising}
	The non-local means filter stands as a watershed in the realm of denoising techniques, reshaping the landscape with its innovative approach. While conventional filters confine their operations within a restricted local neighborhood, the non-local means filter defies this limitation, embarking on a journey through the entire image domain. It meticulously scrutinizes each pixel, forging connections with distant brethren, all in pursuit of clarity and purity. Through a judicious weighting scheme, akin to the discerning eye of an art connoisseur, it evaluates the resemblance between local neighborhoods, assigning significance to each contribution based on this intrinsic similarity.
	
	This audacious methodology yields remarkable dividends, preserving the intricate tapestry of textures and structures that define the essence of the image. By transcending spatial boundaries, the filter unearths hidden correlations, revealing the latent coherence that permeates seemingly disparate regions. It is through this revelation that noise succumbs, vanquished by the collective might of pixels harmonizing in their shared likeness.
	
	The filter's underlying ethos is elegantly simple yet profoundly effective: unity in similarity transcends the constraints of proximity. It operates on the premise that pixels need not be neighbors to commune; rather, it is their kinship in appearance that binds them in a symphony of denoising prowess. This principle forms the cornerstone of its operation, guiding each computational step with unwavering purpose.
	
	In essence, the non-local means filter orchestrates a symphony of pixels, conducting them towards a crescendo of clarity. It beckons forth the latent harmony that lies dormant within the image, coaxing noise into oblivion while nurturing the delicate nuances of its content. Through its visionary approach, it heralds a new era in denoising, where fidelity and detail reign supreme, transcending the constraints of locality to embrace the boundless expanse of similarity.

	\paragraph{Bilateral Filtering for Edge Preservation}
	The bilateral filter further refines the concept of adaptive filtering by combining spatial distance and intensity difference into a single weight. This dual consideration allows the filter to smooth areas of similar intensity while preserving sharp intensity transitions, such as edges. The result is a denoising effect that reduces noise in flat regions and around edges without blurring the edges themselves. Bilateral filtering is particularly useful in applications where edge preservation is critical, such as in medical imaging or feature extraction tasks.
	
	Moreover, the bilateral filter's ability to incorporate both spatial and intensity information distinguishes it from traditional filters, making it particularly adept at handling complex images with varying textures and structures. This versatility enables its widespread use across different domains, including computer vision, image processing, and graphics. Additionally, its computational efficiency has made it a popular choice for real-time applications where processing speed is crucial.
	
	Furthermore, the adaptability of bilateral filtering makes it suitable for a range of scenarios beyond noise reduction. For instance, it can enhance the perceptual quality of images by emphasizing important features while suppressing irrelevant details. This characteristic makes it invaluable in tasks such as image enhancement and stylization.
	
	In medical imaging, where accurate delineation of anatomical structures is paramount, bilateral filtering offers significant advantages. By preserving edge information while reducing noise, it facilitates more precise segmentation and analysis, leading to improved diagnostic accuracy and treatment planning.
	
	Overall, the bilateral filter stands out as a powerful tool for edge preservation and noise reduction, offering a balance between smoothing and preserving important image features. Its effectiveness in various applications underscores its significance in modern image processing pipelines.

	\paragraph{Edge-Aware Properties}
	One of the key strengths of advanced filtering techniques like non-local means and bilateral filters is their edge-aware property. \textbf{Moreover}, by inherently distinguishing between areas of uniform intensity and those featuring significant intensity gradients, these filters adapt their denoising strength, applying less smoothing near edges \textbf{while} more in homogeneous regions. This adaptability ensures that the denoising process does not indiscriminately blur important image details, a common drawback of simpler denoising methods. \textbf{Additionally}, the edge-aware nature of these filters enables them to preserve edge sharpness and enhance overall image quality. \textbf{Furthermore}, this edge-awareness plays a crucial role in various computer vision tasks such as image segmentation and object recognition \textbf{as well as} in medical image processing where preserving fine details is essential for accurate diagnosis. In practical applications, this property \textbf{likewise} facilitates the removal of noise from images without sacrificing their structural integrity, resulting in visually pleasing and informative outcomes.

	\paragraph{Computational Considerations and Optimizations}
	While advanced filtering techniques offer superior denoising performance, they also pose computational challenges due to their complexity, particularly for non-local means, which requires comparing each pixel with every other pixel in the image. This computational burden can lead to impractical processing times, especially for high-resolution images or real-time applications. However, recent advancements have been made in optimizing these algorithms to address these challenges. 
	
	One key optimization strategy involves the development of fast approximation methods. These methods aim to achieve comparable denoising results while significantly reducing the computational overhead. By sacrificing a certain degree of accuracy for speed, these approximations enable the application of advanced denoising techniques in scenarios where real-time processing is critical.
	
	Additionally, parallel computing has emerged as a powerful tool for accelerating denoising algorithms. By distributing the computational workload across multiple processing units or cores, parallelization allows for significant speedups in denoising tasks. This approach is particularly effective for large-scale image processing tasks, where the data can be divided into smaller chunks and processed concurrently.
	
	Furthermore, hardware accelerations, such as specialized GPUs or dedicated hardware accelerators, have been increasingly utilized to further enhance the performance of denoising algorithms. These hardware platforms are specifically designed to execute parallelizable tasks efficiently, making them well-suited for accelerating image processing operations like denoising. Integrating such hardware accelerators into denoising systems can lead to substantial improvements in processing speed and overall efficiency.
	
	Collectively, these optimizations have significantly reduced processing times, making advanced denoising techniques more accessible for a wider range of applications. By leveraging fast approximation methods, parallel computing, and hardware accelerations, researchers and practitioners can now deploy advanced denoising algorithms in real-world scenarios with improved efficiency and effectiveness.

	\paragraph{Integration with Machine Learning Models}
	Integrating machine learning, particularly deep learning models, with advanced filtering techniques has catalyzed a transformative shift in image denoising methodologies. By synergizing machine learning algorithms with sophisticated filtering mechanisms, a multifaceted approach emerges, redefining the landscape of denoising strategies. These fusion methodologies harness the innate adaptability of machine learning models to emulate and surpass the efficacy of conventional filters.
	
	Deep learning architectures, renowned for their capacity to discern intricate patterns within data, serve as dynamic counterparts to traditional denoising filters. Through extensive training, these models acquire an intrinsic understanding of image features and noise characteristics, enabling them to emulate the nuanced behavior of sophisticated filters. Moreover, machine learning algorithms possess the versatility to adapt and optimize filter parameters based on diverse image and noise profiles, thereby tailoring the denoising process to specific contexts.
	
	The integration of machine learning with advanced filtering techniques not only augments denoising efficacy but also mitigates computational overhead. By leveraging learned representations of image structure and noise patterns, these hybrid approaches streamline the denoising pipeline, expediting processing times without compromising on performance. Consequently, the amalgamation of machine learning and filtering methodologies engenders a paradigm shift towards efficient, adaptive, and context-aware image denoising solutions.

	Advanced filtering techniques have fundamentally transformed the landscape of image denoising, offering nuanced and effective solutions that maintain the integrity of the original image. Through continuous development and integration with emerging technologies, these methods remain at the forefront of efforts to enhance image quality in the face of noise.

	\subsubsection{Applications and Limitations}
	\paragraph{Ubiquitous Need for Image Denoising}
	Image denoising finds its relevance in a myriad of applications where the clarity and quality of images are paramount. In digital photography, denoising is essential for enhancing the visual appeal of photos taken in less-than-ideal lighting conditions, where sensor noise can significantly degrade image quality. The application of denoising algorithms ensures that the captured images maintain their integrity and sharpness, allowing photographers to produce professional-grade photographs irrespective of environmental constraints. Moreover, in the realm of medical imaging, denoising plays a pivotal role in enhancing the interpretability of diagnostic images obtained from modalities such as MRI, CT scans, and ultrasound. These modalities often suffer from inherent noise stemming from various sources including hardware imperfections and physiological factors. By employing sophisticated denoising techniques, medical practitioners can extract crucial diagnostic information with higher fidelity, leading to more accurate diagnoses and improved patient care. Furthermore, denoising is indispensable in video processing applications, particularly in scenarios where video streams are captured in low-light environments or under challenging shooting conditions characterized by high ISO settings. By mitigating noise artifacts, denoising algorithms ensure that the visual content remains clear and coherent, thereby enhancing viewer satisfaction and enabling seamless downstream analysis. The ubiquity of image denoising across diverse domains underscores its indispensable nature in modern imaging workflows, where the pursuit of visual clarity and fidelity reigns supreme.

	\paragraph{Preserving Image Integrity}
	The core challenge in image denoising across these applications is to effectively reduce noise without compromising the integrity of the original image. This balance is crucial because the details lost during aggressive denoising processes could be vital for the application at hand. For instance, in medical imaging, fine details in an image might represent critical diagnostic features. In digital photography, the textures and edges contribute to the overall aesthetic and realism of the photograph. Therefore, the goal of denoising extends beyond mere noise reduction to include the preservation of these essential elements that confer meaning and value to the image.
	
	Achieving this balance requires sophisticated algorithms and techniques that not only remove noise but also intelligently preserve important features. And, given the diverse range of applications, each with its own unique requirements and constraints, a one-size-fits-all approach is not viable. Instead, adaptive methodologies must be employed to tailor the denoising process according to the specific needs of each application.
	
	Moreover, it's important to consider the trade-offs involved in the denoising process. While aggressive denoising may lead to sharper images, it could also result in the loss of subtle details. Conversely, a more conservative approach may retain more details but could leave behind noticeable noise artifacts. Thus, striking the right balance between noise reduction and detail preservation is a delicate task that requires careful consideration.
	
	Furthermore, advancements in deep learning have revolutionized the field of image denoising, enabling the development of highly sophisticated neural network architectures capable of learning intricate patterns and structures from large datasets. However, even with these powerful tools at hand, the challenge of preserving image integrity remains at the forefront, emphasizing the importance of ongoing research and innovation in this critical area.

	\paragraph{Trade-offs and Computational Considerations}
	The trade-off between noise reduction and detail preservation is paramount in image denoising. Achieving optimal noise reduction while preserving crucial details is a delicate balance. However, this challenge is just one facet of the complex landscape of image processing. Computational efficiency emerges as another crucial consideration, particularly in scenarios involving high-resolution images or real-time video processing.
	
	Advanced denoising algorithms, particularly those rooted in deep learning methodologies, often exhibit remarkable efficacy in noise reduction. Yet, this efficacy can come at a computational cost. These algorithms, characterized by their intricate neural architectures and sophisticated training processes, frequently demand substantial processing power and time resources. Consequently, in contexts demanding real-time performance or rapid analysis, such as live video streaming or clinical diagnosis, the computational overhead of these algorithms becomes a significant concern.
	
	Moreover, the computational burden extends beyond merely executing the denoising algorithm itself. Preprocessing steps, such as image acquisition and data preparation, as well as post-processing tasks like result visualization or integration into larger systems, further compound the computational demands. Therefore, in addition to evaluating the efficacy of denoising algorithms, it becomes imperative to assess their computational efficiency and scalability, ensuring they align with the specific requirements and constraints of the intended application.
	
	In summary, while advancements in denoising algorithms offer promising solutions for enhancing image quality, the computational considerations loom large in practical implementation. Striking a balance between noise reduction and computational efficiency is essential for realizing the full potential of these techniques across diverse domains.

	\paragraph{Adapting to Noise Variability}
	Another limitation is the variability of noise, which can differ vastly across images depending on the source, the capture device, and environmental conditions. \textbf{Moreover}, a denoising technique effective for Gaussian noise might not perform as well on speckle or salt-and-pepper noise. This discrepancy \textbf{requires} adaptive or hybrid approaches capable of handling diverse noise types. \textbf{Furthermore}, this variability \textbf{calls for} flexible denoising solutions that can be tailored to the specific characteristics of the noise and the requirements of the application. 
	
	Addressing the diverse nature of noise is crucial in real-world applications, as images captured in different scenarios can exhibit unique noise patterns. \textbf{Additionally}, the choice of denoising method must be guided by the understanding that noise can manifest in various forms. \textbf{Consequently}, denoising algorithms need to be robust enough to adapt to these variations. 
	
	Incorporating adaptive mechanisms into denoising algorithms allows them to adjust their parameters dynamically based on the characteristics of the noise present in the image. \textbf{Furthermore}, hybrid approaches that combine multiple denoising techniques offer a versatile solution to tackle different noise types effectively. \textbf{On the other hand}, relying solely on a single denoising method may lead to suboptimal results when confronted with diverse noise scenarios.
	
	In summary, the challenge of noise variability underscores the necessity for adaptive and flexible denoising strategies. \textbf{Therefore}, researchers continue to explore innovative approaches to address this issue and enhance the performance of denoising algorithms across a wide range of applications.

	\paragraph{Emerging Solutions and Future Directions}
	Despite these challenges, the field of image denoising continues to evolve, with research and development efforts focused on overcoming these limitations. Advances in computational hardware, algorithm optimization, and the development of AI-driven approaches are expanding the capabilities of denoising techniques. 
	
	Furthermore, the integration of denoising algorithms into imaging devices and software is making these powerful tools more accessible to end-users. This integration facilitates automatic and real-time noise reduction in various applications, ranging from medical imaging to consumer photography. Moreover, the seamless incorporation of denoising algorithms enhances the overall efficiency and effectiveness of imaging systems, enabling clearer and more precise image capture in diverse environments.
	
	Additionally, the synergy between hardware advancements and algorithmic innovations is paving the way for unprecedented levels of denoising performance. By leveraging the computational power of modern GPUs and specialized hardware accelerators, denoising algorithms can now process high-resolution images with remarkable speed and fidelity. This increased efficiency not only improves the user experience but also enables novel applications in fields such as remote sensing, surveillance, and autonomous navigation.
	
	In summary, the convergence of hardware, software, and algorithmic advancements represents a significant milestone in the evolution of image denoising. As these technologies continue to mature, we can expect further breakthroughs in noise reduction capabilities, ultimately empowering users with enhanced image quality and fidelity across a wide range of applications.

	The applications and limitations of image denoising highlight its critical role in enhancing image quality across diverse domains. As technology advances, the ongoing development of more sophisticated, efficient, and adaptable denoising methods promises to further mitigate these limitations, broadening the scope and effectiveness of image denoising in improving visual communication and analysis.

	\subsubsection{Algorithmic Pseudocode for Basic Image Denoising}
	The Denoising Algorithm presented here provides a straightforward strategy for cleaning up images by reducing noise. Each pixel's value in the resulting image is determined as the average of its neighboring pixel values from the original input image. This process is facilitated by the assumed functionality of the `GetNeighbors` function, which retrieves the immediate neighboring pixels. While this simplistic method effectively smooths out noise, it also runs the risk of blurring significant image features such as edges and textures. Despite its simplicity, the averaging filter serves as a foundational concept in understanding image denoising techniques. However, in practical applications where preserving intricate image details is essential, more sophisticated algorithms are often necessary to achieve optimal results. The pseudo figure referenced in this context illustrates the iterative nature of the denoising algorithm, showcasing its stepwise approach to processing image data. Image Denoising is encapsulated in pseudocode \ref{fig:image-denoising-pseudocode}, illustrating its iterative approach to parameter estimation.
	
	\begin{algorithm}
		\caption{Basic Image Denoising Using Averaging Filter}
		\begin{algorithmic}[1]
			\Procedure{BasicImageDenoising}{Image}
			\State $Width \gets Image.Width$
			\State $Height \gets Image.Height$
			\State $DenoisedImage \gets CreateEmptyImage(Width, Height)$
			\For{$x \gets 1$ to $Width$}
			\For{$y \gets 1$ to $Height$}
			\State $Sum \gets 0$
			\State $Count \gets 0$
			\For{each $neighbor$ in $GetNeighbors(Image, x, y)$}
			\State $Sum \gets Sum + neighbor.Value$
			\State $Count \gets Count + 1$
			\EndFor
			\State $DenoisedPixel \gets Sum / Count$
			\State $SetPixelValue(DenoisedImage, x, y, DenoisedPixel)$
			\EndFor
			\EndFor
			\State \Return $DenoisedImage$
			\EndProcedure
		\end{algorithmic}\label{fig:image-denoising-pseudocode}
	\end{algorithm}
	
	\subsection{Previous Work on ML and AI Interplay with Image Denoising Algorithms}
	
	\paragraph{Overview of Deep Learning for Image Denoising}
	The evolution of deep learning methodologies has influenced the field of image denoising, providing an overview of advancements and challenges within this domain. In 2020, a study in \textit{Neural Networks} discussed the impact of deep learning techniques on image denoising \cite{tian2020deep}. This study reviewed various deep learning frameworks, noting their effectiveness in handling noise patterns while preserving image details. It discussed the shift from manual feature extraction to data-driven approaches and identified limitations such as reliance on large datasets and computational resources, suggesting avenues for future research.
	
	\paragraph{Machine Learning Approaches to Image Denoising}
	In 2021, a review in \textit{IEEE Access} analyzed image denoising techniques, focusing on machine learning models \cite{thakur2021image}. It compared classical algorithms with modern machine learning and deep learning solutions, addressing the increasing complexity of noise in digital images. The review advocated for hybrid models combining different approaches to meet the demands of modern image processing tasks.
	
	\paragraph{Hybrid Deep Learning and Optimization for Image Denoising}
	In 2023, a study proposed a hybrid approach integrating deep learning and optimization techniques to enhance image denoising \cite{jebur2023image}. Published in \textit{Technologies}, the paper introduced a model combining Bidirectional Long Short-Term Memory (Bi-LSTM) and Convolutional Neural Networks (CNN), optimized through the Self-Improved Orca Predation Algorithm (SI-OPA). This hybrid model demonstrated improved denoising efficiency, showcasing the potential of combining deep learning architectures with advanced optimization techniques.

	\subsection{Algogenic Enhancements for Image Denoising}
	\subsubsection{LLM-Based Noise Characterization}
	\paragraph{Harnessing the Power of LLMs for Understanding Noise}
	The integration of Large Language Models for identifying and characterizing noise patterns in images has shown promise in enhancing the specificity of denoising algorithms. By analyzing descriptive metadata and contextual information related to images, LLMs can pinpoint the origins and types of noise, such as Gaussian or salt-and-pepper noise. This approach allows for the development of denoising strategies tailored to specific noise profiles, potentially increasing the precision of noise removal while preserving image integrity. However, the practical implementation of this method requires seamless integration into existing image processing pipelines and a comprehensive understanding of the nuances in different noise types, which may pose challenges in terms of computational complexity and algorithm adaptability.
	
	\paragraph{Analyzing Descriptive Metadata for Noise Insights}
	Utilizing LLMs to analyze image metadata presents a nuanced method for predicting noise characteristics, allowing for the customization of denoising algorithms based on information about the imaging device, environmental conditions, and known imaging artifacts. This metadata-driven approach could lead to more effective noise reduction by enabling algorithms to anticipate and mitigate specific noise types. However, the effectiveness of this strategy largely depends on the availability and accuracy of metadata, which may not always be consistent or comprehensive. Furthermore, translating these insights into actionable denoising strategies requires a sophisticated understanding of the relationship between metadata elements and noise manifestations, underscoring the need for advanced LLM capabilities and in-depth domain knowledge.
	
	\paragraph{Facilitating Targeted Denoising Strategies}
	Employing LLMs to facilitate targeted denoising strategies involves leveraging nuanced noise characterizations to apply denoising techniques selectively across an image. This approach suggests that denoising algorithms could achieve improved efficiency by focusing on areas more likely to be affected by specific types of noise, based on LLM analysis. Implementing such targeted strategies necessitates a deep integration between LLMs and denoising algorithms, ensuring that insights into noise characteristics directly inform the denoising process. While this method holds the potential for enhancing denoising outcomes, it also requires sophisticated algorithmic frameworks capable of dynamically adjusting denoising parameters in response to LLM insights, posing challenges in terms of algorithm complexity and computational resources.
	
	\paragraph{The Role of LLMs in Advanced Denoising Frameworks}
	Incorporating LLMs into advanced denoising frameworks introduces a context-aware approach to image processing, aiming to improve the effectiveness of noise reduction while preserving essential image details. LLMs can offer a nuanced understanding of the image content, which, when integrated into denoising algorithms, allows for adaptive noise reduction strategies that consider the unique characteristics of each image. However, this integration challenges traditional denoising methods, necessitating the development of more sophisticated algorithms that can interpret and act upon LLM-generated insights. Furthermore, the success of LLM-based denoising frameworks relies on continuous learning and adaptation, requiring extensive datasets and computational power to achieve optimal performance.
	
	\paragraph{Mathematical Modeling of Noise Based on LLM Insights}
	Translating LLM insights into mathematical models for noise characterization proposes a method to enhance image denoising algorithms by providing a more accurate representation of noise distributions. By utilizing LLM-derived information, such as the likelihood of specific noise types based on image metadata, denoising algorithms can adjust their parameters more precisely, potentially improving noise reduction effectiveness. Implementing this approach involves complex mathematical formulations that accurately reflect LLM insights, requiring a deep integration between LLM analyses and algorithmic noise models. While promising, the practical application of this strategy is contingent upon the ability to accurately translate linguistic insights into quantitative noise models, highlighting the need for interdisciplinary expertise in both natural language processing and statistical modeling.
	
	\subsubsection{Semantic Understanding for Selective Denoising}
	\paragraph{Elevating Denoising with Deep Semantic Insights}
	Integrating LLMs for semantic analysis in image denoising processes enables a deeper understanding of image content, facilitating selective denoising strategies that prioritize the preservation of semantically important features. By analyzing image-related textual data, LLMs can identify key elements within images that require careful denoising treatment, such as critical details in medical imaging or important textural information in natural scenes. However, leveraging semantic insights for denoising poses challenges in accurately interpreting and applying these insights within denoising algorithms. Moreover, the effectiveness of this approach depends on the quality and depth of the textual data available for analysis, as well as the LLM's ability to understand and translate these insights into actionable denoising strategies.
	
	\paragraph{Selective Denoising Based on Content Relevance}
	Employing LLMs to enable selective denoising based on content relevance suggests a tailored approach to noise reduction, where denoising efforts are concentrated on preserving the integrity of crucial image features. This strategy involves LLMs identifying and prioritizing areas within an image based on their semantic significance, allowing for differential denoising treatment. Implementing selective denoising requires sophisticated algorithms capable of dynamically adjusting their operations in response to LLM insights, highlighting potential challenges in algorithm design and computational efficiency. Furthermore, the success of this approach is dependent on the LLM's ability to accurately discern the semantic importance of different image regions, necessitating advanced natural language understanding and image processing capabilities.
	
	\paragraph{Contextual Analysis for Enhanced Image Processing}
	Integrating contextual analysis through LLMs in the denoising process introduces a method for enhancing image quality by considering broader contextual information related to the image. LLMs can analyze textual descriptions and metadata to gain insights into the conditions under which an image was captured, informing more nuanced denoising strategies. However, effectively leveraging contextual analysis for denoising requires algorithms to interpret and incorporate a wide range of contextual cues, posing challenges in terms of natural language understanding and the translation of these insights into technical denoising parameters. Additionally, the variability and complexity of contextual information associated with images necessitate advanced LLM capabilities and a deep integration between language models and denoising algorithms.
	
	\paragraph{Balancing Noise Reduction and Semantic Integrity}
	Achieving a balance between noise reduction and the preservation of semantic integrity in denoised images involves leveraging LLMs to inform denoising algorithms about the semantic significance of various image regions. This approach aims to ensure that essential features and details are retained during the denoising process, based on their relevance and importance as identified by LLM analysis. Implementing this balance challenges traditional denoising methods by requiring a more sophisticated understanding of image content and its semantic implications. Moreover, the practical application of balancing noise reduction with semantic integrity depends on the LLM's ability to provide accurate and relevant insights, highlighting the need for advanced natural language processing and image analysis techniques.
	
	\paragraph{Mathematical Formulations Informed by Semantic Analysis}
	Translating semantic insights from LLMs into mathematical formulations for denoising algorithms proposes a novel approach to enhancing image quality by preserving semantically significant features. By incorporating LLM-derived understanding of image content into the mathematical modeling of denoising processes, algorithms can adjust their operations to prioritize the preservation of important details and textures. However, the success of this approach hinges on the accurate translation of semantic insights into quantifiable parameters, posing challenges in terms of model complexity and computational demands. Additionally, the effectiveness of mathematical formulations informed by semantic analysis depends on the depth and accuracy of the LLM's interpretation of textual and contextual data, underscoring the importance of advanced natural language understanding capabilities.
	
	\subsubsection{LLM-Guided Anomaly Detection in Noise Patterns}
	\paragraph{Advancing Noise Analysis with Anomaly Detection}
	Employing LLMs for anomaly detection in noise patterns introduces a method for identifying and addressing atypical noise types within images. By analyzing noise characteristics and comparing them with known patterns, LLMs can highlight anomalies that may indicate underlying issues or unique conditions affecting image quality. Implementing LLM-guided anomaly detection challenges conventional denoising methods by requiring algorithms to adapt to a wider range of noise conditions, including rare or unusual patterns. Moreover, the effectiveness of this approach is contingent upon the LLM's ability to accurately recognize and interpret anomalies, necessitating advanced pattern recognition and natural language processing capabilities.
	
	\paragraph{Customized Denoising Solutions Through Anomaly Insights}
	Utilizing LLMs to develop customized denoising solutions based on anomaly detection offers a targeted approach to improving image quality. By identifying specific anomalies in noise patterns, LLMs can inform the development of denoising strategies tailored to address these unique challenges. However, translating anomaly insights into effective denoising solutions requires algorithms to be flexible and adaptable, capable of incorporating a wide range of noise characteristics into their operations. Furthermore, the success of customized denoising solutions depends on the accuracy and relevance of the LLM's anomaly detection, highlighting the need for sophisticated machine learning models and a deep understanding of image noise phenomena.
	
	\paragraph{Semantic and Contextual Analysis for Anomaly Identification}
	Incorporating semantic and contextual analysis through LLMs for anomaly identification in noise patterns introduces a comprehensive approach to understanding and addressing noise in images. By leveraging textual data and image metadata, LLMs can provide insights into potential sources of anomalies and their implications for image quality. Implementing this approach challenges denoising algorithms to interpret and act upon complex semantic and contextual information, necessitating advanced natural language processing and image analysis capabilities. Moreover, the effectiveness of semantic and contextual analysis for anomaly identification depends on the depth and accuracy of the LLM's understanding of the textual and contextual cues associated with images, underscoring the importance of sophisticated language models in the denoising process.
	
	\paragraph{Integrating Anomaly Detection into Denoising Frameworks}
	The integration of LLM-guided anomaly detection into denoising frameworks offers a method for enhancing the adaptability and effectiveness of denoising algorithms. By identifying and addressing anomalies in noise patterns, denoising processes can be optimized to handle a wider range of noise conditions, improving image quality and fidelity. However, effectively integrating anomaly detection into denoising frameworks challenges traditional algorithms to be more flexible and responsive to LLM insights, requiring advanced computational techniques and a deep integration between language models and image processing algorithms. Moreover, the success of this integration depends on the LLM's ability to accurately detect and interpret noise anomalies, highlighting the need for continual improvement and refinement of LLM capabilities.
	
	\paragraph{Enhancing Denoising Efficacy and Precision}
	Enhancing the efficacy and precision of denoising algorithms through LLM-guided anomaly detection involves leveraging the insights provided by LLMs to identify and address specific anomalies in noise patterns. This approach aims to improve the overall quality of denoised images by ensuring that denoising strategies are tailored to the unique characteristics of each image. Implementing this enhancement challenges traditional denoising methods to incorporate LLM insights into their operations, requiring algorithms to be adaptable and capable of handling a diverse range of noise conditions. Furthermore, the effectiveness of enhancing denoising efficacy and precision through LLM-guided anomaly detection depends on the accuracy and relevance of the LLM's analysis, underscoring the importance of advanced machine learning techniques and a deep understanding of noise phenomena in the denoising process.
	
	\subsubsection{Natural Language Processing for Denoising Parameter Optimization}
	\paragraph{Leveraging NLP in Denoising Workflows}
	Integrating natural language processing (NLP) capabilities of large language models into denoising workflows represents a transformative approach to optimizing denoising parameters. By enabling denoising systems to interpret and act upon user preferences and requirements articulated in human language, NLP facilitates a more intuitive and effective communication between users and denoising algorithms. This integration challenges existing denoising pipelines to incorporate NLP techniques seamlessly, enhancing their flexibility and adaptability to user inputs. However, the complexity of processing natural language inputs and mapping them to specific denoising parameters may introduce computational overhead and challenges in ensuring the robustness and reliability of the system. Despite these challenges, NLP's integration into denoising workflows holds the potential to revolutionize user interaction with denoising tools, making advanced image processing techniques more accessible and user-friendly.
	
	\paragraph{Translating Descriptive Inputs into Technical Parameters}
	The capability of LLMs to translate descriptive, often subjective, inputs into technical parameters that guide denoising algorithms is a cornerstone of leveraging NLP in denoising workflows. This process allows users to convey their expectations and preferences in a natural and intuitive manner, enabling the denoising system to adjust its parameters accordingly for optimal results. The success of this approach hinges on the LLM's ability to accurately interpret user inputs and translate them into actionable denoising strategies. While promising, this method poses challenges in terms of the LLM's understanding of the nuances of language and its ability to make appropriate technical adjustments. Addressing these challenges requires ongoing advancements in NLP and machine learning, ensuring that LLMs can effectively bridge the gap between user inputs and technical denoising operations.
	
	\paragraph{Adaptive Parameter Adjustment for Dynamic Scenarios}
	The dynamic nature of image capture and the variability of noise necessitate adaptive parameter adjustments in denoising algorithms, a task well-suited for LLMs equipped with NLP capabilities. By processing updated descriptions or feedback in real-time, LLMs can continually reassess and refine denoising parameters, ensuring that the algorithm remains responsive to evolving user preferences and conditions. This adaptability is crucial for maintaining optimal denoising performance across diverse scenarios. However, achieving real-time adaptability poses challenges in terms of computational efficiency and the ability of the LLM to accurately interpret nuanced feedback. Despite these challenges, the potential for LLM-driven adaptive parameter adjustment to enhance denoising outcomes and user satisfaction makes it a promising area for future research and development.
	
	\paragraph{Enhancing User Interaction with Denoising Tools}
	Incorporating NLP into denoising workflows significantly enhances user interaction with denoising tools, making the process more accessible and engaging. By allowing users to describe their noise reduction goals in natural language, LLMs can interpret these inputs and adjust denoising parameters accordingly, providing a more user-friendly experience. This approach not only democratizes access to sophisticated denoising techniques but also fosters a more dynamic and collaborative denoising process. However, ensuring that LLMs accurately interpret and respond to user inputs requires advanced NLP techniques and a deep understanding of user intentions. Addressing these requirements is essential for realizing the full potential of NLP-enhanced user interaction in denoising workflows.
	
	\paragraph{Optimizing for Human-Perceived Image Quality}
	The ultimate goal of utilizing LLMs for denoising parameter optimization is to align denoising efforts with human-perceived image quality. By grounding parameter adjustments in the rich context provided by textual descriptions, LLMs can help ensure that denoised images not only meet technical quality standards but also satisfy user expectations for visual appeal and realism. Achieving this alignment challenges traditional denoising techniques to incorporate perceptual considerations into their optimization processes, requiring a nuanced understanding of both technical image quality metrics and subjective human preferences. The integration of LLM-driven parameter optimization represents a significant step forward in making denoising algorithms more responsive to human perceptions, enhancing the overall effectiveness and satisfaction of denoising outcomes.
	
	\subsubsection{LLM-Driven Adaptive Filtering Techniques}
	\paragraph{Context-Aware Filtering Through LLM Analysis}
	The integration of LLMs into the image denoising framework enables context-aware filtering strategies that adapt to the unique content and noise characteristics of each image. By analyzing textual metadata and descriptions, LLMs can guide the selection of filtering techniques that are best suited to preserving essential details while effectively reducing noise. This approach promises to improve the precision and efficacy of denoising algorithms but challenges them to interpret and act upon complex LLM-generated insights. Successful implementation requires advanced algorithms capable of dynamic filtering adjustments based on semantic understanding, highlighting the need for ongoing research and development in integrating LLMs with image processing techniques.
	
	\paragraph{Dynamic Selection of Filtering Techniques}
	Utilizing LLMs to dynamically select filtering techniques based on contextual analysis represents a significant advancement in denoising methodologies. By understanding the semantic content of images and associated textual data, LLMs can recommend filtering strategies that optimize noise reduction while preserving critical image features. Implementing this dynamic selection process challenges existing denoising frameworks to be adaptable and responsive to LLM insights, necessitating advancements in algorithm design and computational efficiency. Despite these challenges, the potential for LLM-guided filtering techniques to enhance denoising outcomes and maintain image integrity makes it a promising area for future exploration.
	
	\paragraph{Real-Time Adjustment of Filter Parameters}
	The capability of LLMs to facilitate real-time adjustment of filter parameters in response to ongoing denoising analysis offers a path toward more responsive and effective denoising strategies. By continuously monitoring the denoising process and adjusting parameters based on LLM-generated feedback, denoising algorithms can achieve optimal balance between noise reduction and detail preservation. However, realizing this real-time adaptability poses significant challenges in terms of computational resources and the ability of LLMs to accurately assess denoising efficacy. Addressing these challenges is crucial for developing denoising algorithms that can dynamically optimize filter parameters, enhancing the quality and perceptual fidelity of denoised images.
	
	\paragraph{Enhancing Denoising with Semantic Understanding}
	Incorporating semantic understanding into denoising algorithms through LLM analysis enables a more nuanced approach to noise reduction, prioritizing the preservation of meaningful content and features within images. By leveraging LLMs to interpret image content and its semantic significance, denoising strategies can be tailored to maintain essential details while effectively mitigating noise. Implementing this approach challenges traditional denoising methods to integrate semantic insights into their operations, requiring sophisticated algorithms capable of discerning and preserving semantic content. Despite these challenges, the integration of semantic understanding holds promise for advancing denoising techniques, leading to improved image quality and enhanced visual experiences.
	
	\paragraph{Bridging Human Perception and Technical Processes}
	LLM-driven adaptive filtering techniques represent a bridge between human perception and technical denoising processes, offering a more intuitive and effective approach to image enhancement. By translating human descriptions and preferences into technical denoising strategies, LLMs facilitate a user-centric denoising process that aligns with human visual expectations. However, achieving this bridge poses challenges in accurately interpreting human input and adapting denoising algorithms accordingly. Addressing these challenges is essential for realizing the full potential of LLM-driven denoising, enhancing both the technical performance of denoising algorithms and their alignment with human perceptual standards.
	
	\subsubsection{Textual Feedback Loop for Denoising Refinement}
	\paragraph{Innovative Iterative Refinement Through Textual Feedback}
	Integrating a textual feedback loop into the denoising process, facilitated by LLMs, introduces an innovative approach to refining denoising outcomes. By generating descriptive feedback on denoised images, LLMs enable a continuous cycle of evaluation and adjustment, enhancing the quality of denoising algorithms. Implementing this feedback loop challenges traditional denoising methods to be responsive to textual insights, requiring advancements in natural language processing and image analysis. Despite these challenges, the potential for textual feedback to drive iterative refinement and improve denoising efficacy makes it a promising avenue for enhancing image processing workflows.
	
	\paragraph{LLM as a Collaborative Partner in Image Denoising}
	Positioning LLMs as collaborative partners in the denoising process transforms the traditional approach to image enhancement, fostering a dialogue between users and denoising algorithms. Through conversational interaction, LLMs can interpret user feedback and guide the refinement of denoising strategies, enhancing user involvement and satisfaction. Implementing this collaborative approach challenges denoising systems to integrate conversational AI capabilities, necessitating advancements in user interface design and algorithm adaptability. Despite these challenges, the integration of LLMs as collaborative partners holds promise for creating more responsive and user-centric denoising processes, leading to improved image quality and user experiences.
	
	\paragraph{Translating Textual Feedback into Actionable Adjustments}
	The ability of LLMs to translate textual feedback into actionable adjustments for denoising algorithms represents a significant advancement in image processing. By interpreting user feedback and guiding algorithmic refinements, LLMs enable a more dynamic and responsive denoising process. Implementing this translation process challenges existing algorithms to be adaptable and capable of incorporating feedback-driven adjustments, requiring sophisticated natural language understanding and algorithm design. Despite these challenges, leveraging textual feedback for denoising optimization holds promise for enhancing the effectiveness and user satisfaction of denoising workflows, leading to higher-quality denoised images.
	
	\paragraph{Enhanced User Engagement and Customization}
	Incorporating LLM-based evaluation metrics and feedback mechanisms into the denoising process significantly enhances user engagement and allows for greater customization of denoising outcomes. By enabling users to provide input and receive feedback through natural language, LLMs facilitate a more interactive and personalized denoising experience. Implementing this user-centric approach challenges denoising algorithms to be responsive to individual preferences and adaptable to feedback, necessitating advancements in conversational AI and user interface design. Despite these challenges, the potential for enhanced user engagement and customization through LLM integration offers promising opportunities for improving the accessibility and effectiveness of denoising tools, leading to more satisfying and tailored image enhancement results.
	
	\paragraph{Closing the Loop for Optimal Denoising Outcomes}
	The establishment of a textual feedback loop, powered by LLMs, for refining denoising outcomes represents a holistic approach to image enhancement. By facilitating continuous interaction between users and denoising algorithms, LLMs enable iterative adjustments based on descriptive feedback, driving improvements in denoising efficacy and user satisfaction. Implementing this feedback loop challenges traditional denoising methods to incorporate dynamic user inputs and adapt their operations accordingly, requiring advanced natural language processing and algorithmic flexibility. Despite these challenges, closing the loop through textual feedback holds promise for achieving optimal denoising outcomes, enhancing the quality of denoised images and aligning them more closely with user expectations and perceptual standards.
	
	\subsubsection{Generative Text-to-Image Approaches for Noise Reduction}
	\paragraph{Pioneering Image Restoration with LLMs}
	The application of generative text-to-image approaches, utilizing LLMs for image denoising, introduces a novel method for reconstructing noise-free images based on textual descriptions. This approach challenges traditional denoising techniques by leveraging the generative capabilities of LLMs to visualize and create images that align with described content, potentially bypassing conventional noise reduction limitations. Implementing generative text-to-image methods for denoising requires a deep integration of natural language understanding and image generation technologies, posing challenges in terms of model training and image fidelity. Despite these challenges, the potential for LLM-driven generative approaches to transform image restoration and denoising practices holds promise for significant advancements in image quality enhancement.
	
	\paragraph{Textual Descriptions as Blueprints for Restoration}
	Utilizing textual descriptions as blueprints for image restoration through LLM-driven generative approaches offers a method for precisely tailoring denoising and restoration processes to user-defined content specifications. This strategy allows users to influence the denoising outcome by providing detailed descriptions, which LLMs interpret to guide the generative process. Implementing this approach challenges denoising systems to accurately translate textual inputs into high-quality images, requiring advancements in generative modeling and natural language processing. Despite these challenges, leveraging textual descriptions for image restoration holds promise for enhancing the customizability and effectiveness of denoising workflows, leading to images that more closely align with user expectations and intended content.
	
	\paragraph{Bridging Semantic Gaps in Image Reconstruction}
	The integration of LLMs in bridging semantic gaps during the image reconstruction process represents a significant advancement in denoising methodologies. By leveraging textual descriptions to guide the generation of denoised images, LLMs can ensure that reconstructed images retain the essential content and context lost due to noise interference. Implementing this approach challenges denoising algorithms to interpret and act upon semantic cues provided by LLMs, necessitating advancements in understanding and translating textual data into visual representations. Despite these challenges, the potential for LLMs to bridge semantic gaps in image reconstruction offers promising opportunities for improving the fidelity and contextuality of denoised images, enhancing their usability and interpretability across various applications.
	
	\paragraph{Advancements in Generative AI for Enhanced Fidelity}
	The role of advancements in generative AI, particularly in the context of LLM-driven denoising techniques, is crucial for achieving enhanced fidelity in reconstructed images. By developing models capable of accurately interpreting textual descriptions and generating visually coherent images, generative AI promises to improve the quality and realism of denoised images. Implementing these advancements challenges current generative models to handle complex textual inputs and produce high-resolution images that faithfully represent described content. Despite these challenges, the ongoing evolution of generative AI technologies holds promise for revolutionizing image denoising and restoration practices, leading to significant improvements in image clarity and content preservation.
	
	\paragraph{Expanding the Possibilities of Image Denoising}
	The application of generative text-to-image approaches, facilitated by LLMs, expands the possibilities of image denoising beyond traditional noise reduction techniques. By enabling users to specify desired content through textual descriptions, LLMs can guide the generative process to produce images that not only reduce noise but also enhance visual content and fidelity. Implementing this innovative approach challenges denoising systems to integrate advanced generative models and natural language processing capabilities, necessitating ongoing research and development. Despite these challenges, the potential for LLM-driven generative approaches to transform image denoising and restoration practices offers exciting opportunities for advancing image quality and expanding the creative possibilities of image enhancement.
	
	\subsubsection{LLM-Assisted Edge and Texture Preservation}
	\paragraph{Contextual Analysis for Feature Preservation}
	The integration of LLMs for contextual analysis in the denoising process introduces a nuanced approach to preserving essential image features such as edges and textures. By leveraging LLMs to interpret associated textual data, denoising algorithms can be informed about areas within the image where feature preservation is critical. Implementing this approach challenges traditional denoising methods to adapt to LLM insights, requiring advancements in algorithmic flexibility and natural language processing capabilities. Despite these challenges, the potential for LLM-assisted contextual analysis to enhance feature preservation in denoised images holds promise for improving image quality and maintaining visual integrity across various applications.
	
	\paragraph{Identifying Features Requiring Preservation}
	Employing LLMs to identify specific features within an image that require preservation during the denoising process offers a targeted approach to maintaining image fidelity. By analyzing textual descriptions and metadata, LLMs can highlight areas of significance, guiding denoising algorithms to prioritize the retention of critical details. Implementing this feature identification process challenges denoising systems to integrate LLM-generated insights, necessitating advancements in image analysis and natural language understanding. Despite these challenges, leveraging LLMs for feature identification holds promise for enhancing the precision and effectiveness of denoising algorithms, leading to improved preservation of essential image characteristics.
	
	\paragraph{Adaptive Denoising Strategies}
	The development of adaptive denoising strategies, informed by LLM analysis, represents a significant advancement in image processing. By leveraging LLM-generated insights into image content and context, denoising algorithms can dynamically adjust their operations to preserve important features while effectively reducing noise. Implementing adaptive denoising strategies challenges traditional algorithms to be responsive to LLM insights, requiring advancements in algorithm design and natural language processing. Despite these challenges, the potential for LLM-informed adaptive denoising to improve image quality and fidelity makes it a promising area for future research and development.
	
	\paragraph{LLM-Guided Parameter Optimization}
	Incorporating LLMs for the optimization of denoising parameters introduces a method for enhancing the precision and effectiveness of denoising algorithms. By analyzing textual feedback and descriptions, LLMs can guide the adjustment of denoising parameters to better preserve image features and improve overall image quality. Implementing LLM-guided parameter optimization challenges denoising systems to accurately interpret and act upon LLM insights, necessitating advancements in natural language understanding and algorithmic adaptability. Despite these challenges, leveraging LLMs for parameter optimization holds promise for refining denoising strategies and achieving superior denoising outcomes.
	
	\paragraph{Enhancing Perceptual Quality of Denoised Images}
	The integration of LLMs in denoising algorithms for the purpose of enhancing the perceptual quality of denoised images introduces a holistic approach to image enhancement. By leveraging LLM-generated insights to preserve essential features and maintain visual integrity, denoising processes can produce images that not only exhibit reduced noise but also enhanced visual appeal. Implementing this approach challenges traditional denoising methods to incorporate perceptual considerations into their operations, requiring advancements in natural language processing and image analysis. Despite these challenges, the potential for LLM-enhanced denoising to improve the perceptual quality of images offers promising opportunities for advancing image processing techniques and enhancing visual experiences.
	
	\subsubsection{Interactive Denoising Through Conversational AI}
	\paragraph{Revolutionizing Denoising with User Interaction}
	Integrating user interaction into the denoising process through conversational AI and LLMs represents a transformative approach to image enhancement. By enabling direct communication between users and denoising algorithms, this interactive framework allows for personalized denoising strategies based on user input. Implementing conversational AI in denoising challenges existing algorithms to be responsive to natural language inputs, requiring advancements in conversational AI technologies and user interface design. Despite these challenges, the potential for interactive denoising to enhance user engagement and achieve tailored denoising outcomes holds promise for revolutionizing the field of image processing.
	
	\paragraph{Customized Denoising Based on Dialogue}
	Employing conversational AI for customized denoising based on user dialogue introduces a dynamic and personalized approach to image enhancement. By interpreting user preferences and feedback expressed in natural language, conversational AI can guide denoising algorithms to adjust their operations accordingly. Implementing this dialogue-based customization challenges denoising systems to accurately interpret and act upon user inputs, necessitating advancements in natural language understanding and algorithm adaptability. Despite these challenges, leveraging conversational AI for customized denoising holds promise for improving the relevance and effectiveness of denoising outcomes, leading to enhanced user satisfaction and image quality.
	
	\paragraph{Adapting Denoising in Real-Time}
	The capability of conversational AI to adapt denoising parameters in real-time based on ongoing user interaction represents a significant advancement in image processing. By enabling a continuous dialogue between users and the denoising system, conversational AI ensures that denoising strategies remain aligned with user expectations and evolving preferences. Implementing real-time adaptation challenges denoising algorithms to be agile and responsive to conversational inputs, requiring advancements in AI technologies and computational efficiency. Despite these challenges, the potential for conversational AI-driven adaptation to enhance the responsiveness and user-centricity of denoising processes holds promise for advancing the field of image enhancement.
	
	\paragraph{Enhancing User Experience and Satisfaction}
	Incorporating conversational AI into denoising workflows significantly enhances the user experience, making the denoising process more intuitive and engaging. By facilitating natural language interaction between users and denoising algorithms, conversational AI enables users to express their preferences and receive personalized feedback, leading to improved user satisfaction. Implementing this user-centric approach challenges denoising systems to integrate advanced conversational AI capabilities, necessitating ongoing research and development. Despite these challenges, leveraging conversational AI for enhancing user interaction holds promise for democratizing access to advanced denoising techniques and fostering a more satisfying and personalized image processing experience.
	
	\paragraph{Broadening the Accessibility of Advanced Denoising}
	Interactive denoising through conversational AI broadens the accessibility of advanced denoising techniques to a wider audience. By simplifying the interaction with denoising tools through natural language communication, conversational AI enables users of varying expertise levels to benefit from state-of-the-art denoising technologies. Implementing this approach challenges existing denoising frameworks to be user-friendly and adaptable to conversational inputs, requiring advancements in AI and user interface design. Despite these challenges, the potential for conversational AI to democratize access to sophisticated denoising capabilities offers exciting opportunities for enhancing image quality and user engagement across various applications.
	
	\subsubsection{LLM-Based Evaluation Metrics for Denoised Images}
	\paragraph{Innovating Image Quality Assessment with LLMs}
	Developing LLM-based evaluation metrics for denoised images introduces a novel approach to assessing image quality and the effectiveness of denoising techniques. By leveraging the natural language processing capabilities of LLMs, this method offers a more nuanced and comprehensive evaluation of denoised images, considering both technical and perceptual aspects of image quality. Implementing LLM-based metrics challenges traditional evaluation methods to incorporate linguistic analysis, requiring advancements in LLM technologies and integration with image processing workflows. Despite these challenges, the potential for LLM-based evaluation metrics to provide more relevant and user-centric assessments of denoised images holds promise for advancing the field of image denoising and improving the quality of visual content.
	
	\paragraph{Bridging Technical and Perceptual Quality}
	The integration of LLM-based evaluation metrics for denoised images represents a significant advancement in bridging the gap between technical image quality metrics and perceptual image quality assessments. By leveraging the deep semantic understanding of LLMs, these metrics offer a comprehensive evaluation framework that considers the nuanced aspects of image quality important to human viewers. Implementing LLM-based metrics challenges traditional image quality assessment methods to adapt to semantic analysis, necessitating advancements in natural language processing and image analysis technologies. Despite these challenges, the potential for LLM-based metrics to enhance the accuracy and relevance of image quality evaluations holds promise for improving the effectiveness of denoising algorithms and ensuring that denoised images meet both technical and perceptual quality standards.
	
	\paragraph{Dynamic Feedback for Denoising Optimization}
	Utilizing LLM-based evaluation metrics to provide dynamic feedback for denoising optimization introduces a method for continuously refining denoising processes based on textual feedback. By analyzing the perceived quality of denoised images through natural language, LLMs can guide the adjustment of denoising parameters and strategies for improved outcomes. Implementing this feedback mechanism challenges denoising algorithms to be responsive to LLM-generated insights, requiring advancements in feedback integration and algorithm adaptability. Despite these challenges, the potential for dynamic feedback to drive iterative refinement and enhance denoising efficacy makes it a promising area for advancing image processing techniques and achieving higher-quality denoised images.
	
	\paragraph{Enhanced User Interaction and Satisfaction}
	Incorporating LLM-based evaluation metrics into the denoising process significantly enhances user interaction and satisfaction by enabling a more intuitive and personalized assessment of denoised images. By leveraging natural language to express evaluations and feedback, LLMs facilitate a user-centric approach to image quality assessment, allowing users to convey their perceptions and preferences effectively. Implementing this approach challenges denoising systems to integrate LLM capabilities, necessitating advancements in conversational AI and user interface design. Despite these challenges, leveraging LLM-based metrics for enhanced user interaction holds promise for democratizing access to advanced denoising techniques and fostering a more satisfying and tailored image processing experience.
	
	\paragraph{Setting New Standards for Image Denoising}
	The development of LLM-based evaluation metrics for denoised images sets new standards for image quality assessment in the field of image processing. By offering a more comprehensive and nuanced evaluation framework that considers both technical and perceptual aspects of image quality, LLM-based metrics promise to enhance the accuracy and relevance of denoising evaluations. Implementing these metrics challenges traditional denoising methodologies to incorporate advanced natural language processing and semantic analysis, necessitating ongoing research and development. Despite these challenges, the potential for LLM-based metrics to transform image quality assessment and denoising practices holds promise for advancing the field of image processing and improving the quality of visual content across various applications.

	\subsubsection{Pseudocode for Algogenic Image Denoising}
	The Algogenic image denoising approach harnesses AI to enhance conventional methods by dynamically adjusting parameters and strategies based on the observed behavior of the system and real-time error estimates. This pseudocode, available in \ref{fig:image-denoising-Algogen-pseudocode}, outlines an advanced framework incorporating AI-driven enhancements for adaptive noise reduction, pixel selection, quality assessment, and real-time parameter optimization.
	
	\begin{algorithm}
		\caption{Algogenic Image Denoising Pseudocode}
		\begin{algorithmic}[1]
			\Procedure{AlgogenicImageDenoising}{Image, Metadata}
			
			\Comment{Preprocessing Phase}
			\State noiseType $\gets$ LLMBasedNoiseCharacterization(Metadata)
			\State importantAreas $\gets$ SemanticUnderstandingForSelectiveDenoising(Image, Metadata)
			
			\Comment{Core Denoising Phase}
			\State denoisingAlgorithm $\gets$ SelectDenoisingAlgorithm(noiseType)
			\If{AdaptiveFilteringNeeded(Image, noiseType)}
			\State anomalyDetection $\gets$ LLMGuidedAnomalyDetectionInNoisePatterns(Image)
			\State denoisingAlgorithm $\gets$ AdjustAlgorithmBasedOnAnomaly(anomalyDetection)
			\EndIf
			\State denoisedImage $\gets$ ApplyDenoisingAlgorithm(Image, denoisingAlgorithm, importantAreas)
			\Repeat
			\State feedback $\gets$ InteractiveDenoisingThroughConversationalAI(denoisedImage)
			\State denoisedImage $\gets$ RefineDenoisingBasedOnFeedback(denoisedImage, feedback)
			\Until{SatisfactionAchieved(feedback)}
			\State denoisedImage $\gets$ NLPForDenoisingParameterOptimization(denoisedImage, feedback)
			\State denoisedImage $\gets$ LLMAssistedEdgeAndTexturePreservation(denoisedImage, importantAreas)
			
			\Comment{Postprocessing Phase}
			\State finalImage $\gets$ GenerativeTextToImageApproachesForNoiseReduction(denoisedImage, Metadata)
			\State EvaluateDenoisedImage $\gets$ LLMBasedEvaluationMetricsForDenoisedImages(finalImage)
			
			\EndProcedure
		\end{algorithmic}\label{fig:image-denoising-Algogen-pseudocode}
	\end{algorithm}

	\begin{figure}
		\centering
		\includegraphics[width=0.47\textwidth]{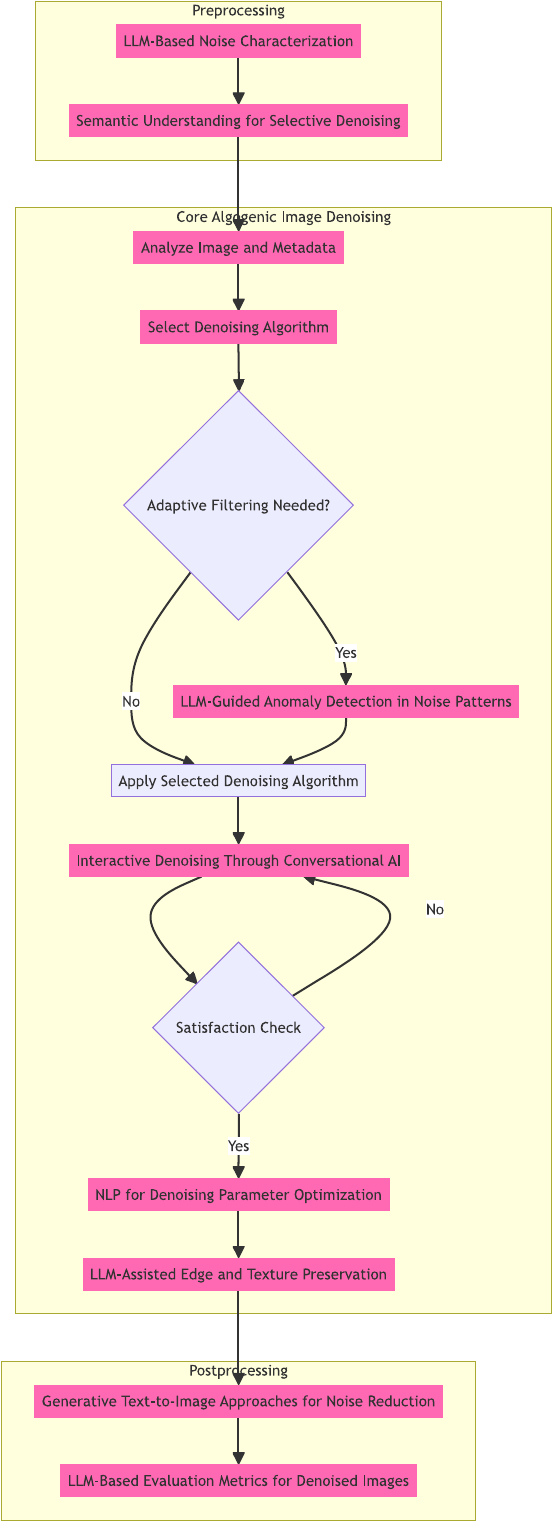}
		\caption{Integrating Algogens with Image Denoising: This diagram visualizes the advanced Algogenic Image Denoising framework, emphasizing the integration of generative AI enhancements at every stage of the process. It highlights the initial LLM-based noise characterization and semantic understanding for selective denoising in the preprocessing phase. The core denoising phase is shown to adaptively select and adjust denoising algorithms based on AI-driven anomaly detection and interactive user feedback, ensuring both noise reduction and preservation of critical image details. Postprocessing leverages generative text-to-image approaches and LLM-based evaluation metrics to refine and assess the denoised images, showcasing the synergy between AI-driven insights and traditional denoising methods for optimal image quality. This integration significantly enhances the denoising process's adaptability, effectiveness, and sensitivity to the semantic content of images.}
		\label{fig:image_denoising}
	\end{figure}

	\section{Super-Resolution}\index{Super-Resolution}
	\subsection{Introduction to Super-Resolution}
	\subsubsection{The Concept of Super-Resolution}
	
	\paragraph{Definition and Objectives}
	Super-resolution (SR) refers to the process of reconstructing a high-resolution (HR) image from one or more low-resolution (LR) observations of the same scene. The primary objective of SR is to recover the finer details and textures that are lost during the downsampling process, which might occur due to limitations in the imaging sensor, compression algorithms, or intentional downscaling. By enhancing the resolution of images, SR aims to improve the visual quality and usability of images for various applications. \textbf{Furthermore}, the advancement in SR techniques has significant implications in fields such as medical imaging, where precise visualization of anatomical structures is crucial for diagnosis and treatment planning. \textbf{Moreover}, in satellite imaging, SR enables the extraction of finer spatial information, enhancing the ability to monitor environmental changes, urban development, and agricultural patterns with greater accuracy. \textbf{Similarly}, in video enhancement, SR can enhance the clarity and detail of video frames, leading to better quality playback and improved analysis in surveillance systems or digital entertainment. \textbf{Additionally}, the utility of SR extends to domains like forensics, where enhancing the resolution of surveillance footage can aid in identifying perpetrators or crucial details in criminal investigations. In essence, SR serves as a vital tool across diverse fields, \textbf{therefore} driving research and development towards more robust and efficient algorithms to address the increasing demand for high-quality imaging solutions.

	\paragraph{Underlying Principles}
	The foundation of SR techniques lies in the exploitation of underlying redundancies and correlations present in the LR images. \textit{Moreover}, through sophisticated mathematical models and algorithms, SR methods aim to infer the high-frequency components that are not captured in the LR images. This process involves a combination of image registration, \textit{which} aligns multiple observations of the same scene, interpolation, \textit{which} estimates the pixel values in the HR grid, \textit{and} reconstruction algorithms that integrate information from all available LR images to produce a single HR image. The effectiveness of SR methods heavily depends on the accuracy of these steps \textit{and} the assumption that the scene contains sufficient information at different scales. \textit{Additionally}, the success of SR techniques relies on the quality of the registration process, ensuring precise alignment of the LR images. \textit{Furthermore}, the interpolation step plays a critical role in estimating the missing high-frequency details by inferring pixel values at higher resolutions. \textit{On the other hand}, reconstruction algorithms, such as convolutional neural networks (CNNs), leverage the information from multiple LR images to reconstruct a high-resolution version, exploiting correlations across different scales and orientations. \textit{Consequently}, the integration of these processes results in the generation of HR images with enhanced visual quality, providing finer details and sharper edges, which are crucial for various applications ranging from medical imaging to satellite imagery analysis.

	\paragraph{Challenges in Super-Resolution}
	One of the significant challenges in SR is dealing with the ill-posed nature of the problem. \textbf{Moreover}, since there are infinitely many HR images that can correspond to a single LR image, constraints and prior knowledge about the image content are necessary to guide the reconstruction process towards a plausible solution. This prior knowledge can be in the form of regularization terms in the optimization framework, which penalize unlikely image features, or learned priors from a dataset of HR images using machine learning techniques. \textbf{Furthermore}, noise in the LR images, inaccuracies in image registration, and computational constraints pose further challenges in achieving high-quality SR reconstructions.
	
	The ill-posed nature of SR problem makes it challenging due to its underdetermined nature, where multiple solutions are possible. \textbf{However}, by incorporating constraints and prior knowledge, the solution space can be constrained, leading to more accurate and visually pleasing HR images. Regularization terms play a crucial role in balancing the fidelity of the reconstructed image with the preservation of important image characteristics. \textbf{Additionally}, machine learning techniques such as deep neural networks have shown promise in learning complex priors directly from data, improving the quality of SR results. 
	
	Addressing noise in LR images is essential \textbf{since} it can degrade the quality of the reconstructed HR images. Various denoising techniques, such as filtering or learning-based methods, are \textbf{therefore} employed to enhance the robustness of SR algorithms against noise. \textbf{In addition}, inaccuracies in image registration, which occur due to motion or alignment errors, need to be compensated for to ensure accurate reconstruction. This requires robust registration algorithms capable of handling various types of distortions.
	
	\textbf{Moreover}, computational constraints present practical challenges in SR, as high-quality reconstructions often require significant computational resources. Efficient algorithms and optimizations are \textbf{furthermore} needed to make SR feasible for real-time applications or resource-limited devices. Despite these challenges, ongoing research and advancements in SR continue to push the boundaries, offering solutions that bridge the gap between LR and HR imagery.

	\paragraph{Mathematical Formulation}
	The Super-Resolution (SR) problem involves the reconstruction of a High-Resolution (HR) image $I_{HR}$ from a Low-Resolution (LR) counterpart $I_{LR}$ and can be conceptualized as an inverse problem. In this scenario, the aim is to estimate $I_{HR}$ based on the information provided by $I_{LR}$. This relationship is succinctly captured in the equation:
	\[I_{LR} = D \left( H \left( I_{HR} \right) \right) + n,\]
	where $H$ symbolizes the blurring operator, $D$ represents the downsampling operator, and $n$ accounts for any additive noise present in the LR image. This formulation implies that the LR image is a result of applying the blurring operation followed by downsampling on the HR image, corrupted by noise.
	
	Resolving the SR task thus entails the recovery of $I_{HR}$ from $I_{LR}$. However, this process is inherently challenging due to the need to reverse the effects of $H$ and $D$ while considering the influence of the additive noise $n$. Typically, this is achieved by formulating an optimization problem that strikes a balance between fidelity to the observed LR data and regularization constraints that encapsulate prior knowledge about the characteristics of HR images.
	
	Various optimization techniques, such as iterative algorithms or deep learning approaches, may be employed to address this problem. However, the overarching objective remains consistent: to reconstruct a faithful HR image that aligns with the observed LR input while leveraging additional information or constraints to enhance the quality of the reconstruction.

	\subsubsection{Key Principles and Mechanisms}
	
	\paragraph{Foundational Aspects}
	The key principles and mechanisms underlying Super-Resolution (SR) technology revolve around the concepts of signal processing, machine learning, and computational imaging. At the core, SR techniques seek to enhance the spatial resolution of images by employing advanced algorithms to infer missing high-frequency details that are absent in the low-resolution (LR) inputs. These methodologies are grounded in a deep understanding of image properties, the nature of image degradation processes, and the potential to reverse these processes to some extent.
	
	Furthermore, the synergy between signal processing and machine learning is particularly noteworthy. While traditional signal processing techniques focus on deterministic algorithms to process and manipulate signals, machine learning introduces a paradigm shift by enabling systems to learn patterns and relationships directly from data. This integration allows SR algorithms to adapt and optimize their performance based on the specific characteristics of input images, thereby achieving unprecedented levels of resolution enhancement.
	
	Moreover, the role of computational imaging cannot be overstated in the context of SR. Computational imaging leverages computational methods to manipulate and reconstruct images, transcending the limitations imposed by physical optics. By exploiting mathematical models and iterative optimization, computational imaging techniques complement traditional imaging systems, paving the way for novel approaches to resolution enhancement.
	
	Additionally, the interdisciplinary nature of SR underscores its significance in various domains, including biomedical imaging, remote sensing, and surveillance. The ability to reconstruct high-resolution images from degraded inputs has profound implications for medical diagnosis, environmental monitoring, and security applications. Consequently, the continued advancement of SR technology promises to revolutionize diverse fields by unlocking new insights and capabilities through enhanced image resolution.

	\paragraph{Image Reconstruction and Upsampling}
	Image reconstruction and upsampling play pivotal roles in Super-Resolution (SR), constituting the process of enhancing the resolution of low-resolution (LR) images to approximate their high-resolution (HR) counterparts. This enhancement involves enlarging the pixel grid of LR images while striving to introduce new details that closely resemble those present in HR images. Various techniques are employed for upsampling, ranging from basic interpolation methods like bicubic or Lanczos resampling to advanced methodologies leveraging deep learning models. 
	
	Interpolation methods, such as bicubic or Lanczos resampling, operate on the principle of approximating new pixel values based on surrounding pixel intensities. While these techniques are computationally efficient and widely used, they often fail to capture intricate details and may result in blurred outputs. Conversely, deep learning-based approaches harness the power of neural networks to learn complex mappings between LR and HR image spaces. These models exploit large datasets to infer high-frequency details and structural information, enabling them to generate visually compelling reconstructions.
	
	Furthermore, deep learning models exhibit adaptability to diverse image types and content complexities, making them suitable for a wide range of SR applications. By learning from abundant data, these models can discern patterns and relationships within images, facilitating the synthesis of realistic textures and fine details. However, it's essential to address challenges such as model complexity, training data quality, and computational requirements when employing deep learning-based upsampling techniques.
	
	In summary, while traditional interpolation methods offer simplicity and efficiency, deep learning-driven upsampling approaches present unparalleled capabilities in generating high-quality HR reconstructions from LR inputs. The integration of advanced algorithms and neural network architectures continues to push the boundaries of image reconstruction, promising further advancements in the field of Super-Resolution.

	\paragraph{Learning from Data}
	A significant advancement in SR has come from the application of machine learning, particularly deep learning. Convolutional Neural Networks (\textbf{CNNs}) and Generative Adversarial Networks (\textbf{GANs}) have been successfully applied to learn complex mappings from LR to HR images. These models are trained on large datasets of image pairs \textbf{or} are trained in an unsupervised manner to generate HR images that are visually pleasing and rich in details. The learning process involves optimizing the models to reduce the difference between the generated image and the ground truth HR image, guided by loss functions that can assess both pixel-wise accuracy and perceptual quality.
	
	\textbf{Furthermore}, the use of CNNs and GANs allows for the exploitation of hierarchical features present in the images, capturing intricate patterns and textures that contribute to the enhancement of resolution. \textbf{Moreover}, the iterative nature of deep learning techniques enables continuous refinement of the models, leading to increasingly accurate and realistic HR image generation. \textbf{Additionally}, the ability of GANs to learn from data distributions and generate novel, high-quality images \textbf{in parallel with} CNNs' capacity to perform end-to-end learning enhances the overall performance of SR systems.
	
	\textbf{On the other hand}, while deep learning methods have shown remarkable success in SR, challenges such as computational complexity and the need for large amounts of annotated data \textbf{remain}. \textbf{Nevertheless}, ongoing research efforts focus on addressing these challenges through innovations in network architectures, optimization algorithms, and data augmentation techniques. The integration of domain-specific knowledge \textbf{also} plays a crucial role in improving the generalization capability of SR models, making them applicable across diverse domains and scenarios.
	
	\textbf{In conclusion}, the synergy between machine learning techniques, particularly CNNs and GANs, has revolutionized the field of SR by enabling the generation of high-quality HR images from LR inputs. \textbf{Consequently}, these advancements have profound implications across various domains, including medical imaging, remote sensing, and entertainment, where high-resolution imagery is essential for accurate analysis and visualization.

	\paragraph{Regularization and Prior Knowledge}
	Incorporating regularization and prior knowledge into SR algorithms is crucial for dealing with the ill-posed nature of the problem. Regularization techniques, such as Tikhonov regularization, Total Variation (TV) minimization, and sparsity-induced norms, are used to impose smoothness, edge preservation, or sparsity constraints on the solution. These constraints help in guiding the reconstruction process towards more plausible HR images by incorporating assumptions about the image content, such as natural image statistics, edge distributions, or texture patterns.
	
	Moreover, regularization methods like Tikhonov regularization contribute significantly to stabilizing the inversion process by balancing the fidelity to the observed data and the imposed prior knowledge, preventing overfitting and enhancing generalization to unseen data. Additionally, Total Variation (TV) minimization aids in preserving edges and discontinuities in the reconstructed image, which is crucial for maintaining the visual quality and perceptual fidelity of the super-resolved output. Furthermore, sparsity-induced norms play a pivotal role in promoting simplicity and structure in the solution space, enabling the extraction of essential features and patterns from the low-resolution input.
	
	Furthermore, incorporating prior knowledge about the underlying image structure and characteristics enhances the robustness and effectiveness of SR algorithms. By leveraging prior information such as statistical regularities in natural images, known distributions of edges, and common texture patterns, SR methods can generate more realistic and visually pleasing HR reconstructions. Additionally, integrating domain-specific knowledge or task-specific constraints, such as geometric transformations or object shapes, can further refine the super-resolved results, tailoring them to specific application domains or objectives.

	\paragraph{Optimization Techniques}
	The SR problem is often formulated as an optimization problem, where the objective is to find the HR image that best explains the observed LR images while satisfying the imposed regularization constraints. This involves solving complex optimization problems that can be computationally intensive. Various optimization techniques, including gradient descent, conjugate gradient methods, and more recently, deep learning-based optimization methods, are employed to efficiently solve these problems. The choice of optimization technique and its implementation details are critical for the performance and effectiveness of SR methods.
	
	Furthermore, the selection of the appropriate optimization method greatly influences the computational efficiency and convergence rate of SR algorithms. Gradient descent, a widely used optimization technique, iteratively updates the parameters in the direction of the negative gradient of a cost function, aiming to minimize the error between the predicted and ground truth HR images. While gradient descent is simple to implement and effective in convex optimization problems, it may suffer from slow convergence in non-convex optimization scenarios encountered in SR tasks.
	
	Moreover, conjugate gradient methods offer an alternative approach by efficiently minimizing quadratic functions without the need for computing the full Hessian matrix. These methods maintain conjugacy between successive search directions, resulting in faster convergence compared to standard gradient descent. However, their applicability may be limited by the requirement of the function being quadratic, which is not always the case in SR optimization problems.
	
	Additionally, the emergence of deep learning-based optimization methods, such as stochastic gradient descent (SGD) and Adam optimization, has revolutionized the field of SR. These techniques leverage neural networks to approximate the mapping between LR and HR images, allowing for end-to-end learning of the SR model parameters. Despite their success, deep learning-based approaches often require substantial computational resources for training and may suffer from overfitting if not properly regularized.
	
	Therefore, the selection of an optimization technique for SR is a nuanced decision, balancing computational efficiency, convergence properties, and the complexity of the problem at hand. Further research into novel optimization algorithms tailored specifically for SR applications is essential to push the boundaries of achievable performance.

	\paragraph{Multi-frame Super-Resolution}
	Beyond single-image SR, multi-frame SR techniques exploit additional information available in sequences of images, such as video frames or multiple photographs of the same scene taken from slightly different viewpoints. These methods involve motion estimation and alignment processes to combine information from multiple LR images to reconstruct the HR image. The use of multiple frames allows for better handling of noise and can provide more information for reconstructing the HR image, making multi-frame SR particularly effective in certain applications.
	
	Furthermore, the incorporation of multiple frames not only aids in noise reduction but also enhances the robustness of the super-resolution process. Moreover, by leveraging the temporal coherence present in consecutive frames, multi-frame SR methods can achieve more accurate motion estimation, thereby enabling more precise alignment and fusion of information from different viewpoints. Additionally, the utilization of multiple LR images facilitates the exploitation of redundant information across frames, leading to improved reconstruction quality and detail preservation. On the other hand, while single-image SR approaches may struggle with noisy input or ambiguous features, multi-frame techniques can mitigate these challenges by aggregating information from multiple sources. Furthermore, the inherent redundancy among consecutive frames allows for the correction of artifacts and inconsistencies that may arise during the super-resolution process.

	In summary, the principles and mechanisms of SR are grounded in a multidisciplinary approach that combines elements of signal processing, computational imaging, and machine learning. The continuous evolution of these techniques, driven by advancements in algorithms, computing power, and data availability, is pushing the boundaries of what can be achieved in image resolution enhancement.

	\subsubsection{The Role of Deep Learning}
	
	\paragraph{Transformation in Super-Resolution}
	Deep Learning has revolutionized the field of Super-Resolution (SR) by introducing models that can learn complex, non-linear mappings from low-resolution (LR) to high-resolution (HR) images. \textbf{Moreover}, the advent of Convolutional Neural Networks (CNNs) and subsequent architectures like Generative Adversarial Networks (GANs) and Residual Neural Networks have shifted the paradigm from traditional algorithmic approaches to data-driven methods. These deep learning models leverage vast amounts of image data to learn the underlying patterns and textures that characterize high-quality images, \textbf{furthermore}, enabling them to reconstruct HR images with remarkable detail and accuracy. \textbf{Additionally}, by exploiting the hierarchical structure of deep networks, these models can capture intricate features at multiple scales, thus enhancing their ability to generate realistic and visually appealing HR images. \textbf{On the other hand}, while these approaches have demonstrated impressive performance, challenges such as computational complexity and the need for large-scale annotated datasets persist. However, \textbf{despite} these challenges, ongoing research efforts continue to refine and improve SR models, \textbf{ultimately} pushing the boundaries of image reconstruction and enhancing various applications ranging from medical imaging to satellite imagery analysis.

	\paragraph{Convolutional Neural Networks (CNNs)}
	CNNs, hailed as pioneering deep learning methodologies, stand as the cornerstone of Super-Resolution (SR) techniques. Their architecture is meticulously crafted to autonomously discern and progressively assimilate spatial hierarchies of features from image data. Specifically tailored for SR tasks, CNNs excel in capturing intricate interdependencies between Low-Resolution (LR) and High-Resolution (HR) images, leveraging multilayered convolutional filters. Through this intricate network, CNNs adeptly learn the intricate mappings required for upscaling images, a feat beyond the capabilities of conventional interpolation techniques.
	
	The intrinsic prowess of CNNs lies in their ability to retain essential edge characteristics and intricate textural nuances while executing the upscaling process. Unlike conventional methods that often result in the loss of such crucial details, CNNs preserve them meticulously, ensuring the fidelity and visual appeal of the generated SR images remain unparalleled.

	\paragraph{Generative Adversarial Networks (GANs)}
	GANs have further enhanced the capabilities of SR by introducing a competitive framework where two networks, a generator and a discriminator, are trained simultaneously. The generator aims to produce HR images that are indistinguishable from real HR images, while the discriminator evaluates the authenticity of the generated images. This adversarial training process encourages the generation of HR images that are not only high in resolution but also realistic in terms of texture and details, pushing the boundaries of perceptual quality in SR.
	
	Additionally, GANs foster innovation in SR through their ability to capture complex high-dimensional data distributions. The adversarial nature of GANs fosters a dynamic equilibrium between the generator and discriminator, allowing for continuous improvement in image quality. Furthermore, GANs address the challenge of generating diverse and photorealistic images, which is crucial for various applications such as image editing and synthesis. Moreover, GANs offer a versatile framework for incorporating additional constraints or objectives, such as style transfer or domain adaptation, enabling tailored solutions for diverse SR tasks. Thus, GANs play a pivotal role in advancing the state-of-the-art in SR, offering both theoretical insights and practical tools for generating high-quality images.

	\paragraph{Loss Functions and Perceptual Quality}
	Deep learning models for SR are trained using loss functions that measure the discrepancy between the generated HR images and the ground truth HR images. Traditional loss functions, such as Mean Squared Error (MSE), have been supplemented with perceptual loss functions that assess similarity in feature space, encouraging models to produce images that are visually similar to human perception. This shift towards perceptual quality has led to the development of SR models that prioritize the generation of natural-looking images over strict pixel accuracy.
	
	Furthermore, incorporating perceptual loss functions enhances the ability of SR models to capture high-level features, such as textures, structures, and semantic content, which are crucial for producing visually pleasing results. Moreover, by leveraging pre-trained deep neural networks, such as VGG or ResNet, as feature extractors within the perceptual loss functions, SR models can exploit rich hierarchical representations learned from large-scale image datasets.
	
	Additionally, perceptual loss functions address the limitations of traditional pixel-wise metrics by focusing on perceptually relevant aspects of image quality, mitigating issues related to artifacts and unrealistic textures often observed in images generated solely based on pixel-wise losses. Furthermore, these perceptual loss functions facilitate the generation of images that are not only quantitatively accurate but also aesthetically pleasing to human observers, thereby advancing the state-of-the-art in image super-resolution.
	
	Therefore, the integration of perceptual loss functions into the training process of SR models signifies a paradigm shift towards optimizing for perceptual quality, ultimately leading to significant improvements in the visual fidelity of super-resolved images.

	\paragraph{Transfer Learning and Pre-trained Models}
	The application of transfer learning, along with the utilization of pre-trained models, has revolutionized the landscape of machine learning, particularly in the realm of super-resolution (SR) models. By exploiting the knowledge encoded in models pre-trained on vast and diverse datasets, the arduous task of training SR models from scratch is alleviated. This paradigm shift significantly diminishes the demand for computational resources and time, rendering SR model development more accessible and cost-effective. Furthermore, the adaptability of pre-trained models facilitates their fine-tuning to suit specific image types or domains, thus empowering the tailoring of SR solutions for specialized applications. For instance, in medical imaging, where image quality is paramount for accurate diagnosis, pre-trained models can be fine-tuned to enhance the resolution of medical scans, improving the clarity of diagnostic images. Similarly, in satellite imagery, pre-trained models can be customized to sharpen details in aerial photographs, aiding in various applications such as urban planning, environmental monitoring, and disaster management. Moreover, in the domain of video upscaling, pre-trained models enable the enhancement of video quality by interpolating intermediate frames, resulting in smoother and more visually appealing footage. Therefore, the symbiotic relationship between transfer learning and pre-trained models not only accelerates the development of SR solutions but also enhances their efficacy and versatility across diverse domains.

	\paragraph{Challenges and Future Directions}
	Despite the remarkable successes of deep learning in Super-Resolution (SR), challenges remain, including the need for large volumes of training data, the computational cost of training and deploying deep models, and the risk of generating artifacts or hallucinated details. While deep learning approaches have shown promise in enhancing image resolution, their reliance on extensive labeled data for training presents a bottleneck, especially in domains where obtaining such data is costly or impractical. Moreover, the computational resources required for training and deploying these models can be prohibitive, hindering their widespread adoption, particularly in resource-constrained environments.
	
	Furthermore, the potential for generating artifacts or hallucinated details poses a significant challenge, especially in critical applications such as medical imaging or satellite imagery analysis, where accuracy is paramount. To address these challenges, future research directions are focused on developing more efficient and lightweight models. These models aim to achieve comparable performance to state-of-the-art approaches while reducing the computational burden, thus enabling their deployment on resource-limited devices or platforms.
	
	In addition to improving model efficiency, enhancing the robustness and generalizability of SR techniques is another critical area of focus. Current deep learning models may struggle with generalizing to unseen data or different imaging conditions, leading to suboptimal performance in real-world scenarios. Therefore, research efforts are directed towards designing more robust architectures and training methodologies that can adapt effectively to diverse input conditions, thereby improving the reliability and applicability of SR techniques across various domains.
	
	Moreover, exploring unsupervised and semi-supervised learning approaches holds promise in alleviating the dependency on high-quality labeled data. By leveraging unannotated or weakly annotated data, these approaches seek to learn meaningful representations of image content without explicit supervision, thus potentially reducing the data acquisition and annotation costs associated with traditional supervised learning paradigms.
	
	Overall, addressing these challenges and pursuing these future research directions is crucial for advancing the field of Super-Resolution, enabling its broader adoption across diverse applications and domains.

	Deep learning has undeniably established itself as a cornerstone in the advancement of SR technology, offering unparalleled improvements in image quality and opening new avenues for research and application. Its role in SR exemplifies the transformative potential of deep learning across the domains of image and signal processing.
	
	\subsubsection{Applications and Limitations}
	
	\paragraph{Broad Spectrum of Applications}
	Super-Resolution (SR) techniques, particularly those powered by deep learning, have found applications across a wide array of fields where image quality is paramount. In the realm of medical imaging, SR helps in enhancing the resolution of images such as MRI and CT scans, providing clinicians with more detailed visual information for diagnosis and treatment planning. \textbf{Moreover}, in satellite and aerial imaging, SR plays a pivotal role by improving the clarity and detail of images captured from space or high altitudes. This enhancement is instrumental in environmental monitoring, urban planning, and military surveillance, \textbf{furthermore} underscoring the versatility of SR technologies. 
	
	In consumer electronics, SR is not merely a luxury but a necessity, \textbf{as} it is employed to upscale video content in real-time. By leveraging deep learning algorithms, SR enhances the viewing experience on high-definition displays, ensuring that consumers enjoy sharper and more immersive visuals. 
	
	\textbf{Additionally}, SR finds crucial applications in forensic science, where it assists in improving the quality of surveillance footage. The ability to reconstruct and enhance details from low-resolution images aids law enforcement agencies in investigations and crime scene analysis. 
	
	In digital archives, SR emerges as a powerful tool for preservation and restoration. \textbf{By the same token}, it helps in restoring old films and photographs, breathing new life into historical artifacts. This application not only preserves cultural heritage but also facilitates research and education by making archival material more accessible and comprehensible.

	\paragraph{Limitations and Challenges}
	Despite the impressive capabilities of SR technologies, they are not without limitations and challenges. One major limitation is the dependence on high-quality, large datasets for training deep learning models, which may not be available for all applications. This can lead to models that perform well on benchmark datasets but struggle with images in the wild, showing a lack of generalizability. Computational efficiency is another challenge, as deep learning models, especially those of higher complexity, require significant computational resources for training and inference, limiting their deployment on low-power devices or in real-time applications. Additionally, there is the issue of artifacts and unrealistic details being introduced in the upscaling process, particularly with aggressive SR factors or when dealing with images that contain complex textures or patterns. The balance between enhancing resolution and preserving natural image characteristics remains a delicate one.
	
	However, it's important to note that researchers are actively addressing these challenges through various approaches. For instance, efforts are being made to develop techniques that reduce the reliance on extensive datasets, such as self-supervised learning methods or transfer learning from related tasks. Moreover, advancements in hardware, like specialized accelerators or distributed computing, are improving the computational efficiency of deep learning models, making them more accessible for a wider range of applications. Furthermore, ongoing research is dedicated to refining SR algorithms to mitigate artifacts and preserve image fidelity, employing strategies like adversarial training or incorporating perceptual loss functions. Despite these challenges, the continuous progress in the field offers promising avenues for overcoming the limitations of SR technologies and unlocking their full potential in diverse real-world scenarios.

	\paragraph{Future Directions}
	Addressing the limitations of current SR techniques involves ongoing research and development efforts. One area of focus is the development of more efficient and lightweight models that can provide high-quality SR on devices with limited computational resources. This necessitates exploring novel architectures and algorithms designed to optimize resource usage while maintaining performance. And, with the proliferation of edge devices and IoT applications, such advancements are crucial for enabling on-device super-resolution capabilities, thereby reducing reliance on cloud computing and improving real-time processing efficiency.
	
	Another critical aspect under investigation is the exploration of unsupervised and semi-supervised learning methods. By leveraging unlabeled or partially labeled data, these approaches aim to alleviate the burden of acquiring large annotated datasets, which can be both costly and time-consuming. Incorporating techniques such as self-supervised learning or generative adversarial networks (GANs) holds promise in enhancing model generalization and adaptability across diverse image domains. But, this avenue also poses challenges in terms of ensuring robustness and avoiding overfitting, necessitating further research into regularization techniques and model regularization.
	
	Furthermore, there's a growing interest in integrating domain-specific knowledge into SR models. In fields like medical imaging and satellite imagery, where image characteristics and quality requirements differ significantly from generic visual data, customizing super-resolution techniques becomes imperative. By incorporating domain-specific priors or constraints, such as anatomical structures in medical images or atmospheric conditions in satellite imagery, tailored SR models can deliver more accurate and clinically relevant results. Moreover, synergizing traditional signal processing techniques with deep learning methodologies can lead to hybrid models capable of capturing both low-level details and high-level semantic information, thereby enhancing the interpretability and diagnostic value of super-resolved images.
	
	Efforts are also underway to develop more robust evaluation metrics that can better assess the perceptual quality of super-resolved images. While traditional metrics like PSNR and SSIM provide quantitative assessments, they often fail to align with human perception, especially in scenarios involving complex textures or structural deformations. Consequently, there's a shift towards adopting perceptual quality metrics inspired by insights from cognitive psychology and visual neuroscience. Metrics like perceptual hashing, structural similarity with distortion perception (SSIM-DP), or learned perceptual image patch similarity (LPIPS) offer more nuanced evaluations, aligning better with human perceptual judgments and guiding the optimization of SR models towards producing visually pleasing results.

	In conclusion, while SR technologies, particularly those utilizing deep learning, have transformed the capabilities of image enhancement and upscaling, they continue to evolve in response to their inherent limitations and the growing demands of diverse applications. The future of SR lies in the advancement of models that are not only powerful and versatile but also efficient, adaptable, and capable of delivering high-quality results across a broad spectrum of real-world scenarios.

	\subsubsection{Algorithmic Pseudocode for Super-Resolution}
	The Super Resolution (SR) Algorithm is a sophisticated framework tailored for enhancing the resolution of images, particularly in scenarios where high-resolution information is desired from low-resolution inputs. It distinguishes itself by iteratively refining the mapping function between low and high-resolution images, incorporating pairs of input-output images to optimize model parameters. The training procedure for a Super-Resolution model begins with the initialization of the model's weights, followed by iterative learning over a specified number of epochs. During each epoch, the model processes pairs of low and high-resolution images, generating a super-resolved image from each low-resolution input. The difference between the generated high-resolution image and the true high-resolution image is quantified using a loss function, guiding the optimization of the model's weights to reduce this discrepancy. Upon completing the training process, the model is capable of enhancing the resolution of new low-resolution images, applying the learned mapping function to produce images with improved detail and clarity. The pseudocode available in \ref{fig:super-resolution-pseudocode} abstracts the complexity of deep learning algorithms into a high-level overview, emphasizing the structured approach to learning and applying Super-Resolution techniques.
	
	\begin{algorithm}
		\caption{Deep Learning-Based Super-Resolution Pseudocode}
		\begin{algorithmic}[1]
			\Procedure{TrainSRModel}{LR\_Images, HR\_Images, Model, Epochs, LossFunction}
			\State Initialize Model with random weights
			\For{each epoch in Epochs}
			\For{each (LR\_image, HR\_image) pair in (LR\_Images, HR\_Images)}
			\State Generate SR\_image from LR\_image using Model
			\State Calculate loss using LossFunction(HR\_image, SR\_image)
			\State Update Model weights to minimize loss
			\EndFor
			\EndFor
			\State \Return Trained Model
			\EndProcedure
			
			\Procedure{GenerateSRImage}{LR\_Image, TrainedModel}
			\State SR\_Image $\gets$ Apply TrainedModel to LR\_Image
			\State \Return SR\_Image
			\EndProcedure
		\end{algorithmic}\label{fig:super-resolution-pseudocode}
	\end{algorithm}
	
\subsection{Previous Work on ML and AI Interplay with Super Resolution Algorithms}

\paragraph{Review on Deep Learning for SISR}
A study providing an overview of the integration of single image super-resolution (SISR) with deep learning methodologies is presented in \cite{bashir2021comprehensive}. This review systematically outlines the progression of SISR techniques within the framework of deep learning. It traces the evolution from early convolutional neural networks (CNNs) to more advanced architectures like generative adversarial networks (GANs) and attention-based models. The review emphasizes the role of deep learning in advancing SISR, shedding light on the algorithmic foundations facilitating these advancements. It highlights the shift from traditional, model-based approaches to data-driven, learning-based strategies, addressing challenges such as scalability and diverse image content handling. The review also discusses performance metrics used for evaluating SISR methods, providing a critical assessment of how deep learning has refined super-resolution quality assessment. This comprehensive review contributes to understanding the impact of deep learning on SISR and provides insights for future research directions.

\paragraph{GANs for SISR}
An approach to single image super-resolution (SISR) utilizing generative adversarial networks (GANs) is introduced in \cite{ledig2017photo}. This work presents an alternative to conventional SISR techniques by leveraging GANs to generate photo-realistic high-resolution images from low-resolution inputs. The SRGAN architecture embodies a novel application of deep learning to enhance spatial resolution while preserving natural image characteristics. The paper details the design and implementation of SRGAN, focusing on its end-to-end mapping capability from low to high-resolution images. It discusses challenges in training GANs for SISR, particularly in balancing the generator and discriminator for optimal performance. The paper provides a thorough evaluation of SRGAN, demonstrating its effectiveness compared to existing SISR methods across different benchmarks. This work showcases the potential of GANs in enhancing image resolution and sets a new standard for SISR techniques, laying the groundwork for future research in adversarial learning for image restoration tasks.

\subsection{Algogenic Enhancements for Super-Resolution}
\subsubsection{Adaptive Learning from Diverse Data Sources}

\paragraph{Integrating Rich Datasets}
We suggest that focusing on adaptive learning from a wide array of data sources specifically enhances the Super-Resolution algorithm's robustness and versatility. By training on varied image types, including landscapes, urban scenes, and complex textures, the model can develop a more comprehensive feature set, improving its ability to upscale images across different contexts effectively. This strategy not only broadens the model's generalization capabilities but also mitigates domain-specific biases, enriching the super-resolved images with realistic details. Furthermore, leveraging insights from diverse fields like photography and medical imaging encourages interdisciplinary knowledge exchange, potentially leading to novel Super-Resolution methodologies.

\paragraph{Cross-Domain Adaptability}
We highlight the importance of cross-domain adaptability, utilizing techniques such as transfer learning and domain adaptation to enhance the Super-Resolution algorithm. This adaptability allows the model to fine-tune its parameters for specific input characteristics, improving performance across varied imaging conditions. For instance, models trained on high-resolution photographs can, through adaptation, enhance images under low-light conditions. This flexibility not only broadens the algorithm's applicability but also opens up new avenues for improving image resolution in diverse fields.

\paragraph{Algorithmic Integration for Enhanced Learning}
The integration of algorithms that dynamically adjust learning processes based on image content is crucial for Super-Resolution. By prioritizing learning on relevant features, such as texture or edge details, the algorithm can optimize its performance for each specific image type. This approach not only improves the clarity and detail of super-resolved images but also enhances computational efficiency, making it particularly relevant for real-time applications.

\subsubsection{Intelligent Upsampling Strategies}

\paragraph{Conceptual Foundation}
Intelligent upsampling, leveraging llms, aims to predict and reconstruct high-frequency details more accurately than traditional methods. This technique enhances image resolution while preserving original integrity, incorporating contextual understanding to achieve coherent and visually pleasing results. By iteratively refining the upsampling process, llms facilitate continuous improvement, offering a promising direction for Super-Resolution algorithms.

\paragraph{Leveraging Deep Learning for Detail Inference}
Deep learning frameworks, such as CNNs and GANs, play a pivotal role in intelligent upsampling by predicting missing details in low-resolution images. These models excel in capturing and replicating textures and patterns, ensuring the upscaled images maintain visual fidelity. The combination of CNNs and GANs enables the creation of super-resolved images with remarkable clarity and realism, pushing the boundaries of current Super-Resolution technologies.

\paragraph{Incorporating Contextual Awareness}
Incorporating contextual awareness allows Super-Resolution algorithms to adapt the upsampling process to the image's content, enhancing the preservation of important features. This strategy ensures that the upscaled images not only exhibit higher resolution but also retain essential visual details, significantly advancing the field of Super-Resolution.

\subsubsection{Automated Parameter Optimization}

\paragraph{Essence and Importance}
Automated parameter optimization is pivotal for refining Super-Resolution models, focusing on hyperparameter adjustments to enhance image quality. This approach streamlines the tuning process, improving efficiency and model adaptability across different datasets and applications. By employing techniques like Bayesian optimization, this strategy accelerates the development of high-quality Super-Resolution models, emphasizing the need for adaptability in dynamic imaging environments.

\paragraph{Utilizing Advanced Optimization Techniques}
The adoption of techniques such as Bayesian optimization and genetic algorithms facilitates efficient hyperparameter tuning, crucial for Super-Resolution models. These methods expedite the identification of optimal configurations, enhancing model performance and accelerating the development process. This streamlined approach to model refinement underscores the importance of advanced optimization techniques in achieving superior Super-Resolution results.

\paragraph{Impact on Model Performance and Generalization}
Automated optimization significantly improves model generalization, enabling Super-Resolution algorithms to perform consistently across varied image types and conditions. This adaptability is essential for applications requiring robust performance, highlighting the transformative potential of automated parameter optimization in the Super-Resolution domain.

\subsubsection{Content-Aware Fill and Detail Synthesis}

\paragraph{Innovative Approach to Super-Resolution}
Content-aware fill and detail synthesis offer a novel approach to Super-Resolution, focusing on generating missing details in a contextually consistent manner. This method enhances image realism and integrity, ensuring seamless integration of enhanced details. The adaptability of content-aware techniques, coupled with their efficiency in resource allocation, marks a significant advancement in Super-Resolution technologies, expanding their applicability beyond traditional tasks.

\paragraph{Leveraging Deep Learning for Contextual Understanding}
Deep learning models, particularly CNNs and GANs, are instrumental in content-aware fill and detail synthesis, enabling the Super-Resolution algorithm to capture and replicate complex image details accurately. This integration facilitates the production of super-resolved images that maintain high fidelity to the original content, enhancing the perceptual quality of upscaled images.

\paragraph{Enhancing Image Realism and Quality}
Content-aware fill and detail synthesis significantly improve the realism and quality of super-resolved images, addressing the limitations of traditional upscaling methods. By intelligently filling in details based on the image's context, this approach produces sharper, more detailed images, advancing the field of Super-Resolution.

\subsubsection{Multi-modal Data Integration}

\paragraph{Expanding the Data Horizon}
Integrating diverse data types, including textual descriptions and sensor data, enhances Super-Resolution models by providing additional context and information. This multi-modal approach improves image reconstruction accuracy and detail, pushing the boundaries of traditional Super-Resolution techniques and enabling more robust algorithms capable of addressing complex real-world scenarios.

\paragraph{Synergistic Use of Varied Data Types}
The synergistic integration of varied data types, such as depth information and textual data, enriches the Super-Resolution model's input, enhancing its precision and adaptability. This approach allows for the generation of high-quality upscaled images that maintain visual and semantic coherence, highlighting the benefits of multi-modal data integration in Super-Resolution algorithms.

\paragraph{Enhancing Perceptual Quality and Detail Accuracy}
Multi-modal data integration significantly improves the perceptual quality and detail accuracy of super-resolved images, enabling SR models to produce results that are both higher in resolution and more contextually accurate. This advancement underscores the importance of leveraging diverse data sources in enhancing the capabilities of Super-Resolution technologies.

\subsubsection{Real-time Adaptation and Learning}

\paragraph{Adapting to Evolving Conditions}
Real-time adaptation and learning enable Super-Resolution models to continually adjust and improve in response to new images and changing conditions. This dynamic enhancement ensures that SR algorithms remain effective across diverse environments, highlighting the importance of adaptability and continuous learning in maintaining high-quality super-resolution.

\paragraph{Leveraging Incremental Learning Techniques}
Incorporating incremental learning techniques allows Super-Resolution models to update their knowledge base efficiently, adapting to new data without extensive retraining. This adaptability is crucial for applications requiring immediate and accurate image enhancement, emphasizing the role of incremental learning in enhancing the robustness and responsiveness of SR algorithms.

\paragraph{Balancing Stability and Plasticity}
Achieving a balance between stability and plasticity is essential for real-time adaptation in Super-Resolution models. Techniques like elastic weight consolidation and experience replay help maintain this balance, ensuring that models can learn from new data without forgetting previous knowledge. This balance is crucial for developing versatile and reliable Super-Resolution algorithms capable of adapting to a wide range of scenarios.

\subsubsection{Ethical and Bias Mitigation Mechanisms}

\paragraph{Addressing Ethical Considerations}
Incorporating ethical and bias mitigation mechanisms is essential for developing fair and unbiased Super-Resolution models. These strategies ensure that SR technologies respect privacy, representation, and equity, addressing potential ethical concerns and biases. By prioritizing ethical considerations and incorporating diverse perspectives, developers can create more responsible and trustworthy Super-Resolution technologies.

\paragraph{Mitigating Bias in Super-Resolution Models}
Addressing bias in Super-Resolution models involves diversifying training datasets and implementing algorithmic interventions. These measures help reduce the risk of biased predictions, ensuring that SR algorithms perform equitably across different scenarios. Ongoing efforts to mitigate bias are crucial for developing fair and accurate Super-Resolution models.

\paragraph{Implementing Ethical Frameworks}
The implementation of ethical frameworks guides the development and deployment of Super-Resolution technologies, emphasizing transparency, accountability, and inclusivity. Regular ethical audits and impact assessments help identify and address potential issues, fostering trust and confidence in SR technologies. Ethical frameworks play a vital role in promoting responsible and sustainable Super-Resolution practices.

\subsubsection{Cross-Domain Knowledge Transfer}

\paragraph{Enhancing Versatility Through Knowledge Sharing}
Cross-domain knowledge transfer enhances the versatility and efficiency of Super-Resolution models by leveraging insights from diverse fields. This approach fosters interdisciplinary collaboration and innovation, broadening the applicability of SR technologies and encouraging the development of novel image enhancement methodologies.

\paragraph{Utilizing Transfer Learning Techniques}
Transfer learning techniques are central to cross-domain knowledge transfer, enabling Super-Resolution models to apply knowledge from one domain to improve performance in another. This strategy accelerates model development, enhances performance, and expands the applicability of SR algorithms, demonstrating the transformative potential of transfer learning in Super-Resolution.

\paragraph{Overcoming Domain-Specific Challenges}
Cross-domain knowledge transfer helps overcome domain-specific challenges in Super-Resolution, enabling models to learn unique features from specialized datasets. This adaptability enhances the algorithm's ability to reconstruct high-resolution images across diverse domains, underscoring the importance of knowledge transfer in advancing Super-Resolution technologies.

	\subsubsection{Pseudocode for Algogenic Super-Resolution}
	The Algogenic super-resolution approach harnesses AI to enhance traditional super-resolution methods by dynamically adjusting algorithmic parameters and strategies based on observed system behavior and real-time error estimates. This pseudocode, available in \ref{fig:super-resolution-Algogen-pseudocode}, delineates an advanced framework integrating AI-driven enhancements for adaptive resolution scaling, image patch selection, quality assessment, and real-time parameter optimization.
	
	\begin{algorithm}
		\caption{Algogenic Super-Resolution Pseudocode}
		\begin{algorithmic}[1]
			\Procedure{AlgogenicSuperResolution}{LowResImage}
			
			\Comment{Preprocessing Phase}
			\State IntegrateRichDatasets()
			\State PerformCrossDomainAdaptability()
			
			\Comment{Core Denoising Phase}
			\State enhancedLearning $\gets$ AlgorithmicIntegrationForEnhancedLearning()
			\State upsamplingStrategy $\gets$ SelectIntelligentUpsamplingStrategy(enhancedLearning)
			\If{AdaptiveParameterOptimizationNeeded()}
			\State PerformAutomatedParameterOptimization(upsamplingStrategy)
			\EndIf
			\State UpsampledImage $\gets$ ApplyUpsampling(LowResImage, upsamplingStrategy)
			\State UpsampledImage $\gets$ ContentAwareFillAndDetailSynthesis(UpsampledImage)
			\State UpsampledImage $\gets$ MultiModalDataIntegration(UpsampledImage)
			\While{Not SatisfactoryQuality(UpsampledImage)}
			\State UpsampledImage $\gets$ RealTimeAdaptationAndLearning(UpsampledImage)
			\EndWhile
			\State UpsampledImage $\gets$ CrossDomainKnowledgeTransfer(UpsampledImage)
			
			\Comment{Postprocessing Phase}
			\State ApplyEthicalAndBiasMitigationMechanisms(UpsampledImage)
			
			\EndProcedure
		\end{algorithmic}\label{fig:super-resolution-Algogen-pseudocode}
	\end{algorithm}

	\begin{figure}
		\centering
		\includegraphics[width=0.4\textwidth]{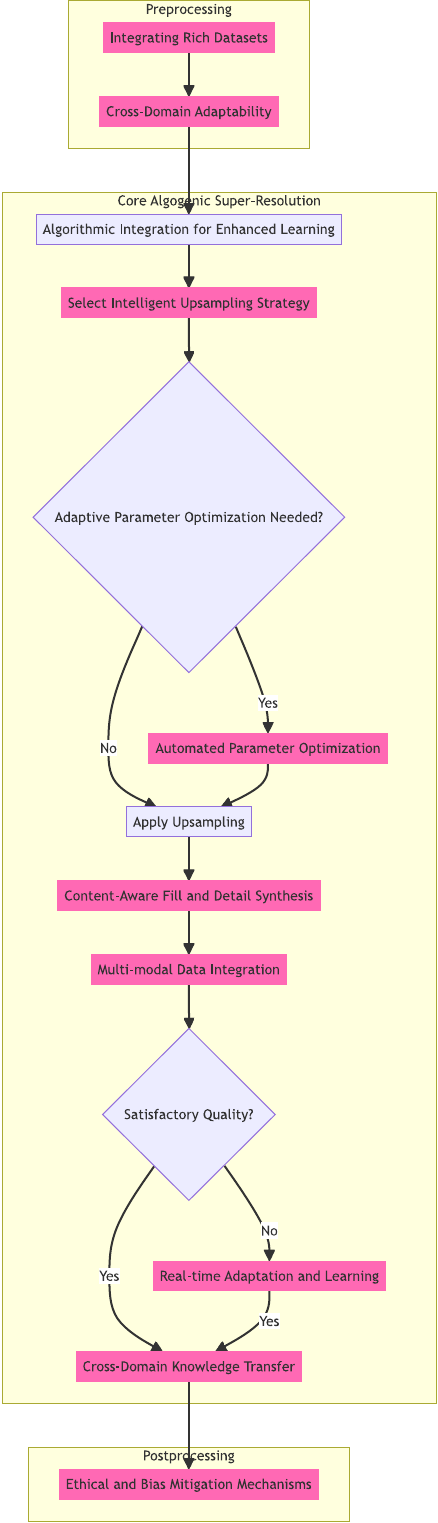}
		\caption{Enhancing Super-Resolution through Algogenic Frameworks: This figure conceptualizes the integration of Algogenic enhancements with Super-Resolution techniques, highlighting the process from preprocessing with diverse data sources to intelligent, AI-driven upsampling strategies in the core phase, and concluding with postprocessing that ensures ethical standards. It illustrates the dynamic selection and application of upsampling strategies based on algorithmic integration and enhanced learning, enriched by cross-domain adaptability and real-time learning adjustments. This comprehensive approach signifies a leap in image enhancement capabilities, showcasing how generative AI not only optimizes the technical process of Super-Resolution but also ensures adaptability, quality, and ethical integrity throughout.}
		\label{fig:super_resolution}
	\end{figure}

	\section{Image Inpainting}\index{Image Inpainting}
	\subsection{Introduction to Image Inpainting}
	\subsubsection{The Concept of Image Inpainting}
	
	\paragraph{Definition and Purpose}
	Image inpainting, a technique deeply rooted in the domain of digital image processing, refers to the process of reconstructing missing or damaged parts of images in a manner that is seamless and undetectable to the observer. The primary purpose of image inpainting is to restore images to their original state or to fill in the gaps in a way that maintains the continuity of visual information. This process is not merely about pixel interpolation; it is an art of guessing the missing details based on the available contextual and structural information in the image, making the restored parts blend naturally with the intact regions.
	
	Furthermore, image inpainting plays a crucial role in various applications such as image restoration, object removal, and image editing. In the realm of restoration, it aids in the preservation of historical artifacts and documents by repairing damaged areas without altering the original appearance significantly. Moreover, in the context of object removal, image inpainting enables the elimination of unwanted elements or distractions from photographs, enhancing their aesthetic appeal and visual coherence. Additionally, in image editing tasks, inpainting facilitates the creation of seamless composites by filling in regions where objects have been removed or rearranged, ensuring a smooth transition between different elements of the composition.
	
	Furthermore, the advancement of image inpainting techniques has been driven by the increasing demand for realistic and high-quality image manipulation tools in various fields such as entertainment, forensics, and medical imaging. Moreover, with the proliferation of digital content creation platforms and social media, there is a growing need for efficient and effective inpainting algorithms capable of producing visually appealing results in real-time scenarios. Thus, image inpainting continues to be a vibrant area of research and development, with ongoing efforts aimed at pushing the boundaries of inpainting capabilities and applications.

	\paragraph{Historical Context and Evolution}
	Historically, the concept of inpainting has its origins in the restoration of artworks, where physical damages to paintings and frescoes were meticulously repaired by skilled conservators. \textbf{Moreover}, in the digital realm, this concept has evolved to encompass a wide array of algorithms and techniques aimed at automating the restoration of damaged photographs, \textbf{removing unwanted objects, and filling in missing areas in images} captured by cameras or other imaging devices. The evolution of image inpainting from manual art restoration to automated digital processes highlights the advancements in computational algorithms and the growing understanding of how to replicate human perception of continuity and completeness in visual content.

	\paragraph{Technological Foundations and Approaches}
	The field of image inpainting rests upon a robust amalgamation of diverse technological foundations and approaches, each contributing uniquely to its advancement. Initially rooted in signal processing theories, image inpainting sought to reconstruct missing or damaged regions within images by employing techniques akin to interpolation, extrapolation, and diffusion. These methodologies, while effective to some extent, often struggled to produce convincing results, particularly in scenarios involving intricate textures or complex structures.
	
	Subsequently, the integration of computer vision techniques marked a significant stride forward in the evolution of image inpainting. Patch-based methods emerged as a prominent strategy, wherein the missing portions of an image were inferred by analyzing neighboring patches and synthesizing plausible content based on similarity metrics. This approach fostered a semblance of coherence and continuity in the inpainted regions, striving to preserve the intrinsic characteristics of the image.
	
	However, the landscape of image inpainting underwent a paradigm shift with the advent of deep learning algorithms, notably Convolutional Neural Networks (CNNs) and Generative Adversarial Networks (GANs). By harnessing the power of large-scale datasets, these neural architectures revolutionized the field by endowing algorithms with the capacity to discern intricate patterns and semantic features inherent in images. Leveraging the hierarchical representations learned through convolutional layers, CNNs excel in capturing local spatial dependencies and contextual information, thus facilitating more nuanced inpainting results.
	
	Moreover, the introduction of GANs ushered in a new era of realism and fidelity in image inpainting. By pitting a generator against a discriminator in a competitive training framework, GANs engendered a process akin to creative collaboration, wherein the generator endeavors to produce increasingly convincing forgeries while the discriminator refines its discernment abilities. This adversarial interplay culminates in the generation of inpainted images that exhibit remarkable visual fidelity, indistinguishable from authentic imagery to the human eye.
	
	In conclusion, the convergence of signal processing theories, computer vision techniques, and deep learning algorithms has propelled image inpainting to unprecedented heights of efficacy and realism. The symbiotic synergy between these technological foundations continues to underpin advancements in the field, promising ever more sophisticated solutions to the challenge of seamlessly restoring missing visual content.

	\paragraph{Mathematical Modeling}
	Mathematically, image inpainting can be formulated as an optimization problem where the goal is to find the best approximation of the missing or damaged part of an image. This involves minimizing a loss function that measures the difference between the inpainted image and the original image, given the known pixels. The problem can be expressed as
	\[
	\min_{I} \; L\left( I, I_{orig} \right) \; \text{subject to} \; I_{known},
	\]
	where \(I\) represents the inpainted image, \(I_{orig}\) is the original image, \(L\) is the loss function, typically involving terms for data fidelity and regularization to enforce smoothness or texture continuity, and \(I_{known}\) denotes the portion of the image that is known or unaffected by damage. Advanced inpainting techniques incorporate additional constraints or terms in the loss function to ensure that the inpainted area is consistent not only in appearance but also in semantic content with the rest of the image.
	
	The concept of image inpainting encapsulates a blend of artistic intuition and algorithmic precision, aiming to restore or reconstruct images in a way that is both visually pleasing and faithful to the original content. As technology advances, the methods and approaches to inpainting continue to evolve, offering ever more sophisticated tools for image restoration and manipulation.

	\subsubsection{Key Principles and Mechanisms}
	
	\paragraph{Underlying Principles of Image Inpainting}
	The process of image inpainting rests upon fundamental principles that ensure the coherence and fidelity of the restored image. Continuity, as a foundational concept, dictates that the replenished regions exhibit a seamless integration with the surrounding environment. This entails meticulous attention to details such as texture, color, and pattern, ensuring that the transition between the original content and the inpainted area remains imperceptible. By adhering to the principle of continuity, inpainting algorithms can effectively reconstruct missing portions without introducing artifacts that could disrupt the visual harmony of the image.
	
	Consistency, another pivotal aspect, governs the preservation of structural and semantic coherence throughout the inpainting procedure. It necessitates that the generated content not only matches the appearance of the surrounding elements but also aligns with the underlying context of the image. This entails maintaining the spatial relationships and semantic meaning inherent in the original scene, thereby upholding the perceptual integrity of the inpainted regions. By upholding consistency, inpainting techniques can produce results that not only appear visually plausible but also uphold the narrative and compositional coherence of the image.
	
	Contextuality serves as a guiding principle that empowers inpainting algorithms to discern the broader context of the image and make informed decisions regarding the reconstruction of missing content. By analyzing the spatial and semantic cues present in the surrounding regions, the algorithm can infer the likely appearance of the obscured content, thus enhancing the realism and appropriateness of the inpainted result. Contextuality enables the algorithm to leverage contextual information such as object boundaries, texture gradients, and scene semantics to guide the inpainting process, resulting in coherent and contextually appropriate restorations.
	
	In summary, the efficacy of image inpainting techniques hinges upon their adherence to the principles of continuity, consistency, and contextuality. By seamlessly blending inpainted regions with their surroundings, preserving structural and semantic coherence, and leveraging contextual cues, these principles underpin the generation of high-quality inpainted images that seamlessly integrate with the original content.

	\paragraph{Patch-Based Inpainting Mechanisms}
	Patch-based inpainting mechanisms, while rudimentary in nature, represent a foundational approach in image restoration. Through the utilization of patches extracted from intact regions of the image, these mechanisms endeavor to seamlessly fill the lacunae, thereby restoring visual continuity. The underlying premise hinges upon the intrinsic redundancy prevalent in natural images, where comparable textures and structures frequently recur. Consequently, the algorithm's efficacy pivots on its adeptness in discerning and selecting patches that closely resemble the missing content, a process mediated by a predefined similarity metric. This metric serves as the yardstick for evaluating patch congruity, ensuring that the selected patches seamlessly integrate with the surrounding context. Moreover, the blending operation executed thereafter is pivotal in harmonizing the amalgamated patches with the surrounding environment, obviating conspicuous demarcations and fostering a perceptually cohesive outcome. Nevertheless, the efficacy of patch-based inpainting is contingent upon the availability of suitable matches within the image domain. In scenarios where distinctive or irregular features dominate, the algorithm's performance may falter, necessitating alternative strategies for artifact concealment. Despite its elementary nature, patch-based inpainting remains a cornerstone in the pantheon of image restoration techniques, often serving as the foundation upon which more sophisticated methodologies are built.

	\paragraph{Structural and Texture Synthesis}
	Advancements in inpainting have led to the development of techniques that explicitly separate the reconstruction process into structural synthesis and texture synthesis. \textbf{Furthermore}, this separation enhances the overall quality and realism of the reconstructed images. 
	
	\textbf{Structural synthesis} can be likened to the skeletal framework of an image, focusing on reinstating the fundamental elements such as edges, contours, and major shapes. This stage is crucial as it establishes the underlying structure upon which the finer details will be added. \textbf{Consequently}, the reconstructed region maintains the coherence and continuity of the overall scene, even in the absence of original information. 
	
	On the other hand, \textbf{texture synthesis} addresses the intricate patterns and details that contribute to the visual richness of an image. \textbf{Moreover}, this aspect ensures that the filled regions seamlessly blend with the surrounding areas, preserving the visual harmony and avoiding perceptible discrepancies. Techniques such as patch-based synthesis or deep learning-based approaches are commonly employed in this stage to capture the complex texture variations present in natural scenes.
	
	This distinction between structural and texture synthesis allows for a more nuanced approach to image inpainting. \textbf{In addition}, it enables the preservation of both global structure and local texture characteristics, leading to more visually pleasing results. This approach is particularly beneficial in scenarios where images contain diverse textures or intricate details, such as landscapes or portraits. \textbf{Furthermore}, it facilitates better handling of challenging inpainting tasks, where maintaining consistency and realism are paramount.

	\paragraph{Deep Learning-Based Approaches}
	Deep learning-based approaches have revolutionized image inpainting by leveraging neural networks to learn complex representations of images. These methods use large datasets to train models that can predict the content of missing regions with a high level of detail and accuracy. Generative models, such as Generative Adversarial Networks (GANs), have been particularly effective, as they can generate new image content that is both diverse and realistic. The training process involves a discriminator network that challenges the generator by distinguishing between real and inpainted images, pushing the generator to produce increasingly convincing results.
	
	Furthermore, these deep learning techniques have opened up avenues for inpainting in various domains beyond just images. Moreover, the utilization of large datasets not only enhances the robustness of the models but also enables them to capture intricate features within the images, leading to more accurate predictions. Additionally, the iterative nature of the training process, with the discriminator providing feedback to the generator, facilitates the refinement of inpainted results over time. On the other hand, while GANs have shown remarkable success, they also pose challenges such as mode collapse and training instability, which researchers continue to address through novel architectures and optimization strategies. Nonetheless, the advancements in deep learning-based inpainting have significantly improved the quality and efficiency of content completion in images, promising further developments in this field.

	\paragraph{Mathematical Formulations and Loss Functions}
	In deep learning-based inpainting, the mathematical formulation often involves optimizing a composite loss function that includes terms for content fidelity, perceptual quality, and adversarial loss. The content fidelity term ensures that the inpainted image closely matches the known parts of the original image, the perceptual quality term encourages the inpainted image to have similar feature representations as real images, and the adversarial loss, derived from the discriminator network, promotes realism in the generated content. This can be represented as
	\[
	\min_{G} \max_{D} \; L_{content}(G) + \lambda_{perceptual} L_{perceptual}(G) + \lambda_{adv} L_{adv}(G, D),
	\]
	where \(G\) is the generator, \(D\) is the discriminator, \(L_{content}\), \(L_{perceptual}\), and \(L_{adv}\) are the respective loss terms, and \(\lambda_{perceptual}\) and \(\lambda_{adv}\) are weights that balance the contributions of each term.
	
	The principles and mechanisms of image inpainting cover a wide range of techniques, from simple patch-based methods to complex deep learning algorithms, all aimed at restoring missing or damaged areas of images in a visually plausible and contextually coherent manner.

	\subsubsection{The Role of Contextual Information}
	
	\paragraph{Foundation of Contextual Relevance}
	The role of contextual information in image inpainting is pivotal, serving as the foundation for making informed decisions about how to accurately fill in missing or damaged areas of an image. Contextual information encompasses a broad range of data, including but not limited to, the spatial relationships within the image, the semantic content (what the image depicts), and even external data related to the image. 
	
	Utilizing this contextual information enables inpainting algorithms to reconstruct images in a manner that seamlessly integrates the inpainted regions with the surrounding content. \textbf{Moreover}, by considering the spatial relationships within the image, such as edge continuity and texture coherence, inpainting algorithms can generate visually plausible results that maintain the structural integrity of the scene. 
	
	\textbf{Furthermore}, incorporating semantic understanding allows the algorithm to prioritize important image elements during inpainting, ensuring that crucial objects or features are accurately restored. This ensures that the inpainted regions not only blend seamlessly into the image but also contribute meaningfully to the overall visual narrative. 
	
	\textbf{Additionally}, leveraging external data sources, such as context from related images or domain-specific knowledge, can enhance the inpainting process by providing additional cues for accurate reconstruction. By considering a diverse range of contextual cues, inpainting algorithms can produce results that are not only visually appealing but also contextually relevant, enhancing the overall quality of the inpainted images. 
	
	In summary, the effective utilization of contextual information is crucial for image inpainting algorithms to generate realistic and contextually appropriate results, ensuring that the inpainted regions align harmoniously with the rest of the image.

	\paragraph{Spatial and Semantic Context}
	Spatial context plays a fundamental role in the inpainting process by providing guidance on how objects and features are spatially arranged within the image space. This spatial understanding enables the inpainting algorithm to maintain geometric consistency and alignment when filling in missing or damaged regions of the image. For instance, when inpainting a portion of a road in an urban scene, the algorithm needs to consider the surrounding context, such as the presence of buildings, sidewalks, and other road features, to accurately reconstruct the missing segment while preserving the road's width and trajectory.
	
	On the other hand, semantic context involves a deeper understanding of the content and meaning conveyed by the image. It encompasses the recognition of objects, scenes, and their semantic interrelationships. This semantic understanding is essential for inpainting algorithms to generate coherent and contextually appropriate content. For example, when inpainting a damaged section of a natural landscape, the algorithm must recognize the presence of trees, mountains, and sky and ensure that the inpainted region seamlessly integrates with the surrounding environment in terms of color, texture, and spatial configuration.
	
	Furthermore, semantic context enables inpainting algorithms to adhere to domain-specific constraints and conventions. For instance, when inpainting architectural elements like buildings, the algorithm needs to conform to architectural norms regarding building design, structure, and symmetry. By incorporating semantic context, the inpainting process can produce visually plausible results that align with human expectations and perceptual understanding of the scene.

	\paragraph{Incorporating External Contextual Data}
	In the realm of image inpainting, the utilization of external contextual data marks a significant advancement, transcending the traditional reliance solely on the visual content of the image. By tapping into textual descriptions associated with the image, metadata detailing its origin or content, or even a repository of analogous images from comparable settings, inpainting algorithms gain a broader understanding of the scene at hand. This holistic approach enhances the accuracy and relevance of the inpainted results by supplementing the inherent visual context with additional semantic information.
	
	Textual descriptions serve as rich sources of context, offering insights into the scene's composition, its historical significance, or the intended narrative. Metadata, including geographical coordinates, timestamps, or user-generated tags, provide further context, aiding in the interpretation of the image's content and spatial-temporal context. Moreover, leveraging related images from similar contexts furnishes the algorithm with a diverse array of reference points, enabling it to extrapolate missing information more intelligently.
	
	Consider, for instance, an inpainting scenario involving a damaged image depicting a renowned landmark. While the internal visual cues within the image may be inadequate for accurate reconstruction, incorporating external data specifying the landmark's identity, architectural style, or historical period can substantially refine the inpainting process. This integration empowers the algorithm to infer missing details with greater precision, ensuring that the restored image aligns faithfully with the expected characteristics of the depicted landmark.
	
	Ultimately, the amalgamation of external contextual data with intrinsic visual information engenders a more comprehensive inpainting framework, capable of discerning and reproducing nuanced details that transcend the confines of the image itself. By embracing a multifaceted approach that assimilates textual, metadata, and analogous image data, inpainting algorithms transcend mere pixel-level reconstruction, striving towards a deeper understanding and faithful restoration of the depicted scene.

	\paragraph{Deep Learning and Contextual Modeling}
	Deep learning approaches, such as Convolutional Neural Networks (CNNs) and Generative Adversarial Networks (GANs), have revolutionized the field of image inpainting by profoundly enhancing the capacity to model contextual information. CNNs and GANs, owing to their hierarchical architecture, exhibit exceptional prowess in not only discerning spatial intricacies but also grasping semantic nuances within images. Through multiple layers of abstraction, CNNs extract increasingly abstract features, thereby encapsulating both local and global contextual cues vital for accurate inpainting. Conversely, GANs leverage adversarial training to generate highly realistic content by discerning the intricate interplay of context and content. This adversarial framework fosters the creation of inpainted regions that seamlessly blend with the surrounding context, ensuring visually coherent results. Moreover, the dynamic nature of GANs enables them to adaptively learn from the contextual cues present in the input image, refining the inpainting process iteratively. Despite their computational intensity, CNNs and GANs synergistically exploit the rich contextual information inherent in images to infer missing content with remarkable fidelity. Furthermore, the fusion of these deep learning paradigms offers a holistic approach to contextual modeling, where CNNs decode contextual semantics while GANs synthesize content-aware completions. This amalgamation empowers inpainting systems to effectively navigate diverse image contexts, yielding compelling results across a spectrum of inpainting tasks. Thus, the symbiotic integration of CNNs and GANs underscores the transformative potential of deep learning in contextual modeling, propelling the boundaries of image inpainting towards unprecedented realism and accuracy.

	\paragraph{Mathematical Consideration of Context}
	The mathematical consideration of context in image inpainting involves designing loss functions and network architectures that can effectively encode and leverage contextual information. This often includes terms in the loss function that specifically reward spatial coherence and semantic accuracy, such as
	\[
	L = \alpha L_{pixel}(I, \hat{I}) + \beta L_{context}(I, \hat{I}) + \gamma L_{semantic}(I, \hat{I}),
	\]
	where \(I\) is the original image, \(\hat{I}\) is the inpainted image, \(L_{pixel}\) measures pixel-level accuracy, \(L_{context}\) evaluates the alignment with spatial context, \(L_{semantic}\) assesses semantic coherence, and \(\alpha\), \(\beta\), and \(\gamma\) are weights that balance the importance of each component.
	
	Contextual information is thus an essential element in the image inpainting process, enabling algorithms to produce reconstructions that are not just visually plausible but fully integrated into the visual and semantic narrative of the original image. This comprehensive approach ensures that inpainted images are both aesthetically pleasing and contextually coherent.

	\subsubsection{Applications and Limitations}
	
	\paragraph{Diverse Applications of Image Inpainting}
	Image inpainting, a technique that fills in missing or damaged areas of an image, has gained widespread utility across diverse fields owing to its adaptability and effectiveness in image restoration and enhancement tasks. Its application spans various domains, each benefiting from its unique capabilities.
	
	In the realm of art restoration, digital inpainting emerges as a pivotal tool, facilitating the meticulous reconstruction of damaged or deteriorated artworks. Conservators leverage these techniques to visualize potential restorations before undertaking physical interventions, ensuring a balanced approach between preservation and enhancement of cultural heritage.
	
	Moreover, within the sphere of photography and film, image inpainting serves as a formidable ally in the pursuit of flawless visual storytelling. By seamlessly removing unwanted elements or imperfections from images and videos, it aids in refining the narrative, enhancing visual appeal, and preserving the authenticity of historical or sentimental visual content.
	
	In the context of surveillance and law enforcement, image inpainting assumes a crucial role in forensic analysis. It enables investigators to reconstruct crucial details from partially obscured or damaged footage, potentially unveiling vital clues and aiding in criminal investigations.
	
	Furthermore, in the consumer electronics landscape, the integration of inpainting capabilities empowers users with powerful photo editing tools. By seamlessly removing unwanted objects or blemishes and filling in gaps intelligently, these features elevate the quality and aesthetics of personal photographs, empowering users to create visually captivating memories.
	
	The versatility of image inpainting transcends disciplinary boundaries, offering innovative solutions to diverse challenges across art, media, security, and personal expression.

	\paragraph{Expanding to Medical Imaging and Research}
	Beyond these applications, image inpainting is increasingly important in medical imaging, where it assists in reconstructing missing or corrupted parts of medical scans, potentially aiding in diagnosis and research. This application is particularly significant, as it can lead to better patient outcomes by providing more complete and accurate visual data for analysis. 
	
	Moreover, in medical imaging, the ability to accurately reconstruct missing information can be crucial for identifying abnormalities or lesions that might otherwise be overlooked. For instance, in MRI scans, where small lesions or artifacts can occur due to motion artifacts or limitations in imaging techniques, inpainting algorithms can help enhance the clarity of images, allowing radiologists to make more confident diagnoses. Furthermore, in fields such as pathology and histology, where detailed examination of tissue samples is essential for diagnosing diseases, inpainting techniques can aid in reconstructing missing sections of samples, improving the accuracy of diagnoses and facilitating research into various medical conditions.
	
	Additionally, in scientific research, especially in fields like astronomy or environmental science, inpainting plays a vital role in data analysis and interpretation. By filling gaps in observational data or satellite imagery, researchers can obtain a more comprehensive understanding of complex phenomena such as atmospheric dynamics or celestial events. This not only enhances our knowledge of the universe and the environment but also contributes to the development of predictive models and strategies for addressing environmental challenges.
	
	Overall, the integration of image inpainting techniques into medical imaging and scientific research holds immense promise for advancing diagnostic capabilities, deepening our understanding of natural processes, and ultimately improving human health and well-being.

	\paragraph{Limitations and Challenges}
	Despite its vast potential, image inpainting is not without limitations and challenges. One major limitation is the dependence on the context and quality of the available information within the image. In cases where significant portions of an image are missing or the available data does not provide enough context, inpainting algorithms might struggle to generate accurate and plausible fill-ins. Moreover, the risk of semantic incoherence increases with the complexity of the image and the size of the area to be inpainted, potentially leading to reconstructions that are visually plausible but semantically incorrect. While inpainting techniques have advanced considerably, they often face difficulties when dealing with large missing regions or when the surrounding information is ambiguous. Consequently, despite continuous improvements, achieving robust performance in inpainting remains a challenging task. Furthermore, as images become more intricate and diverse in content, it becomes increasingly challenging for inpainting algorithms to accurately infer missing information. Additionally, the presence of noise or artifacts in the image can further exacerbate the inpainting process, resulting in distorted or unrealistic fill-ins. Nevertheless, ongoing research in this field aims to address these limitations and develop more effective inpainting methods capable of handling diverse scenarios and producing high-quality results.

	\paragraph{Ethical Considerations and Misuse}
	Another critical challenge is the ethical consideration regarding the misuse of inpainting technology. While this innovation offers tremendous potential for enhancing and restoring damaged images, its misuse can lead to significant ethical dilemmas. In particular, the ease with which inpainting can be used to alter images raises concerns about authenticity and trustworthiness. This concern is especially pertinent in sensitive contexts like journalism, where the integrity of visual evidence is paramount for informing the public accurately. In legal proceedings, the manipulation of images through inpainting could potentially undermine the fairness of trials by presenting fabricated or misleading evidence. Similarly, in historical documentation, the intentional alteration of images may distort our understanding of the past, eroding the integrity of historical records. Moreover, the ethical implications extend to broader societal issues, such as the perpetuation of false narratives or the manipulation of public opinion through deceptive imagery. Therefore, while inpainting technology holds promise for various applications, its potential for misuse underscores the importance of robust ethical frameworks and regulatory measures to safeguard against deceptive practices. Furthermore, fostering awareness among users about the ethical responsibilities associated with this technology is crucial for promoting its responsible and beneficial use.

	\paragraph{Technical Limitations and Future Directions}
	From a technical perspective, the computational cost of advanced inpainting algorithms, particularly those based on deep learning, can be prohibitive for real-time applications or devices with limited processing power. These algorithms often require significant computational resources due to the complexity of neural network architectures and the large amount of data involved in training. Consequently, real-time inpainting on resource-constrained devices remains a challenge, necessitating the exploration of lightweight models or optimization techniques to reduce computational overhead.
	
	Moreover, the quality of inpainting can be highly variable depending on the algorithm, the model's training data, and the specific characteristics of the image being processed. While deep learning-based methods have shown remarkable performance in certain scenarios, they may struggle with complex scenes, irregular object shapes, or rare data distributions. This variability underscores the importance of robustness testing across diverse datasets and the development of adaptive algorithms capable of handling a wide range of inpainting scenarios.
	
	Looking forward, addressing these limitations and challenges will require ongoing research and development, focusing on improving the robustness, efficiency, and ethical governance of inpainting technologies. Efforts to enhance robustness may involve exploring multi-modal approaches that combine deep learning with traditional image processing techniques to leverage their respective strengths in different inpainting scenarios. Additionally, optimizing algorithms for parallel processing or hardware acceleration could help mitigate computational bottlenecks and enable real-time or near-real-time performance on a broader range of devices.
	
	The future of image inpainting lies in the balance between advancing technological capabilities and navigating the ethical implications of altering visual content. As inpainting tools become more sophisticated and accessible, there is a growing need for transparent and accountable frameworks to govern their use. This includes considerations of privacy, consent, and potential misuse, highlighting the importance of interdisciplinary collaboration between technologists, ethicists, policymakers, and other stakeholders to ensure responsible innovation in this field.

	\subsubsection{Algorithmic Pseudocode for Image Inpainting}
	The Image Inpainting Algorithm is a sophisticated framework tailored for filling missing or damaged regions within an image, identified by a Mask. This pseudocode outlines a basic procedure for image inpainting that focuses on filling missing or damaged regions identified by a Mask. The process begins by creating an initial copy of the original image, which will be incrementally updated. The algorithm iteratively identifies the boundary pixels that lie at the edge of the missing regions and searches for the best matching patches from the known regions of the image to copy or blend into the missing areas. This step is repeated until no missing regions remain, at which point a post-processing step is applied to ensure that the inpainted areas blend seamlessly with the rest of the image. The pseudocode in \ref{fig:image-inpainting-pseudocode} captures the operational mechanics of a patch-based image inpainting approach, emphasizing the systematic procedure of filling in missing areas based on contextual information from surrounding pixels.
	
	\begin{algorithm}
		\caption{Algorithmic Pseudocode for Image Inpainting}
		\begin{algorithmic}[1]
			\Procedure{ImageInpainting}{Image, Mask}
			\State Identify damaged or missing regions using Mask
			\State Initialize inpainted image as a copy of the original Image
			\While{there are still missing regions}
			\State Find boundary pixels between known and unknown regions
			\For{each boundary pixel}
			\State Identify best matching patch from known region
			\State Copy or blend best matching patch into the missing region
			\EndFor
			\State Update Mask to reflect filled regions
			\EndWhile
			\State Apply post-processing for seamless blending
			\State \Return Inpainted image
			\EndProcedure
		\end{algorithmic}\label{fig:image-inpainting-pseudocode}
	\end{algorithm}

\subsection{Previous Work on ML and AI Interplay with Image Inpainting Algorithms}

\paragraph{Theoretical Foundation for Image Inpainting with DDPMs}
In 2023, a study provided a theoretical justification for utilizing Denoising Diffusion Probabilistic Models (DDPMs) in image inpainting. This research elucidated the mathematical principles and effectiveness of DDPMs for generating inpainted images of high quality. By integrating DDPMs, the study demonstrated an enhancement in the quality of inpainted areas and contributed to understanding the behavior of probabilistic models in image restoration tasks. The findings highlighted the potential of DDPMs to address complex inpainting scenarios where conventional methods may struggle due to the absence of structural or textural information. This work underscored the importance of a solid theoretical foundation in advancing AI-driven image inpainting techniques, setting a new benchmark for subsequent studies and encouraging a shift towards more theoretically grounded methods in the field \cite{rout2023theoretical}.

\paragraph{Efficiency through GAN-based Image Inpainting}
In another contribution, a 2023 paper introduced an efficient GAN-based algorithm tailored for image inpainting tasks. This study tackled challenges such as computational efficiency and the generation of visually plausible and coherent inpainted images. By employing Generative Adversarial Networks (GANs), the proposed algorithm exhibited notable improvements in processing speed and image quality, making it useful for real-time applications and large-scale image editing tasks. The GAN-based approach capitalized on adversarial training to produce detailed and context-consistent inpainted images, potentially transforming image editing software by offering enhanced flexibility and creativity \cite{han2023gan}.

\paragraph{Comprehensive Review of Deep Learning-based Inpainting}
In 2023, a comprehensive review synthesized deep learning-based image inpainting methods. This work served as a critical resource, offering an analysis of the current state of the art and identifying directions for future research. By examining various deep learning approaches, the review highlighted trends, challenges, and advancements in the field. It emphasized the increasing reliance on convolutional neural networks (CNNs), attention mechanisms, and other ML techniques to achieve accurate and visually appealing inpainting results. The review also stressed the importance of large-scale datasets and innovative training strategies in improving inpainting model performance. Overall, the review facilitated a deeper understanding of technological advancements in image inpainting and outlined opportunities for future research and development \cite{xu2023review}.

\subsection{Algogenic Enhancements for Image Inpainting}
\subsubsection{Contextual Understanding and Semantic Coherence}

\paragraph{Refining Inpainting through Enhanced Deep Contextual Insights}
Embedding generative AI specifically within the framework of image inpainting algorithms facilitates a much deeper and nuanced understanding of the image's overall context, thereby significantly enhancing the capability to reconstruct missing or damaged portions with a high degree of semantic integrity and visual coherence. This sophisticated approach involves a comprehensive analysis of the entire visible spectrum of an image using advanced Convolutional Neural Networks (CNNs) while integrating sophisticated Natural Language Processing (NLP) techniques to enrich the contextual comprehension, which in turn, guides the generative process to produce content that is not only visually plausible but also semantically coherent with the rest of the image. The application of these cutting-edge technologies in the realm of image inpainting empowers the algorithms to more effectively tackle complex scenarios such as extensive occlusions, intricate background details, and the nuanced textures of various objects, suggesting that a profound understanding of context could substantially elevate the quality, realism, and overall aesthetic of inpainted images. However, the practical implementation of such advanced techniques necessitates careful consideration towards the allocation of computational resources and the specific complexities inherent in diverse inpainting tasks.

\paragraph{Implementing Semantic Coherence with Comprehensive Training and Caution}
Achieving a high level of semantic coherence in image inpainting, in practical terms, involves meticulously training Large Language Models on exceptionally diverse and comprehensive datasets that encompass a broad spectrum of contexts, scenarios, and visual narratives. These models are equipped with the extraordinary capability to discern complex patterns, relationships, and semantic cues within images, enabling them to accurately predict and generate the most plausible content for the missing parts, thereby ensuring both visual and semantic consistency across the inpainted imagery. For instance, the ability to recognize and contextually fill in gaps with appropriate elements such as clouds in a sky scene, or foliage in a landscape, exemplifies the model's capacity to apply its extensive learned knowledge to real-world inpainting challenges, ensuring a seamless blend between the generated and original content. However, the effectiveness and reliability of this approach are heavily dependent on the breadth and depth of the training data, as well as the model's inherent capacity to generalize from its training experiences to novel, unseen inpainting challenges.

\paragraph{Harnessing Adaptive Learning for Enhanced Contextual Relevance in Inpainting}
Adaptive learning techniques significantly amplify the contextual relevance of inpainting solutions by enabling the models to continuously refine and enhance their understanding based on an ongoing influx of new data, feedback, and real-world application scenarios. This dynamic adjustment process significantly boosts the model's ability to adeptly handle a diverse array of inpainting scenarios with increased effectiveness. Implementing adaptive learning in the specific context of image inpainting implies engaging in an iterative refinement process, wherein the model's predictions for inpainted areas progressively improve over time, ensuring that the generated content remains not only relevant and coherent with the evolving standards and expectations of visual content but also enhances the overall aesthetic and narrative coherence of the inpainted images. This approach underscores the immense potential for models to adapt and evolve in response to emerging challenges and scenarios, although it also highlights the essential need for ongoing data collection, model training, and refinement efforts to sustain and enhance performance levels over time.

\subsubsection{Adaptive Texture Synthesis}

\paragraph{Elevating Inpainting with Advanced Adaptive Texture Synthesis}
Adaptive texture synthesis, specifically within the context of Algogenic enhancements for image inpainting, represents a significant leap towards producing inpainted images that not only seamlessly blend with the original in terms of visual appeal but are virtually indistinguishable in texture detail and fidelity. This advanced technique harnesses the power of generative AI to meticulously analyze and understand the texture patterns surrounding the missing or damaged areas, subsequently synthesizing new textures that adaptively match and integrate with the contextual patterns and aesthetic nuances of the existing image content. This progressive approach enables inpainting to transcend traditional limitations, offering reconstructions that not only preserve but enhance the unique textures of each image, be it the intricate details of human skin, the rugged textures of natural terrains, or the smooth surfaces of man-made objects, suggesting a profound advancement in the capability to maintain texture consistency and coherence. The practical implementation of this enhancement hinges on the model's ability to discern a wide range of texture patterns and effectively replicate them, indicating a deep understanding of texture dynamics. However, the successful application of adaptive texture synthesis in real-world scenarios is contingent upon the diversity of the training data and the availability of substantial computational resources to process complex, high-resolution imagery.

\paragraph{Leveraging Advanced Deep Learning for Unprecedented Textural Realism}
Utilizing deep learning models for achieving an unprecedented level of textural realism within the domain of image inpainting involves leveraging sophisticated adaptive texture synthesis techniques, which are predicated on training these models on a highly diverse set of images to capture an extensive range of texture variations. This exhaustive training process enables the models to adeptly fill in missing or compromised sections of images with textures that not only blend seamlessly with the original content but also enhance the overall visual and aesthetic quality of the inpainted areas. The key to successfully implementing this enhancement lies in the utilization of vast, varied datasets that provide the models with ample examples of different textures and patterns, thereby allowing for the accurate replication of complex textural nuances in the inpainting process. This approach holds the potential to significantly elevate the visual quality and realism of inpainted images, though it also necessitates the deployment of substantial computational power and the development of sophisticated model architectures capable of achieving the desired levels of detail, texture fidelity, and overall aesthetic congruence.

\subsubsection{Cross-Modal Data Integration}

Integrating cross-modal data into the inpainting process represents a paradigm shift in enhancing the contextual depth and semantic coherence of inpainted images by incorporating a wide array of information beyond the visual domain. This innovative approach leverages diverse types of data, including text, audio, and metadata, to imbue the inpainting process with additional layers of context, insights, and interpretative depth. Large Language Models play a critical role in this integration, offering unparalleled capabilities to comprehend and process textual descriptions, annotations, or any related textual data that can provide further insights about the missing or damaged areas in an image. Moreover, LLMs excel not only in interpreting textual information but also in generating coherent and contextually relevant content, enabling them to propose highly plausible replacements for missing visual elements based on the rich textual cues provided. This fusion of textual and visual information not only significantly enhances the accuracy and semantic consistency of inpainting but also facilitates the generation of more complex, nuanced, and semantically enriched results. Additionally, LLMs aid in the seamless handling of multimodal data, allowing for a holistic understanding of the inpainting task by incorporating diverse sources of information simultaneously. Furthermore, the utilization of LLMs enables the adoption of adaptive inpainting strategies, where the model dynamically adjusts its inpainting process based on the rich content and context provided by the auxiliary data, thereby enhancing the quality of reconstructed images and opening new avenues for innovative applications in areas such as multimedia editing, content generation, and assistive technologies, where the integration of diverse data sources is crucial for achieving highly desirable and contextually enriched outcomes.

\subsubsection{Generative Model Fine-Tuning}

Fine-tuning generative models for highly specific inpainting tasks entails an intricate process of adjusting and optimizing a pre-trained model's parameters to better align with the unique characteristics, requirements, and contextual nuances of the target images. This meticulous customization process aims to significantly improve the model's ability to generate content that is not only visually plausible but also exhibits a high degree of contextual appropriateness and semantic coherence for the specific application or type of image in question. By engaging in this fine-tuning process, models are empowered to achieve a heightened level of accuracy, realism, and overall aesthetic congruence in the inpainted areas, thereby ensuring that the synthesized content seamlessly integrates with the original image's context. The practical implementation of this enhancement requires a deep and nuanced understanding of the model's architecture, the specific challenges and intricacies of the inpainting task at hand, and a strategic approach to optimizing the model's performance to meet the exacting demands of high-quality inpainting outcomes. While fine-tuning offers a promising pathway to more effective and contextually aligned inpainting solutions, it also demands a comprehensive grasp of the underlying technologies, a thoughtful consideration of the inpainting challenge's unique aspects, and a commitment to ongoing refinement and optimization efforts to realize the full potential of these advanced generative models in real-world applications.

\subsubsection{Advanced Damage Detection and Assessment}

Integrating advanced damage detection and assessment mechanisms at the outset of the inpainting process represents a critical advancement in Algogenic enhancements, setting the stage for a more targeted, efficient, and effective approach to image restoration. This preliminary phase involves deploying state-of-the-art algorithms, potentially augmented with the capabilities of Large Language Models, to conduct a thorough and meticulous analysis of images. This analysis aims to accurately identify the extent, type, and specific characteristics of damage or missing regions, thereby laying the groundwork for a more informed and precisely tailored inpainting process. By gaining a deep understanding of the nature and intricacies of the damage, the generative models are better equipped to customize their reconstruction strategies to address the unique challenges and requirements presented by each image. This advanced assessment of damage not only enhances the strategic focus of inpainting efforts, leading to improvements in efficiency and the quality of outcomes but also significantly reduces the risk of overfitting by providing a clearer contextual framework for the inpainting task, thereby ensuring the generated content is both natural and contextually appropriate. Furthermore, incorporating this comprehensive analysis into the inpainting workflow fosters greater adaptability, as the algorithms can dynamically adjust their approaches based on the complexity and nuances of the identified damage, ensuring robust performance across a diverse array of image datasets. This holistic and informed approach to inpainting, which integrates preliminary analysis with LLMs-powered reconstruction, promises to revolutionize the field of image restoration, achieving superior results in terms of fidelity, coherence, and computational efficiency, and setting a new benchmark for quality in the domain of digital image processing and enhancement.

	\subsubsection{Pseudocode for Algogenic Image Inpainting}
	The Algogenic image inpainting approach harnesses AI to enhance conventional algorithms by dynamically adapting inpainting parameters and strategies according to the system's behavior and real-time error assessments. This pseudocode, accessible in \ref{fig:image-inpainting-Algogen-pseudocode}, delineates a sophisticated framework integrating AI-driven improvements for adaptive completion control, patch selection, acceptance standards, and real-time parameter refinement.
	
	\begin{algorithm}
		\caption{Algogenic Image Inpainting Process}
		\begin{algorithmic}[1]
			\Procedure{AlgogenicImageInpainting}{Image}
			
			\Comment{Preprocessing Phase}
			\State damageMap $\gets$ AdvancedDamageDetectionAndAssessment(Image)
			\State contextualData $\gets$ CrossModalDataIntegration(Image)
			
			\Comment{Core Inpainting Phase}
			\State semanticContext $\gets$ ContextualUnderstandingAndSemanticCoherence(Image, contextualData)
			\State textureSynthesis $\gets$ AdaptiveTextureSynthesis(Image, damageMap, semanticContext)
			\If{GenerativeModelFineTuningNeeded()}
			\State fineTunedModel $\gets$ GenerativeModelFineTuning(textureSynthesis, semanticContext)
			\EndIf
			\State inpaintedImage $\gets$ ApplyInpainting(Image, fineTunedModel $\lor$ textureSynthesis)
			\State inpaintedImage $\gets$ IncorporateLLMsForEnhancedContextualization(inpaintedImage, contextualData)
			\While{Not SatisfactoryInpainting(inpaintedImage)}
			\State inpaintedImage $\gets$ IterativeRefinementBasedOnFeedback(inpaintedImage)
			\EndWhile
			
			\Comment{Postprocessing Phase}
			\State FinalImage $\gets$ ApplyEthicalAndBiasMitigationMechanisms(inpaintedImage)
			
			\EndProcedure
		\end{algorithmic}\label{fig:image-inpainting-Algogen-pseudocode}
	\end{algorithm}

	\begin{figure}
		\centering
		\includegraphics[width=0.58\textwidth]{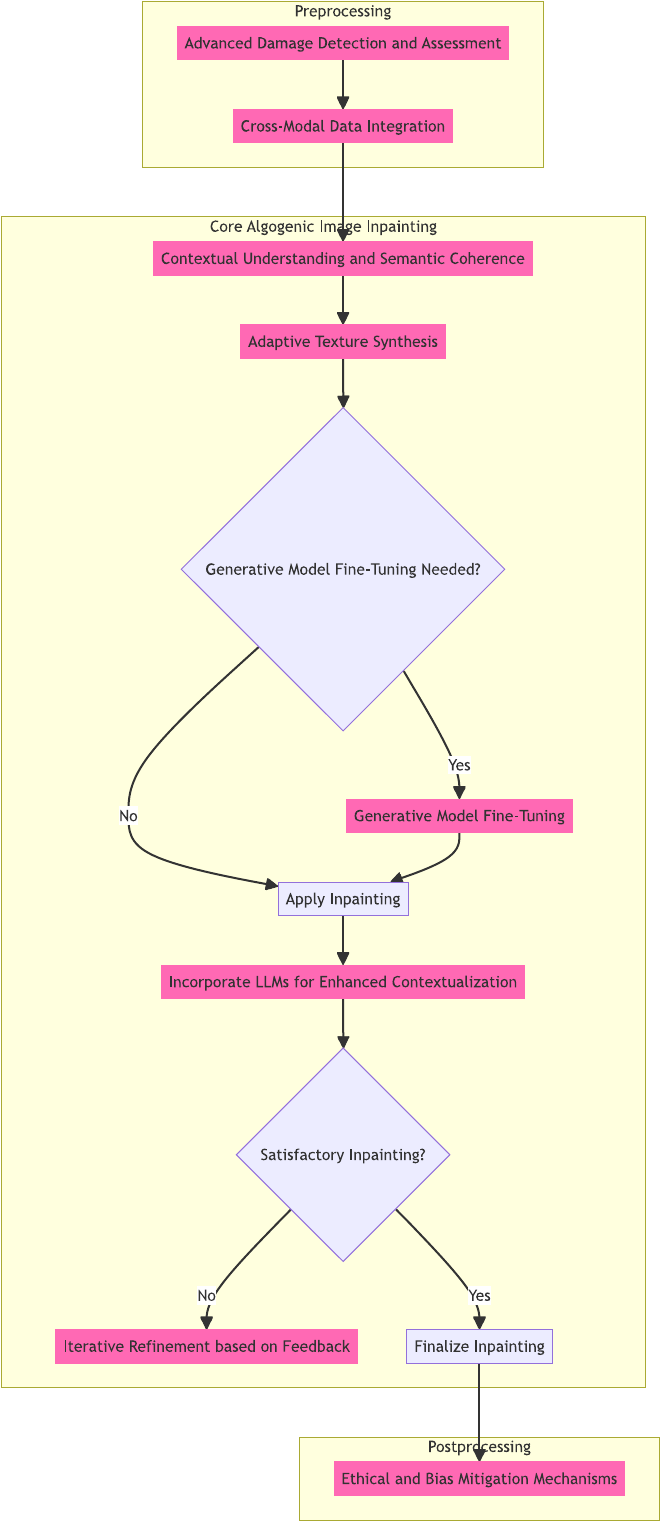}
		\caption{Advancing Image Inpainting through Algogenic Enhancements: This figure would conceptualize the integration of Algogenic enhancements within the image inpainting process, highlighting the transformation from advanced damage detection and cross-modal data integration in the preprocessing phase to the core application of contextual understanding, semantic coherence, and adaptive texture synthesis. It illustrates the dynamic, iterative refinement based on generative AI insights and the application of ethical and bias mitigation mechanisms in postprocessing. This comprehensive approach underlines how generative AI not only optimizes the technical aspects of image inpainting but also ensures the generated content is contextually coherent and semantically accurate, thereby significantly enhancing the realism and quality of inpainted images.}
		\label{fig:image_inpainting}
	\end{figure}

	\section{Style Transfer}\index{Style Transfer}
	\subsection{Introduction to Style Transfer}
	\subsubsection{The Concept of Style Transfer}
	
	\paragraph{Defining Style Transfer}
	The concept of Style Transfer in image processing refers to the computational technique of applying the visual style of one image, known as the style image, onto the content of another image, referred to as the content image. This process results in a new image that retains the original content but is rendered in the aesthetic style of the style image. Style transfer bridges the gap between content and expression, allowing for the creation of images that combine the structural elements of one picture with the texture, color, and stroke patterns of another. This technique has its roots in the field of computer vision and deep learning, where neural networks learn to encapsulate and transfer artistic styles from one image to another.
	
	Furthermore, this process of style transfer is not merely a superficial overlay of one image onto another; rather, it involves complex computational algorithms that analyze and extract the underlying stylistic features of the style image. Moreover, these algorithms leverage deep neural networks, particularly convolutional neural networks (CNNs), to capture intricate details of both content and style images. Additionally, the integration of perceptual loss functions, such as those derived from pretrained CNNs like VGG or ResNet, enhances the fidelity of style transfer by ensuring that the generated images not only mimic the style but also preserve semantic content. Consequently, style transfer algorithms can produce visually appealing results that exhibit a harmonious blend of content and style, facilitating creative expression and artistic exploration in digital media. Hence, style transfer serves as a powerful tool in various domains, including graphic design, photography, and visual arts, enabling artists and designers to manipulate and transform images in innovative ways.

	\paragraph{Technical Foundations}
	At the heart of style transfer is the use of Convolutional Neural Networks (CNNs), which analyze and understand the distinct features that define the style of an artwork and the content of a photograph or any other type of image. The process typically involves three images: a content image, a style reference image, and an output image that starts as a white noise image and is iteratively updated. The style transfer algorithm optimizes the output image so that its content resembles the content image while stylistically mirroring the style reference image. This optimization is achieved by minimizing a loss function that comprises two main components: a content loss that ensures content fidelity to the original image, and a style loss that ensures stylistic similarity to the style reference.
	
	Furthermore, this process can be computationally intensive due to the iterative nature of optimization and the complex computations involved in CNNs. Moreover, the choice of the CNN architecture and the layers used for feature extraction greatly influence the quality of the style transfer. Additionally, hyperparameters such as learning rate and number of iterations play crucial roles in determining the convergence and effectiveness of the algorithm. Hence, thorough experimentation and tuning are essential for achieving desirable results in style transfer applications. On the other hand, advancements in hardware acceleration and parallel processing have significantly improved the efficiency of style transfer algorithms, enabling real-time or near-real-time performance in some cases. Consequently, style transfer has found applications not only in artistic image generation but also in various fields such as photography, graphic design, and augmented reality.

	\paragraph{Mathematical Representation}
	Mathematically, the style transfer process can be represented by the optimization problem
	\[
	\min_{I_{out}} \alpha L_{content}(I_{out}, I_{content}) + \beta L_{style}(I_{out}, I_{style}),
	\]
	where \(I_{out}\) is the output image, \(I_{content}\) is the content image, \(I_{style}\) is the style image, \(L_{content}\) and \(L_{style}\) are the content and style loss functions respectively, and \(\alpha\) and \(\beta\) are weighting factors that balance the influence of content and style in the final output. The content loss ensures that the high-level features of the content image are preserved in the output image, while the style loss ensures that the textures, colors, and visual patterns of the style image are captured.
	
	\paragraph{Evolution and Applications}
	Since its inception, style transfer has evolved significantly, with advancements in AI and machine learning further enhancing its capabilities and applications. The synergy between deep learning techniques and image processing has propelled style transfer beyond mere artistic endeavors, enabling its integration into various fields. For instance, in film production, style transfer algorithms are employed to mimic the visual aesthetics of renowned filmmakers or to create distinctive visual signatures for different scenes, enhancing the cinematic experience. In video games, style transfer algorithms contribute to creating immersive environments by dynamically adapting the visual style to match the game's narrative or genre, thereby enhancing player engagement and realism. Moreover, in virtual reality applications, style transfer techniques help in rendering realistic and visually appealing virtual worlds, enriching the user experience and blurring the lines between reality and simulation. Furthermore, the therapeutic potential of style transfer in art creation has gained attention, with studies suggesting its effectiveness in promoting relaxation, stress relief, and self-expression. By enabling individuals without formal artistic training to produce aesthetically pleasing artworks, style transfer democratizes art creation and fosters inclusivity in the creative process. This democratization not only empowers individuals but also fosters a culture of experimentation and exploration, driving innovation in digital content creation. Thus, the evolution of style transfer from its artistic origins to its multifaceted applications underscores its transformative impact on diverse domains, revolutionizing the way we create, perceive, and interact with visual content.

	\subsubsection{Key Principles and Mechanisms}
	
	\paragraph{Core Principles of Style Transfer}
	The core principles of style transfer revolve around the decomposition of images into content and style representations, leveraging the capabilities of neural networks. The content of an image refers to the basic shapes and structure that define the image's recognizable features, such as the outline of objects, the spatial arrangement of major elements, and the overall composition. The style of an image, on the other hand, encompasses the visual aesthetic that characterizes the image, including textures, colors, brush strokes, and patterns. The primary goal of style transfer is to synthesize these two aspects from different sources into a single cohesive image that maintains the original content's integrity while adopting the artistic style of another image.
	
	Furthermore, it is essential to note that style transfer algorithms typically employ convolutional neural networks (CNNs) to extract both content and style representations efficiently. These networks are pretrained on large datasets to capture high-level features, enabling them to discern semantic content and abstract style elements effectively. Moreover, the process often involves defining loss functions that quantify the disparity between the content and style representations of the input and target images, respectively. These loss functions guide the optimization process, allowing the generated image to gradually converge towards a visually pleasing combination of content and style.
	
	Additionally, advancements in style transfer techniques have led to the development of various approaches, including neural style transfer, which utilizes pretrained CNNs to extract style features from a reference image and apply them to a content image. Likewise, style transfer methods based on generative adversarial networks (GANs) introduce adversarial training strategies to enhance the realism and coherence of synthesized images. Moreover, researchers continue to explore novel architectures and loss formulations to address challenges such as preserving fine details in the content while faithfully capturing the desired style characteristics.

	\paragraph{Mechanisms of Neural Style Transfer}
	The mechanisms underlying neural style transfer leverage deep learning models, particularly Convolutional Neural Networks (CNNs), as pivotal tools trained to comprehend and manipulate both content and style aspects within images. CNN architectures designated for style transfer typically comprise multiple layers, each progressively extracting intricate features from the input image. Initially, lower layers capture fundamental features like edges and textures, essential for accurate style representation, while higher layers discern more abstract features defining the content. The style transfer procedure entails delineating and optimizing a loss function encompassing two primary constituents: the content loss and the style loss. The content loss ensures that the resultant image bears resemblance to the original content, while the style loss ensures that the output image exhibits a stylistic resemblance to the reference image. This optimization is often executed via backpropagation, wherein the pixel values of the output image are iteratively adjusted to minimize the overall loss. This iterative process involves updating the output image until convergence is achieved, facilitating the generation of visually compelling stylized images. Furthermore, advancements in neural style transfer techniques have led to the development of various algorithms optimizing style transfer efficiency and enabling real-time applications.

	\paragraph{Mathematical Formulation of Loss Functions}
	The mathematical formulation of the style transfer involves defining the content loss and style loss in a way that allows their joint minimization. The content loss is usually defined as the Mean Squared Error (MSE) between the feature representations of the content image and the output image at certain layers within the CNN, expressed as
	\[
	L_{content} = \frac{1}{2} \sum_{i,j} \left( F_{ij}^{l} - P_{ij}^{l} \right)^2,
	\]
	where \(F_{ij}^{l}\) and \(P_{ij}^{l}\) represent the feature representations of the output and content images at layer \(l\), respectively. The style loss, on the other hand, is often calculated using the Gram matrix of the feature maps from the style image and the output image to capture the style information, defined as
	\[
	L_{style} = \sum_{l} \frac{1}{4N_l^2M_l^2} \sum_{i,j} \left( G_{ij}^{l} - A_{ij}^{l} \right)^2,
	\]
	where \(G_{ij}^{l}\) and \(A_{ij}^{l}\) are the Gram matrices of the output and style images at layer \(l\), \(N_l\) is the number of feature maps, and \(M_l\) is the size of the feature map. The overall loss is a weighted sum of the content and style losses, which the optimization process seeks to minimize.
	
	\paragraph{Evolution of Techniques and Real-time Applications}
	Over time, style transfer techniques have evolved from iterative optimization-based approaches to more efficient methods that utilize feed-forward networks for real-time style transfer. Initially, style transfer heavily relied on iterative optimization algorithms, which often resulted in time-consuming processes, hindering its application in real-time scenarios. However, with the emergence of feed-forward networks, such as Convolutional Neural Networks (CNNs) and Generative Adversarial Networks (GANs), style transfer has undergone a significant transformation. These feed-forward networks enable instantaneous transformation of images, allowing for real-time application of style transfer in various domains.
	
	Moreover, the evolution of style transfer techniques has facilitated its integration into diverse real-time applications. For instance, in video processing, real-time style transfer enables the seamless application of artistic styles to live video streams, enhancing visual aesthetics and creativity. Similarly, in interactive web and mobile applications, users can dynamically apply different styles to their photos or videos in real-time, enriching user experiences and engagement. Furthermore, in augmented reality environments, real-time style transfer can overlay virtual elements with consistent stylistic coherence onto the real world, enhancing immersion and blending virtual and physical realities.
	
	The advancements in style transfer techniques have not only democratized artistic expression but also opened new avenues for commercial applications. Industries ranging from entertainment and advertising to fashion and gaming are leveraging real-time style transfer to create captivating visual content, personalized user experiences, and innovative marketing campaigns. As style transfer mechanisms and models continue to evolve, driven by advancements in deep learning and computational efficiency, the possibilities for its creative and commercial utilization are boundless, promising a future where visual content transcends conventional boundaries and sparks new forms of expression and interaction.

	\subsubsection{The Role of Deep Learning}
	
	\paragraph{Foundational Impact on Style Transfer}
	Deep learning has played a pivotal role in the advancement and widespread adoption of style transfer techniques in image processing. The introduction of Convolutional Neural Networks (CNNs) to this field marked a significant departure from traditional methods, enabling the automated and sophisticated analysis of both content and style elements within images. Deep learning models, particularly those pre-trained on vast datasets of images, have the unique ability to extract and abstract complex features from an image, ranging from simple textures and patterns to intricate compositions and structures that define the image's content and style.
	
	Furthermore, the utilization of CNNs in style transfer has not only revolutionized the field but has also democratized the creation of artistic and visually appealing images. Moreover, the iterative refinement of deep learning architectures, such as Generative Adversarial Networks (GANs) and Variational Autoencoders (VAEs), has led to the development of more sophisticated and nuanced style transfer methods. Additionally, the integration of attention mechanisms and multi-scale processing has further enhanced the fidelity and realism of stylized images.
	
	Consequently, the synergy between deep learning and style transfer has paved the way for a myriad of applications across various domains, including digital art, graphic design, and augmented reality. Moreover, the accessibility of pre-trained models and open-source frameworks has lowered the barrier to entry for researchers and practitioners, fostering innovation and experimentation in the field of image stylization. Thus, deep learning continues to serve as the cornerstone of modern style transfer, driving advancements and pushing the boundaries of creative expression in image processing.

	\paragraph{Mechanisms Enabled by Deep Learning}
	The mechanisms of style transfer powered by deep learning involve several key processes. Initially, a CNN analyzes the style reference image to capture its distinctive stylistic features, such as brush strokes, color palettes, and texture patterns, often encoded in the activations of the network's earlier layers. \textbf{Furthermore}, the same network processes the content image to extract its high-level content features, typically represented in the deeper layers of the network. The style transfer \textbf{then} occurs through the optimization of a target image that simultaneously minimizes the distance from the content features of the content image and the style features of the style image, effectively merging the two sets of features into a single coherent output. \textbf{Moreover}, the optimization process involves iterative updates to the target image guided by loss functions that measure the differences in content and style features between the target image and the content/style reference images. This iterative process ensures that the generated output maintains the content of the content image while adopting the stylistic characteristics of the style reference image. Additionally, the deep learning framework allows for the customization of style transfer models, enabling users to control the strength of style transfer, adjust the balance between content and style preservation, and even explore novel artistic creations through the manipulation of input parameters. The versatility and effectiveness of deep learning-based style transfer have led to its widespread adoption in various applications, including image editing, artistic rendering, and visual content generation.

	\paragraph{Advancements in Optimization Techniques}
	Deep learning has also facilitated advancements in optimization techniques for style transfer. Early methods relied on slow, iterative processes to adjust the output image gradually. These iterative approaches necessitated numerous iterations to achieve the desired balance between content and style, resulting in considerable processing time. However, recent developments in the field have led to the adoption of feed-forward neural networks, marking a significant departure from iterative optimization. These networks are capable of performing style transfer in a single pass, resulting in a substantial reduction in processing time. By leveraging feed-forward neural networks, style transfer can now be accomplished almost in real-time, enabling applications with stringent latency requirements.
	
	Furthermore, the transition to feed-forward networks has fundamentally altered the approach to style transfer. Traditionally, style transfer algorithms required pairs of content and style images to be optimized iteratively, adjusting the output image until convergence. In contrast, feed-forward networks are trained on such pairs to directly learn the transformation. This approach eliminates the need for iterative optimization during inference, allowing for the instantaneous application of learned styles to new content images. Consequently, the computational burden associated with style transfer has been significantly alleviated, opening doors to a wide range of applications previously constrained by processing limitations.
	
	Moreover, the efficiency gains achieved through feed-forward neural networks have democratized style transfer, making it more accessible to a broader audience. Previously, the computational demands of iterative optimization restricted style transfer to high-performance computing environments. However, with the advent of efficient feed-forward networks, style transfer can now be performed on a variety of devices, including smartphones and tablets, enabling on-the-go creativity and expanding the reach of artistic expression.
	
	In summary, the integration of feed-forward neural networks has revolutionized optimization techniques for style transfer, offering unprecedented speed and efficiency while democratizing access to this creative tool.

	\paragraph{Mathematical Framework and Loss Optimization}
	The deep learning approach to style transfer relies on a sophisticated mathematical framework to achieve its objectives. At its core, this framework defines a comprehensive loss function, comprising content loss, style loss, and often additional regularization terms. The content loss ensures that the generated output faithfully captures the essence and structure of the target image. On the other hand, the style loss ensures that the artistic characteristics and textures from the style reference are effectively infused into the output. These losses are carefully balanced to strike a harmonious blend between content preservation and style replication. Moreover, the incorporation of regularization terms serves to maintain coherence and smooth transitions within the stylized output, preventing artifacts and enhancing visual fidelity.
	
	The optimization of this multifaceted loss function is pivotal in achieving desirable style transfer results. Through iterative optimization techniques, such as backpropagation, the parameters of the model are adjusted to minimize the loss. Backpropagation computes gradients of the loss function with respect to the model parameters, enabling efficient updates that steer the output image towards the desired aesthetic representation. This iterative refinement process iteratively adjusts the pixel values of the output image, gradually aligning it with both the content of the target image and the artistic style of the reference. Consequently, the stylized output emerges as a seamless fusion of content and style, reflecting the intricacies of the original image while encapsulating the desired artistic flair. The elegance of this approach lies in its ability to harness the power of deep learning and optimization algorithms to seamlessly blend artistic vision with computational prowess.

	\paragraph{Future Directions and Potential}
	The integration of deep learning into style transfer continues to open new avenues for research and application. \textbf{Furthermore}, it fosters the development of models capable of understanding and applying multiple styles simultaneously, \textbf{as well as} adapting styles based on semantic understanding of content. \textbf{Moreover}, it enables the creation of dynamic styles that evolve in real-time. As deep learning models become more sophisticated and capable of processing more complex and nuanced data, the potential for innovative style transfer applications expands \textbf{furthermore}, promising to revolutionize the ways in which we create, interact with, and interpret visual content. \textbf{Additionally}, the intersection of deep learning with style transfer opens up possibilities for exploring interdisciplinary collaborations. \textbf{Similarly}, the integration of stylistic elements with deep learning architectures can lead to advancements in fields such as fashion design, architecture, and digital art. \textbf{Moreover}, as the computational resources continue to advance, the scalability of deep learning models for style transfer can \textbf{likewise} increase, facilitating their deployment in real-world scenarios. \textbf{Consequently}, this convergence of deep learning and style transfer not only enhances the creativity and expressiveness of visual content but also paves the way for novel applications across various domains.

	\subsubsection{Applications and Limitations}
	
	\paragraph{Broad Spectrum of Applications}
	The advent of style transfer, powered by deep learning, has unleashed a plethora of applications that span across various domains. In the realm of digital art and design, artists and designers harness style transfer to create stunning artworks that blend classic styles with contemporary subjects, bridging the gap between traditional art forms and modern digital expression. The film and entertainment industry benefits from style transfer by applying unique visual effects in post-production, offering viewers an immersive experience that traditional filming techniques cannot achieve. Additionally, in the domain of fashion and interior design, style transfer aids in visualizing patterns and textures in new contexts, enabling designers to experiment with creative combinations effortlessly.
	
	The educational sector also sees value in style transfer, using it as a tool to engage students in art history and computer science by demonstrating the fusion of technology and creativity. Furthermore, the technology extends to social media and mobile applications, where users can personalize their photos and videos with various artistic styles, enhancing social interaction and content creation. Another innovative application is in therapy and mental health, where style transfer provides a means for emotional expression and relaxation through art creation, even for individuals without formal art training.
	
	\paragraph{Encountering Limitations and Challenges}
	Despite its impressive capabilities, style transfer is not without limitations. One of the primary challenges is maintaining the balance between content preservation and style application. In some instances, the application of a style can overly dominate or distort the original content, leading to results that may not faithfully represent the intended subject matter. Moreover, the quality of style transfer is heavily dependent on the complexity and compatibility of the chosen style and content images, with some combinations yielding less coherent or aesthetically pleasing outcomes than others.
	
	Another significant limitation is the computational demand of deep learning-based style transfer, especially for high-resolution images or real-time video processing. The requirement for substantial computational resources can limit accessibility for individuals and organizations without access to advanced hardware. Additionally, the automated nature of style transfer raises concerns about the potential for copyright infringement and ethical considerations, particularly when styles characteristic of specific artists are applied without permission.
	
	\paragraph{Navigating Towards Future Developments}
	Looking forward, the field of style transfer continues to evolve, with ongoing research focused on addressing its current limitations and expanding its applications. Efforts to develop more efficient and versatile models aim to reduce computational demands and improve the accessibility of style transfer technology. Advances in understanding the nuanced interplay between content and style at a deeper level promise to enhance the quality and fidelity of stylized outputs. Moreover, the integration of interactive and adaptive learning elements into style transfer systems is poised to offer users greater control over the stylization process, tailoring results more closely to individual preferences and objectives.
	
	As the technology matures, it is also anticipated that ethical frameworks and copyright guidelines specific to digital art and style transfer will emerge, guiding the responsible use and commercialization of this innovative tool. The future of style transfer lies in the harmonious integration of art, technology, and ethics, opening new horizons for creative expression and digital innovation.

	\subsubsection{Algorithmic Pseudocode for Style Transfer}
	Style transfer in image processing involves a sophisticated framework designed to seamlessly merge the aesthetic qualities of one image onto the content structure of another. This process, outlined in the provided pseudocode (see Figure \ref{fig:style-transfer-pseudocode}), leverages a pre-trained Convolutional Neural Network (CNN) to achieve its objectives. Initially, the output image is initialized with the content image, serving as the canvas upon which the stylistic elements will be imposed. Subsequently, features are extracted from both the content and style images using the CNN. To ensure fidelity to the original content, a content loss is computed, while a style loss ensures faithful replication of the style image's aesthetic attributes. The optimization objective is the minimization of a total loss, which is a weighted sum of these two losses. Through iterative backpropagation, the output image is refined until convergence, signifying successful style transfer. Post-processing techniques may then be employed to enhance the visual quality of the stylized image, ensuring it remains visually appealing while preserving the integrity of the original content's structure. For reference see figure \ref{fig:style-transfer-pseudocode}.
	
	\begin{algorithm}
		\caption{Algorithmic Pseudocode for Style Transfer}
		\begin{algorithmic}[1]
			\Procedure{StyleTransfer}{ContentImage, StyleImage, OutputImage}
			\State Initialize OutputImage with ContentImage
			\State Load pre-trained Convolutional Neural Network
			\While{not converged}
			\State Extract features from ContentImage and StyleImage
			\State Compute content loss between OutputImage and ContentImage
			\State Compute style loss between OutputImage and StyleImage
			\State Total loss $\gets$ weighted sum of content and style losses
			\State Use backpropagation to minimize total loss
			\State Update OutputImage to reduce total loss
			\EndWhile
			\State Apply post-processing to enhance visual quality of OutputImage
			\State \Return OutputImage
			\EndProcedure
		\end{algorithmic}\label{fig:style-transfer-pseudocode}
	\end{algorithm}

\subsection{Previous Work on ML and AI Interplay with the Algorithm}

\paragraph{Contextual Style Transfer in Medical Imaging}
Recent work by Xu et al. (2020) explores the integration of machine learning and artificial intelligence into medical image processing, focusing on contextual style transfer. This method utilizes deep learning techniques to enhance the clarity and interpretability of medical images, aiding in diagnosis and treatment planning. By adapting style transfer from artistic imagery to medical contexts, the approach demonstrates efficacy in improving the visual representation of complex biological structures. This facilitates a more intuitive understanding of medical imagery by healthcare professionals and opens avenues for automated image analysis systems, showcasing potential for future research in medical image enhancement through AI-driven methodologies.

\paragraph{Comprehensive Review on Neural Style Transfer}
Singh et al. (2021) provide a comprehensive review on neural style transfer, discussing its advancements and challenges. The review highlights the technique's versatility across various applications beyond the arts, including fashion and advertising. It analyzes the mechanisms enabling the blending of content and style from different images and identifies areas for future exploration, such as improving algorithm efficiency and maintaining semantic integrity. By synthesizing insights from various studies, the review guides research efforts toward addressing limitations and unlocking neural style transfer's potential in artistic and practical applications.

\paragraph{Exploring Texture and Art with Deep Neural Networks}
Gatys et al. (2017) explore texture synthesis and artistic style transfer using deep neural networks, representing a foundational milestone in the convergence of art and artificial intelligence. Their work demonstrates how convolutional neural networks decode and replicate stylistic elements of artwork, generating new images that combine original content with aesthetic qualities of famous paintings. This research broadens understanding of computational representation of artistic styles and sets the stage for AI applications in creative industries such as graphic design and digital media. The methodologies developed have implications beyond art reproduction, influencing fields like therapeutic visual arts, showcasing AI's potential to augment human creativity.

\subsection{Algogenic Enhancements for Style Transfer}
\subsubsection{Semantic Understanding for Style Application}

\paragraph{Refining Contextual Understanding in Style Transfer}
The application of Large Language Models in refining the semantic understanding crucial for nuanced style application in images represents a thoughtful Algogenic enhancement. By analyzing textual narratives related to images or interpreting nuanced user inputs, LLMs facilitate a deeper grasp of both the explicit and implicit attributes desired in the style transfer outcome. This process ensures that the integration of content and style not only maintains aesthetic harmony but also aligns with the contextual underpinnings of the source material. The adaptability of LLMs to subtle contextual variations enhances the precision of style application, offering dynamic fine-tuning capabilities based on evolving user feedback or input context. Such integration opens new creative pathways, enabling the generation of images that resonate more deeply with the semantic intent behind the style application, thereby offering a more tailored and expressive user experience.

\paragraph{Elevating Style Relevance through Deeper Semantic Insights}
Incorporating LLMs for enhanced semantic analysis fundamentally transforms the process of style transfer, moving beyond simple pixel manipulation. This approach leverages LLMs to uncover the deeper meaning within image content, enabling the selection and application of styles that are inherently more coherent with the content's essence. By identifying and interpreting key elements like mood, thematic underpinnings, and significant motifs through a comprehensive semantic analysis, the style transfer process becomes more informed and intentional. For example, an LLM could recognize the calmness of a landscape, suggesting a style transfer that accentuates its serene beauty, or detect the bustling energy of a city scene, choosing a style that reflects its dynamic vibrancy. This method ensures that the style application is not only visually appealing but also deeply resonant with the original content's semantic context, enhancing the perceptual experience.

\paragraph{Aligning Style Transfer with User Preferences}
Utilizing LLMs in style transfer facilitates a shift towards a more interactive and user-driven approach. By interpreting users' descriptive inputs about their stylistic preferences, LLMs act as a bridge between abstract artistic desires and their tangible realization. This capability allows for a nuanced interpretation of user inputs, translating them into specific directives for the style transfer algorithm, thus offering a customized artistic experience. Such a user-centric approach democratizes the art creation process, enabling individuals without technical or artistic backgrounds to express their creative visions effortlessly. This not only makes the technology more accessible but also encourages a collaborative creative process, where users feel more connected to the outcomes, viewing them as true reflections of their personal artistic intentions.

\subsubsection{Natural Language-Driven Style Modification}

\paragraph{Empowering User Engagement through Descriptive Inputs}
The integration of LLMs for Natural Language-Driven Style Modification significantly enhances user interaction with style transfer technology. By understanding and acting on user descriptions, this approach allows for a personalized style application, reflective of individual creative visions. This direct interaction via natural language enriches the user experience by providing a platform for expressing detailed stylistic preferences and intentions, which the algorithm then interprets to tailor the style transfer process. Such empowerment encourages exploration and personal expression, bridging the gap between complex image processing techniques and user creativity.

\paragraph{Translating Linguistic Descriptions into Visual Styles}
LLMs play a transformative role in decoding user-provided natural language descriptions into specific stylistic adjustments within the style transfer process. This capability to understand and translate descriptive language into visual modifications enables a nuanced and personalized approach to style application. Users can articulate their vision in familiar terms, guiding the algorithm to achieve desired aesthetic outcomes. This symbiosis between linguistic input and visual output fosters a more accessible and engaging creative process, allowing users to explore and refine their artistic preferences with unprecedented ease.

\paragraph{Facilitating Personalized Artistic Creation}
By enabling descriptive control over the style transfer process, this Algogenic enhancement democratizes digital art creation, allowing for a deeper personal connection with the artwork. This approach not only makes advanced image processing techniques more accessible but also encourages a diverse range of creative expressions. The iterative nature of this process, supported by LLM feedback, allows for continuous refinement, enabling users to develop their artistic visions in a collaborative manner with the algorithm, thereby enhancing the personal value and emotional engagement with the created artwork.

\subsubsection{Cross-Modal Style Synthesis}

\paragraph{Integrating Textual Concepts with Visual Artistry}
Cross-Modal Style Synthesis represents a forward-thinking enhancement in style transfer, utilizing LLMs to bridge the conceptual gap between textual descriptions and visual style applications. This innovative approach allows users to express styles in descriptive language, which the system translates into visual representations, thereby expanding the creative boundaries of style transfer. The ability to generate visual styles from textual descriptions not only enriches the creative palette but also underscores the potential for a more personalized and concept-driven approach to digital artistry, pushing the limits of traditional style application methods.

\paragraph{Implementing Text-to-Visual Style Translation Mechanisms}
The mechanism behind Cross-Modal Style Synthesis showcases the critical role of LLMs in translating textual descriptions into actionable stylistic elements. By parsing descriptive inputs, LLMs identify and interpret the essence of the requested style, facilitating a nuanced transformation of visual elements to align with the described aesthetics. This dynamic translation process, from textual narrative to visual style, underscores the potential for creating artwork that precisely reflects complex conceptual inspirations, offering a new dimension of creative expression in digital art.

\paragraph{Enhancing Accessibility and Expressiveness in Digital Art}
Cross-Modal Style Synthesis enriches the landscape of digital art by making style transfer more accessible and expressive. This approach allows for the exploration of unique, conceptual styles derived from textual descriptions, broadening the horizons of artistic creation. By fostering an environment where creativity is not bound by pre-existing visual references, LLMs enable a more inclusive and exploratory approach to art creation, encouraging diversity in artistic expression and facilitating personalized storytelling through visual media.

\subsubsection{Algogenic Approach to Adaptive Style Application}

\paragraph{Ensuring Stylistic Coherence Across Dynamic Content}
The Algogenic Approach to Adaptive Style Application leverages the analytical capabilities of LLMs to maintain stylistic coherence across dynamic media, such as video content. By dynamically adjusting stylistic applications in response to content changes, this approach ensures that the artistic intent remains consistent throughout the piece. The integration of LLMs allows for real-time interpretation of content dynamics, facilitating adaptive style modifications that preserve the aesthetic and narrative continuity of the media, thereby enhancing the overall viewing experience.

\paragraph{Adapting Style Application Through Contextual Insights}
Incorporating LLMs for dynamic adaptation within style transfer processes enables a sophisticated understanding of content changes and user intentions. By analyzing various data forms related to the content, LLMs provide actionable insights that guide the adaptive modulation of stylistic applications, ensuring relevance and consistency across changing scenes or themes. This capacity for nuanced understanding and adaptation enhances the personalization and engagement of the style transfer process, catering to the evolving preferences of users and the dynamic nature of the content.

\paragraph{Optimizing Style Transfer for Dynamic Media}
The formulation of an adaptive style transfer model highlights the integration of LLM insights to dynamically adjust stylistic applications in response to content evolution. This model utilizes a novel approach to ensure continuity and coherence in dynamically changing content, leveraging both visual and textual cues. The ability to dynamically adjust style parameters based on LLM-derived insights marks a significant advancement in the field, offering new creative possibilities for dynamic media production and storytelling.

\subsubsection{Algogenic Enhancement for Enhanced Realism and Contextual Coherence}

\paragraph{Improving Realism and Contextual Coherence in Stylization}
The Algogenic Enhancement for Enhanced Realism and Contextual Coherence capitalizes on LLMs to assess and refine the realism and contextual alignment of stylized outputs. This enhancement ensures that applied styles are not only visually appealing but also contextually appropriate, enhancing the believability and immersive quality of the stylized images. By integrating LLMs for contextual analysis, the style transfer process becomes more adaptable and sensitive to the nuances of the source content, facilitating stylizations that are both aesthetically captivating and semantically coherent.

\paragraph{Utilizing LLMs for Enhanced Realism and Coherence Evaluation}
LLMs function as evaluators within this Algogenic framework, scrutinizing stylized outputs against real-world norms and contextual expectations. This evaluation ensures that the stylization process respects the inherent logic and structure of the original content, enhancing the coherence and realism of the final output. The role of LLMs in maintaining the semantic integrity and narrative coherence of the stylized images underscores the potential for creating more immersive and contextually aligned visual experiences, elevating the standard of digital art creation.

\paragraph{Expanding Creative Potential through Contextually Coherent Stylization}
This enhancement extends the creative possibilities of style transfer, enabling the production of artworks that are not only visually striking but also deeply grounded in their contextual narrative. By prioritizing realism and contextual coherence, this approach addresses the challenge of balancing artistic expression with the logical constraints of the depicted scenes, thereby enriching the user experience and expanding the applicability of style transfer technology across diverse digital media landscapes.

	\subsubsection{Pseudocode for Algogenic Style Transfer}
	The Algogenic style transfer in image processing approach harnesses AI to enhance conventional style transfer methods by dynamically adapting parameters and strategies based on the system's behavior and real-time error estimates. This pseudocode, accessible in \ref{fig:style-transfer-Algogen-pseudocode}, delineates a sophisticated framework integrating AI-driven improvements for adaptive style blending, feature selection, similarity metrics, and real-time parameter optimization.

	\begin{algorithm}
		\caption{Algogenic Style Transfer Process}
		\begin{algorithmic}[1]
			\Procedure{AlgogenicStyleTransfer}{ContentImage, StyleDescription}
			
			\Comment{Preprocessing Phase}
			\State contextualInsights $\gets$ LeverageLLMsForContextualInsights(ContentImage, StyleDescription)
			\State styleModifications $\gets$ NaturalLanguageDrivenStyleModification(StyleDescription)
			
			\Comment{Core Style Transfer Phase}
			\State semanticApplication $\gets$ ApplySemanticUnderstandingForStyle(contextualInsights, styleModifications)
			\State styleSynthesis $\gets$ CrossModalStyleSynthesis(StyleDescription, semanticApplication)
			\If{AdaptiveStyleApplicationNeeded(ContentImage)}
			\State adaptiveStyle $\gets$ AlgogenicApproachToAdaptiveStyleApplication(ContentImage, styleSynthesis)
			\Else
			\State adaptiveStyle $\gets$ styleSynthesis
			\EndIf
			\State outputImage $\gets$ ApplyStyleTransfer(ContentImage, adaptiveStyle)
			\State outputImage $\gets$ EnhanceRealismAndContextualCoherence(outputImage, contextualInsights)
			\While{Not SatisfactoryOutcome(outputImage)}
			\State outputImage $\gets$ IterativeRefinementBasedOnFeedback(outputImage)
			\EndWhile
			
			\Comment{Postprocessing Phase}
			\State FinalImage $\gets$ ApplyEthicalAndBiasConsiderations(outputImage)
			
			\EndProcedure
		\end{algorithmic}\label{fig:style-transfer-Algogen-pseudocode}
	\end{algorithm}

	\begin{figure}
		\centering
		\includegraphics[width=0.6\textwidth]{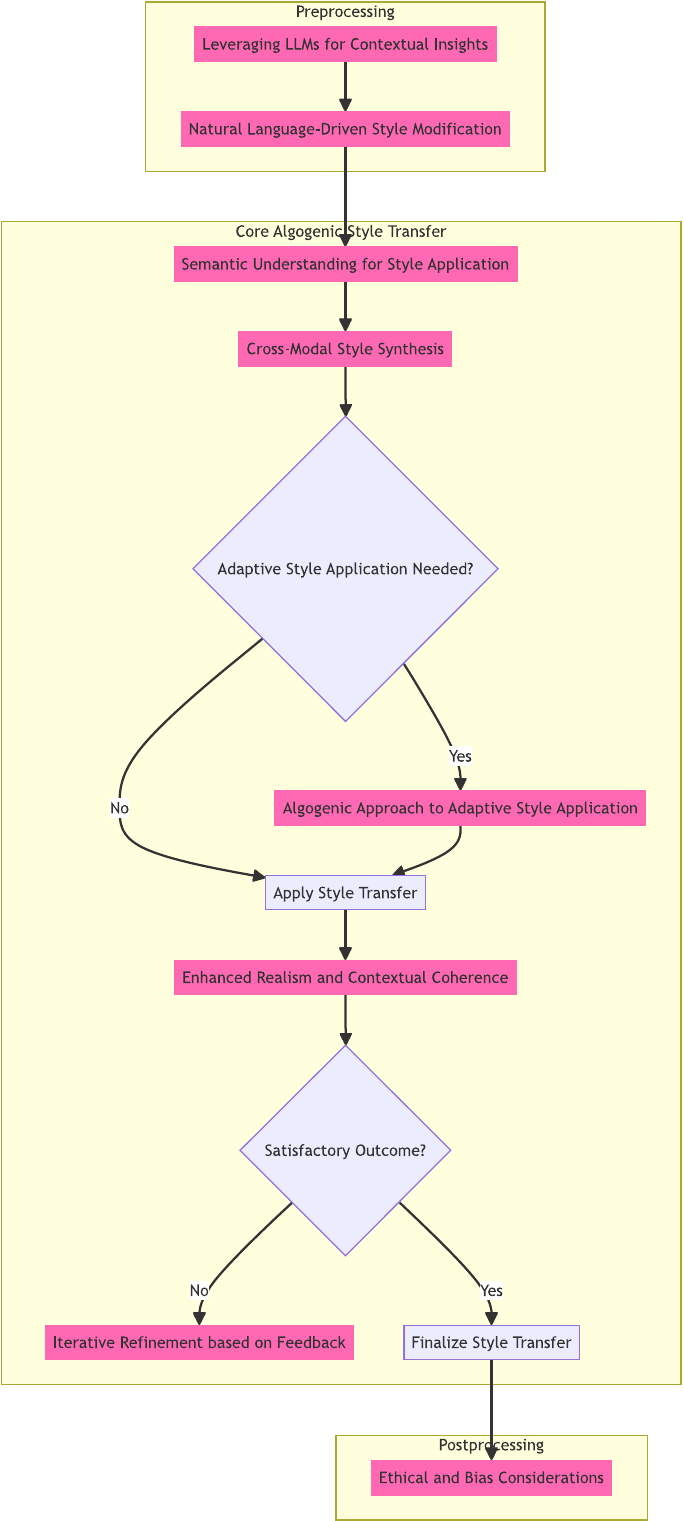}
		\caption{Innovating Artistic Expression through Algogenic Style Transfer: This diagram visualizes the Algogenic framework's application to the style transfer process, highlighting the fusion of generative AI with algorithmic precision. Beginning with preprocessing that leverages Large Language Models for deep contextual insights and natural language-driven style modifications, the framework progresses through a core phase that intricately applies semantic understanding and cross-modal style synthesis. Adaptive strategies ensure the style remains consistent across dynamic content, enhanced by iterative refinements for realism and contextual coherence. This sophisticated integration results in style transfers that not only achieve aesthetic excellence but also respect the semantic integrity and user's creative vision, pushing the boundaries of digital artistry.}
		\label{fig:style_transfer}
	\end{figure}


	\chapterimage{pngs/time_series_analysis.png} 
	\chapter{Time Series Analysis Algogens}\index{Time Series Analysis Algogens}
	
	\section{Time Series Forecasting}\index{Time Series Forecasting}
	\subsection{Introduction to Time Series Forecasting}
	\subsubsection{The Concept of Time Series Forecasting}
	
	\paragraph{Foundational Overview}
	Time series forecasting stands as a pivotal analytical process, leveraging observed historical data to predict future values. This methodological approach is foundational across a myriad of fields such as economics, meteorology, and business planning, underscoring its critical role in informed decision-making processes. The essence of time series forecasting lies in its ability to model the temporal dynamics inherent in the data, capturing trends, seasonality, and other patterns that evolve over time.
	
	Furthermore, time series forecasting serves as a bridge between past observations and future predictions, facilitating proactive measures in various domains. Moreover, it enables organizations to anticipate market trends, manage resources efficiently, and strategize for potential challenges. Additionally, by analyzing historical data points alongside current trends, stakeholders gain insights into potential future scenarios, empowering them to make data-driven decisions. 
	
	Moreover, the importance of time series forecasting extends beyond predicting trends; it aids in risk management, where anticipating future events is crucial for mitigating potential losses. Additionally, in meteorology, accurate weather forecasts rely heavily on time series analysis, allowing for timely warnings and preparations for severe weather events. 
	
	In business, time series forecasting informs inventory management, budget planning, and sales projections, optimizing operational efficiency and enhancing profitability. Additionally, economists utilize time series models to forecast economic indicators, aiding policymakers in making informed decisions to stabilize economies and mitigate financial risks.
	
	Therefore, the significance of time series forecasting cannot be overstated, as it provides a systematic framework for understanding and predicting future outcomes based on past data, driving informed decision-making and strategic planning across diverse disciplines.

	\paragraph{Underpinning Mathematical Principles}
	At its core, the mathematical formulation of time series forecasting seeks to construct a model that can effectively map past observations to future outcomes. This is typically represented by the function \( f(X_{t-1}, X_{t-2}, \ldots, X_{t-n}) = \hat{X}_t \), where \( \hat{X}_t \) denotes the predicted value at time \( t \), and \( X_{t-1}, X_{t-2}, \ldots, X_{t-n} \) are the observed values at preceding time points. The selection of \( n \), the number of past observations considered, reflects the model's complexity and its sensitivity to the historical data's temporal structure.
	
	Moreover, the choice of \( n \) is crucial as it directly impacts the forecasting accuracy and computational complexity of the model. Increasing \( n \) allows the model to capture more intricate temporal dependencies within the data, potentially leading to more accurate predictions. However, a higher \( n \) also increases the computational burden and the risk of overfitting, where the model learns noise or idiosyncrasies in the data rather than genuine patterns. Conversely, a smaller \( n \) may result in a simpler model with lower computational costs, but it might overlook important long-term trends or seasonality present in the data.
	
	Furthermore, the mathematical underpinnings of time series forecasting often involve various statistical techniques such as autoregressive (AR), moving average (MA), or a combination of both (ARMA). These techniques leverage the temporal dependencies present in the data to make predictions. AR models capture the linear relationship between the current observation and its past values, while MA models focus on modeling the dependency between the current observation and a stochastic term based on past errors. Combining these approaches in ARMA models allows for a more comprehensive representation of the underlying processes driving the time series data.
	
	Additionally, the performance of time series forecasting models heavily relies on the assumption of stationarity, where the statistical properties of the data remain constant over time. Non-stationary data, characterized by trends or seasonality, requires preprocessing techniques such as differencing or transformation to make it stationary before applying forecasting models. Understanding the underlying mathematical principles and assumptions is essential for developing robust and accurate time series forecasting models.

	\paragraph{Strategies and Methodologies}
	Various strategies and methodologies have been developed to tackle the complexities of time series forecasting. Simple moving averages and exponential smoothing techniques are foundational methods in this realm. They prioritize recent observations, providing a quick overview of the data's behavior over time. While effective for capturing short-term trends, their simplicity limits their ability to handle intricate patterns and seasonality.
	
	Moving beyond these basic approaches, more advanced models like ARIMA (Autoregressive Integrated Moving Average) and its seasonal variant SARIMA have gained prominence. These models incorporate autoregressive and moving average components along with differencing to address trends and seasonality. By analyzing historical data and iteratively adjusting parameters, ARIMA and SARIMA can make accurate predictions while accommodating various time series characteristics.
	
	The emergence of deep learning has revolutionized time series forecasting with neural network-based methodologies such as Long Short-Term Memory (LSTM) networks. Unlike traditional methods, LSTM networks excel at capturing complex temporal dependencies and nonlinear relationships within the data. Their architecture, featuring recurrent connections and memory cells, enables them to retain information over extended time periods, making them well-suited for modeling time series data with intricate dynamics.
	
	Each of these methodologies offers distinct advantages and trade-offs. While simple techniques provide quick insights and are computationally efficient, they may lack the sophistication to capture nuanced patterns. Conversely, models like ARIMA and SARIMA offer greater flexibility and accuracy but require careful parameter tuning and may struggle with extremely noisy or irregular data. Deep learning approaches like LSTM networks excel in handling complex data structures but demand significant computational resources and data volume for training.
	
	In practice, selecting the most suitable forecasting methodology depends on factors such as data characteristics, forecasting horizon, computational resources, and the level of interpretability required for decision-making.

	\paragraph{Application Domains and Impact}
	The application domains of time series forecasting are vast and profound, influencing various sectors with their predictive capabilities. Predicting stock market trends stands as one of the primary applications, where accurate forecasts guide investment strategies, enabling investors to make informed decisions. Similarly, in the realm of meteorology, forecasting weather patterns holds paramount importance, aiding agricultural planning, disaster preparedness, and resource allocation. Moreover, within the business landscape, precise demand forecasting plays a pivotal role. By accurately predicting customer demand, companies can streamline their supply chain management processes, optimize inventory levels, and minimize operational costs, thus enhancing overall efficiency and profitability.
	
	However, the efficacy of time series forecasting hinges upon several crucial factors. Firstly, the quality and granularity of historical data significantly impact the forecasting accuracy. High-quality, relevant data sets provide a solid foundation for model training and validation. Additionally, the selection of an appropriate forecasting model is paramount. Different time series patterns necessitate different modeling techniques, ranging from classical statistical methods like ARIMA to more advanced machine learning algorithms such as recurrent neural networks (RNNs) or Long Short-Term Memory (LSTM) networks. Furthermore, the ability of forecasting models to adapt to sudden, unforeseen changes in the underlying process is crucial. Whether it's abrupt shifts in consumer behavior or unexpected market dynamics, models must possess robust mechanisms to incorporate new information and adjust predictions accordingly, ensuring their relevance and reliability in dynamic environments.

	\paragraph{Challenges and Evolution}
	The primary challenges in time series forecasting arise from data quality issues, such as missing values or noise, and the inherent difficulty of predicting the future in the presence of unexpected events or shocks. While advancements have been made in data preprocessing techniques to handle missing values and reduce noise, the unpredictable nature of real-world events poses a persistent obstacle to accurate forecasting. Yet, despite these challenges, the field continues to evolve rapidly. Ongoing research endeavors focus on enhancing model accuracy by leveraging advanced machine learning algorithms and incorporating external data sources for context. Moreover, the development of algorithms that can adapt to changing dynamics is crucial in ensuring robust forecasting performance. Large Language Models stand out among the latest innovations driving this evolution. These models, with their ability to comprehend and generate human-like text, offer promising avenues for improving forecasting accuracy by capturing subtle contextual nuances that traditional models may overlook. Furthermore, the integration of LLMs into forecasting frameworks not only enhances predictive capabilities but also opens new avenues for automated analysis and decision-making processes. Consequently, the convergence of machine learning and artificial intelligence technologies with traditional forecasting methodologies marks a significant paradigm shift in the field, paving the way for more sophisticated and adaptable forecasting systems capable of addressing the complexities of real-world data.

	\subsubsection{Key Principles and Mechanisms}
	
	\paragraph{Identifying Temporal Patterns}
	The discipline of time series forecasting is fundamentally anchored in the identification of temporal patterns within historical data. These patterns, which include trends, seasonality, and cyclic behaviors, form the bedrock upon which forecasting models are built. A trend represents a long-term increase or decrease in the data, indicating a general direction over time. Moreover, trends are pivotal in understanding the underlying dynamics of the data, serving as crucial indicators for future projections. Seasonality, on the other hand, refers to patterns that repeat at regular intervals, such as daily, monthly, or quarterly fluctuations, often driven by external factors like weather or holidays. Additionally, seasonality holds paramount importance in forecasting accuracy, as it necessitates the incorporation of seasonal adjustments to capture recurring patterns effectively. Cyclic behavior, while akin to seasonality, exhibits distinctive characteristics as it occurs over irregular, often longer periods. Furthermore, cyclic patterns can be influenced by broader economic or environmental factors, necessitating sophisticated modeling techniques to discern and incorporate such complexities. Understanding these temporal patterns is essential not only for constructing robust forecasting models but also for extracting meaningful insights from historical data to inform decision-making processes.

	\paragraph{Statistical Models in Forecasting}
	Among the plethora of tools available for time series analysis, traditional statistical models like ARIMA and its variant SARIMA stand out for their robustness and efficacy. The ARIMA model, short for Autoregressive Integrated Moving Average, combines autoregressive features, where predictions are based on past values, with moving average components, which account for lags in the error terms. This model is particularly adept at handling data where trends and non-seasonal patterns are prevalent. 
	
	Furthermore, the integration aspect of ARIMA addresses the need to stabilize non-stationary time series data, making it suitable for a wide range of forecasting tasks. Additionally, ARIMA models offer the flexibility to incorporate seasonal variations, providing a comprehensive framework for forecasting in various industries such as finance, economics, and meteorology. 
	
	Moreover, the iterative process of model identification, estimation, and validation in ARIMA modeling allows for fine-tuning the parameters to achieve optimal forecasting performance. Likewise, the diagnostic tools available for ARIMA models enable analysts to assess the adequacy of the model and identify areas for improvement. 
	
	However, it's important to note that ARIMA models have limitations, particularly in capturing complex nonlinear relationships and abrupt changes in the data. Nonetheless, by complementing ARIMA with other techniques such as machine learning algorithms, analysts can enhance forecasting accuracy and address these limitations effectively. 
	
	In summary, ARIMA and SARIMA models serve as foundational tools in time series forecasting, offering a reliable framework for analyzing and predicting various temporal phenomena with practical applications across diverse domains.

	\paragraph{Seasonal Variations with SARIMA}
	The SARIMA model extends ARIMA by incorporating seasonal elements, making it invaluable for datasets exhibiting strong periodic patterns. It adds seasonal autoregressive (AR) and moving average (MA) terms to model seasonal effects, effectively capturing the intricacies of seasonality alongside the ARIMA model's capabilities. This dual approach allows SARIMA to provide accurate forecasts for time series where both non-seasonal and seasonal patterns exist, offering a comprehensive framework for analyzing complex datasets.
	
	Furthermore, SARIMA addresses the limitations of traditional ARIMA models when dealing with data that display clear seasonal fluctuations. By introducing seasonal AR and MA components, SARIMA accounts for periodic variations that occur at fixed intervals, enabling it to better capture the underlying patterns in the data. Additionally, SARIMA facilitates the modeling of complex seasonal behaviors, such as those observed in economic indicators, weather patterns, or retail sales data.
	
	Moreover, SARIMA's ability to handle both non-seasonal and seasonal variations simultaneously enhances its utility in real-world applications. This flexibility allows analysts to explore and interpret time series data more comprehensively, leading to more accurate forecasts and informed decision-making processes. SARIMA's incorporation of seasonal elements alongside its non-seasonal components makes it a versatile tool for time series analysis across diverse domains.
	
	In conclusion, SARIMA stands as a powerful extension of the ARIMA model, specifically designed to tackle time series data characterized by seasonal fluctuations. Its integration of seasonal AR and MA terms enables it to capture complex seasonal patterns, providing analysts with a robust framework for forecasting and understanding temporal data trends.

	\paragraph{Mathematical Foundations}
	The mathematical foundation underpinning these models is fundamental to understanding their behavior and applicability in time series analysis. In the context of ARIMA (Autoregressive Integrated Moving Average) models, their formulation is expressed as ARIMA(p,d,q), where \(p\) signifies the order of autoregressive terms, \(d\) represents the degree of differencing or integration, and \(q\) denotes the order of moving average terms. This structure allows ARIMA models to capture the temporal dependencies present in a time series by incorporating lagged observations, differencing to stabilize non-stationary series, and modeling residual errors through moving average terms. 
	
	Additionally, the extension to seasonal time series analysis is facilitated through SARIMA (Seasonal Autoregressive Integrated Moving Average) models, which augment the basic ARIMA framework with seasonal parameters. SARIMA models are denoted as SARIMA(p,d,q)(P,D,Q)s, where \(P\), \(D\), and \(Q\) represent the seasonal autoregressive, differencing, and moving average orders, respectively. The inclusion of these seasonal parameters enables SARIMA models to effectively capture seasonal patterns inherent in many real-world time series data, providing a more comprehensive representation of the underlying processes. The parameter \(s\) specifies the seasonality period, allowing for the modeling of periodic fluctuations such as monthly or quarterly patterns.

	\paragraph{Mechanisms for Forecast Accuracy}
	The accuracy of time series forecasting models heavily relies on the correct identification of underlying patterns and the precise selection of model parameters. Model diagnostics, such as autocorrelation and partial autocorrelation analyses, serve as crucial tools in guiding the process of parameter selection. \textbf{Moreover}, the iterative refinement of models through validation and testing ensures that the forecasts maintain their reliability and relevance to the specific characteristics of the time series data under analysis. As time series forecasting continues to evolve, the integration of machine learning and AI technologies offers \textbf{new avenues} to enhance these traditional models. This integration promises even greater accuracy and deeper insights into future trends, thus advancing the field's capabilities \textbf{furthermore}. The incorporation of machine learning techniques allows models to adapt to changing data dynamics and capture complex relationships that may not be apparent through conventional statistical methods. This adaptability is particularly valuable in domains where data patterns evolve rapidly or exhibit nonlinear behavior. \textbf{In addition}, leveraging AI technologies enables the automation of certain aspects of the forecasting process, streamlining operations and freeing up analysts' time for more strategic tasks. By harnessing the power of machine learning and AI, forecasters can unlock new opportunities to improve forecast accuracy and make more informed decisions based on predictive insights.

	\subsubsection{The Role of Deep Learning and Traditional Models}
	
	\paragraph{Synergy Between Deep Learning and Traditional Approaches}
	The forecasting landscape is enriched by the coexistence of deep learning models, such as Recurrent Neural Networks (RNNs) and Long Short-Term Memory (LSTM) networks, alongside traditional statistical models like ARIMA and SARIMA. Deep learning models have carved a niche for themselves in the realm of time series forecasting due to their unparalleled ability to model complex, nonlinear sequences and their capacity to learn from vast amounts of data. Moreover, these models are particularly adept at capturing intricate patterns in time series data that might elude traditional statistical methods, making them a powerful tool for a wide range of forecasting tasks. However, while deep learning models excel in capturing complex patterns, they often require large amounts of data for training and can be computationally expensive. On the other hand, traditional statistical models like ARIMA and SARIMA are often more interpretable and require less computational resources, making them suitable for scenarios with limited data or computational constraints. Nevertheless, the synergy between deep learning and traditional approaches presents an opportunity for leveraging the strengths of both paradigms. By combining the interpretability and efficiency of traditional statistical models with the predictive power of deep learning models, practitioners can develop robust forecasting systems that are both accurate and interpretable. Thus, the integration of deep learning and traditional approaches represents a promising direction for advancing the field of time series forecasting.

	\paragraph{Continued Relevance of ARIMA and SARIMA}
	Despite the advancements in deep learning, traditional models such as ARIMA and SARIMA maintain their significance in the forecasting toolkit. Their enduring value stems from their simplicity, interpretability, and effectiveness, particularly in scenarios characterized by clear seasonal patterns or when the available data is limited. These models offer a structured approach to forecasting, with well-defined parameters that can be systematically identified and optimized. The transparency and ease of understanding associated with ARIMA and SARIMA models make them especially appealing for applications where model interpretability is crucial.
	
	Moreover, while deep learning techniques like neural networks often require extensive computational resources and large datasets for training, ARIMA and SARIMA models are relatively lightweight in terms of computational demand and can be trained efficiently even with smaller datasets. Additionally, the interpretability of ARIMA and SARIMA models allows analysts to gain insights into the underlying trends and patterns driving the forecast, facilitating informed decision-making processes.
	
	Furthermore, in practical applications where stakeholders require explanations for forecasting outcomes or where regulatory requirements mandate transparent models, the simplicity and interpretability of ARIMA and SARIMA models become invaluable. These models provide a clear framework for understanding how past observations influence future predictions, enabling users to validate the forecast assumptions and make adjustments as necessary.
	
	Consequently, while deep learning methods continue to push the boundaries of predictive accuracy, the enduring relevance of ARIMA and SARIMA lies in their ability to offer interpretable, reliable forecasts in a wide range of scenarios, making them indispensable tools in the forecasting practitioner's arsenal.

	\paragraph{Choosing the Right Model}
	The decision to employ deep learning models versus traditional statistical models hinges on several factors, including the complexity of the time series data, the volume of available data, and the specific requirements of the forecasting task. Deep learning models, with their flexibility and learning capabilities, are well-suited for handling large datasets and capturing complex relationships within the data. However, these models require substantial computational resources and can be challenging to interpret. On the other hand, traditional models like ARIMA and SARIMA are more straightforward to implement and can provide robust forecasts even with relatively small datasets, offering a pragmatic solution in many practical forecasting scenarios.
	
	While deep learning models offer unparalleled flexibility in capturing intricate patterns within the data, their reliance on massive computational resources can pose significant challenges, particularly for organizations with limited computing infrastructure. Conversely, traditional statistical models such as ARIMA and SARIMA may lack the sophistication of deep learning architectures but offer a computationally efficient alternative, making them more accessible to a broader range of users. Additionally, the interpretability of traditional models can be advantageous in situations where understanding the underlying factors driving the forecasts is crucial for decision-making.
	
	Moreover, the choice between deep learning and traditional statistical models should also consider the availability of historical data. Deep learning models typically thrive on extensive datasets, leveraging their capacity to extract intricate patterns and dependencies. However, in scenarios where historical data is limited, traditional models can still deliver reliable forecasts, thanks to their ability to capture temporal dependencies and seasonal variations effectively.
	
	Furthermore, the specific requirements of the forecasting task play a pivotal role in model selection. For instance, if the forecasting task involves short-term predictions with a limited amount of historical data, traditional statistical models may suffice, offering quick and interpretable solutions. Conversely, for long-term forecasting tasks with complex data patterns, deep learning models may provide superior performance, albeit at the cost of increased computational overhead and potential interpretability challenges.
	
	In conclusion, the choice between deep learning and traditional statistical models necessitates a careful evaluation of various factors, including data complexity, volume, computational resources, interpretability, and forecasting requirements. While deep learning models excel in capturing intricate relationships within large datasets, traditional models offer a pragmatic and interpretable alternative, particularly in resource-constrained environments or when historical data is limited.

	\paragraph{Complementary Strengths and Integrated Approaches}
	An emerging trend in time series forecasting is the integration of deep learning and traditional statistical models to leverage their complementary strengths. This hybrid approach can combine the deep learning models' ability to learn from data and capture complex patterns with the traditional models' effectiveness in modeling seasonality and trends. By integrating these methodologies, forecasters can create more accurate and robust forecasting systems that capitalize on the unique advantages of both deep learning and traditional models.
	
	Furthermore, this integration addresses some of the limitations inherent in standalone approaches. Deep learning models, while powerful in capturing intricate patterns, may struggle with interpreting the underlying dynamics of the time series data. Conversely, traditional statistical models often lack the flexibility to adapt to rapidly changing patterns or anomalies in the data. However, by merging these methodologies, practitioners can mitigate these weaknesses. 
	
	Moreover, the integrated approach allows for enhanced interpretability and explainability of the forecasts. Traditional statistical models provide a clear framework for understanding the relationships between variables and the impact of various factors on the forecasted outcomes. This interpretability is crucial, especially in domains where transparency and accountability are paramount.
	
	Additionally, the combined approach can improve the overall reliability and robustness of forecasts by reducing the risk of overfitting, a common challenge in deep learning models. By incorporating domain knowledge and expert insights into the modeling process, forecasters can ensure that the resulting forecasts are not only accurate but also actionable and reliable for decision-making purposes.
	
	In conclusion, the integration of deep learning and traditional statistical models offers a promising avenue for advancing the field of time series forecasting. By harnessing the complementary strengths of these methodologies, practitioners can develop more accurate, interpretable, and reliable forecasting systems that are better equipped to handle the complexities of real-world data.

	\paragraph{Future Directions in Forecasting}
	The interplay between deep learning and traditional models in time series forecasting is an area of active research and development. Traditional models, such as ARIMA and exponential smoothing techniques, have long been the cornerstone of forecasting due to their simplicity and interpretability. These models provide a clear understanding of the underlying patterns in data and are often favored when transparency is crucial, such as in financial forecasting or demand planning. Moreover, their reliance on fundamental statistical principles makes them robust and reliable even in scenarios with limited data availability or noisy signals.
	
	On the other hand, deep learning approaches, particularly recurrent neural networks (RNNs) and convolutional neural networks (CNNs), offer unparalleled flexibility and predictive power. They excel in capturing complex nonlinear relationships and can automatically extract features from raw data, eliminating the need for manual feature engineering. This makes them particularly well-suited for tasks involving high-dimensional data or unstructured inputs, such as image or text-based forecasting. Additionally, their ability to learn from vast amounts of data enables them to adapt to changing patterns and handle non-stationary time series more effectively.
	
	However, despite their promise, deep learning models often lack interpretability, which can be a significant drawback in domains where understanding the reasoning behind predictions is crucial, such as healthcare or legal contexts. Furthermore, they are computationally intensive and require large amounts of data for training, which may pose challenges in resource-constrained environments.
	
	Nevertheless, ongoing research aims to bridge the gap between traditional and deep learning approaches, leveraging the strengths of each paradigm while mitigating their respective limitations. Hybrid models that combine the interpretability of traditional methods with the predictive power of deep learning are gaining traction, offering a compromise between accuracy and transparency. Additionally, techniques such as transfer learning and ensemble methods are being explored to enhance the generalization and robustness of deep learning models, making them more suitable for real-world forecasting tasks.
	
	In the future, the choice of forecasting method is likely to be guided by a nuanced understanding of the trade-offs between different approaches, taking into account factors such as the complexity of the underlying data, the interpretability requirements, and the computational resources available. This necessitates a holistic approach to forecasting, where practitioners are equipped with a diverse toolkit of methods and techniques to tackle a wide range of forecasting challenges effectively.

	\subsubsection{Applications and Limitations}
	
	\paragraph{Diverse Applications Across Sectors}
	Time series forecasting serves as a cornerstone analytical technique with widespread applications across various sectors, reflecting its versatility and critical importance. In the financial sector, forecasting models are indispensable for predicting stock market trends, currency exchange rates, and economic indicators, aiding investors and policymakers in making informed decisions. \textbf{Moreover}, these models play a vital role in risk management strategies, allowing financial institutions to assess and mitigate potential market fluctuations and optimize investment portfolios. The meteorological domain \textbf{likewise} relies heavily on time series forecasting for weather predictions, storm tracking, and climate modeling. These forecasts are crucial for agricultural planning, enabling farmers to optimize planting schedules, irrigation, and pest management, thus enhancing crop yields and food security. \textbf{Furthermore}, accurate weather forecasts are imperative for disaster preparedness, as they enable timely evacuation plans and resource allocation in the event of natural calamities such as hurricanes, floods, or wildfires. In the realm of business and supply chain management, accurate demand forecasting \textbf{also} plays a pivotal role. By predicting consumer demand patterns, businesses can optimize inventory levels, reduce stockouts, and minimize excess inventory, leading to cost savings and improved profitability. Additionally, precise forecasts aid in planning production schedules and managing logistics, ensuring efficient operations and customer satisfaction. \textbf{Consequently}, time series forecasting contributes significantly to enhancing overall business performance and competitiveness in the global market.

	\paragraph{Limitations and Challenges}
	Despite its broad applicability and proven success, time series forecasting is not without its limitations and challenges. The unpredictability of future events stands as a primary constraint, especially in systems influenced by complex external factors or those subject to abrupt changes, such as financial markets or environmental conditions. Data quality also plays a crucial role in the accuracy of forecasts; incomplete, noisy, or biased data sets can lead to misleading predictions, underscoring the importance of robust data collection and preprocessing methods.
	
	Another significant challenge arises from the inherent assumptions underlying many forecasting models. For instance, models like ARIMA assume a level of stationarity in the data, which may not always hold true in real-world scenarios. Similarly, deep learning models require large volumes of data to train effectively, which may not be available in all contexts. The complexity and "black box" nature of some deep learning models further complicate their interpretability, making it difficult for users to understand and trust the forecasting process.
	
	\paragraph{Navigating Limitations Through Innovative Approaches}
	The field of time series forecasting continues to evolve, with ongoing research aimed at addressing these limitations. Hybrid models that combine the strengths of traditional statistical methods and deep learning are emerging as a promising approach to enhance forecast accuracy and robustness. Additionally, advancements in data collection and processing technologies are improving the quality and availability of data, mitigating some of the challenges associated with data-driven forecasting.
	
	The integration of domain-specific knowledge and contextual information, facilitated by techniques like feature engineering and the use of external data sources, is also proving valuable in overcoming the limitations of purely data-driven models. As the field progresses, the development of more adaptive, interpretable, and reliable forecasting models remains a key focus, promising to extend the applicability and effectiveness of time series forecasting across an even broader array of domains and challenges.

	\subsubsection{Algorithmic Pseudocode for Time Series Forecasting}
	The Time Series Forecasting Algorithm is a sophisticated framework tailored for accurately predicting future values in time series data. It operates by iteratively refining parameter estimates to maximize the predictive accuracy of the model. This iterative process incorporates both historical data and trend patterns to forecast future values effectively. The operational procedure of the algorithm is illustrated in pseudocode \ref{fig:time-series-forecasting-pseudocode}, highlighting its iterative nature in parameter estimation.
	
	\begin{algorithm}
		\caption{Algorithmic Time Series Forecasting Pseudocode}
		\begin{algorithmic}[1]
			\Procedure{TraditionalForecasting}{TimeSeriesData}
			\State Split TimeSeriesData into TrainingSet, TestSet
			\State Identify trend, seasonality, and autocorrelation in TrainingSet
			\State Select appropriate ARIMA model parameters (p, d, q) based on ACF and PACF plots
			\State Train ARIMA model on TrainingSet
			\For{each period in TestSet}
			\State Forecast future values using the trained ARIMA model
			\State Compare forecasted values with actual values to calculate error
			\EndFor
			\State Adjust ARIMA model parameters if necessary, based on error metrics
			\State Forecast future values beyond TestSet using the final ARIMA model
			\State \Return ForecastedValues
			\EndProcedure
		\end{algorithmic}\label{fig:time-series-forecasting-pseudocode}
	\end{algorithm}
	
\subsection{Previous Work on ML and AI Interplay with Time Series Forecasting Algorithms}

\paragraph{Machine Learning Enhancements in Time Series Analysis (2022)}
The study conducted in 2022 investigates the integration of Machine Learning (ML) techniques with traditional time series analysis and forecasting methods. It emphasizes the evolving landscape of the field, showcasing the adaptability and efficiency of ML algorithms in capturing complex patterns within time series data. By leveraging advanced ML algorithms, the study illustrates a shift towards more dynamic, data-driven approaches in forecasting. The incorporation of ML not only improves predictive accuracy but also introduces flexibility and scalability previously unattainable with conventional methods. This work represents a step forward in the evolution of time series analysis, providing a foundation for further research and application in the domain \cite{garg2022machine}.

\paragraph{Survey on Time Series Forecasting Models (2021)}
The 2021 survey evaluates time series analysis and modeling techniques, particularly focusing on forecasting capabilities enhanced by Machine Learning and Artificial Intelligence. It offers an overview of the landscape, identifying trends, methodologies, and challenges within time series forecasting. The integration of AI and ML is recognized as transformative, enabling more accurate and efficient analysis of temporal data. The survey examines various ML-driven models, highlighting their strengths and weaknesses, and providing insight into the future direction of time series forecasting. It emphasizes the need for a synergistic approach combining traditional statistical techniques with ML and AI. This research serves as a reference for understanding the impact of AI and ML on time series analysis, suggesting potential directions for future innovation \cite{dama2021time}.

\subsection{Algogenic Enhancements for Time Series Forecasting}
\subsubsection{Contextual Data Integration with Generative AI}

\paragraph{Enhancing Forecasting Models with Contextual Insights}
In enhancing time series forecasting through Algogenic methods, the specific inclusion of contextual data via Large Language Models plays a critical role. This approach enables the enrichment of traditional forecasting algorithms with nuanced insights from textual data sources, such as news feeds, social media, and economic reports. The essence of this integration lies in its capacity to augment numerical time series data with external, textually derived factors that might influence forecasting outcomes. We suggest that this methodology could potentially enhance the models' accuracy by incorporating a broader array of influencing factors beyond the historical data typically used in time series analysis.

\paragraph{Operationalizing LLMs for Contextual Analysis}
The practical implementation of LLMs for contextual data integration involves processing extensive textual data to extract relevant insights, such as market sentiment or significant events, that might impact forecasting variables. This operation hinges on LLMs' ability to interpret large volumes of unstructured data, employing natural language processing to discern sentiment, trends, and key events. In practice, this could mean analyzing financial news to identify market-moving events or social media to gauge consumer sentiment, with these insights then systematically integrated into forecasting models to refine their predictions.

\paragraph{Transforming Forecasting with Generative AI}
The application of LLMs in time series forecasting significantly shifts the paradigm from a traditionally quantitative analysis to one that also considers qualitative, contextual data. This Algogenic enhancement is designed to not only improve forecast accuracy but also to offer a more dynamic, responsive modeling approach that adapts to new information. The integration of generative AI facilitates the identification of complex patterns and dependencies that are not evident through traditional methods alone, enabling a more informed and nuanced forecast.

\subsubsection{Natural Language-Based Model Configuration}

\paragraph{Democratizing Access to Sophisticated Models}
Utilizing natural language interfaces to configure forecasting models represents a significant step towards making complex time series forecasting methodologies accessible to a broader audience. This approach, powered by LLMs, allows users to specify their forecasting needs in plain language, which the system then translates into technical configurations. This method suggests that even those without deep technical expertise in statistical modeling could effectively employ advanced forecasting techniques, potentially democratizing the use of such models.

\paragraph{Operational Mechanism and User Interaction}
The mechanism behind natural language-based configuration involves LLMs interpreting user inputs and translating these into model specifications or adjustments. For example, a user's request to forecast based on seasonal patterns is automatically mapped to the appropriate model configuration, such as selecting a SARIMA model. This process simplifies the interaction with complex forecasting tools, making them more accessible and user-friendly, and it underscores the potential for natural language processing to bridge the gap between sophisticated algorithmic models and end-user applications.

\paragraph{Enhancing Usability and Adoption}
By facilitating natural language interactions with forecasting models, this Algogenic enhancement significantly lowers the barriers to entry for utilizing advanced time series forecasting techniques. The approach not only simplifies the model configuration process but also broadens the potential user base to include non-experts, thereby encouraging wider adoption and innovative uses of time series forecasting across different domains.

\subsubsection{Adaptive Forecasting with Real-Time Data}

\paragraph{Enhancing Responsiveness to Market and Environmental Changes}
Integrating real-time data analysis into time series forecasting models, with the support of LLMs, markedly increases the models' ability to adjust forecasts in light of new information. This enhancement is particularly valuable in rapidly changing environments, where it can significantly improve the responsiveness and accuracy of forecasts. The implementation involves continuously updating the model's inputs with the latest data, thereby ensuring that the forecasting remains relevant and reflective of current conditions.

\paragraph{Mechanism of Real-Time Data Integration}
The core of real-time data integration involves LLMs processing incoming data streams to detect and incorporate relevant changes into the forecasting model. This dynamic approach allows for the continual refinement of forecasts, based on the most current data, enhancing the model's adaptability and accuracy in the face of changing market conditions or unexpected events.

\paragraph{Broadening the Scope of Forecasting Applications}
By enabling adaptive forecasting with real-time data, this Algogenic enhancement broadens the applicability of time series forecasting, making it more relevant for sectors characterized by rapid changes. The approach enhances the models' capability to provide timely, accurate forecasts across various applications, from financial markets to disaster response, thereby significantly improving decision-making processes in these contexts.

\subsubsection{Explainable Forecasts}

\paragraph{Bridging Complexity and User Understanding}
The integration of explainable features into time series forecasting, facilitated by LLMs, aims to demystify the forecasting process for users. By providing clear, understandable explanations of forecast outcomes, this enhancement makes sophisticated forecasting models more accessible and transparent, thereby fostering trust and enabling users to make more informed decisions based on the model's predictions.

\paragraph{Operationalizing Explanation Generation}
The generation of explanations within time series forecasting involves LLMs analyzing the model's workings and translating this analysis into natural language descriptions. This process aims to elucidate the reasoning behind forecasts, offering users insights into the factors influencing the predictions and enhancing the overall transparency and trustworthiness of the forecasting process.

\paragraph{Enhancing User Engagement and Model Transparency}
By making forecasts explainable, this enhancement not only improves user engagement by ensuring that the model's outputs are accessible and understandable but also promotes transparency. This approach allows users to grasp the rationale behind predictions, fostering a more informed and confident use of forecasting models in decision-making processes.

\subsubsection{Scenario Analysis and Simulation}

\paragraph{Expanding Forecasting Horizons through Simulation}
Incorporating scenario analysis and simulation into time series forecasting, facilitated by LLMs, significantly extends the forecasting capabilities by allowing the exploration of various "what-if" scenarios. This enhancement enables forecasters to simulate the potential impacts of different factors on future trends, thereby offering valuable insights for strategic planning and risk management.

\paragraph{Mechanism for Descriptive Scenario Processing}
The process of scenario analysis with LLMs involves interpreting natural language descriptions of potential scenarios and translating these into model adjustments or simulations. This capability allows forecasters to incorporate a wide range of qualitative inputs into their models, enhancing the forecasting process's adaptability and robustness in the face of uncertainty.

\paragraph{Strategic Advantages of Scenario-Based Forecasting}
Scenario-based forecasting provides strategic advantages by enabling a proactive approach to decision-making, allowing organizations to prepare for a variety of potential futures. This method enhances the flexibility and resilience of forecasting efforts, improving organizations' ability to navigate uncertainty and capitalize on opportunities.

\subsubsection{Cross-Domain Insights for Forecast Enhancement}

\paragraph{Harnessing the Power of Diverse Data Sources}
The application of LLMs to integrate cross-domain insights into time series forecasting represents a significant Algogenic enhancement. By analyzing diverse data sources, LLMs can uncover hidden relationships and trends that traditional methods might miss, thereby enriching the forecasting models with a deeper understanding of the factors influencing the forecasted outcomes.

\paragraph{Mechanism for Cross-Domain Analysis}
The mechanism for integrating cross-domain insights involves LLMs synthesizing information from various textual sources and translating these insights into factors that can be incorporated into forecasting models. This process allows for a more comprehensive analysis that captures the complex interplay of factors influencing the forecasted variables.

\paragraph{Expanding Forecasting Horizons with Cross-Domain Insights}
Incorporating cross-domain insights into time series forecasting models, facilitated by LLMs, expands the horizons of forecasting by providing a more nuanced and comprehensive view of the factors influencing future trends. This approach enhances the predictive accuracy of the models and offers strategic advantages in anticipating and responding to changes, thereby improving decision-making processes across various domains.

	\subsubsection{Pseudocode for Algogenic Time Series Forecasting}
	The Algogenic time series forecasting approach harnesses AI to enhance conventional forecasting methods by dynamically adjusting parameters and strategies based on observed data patterns and real-time error estimates. This pseudocode, available in \ref{fig:time-series-forecasting-Algogen-pseudocode}, outlines a sophisticated framework integrating AI-driven enhancements for adaptive model adjustments, data selection, prediction evaluation, and real-time parameter optimization.
	
	\begin{algorithm}
		\caption{Algogenic Time Series Forecasting Pseudocode}
		\begin{algorithmic}[1]
			\Procedure{AlgogenicTimeSeriesForecasting}{TimeSeriesData, UserInput}
			
			\Comment{Preprocessing Phase}
			\State Analyze contextual data with LLM to extract insights.
			\State Interpret UserInput with LLM for model configuration.
			\State Integrate cross-domain insights into TimeSeriesData.
			
			\Comment{Core Forecasting Phase}
			\State Initialize forecasting model based on LLM-configured parameters.
			\State Adaptively update model with real-time data and insights.
			\State Perform scenario analysis to simulate various futures.
			\State Generate forecast using the dynamically updated model.
			
			\Comment{Postprocessing Phase}
			\State Generate explainable forecasts with LLM.
			\State Provide scenario analysis results and their implications.
			\State Adjust forecasting strategy based on scenario outcomes.
			
			\EndProcedure
		\end{algorithmic}\label{fig:time-series-forecasting-Algogen-pseudocode}
	\end{algorithm}

	\begin{figure}
		\centering
		\includegraphics[width=0.58\textwidth]{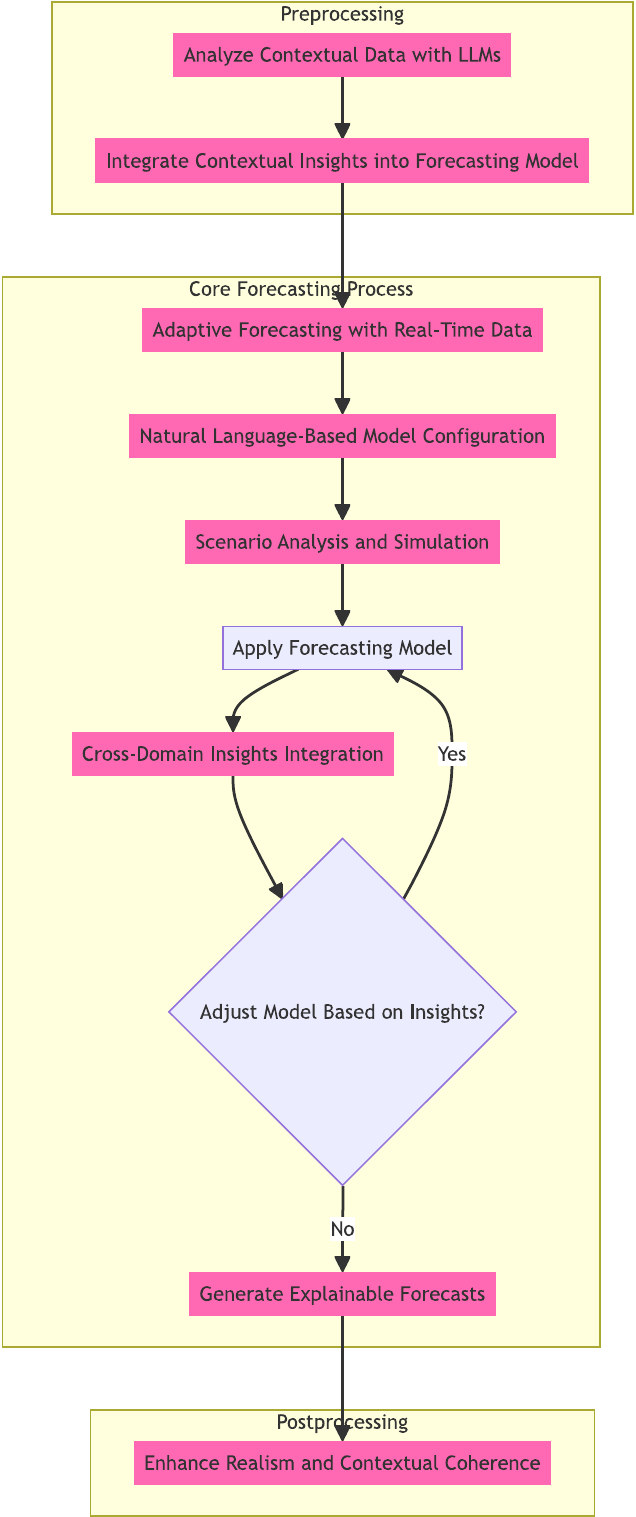}
		\caption{Innovations in Algogenic Time Series Forecasting: This visualization represents the Algogenic approach to time series forecasting, illustrating the integration of generative AI with traditional forecasting models. It highlights the use of contextual data analysis, natural language-based model configuration, adaptive forecasting with real-time data, explainable forecasts, and scenario analysis to improve forecasting accuracy and user interaction. The framework demonstrates how generative AI enhances the forecasting process by dynamically incorporating external insights, user inputs, and real-time adjustments, leading to more accurate, adaptable, and understandable forecasts.}
		\label{fig:time_series_forecasting}
	\end{figure}

	\section{Synthetic Time Series Generation}\index{Synthetic Time Series Generation}
	\subsection{Introduction to Synthetic Time Series Generation}
	\subsubsection{The Concept of Synthetic Time Series Generation}
	
	\paragraph{Defining Synthetic Time Series Generation}
	Synthetic time series generation constitutes a pivotal methodology in the realm of data science and analytics, centering on the creation of artificial data sequences. These sequences are meticulously crafted to mirror the statistical properties of real-world time series data, encompassing aspects such as trends, seasonality, and autocorrelation. The essence of this technique lies in its ability to produce data that, while not real, behaves and appears statistically similar to genuine datasets.
	
	Furthermore, synthetic time series generation plays a crucial role in various applications, including but not limited to, algorithm development, model validation, and scenario analysis. By simulating diverse temporal patterns and structures, it enables researchers and practitioners to assess the robustness and efficacy of their analytical tools under different conditions. Moreover, it facilitates the exploration of hypothetical scenarios and what-if analyses, providing insights into potential outcomes and decision-making processes.
	
	Additionally, synthetic time series generation serves as a valuable tool for education and training purposes. It allows students and professionals alike to experiment with data-driven techniques in a controlled environment, fostering a deeper understanding of time series analysis concepts and methodologies. Through hands-on experience with synthetic datasets, learners can gain practical skills in data generation, manipulation, and interpretation, enhancing their proficiency in data science and analytics.
	
	Therefore, synthetic time series generation stands as a versatile and indispensable tool in the arsenal of data scientists and analysts, offering opportunities for innovation, exploration, and learning in the dynamic landscape of time series data analysis.

	\paragraph{Utility and Applications}
	The utility of synthetic time series generation spans several critical areas. In model testing, it offers a robust framework for evaluating the performance and resilience of predictive models under a wide array of conditions, including those rarely observed in the actual data. Moreover, it provides insights into how models behave in extreme scenarios, thereby enabling developers to fine-tune algorithms and improve overall performance. For dataset augmentation, synthetic data generation addresses the challenges of limited or imbalanced data by generating additional samples, thereby enhancing the diversity and volume of data available for training machine learning models. Furthermore, synthetic data can help mitigate overfitting by introducing variability into the training set, leading to more robust and generalizable models. 
	
	In scenarios where data privacy is of paramount concern, synthetic data provides a means to share and analyze datasets without exposing sensitive information, ensuring privacy and confidentiality are maintained. This is particularly crucial in industries such as healthcare and finance, where strict regulations govern the handling of personal or proprietary data. By generating synthetic data that closely resembles the original dataset in statistical properties but does not contain any real information, organizations can comply with regulatory requirements while still leveraging the power of data-driven insights. Additionally, synthetic data can be used for educational purposes, allowing researchers and students to experiment with different datasets and algorithms without risking the integrity or privacy of real data. Overall, the versatility and utility of synthetic time series generation make it a valuable tool across various domains, offering solutions to data-related challenges while safeguarding privacy and enabling innovation.

	\paragraph{Technical Foundations and Methodologies}
	The generation of synthetic time series relies on a multifaceted array of technical methodologies, spanning from fundamental statistical approaches to sophisticated machine learning paradigms. At its essence, this endeavor necessitates a profound comprehension and emulation of the inherent statistical distributions and temporal interdependencies intrinsic to time series data. \textbf{Furthermore}, it mandates the utilization of methodologies adept at capturing the nuanced intricacies embedded within temporal sequences. Autoregressive models stand as stalwarts in this domain, leveraging past observations to forecast future states, thereby encapsulating short-term dependencies within the data. Conversely, generative adversarial networks (GANs) \textbf{and} variational autoencoders (VAEs) offer \textbf{an} alternative avenue, enabling the synthesis of time series data through adversarial training or probabilistic encoding, respectively. \textbf{Moreover}, these advanced techniques excel in encapsulating long-range dependencies and non-linear dynamics present in real-world time series. By training these models on authentic time series datasets, they \textbf{can} effectively discern underlying patterns and distributions, enabling the generation of synthetic instances that faithfully preserve the statistical characteristics of the original data. Through a judicious selection and orchestration of these methodologies, practitioners can harness the power of synthetic time series generation \textbf{to} facilitate various applications, ranging from data augmentation for machine learning tasks to stress testing financial models.

	\paragraph{Mathematical Representation}
	Mathematically, the generation of a synthetic time series can be represented as a function \( G \) that transforms a noise vector \( \mathbf{z} \), drawn from a known distribution, into a synthetic time series \( \mathbf{x'} \) that approximates the distribution of the real time series \( \mathbf{x} \). This can be expressed as:
	\[
	\mathbf{x'} = G(\mathbf{z}), \quad \text{where} \quad \mathbf{z} \sim P_z.
	\]
	Here, \( P_z \) denotes the distribution of the noise vector, and the goal of \( G \) is to capture the complex temporal dynamics and statistical properties of \( \mathbf{x} \), thereby generating \( \mathbf{x'} \) that is statistically indistinguishable from \( \mathbf{x} \).
	
	\paragraph{Challenges and Advancements}
	Despite the promise of synthetic time series generation, it is fraught with challenges. The fidelity of synthetic data to real-world data is a paramount concern, particularly when confronted with intricate temporal dependencies and rare occurrences. Such complexities pose a formidable obstacle, demanding innovative solutions for achieving verisimilitude in synthetic datasets. However, recent strides in machine learning offer a glimmer of hope amidst these obstacles. The infusion of Algogenic enhancements, coupled with the integration of Large Language Models, represents a paradigm shift in the landscape of synthetic data generation. These advancements not only bolster the fidelity of synthetic data but also facilitate the preservation of semantic coherence, thereby addressing longstanding concerns regarding the meaningfulness of generated data. By leveraging LLMs, practitioners can imbue synthetic time series with contextual richness, ensuring that generated sequences adhere to realistic patterns and distributions. Moreover, the adoption of scenario-based generation techniques further augments the utility of synthetic data, enabling the creation of bespoke datasets tailored to specific use cases. Consequently, these developments propel synthetic time series generation beyond mere replication, ushering in an era of nuanced, privacy-preserving data synthesis. In doing so, they pave the way for a plethora of applications, spanning domains as diverse as finance, healthcare, and environmental science, where the availability of high-fidelity synthetic data is indispensable for driving innovation and discovery.

	\subsubsection{Key Principles and Mechanisms}
	
	\paragraph{Foundation of Synthetic Generation}
	The creation of synthetic time series data is firmly rooted in the comprehensive analysis and understanding of the statistical characteristics intrinsic to the original dataset. This foundational step involves meticulously examining the statistical distributions, correlations, and recurrent patterns that define the time series, ensuring that the synthetic data faithfully reflects these aspects. 
	
	Moreover, the process necessitates a deep understanding of the underlying mechanisms driving the observed phenomena. While statistical analysis provides crucial insights, an in-depth comprehension of the domain-specific knowledge is indispensable. Therefore, domain expertise plays a pivotal role in guiding the generation process, allowing for the incorporation of contextual nuances and domain-specific constraints into the synthetic data.
	
	Additionally, the synthesis of time series data demands a judicious selection of appropriate generative models and techniques. Various methodologies such as autoregressive models, recurrent neural networks (RNNs), and generative adversarial networks (GANs) offer distinct advantages and trade-offs, depending on the characteristics of the original data and the intended application. Careful consideration of these factors is paramount to ensure the fidelity and relevance of the synthetic data.
	
	Furthermore, the generation process must account for temporal dependencies and dynamics inherent in the time series. Techniques such as temporal convolutional networks (TCNs) and long short-term memory (LSTM) networks are adept at capturing sequential patterns and temporal correlations, facilitating the generation of realistic synthetic sequences that preserve the temporal structure of the original data.
	
	In conclusion, the foundation of synthetic generation rests upon a multifaceted approach encompassing statistical analysis, domain expertise, selection of appropriate generative models, and consideration of temporal dynamics. By intricately weaving together these elements, synthetic time series data can faithfully emulate the statistical characteristics and temporal behaviors of real-world datasets, thereby serving as invaluable assets for a myriad of applications in diverse domains.

	\paragraph{Spectrum of Methodologies}
	The methodologies employed in synthetic time series generation span a broad spectrum, from relatively straightforward statistical models to sophisticated machine learning techniques. Early approaches might rely on autoregressive (AR) models or moving averages (MA) to simulate data based on observed trends and seasonality. However, the advent of machine learning and, more recently, deep learning, has introduced more complex approaches capable of modeling intricate temporal dependencies. Generative adversarial networks (GANs) and variational autoencoders (VAEs) exemplify this shift, offering powerful tools for learning and generating data that mimics the multi-dimensional distributions of time series datasets.
	Furthermore, in recent years, researchers have explored hybrid methodologies that combine traditional statistical techniques with modern machine learning algorithms to capitalize on the strengths of both. Moreover, the integration of domain knowledge and expert insights into the modeling process has become increasingly prevalent, enhancing the interpretability and robustness of generated time series data. Additionally, advancements in computational resources have enabled the application of computationally intensive methods, such as deep learning architectures, to larger and more complex datasets, further expanding the horizons of synthetic time series generation. Consequently, the field continues to evolve, driven by a diverse array of methodologies and interdisciplinary collaborations aimed at addressing emerging challenges and pushing the boundaries of what is possible in generating realistic time series data.

	\paragraph{Capturing Temporal Dynamics}
	At the heart of synthetic time series generation lies the formidable challenge of accurately capturing temporal dynamics and dependencies within the data. It transcends mere replication of statistical distributions; it entails a profound comprehension of how values at one temporal juncture interlace with those at another. Herein, techniques like Long Short-Term Memory (LSTM) networks manifest their prowess. LSTM, a variant of recurrent neural networks (RNN), stands out for its intrinsic capacity to retain information over extended temporal spans. This attribute renders LSTM networks eminently suitable for the intricate task of generating time series data, wherein antecedent values wield substantial influence over subsequent ones. These networks, by virtue of their recurrent architecture, excel at grasping the sequential patterns inherent in time series data, thereby facilitating the creation of synthetic time series that faithfully capture the nuanced temporal dependencies present in the original data. Thus, LSTM networks emerge as indispensable tools in the realm of synthetic time series generation, bridging the chasm between raw data and meticulously crafted synthetic counterparts.

	\paragraph{Algorithmic Implementation}
	Mathematically, the generation process often relies on defining a model \( G \) that, given a set of parameters \( \theta \) learned from the real data \( X \), generates new data \( X' \) such that the statistical properties of \( X' \) closely match those of \( X \). The learning process can be represented as:
	\[
	\theta^* = \arg\min_{\theta} D(F(X), F(G(X'; \theta))),
	\]
	where \( D \) is a measure of divergence between the distribution of the real data \( F(X) \) and the distribution of the synthetic data \( F(G(X'; \theta)) \), with the goal of minimizing this divergence to ensure fidelity to the original data's statistical properties.
	
	\paragraph{Advancements and Evolution}
	The field of synthetic time series generation continues to evolve, driven by advancements in computational techniques and theoretical understanding. The integration of deep learning models has significantly expanded the potential to generate high-fidelity synthetic data. Furthermore, recent developments in Algogenic enhancements, particularly the use of Large Language Models for enriching synthetic data with contextual and domain-specific insights, represent a promising frontier. These advancements not only improve the quality of synthetic data but also broaden its applicability across diverse domains, from finance to healthcare, where accurate and realistic synthetic time series can provide invaluable benefits.
	
	\textbf{Deep Learning Advancements:} The incorporation of deep learning architectures such as recurrent neural networks (RNNs) and generative adversarial networks (GANs) has revolutionized synthetic data generation. These models excel at capturing intricate patterns and dependencies within temporal data, enabling the creation of time series that closely mimic real-world dynamics.
	
	\textbf{Enhanced Contextual Understanding:} Large Language Models like GPT-3 have emerged as powerful tools for contextual understanding and natural language processing. Leveraging these models, researchers can infuse synthetic time series with nuanced contextual information, ensuring that generated data reflects the complexities and subtleties observed in real-world scenarios.
	
	\textbf{Cross-Domain Applicability:} The impact of these advancements transcends traditional boundaries, extending into various sectors including finance, healthcare, and climate science. In finance, synthetic time series facilitate risk assessment, portfolio optimization, and algorithmic trading strategies. Similarly, in healthcare, they aid in predictive modeling, drug discovery, and personalized treatment planning. The versatility of synthetic time series generated through advanced techniques underscores their relevance and significance across diverse domains.
	
	\textbf{Future Prospects:} As research in synthetic data generation continues to advance, further breakthroughs are anticipated. The exploration of novel architectures, refinement of training methodologies, and integration of multi-modal data sources hold promise for even more realistic and utility-driven synthetic time series. These developments are poised to catalyze innovation across industries, shaping the future landscape of data-driven decision-making and computational modeling.

	\subsubsection{The Role of Deep Learning}
	
	\paragraph{Revolutionizing Synthetic Data Generation}
	Deep learning has markedly transformed the landscape of synthetic data generation, introducing a paradigm shift in how artificial data sequences are produced. Particularly, generative adversarial networks (GANs) and variational autoencoders (VAEs) stand at the forefront of this revolution. These models harness the potent capabilities of deep learning architectures to learn and replicate the intricate, multi-dimensional distributions characteristic of real-world data. Their advent has not only expanded the horizons of synthetic data generation but also significantly enhanced the fidelity and utility of the generated datasets.
	
	Moreover, GANs and VAEs have demonstrated exceptional adaptability across various domains, from computer vision to natural language processing. Their ability to capture and model complex data distributions enables them to generate synthetic data that closely mimics real-world scenarios, offering invaluable benefits in scenarios where access to large, diverse datasets is limited or impractical.
	
	Furthermore, the advancements in synthetic data generation facilitated by GANs and VAEs have profound implications for applications such as data augmentation, domain adaptation, and privacy-preserving data sharing. By producing synthetic data that preserves the statistical characteristics of the original dataset, these models enable researchers and practitioners to augment training data, mitigate domain shift, and anonymize sensitive information without compromising the utility of the data.
	
	Additionally, the synergy between deep learning and synthetic data generation has sparked interdisciplinary collaborations, fostering innovations in fields ranging from healthcare and finance to robotics and autonomous systems. This convergence of expertise not only accelerates the pace of research and development but also cultivates novel solutions to longstanding challenges, ultimately driving transformative advancements across industries.
	
	In summary, the fusion of deep learning methodologies with synthetic data generation techniques, particularly through GANs and VAEs, has revolutionized the way artificial data sequences are created. This transformative shift not only broadens the applicability of synthetic data but also paves the way for groundbreaking discoveries and innovations across diverse domains.

	\paragraph{Generative Adversarial Networks (GANs)}
	Generative Adversarial Networks (GANs) represent a paradigm shift in the realm of artificial intelligence and machine learning. These networks, comprising two key components – a generator and a discriminator, revolutionize synthetic data generation. The crux of GANs' functionality lies in their adversarial training mechanism. The generator, analogous to a skilled counterfeiter, fabricates data samples intended to be indistinguishable from genuine instances. Conversely, the discriminator, akin to a vigilant detective, scrutinizes these samples, discerning real from synthetic. This adversarial interplay creates a dynamic feedback loop, propelling both networks towards refinement. Iteration by iteration, the generator hones its craft, crafting increasingly realistic data, while the discriminator evolves to become more discerning. Consequently, GANs possess an inherent capacity to capture the intricate nuances of complex data distributions. Their versatility extends to various domains, with applications ranging from image synthesis to natural language processing. Particularly noteworthy is their adeptness in synthesizing time series data. By mimicking the statistical intricacies of real-world datasets, GANs furnish researchers and practitioners with invaluable tools for data augmentation, anomaly detection, and predictive modeling. The symbiotic relationship between the generator and discriminator fosters a continual pursuit of realism, rendering GANs indispensable in contemporary machine learning endeavors.

	\paragraph{Variational Autoencoders (VAEs)}
	Variational Autoencoders (VAEs) represent a pivotal advancement in synthetic data generation, bridging the realms of probabilistic graphical models and deep learning. They operate by encoding input data into a latent space, where the latent variables capture meaningful features of the data distribution. Unlike traditional autoencoders, VAEs enforce a probabilistic interpretation, where the encoder network learns a probability distribution over the latent space. This probabilistic formulation enables VAEs to sample from the latent space, facilitating the generation of new instances that closely mimic the original data distribution.
	
	While traditional autoencoders map the input directly to a fixed encoding, VAEs introduce a stochastic element by parameterizing the latent space with mean and variance vectors, enabling the generation of diverse outputs for a given input. This stochasticity encourages the model to learn a more structured and continuous latent space representation, fostering better generalization and interpolation capabilities.
	
	Moreover, VAEs adopt a two-fold objective during training: reconstruction loss and a regularization term, typically the Kullback-Leibler (KL) divergence. The reconstruction loss ensures that the generated samples faithfully reconstruct the input data, preserving its statistical properties. Simultaneously, the KL divergence regularizes the latent space distribution, encouraging it to approximate a predefined prior distribution, such as a standard Gaussian. This dual optimization strategy enables VAEs to learn a disentangled representation of the input data, where each dimension of the latent space corresponds to a meaningful attribute or feature.
	
	Furthermore, VAEs offer inherent scalability and flexibility, making them suitable for various applications, including the generation of synthetic time series data. By leveraging the compact latent space representation, VAEs can capture the temporal dependencies and underlying patterns present in the original time series, facilitating the generation of realistic and diverse synthetic samples.
	
	In addition to data generation, VAEs find utility in tasks such as denoising, anomaly detection, and semi-supervised learning, underscoring their versatility and effectiveness across diverse domains. Consequently, VAEs have emerged as a cornerstone in the arsenal of generative modeling techniques, empowering researchers and practitioners to tackle complex data generation challenges with unparalleled efficacy and flexibility.

	\paragraph{Impact on Synthetic Time Series Generation}
	The role of deep learning, particularly through GANs and VAEs, in synthetic time series generation is profound. These models excel in capturing temporal dependencies and patterns within time series data, a task that traditional statistical methods may find challenging. By learning the sequential structure and variability inherent in time series, deep learning models can generate synthetic sequences that not only statistically resemble the original data but also exhibit realistic temporal dynamics.
	
	Furthermore, deep learning models offer a remarkable advantage over conventional statistical approaches by implicitly capturing intricate non-linear relationships and complex dependencies present in time series data. Unlike traditional methods, which often rely on predefined assumptions or parametric models, deep learning techniques like GANs and VAEs can adaptively learn from data, allowing for more flexible and adaptive modeling of temporal dynamics.
	
	Moreover, the ability of deep learning models to incorporate contextual information and high-dimensional representations enhances their capacity to generate diverse and realistic synthetic time series across various domains. This capability is particularly crucial in applications where the generated data must accurately reflect the underlying complexities and uncertainties inherent in real-world time series.
	
	Additionally, the generative nature of deep learning models enables the exploration of latent spaces, facilitating the generation of novel time series data that extends beyond the boundaries of the observed data distribution. This property opens up avenues for generating synthetic data for scenarios where limited or incomplete data are available, thereby supporting tasks such as data augmentation, anomaly detection, and scenario analysis in time series forecasting.
	
	Consequently, the integration of deep learning techniques in synthetic time series generation not only advances the state-of-the-art in data-driven modeling but also fosters innovation in various domains reliant on accurate and realistic synthetic data.

	\paragraph{Future Directions and Challenges}
	Despite the significant advancements brought about by deep learning in synthetic data generation, challenges remain. Ensuring the generated data's diversity, managing model complexity, and addressing training stability, especially with GANs, are ongoing areas of research. Moreover, the interpretability of deep learning models and the ethical considerations in generating synthetic data require careful attention. As the field progresses, continued innovation in deep learning methodologies and the integration of Algogenic enhancements, including the application of Large Language Models for improved contextual relevance and domain-specificity, are expected to further refine and expand the capabilities of synthetic time series generation. \par
	Additionally, the fusion of diverse data modalities presents both opportunities and complexities. Whereas traditional approaches often focus on single-modal data, such as images or text, emerging research explores the synergy of multiple modalities, such as combining textual and visual information. Inasmuch as this convergence facilitates richer data representations, it also introduces challenges in harmonizing disparate data formats and ensuring coherence across modalities. However, such integrative efforts promise enhanced data realism and a more comprehensive understanding of complex real-world phenomena. Furthermore, leveraging domain knowledge and incorporating expert insights can bolster the fidelity of synthetic data, ensuring its relevance and utility across diverse applications. Hence, interdisciplinary collaborations between domain experts and machine learning practitioners are crucial for advancing the frontier of synthetic data generation.

	\subsubsection{Applications and Limitations}
	
	\paragraph{Expansive Applications of Synthetic Time Series}
	The generation of synthetic time series data finds its utility in a broad spectrum of applications, each leveraging the unique capability to mimic real-world data while offering flexibility and scalability. Data augmentation stands as a primary application, where synthetic time series are utilized to enhance machine learning models by providing additional training data, particularly in scenarios where real data is scarce, imbalanced, or exhibits rare events. This augmentation is crucial in improving model robustness and generalization \textbf{Moreover}, as synthetic data can be generated to represent various scenarios and outliers that might not be present in the original dataset, thereby enriching the learning process and ensuring better performance in diverse real-world situations. Financial modeling also benefits significantly from synthetic time series, enabling the simulation of market conditions, stress testing, and scenario analysis without relying on sensitive or proprietary financial data. Additionally, privacy-preserving data sharing emerges as a critical application, where synthetic data allows for the analysis and sharing of datasets that resemble real statistical properties without compromising individual privacy or violating data protection regulations \textbf{Furthermore}. \textbf{In addition}, synthetic time series can be tailored to match specific statistical properties, ensuring that the shared data retains its integrity and utility for analysis while safeguarding the privacy of individuals. These applications underscore the versatility and importance of synthetic time series in various domains, offering solutions to challenges related to data scarcity, privacy concerns, and model generalization.

	\paragraph{Navigating the Challenges}
	Despite the wide-ranging applications, the generation of synthetic time series data is not without its limitations. One of the foremost challenges lies in accurately capturing the complexity and nuances of the original data. This includes replicating the intricate temporal dependencies, seasonal patterns, and potential non-linearities present in real-world time series. Ensuring the usefulness and relevance of the generated data is another challenge, as synthetic data must be sufficiently realistic and representative to be valuable for training models or making informed decisions. Moreover, privacy compliance represents a further limitation, particularly in ensuring that synthetic data does not inadvertently reveal sensitive information about individuals or entities represented in the original dataset. Achieving a balance between data utility and privacy is a complex task that requires careful consideration of the generation process and the application of sophisticated techniques to mitigate re-identification risks. Furthermore, there is the challenge of maintaining the diversity and variability of the generated data, as overly simplistic models may fail to capture the full range of behaviors present in the real-world data. Thus, constant refinement and validation of the synthetic data generation process are necessary to address these challenges effectively.

	\paragraph{Future Directions and Ethical Considerations}
	As the field of synthetic time series generation continues to evolve, addressing these limitations remains a focal point of research and development. Advancements in deep learning, particularly in refining generative models like GANs and VAEs, offer promising avenues for enhancing the realism and applicability of synthetic data. Moreover, the integration of Algogenic enhancements, such as the application of Large Language Models for context-aware generation and semantic consistency checks, is poised to further bridge the gap between synthetic and real data. Ethical considerations, especially regarding privacy and the potential for misuse of synthetic data, underscore the importance of developing transparent, responsible practices for data generation. As applications of synthetic time series expand, ongoing innovation and ethical vigilance will be crucial in harnessing the full potential of this technology while navigating its inherent challenges.

	\subsubsection{Pseudocode for Algorithmic Synthetic Time Series Generation}
	The Synthetic Time Series Generation Algorithm is a robust framework crafted for generating realistic time series data efficiently. It excels in simulating time series with complex patterns and dependencies. This algorithm iteratively refines its parameter estimates by maximizing the likelihood function using both observed data and internal model dynamics. The procedural essence of this algorithm is captured in the pseudocode provided in Figure \ref{fig:time-series-gen-pseudocode}, demonstrating its iterative nature in generating synthetic time series.
	
	\begin{algorithm}
		\caption{Algorithmic Pseudocode for Synthetic Time Series Generation}
		\begin{algorithmic}[1]
			\Procedure{GenerateSyntheticTimeSeries}{OriginalTimeSeries}
			\State Analyze statistical properties of OriginalTimeSeries
			\State Determine model type (e.g., ARIMA, GAN, VAE) based on data characteristics
			\State Initialize model parameters
			\State Train model on OriginalTimeSeries to learn data distribution
			\While{not reached desired level of statistical similarity}
			\State Generate SyntheticSeries from trained model
			\State Compare statistical properties of SyntheticSeries with OriginalTimeSeries
			\State Adjust model parameters to improve similarity
			\EndWhile
			\State \Return SyntheticSeries
			\EndProcedure
		\end{algorithmic}\label{fig:time-series-gen-pseudocode}
	\end{algorithm}

\subsection{Previous Work on ML and AI Interplay with Synthetic Time Series Generation Algorithms}

\paragraph{Towards Generating Real-World Time Series Data}
The 2021 IEEE International Conference on Data Mining presented advancements in generating real-world time series data using machine learning and artificial intelligence techniques \cite{pei2021towards}. This work focused on developing data generation models capable of producing time series data resembling real-world phenomena. By employing advanced algorithms and deep learning frameworks, the study addressed challenges in time series data generation, such as capturing complex temporal dependencies and variability. The proposed models showed the ability to generate synthetic time series data with statistical properties similar to genuine datasets, aiding analyses in various domains such as finance, healthcare, and environmental science. The improved capacity for generating realistic time series datasets provides opportunities for testing and enhancing machine learning models, addressing the scarcity of real-world datasets.

\paragraph{Machine Learning Algorithms for Time Series Analysis and Forecasting}
In 2022, a review on machine learning algorithms for time series analysis and forecasting was published, covering methodologies and applications across sectors \cite{garg2022machine}. This study highlighted the evolution and impact of machine learning techniques in improving the accuracy and efficiency of time series forecasting. It emphasized the importance of selecting suitable algorithms based on data characteristics and forecasting objectives, ranging from traditional statistical methods to deep learning approaches. The review critically analyzed various machine learning algorithms, discussing their strengths, limitations, and applicability to different types of time series data. It also stressed the significance of preprocessing and feature engineering in enhancing model performance, showcasing how AI techniques can reveal complex patterns and trends. The insights from this review are beneficial for practitioners and researchers interested in employing machine learning for predictive analytics, offering guidance for navigating time series analysis and forecasting complexities. The convergence of machine learning and time series data analysis enables a nuanced understanding of temporal data, improving predictive capabilities.

\subsection{Algogenic Enhancements for Synthetic Time Series Generation}

\subsubsection{Semantic Consistency Checking}

\paragraph{Ensuring Domain-Specific Realism}
We suggest incorporating Large Language Models for enhancing the realism of synthetic time series data through Semantic Consistency Checking. This method involves the critical evaluation of synthetic time series data by LLMs to ensure that the generated sequences not only statistically resemble the original dataset but also conform to domain-specific knowledge and realistic scenarios. For example, in the context of financial time series generation, LLMs could assess if synthetic data mirrors realistic economic behaviors and market conditions. This process includes examining trends and volatilities against established economic theories and market behaviors, thereby enhancing the domain-specific realism of synthetic financial datasets.

Moreover, LLMs' deep textual analysis capability allows for a nuanced understanding of various domains, aiding in the generation of synthetic data that reflects real-world complexities. This is particularly beneficial in sectors like healthcare, where synthetic patient data must be consistent with medical standards and physiological realities, thus improving the utility and applicability of synthetic datasets across different industries.

\paragraph{Operational Mechanism}
The operation of Semantic Consistency Checking involves LLMs analyzing synthetic time series data against extensive domain-specific knowledge bases to detect and rectify inconsistencies. This process might include comparing the synthetic data's trends, patterns, and anomalies against a vast corpus of textual information, ensuring that the generated data is not only statistically coherent but also semantically aligned with real-world contexts. The use of advanced natural language processing techniques allows LLMs to refine their evaluation, enhancing the semantic consistency and applicability of synthetic datasets for practical uses.

\paragraph{Advancing Synthetic Data Generation}
Semantic Consistency Checking is posited as a modest yet impactful advancement in synthetic data generation, prioritizing the contextual and semantic alignment of synthetic datasets with their intended application domains. This cautious approach ensures that synthetic time series data is both statistically representative and semantically rich, addressing the practical needs of various industries for reliable and applicable synthetic data. The emphasis on domain-specific realism and practical applicability underscores the potential of this method to improve synthetic data generation processes significantly.

\subsubsection{Scenario-Based Generation}

\paragraph{Tailoring Synthetic Data to Specified Conditions}
We propose utilizing LLMs for Scenario-Based Generation, where synthetic time series data is crafted to reflect specific, descriptive scenarios provided by users. This method enables the generation of data that not only aligns with statistical patterns found in historical data but also incorporates hypothetical or future conditions articulated through natural language descriptions. This capability is especially useful for stress-testing and strategic planning, allowing for a nuanced exploration of potential future events and their impacts on data-driven models.

\paragraph{Mechanism for Interpreting and Implementing Scenarios}
The mechanism behind Scenario-Based Generation involves LLMs interpreting natural language descriptions of desired scenarios and translating these descriptions into quantitative data generation parameters. This process ensures that the generated synthetic data accurately reflects the specified conditions, enhancing the utility of synthetic datasets for predictive modeling and decision-making processes.

\paragraph{Enhancing Predictive Modeling and Risk Management}
Scenario-Based Generation enhances predictive modeling and risk management by enabling the generation of synthetic data tailored to specific scenarios. This approach allows stakeholders to explore the potential impacts of various future events, improving the robustness and resilience of decision-making processes. The focus on generating scenario-specific synthetic data underscores its value in strategic planning and risk assessment activities across industries.

\subsubsection{Data Augmentation for Rare Events}

\paragraph{Addressing Data Imbalance and Enhancing Model Robustness}
We suggest the augmentation of datasets with synthetic instances of rare events using LLMs to address data imbalances and enhance the robustness of predictive models. This strategy is aimed at enriching datasets with underrepresented events, thus providing a more balanced foundation for model training and improving the predictive accuracy of models in capturing rare but significant occurrences.

\paragraph{Identifying and Generating Rare Event Data}
The process involves LLMs identifying gaps in datasets regarding rare events and generating synthetic data to fill these gaps. This targeted data augmentation ensures that the synthetic datasets are both comprehensive and realistic, enhancing the models' ability to predict rare events accurately.

\paragraph{Broadening the Impact of Synthetic Data}
Augmenting datasets with synthetic instances of rare events broadens the applicability and impact of synthetic data generation, improving the performance of predictive models on rare occurrences. This approach not only addresses the challenge of data scarcity but also contributes to the development of more resilient and accurate data-driven models across various domains.

\subsubsection{Automated Feature Engineering for Synthetic Data}

\paragraph{Elevating Data Quality with Intelligent Feature Creation}
We advocate for the use of LLMs in the automated generation and selection of features for synthetic datasets, known as Automated Feature Engineering for Synthetic Data. This enhancement aims to dynamically identify and incorporate relevant features into synthetic datasets, thereby enhancing the datasets' statistical and semantic quality. The process leverages LLMs to analyze data and generate features that capture essential patterns and relationships, improving the utility of synthetic datasets for complex model training and analysis.

\paragraph{Operationalizing Intelligent Feature Selection}
The process of operationalizing intelligent feature selection involves LLMs analyzing datasets to identify key features and relationships critical for understanding the data's underlying dynamics. By automating the feature selection and generation process, LLMs can enhance the relevance and quality of synthetic datasets, supporting more accurate and efficient data-driven decision-making.

\paragraph{Advancing Predictive Modeling and Analysis}
Automated Feature Engineering for Synthetic Data advances predictive modeling and analysis by enriching synthetic datasets with intelligently selected features. This enhancement supports the development of more sophisticated models by providing datasets that more accurately reflect complex real-world phenomena, thereby improving predictive accuracy and analytical insights.

\subsubsection{Natural Language to Time Series Conversion}

\paragraph{Bridging Linguistic Descriptions and Quantitative Data}
We propose leveraging LLMs to convert natural language descriptions into quantitative time series data, facilitating a more intuitive and accessible approach to data generation. This method allows users to specify desired data characteristics in natural language, which LLMs then interpret to generate synthetic time series data that meets these specifications. This approach democratizes data generation, making it more accessible to a wider audience and enhancing interdisciplinary collaboration.

\paragraph{Operationalizing Descriptive Inputs for Data Generation}
The operationalization of descriptive inputs for data generation involves LLMs interpreting natural language inputs and translating them into specific data generation parameters. This process ensures that the generated synthetic time series data accurately reflects the user's described scenarios and conditions, enhancing the relevance and utility of synthetic datasets for various analytical and decision-making purposes.

\paragraph{Enhancing Accessibility and Creativity in Data Generation}
The conversion of natural language descriptions to time series data enhances the accessibility and creativity of synthetic data generation. By enabling users to specify data generation criteria in natural language, this approach supports a more user-friendly and flexible data generation process, encouraging innovation and exploration in data-driven research and applications.

\subsubsection{Ethical and Privacy Considerations}

\paragraph{Safeguarding Data Integrity and Privacy}
We emphasize the importance of using LLMs to assess synthetic datasets for compliance with ethical standards and privacy regulations. This approach ensures that synthetic data generation processes are responsible and do not compromise individuals' privacy or perpetuate biases. By integrating ethical and privacy considerations into the data generation process, we can maintain the integrity and trustworthiness of synthetic datasets, supporting their responsible use in various applications.

\paragraph{Operational Mechanisms for Ethical Compliance}
The operational mechanisms for ensuring ethical compliance involve LLMs evaluating synthetic datasets against established ethical and privacy standards. This process includes identifying potential biases, ensuring data anonymization, and assessing compliance with legal frameworks. By embedding ethical considerations into the data generation process, we can enhance the ethical integrity of synthetic datasets and promote their responsible use.

\paragraph{Promoting Responsible Synthetic Data Use}
Promoting responsible use of synthetic data involves integrating ethical and privacy considerations into the data generation process. By employing LLMs to assess and ensure the ethical integrity of synthetic datasets, we can foster trust and accountability in synthetic data generation, supporting its application in a manner that respects privacy rights and ethical norms.

\subsubsection{Interactive Generation with Feedback Loops}

\paragraph{Empowering Users with Interactive Data Customization}
We propose the development of interactive data generation processes with feedback loops, enabling users to iteratively refine synthetic datasets based on their feedback. This approach leverages LLMs to adjust data generation parameters in response to user input, enhancing the relevance and accuracy of synthetic datasets. By fostering a collaborative and iterative data generation process, we can ensure that the generated data meets the specific needs and expectations of users.

\paragraph{Mechanism for Feedback-Driven Refinement}
The mechanism for feedback-driven refinement involves an iterative process where LLMs adjust data generation based on user feedback. This process enables the continuous improvement of synthetic datasets, ensuring that they accurately reflect users' requirements and enhance the overall quality and relevance of the generated data.

\paragraph{Enhancing Data Generation Outcomes and User Experience}
Interactive generation with feedback loops enhances the outcomes of data generation processes and the user experience by making data generation more responsive to user needs. This approach supports the creation of synthetic datasets that are closely aligned with specific research or operational goals, improving the utility and effectiveness of synthetic data for various applications.

	\subsubsection{Pseudocode for Algogenic Synthetic Time Series Generation}
	The Algogenic synthetic time series generation approach harnesses AI to enhance traditional methods by dynamically adjusting generation parameters and strategies based on the observed behavior of the system and real-time error estimates. This pseudocode, available in \ref{fig:synthetic-time-series-Algogen-pseudocode}, outlines an advanced framework incorporating AI-driven enhancements for adaptive pattern generation, data point selection, validation criteria, and real-time parameter optimization.
	
	\begin{algorithm}
		\caption{Algogenic Synthetic Time Series Generation Pseudocode}
		\begin{algorithmic}[1]
			\Procedure{GenerateSyntheticTimeSeries}{InputData, UserScenarios}
			
			\Comment{Preprocessing Phase}
			\State $SemanticInsights \gets AnalyzeContextWithLLM(InputData)$
			\State $ScenarioAdjustments \gets InterpretScenarios(UserScenarios)$
			\State $DataAugmentation \gets IdentifyRareEvents(InputData)$
			
			\Comment{Core Generation Process}
			\For{each $Segment$ in $InputData$}
			\State $SyntheticSegment \gets GenerateDataSegment(Segment, SemanticInsights)$
			\State $SyntheticSegment \gets ApplyScenarioAdjustments(SyntheticSegment, ScenarioAdjustments)$
			\State $SyntheticSegment \gets AugmentDataForRareEvents(SyntheticSegment, DataAugmentation)$
			\State $FeatureEngineering \gets AutoFeatureEngineering(SyntheticSegment)$
			\State $SyntheticData \gets SyntheticData + SyntheticSegment$
			\EndFor
			
			\Comment{Postprocessing Phase}
			\State $SyntheticData \gets ValidateSemanticConsistency(SyntheticData, SemanticInsights)$
			\State $SyntheticData \gets ApplyPrivacyAndEthicalChecks(SyntheticData)$
			\State $Feedback \gets CollectUserFeedback(SyntheticData)$
			\While{$Feedback$ indicates adjustments}
			\State $SyntheticData \gets RefineBasedOnFeedback(SyntheticData, Feedback)$
			\State $Feedback \gets CollectUserFeedback(SyntheticData)$
			\EndWhile
			
			\State \Return $SyntheticData$
			\EndProcedure
		\end{algorithmic}\label{fig:synthetic-time-series-Algogen-pseudocode}
	\end{algorithm}

	\begin{figure}
		\centering
		\includegraphics[width=0.56\textwidth]{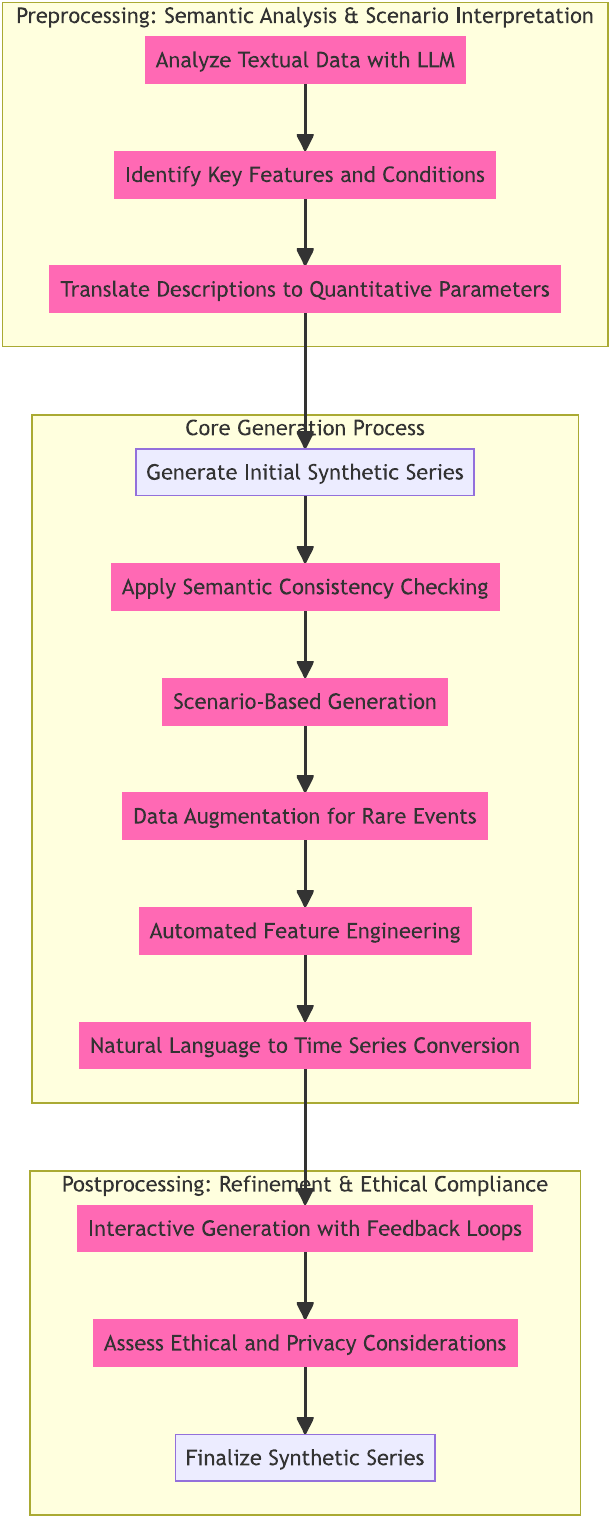}
		\caption{Integrating AI in Synthetic Time Series Generation: This diagram showcases the Algogenic enhancements applied to synthetic time series generation, emphasizing the role of Large Language Models in enriching the process. Key enhancements include semantic consistency checking to ensure domain-specific realism, scenario-based generation for tailored synthetic data, data augmentation for rare events to improve model robustness, and ethical and privacy considerations to safeguard data integrity. The framework highlights the dynamic interaction between AI-driven insights and traditional data generation techniques, facilitating the production of synthetic time series data that is not only statistically representative but also contextually accurate and ethically sound.}
		\label{fig:synthetic_time_series_generation}
	\end{figure}

	
	\chapterimage{pngs/classical_algorithms.png} 
	
	\chapter{Other Classical Algogens}\index{Classical Algogens}
	
	\section{QuickSort}\index{QuickSort}
	\subsection{Introduction to QuickSort}
	\subsubsection{The Concept of QuickSort}
	
	\paragraph{Fundamental Overview}
	QuickSort, a cornerstone in the realm of sorting algorithms, operates on the divide-and-conquer principle, strategically breaking down a problem into smaller, more manageable parts to achieve efficiency and speed in data organization. Developed by Tony Hoare in 1960, QuickSort has become one of the most widely used sorting methods due to its superior average-case performance and its simplicity in implementation.
	
	Furthermore, QuickSort's efficiency lies in its ability to pivot on elements within the data set, swiftly partitioning them into segments while recursively sorting each partition. This approach contrasts with simpler sorting algorithms like bubble sort or insertion sort, which often exhibit poorer performance on large datasets due to their less sophisticated methodologies. 
	
	Moreover, unlike other divide-and-conquer algorithms, QuickSort's partitioning step doesn't require extra space, making it more memory-efficient. Additionally, its adaptability to different data distributions contributes to its versatility, as it performs well on both randomly ordered and partially sorted data. 
	
	However, despite its numerous advantages, QuickSort does have limitations. In the worst-case scenario, particularly with already sorted or nearly sorted data, it can degrade to quadratic time complexity, although such instances are rare in practice. Nonetheless, researchers have developed various optimizations, such as randomized pivot selection and hybrid approaches like Introsort, to mitigate these drawbacks, ensuring QuickSort remains a powerful tool in the sorting algorithm arsenal.

	\paragraph{Operational Mechanics}
	The essence of QuickSort lies in its selection of a 'pivot' element from the array to be sorted. This pivotal choice determines the efficiency and performance of the sorting algorithm. \textbf{Moreover}, once a pivot is chosen, QuickSort dynamically partitions the array into two subsets: elements less than the pivot and elements greater than the pivot. This \textbf{dynamic partitioning} process ensures that at each step, the problem size is effectively reduced, paving the way for a swift sorting process. QuickSort then \textbf{recursively} applies the same strategy to the subarrays formed by the partition, \textbf{consequently} facilitating a divide-and-conquer approach. This recursive nature of QuickSort, while seemingly straightforward, leads to a highly efficient sorting algorithm with an average-case time complexity of $O(n \log n)$. The elegance of QuickSort's efficiency lies in its ability to intelligently divide the problem space, tackling smaller subproblems iteratively until the entire array is sorted. The \textbf{recursive partitioning} mechanism not only contributes to its efficiency but also makes it adaptable to various data distributions and sizes. Thus, QuickSort's operational mechanics underscore a balance between simplicity and effectiveness, rendering it a staple in sorting algorithms.

	\paragraph{Pivot Selection Strategies}
	The choice of pivot selection strategy in QuickSort profoundly influences its efficiency and performance characteristics. Traditional methods, such as selecting the first or last element of the array, are straightforward but can lead to suboptimal outcomes in certain scenarios. For instance, choosing the first element may result in poor performance when the input array is already sorted or nearly sorted, as it can cause the algorithm to exhibit its worst-case time complexity of $O(n^2)$ due to highly unbalanced partitions.
	
	Conversely, selecting the last element as the pivot may mitigate some of the issues encountered with the first element strategy but still leaves the algorithm susceptible to poor performance with certain input distributions. Randomized pivot selection, where a pivot is chosen at random from the array, offers a degree of unpredictability that can improve average-case performance, particularly for uniformly distributed data. However, this approach does not guarantee optimal partition balance and may still encounter worst-case scenarios.
	
	An alternative strategy, known as the 'median-of-three', aims to address the shortcomings of deterministic and purely random pivot selection methods. By selecting the median value among the first, middle, and last elements of the array as the pivot, this approach seeks to achieve better partition balance and, consequently, more efficient sorting. This strategy tends to perform well across a wide range of input distributions, mitigating the risk of worst-case time complexity while maintaining good average-case performance.
	
	Additionally, advanced techniques such as hybrid approaches combining deterministic and randomized strategies have been proposed to further enhance QuickSort's robustness and adaptability to different types of input data. These methods intelligently select pivots based on the characteristics of the input array, aiming to achieve a balance between determinism and randomness to optimize sorting performance.
	
	In summary, the selection of pivot strategy in QuickSort is a crucial determinant of its efficiency and ability to handle various input scenarios. While traditional methods offer simplicity, more sophisticated approaches such as the median-of-three strategy and hybrid techniques provide avenues for improving sorting performance across a wide range of input distributions.

	\paragraph{Algorithmic Efficiency}
	The theoretical efficiency of QuickSort shines in its average-case scenario, where the choice of pivot leads to balanced partitions. This characteristic allows QuickSort to achieve a time complexity of $O(n \log n)$ on average, making it one of the fastest sorting algorithms. However, in its worst-case scenario, particularly when the smallest or largest element is consistently chosen as the pivot, the time complexity degrades to $O(n^2)$. This issue arises due to unbalanced partitions, where one partition contains significantly more elements than the other, leading to inefficient sorting. Despite this drawback, practical implementations often incorporate various strategies to mitigate the impact of worst-case scenarios. For instance, hybrid models combine QuickSort with other sorting algorithms like Insertion Sort. This hybrid approach leverages the strengths of both algorithms, utilizing QuickSort's efficiency for larger partitions while resorting to Insertion Sort for smaller arrays. By doing so, the overhead associated with the worst-case scenario is minimized, enhancing the overall performance of the sorting process. These optimizations underscore the importance of considering practical implications alongside theoretical analyses when evaluating algorithmic efficiency.

	\paragraph{Significance and Applications}
	QuickSort's significance extends beyond its algorithmic elegance to practical applications across software development, database management, and data analysis. Its in-place sorting capability, requiring minimal additional memory allocation, makes it particularly suitable for large datasets. Furthermore, QuickSort's adaptability to parallel processing allows it to remain relevant and efficient in the age of multi-core and distributed computing, solidifying its position as a versatile and powerful tool in the algorithmic toolbox.
	
	Moreover, QuickSort's efficiency is notable in scenarios where computational resources are limited. Unlike some sorting algorithms that demand significant memory overhead or incur high time complexity, QuickSort's divide-and-conquer strategy enables it to achieve impressive speed and performance. Consequently, it is widely employed in real-time systems, where quick response times are critical.
	
	Additionally, the simplicity of QuickSort's implementation contributes to its widespread adoption. Its recursive nature and intuitive partitioning steps make it accessible to developers across different skill levels. As a result, QuickSort finds applications not only in high-performance computing environments but also in educational settings, serving as a pedagogical tool to teach fundamental concepts of algorithm design and analysis.
	
	In summary, QuickSort's blend of efficiency, adaptability, and simplicity renders it indispensable in various domains, ranging from handling big data in enterprise applications to serving as a teaching aid in computer science education.

	\subsubsection{Key Principles and Mechanisms}
	
	\paragraph{Divide and Conquer Strategy}
	At the heart of QuickSort's methodology is the divide and conquer strategy, a powerful algorithmic approach that solves a problem by breaking it down into smaller sub-problems, solving each sub-problem independently, and then combining their solutions to solve the original problem. QuickSort applies this strategy by dividing the array into two partitions based on a pivot element, ensuring that elements in the left partition are less than the pivot, while those in the right are greater. This division process is recursive, with QuickSort continually applied to smaller and smaller partitions until the entire array is sorted.
	
	Moreover, the effectiveness of the divide and conquer strategy lies in its ability to optimize both time and space complexities. By breaking the problem into smaller, more manageable parts, QuickSort reduces the overall time complexity to an average-case scenario of $O(n log n)$. This efficiency is crucial for large datasets, where traditional sorting methods like insertion sort or bubble sort become impractical due to their higher time complexities. Additionally, the recursive nature of the divide and conquer approach minimizes the need for additional memory allocation, making QuickSort more memory-efficient compared to other sorting algorithms.
	
	Furthermore, the choice of the pivot element significantly influences the performance of QuickSort. While a poorly chosen pivot can lead to inefficient partitioning and degrade the algorithm's time complexity to $O(n^2)$, a well-selected pivot ensures balanced partitions, maintaining the average-case time complexity. Consequently, various strategies for pivot selection have been devised, such as selecting the first or last element, median-of-three method, or random selection. Each strategy aims to mitigate the risk of worst-case scenarios and improve the overall efficiency of QuickSort.
	
	In addition to its effectiveness in sorting arrays, the divide and conquer strategy employed by QuickSort is applicable to a wide range of computational problems beyond sorting. Tasks such as searching, data compression, and numerical computations often benefit from the divide and conquer paradigm, highlighting the versatility and importance of this algorithmic approach in computer science and beyond.

	\paragraph{Partitioning Process}
	The partitioning process, fundamental to QuickSort's efficiency, entails reorganizing the array to position all elements with values less than the pivot before it, while those with greater values follow. The selection of the pivot profoundly impacts the effectiveness of this process, as an ideal pivot divides the array into two nearly equal partitions. Various strategies exist for selecting the pivot, such as choosing the first element, the last element, the middle element, or employing a median-of-three approach to mitigate the risk of worst-case scenarios. Regardless of the pivot selection strategy, the essence of the partitioning mechanism remains consistent across implementations.
	
	Typically, the partitioning process involves traversing the array, comparing each element to the pivot, and swapping elements as necessary to ensure they reside on the correct side of the pivot. This iterative process effectively organizes the array around the pivot point, contributing to the overall efficiency of the QuickSort algorithm. However, it's important to note that the efficiency of the partitioning process is contingent not only on the choice of pivot but also on the underlying implementation details, such as the method used for element swapping and boundary conditions.
	
	Moreover, the partitioning process is not only crucial for the efficiency of QuickSort but also lays the groundwork for the subsequent recursive sorting steps. By partitioning the array into smaller segments based on the pivot, QuickSort efficiently sorts each partition independently, eventually achieving a fully sorted array. Thus, the partitioning process serves as a pivotal step in the QuickSort algorithm, showcasing its elegance and effectiveness in sorting large datasets efficiently.

	\paragraph{Recursive Sorting}
	Following partitioning, QuickSort recursively applies the same logic to the subarrays formed on either side of the pivot. This recursion continues until the base case is reached, usually when a subarray contains fewer than two elements, implying that it is already sorted. The recursive nature of QuickSort allows it to efficiently break down the sorting problem into manageable pieces, applying the divide and conquer strategy in a manner that significantly reduces the overall number of comparisons and swaps needed to sort the array.
	
	Moreover, the recursive approach enables QuickSort to exploit the principle of locality, as it focuses on sorting smaller subarrays before moving on to larger ones. This localized sorting contributes to better cache performance, especially in situations where data access patterns exhibit spatial locality. Additionally, the recursive nature of QuickSort lends itself well to parallelization, as independent subarrays can be sorted concurrently, leading to potential performance gains on multi-core architectures.
	
	Furthermore, the elegance of the divide and conquer strategy in QuickSort lies in its simplicity and effectiveness. By repeatedly partitioning the array into smaller subarrays and sorting them independently, QuickSort achieves a time complexity of O(n log n) on average, making it one of the fastest sorting algorithms in practice. Furthermore, its in-place partitioning scheme minimizes the need for extra memory, making it memory efficient as well.
	
	Consequently, QuickSort stands out as a versatile and efficient sorting algorithm suitable for a wide range of applications, from small-scale sorting tasks to large-scale data processing scenarios. Its recursive nature, coupled with its divide and conquer approach, allows it to handle diverse data sets efficiently, making it a popular choice among programmers and software engineers.

	\paragraph{In-place Sorting}
	QuickSort is characterized by its in-place sorting capability, meaning it requires only a small, constant amount of additional storage space. This efficiency makes QuickSort particularly appealing for sorting large datasets where memory usage is a concern. By performing all sorting operations within the original array and using only a small stack to keep track of the subarray boundaries, QuickSort minimizes its memory footprint, distinguishing it from other sorting algorithms that may require significant additional space.
	
	Furthermore, the in-place nature of QuickSort contributes to its speed and practicality. Unlike algorithms that rely on auxiliary data structures like heaps or linked lists for sorting, QuickSort operates directly on the input array, eliminating the overhead associated with managing separate data structures. Moreover, the ability to sort in-place enhances cache locality, as the algorithm accesses contiguous memory locations, reducing the number of cache misses and improving overall performance.
	
	Additionally, the simplicity of the in-place approach facilitates implementation and reduces the complexity of the algorithm. Since QuickSort modifies the input array itself, there is no need to allocate additional memory or manage external data structures, simplifying the code and making it easier to understand and maintain.
	
	In contrast to sorting methods that allocate extra memory for temporary storage or auxiliary data structures, QuickSort's in-place strategy optimizes memory usage and computational efficiency, making it a versatile choice for a wide range of applications.

	\paragraph{Algorithmic Complexity and Performance}
	The performance of QuickSort is most often represented by its average-case time complexity of $O(n \log n)$, making it one of the fastest sorting algorithms for average scenarios. \textbf{However}, its performance can degrade to $O(n^2)$ \textbf{in the worst case}, particularly when partitioning results in one very small and one very large subarray. Various strategies, such as randomizing pivot selection \textbf{or} using median-of-three, can mitigate this risk, ensuring that QuickSort remains efficient across a wide range of input data scenarios. These key principles and mechanisms underpin QuickSort's robustness, versatility, and enduring popularity in computational applications. \textbf{Moreover}, QuickSort's simplicity and ease of implementation contribute to its widespread adoption and integration into various programming languages and frameworks. \textbf{Additionally}, its in-place partitioning and recursive divide-and-conquer strategy make it particularly suitable for applications with limited memory resources, enhancing its practical utility in real-world scenarios.

	\subsubsection{The Role of Pivot Selection}
	
	\paragraph{Pivotal to Performance}
	The role of pivot selection in QuickSort cannot be overstated, as it is the crux upon which the efficiency and effectiveness of the algorithm hinge. Selecting an optimal pivot is crucial for achieving balanced partitions, which in turn minimizes the depth of recursion required to sort the array. An ideal pivot would split the array into two equal parts, ensuring that the recursive calls operate on increasingly smaller subsets of the array, thus optimizing the sorting process.
	
	Furthermore, an ill-chosen pivot can lead to skewed partitions, causing QuickSort to degrade into a less efficient algorithm, approaching its worst-case time complexity of \(O(n^2)\). Thus, the pivotal decision of pivot selection significantly influences the algorithm's runtime behavior and sorting performance. Moreover, the choice of pivot also impacts the stability of the sorting process, as an unbalanced partition may result in unnecessary swaps and comparisons, potentially introducing errors in the sorted output. Consequently, the careful consideration of pivot selection is indispensable for harnessing the full potential of QuickSort and ensuring its reliable and swift operation in various sorting scenarios.

	\paragraph{Strategies for Pivot Selection}
	There are several strategies for selecting a pivot in QuickSort, each with its own advantages and potential drawbacks. The simplest method is to choose the first or last element of the array as the pivot. This approach is straightforward to implement and requires no additional computation. However, it can lead to poor performance on already sorted or nearly sorted data since it tends to create highly unbalanced partitions, resulting in suboptimal sorting times. 
	
	Another strategy involves selecting a random element as the pivot. This method aims to introduce randomness into the pivot selection process, which can help avoid predictable patterns in the input data that might otherwise lead to worst-case time complexity scenarios. By choosing a pivot randomly from the array, the algorithm becomes less susceptible to deliberate manipulation by adversaries seeking to exploit its weaknesses.
	
	The median-of-three strategy offers a balanced compromise between simplicity and effectiveness. It selects the pivot as the median of the first, middle, and last elements of the array. This approach addresses the potential pitfalls of always selecting the first or last element while avoiding the computational overhead associated with fully sorting the array to find the true median. By using this strategy, QuickSort can achieve more evenly distributed partitions, reducing the likelihood of encountering scenarios where one partition dominates the sorting process, thereby improving overall performance.
	
	Each pivot selection strategy comes with its trade-offs, and the choice of method depends on the specific characteristics of the input data and the desired performance goals. While the median-of-three strategy offers a good balance between simplicity and effectiveness, it may still struggle with certain types of input data, such as those with many duplicate elements or extreme outliers. Therefore, it's essential to consider the nature of the data and the expected workload when selecting the most appropriate pivot selection strategy for QuickSort.

	\paragraph{Impact on Algorithmic Complexity}
	The choice of pivot significantly affects QuickSort's time complexity. In the best-case scenario, where each pivot selection results in perfectly balanced partitions, the algorithm achieves a time complexity of $O(n \log n)$. This scenario occurs when the pivot divides the array into two nearly equal halves, minimizing the number of comparisons needed to sort each partition. Conversely, in the worst-case scenario, such as when the smallest or largest element is consistently chosen as the pivot, the time complexity degrades to $O(n^2)$. In this situation, the partitions created are highly imbalanced, leading to inefficient sorting as one partition becomes significantly larger than the other, requiring more comparisons to sort. Despite this worst-case scenario, QuickSort's average time complexity remains at $O(n \log n)$, making it efficient for most practical cases. However, the variability in performance based on pivot selection underscores the importance of employing a robust strategy that adapts to the dataset's characteristics. Various strategies exist to mitigate this variability, such as randomized pivot selection or using median-of-three pivot selection, which selects the median value among the first, middle, and last elements. These strategies aim to achieve better partition balance and, consequently, improve overall performance. Therefore, understanding the impact of pivot selection on QuickSort's time complexity is crucial for implementing efficient sorting algorithms in real-world applications.

	\paragraph{Adaptive and Hybrid Approaches}
	To enhance QuickSort's performance across diverse datasets, adaptive and hybrid pivot selection approaches have been developed. These methods analyze the dataset to dynamically choose the most appropriate pivot selection strategy. This adaptive strategy allows QuickSort to adjust its behavior based on the characteristics of the input data, ensuring efficient sorting regardless of varying patterns or distributions within the dataset. \textbf{Moreover}, these approaches often incorporate machine learning algorithms or heuristics informed by the data's properties. By leveraging machine learning techniques, QuickSort can learn from past sorting experiences and adapt its pivot selection strategy accordingly, leading to improved performance on similar datasets in the future. \textbf{Furthermore}, hybrid approaches may combine QuickSort with other sorting algorithms, such as Insertion Sort for small arrays. This combination capitalizes on the strengths of each algorithm, \textbf{thus} improving overall performance. \textbf{On the other hand}, hybrid strategies also aim to mitigate the weaknesses inherent in individual algorithms. For instance, while QuickSort excels in average-case scenarios, it may degrade to quadratic time complexity in worst-case scenarios. By incorporating Insertion Sort for smaller partitions, the hybrid approach mitigates this risk and ensures more consistent performance across different dataset characteristics.

	\paragraph{Conclusion}
	In summary, pivot selection stands as a pivotal determinant within the QuickSort algorithm, shaping its operational efficiency and overall performance. The choice of pivot significantly influences the partitioning process, directly impacting the number of comparisons and swaps executed during sorting. Moreover, it plays a critical role in mitigating the worst-case scenario, where QuickSort's performance can degrade to quadratic time complexity due to poorly chosen pivots. Consequently, the ongoing pursuit of sophisticated pivot selection strategies underscores the quest for enhancing QuickSort's efficacy. Adaptive techniques, which dynamically adjust pivot selection based on the input data's characteristics, offer promise in optimizing performance across diverse datasets. Likewise, hybrid approaches that combine multiple pivot selection methods aim to leverage the strengths of each strategy, potentially yielding superior sorting outcomes. These research endeavors epitomize the relentless pursuit of algorithmic refinement, striving to unlock QuickSort's full potential and sustain its relevance in addressing sorting challenges across various computational domains. Through continuous innovation and refinement, QuickSort maintains its stature as a cornerstone sorting algorithm, adeptly balancing efficiency and versatility in tackling diverse sorting tasks.

	\subsubsection{Applications and Limitations}
	
	\paragraph{Wide-ranging Applications}
	QuickSort's efficiency, especially in its average-case performance, has made it a preferred sorting algorithm in various applications. It is extensively used in systems where time complexity is critical, such as in database algorithms for sorting large datasets, in search algorithms where sorted data can significantly speed up query response times, and in embedded systems where memory efficiency is paramount. The algorithm's in-place sorting capability, requiring minimal additional memory, makes it particularly suitable for high-volume data processing tasks. Moreover, QuickSort's versatility and efficiency have led to its adoption in programming libraries and frameworks, serving as the underlying mechanism for data sorting functions in languages like C (qsort in stdlib.h), Java (Arrays.sort for primitive types), and Python (used in the Timsort algorithm as a fallback method).
	
	In database systems, where sorting large datasets is a common operation, QuickSort's average-case time complexity of O(n log n) ensures efficient data retrieval and manipulation, contributing to overall system performance. Similarly, in search algorithms, QuickSort's ability to quickly organize data enhances the speed of query responses, crucial in applications ranging from web search engines to information retrieval systems. In embedded systems with limited memory resources, QuickSort's in-place sorting minimizes memory overhead, allowing efficient processing of large datasets without excessive memory consumption. Furthermore, its integration into popular programming languages facilitates developers in implementing sorting functionalities without reinventing the wheel, contributing to code efficiency and maintainability across diverse software projects.

	\paragraph{Limitations and Considerations}
	Despite its widespread use and performance advantages, QuickSort is not without limitations. Its worst-case time complexity of $O(n^2)$, although rare in practice with good pivot selection strategies, can be a concern for datasets that might trigger these scenarios, such as nearly sorted arrays or arrays with many duplicate values. This variability in performance necessitates careful consideration of the data characteristics and potentially the use of hybrid sorting strategies in critical applications. Moreover, the recursive nature of QuickSort can lead to stack overflow errors in environments with limited stack size, especially for very large datasets unless tail recursion optimizations are employed.
	
	However, despite these limitations, QuickSort remains a popular choice due to its average-case time complexity of $O(n \log n)$ and efficient performance on a wide range of datasets. Additionally, its simplicity in implementation and low memory overhead make it an attractive option for many applications. Furthermore, advancements in pivot selection techniques, such as the median-of-three method, have mitigated the occurrence of worst-case scenarios, further enhancing its practicality. Consequently, while understanding its limitations is crucial, QuickSort's benefits often outweigh its drawbacks in real-world scenarios.

	\paragraph{Mitigation Strategies}
	To mitigate its limitations, various optimizations and strategies have been developed. Implementing a hybrid approach that switches to a different sorting algorithm, like Insertion Sort, for small partitions can avoid the degradation of performance in nearly sorted datasets and reduce the overhead of recursion. \textbf{Moreover}, the choice of a pivot selection strategy, as previously discussed, is crucial in minimizing the likelihood of encountering the worst-case performance. \textbf{Additionally}, parallel implementations of QuickSort have been explored to leverage multi-core processors, significantly reducing sorting times for large datasets by distributing the workload across multiple processing units. \textbf{Furthermore}, the incorporation of advanced data structures such as balanced trees or heaps can enhance the efficiency of QuickSort by providing better pivot selection mechanisms or reducing the complexity of certain operations. \textbf{On the other hand}, dynamic programming techniques can be employed to optimize subproblems encountered during the sorting process, effectively reducing redundant computations and improving overall time complexity. \textbf{In contrast}, while QuickSort excels in average-case performance, alternative sorting algorithms like Merge Sort or Heap Sort may be more suitable for datasets with predictable patterns or where worst-case scenarios are frequent, offering a more consistent runtime. \textbf{Consequently}, a comprehensive understanding of the dataset characteristics and computational environment is essential in selecting and tailoring mitigation strategies to maximize sorting efficiency.

	\paragraph{Mitigation Strategies}
	To mitigate its limitations, various optimizations and strategies have been developed. Implementing a hybrid approach that switches to a different sorting algorithm, like Insertion Sort, for small partitions can avoid the degradation of performance in nearly sorted datasets and reduce the overhead of recursion. \textbf{Moreover}, the choice of a pivot selection strategy, as previously discussed, is crucial in minimizing the likelihood of encountering the worst-case performance. \textbf{Additionally}, parallel implementations of QuickSort have been explored to leverage multi-core processors, significantly reducing sorting times for large datasets by distributing the workload across multiple processing units. \textbf{Furthermore}, the incorporation of advanced data structures such as balanced trees or heaps can enhance the efficiency of QuickSort by providing better pivot selection mechanisms or reducing the complexity of certain operations. \textbf{On the other hand}, dynamic programming techniques can be employed to optimize subproblems encountered during the sorting process, effectively reducing redundant computations and improving overall time complexity. \textbf{In contrast}, while QuickSort excels in average-case performance, alternative sorting algorithms like Merge Sort or Heap Sort may be more suitable for datasets with predictable patterns or where worst-case scenarios are frequent, offering a more consistent runtime. \textbf{Consequently}, a comprehensive understanding of the dataset characteristics and computational environment is essential in selecting and tailoring mitigation strategies to maximize sorting efficiency.

	\paragraph{Future Directions}
	As computational needs evolve and datasets grow in size and complexity, the continued refinement of QuickSort and the development of new strategies to enhance its performance and reliability remain areas of active research and interest. QuickSort, a fundamental algorithm in computer science, has demonstrated remarkable efficiency in sorting large datasets, yet its performance can degrade under certain conditions, such as when the dataset is nearly sorted or contains a high number of duplicates. Therefore, exploring alternative partitioning strategies or incorporating adaptive mechanisms into QuickSort could mitigate these challenges and improve its robustness across a wider range of scenarios.
	
	Furthermore, the integration of machine learning models holds significant promise for advancing sorting algorithms like QuickSort. By leveraging historical performance data and dataset characteristics, machine learning algorithms can effectively learn patterns and relationships to predict the most suitable sorting strategy for a given dataset. This predictive capability can lead to more efficient sorting operations, especially for datasets with varying properties. Additionally, the exploration of hybrid approaches that combine traditional sorting algorithms with machine learning-based decision-making can further enhance sorting performance and adaptability.
	
	Moreover, research efforts can also focus on optimizing QuickSort for specific application domains or hardware architectures. Tailoring the algorithm's implementation to leverage parallel processing capabilities, cache hierarchies, or specialized hardware accelerators can lead to substantial performance gains in real-world scenarios. Additionally, investigating the impact of emerging technologies such as quantum computing on sorting algorithms could uncover novel opportunities for accelerating sorting operations on future computing platforms.
	
	In summary, the future directions of QuickSort and sorting algorithms, in general, encompass a multidimensional approach that involves refining traditional techniques, leveraging machine learning advancements, tailoring implementations to specific contexts, and exploring the implications of evolving computing technologies. By addressing these challenges and opportunities, researchers can continue to push the boundaries of sorting efficiency and scalability, enabling more effective data processing in diverse computational environments.

	\subsubsection{Algorithmic Pseudocode for QuickSort}
	The QuickSort Algorithm is a powerful sorting technique renowned for its efficiency in organizing data. Similar to other divide-and-conquer algorithms, QuickSort operates by selecting a 'pivot' element from the array and partitioning the remaining elements based on whether they are less than or greater than the pivot. This process is facilitated by the \texttt{Partition} function, which plays a crucial role in determining the final position of the pivot and arranging elements accordingly. By recursively applying this partitioning step to sub-arrays on either side of the pivot, QuickSort achieves optimal sorting performance. Its in-place sorting mechanism and recursive approach contribute to an average-case time complexity of $O(n \log n)$, making it particularly well-suited for handling large datasets. The operational details of QuickSort are succinctly depicted in the accompanying pseudocode (see Figure \ref{fig:quicksort-pseudocode}).
	
	\begin{algorithm}
		\caption{QuickSort Algorithm Pseudocode}
		\begin{algorithmic}[1]
			\Procedure{QuickSort}{Array, Low, High}
			\If{Low < High}
			\State $PivotIndex \gets$ Partition(Array, Low, High)
			\State Call \Call{QuickSort}{Array, Low, PivotIndex - 1}
			\State Call \Call{QuickSort}{Array, PivotIndex + 1, High}
			\EndIf
			\EndProcedure
			
			\Function{Partition}{Array, Low, High}
			\State $Pivot \gets$ Array[High]
			\State $i \gets$ Low - 1
			\For{$j \gets$ Low to High-1}
			\If{Array[$j$] $\leq$ Pivot}
			\State $i \gets$ $i + 1$
			\State Swap Array[$i$] with Array[$j$]
			\EndIf
			\EndFor
			\State Swap Array[$i + 1$] with Array[High]
			\State \Return $i + 1$
			\EndFunction
		\end{algorithmic}\label{fig:quicksort-pseudocode}
	\end{algorithm}
	
\subsection{Previous Work on ML and AI Interplay with Sorting Algorithms}
Efforts to optimize computational algorithms have recently seen advancements through the application of machine learning techniques, notably deep reinforcement learning (DRL) as discussed in \cite{mankowitz2023faster}. In a study published in Nature, researchers introduced an approach utilizing DRL to explore faster sorting algorithms. This study represents an exploration at the intersection of artificial intelligence and algorithmic design, illustrating AI's potential not only to comprehend and implement existing algorithms but also to refine and enhance them. Sorting, being a fundamental operation in computer science with implications spanning from database management to search engine functionality, underscores the significance of this research. Employing a sophisticated DRL model, researchers iteratively evaluated and improved sorting strategies, resulting in the identification of algorithms exhibiting superior efficiency compared to traditional methods. This accomplishment emphasizes the transformative nature of amalgamating AI with algorithmic research, offering new avenues for computational process optimization. Utilizing DRL for algorithmic innovation signifies a shift in problem-solving approaches within computer science, emphasizing the potential synergy between AI methodologies and classical algorithmic frameworks. Beyond sorting, the implications of this research suggest a methodology applicable to various algorithmic challenges, thus establishing a benchmark for future investigations in the field.

	\subsection{Algogenic Enhancements for QuickSort}
	\subsubsection{Adaptive Pivot Selection}
	
	\paragraph{Specificity in Pivot Strategy} The effectiveness of QuickSort pivots significantly on the method used for pivot selection, directly impacting its performance efficiency. Adaptive Pivot Selection, as a specific Algogenic enhancement, employs Large Language Models to discern and predict the most suitable pivot selection strategy based on the dataset's unique characteristics. This method, by leveraging the analytical depth of LLMs, sifts through extensive historical data to identify patterns that could inform a strategic pivot choice, thus aiming to optimize QuickSort's efficiency. The potential of LLMs to dynamically adjust to varying datasets underscores the adaptability of this approach, incorporating a broad spectrum of factors such as data distribution and outliers. However, the practical implementation of such a strategy demands careful consideration of computational resources and potential impacts on algorithm complexity.
	
	\paragraph{Dynamic Adjustment for Data Specifics} This enhancement's essence lies in its capacity for real-time adjustment of pivot strategy, informed by ongoing analysis of the dataset's evolving characteristics. The LLMS guide QuickSort to adapt its pivot selection dynamically, a capability crucial for averting inefficient sorting scenarios. This adjustment is predicated on recognizing dataset patterns, such as uniform distribution or the presence of outliers, to tailor pivot selection accordingly. While promising, the implementation of dynamic adjustments necessitates a sophisticated understanding of LLMs' operational dynamics and the computational overhead involved.
	
	\paragraph{Integration and Algorithmic Execution} Implementing this Algogenic enhancement entails incorporating LLM-driven recommendations into QuickSort's partition function and possibly its recursive logic. This integration aims to refine the sorting process based on LLM insights, enhancing QuickSort's adaptability and efficiency. However, the practicality of these modifications requires an evaluation of their impact on the algorithm's complexity and runtime performance. The dynamic nature of this approach, while beneficial, introduces additional layers of computational demand, underscoring the need for a balanced assessment of its feasibility.
	
	\paragraph{Impact and Potential for Efficiency} The introduction of Adaptive Pivot Selection aims to bolster QuickSort's performance, pushing it towards optimal efficiency across varied datasets. This enhancement reflects the broader potential of Algogenic approaches to refine traditional algorithms, suggesting a pathway to more intelligent and responsive computational solutions. Nonetheless, the optimistic view of its impact should be tempered with caution, acknowledging the challenges in practical implementation and the incremental nature of potential improvements.
	
	\subsubsection{Data Distribution Analysis for Optimal Partitioning}
	
	\paragraph{Insight-Driven Partitioning Enhancement} QuickSort's efficiency is intricately linked to its partitioning efficacy. By leveraging LLMs for Data Distribution Analysis, this enhancement seeks to identify optimal partitioning strategies that yield balanced partitions, directly influencing QuickSort's performance. The integration of data insights into the partitioning process allows QuickSort to dynamically adapt to the dataset's distribution, potentially enhancing its robustness against skewed distributions. While the concept is promising, the translation of LLM-driven insights into tangible partitioning strategies requires a deep integration of AI insights with algorithmic logic, presenting both technical and computational challenges.
	
	\paragraph{Strategic Formulation via LLM Analysis} The use of LLMs to formulate partitioning strategies involves a detailed analysis of the dataset, enabling the identification of optimal pivot points and partitioning approaches. This strategic formulation is aimed at achieving equitable partition distributions, a factor critical to QuickSort's performance. The adaptability facilitated by LLMs suggests a potential for QuickSort to remain efficient across diverse datasets. However, the complexity of dynamically adapting partitioning strategies based on LLM recommendations introduces considerations regarding the scalability of this approach and its impact on overall algorithm performance.
	
	\paragraph{Implementation and Algorithmic Adaptation} The practical implementation of this enhancement involves the integration of LLM-guided partitioning strategies into QuickSort, demanding modifications that allow for dynamic adjustments based on the dataset characteristics. While aiming to enhance efficiency, this approach must carefully navigate the trade-offs between improved partitioning and the added computational complexity. The execution of such dynamic strategies requires a robust framework for real-time analysis and adaptation, highlighting the need for efficient LLM integration mechanisms.
	
	\paragraph{Sorting Efficiency and Performance Gains} Incorporating Data Distribution Analysis aims to directly improve QuickSort's sorting efficiency by facilitating more balanced partitions. This strategy holds the promise of enhancing QuickSort's reliability and performance across varying datasets. However, the realization of these benefits hinges on the effective translation of LLM insights into practical sorting optimizations, a process that may face limitations in computational resource availability and algorithm adaptability.
	
	\subsubsection{Recursive Depth Optimization}
	
	\paragraph{Optimization of Recursive Calls} This enhancement focuses on optimizing QuickSort's recursive depth through LLM analysis, aiming to reduce unnecessary recursive calls and thereby enhance performance efficiency. The potential for LLMs to dynamically adjust recursion depth based on dataset characteristics offers a pathway to improved sorting efficiency. However, the practical implementation of such dynamic adjustments requires a delicate balance between recursion optimization and the maintenance of algorithm integrity, presenting challenges in the seamless integration of LLM recommendations.
	
	\paragraph{Adapting Recursive Strategies} Leveraging LLM insights for recursive strategy adaptation involves determining optimal recursion depths and potentially integrating non-recursive sorting methods for small datasets. This hybrid approach seeks to combine QuickSort's efficiency with the practical benefits of simpler sorting methods for specific scenarios. While the concept of adaptive recursion is promising, its practical application demands a comprehensive framework for real-time LLM analysis and strategy adaptation, considering the computational overhead involved.
	
	\paragraph{Implementation Dynamics and Efficiency} Integrating Recursive Depth Optimization into QuickSort requires modifications that enable dynamic decision-making regarding recursive calls, based on LLM insights. This adaptive approach aims to enhance QuickSort's performance, particularly in handling complex datasets. However, the dynamic nature of this enhancement introduces additional computational demands, necessitating a careful evaluation of the trade-offs between adaptive efficiency and algorithm complexity.
	
	\paragraph{Balancing Performance with Resource Utilization} The goal of optimizing recursive depth is to improve QuickSort's performance while managing resource utilization effectively. This enhancement promises to mitigate performance bottlenecks and enhance algorithm adaptability. However, achieving a balance between performance gains and the computational cost of implementing recursive depth optimization poses significant challenges, emphasizing the need for efficient LLM integration and algorithmic refinement.
	
	\subsubsection{Real-time Monitoring for Adaptive Strategy Switching}
	
	\paragraph{Dynamic Strategy Adaptation} This enhancement introduces the concept of real-time monitoring and adaptive strategy switching based on LLM insights, aiming to optimize QuickSort's performance dynamically. By enabling QuickSort to adjust its sorting strategy in response to evolving dataset characteristics, this approach seeks to enhance sorting efficiency and adaptability. However, the practicality of implementing real-time adaptive switching demands a robust framework for LLM integration and strategy execution, challenging the algorithm's responsiveness and computational efficiency.
	
	\paragraph{Mechanisms of Adaptive Switching} The implementation of adaptive strategy switching involves real-time analysis and decision-making processes guided by LLMs, allowing QuickSort to dynamically adjust its sorting approach. While this strategy promises enhanced performance through tailored sorting adaptations, the complexity of real-time LLM integration and the potential computational overhead pose significant challenges to its effective implementation.
	
	\paragraph{Integrating Adaptive Strategies} The integration of real-time monitoring and adaptive strategy switching into QuickSort necessitates modifications that incorporate LLM-guided decision-making into the sorting process. This approach aims to leverage the adaptability of LLMs to optimize QuickSort's performance across diverse scenarios. However, achieving seamless integration while maintaining computational efficiency and algorithm stability presents a complex challenge, requiring careful consideration of algorithmic and computational trade-offs.
	
	\paragraph{Efficiency and Adaptability Enhancements} By employing real-time adaptive strategy switching, this enhancement seeks to elevate QuickSort's efficiency and adaptability, ensuring robust performance across varying datasets. The promise of dynamic adaptability through LLM integration highlights the potential for significant performance improvements. However, the realization of these benefits is contingent upon overcoming the challenges associated with real-time LLM analysis and adaptive strategy implementation, underscoring the need for continued innovation in Algogenic algorithm design.
	
	\subsubsection{Parallel Partitioning Strategy}
	
	\paragraph{Leveraging Parallel Processing} The Parallel Partitioning Strategy aims to enhance QuickSort's performance through the parallel execution of partitioning tasks, leveraging modern multi-core and distributed computing environments. By employing LLMs to assess the feasibility of parallel partitioning, this strategy seeks to optimize sorting efficiency for large datasets. However, the practical implementation of parallel partitioning demands sophisticated coordination mechanisms and efficient resource utilization, presenting challenges in achieving scalable performance improvements.
	
	\paragraph{LLM-Guided Parallelization Decisions} The decision-making process for implementing parallel partitioning involves LLM analysis of dataset characteristics and computing environments, aiming to optimize the parallelization strategy for QuickSort. While the potential for performance gains through parallel execution is significant, the complexity of dynamically adjusting partitioning strategies based on LLM recommendations introduces challenges in computational efficiency and algorithm adaptability.
	
	\paragraph{Implementation Considerations and Scalability} Integrating a Parallel Partitioning Strategy into QuickSort requires modifications that enable the efficient parallel execution of partitioning tasks, considering the characteristics of the hardware architecture and dataset. The scalability of this approach promises enhanced sorting performance in diverse computing environments. However, addressing the challenges associated with parallel processing, such as data dependencies and synchronization overhead, is crucial for realizing the potential benefits of this enhancement.
	
	\paragraph{Performance Optimization Across Environments} The adoption of a Parallel Partitioning Strategy seeks to optimize QuickSort's performance in environments with high computational resources, offering scalable improvements through parallel execution. This strategy underscores the potential for leveraging modern computing architectures to enhance algorithm efficiency. However, the practicality of achieving significant performance gains through parallel partitioning depends on overcoming the inherent challenges of parallel algorithm implementation, emphasizing the need for advanced programming techniques and efficient resource management.

	\subsubsection{Pseudocode for Algogenic QuickSort}
	The Algogenic QuickSort approach integrates AI to enhance the conventional QuickSort algorithm by dynamically adapting sorting parameters and strategies according to the system's behavior and real-time performance evaluations. This pseudocode, accessible in \ref{fig:quicksort-Algogen-pseudocode}, illustrates a sophisticated framework that incorporates AI-driven improvements for adaptive pivot selection, partitioning strategies, termination conditions, and real-time parameter optimization.
	
	\begin{algorithm}
		\caption{Algogenic QuickSort Pseudocode}
		\begin{algorithmic}[1]
			\Procedure{AlgogenicQuickSort}{$Array$, $Low$, $High$}
			
			\If{$Low < High$}
			\State $PivotIndex \gets$ \Call{AdaptivePivotSelection}{$Array$, $Low$, $High$}
			\State $PivotNewIndex \gets$ \Call{Partition}{$Array$, $Low$, $High$, $PivotIndex$}
			\State \Call{AlgogenicQuickSort}{$Array$, $Low$, $PivotNewIndex - 1$}
			\State \Call{AlgogenicQuickSort}{$Array$, $PivotNewIndex + 1$, $High$}
			\EndIf
			
			\EndProcedure
			
			\Function{AdaptivePivotSelection}{$Array$, $Low$, $High$}
			\State Analyze $Array[Low:High]$ with LLM for optimal pivot strategy
			\State \Return PivotIndex based on LLM recommendations
			\EndFunction
			
			\Function{Partition}{$Array$, $Low$, $High$, $PivotIndex$}
			\State Perform data distribution analysis with LLM
			\State Optimize partitioning based on LLM insights
			\State \Return New pivot position after partitioning
			\EndFunction
			
		\end{algorithmic}\label{fig:quicksort-Algogen-pseudocode}
	\end{algorithm}
	
	\begin{figure}
		\centering
		\includegraphics[width=0.6\textwidth]{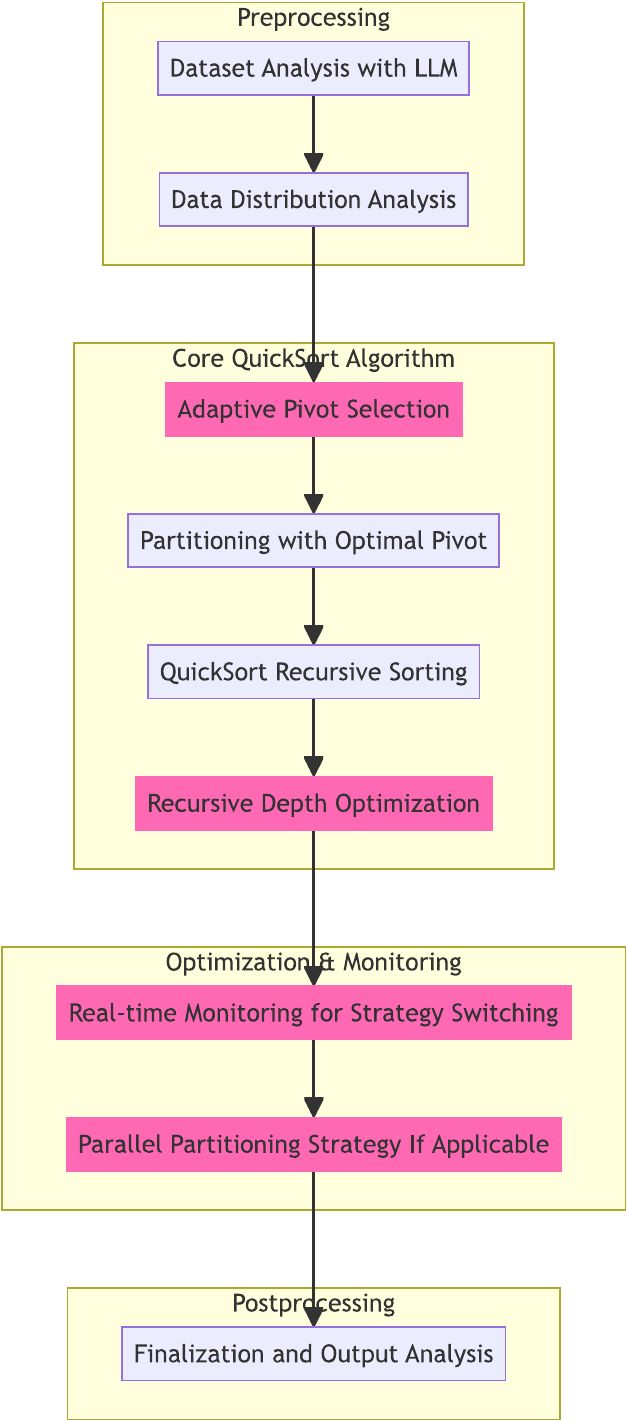}
		\caption{Algogenic Enhancements in QuickSort: This diagram showcases the integration of Algogenic enhancements with the QuickSort algorithm. It highlights the dynamic adaptation of pivot selection strategies and partitioning methods, informed by generative AI analysis of the dataset's characteristics. The adaptive pivot selection is implemented prior to partitioning, optimizing sorting efficiency based on real-time insights. Furthermore, data distribution analysis and recursive depth optimization are utilized to ensure balanced partitioning and efficient recursive calls, significantly improving QuickSort's performance across diverse datasets. This integration of Algogenic enhancements and traditional QuickSort algorithm demonstrates a significant leap in sorting algorithms' adaptability, efficiency, and overall effectiveness.}
		\label{fig:quicksort}
	\end{figure}

	\section{RSA}\index{RSA}
	\subsection{Introduction to RSA}
	\subsubsection{The Concept of the RSA Algorithm}
	
	\paragraph{Introduction to RSA}
	The RSA algorithm, named after its inventors Rivest, Shamir, and Adleman, is a cornerstone of modern cryptographic systems, providing a foundation for secure communication in an increasingly digital world. Introduced in 1977, RSA represents one of the first practical implementations of public key cryptography, a system where encryption and decryption are performed using separate keys—a public key for encryption and a private key for decryption. This innovative approach to cryptography has revolutionized the way sensitive information is protected online.
	
	Moreover, RSA has become a fundamental component of various security protocols, including HTTPS, SSH, and SSL/TLS, ensuring the confidentiality, integrity, and authenticity of data transmitted over networks. Furthermore, the security of RSA relies on the difficulty of factoring large prime numbers, a problem widely believed to be computationally infeasible with current technology. Additionally, the mathematical foundation of RSA lies in modular arithmetic and number theory, where the security of the algorithm stems from the challenge of deriving the private key from the public key without knowledge of the prime factors. 
	
	Furthermore, RSA's versatility extends beyond traditional encryption and decryption tasks; it also facilitates digital signatures, enabling entities to authenticate the origin and integrity of messages. This capability is crucial for establishing trust in electronic transactions and communication channels. Furthermore, the robustness and widespread adoption of RSA have contributed to its longevity and resilience in the face of evolving cyber threats.
	
	In conclusion, RSA stands as a testament to the power of innovative cryptographic techniques in safeguarding sensitive information in the digital age. Its impact on cybersecurity and secure communication cannot be overstated, making it an indispensable tool for ensuring privacy and security in an interconnected world.

	\paragraph{Mathematical Foundations}
	At the heart of the RSA algorithm lies a beautiful interplay of number theory and computational complexity. The security of RSA is based on the practical difficulty of factorizing a large composite number into its prime factors, a problem known in mathematics as prime factorization. Specifically, the algorithm utilizes two large prime numbers, often hundreds of digits long, to generate the public and private keys. These keys are linked mathematically; moreover, the private key cannot be feasibly derived from the public key due to the computational intractability of the prime factorization problem for large numbers.
	
	The RSA algorithm leverages the concept of modular arithmetic and the properties of exponentiation to ensure secure communication over insecure channels. It relies on the fact that while multiplying two large prime numbers together to obtain a composite number is relatively easy, the reverse operation, namely, factorizing the composite number back into its prime factors, becomes exceedingly difficult as the size of the numbers increases. This asymmetry in computational complexity forms the basis of RSA's security.
	
	Furthermore, the security of RSA rests on the assumption that there are no efficient algorithms capable of factoring large numbers in polynomial time, a conjecture that remains unproven but widely believed in the mathematical community. Additionally, the security margin of RSA depends on the size of the keys chosen; larger key sizes offer greater resistance against potential attacks, as they increase the computational effort required to factorize the keys.
	
	Hence, the RSA algorithm stands as a testament to the profound connection between number theory, computational complexity, and cryptography, showcasing how seemingly simple mathematical concepts can be ingeniously applied to create robust encryption schemes that underpin the security of modern communication systems.

	\paragraph{Key Generation Process}
	The process of key generation in RSA involves several crucial steps. Firstly, two large random prime numbers, denoted as $p$ and $q$, are carefully selected. These primes serve as the building blocks for the RSA keys. Their product, $n = p \times q$, establishes the modulus for both the public and private keys, thereby determining the key length, which is vital for the security of the encryption.
	
	Subsequently, the totient of $n$, $\phi(n) = (p-1) \times (q-1)$, is computed. This totient holds significant importance as it is employed in the generation of both the public and private key components. From $\phi(n)$, the public key exponent $e$ is derived. It is chosen such that it satisfies two essential conditions: $1 < e < \phi(n)$ and $e$ is coprime to $\phi(n)$. This careful selection ensures the effectiveness of the RSA encryption process.
	
	Conversely, the private key exponent $d$ is computed as the modular multiplicative inverse of $e$ modulo $\phi(n)$. This calculation guarantees that $d \times e \equiv 1 \mod \phi(n)$, a fundamental property essential for the decryption process to function correctly.
	
	It is noteworthy that the public key, comprising the pair $(n, e)$, is shared openly, while the private key, consisting of $(n, d)$, is kept securely by the key holder. This clear distinction between the public and private components ensures the integrity and confidentiality of data transmission in RSA encryption.

	\paragraph{Encryption and Decryption Mechanisms}
	Encryption using RSA involves raising the plaintext message $m$, treated as an integer less than $n$, to the power of the public key exponent $e$ and then taking the modulus $n$ of the result to produce the ciphertext $c$, where $c = m^e \mod n$. This process underscores the foundational principle of RSA encryption, which relies on the computational complexity of factoring large numbers into their prime components to ensure security. The choice of the public key exponent $e$ is typically a small prime number, often 65537, chosen for its efficiency in the encryption process. Conversely, the private key exponent $d$ is derived from $e$ using modular arithmetic, ensuring that it possesses the inverse relationship necessary for decryption. This elegant symmetry between encryption and decryption, facilitated by the properties of modular arithmetic, ensures that only the holder of the private key can decrypt messages encrypted with the corresponding public key. Moreover, the security of RSA encryption hinges on the difficulty of factoring the modulus $n$ into its constituent prime factors, a task believed to be computationally infeasible for sufficiently large primes. Therefore, the strength of RSA encryption lies in the impracticality of factoring large numbers, which serves as the foundation of its security. Additionally, RSA encryption is widely used in various secure communication protocols, such as HTTPS, SSH, and PGP, due to its robustness and versatility in securing sensitive information.

	\paragraph{RSA's Role in Digital Security}
	The RSA algorithm has played a pivotal role in securing digital communications, enabling not only the encryption of sensitive information but also the authentication of digital signatures, thereby ensuring data integrity and non-repudiation. Its widespread adoption in protocols like SSL/TLS for secure web browsing, email encryption standards, and digital certificates underpins much of the trust model of the internet. Despite challenges from emerging technologies like quantum computing, RSA's legacy as a fundamental building block of cryptographic security endures, exemplifying the profound impact of mathematical theory on practical applications in digital security.
	
	Moreover, the elegance of the RSA algorithm lies in its reliance on the computational complexity of prime factorization, which forms the basis of its security. This reliance ensures that even with advances in computing power, breaking RSA encryption remains a formidable task. Furthermore, the versatility of RSA extends beyond traditional digital communication channels; it finds applications in various domains, including secure messaging apps, financial transactions, and authentication mechanisms for IoT devices.
	
	Additionally, RSA's resilience stems from its robustness against known cryptographic attacks when implemented correctly. This resilience is further augmented by continuous advancements in key management practices and encryption techniques, bolstering its effectiveness in thwarting cyber threats. Furthermore, ongoing research and development efforts in the field of post-quantum cryptography aim to fortify RSA's defenses against potential vulnerabilities posed by quantum adversaries.
	
	In conclusion, the enduring relevance of the RSA algorithm underscores its indispensable role in contemporary digital security landscapes. Its steadfastness in the face of evolving threats and its adaptability to diverse use cases highlight its significance as a cornerstone of cryptographic infrastructure. Thus, RSA stands as a testament to the enduring synergy between mathematical innovation and practical cybersecurity solutions.

	\subsubsection{Key Principles and Mechanisms}
	
	\paragraph{Underlying Principles of RSA}
	The RSA algorithm, a cornerstone of modern cryptography, draws its strength from the intricate interplay of number theory and computational complexity. Central to its robustness is the formidable challenge posed by the factorization of large composite numbers into their prime constituents. This formidable task forms the linchpin of RSA's security framework, rendering the extraction of the private key from its corresponding public counterpart a computationally prohibitive endeavor.
	
	At the heart of RSA lie the elegant principles of modular arithmetic, deftly harnessed to fashion a cryptographic scheme of unparalleled resilience. Leveraging the inherent properties of prime numbers, RSA orchestrates a delicate dance of mathematical operations, ingeniously crafted to withstand the relentless onslaught of cryptographic attacks. Through judicious manipulation of modular exponentiation and the Chinese Remainder Theorem, RSA constructs an impregnable fortress of encryption and decryption, where the sanctuary of data remains inviolate against the prying eyes of adversaries.
	
	In essence, the RSA algorithm operates on the premise of harnessing the inherent complexity of prime factorization as a bulwark against the incursions of unauthorized access. By entrusting the security of digital communications to the enigmatic labyrinth of prime numbers and modular arithmetic, RSA bequeaths a legacy of cryptographic fortitude that stands as a testament to the enduring elegance of mathematical principles in the realm of cybersecurity.

	\paragraph{Mechanisms of RSA Encryption and Decryption}
	RSA's encryption and decryption mechanisms rely on the mathematical properties of modular exponentiation. For encryption, a message $m$, represented as an integer smaller than the modulus $n$, is encrypted by raising it to the power of the public exponent $e$ and then taking the result modulo $n$. This process ensures that the original message cannot be easily derived from the ciphertext without knowledge of the private key. The choice of $e$ as the public exponent is crucial for security; it must be a relatively prime number to Euler's totient function of $n$, ensuring that the encryption process is one-way and irreversible without the private key. The decryption process, on the other hand, utilizes the private exponent $d$ to reverse the encryption operation. This is achieved by raising the ciphertext, obtained through the encryption process, to the power of $d$ modulo $n$. Euler's theorem is fundamental here, stating that for any integer $a$ coprime to $n$, $a^{\phi(n)} \equiv 1 \pmod{n}$, where $\phi(n)$ is Euler's totient function. By extension, this implies that $(a^e)^d \equiv a^{ed} \equiv a \pmod{n}$, allowing for accurate retrieval of the original message. The security of RSA encryption lies in the difficulty of factoring the large composite number $n$ into its prime factors, which is a computationally intensive task even for modern computers, particularly when $n$ is chosen to be sufficiently large. Therefore, RSA encryption provides a robust method for secure communication over untrusted channels, ensuring confidentiality and integrity of transmitted data.

	\paragraph{Key Generation and the Role of Euler's Totient Function}
	The security of RSA hinges on the generation of robust public and private keys through a meticulous process involving Euler's totient function. The selection of two large prime numbers $p$ and $q$ and their product $n = p \times q$ establish the basis for the keys. Euler's totient function $\phi(n) = (p-1) \times (q-1)$ plays a critical role in determining a suitable public exponent $e$. Furthermore, it ensures the derivation of its corresponding private exponent $d$, thereby completing the key pair. 
	
	And, crucially, the relationship between $e$ and $\phi(n)$ is pivotal. They must be coprime, meaning they share no common factors other than 1, to guarantee the security of the encryption process. Moreover, the computation of $d$ as the modular multiplicative inverse of $e$ modulo $\phi(n)$ ensures that only the possessor of the private key can decipher messages encrypted with the public key. 
	
	Additionally, the values of $p$ and $q$ must be chosen carefully to prevent attacks like the factorization of $n$ using methods like Fermat's factorization algorithm or Pollard's rho algorithm. Likewise, ensuring the secrecy of $p$ and $q$ is paramount, as the disclosure of either prime could compromise the security of the entire RSA system. Hence, meticulous key generation procedures, guided by Euler's totient function, are indispensable in maintaining the robustness of RSA encryption.

	\paragraph{Public and Private Keys: Dual Functions}
	RSA keys serve dual functions: the public key $(n, e)$ is used for encrypting messages and verifying digital signatures, while the private key $(n, d)$ is used for decrypting messages and creating digital signatures. This duality allows for secure communication channels where only the intended recipient can decrypt the message, and the authenticity of the sender can be verified through digital signatures.
	
	Furthermore, this system operates on the principle of asymmetric cryptography. Unlike symmetric cryptography, where the same key is used for both encryption and decryption, RSA employs two distinct keys, enhancing security. Moreover, the public key can be freely shared with anyone, facilitating secure communication with multiple parties simultaneously. Additionally, the private key remains confidential to its owner, ensuring exclusive access to decrypt messages and create signatures.
	
	Moreover, RSA encryption provides a robust solution against eavesdropping and message tampering. Even if an attacker intercepts the encrypted message, they cannot decipher it without the corresponding private key. Similarly, digital signatures generated using the private key are unique to the sender, preventing unauthorized parties from forging messages.
	
	Hence, RSA encryption offers a versatile framework for establishing secure communication channels and authenticating digital identities. By leveraging the dual nature of public and private keys, it addresses key challenges in data privacy and integrity in the digital domain.

	\paragraph{Computational Complexity and Security}
	The security of the RSA algorithm is directly correlated with the key size; larger keys offer higher security but require more computational power for encryption and decryption processes. This trade-off underscores the critical balance between security and computational efficiency in cryptographic systems. With increasing key size, the computational complexity of RSA encryption and decryption grows polynomially, resulting in longer processing times. Consequently, while RSA remains a widely used encryption method due to its proven security, the scalability of its computational demands raises concerns, particularly in resource-constrained environments.
	
	The infeasibility of solving the prime factorization problem for large $n$ within a practical timeframe without prior knowledge of $p$ and $q$ ensures the robustness of RSA against brute-force attacks. The reliance on the difficulty of factoring large semiprime numbers underpins the security of RSA, as it necessitates a computationally prohibitive effort to derive the private key from the public key. This fundamental property has solidified RSA's position as a cornerstone of modern encryption standards.
	
	However, advancements in computational resources, including the potential development of quantum computers, pose challenges to RSA's security. Quantum computing's capacity for efficiently solving certain mathematical problems, such as integer factorization, threatens RSA's security model. The prospect of quantum algorithms, such as Shor's algorithm, undermining RSA's security underscores the need for cryptographic agility and the development of quantum-resistant algorithms. Consequently, ongoing research endeavors focus on exploring post-quantum cryptography approaches that can withstand the computational power of quantum adversaries while ensuring the integrity and confidentiality of sensitive information.

	\paragraph{RSA in Practice: Protocols and Standards}
	In practical applications, RSA is a cornerstone of various security protocols and standards, ensuring the confidentiality, integrity, and authenticity of digital communications. Its implementation in secure web protocols, digital certificate frameworks, and encrypted messaging services exemplifies RSA's critical role in modern cryptography. \textbf{Moreover}, the ubiquity of RSA in cryptographic applications underscores its significance in safeguarding sensitive information transmitted over networks. Secure web protocols such as HTTPS rely on RSA encryption to establish secure communication channels between clients and servers, \textbf{ensuring} that data exchanged remains confidential and cannot be intercepted by malicious actors. \textbf{Additionally}, digital certificate frameworks like X.509 leverage RSA-based digital signatures to verify the authenticity and integrity of digital identities and documents, \textbf{further enhancing} trust and security in online transactions and communications. Encrypted messaging services, \textbf{such as} Signal and WhatsApp, utilize RSA encryption to protect the privacy of user communications, \textbf{thereby} preventing unauthorized access to sensitive information. The widespread adoption of RSA across these protocols and services highlights its versatility and effectiveness in addressing diverse security requirements. \textbf{Furthermore}, ongoing advancements in RSA implementations and optimizations continually \textbf{strengthen} the security posture of digital systems, \textbf{ensuring} that they remain resilient against evolving cyber threats. In summary, RSA's integration into various protocols and standards not only reinforces the foundation of modern cryptography but also underscores its indispensable role in shaping the secure digital ecosystem.

	\subsubsection{The Role of Prime Numbers}
	
	\paragraph{Foundation of RSA's Security}
	The security of the RSA algorithm is deeply rooted in the properties of prime numbers and their pivotal role in the field of number theory. Prime numbers, which are integers greater than 1 that have no divisors other than 1 and themselves, serve as the foundation for the RSA algorithm's key generation process. The selection of two large, random prime numbers, $p$ and $q$, and their subsequent use in generating the public and private keys, encapsulates the core of RSA's encryption and decryption mechanisms. \textbf{Moreover}, the unique properties of prime numbers, such as their computational complexity in determining their factors, contribute significantly to RSA's robustness against attacks. The \textbf{difficulty associated with factorizing} the product of these two primes, especially as their size increases, forms the bedrock of RSA's cryptographic strength. \textbf{Furthermore}, the reliance on prime factorization as the basis for RSA's security aligns with the difficulty of this problem in the realm of computational complexity theory. This inherent difficulty forms a barrier for adversaries attempting to break RSA encryption through brute force or mathematical algorithms. \textbf{Consequently}, RSA's security rests not only on the complexity of prime factorization but also on the absence of efficient algorithms for solving this problem in polynomial time, thus ensuring the viability of RSA as a secure encryption scheme.

	\paragraph{Prime Number Selection and RSA Key Generation}
	The initial step in RSA key generation involves choosing two distinct large prime numbers. These primes serve as the building blocks for generating the modulus $n = p \times q$, which is a pivotal element in both the public and private keys. The magnitude of $n$ not only dictates the length of the key but also significantly impacts the security and efficiency of the RSA algorithm. The careful selection of these prime numbers is paramount; they must be sufficiently large to resist brute-force attacks while ensuring computational feasibility for encryption and decryption processes.
	
	The prime nature of $p$ and $q$ plays a fundamental role in RSA's security architecture. By utilizing prime numbers, the factorization of $n$ becomes exceedingly challenging, thus safeguarding the confidentiality and integrity of encrypted data. This cryptographic hardness assumption forms the bedrock of RSA's resilience against attacks such as integer factorization, which seeks to unravel the prime factors of $n$. Furthermore, the unique properties of prime numbers, including their indivisibility except by themselves and 1, contribute to the robustness of RSA encryption.
	
	In RSA key generation, the reliance on prime numbers underscores the importance of leveraging mathematical principles to fortify cryptographic systems. The meticulous choice of primes ensures that the resulting modulus $n$ remains resistant to factorization attempts, thereby upholding the security guarantees provided by the RSA algorithm.

	\paragraph{Euler's Totient Function and Its Role}
	Euler's totient function, $\phi(n)$, calculated as $(p-1) \times (q-1)$ for RSA, measures the count of integers less than $n$ that are coprime to $n$. This function plays a crucial role in determining the public exponent $e$ and its corresponding private exponent $d$. The relationship between $e$ and $d$ as modular multiplicative inverses under $\phi(n)$ ensures that the encryption and decryption operations are inverses of each other, enabling the recovery of the original message after encryption.
	
	Furthermore, the significance of Euler's totient function extends beyond RSA encryption. It serves as a fundamental tool in various branches of number theory and cryptography. Additionally, the efficiency of RSA encryption heavily relies on the computational complexity of calculating $\phi(n)$, which is challenging for large prime numbers $p$ and $q$. However, advancements in algorithms and computing power continue to address this challenge.
	
	Moreover, Euler's totient function forms the basis for other cryptographic protocols and algorithms, such as ElGamal encryption and digital signature schemes. Its role in ensuring the security and integrity of cryptographic systems cannot be overstated. Additionally, understanding the properties of $\phi(n)$ allows cryptographers to devise stronger encryption methods and protocols.
	
	In summary, Euler's totient function serves as a cornerstone in modern cryptography, providing the mathematical foundation for secure communication and data protection. Its application in RSA encryption exemplifies its importance in ensuring the confidentiality and integrity of sensitive information in digital environments.

	\paragraph{Uniqueness and Predictability Challenges}
	While prime numbers offer robust security advantages, their selection process in RSA must be handled with care to avoid predictability and ensure uniqueness. The generation of primes $p$ and $q$ requires the use of cryptographic-quality random number generators to mitigate the risk of key predictability and subsequent security vulnerabilities. 
	
	Moreover, the primes must be sufficiently large and chosen independently to prevent efficient factorization of $n$, maintaining RSA's resistance to various cryptographic attacks. Additionally, ensuring that $p$ and $q$ are distinct primes is crucial. Failure to select distinct primes could lead to vulnerabilities such as Carmichael numbers, which are composite numbers erroneously identified as primes due to their behavior in certain modular arithmetic operations.
	
	Consequently, the selection of primes in RSA is a meticulous process, involving stringent criteria to uphold both uniqueness and unpredictability. The significance of this process cannot be overstated, as the security of RSA encryption hinges upon the robustness of the chosen primes. Hence, meticulous attention to detail in prime selection is imperative for the continued efficacy of RSA as a secure encryption algorithm.

	\paragraph{Advancements in Prime Number Research}
	Ongoing research in prime number theory and computational methods for prime generation continues to influence RSA's application and development. Improvements in algorithms for prime testing and the exploration of new mathematical insights into the distribution and properties of primes contribute to enhancing RSA's security and efficiency. 
	
	Furthermore, as computational capabilities evolve, so too does the need for adapting the prime number selection process. Consequently, ensuring that RSA remains a secure method for public-key cryptography is imperative in the face of emerging computational and cryptographic challenges. Moreover, advancements in prime number research not only bolster the security of RSA but also pave the way for innovations in related cryptographic protocols.
	
	While traditional methods such as sieving and primality testing algorithms have long been foundational in prime number research, novel approaches, including elliptic curve methods and probabilistic algorithms, offer promising avenues for efficiently generating and verifying large prime numbers. Additionally, the exploration of deeper mathematical connections, such as those between prime numbers and other areas of number theory, sheds light on previously unexplored facets of prime distribution and behavior.
	
	In parallel, the quest for ever-larger prime numbers persists, driven both by theoretical curiosity and practical cryptographic concerns. Moreover, the investigation of special classes of primes, such as Sophie Germain primes or Mersenne primes, not only deepens our understanding of prime number theory but also presents intriguing possibilities for cryptographic applications.
	
	Hence, ongoing advancements in prime number research play a pivotal role in shaping the landscape of modern cryptography, ensuring the continued robustness and relevance of RSA and its derivatives in safeguarding sensitive information in an increasingly interconnected digital world.

	\subsubsection{Applications and Limitations}
	
	\paragraph{Widespread Applications of RSA}
	The RSA algorithm, since its inception, has been a cornerstone of secure digital communications, serving as the backbone for various encryption, digital signature, and authentication protocols. Its applications span secure email communications, ensuring confidentiality and integrity in electronic correspondence. \textbf{Moreover}, RSA is instrumental in establishing SSL/TLS protocols for secure web browsing, where sensitive information like credit card details, login credentials, and personal data are transmitted securely over the internet. Additionally, RSA plays a crucial role in the realm of VPNs (Virtual Private Networks), offering encrypted internet connections \textbf{as well as} secure access to corporate networks from remote locations. The secure exchange of keys facilitated by RSA enhances the confidentiality and integrity of data in many cryptographic systems, \textbf{furthermore}, its widespread adoption underscores its efficacy and trustworthiness in safeguarding sensitive information. \textbf{On the other hand}, RSA's versatility and robust security measures make it a preferred choice not only for commercial applications but also for government, military, and financial sectors where the stakes are particularly high, and stringent security measures are paramount. The adoption of RSA encryption ensures that data transmitted across networks remains confidential, \textbf{in contrast} to vulnerable plaintext transmissions susceptible to eavesdropping and unauthorized access. Therefore, the enduring prominence of RSA in diverse domains underscores its significance in modern cryptography, where secure communication is indispensable for protecting sensitive information.

	\paragraph{Digital Signatures and Authentication}
	One of the significant applications of RSA is in the creation and verification of digital signatures, which are essential for ensuring the integrity and authenticity of digital documents and software distributions. RSA enables the sender of a message to generate a unique signature that can be verified by the recipient using the sender's public key, thereby confirming the message's origin and integrity. This application is crucial in legal, financial, and commercial transactions where non-repudiation and authenticity are paramount.
	
	Furthermore, digital signatures play a pivotal role in establishing trust in electronic communication and transactions. In a world where cyber threats loom large, digital signatures offer a robust mechanism for ensuring the validity and integrity of data exchanged over insecure channels. Moreover, the reliance on RSA for digital signatures underscores its widespread adoption and trust in the security community.
	
	Additionally, digital signatures provide a means for verifying the identity of the sender without the need for physical presence or reliance on traditional paper-based signatures. This streamlines processes, reduces overhead costs, and facilitates seamless electronic transactions. Moreover, the mathematical underpinnings of RSA ensure that even a slight alteration in the message would result in a completely different signature, making it virtually impossible for malicious actors to forge signatures undetected.
	
	Hence, the use of RSA for digital signatures not only enhances security but also fosters efficiency and convenience in today's digital age. As a result, it has become an indispensable tool for ensuring trust and integrity in a wide range of online interactions.

	\paragraph{Limitations in Computational Efficiency}
	Despite its security strengths, RSA faces limitations related to computational efficiency, particularly when compared to symmetric key cryptographic algorithms. The mathematical operations involved in RSA encryption and decryption, especially with large key sizes necessary for strong security, can be significantly more computationally intensive than those required by symmetric key algorithms. This can lead to performance bottlenecks in environments where fast encryption and decryption of large volumes of data are required. Furthermore, the asymmetry inherent in RSA, where encryption and decryption keys are different, adds to the computational overhead, as it requires separate operations for encryption and decryption, unlike symmetric algorithms where the same key is used for both operations. Consequently, in scenarios where computational resources are limited, or where real-time processing of data is crucial, RSA may not be the most suitable choice. Nevertheless, despite these computational challenges, RSA remains widely used in various applications, especially where its security guarantees outweigh the performance considerations. Moreover, advancements in hardware and algorithms continually strive to mitigate these efficiency concerns, but the fundamental computational overhead of RSA encryption and decryption persists, shaping the choice of cryptographic algorithms in practical implementations.

	\paragraph{Challenges Posed by Quantum Computing}
	The advent of quantum computing poses a theoretical threat to RSA and other public-key cryptographic algorithms based on the difficulty of factorizing large numbers. Quantum algorithms, such as Shor's algorithm, could potentially factorize the large composite numbers central to RSA's security much more efficiently than classical algorithms, undermining the algorithm's security foundation. Additionally, the vulnerability of RSA to quantum attacks raises concerns about the security of encrypted communications, financial transactions, and sensitive data stored online. Moreover, the potential breakthrough in quantum computing not only jeopardizes RSA but also challenges the security of various encryption standards widely used today, including ECC (Elliptic Curve Cryptography) and DH (Diffie-Hellman). Furthermore, the emergence of practical quantum computers capable of breaking RSA could lead to significant disruptions across industries that rely on secure communication protocols and data encryption for operations and transactions. Nevertheless, the cryptographic community is actively researching quantum-resistant algorithms to prepare for future developments, aiming to mitigate the risks posed by quantum computing to cybersecurity.

	\paragraph{Key Size and Security Trade-offs}
	The security of RSA is directly related to the key size; larger keys offer greater security but also require more processing power for encryption and decryption operations. \textbf{Moreover}, as computational capabilities advance, the key sizes considered secure have increased, \textbf{therefore} leading to a continuous need to balance security with computational efficiency. This trade-off highlights the need for ongoing evaluation of RSA's parameters \textbf{and} the exploration of alternative or complementary cryptographic approaches to ensure robust security in the evolving technological landscape. 
	
	Increasing the key size in RSA enhances security by making it computationally infeasible for an adversary to factorize the modulus into its prime factors. However, this comes at the cost of increased computational overhead. \textbf{Additionally}, advancements in computing technologies, such as quantum computing, pose potential threats to the security of RSA, further emphasizing the importance of exploring alternative cryptographic schemes. 
	
	\textbf{On the other hand}, reducing the key size for efficiency purposes may compromise security, leaving systems vulnerable to attacks exploiting weaknesses in smaller key sizes. \textbf{Nevertheless}, ongoing research focuses on developing efficient implementations and algorithms to mitigate the computational burden associated with larger key sizes, allowing for a more balanced approach between security and efficiency in RSA encryption and decryption operations.

	\paragraph{Future Directions and Adaptability}
	In response to these challenges and limitations, the cryptographic community continues to innovate, developing more efficient algorithms, enhancing key generation methods, and exploring post-quantum cryptography solutions. The adaptability of RSA, combined with its proven track record, suggests that it will continue to play a crucial role in cryptographic applications, either in its classic form or through integration with new cryptographic paradigms designed to address its limitations and ensure the security of digital communications in the quantum era.
	
	Furthermore, as quantum computing capabilities advance, the need for robust cryptographic solutions becomes increasingly urgent. Therefore, researchers are not only focusing on improving existing algorithms but also investigating entirely new cryptographic primitives that can withstand the threats posed by quantum adversaries. Additionally, there is a growing emphasis on developing cryptographic techniques that offer provable security guarantees, as traditional approaches may be vulnerable to emerging attacks.
	
	Moreover, the evolution of cryptographic standards is closely intertwined with advancements in hardware and software technologies. Consequently, collaborations between cryptographic experts, hardware manufacturers, and software developers are essential to ensure the seamless integration of cryptographic solutions into various platforms and applications. This interdisciplinary approach facilitates the rapid deployment of secure systems and fosters innovation in cryptographic design.
	
	In the same way, interdisciplinary research efforts are crucial for addressing the practical challenges associated with key management, secure communication protocols, and the integration of cryptographic mechanisms into emerging technologies such as the Internet of Things (IoT), cloud computing, and blockchain. By leveraging insights from diverse fields, researchers can develop holistic solutions that not only enhance security but also promote usability, scalability, and interoperability.
	
	Overall, the future of cryptography lies in its ability to adapt to evolving threats and technological landscapes. By embracing innovation, collaboration, and interdisciplinary approaches, the cryptographic community can continue to safeguard digital assets and enable trust in an increasingly interconnected world.

	\subsubsection{Pseudocode for the RSA Algorithm}
	The RSA Algorithm is a foundational method utilized in modern cryptography for securely exchanging information over insecure channels. Its core procedures are outlined in a pseudocode, as depicted in pseudo figure \ref{fig:rsa-pseudocode}. Key generation initiates the process by selecting two large prime numbers to establish both public and private keys. Subsequently, encryption involves raising the plaintext message, represented as an integer, to the power of the public exponent $e$ and then taking the result modulo $n$ to generate the ciphertext. Decryption, on the other hand, reverses this operation using the private exponent $d$ to recover the original message from the ciphertext. The security of the RSA algorithm is grounded in the computational complexity associated with factoring the product of two large prime numbers, which forms the basis of its resistance against cryptographic attacks.
	
	\begin{algorithm}
		\caption{RSA Algorithm Pseudocode}
		\begin{algorithmic}[1]
			\Procedure{GenerateKeys}{}
			\State Choose two large random primes $p$ and $q$
			\State Compute $n = p \times q$
			\State Compute $\phi(n) = (p-1) \times (q-1)$
			\State Choose an integer $e$ such that $1 < e < \phi(n)$ and $\gcd(e, \phi(n)) = 1$
			\State Compute $d$ as $d \equiv e^{-1} \mod \phi(n)$
			\State \textbf{return} $(n, e)$ as public key and $(n, d)$ as private key
			\EndProcedure
			
			\Procedure{Encrypt}{plaintext, public\_key}
			\State Parse the plaintext into an integer $m < n$
			\State Compute ciphertext $c \equiv m^e \mod n$
			\State \textbf{return} $c$
			\EndProcedure
			
			\Procedure{Decrypt}{ciphertext, private\_key}
			\State Compute the original message $m \equiv c^d \mod n$
			\State Parse $m$ back into plaintext
			\State \textbf{return} plaintext
			\EndProcedure
		\end{algorithmic}\label{fig:rsa-pseudocode}
	\end{algorithm}
	
\subsection{Previous Work on ML and AI Interplay with Security Algorithms}

\paragraph{Enhancing Cloud Computing Security through Machine Learning Approaches}
A review by Butt et al. \cite{butt2020review} discusses the integration of machine learning algorithms to enhance cloud computing security. The paper examines the role of AI and ML in identifying, mitigating, and preventing security threats within cloud environments. Various machine learning algorithms applied to enhance the security mechanisms of cloud services are categorized. The effectiveness of these algorithms in detecting anomalies, recognizing patterns in data traffic, and predicting potential security breaches is explored. This work emphasizes the importance of machine learning in fortifying cloud security and showcases the evolution of these algorithms to adapt to dynamic cyber threats. By leveraging AI and ML capabilities, cloud services can achieve a higher level of security, ensuring data integrity, confidentiality, and availability. The insights from this review suggest that continuous advancement and application of machine learning algorithms are critical for developing robust security solutions to combat the evolving landscape of cyber threats in cloud computing.

\paragraph{Cybersecurity Data Science: A Machine Learning Perspective}
Sarker et al. \cite{sarker2020cybersecurity} provide an overview of how machine learning perspectives are influencing cybersecurity data science. The paper discusses the synergy between machine learning and cybersecurity, illustrating how ML algorithms improve the detection of cyber threats and vulnerabilities. It analyzes different machine learning techniques applicable in cybersecurity, including supervised and unsupervised learning, deep learning, and reinforcement learning. The discussion extends to how these methods can analyze and interpret vast amounts of data for identifying sophisticated cyber-attacks. Moreover, the paper highlights challenges and opportunities in integrating machine learning with cybersecurity, such as the need for large and representative datasets, developing models adaptable to changing threat landscapes, and potential for AI to automate threat response. Through this exploration, it becomes evident that machine learning is fundamental for advancing cybersecurity measures, offering promising avenues to enhance system resilience against malicious activities.

	\subsection{Algogenic Enhancements for RSA}
	
	\subsubsection{Optimized Prime Number Generation}
	
	\paragraph{Enhancing Efficiency with Adaptive Prime Prediction Techniques}
	In the specific application to RSA's prime number generation, leveraging Large Language Models presents a nuanced approach towards enhancing efficiency. By focusing on adaptive prime prediction techniques, LLMs can refine the process of identifying prime numbers through pattern recognition and predictive analytics. This method, while promising, requires careful consideration to ensure that the predictive models do not compromise the randomness and security inherent in prime number selection for RSA keys. The practical implementation involves training LLMs on number theory principles and prime distribution patterns, enabling them to suggest number ranges where primes are likely denser. However, the skepticism towards this approach lies in its reliance on the models' ability to continuously learn and adapt without introducing biases or vulnerabilities that could be exploited.
	
	\paragraph{Skeptical View on Predictive Modelling for Prime Generation}
	While the incorporation of predictive models into RSA's prime generation process could theoretically streamline the identification of suitable primes, there is a need for a cautious and skeptical perspective. The effectiveness of LLMs in accurately predicting prime-rich intervals hinges on their training data and algorithms. If not meticulously designed, these models could inadvertently overlook certain prime candidates or bias the selection process, potentially weakening the cryptographic strength of RSA keys. The practicality of implementing these predictive models involves complex algorithmic adjustments and continuous refinement to ensure that the security and randomness of prime number generation are not compromised.
	
	\paragraph{Adaptive Learning for Enhanced Prime Verification}
	The integration of adaptive learning mechanisms into RSA's prime verification process offers a potential for efficiency gains. LLMs could be employed to dynamically adjust verification strategies based on previous outcomes, potentially reducing the computational overhead. However, the practical implementation of such adaptive learning systems must be approached with caution. Ensuring that these systems do not develop predictable patterns that could be exploited by adversaries is paramount. The value of this enhancement lies in its ability to potentially streamline the prime verification process, but it must be underpinned by robust security measures to prevent any compromise in RSA's integrity.
	
	\paragraph{Critical Analysis of Security Implications}
	The suggestion that LLMs could improve RSA's cryptographic strength by generating larger, more robust primes carries with it significant security implications. A critical analysis reveals that while larger primes do indeed enhance security, the method of generation and verification by LLMs must be transparent and verifiable. The implementation of such LLM-enhanced prime generation processes must include rigorous security assessments to ensure that the introduction of machine learning does not inadvertently introduce vulnerabilities. The practical translation of this enhancement into RSA's framework requires a careful balance between innovation and the fundamental principles of cryptographic security.
	
	\paragraph{Future Perspectives on LLM Integration with RSA}
	Looking towards the future, the integration of LLMs into RSA's framework, particularly in the domain of prime number generation, presents both opportunities and challenges. The potential for LLMs to continually adapt and improve prime generation and verification methods could pave the way for more efficient and secure RSA implementations. However, the practical realization of these enhancements necessitates a commitment to ongoing research, interdisciplinary collaboration, and a cautious approach to integrating AI technologies into cryptographic systems. The value of such advancements will ultimately be measured by their ability to enhance RSA's security without compromising its foundational principles.
	
	\subsubsection{Dynamic Key Length Adjustment}
	
	\paragraph{Adapting to Computational Advances with LLMs}
	The application of Algogens for RSA's dynamic key length adjustment specifically addresses the challenge of adapting to rapid advances in computational power and emerging threats like quantum computing. The use of LLMs in predicting future computational capabilities and recommending timely adjustments to key lengths can significantly enhance RSA's resilience. However, skepticism arises from the complexity of accurately forecasting computational advancements and the practicality of implementing these recommendations without disrupting existing cryptographic operations. The value of this enhancement lies in its potential to maintain RSA's effectiveness, but it requires a sophisticated implementation strategy that ensures seamless integration with existing systems and does not impose excessive computational overhead.
	
	\paragraph{Challenges in LLM-driven Predictive Analysis}
	While LLM-driven predictive analysis for dynamic key length adjustment in RSA offers a promising avenue for proactive security, several challenges need to be addressed. The accuracy of such predictive models is paramount; incorrect predictions could either lead to unnecessary computational burdens or, worse, insufficient encryption strength against future threats. Implementing these models within the RSA framework requires not just technical adjustments but also a strategic approach to data collection and analysis, ensuring that predictions are based on comprehensive and up-to-date information. The practicality and value of this approach hinge on achieving a balance between foresight and adaptability, ensuring RSA remains robust in the face of computational advancements.
	
	\paragraph{Integrating Predictive Adjustments into RSA}
	The integration of LLM-driven predictive adjustments for key length within the RSA algorithm involves significant technical and logistical considerations. This process must ensure that key length adjustments are made in a timely and secure manner, without compromising the operational efficiency or security of RSA-based systems. The practical challenges include developing algorithms that can seamlessly incorporate LLM recommendations into RSA key generation processes, as well as ensuring that these adjustments are transparent and verifiable. While this approach promises to enhance RSA's adaptability to future threats, its implementation must be carefully managed to maintain the trust and reliability of RSA encryption.
	
	\paragraph{Balancing Performance with Security Enhancements}
	The task of dynamically adjusting RSA key lengths with LLM assistance involves a delicate balance between enhancing security and maintaining performance. While longer key lengths offer greater security, they also impose higher computational demands. The practical implementation of this enhancement must consider the operational contexts in which RSA is used, ensuring that performance is not unduly impacted, especially in resource-constrained environments. This balance is crucial for the practical applicability of dynamic key length adjustments, requiring a nuanced approach that leverages LLM insights without compromising the efficiency and usability of RSA encryption.
	
	\paragraph{Future Directions in Achieving Quantum Resilience}
	The role of LLMs in guiding RSA towards quantum resilience through dynamic key length adjustment presents an intriguing future direction. As the quantum computing era approaches, the ability of LLMs to predict and adapt to quantum threats in real-time could be pivotal. However, translating this potential into practical, quantum-resistant RSA implementations poses significant challenges, requiring advances in LLM capabilities, quantum computing understanding, and cryptographic research. The opportunity lies in developing a framework that not only anticipates quantum computational advancements but also implements these insights in a way that RSA encryption remains secure and viable in the post-quantum world.
	
	\subsubsection{Encryption Efficiency Optimization}
	
	\paragraph{Tailoring RSA Encryption Strategies with LLM Insights}
	The specific application of Algogens to optimize RSA encryption efficiency involves tailoring encryption strategies based on LLM insights into the nature of the data being encrypted. This approach promises to enhance RSA's performance by adjusting encryption parameters dynamically, ensuring a more efficient processing of varying data types. However, the skepticism here revolves around the complexity of accurately categorizing data types and determining the optimal encryption strategy for each category. The practical implementation of this enhancement requires sophisticated LLM models capable of understanding data at a granular level and a robust framework for applying these insights in real-time without compromising security.
	
	\paragraph{Challenges in LLM-driven Process Customization}
	The customization of RSA encryption processes through LLM analysis faces several challenges, notably the accuracy of data categorization and the determination of encryption parameters. While LLMs can theoretically identify optimal encryption strategies for different data types, the practical application of these insights must navigate the nuances of data sensitivity, regulatory requirements, and computational constraints. The value of LLM-driven process customization lies in its potential to enhance efficiency, but its implementation requires a careful balance between customization and the overarching need for cryptographic security and compliance.
	
	\paragraph{Integrating Efficiency Optimizations into RSA's Workflow}
	Integrating efficiency optimizations into RSA's encryption workflow, driven by LLM recommendations, involves significant algorithmic and operational adjustments. This integration aims to dynamically adjust encryption parameters based on the characteristics of the data, promising to improve RSA's performance. However, the practical challenges include ensuring that these adjustments do not compromise the security or integrity of the encrypted data and that the system can adapt in real-time to varying data types and encryption needs. The implementation strategy must focus on developing flexible, secure algorithms that can seamlessly incorporate LLM insights, enhancing RSA's efficiency without sacrificing its foundational security principles.
	
	\paragraph{Balancing Efficiency with Cryptographic Security}
	The pursuit of enhanced encryption efficiency through LLM-driven optimizations within RSA necessitates a careful balance between performance gains and the maintenance of cryptographic security. While optimizing encryption processes can improve RSA's usability, especially in high-volume or resource-constrained environments, the integrity and security of the encryption must remain paramount. The practical application of these optimizations requires a meticulous approach to algorithm design, ensuring that efficiency gains do not inadvertently introduce vulnerabilities or weaken RSA's security posture. This balance is critical for the practical viability of efficiency optimizations, underscoring the need for a cautious, security-first approach to integrating LLM recommendations into RSA's encryption processes.
	
	\paragraph{Adapting to Future Encryption Needs}
	As RSA looks to adapt to the evolving needs of digital encryption, leveraging LLMs for efficiency optimization presents a forward-looking strategy. This approach aims to tailor RSA encryption to the diverse requirements of modern data types and communication protocols, enhancing both performance and security. However, the practical translation of this strategy into RSA implementations must account for the rapidly changing landscape of digital threats and the continuous advancements in encryption technology. The future of RSA encryption, augmented by LLM-driven efficiency optimizations, hinges on the ability to innovate and adapt, ensuring that RSA remains a robust, versatile tool for secure communication in the digital age.
	
	\subsubsection{Security Vulnerability Prediction}
	
	\paragraph{Proactive Defense Through LLM-driven Threat Analysis}
	In applying Algogens specifically to RSA's security vulnerability prediction, the focus shifts towards leveraging LLMs for proactive defense through advanced threat analysis. This approach involves LLMs analyzing vast datasets to identify emerging threats and potential vulnerabilities within RSA implementations. While the concept is promising, skepticism arises from the models' ability to accurately predict threats that have not yet been observed or to adapt quickly to the rapid evolution of cyber threats. The practical implementation of this enhancement requires not just sophisticated LLM models but also a comprehensive framework for integrating their insights into RSA's security protocols, ensuring that predictions translate into actionable security measures without causing unnecessary alarms or overlooking subtle vulnerabilities.
	
	\paragraph{Challenges in LLM-driven Vulnerability Identification}
	The integration of LLM-driven vulnerability identification into RSA's security practices faces several challenges. The accuracy and reliability of LLM predictions are paramount, as incorrect identification could either lead to overlooked vulnerabilities or unnecessary resource allocation to non-issues. Implementing these insights within RSA's framework involves developing mechanisms for the timely and secure application of recommended security measures, ensuring that LLM-driven predictions effectively enhance RSA's resilience against attacks. The value of this approach lies in its potential to preemptively strengthen RSA's security posture, but its practical application demands rigorous validation and a flexible, responsive security infrastructure.
	
	\paragraph{Integrating Predictive Security Insights into RSA Practices}
	Integrating LLM-driven predictive security insights into RSA's operational framework requires a multifaceted approach, combining advanced AI analytics with traditional cryptographic security practices. This integration aims to enable RSA systems to dynamically adjust security measures based on predictive analyses, potentially enhancing their ability to preempt and mitigate emerging threats. However, the practical challenges include ensuring that these integrations do not compromise the fundamental security principles of RSA and that the system can respond in real-time to evolving threat landscapes. The implementation strategy must focus on developing secure, adaptive algorithms and processes that can incorporate LLM insights without introducing new vulnerabilities into the RSA framework.
	
	\paragraph{Enhancing RSA's Resilience Against Emerging Threats}
	The goal of enhancing RSA's resilience against emerging threats through LLM-driven security vulnerability prediction involves a strategic blend of AI-driven insights and cryptographic rigor. This approach promises to provide RSA implementations with a proactive defense mechanism, potentially identifying and mitigating vulnerabilities before they can be exploited. However, skepticism surrounds the capacity of LLMs to accurately predict future threats and the practicality of translating these predictions into effective security measures. The practical implementation of this enhancement requires a robust, agile framework that can swiftly adapt to LLM recommendations, ensuring that RSA's security is continuously aligned with the latest threat intelligence and cryptographic standards.
	
	\paragraph{Future Directions in Security Vulnerability Prediction}
	Looking ahead, the role of LLMs in RSA's security vulnerability prediction presents significant opportunities for advancing cryptographic security. The continuous evolution of cyber threats necessitates innovative approaches to defense, and LLM-driven predictive analysis could provide RSA with a cutting-edge tool for anticipating and neutralizing threats. However, the future effectiveness of this strategy depends on advancements in LLM technology, improvements in threat intelligence analysis, and the cryptographic community's ability to integrate these insights into practical, resilient security measures. The ongoing development of LLM-driven security vulnerability prediction promises to not only enhance RSA's defensive capabilities but also to redefine the landscape of cryptographic security in the face of evolving digital threats.
	
	\subsubsection{Adaptive Protocol Selection}
	
	\paragraph{Intelligent Selection of Cryptographic Protocols with LLMs}
	In the realm of RSA, the specific application of Algogens for adaptive protocol selection involves utilizing LLMs to intelligently determine when RSA is the most suitable encryption method based on the security requirements, operational context, and potential threats. While the concept of dynamically selecting cryptographic protocols based on LLM recommendations is promising, skepticism arises regarding the models' ability to accurately assess the nuanced trade-offs between different encryption methods. The practical implementation of this enhancement requires a sophisticated understanding of both cryptographic principles and machine learning, ensuring that LLM recommendations enhance RSA's applicability without compromising security or efficiency.
	
	\paragraph{Challenges in LLM-driven Decision Making for Protocol Selection}
	The integration of LLM-driven decision-making processes for adaptive protocol selection in RSA encryption systems faces several challenges. The accuracy and reliability of LLM recommendations are crucial, as incorrect protocol selection could compromise security or efficiency. Implementing these decisions within the RSA framework involves developing algorithms and interfaces that can seamlessly incorporate LLM insights, ensuring that the selection process is transparent, verifiable, and aligned with the overarching security goals. The value of this approach lies in its potential to optimize RSA's performance and security across varying contexts, but its practical application demands careful consideration of the complexities involved in cryptographic protocol selection.
	
	\paragraph{Seamless Integration of Adaptive Protocol Selection into RSA Systems}
	Integrating adaptive protocol selection into RSA encryption systems, guided by LLM recommendations, requires a strategic approach that balances innovation with the fundamental requirements of cryptographic security. This integration aims to enhance RSA's flexibility and applicability by dynamically adjusting to the optimal encryption protocol based on contextual analysis. However, the practical challenges include ensuring that these adjustments are made securely and efficiently, without introducing vulnerabilities or performance bottlenecks. The implementation strategy must focus on developing robust, adaptable cryptographic frameworks that can effectively utilize LLM insights to optimize protocol selection, enhancing RSA's utility in diverse operational scenarios.
	
	\paragraph{Balancing Security and Performance in Protocol Selection}
	The task of balancing security and performance in the selection of cryptographic protocols for RSA, guided by LLM recommendations, involves a nuanced understanding of the trade-offs involved. While adaptive protocol selection promises to optimize RSA's application based on the specific requirements of each use case, skepticism arises from the complexity of accurately assessing and implementing these trade-offs. The practical implementation of this enhancement requires a careful approach that prioritizes cryptographic security while seeking efficiency gains, ensuring that LLM-driven protocol selection enhances RSA's effectiveness without compromising its foundational security principles.
	
	\paragraph{Navigating Future Cryptographic Challenges with Adaptive RSA}
	Looking towards the future, the application of Algogens for adaptive protocol selection in RSA presents a forward-looking strategy for navigating the evolving landscape of cryptographic challenges. The integration of LLM-driven insights into RSA's protocol selection process promises to provide a dynamic, responsive framework capable of adapting to new threats, computational advances, and encryption needs. However, the practical realization of this strategy requires ongoing innovation, interdisciplinary collaboration, and a commitment to maintaining the security, integrity, and performance of RSA encryption. The future of adaptive RSA, guided by the intelligent analysis and recommendations of LLMs, hinges on the cryptographic community's ability to leverage these advancements while upholding the highest standards of digital security.
	
	\subsubsection{Automated Security Audits}
	
	\paragraph{Elevating RSA Security with Comprehensive Automated Reviews}
	The specific application of Algogens to RSA's automated security audits involves leveraging LLMs to conduct comprehensive, automated reviews of RSA implementations. This approach promises to enhance the security and compliance of RSA-based systems by identifying vulnerabilities, deviations from best practices, and areas for improvement. However, skepticism arises from the models' ability to accurately and comprehensively assess complex cryptographic implementations. The practical implementation of this enhancement requires not just advanced LLM capabilities but also a robust framework for integrating audit findings into RSA's security practices, ensuring that automated reviews translate into actionable, effective security enhancements.
	
	\paragraph{Challenges in Implementing LLM-based Automated Security Audits}
	Implementing LLM-based automated security audits within the RSA framework presents several challenges. The accuracy and depth of LLM-driven audits are critical, as incomplete or incorrect assessments could lead to overlooked vulnerabilities or misdirected resources. Integrating the findings of these audits into RSA's security practices involves developing processes for the timely and effective application of recommended improvements, ensuring that LLM-driven audits genuinely enhance RSA's security posture. The value of automated security audits lies in their potential to provide ongoing, in-depth assessments, but their practical application demands careful consideration of the complexities involved in cryptographic security auditing.
	
	\paragraph{Integrating Comprehensive Security Insights into RSA Development}
	Integrating comprehensive security insights from automated LLM-driven audits into RSA's development and deployment workflows requires a strategic approach that ensures these insights are effectively utilized to enhance security. This integration aims to embed a culture of continuous improvement within RSA's development processes, leveraging LLM capabilities to identify and mitigate vulnerabilities proactively. However, the practical challenges include ensuring that the integration of audit findings does not disrupt development workflows and that recommended security enhancements are implemented in a manner that strengthens RSA's overall security architecture. The implementation strategy must focus on developing mechanisms for seamless, secure incorporation of LLM-driven insights into RSA's development lifecycle, promoting a proactive stance towards cryptographic security.
	
	\paragraph{Advancing RSA's Compliance and Security Posture through Automated Audits}
	The application of automated security audits to RSA, powered by LLMs, offers a promising avenue for advancing RSA's compliance with cryptographic standards and enhancing its overall security posture. By providing a mechanism for continuous, in-depth security evaluations, automated audits can help ensure that RSA implementations adhere to evolving security protocols and best practices. However, skepticism surrounds the ability of automated audits to fully capture the complexities of RSA encryption and the dynamic nature of cybersecurity threats. The practical implementation of this enhancement requires a robust, adaptable framework that can integrate LLM-driven insights into RSA's security practices, ensuring that automated audits contribute to a proactive, resilient approach to cryptographic security.
	
	\paragraph{Future Directions in Enhancing RSA Security with Automated Audits}
	Looking ahead, the role of automated security audits in enhancing RSA's security framework presents significant opportunities for advancing cryptographic practices. The integration of LLM-driven insights into RSA's security audits promises to provide a dynamic, intelligent mechanism for identifying and mitigating vulnerabilities, adapting to new threats, and ensuring compliance with cryptographic standards. However, the future effectiveness of automated security audits depends on advancements in LLM technology, the development of sophisticated audit methodologies, and the cryptographic community's ability to integrate these insights into practical, resilient security measures. The ongoing evolution of automated security audits, powered by the analytical capabilities of LLMs, promises to not only enhance the security of RSA implementations but also to redefine the standards of cryptographic security in the digital age.

	\subsubsection{Pseudocode for Algogenic RSA}
	The Algogenic RSA approach utilizes AI to enhance the traditional RSA encryption and decryption methods by dynamically adjusting encryption parameters, optimizing key generation, and ensuring robust security measures based on real-time data and predictive analysis. This pseudocode outlines an advanced framework incorporating AI-driven enhancements for optimized prime number generation, dynamic key length adjustment, encryption efficiency optimization, and comprehensive security vulnerability predictions. See figure \ref{rsa-Algogen-pseudocode}.
	
	\begin{algorithm}
		\caption{Algogenic RSA Encryption and Decryption Pseudocode}
		\begin{algorithmic}[1]
			\Procedure{AlgogenicRSA}{}
			
			\Comment{Preprocessing Phase}
			\State Analyze numerical data with LLM for optimized prime number generation.
			\State Adjust RSA key length dynamically based on LLM insights.
			
			\Comment{Key Generation Phase}
			\State Generate prime numbers $p$ and $q$ with LLM-guided optimization.
			\State Compute $n = p \times q$ and $\phi(n) = (p-1) \times (q-1)$.
			\State Choose $e$ such that $1 < e < \phi(n)$ and $gcd(e, \phi(n)) = 1$.
			\State Compute $d$ to be the modular multiplicative inverse of $e$ mod $\phi(n)$.
			
			\Comment{Encryption Phase}
			\If{Encrypting large or diverse data types}
			\State Apply LLM-guided efficiency optimization to encryption process.
			\EndIf
			\State Encrypt message $M$ using public key $(n, e)$ to get ciphertext $C$.
			
			\Comment{Decryption Phase}
			\State Decrypt ciphertext $C$ using private key $(n, d)$ to retrieve message $M$.
			
			\Comment{Postprocessing Phase}
			\State Perform LLM-driven security vulnerability prediction.
			\State Select cryptographic protocol adaptively with LLM recommendations.
			\State Conduct automated security audits using LLM insights.
			
			\EndProcedure
		\end{algorithmic}\label{rsa-Algogen-pseudocode}
	\end{algorithm}

	\begin{figure}
		\centering
		\includegraphics[width=0.7\textwidth]{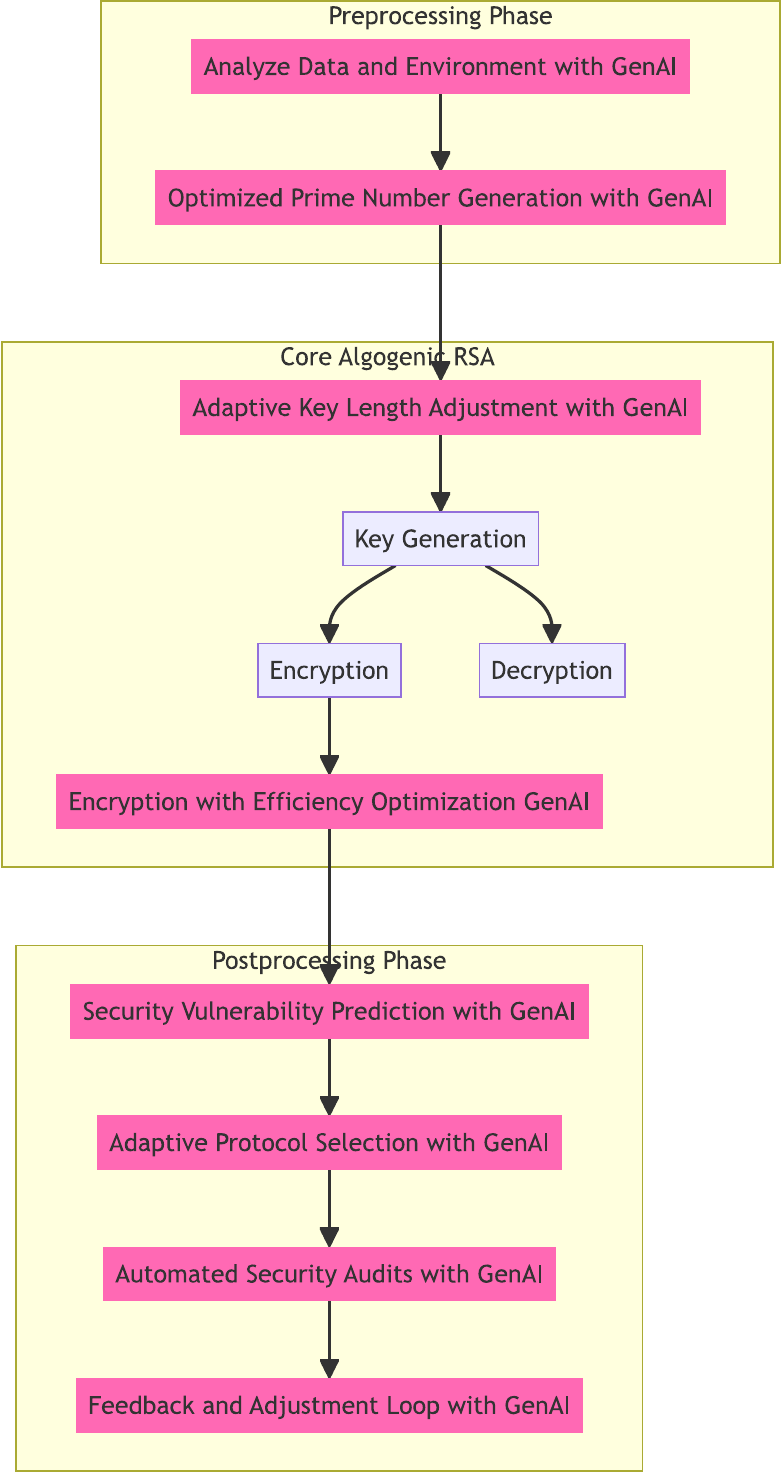}
		\caption{Enhancing RSA with Algogenic Approaches: This figure showcases the integration of Algogenic enhancements with the traditional RSA encryption algorithm. It highlights the innovative application of Large Language Models to optimize prime number generation, dynamically adjust key lengths, and optimize encryption efficiency for diverse data types. Additionally, it illustrates the proactive role of LLMs in security vulnerability prediction and the adaptive selection of cryptographic protocols. These enhancements not only bolster the RSA algorithm's security and efficiency but also ensure its adaptability to evolving computational and cryptographic landscapes.}
		\label{fig:rsa}
	\end{figure}

	\section{Apriori}\index{Apriori}
	\subsection{Introduction to the Apriori Algorithm}
	\subsubsection{The Concept of the Apriori Algorithm}
	
	\paragraph{Introduction to Apriori}
	The Apriori Algorithm stands as a foundational approach in the field of data mining for discovering frequent itemsets in large databases and for deriving association rules. Introduced by Agrawal and Srikant in 1994, this algorithm has paved the way for analyzing vast datasets to uncover patterns, correlations, and structures that inform decision-making processes in retail, marketing, and beyond. Its name, "Apriori," signifies the core principle it leverages: the algorithm proceeds with the assumption (a priori) that any subset of a frequent itemset must also be frequent.
	
	Moreover, the Apriori Algorithm operates in a series of iterations, progressively extending the length of itemsets until no further expansion is possible or until a user-defined threshold is met. This iterative process efficiently prunes the search space by eliminating candidate itemsets that do not meet the minimum support threshold, thereby reducing computational overhead. Additionally, Apriori employs a breadth-first search strategy, systematically exploring itemsets of increasing size, which contributes to its scalability in handling large databases.
	
	Furthermore, association rule mining, a key task enabled by the Apriori Algorithm, uncovers relationships between items based on their co-occurrence in transactions. These association rules, often represented in the form of "if-then" statements, provide valuable insights into consumer behavior, allowing businesses to optimize product placement, promotions, and inventory management strategies.

	\paragraph{Operational Mechanics}
	The Apriori Algorithm, a fundamental technique in association rule mining, employs an iterative approach pivotal in extracting frequent itemsets from large datasets efficiently. Initially, the algorithm scrutinizes the dataset to identify the frequency of individual items, laying the groundwork for subsequent iterations. This foundational step is crucial as it establishes a baseline comprehension of item occurrence, essential for determining significance. By setting a minimum support threshold, often user-defined, the algorithm discerns items that meet the criteria for being labeled as frequent itemsets.
	
	Subsequently, leveraging this foundational knowledge, the algorithm proceeds iteratively, systematically extending the identified itemsets. This process involves amalgamating smaller itemsets to form larger ones, thereby broadening the scope of exploration. At each iteration, the algorithm meticulously evaluates the frequency of these enlarged itemsets against the dataset, a pivotal step in discerning their significance and relevance.
	
	Integral to the efficiency of the Apriori Algorithm is its adept pruning mechanism. By swiftly eliminating non-frequent itemsets at each iteration, the algorithm discerns a streamlined path, focusing computational resources solely on plausible candidates. This pruning strategy is instrumental in curtailing computational overhead, confining the exploration to itemsets with the potential to meet the stipulated support threshold.
	
	Consequently, the Apriori Algorithm orchestrates a meticulous balance between exploration and exploitation. Through iterative refinement and judicious pruning, it navigates the vast landscape of potential itemsets, culminating in the extraction of meaningful associations. This iterative approach, coupled with strategic pruning, underpins the algorithm's efficacy in discerning patterns amidst voluminous datasets, making it a cornerstone in data mining and exploratory analysis.

	\paragraph{Generation of Association Rules}
	Following the identification of frequent itemsets, Apriori proceeds to generate association rules that predict the occurrence of an item based on the presence of other items within the same transaction. These rules are evaluated based on their support, indicating how frequently the itemset appears in the database, and their confidence, which measures the likelihood of the rule being correct. The Apriori Algorithm's strength lies in its simplicity and robustness, enabling the discovery of meaningful patterns within transactional datasets that are otherwise hidden in the volume of data.
	
	Moreover, the process of generating association rules serves as a crucial step in data mining, providing insights into the relationships between different items purchased together. This allows businesses to understand customer behavior better and tailor marketing strategies accordingly. Additionally, association rules can uncover dependencies between items, facilitating inventory management and cross-selling opportunities. Furthermore, by examining both support and confidence metrics, analysts can filter out spurious associations and focus on those with significant relevance. Hence, Apriori's ability to sift through vast datasets and extract actionable insights makes it a valuable tool for businesses seeking to optimize their operations and enhance decision-making processes.

	\paragraph{Algorithmic Efficiency and Challenges}
	While Apriori is celebrated for its straightforward approach to pattern recognition, its efficiency is closely tied to the size of the dataset and the density and diversity of itemsets within. The algorithm's performance can be hindered by the necessity to scan the database multiple times, a challenge that becomes more pronounced with the increase in data volume and itemset complexity. Moreover, the generation of candidate itemsets can lead to exponential growth in computational requirements, necessitating optimizations and enhancements to maintain scalability and practicality in real-world applications. However, despite these challenges, Apriori remains a foundational algorithm in association rule mining due to its simplicity and effectiveness in discovering frequent itemsets. Additionally, advancements in parallel computing architectures and distributed computing frameworks offer opportunities to mitigate the computational burden associated with Apriori's iterative nature. Furthermore, the integration of pruning strategies and data reduction techniques can help alleviate the exponential growth of candidate itemsets, thereby improving the algorithm's efficiency and scalability. Nevertheless, the choice of minimum support threshold and the handling of sparse datasets continue to be areas of research focus, as they significantly impact Apriori's performance and applicability across different domains and use cases.

	\paragraph{Applications and Impact}
	Despite these challenges, the Apriori Algorithm's contribution to data mining and knowledge discovery remains significant. Its applications range from market basket analysis, where it helps retailers understand purchase patterns, to bioinformatics, fraud detection, and recommendation systems. By unveiling the underlying associations between items in large datasets, Apriori helps transform raw data into actionable insights, driving strategic decisions and fostering innovative solutions across various industries. Furthermore, in the realm of market basket analysis, Apriori serves as a cornerstone tool, enabling businesses to comprehend consumer behavior by identifying frequently co-occurring items in transactions. Moreover, in bioinformatics, the algorithm aids in genomic data analysis, facilitating the discovery of meaningful associations between genes and phenotypes, thus advancing our understanding of biological systems. Likewise, in fraud detection, Apriori assists financial institutions in identifying suspicious patterns of transactions, mitigating risks, and safeguarding against fraudulent activities. Additionally, in recommendation systems, it enhances user experience by analyzing past preferences and suggesting relevant items or content, thereby improving customer satisfaction and engagement. Consequently, the widespread adoption of Apriori underscores its pivotal role in leveraging data-driven insights to drive innovation, optimize processes, and gain competitive advantages in today's dynamic business landscape.

	\subsubsection{Key Principles and Mechanisms}
	
	\paragraph{Foundation of Frequent Itemset Generation}
	The Apriori Algorithm, a pioneering method in data mining, rests upon two fundamental principles: the identification of frequent itemsets and the subsequent generation of association rules from these itemsets. At its core, the algorithm embarks on the journey of discovering patterns within vast datasets, rendering it indispensable in various domains like market basket analysis and recommendation systems.
	
	The initial step of the algorithm entails the meticulous determination of itemsets that manifest with a frequency surpassing a predefined threshold, aptly termed the minimum support. This threshold serves as a filter, sieving out itemsets that lack significance in the dataset. By focusing solely on itemsets that meet this criterion, Apriori ensures that the ensuing patterns are not mere artifacts of chance but rather indicative of meaningful associations.
	
	The significance of identifying frequent itemsets cannot be overstated, as they serve as the cornerstone for extracting actionable insights from data. These itemsets encapsulate recurring combinations of items, shedding light on the intrinsic relationships between different elements within the dataset. Consequently, the subsequent generation of association rules from these frequent itemsets becomes a streamlined process, facilitating the extraction of valuable knowledge from the data.
	
	Moreover, the concept of support extends beyond mere frequency; it embodies the notion of significance within the dataset. As such, by setting an appropriate minimum support threshold, analysts can fine-tune the sensitivity of the algorithm, tailoring it to unearth patterns that are both relevant and meaningful within the context of their specific application.

	\paragraph{The Apriori Principle}
	At the heart of the algorithm's efficiency is the Apriori principle, which posits that all subsets of a frequent itemset must also be frequent. This principle serves as the cornerstone for pruning the search space effectively, leading to substantial computational savings. By leveraging the Apriori principle, the algorithm strategically avoids exploring itemsets that are unlikely to meet the minimum support threshold, thus circumventing the need for exhaustive computations.
	
	Furthermore, the Apriori principle enables a systematic approach to itemset generation and evaluation. It provides a clear guideline for determining the potential frequency of itemsets based on the frequency of their subsets. Consequently, the algorithm can prioritize its search efforts on promising itemsets, streamlining the process of identifying frequent itemsets within large datasets.
	
	Moreover, the principle's bidirectional nature ensures that not only frequent itemsets but also infrequent ones are efficiently identified. When an itemset is determined to be infrequent, the Apriori principle guarantees that all of its supersets are also infrequent. This eliminates the need to evaluate each individual itemset separately, as their infrequency propagates upwards through the hierarchy of itemsets.
	
	Thus, by adhering to the Apriori principle, the algorithm navigates through the combinatorial explosion of possible itemsets with precision and efficiency. It optimizes the utilization of computational resources by focusing on the most promising candidates, ultimately accelerating the process of association rule mining and enabling the extraction of meaningful insights from transactional data.

	\paragraph{Iterative Approach and Candidate Generation}
	Apriori operates through an iterative level-wise search, where $k$-itemsets are used to explore $(k+1)$-itemsets. This systematic progression is crucial for identifying frequent itemsets efficiently. At the outset, the algorithm initializes with frequent individual items, which serve as the foundation for subsequent iterations. \textbf{Moreover}, by leveraging the knowledge of frequent $k$-itemsets, the algorithm strategically generates candidate itemsets of size $(k+1)$. This incremental approach ensures that the search space is efficiently traversed, mitigating computational complexity. 
	
	During each iteration, \textbf{the algorithm} meticulously crafts candidate itemsets, employing various pruning techniques to optimize the process. \textbf{However}, not all candidate itemsets survive the rigorous evaluation process. Those failing to surpass the predefined minimum support threshold are promptly discarded, \textbf{thus} focusing computational resources on the most promising candidates. \textbf{Consequently}, this judicious pruning mechanism prevents unnecessary computations, contributing to the overall efficiency of the algorithm.
	
	After candidate generation, the algorithm performs a database scan to count the occurrences of these candidates, a step \textbf{critical} for determining their support. This meticulous evaluation ensures that only the most relevant itemsets progress to subsequent iterations, \textbf{thus} streamlining the search process. By systematically expanding from frequent $k$-itemsets to explore $(k+1)$-itemsets, Apriori optimizes the discovery of frequent itemsets \textbf{accordingly}, facilitating efficient association rule mining in large datasets.

	\paragraph{Rule Generation and Evaluation}
	Once frequent itemsets are identified, Apriori shifts focus to generating association rules, which are implications of the form $X \Rightarrow Y$, where $X$ and $Y$ are disjoint itemsets. The confidence of each rule, defined as the ratio of the support of the union of $X$ and $Y$ to the support of $X$, is calculated to measure the rule's strength. \textbf{Moreover}, only rules meeting a user-defined minimum confidence threshold are considered significant. This mechanism ensures that derived rules are not only frequent across the dataset but also possess a strong predictive power.
	
	The association rule mining process involves not just identifying frequent itemsets but also evaluating the strength of associations between items. \textbf{Consequently}, by measuring the confidence of each rule, analysts can assess the likelihood of the consequent item(s) appearing in transactions given the presence of the antecedent item(s). This evaluation step is crucial in filtering out spurious associations and identifying meaningful patterns in the data. \textbf{Furthermore}, the user-defined confidence threshold allows for a flexible approach, enabling analysts to focus on the most reliable rules based on their specific application requirements.
	
	In addition to confidence, other metrics such as lift and conviction can also be used to evaluate the quality of association rules. \textbf{However}, confidence remains a fundamental measure, providing a straightforward indication of the predictive power of the rules. By setting a minimum confidence threshold, analysts can control the trade-off between rule significance and rule abundance, ensuring that only the most reliable rules are considered for further analysis and interpretation. This selective approach enhances the efficiency and effectiveness of association rule mining in discovering actionable insights from transactional data.

	\paragraph{Optimizations and Scalability Concerns}
	Despite its foundational role in association rule mining, the Apriori Algorithm faces challenges related to scalability and efficiency, particularly with large datasets and a high number of itemsets. One key strategy to mitigate these challenges is to reduce the number of database scans required during the algorithm's execution. This can be achieved by employing various techniques such as vertical data format, where the database is transposed to reduce I/O overhead, and horizontal data format, which involves partitioning the database horizontally and processing each partition independently. These optimizations help minimize disk I/O operations, thereby improving the algorithm's scalability.
	
	Another crucial aspect of optimizing the Apriori Algorithm is the efficient management of candidate itemsets. Utilizing compact data structures like hash trees or bitmaps can significantly reduce memory usage and enhance computational efficiency. By storing candidate itemsets in a compressed format, the algorithm can operate more efficiently, especially when dealing with large datasets containing numerous frequent itemsets.
	
	Parallelizing computations is another avenue for addressing scalability concerns. By distributing the workload across multiple processing units or nodes, parallel implementations of the Apriori Algorithm can exploit the computational resources more effectively, leading to faster execution times and improved scalability.
	
	Additionally, variations of the Apriori algorithm aim to refine the candidate generation and pruning processes to enhance overall performance. Techniques such as dynamic itemset counting, where the support count of itemsets is updated incrementally, and hash-based pruning, which efficiently prunes the search space by leveraging hash functions, contribute to better scalability and efficiency.

	The interplay of these principles and mechanisms within Apriori exemplifies the algorithm's capacity to transform extensive transactional data into insightful and actionable patterns, underscoring its enduring relevance in the data mining domain.
	
	\subsubsection{The Role of Support and Confidence in Apriori}
	
	\paragraph{Defining Support and Confidence}
	In the realm of the Apriori Algorithm, support and confidence serve as cornerstone metrics, essential for navigating the intricate landscape of frequent itemset mining and association rule generation. Support stands as a beacon, illuminating the prevalence of a specific itemset within the dataset, thereby offering invaluable insight into its significance. It quantifies the frequency with which an itemset manifests in the transactions, painting a vivid picture of its occurrence relative to the entirety of the database. Computed as the ratio of transactions containing the itemset of interest to the total number of transactions, support lays the foundation upon which the algorithm operates, guiding its pursuit of meaningful patterns amidst the sea of data.
	
	On the parallel track, confidence emerges as a stalwart measure, fortifying the credibility of association rules unearthed during the algorithmic traversal. Like a steadfast compass, confidence navigates the terrain of association rule mining, gauging the strength of relationships between different items. Specifically, it delineates the likelihood that a transaction embracing a certain itemset $X$ will also encompass another item $Y$, thus encapsulating the essence of association between disparate elements. Through meticulous calculation, confidence unveils the reliability of these associations, offering a numerical testament to the trustworthiness of inferred rules. By computing the ratio of the support of the combined itemset $X \cup Y$ to the support of the individual itemset $X$, confidence elucidates the degree of dependency between items, fostering the identification of robust and actionable patterns.
	
	In essence, support and confidence intertwine to form the bedrock of association rule mining, furnishing practitioners with the tools necessary to sift through vast datasets in pursuit of actionable insights. Their symbiotic relationship underpins the efficacy of the Apriori Algorithm, empowering it to navigate the labyrinth of transactional data and extract meaningful associations that underlie the fabric of consumer behavior and market dynamics.

	\paragraph{Application in Frequent Itemset Generation}
	Support plays a critical role in the initial phase of the Apriori Algorithm, where the goal is to identify itemsets that occur frequently across the dataset. By setting a minimum support threshold, users can filter out itemsets that are too rare to be of interest, focusing computational efforts on those itemsets that significantly impact the dataset. This thresholding not only enhances the efficiency of the algorithm by reducing the search space but also ensures that the patterns discovered are relevant and potentially actionable.
	
	Furthermore, the support threshold serves as a mechanism for controlling the trade-off between computational complexity and the quality of discovered patterns. Adjusting this threshold allows users to fine-tune the algorithm's performance according to their specific requirements and constraints. Additionally, by incorporating domain knowledge or business expertise, users can set support thresholds that reflect the significance of certain itemsets in the context of their application domain.
	
	Moreover, the Apriori Algorithm's reliance on support underscores its suitability for association rule mining tasks in various domains, ranging from market basket analysis to biomedical data mining. The algorithm's ability to efficiently handle large datasets while providing interpretable and actionable results has contributed to its widespread adoption in both research and industry settings.
	
	Therefore, understanding the role of support in the Apriori Algorithm is essential for effectively applying this technique to discover meaningful patterns in data, ultimately leading to informed decision-making and improved business outcomes.

	\paragraph{Influence on Association Rule Mining}
	Once frequent itemsets are identified, the Apriori Algorithm employs confidence to generate association rules that predict the occurrence of an item based on the presence of other items in a transaction. By establishing a minimum confidence threshold, the algorithm can discard rules that do not meet the criteria for reliability, thereby ensuring that the resulting rules are strong predictors of item associations. This selective process is crucial for extracting meaningful rules from the vast number of possible combinations present in the frequent itemsets.
	
	Moreover, the utilization of confidence in association rule mining enables the algorithm to sift through the myriad of potential associations and focus only on those with a high likelihood of occurrence. This filtering mechanism, coupled with the establishment of a confidence threshold, ensures that the generated rules have a substantial basis in the underlying data. Furthermore, the incorporation of confidence allows for the identification of relationships between items that exhibit consistent behavior across transactions, thus enhancing the robustness of the association rules generated by the algorithm.
	
	Additionally, confidence serves as a measure of the strength of association between items, enabling practitioners to prioritize rules that are most likely to be actionable in real-world scenarios. By emphasizing rules with higher confidence levels, the Apriori Algorithm aids decision-making processes by highlighting associations that are statistically significant and reliable. Furthermore, the establishment of a confidence threshold empowers users to customize the trade-off between rule quality and quantity, allowing for flexibility in rule generation based on specific application requirements.
	
	Consequently, the integration of confidence into the Apriori Algorithm fundamentally shapes the nature of the association rules produced, ensuring that only the most meaningful and reliable rules are retained for further analysis and interpretation. This refinement process enhances the efficacy of association rule mining by enabling practitioners to focus their attention on actionable insights derived from the data.

	\paragraph{Balancing Support and Confidence}
	The interplay between support and confidence in Apriori highlights the delicate equilibrium required to navigate between two essential facets: identifying itemsets prevalent enough in the dataset (\textit{support}) and ensuring that resultant association rules possess robust predictive capability (\textit{confidence}). \textbf{Moreover}, setting thresholds too high risks overlooking significant patterns, potentially resulting in missed insights. Conversely, excessively lenient criteria may inundate analysts with an avalanche of rules, diluting the meaningfulness of discovered associations. \textbf{Furthermore}, a judicious balance is imperative as it directly impacts the interpretability and actionable utility of mined rules. \textbf{On the other hand}, overly stringent support thresholds might prune away infrequent but critical associations, while excessively high confidence requirements could yield trivial or spurious rules. \textbf{In contrast}, relaxed confidence thresholds might allow the inclusion of weak associations, potentially cluttering the rule set with noise. Striking a balance \textbf{therefore} necessitates careful consideration of the specific domain, dataset characteristics, and the overarching objectives of the analysis. \textbf{As a result}, practitioners must employ iterative refinement, adjusting support and confidence thresholds iteratively to achieve a harmonious blend of comprehensiveness and relevance in the derived association rules.

	\paragraph{Challenges and Strategic Considerations}
	Choosing appropriate thresholds for support and confidence is a non-trivial task that requires domain knowledge and an understanding of the dataset's characteristics. Moreover, the reliance on these thresholds introduces challenges in terms of scalability and computational efficiency, as varying these parameters can significantly impact the number of candidate itemsets generated and the subsequent workload for rule evaluation. Additionally, strategic considerations, such as incremental threshold adjustment and the exploration of alternative metrics like lift, which considers the rule's improvement over the baseline probability of the consequent, are employed to navigate these challenges, enhancing the algorithm's applicability and effectiveness in uncovering meaningful patterns in data. Furthermore, the incorporation of domain expertise in threshold determination mitigates the risk of overlooking important associations due to overly stringent or lenient criteria. By adjusting thresholds iteratively based on feedback from domain experts or through automated methods, the algorithm can adapt to the evolving nature of the data and the underlying business objectives. Hence, a dynamic approach to threshold setting is crucial for maintaining the relevance and reliability of the association rule mining process in various application domains.

	\subsubsection{Applications and Limitations}
	
	\paragraph{Diverse Applications Across Domains}
	The Apriori Algorithm, with its ability to unearth frequent itemsets and derive association rules, finds applications across a multitude of domains. In retail and market basket analysis, it helps businesses identify products that frequently co-occur in transactions, guiding marketing strategies such as product placement and cross-selling opportunities. In the healthcare sector, Apriori can analyze patient data to find associations between symptoms and diagnoses, aiding in the development of diagnostic tools and treatment plans. Moreover, in web usage mining, it assists in understanding user navigation patterns, improving website design and personalized content delivery. The algorithm's adaptability is particularly evident in its ability to handle various data types, from transactional records in retail to patient profiles in healthcare and clickstream data on the web. This versatility underscores its significance as a fundamental tool in data mining and machine learning. Furthermore, Apriori's capacity to efficiently mine frequent itemsets and association rules from large datasets enables businesses and organizations to derive actionable insights swiftly and accurately. These applications demonstrate Apriori's versatility in extracting insights from large datasets, providing a foundation for data-driven decision-making across industries.

	\paragraph{Limitations and Computational Challenges}
	Despite its widespread utility, the Apriori Algorithm encounters limitations, primarily due to its computational complexity and efficiency concerns. The requirement to scan the database multiple times to compute the support of itemsets can become a bottleneck, especially with large datasets. Additionally, the exponential growth in the number of candidate itemsets as the itemset size increases poses significant challenges in terms of memory usage and processing time. 
	
	Furthermore, while the algorithm's basic structure relies on a simple breadth-first search strategy, the need to generate and evaluate numerous candidate itemsets can lead to a combinatorial explosion, exacerbating the computational burden. Moreover, the algorithm's performance may degrade when dealing with sparse datasets, where the majority of itemsets have low support, leading to inefficiencies in memory utilization and computational resources.
	
	However, advancements in parallel and distributed computing paradigms offer promising avenues to mitigate these challenges. Techniques such as parallelizing the candidate generation process or employing distributed computing frameworks like MapReduce can enhance the algorithm's scalability and efficiency, enabling its application to larger datasets with improved performance.
	
	Additionally, the integration of pruning strategies, such as the adoption of effective pruning criteria like the "hash-based" pruning technique, can alleviate the computational overhead by reducing the number of candidate itemsets that need to be explored. Moreover, leveraging hardware acceleration techniques, such as utilizing Graphics Processing Units (GPUs) or Field-Programmable Gate Arrays (FPGAs), can further expedite the execution of critical algorithmic components, enhancing overall performance.
	
	Consequently, while the Apriori Algorithm exhibits certain limitations in handling large-scale datasets efficiently, ongoing research efforts focusing on algorithmic optimizations, parallelization, and hardware acceleration hold promise for addressing these challenges and enhancing its applicability in real-world scenarios.

	\paragraph{Efficiency and Scalability Concerns}
	Efficiency and scalability pose significant challenges for the Apriori Algorithm. As datasets expand in both size and complexity, the algorithm's performance may suffer, constraining its utility in the context of burgeoning big data scenarios. Despite its foundational role in association rule mining, Apriori's reliance on frequent itemset generation and candidate pruning leads to increased computational overhead, particularly with larger datasets. Consequently, its practicality diminishes when confronted with modern data volumes characterized by millions or even billions of records.
	
	Moreover, the iterative nature of the algorithm, involving multiple database scans, exacerbates its scalability concerns. Each pass over the dataset incurs computational costs, which become prohibitive as dataset sizes escalate. Consequently, the algorithm struggles to maintain efficiency and effectiveness when confronted with the scale and diversity of contemporary data sources.
	
	To address these challenges, researchers and practitioners have explored alternative approaches, such as the FP-Growth algorithm. By employing a different strategy that constructs an FP-tree to encode transaction information, FP-Growth mitigates the need for repeated database scans, thereby enhancing efficiency and scalability. This improvement is particularly pronounced in scenarios with large, sparse datasets, where FP-Growth's compact data structure and streamlined processing offer substantial performance gains over traditional Apriori implementations.
	
	In essence, while Apriori laid the groundwork for association rule mining, its scalability limitations necessitate adaptation and innovation to meet the demands of modern data analytics. The evolution from Apriori to FP-Growth exemplifies this imperative, demonstrating how advancements in algorithmic design can unlock new possibilities for analyzing vast and diverse datasets.

	\paragraph{Adapting to Modern Data Needs}
	The evolution of data mining techniques continues to push the boundaries of what algorithms like Apriori can achieve. Integration with machine learning models and the incorporation of parallel processing and cloud computing resources are among the strategies being explored to enhance Apriori's capability to handle big data. 
	
	Moreover, these adaptations seek to overcome the inherent limitations of the original algorithm, which primarily revolved around its scalability and efficiency in processing vast amounts of data. By integrating machine learning models, such as neural networks or decision trees, Apriori can benefit from enhanced pattern recognition capabilities, enabling it to extract more meaningful and complex association rules from large datasets. 
	
	Furthermore, the incorporation of parallel processing and cloud computing resources addresses the computational challenges posed by big data. Parallelizing the execution of the algorithm across multiple processing units allows for faster execution times and improved scalability, ensuring that Apriori remains efficient even when dealing with massive datasets distributed across various nodes in a cloud infrastructure.
	
	Additionally, these adaptations expand Apriori's application domain, enabling it to tackle a broader range of data mining tasks beyond traditional association rule mining. For example, by leveraging machine learning techniques, Apriori can be applied to tasks such as classification, regression, and clustering, further capitalizing on the wealth of information contained within large datasets.
	
	Thus, by embracing these advancements in technology and methodology, Apriori evolves from a traditional association rule mining algorithm into a versatile tool capable of meeting the diverse and evolving data mining needs of modern applications.

	\paragraph{Conclusion}
	In summary, while the Apriori Algorithm has proven to be a valuable tool in the discovery of association rules and frequent itemsets, its applications are balanced by considerations of computational efficiency and scalability. The Apriori Algorithm, with its bottom-up approach, efficiently identifies frequent itemsets by utilizing the downward closure property, which guarantees that if an itemset is infrequent, all of its supersets are also infrequent. However, this efficiency comes at the cost of high memory consumption and computational overhead, particularly when dealing with large datasets or itemsets with high cardinality. Additionally, the algorithm's reliance on multiple passes over the dataset can lead to performance degradation in real-time or streaming scenarios. Nevertheless, the ongoing development of algorithmic enhancements, such as advanced pruning techniques and parallelization strategies, aims to mitigate these limitations and improve the algorithm's scalability. Furthermore, the exploration of alternative methodologies, including tree-based approaches like FP-Growth and emerging techniques like deep learning-based association rule mining, underscores the dynamic nature of the data mining field. These efforts reflect a continuous endeavor to adapt and optimize foundational techniques like the Apriori Algorithm to meet the evolving demands of data analysis in various domains, ensuring their relevance and effectiveness in contemporary data-driven applications.

	\subsubsection{Pseudocode for the Apriori Algorithm}
	The Apriori Algorithm is a systematic approach developed to identify frequent itemsets within a dataset and derive association rules from them. It operates by iteratively exploring candidate itemsets, starting from the initial set ($C_1$) comprising all unique items found in the transactions. With each iteration, the algorithm incrementally increases the size of itemsets ($k$) to be examined, filtering out those that fail to meet the predefined minimum support threshold. This iterative refinement process persists until no new frequent itemsets can be discovered. Following this, association rules are generated from the aggregated frequent itemsets, considering only those rules that satisfy a specified minimum confidence level. Apriori's efficiency stems from its utilization of support and confidence thresholds, which effectively prune the search space, concentrating computational efforts on promising itemsets and rules. This streamlined approach facilitates the identification of significant patterns within transactional datasets. The operational details of the Apriori Algorithm are outlined in pseudocode \ref{fig:apriori-pseudocode}.
	
	\begin{algorithm}
		\caption{Apriori Algorithm Pseudocode}
		\begin{algorithmic}[1]
			\Procedure{Apriori}{Transactions, minSupport, minConfidence}
			\State $C_1 \gets$ GenerateCandidateItemsets($1$)
			\State $L_1 \gets$ FilterCandidatesBySupport($C_1$, minSupport)
			\State $k \gets 2$
			\While{not $L_{k-1}$ is empty}
			\State $C_k \gets$ GenerateCandidateItemsets($k$, $L_{k-1}$)
			\State $L_k \gets$ FilterCandidatesBySupport($C_k$, minSupport)
			\State $k \gets k + 1$
			\EndWhile
			\State $AllFrequentItemsets \gets$ Aggregate($L_1, L_2, ..., L_{k-1}$)
			\State $Rules \gets$ GenerateAssociationRules($AllFrequentItemsets$, minConfidence)
			\State \Return $Rules$
			\EndProcedure
			
			\Function{GenerateCandidateItemsets}{$k$, $L_{k-1}$}
			\If{$k = 1$}
			\State \Return All unique items in Transactions
			\Else
			\State \Return GenerateNewCandidates($L_{k-1}$)
			\EndIf
			\EndFunction
			
			\Function{FilterCandidatesBySupport}{$C_k$, minSupport}
			\State $L_k \gets$ Items in $C_k$ with support $\geq$ minSupport
			\State \Return $L_k$
			\EndFunction
			
			\Function{GenerateAssociationRules}{$AllFrequentItemsets$, minConfidence}
			\State $Rules \gets$ []
			\For{each itemset in $AllFrequentItemsets$}
			\State $Rules \gets Rules +$ DeriveRules(itemset, minConfidence)
			\EndFor
			\State \Return $Rules$
			\EndFunction
		\end{algorithmic}\label{fig:apriori-pseudocode}
	\end{algorithm}
	
\subsection{Previous Work on ML and AI Interplay with Association Rule Mining Algorithms}
The integration of auto-encoders in association rule mining, as discussed in \cite{berteloot2023association}, combines deep learning with conventional data mining methods. This approach utilizes auto-encoders to distill complex data patterns into abstract representations, facilitating the discovery of association rules. By leveraging auto-encoders' capability to learn dense data representations, the methodology improves the efficiency and accuracy of association rule mining. This integration offers insights into data processing and interpretation, extending its applicability across domains such as e-commerce and bioinformatics. The research suggests potential avenues for enhancing algorithmic processes through the synergy of machine learning and AI, contributing to advancements in the field.

	\subsection{Algogenic Enhancements for the Apriori Algorithm}
	\subsubsection{Intelligent Itemset Pruning}
	
	\paragraph{Streamlining Search with Predictive Pruning}
	The Apriori algorithm's efficiency in mining frequent itemsets faces challenges as dataset sizes increase, necessitating significant computational resources. Intelligent Itemset Pruning, leveraging predictive insights from Large Language Models, aims to refine the search process by preemptively eliminating itemsets unlikely to meet the minimum support threshold. This enhancement focuses on analyzing patterns within data to identify and disregard less promising itemsets early, aiming to reduce the computational burden and enhance the algorithm's scalability in handling extensive datasets. By integrating LLMs to discern and prune itemsets, this approach seeks to streamline the search process, potentially improving the algorithm's performance in identifying frequent itemsets across diverse data environments.
	
	\paragraph{Incorporating LLM Insights into Pruning Decisions}
	This method employs LLMs to scrutinize data for complex patterns, predicting itemset relevance beyond the basic application of minimum support criteria. LLMs can evaluate potential frequency based on trends and inter-item dependencies, enabling dynamic adjustment of pruning decisions. This not only refines the pruning process but also aids in uncovering hidden associations, enhancing the Apriori algorithm's capability to efficiently mine frequent itemsets. By dynamically adjusting pruning strategies based on LLM assessments, the algorithm aims for a more nuanced and effective search process.
	
	\paragraph{Adapting the Apriori Algorithm}
	To accommodate Intelligent Itemset Pruning, the Apriori algorithm integrates a preliminary evaluation step using LLM predictions to filter out unlikely candidates. This integration aims at maintaining the algorithm's integrity while reducing computational load, enhancing efficiency by focusing on itemsets with higher success probabilities. The adaptation includes continuous refinement of pruning criteria, leveraging real-time insights from LLMs to navigate evolving data patterns, aiming for a balance between efficiency and thorough exploration of the itemset space.
	
	\paragraph{Enhancements in Efficiency and Discovery Potential}
	By focusing computational efforts on itemsets with a higher likelihood of relevance, Intelligent Itemset Pruning seeks to improve both the efficiency of the Apriori algorithm and the potential for discovering significant patterns. This approach not only aims to reduce the computational resources required for data mining but also enriches the insights derived from the analysis, enhancing the algorithm's utility in extracting valuable knowledge from large datasets.
	
	\paragraph{Navigating Challenges and Maximizing Impact}
	While the integration of LLMs in Intelligent Itemset Pruning presents a forward step in data mining, it brings challenges related to the accuracy and biases of LLM predictions. The success of this approach depends on robust training and continuous refinement of LLMs, ensuring their predictions remain relevant and unbiased. Emphasizing adaptability and the effective use of AI insights, this strategy aims to enhance the Apriori algorithm's performance, underlining the importance of addressing technical and ethical considerations in AI integration.
	
	\subsubsection{Dynamic Support Threshold Adjustment}
	
	\paragraph{Optimizing Pattern Discovery through Support Adjustment}
	The Apriori algorithm traditionally employs a static minimum support threshold, which may not always efficiently capture meaningful patterns. Dynamic Support Threshold Adjustment, leveraging LLMs, proposes dynamically tailoring the support threshold to the dataset's specific characteristics, aiming for an optimal balance that maximizes the algorithm's ability to identify significant patterns without overwhelming computational resources. This approach seeks to adapt the algorithm to varying data distributions, enhancing its scalability and robustness in discovering accurate patterns.
	
	\paragraph{LLM-driven Analysis for Threshold Determination}
	This enhancement involves LLMs analyzing the dataset to predict the most effective minimum support threshold, allowing for better adaptation to the dataset's nuances. The dynamic adjustment of the support threshold, informed by LLM analysis, aims to fine-tune the algorithm's sensitivity to meaningful associations, enhancing its efficiency and effectiveness in mining frequent itemsets.
	
	\paragraph{Incorporation into the Apriori Workflow}
	Incorporating Dynamic Support Threshold Adjustment into the Apriori algorithm requires modifications to include consultations with LLM recommendations before each itemset generation level. This strategy aims to ensure that adjustments in the support threshold are informed by an ongoing analysis of the dataset, enhancing the algorithm's adaptability and performance in diverse data environments.
	
	\paragraph{Balancing Discovery and Efficiency}
	Dynamic Support Threshold Adjustment aims to refine the search process for frequent itemsets, striking a balance between exploring meaningful patterns and maintaining computational efficiency. This equilibrium is crucial for handling large datasets efficiently, enhancing the algorithm's scalability and its ability to uncover valuable insights.
	
	\paragraph{Future Directions and Potential Challenges}
	The success of Dynamic Support Threshold Adjustment hinges on the accuracy and relevance of LLM analysis, highlighting the need for ongoing refinement and consideration of scalability and interpretability issues. Future developments will likely focus on enhancing LLM predictive accuracy and exploring alternative adjustment methods to improve the robustness and adaptability of the algorithm.
	
	\subsubsection{Rule Confidence Prediction}
	
	\paragraph{Enhancing Rule Generation with Predictive Analytics}
	Rule Confidence Prediction introduces an Algogenic enhancement to predict the confidence of potential association rules using LLMs before full evaluation. This strategy aims to streamline the rule generation phase by focusing on rules likely to meet the confidence threshold, leveraging historical data patterns. By prioritizing high-potential rules, this approach seeks to enhance computational efficiency and the quality of insights derived from the data mining process.
	
	\paragraph{Predictive Modeling for Rule Confidence}
	Integrating predictive modeling with LLMs enhances the Apriori algorithm's capability to prioritize the evaluation of high-confidence rule candidates, optimizing computational resources and uncovering hidden relationships within the dataset. This targeted approach aims to accelerate rule discovery and enhance the algorithm's utility in extracting meaningful insights.
	
	\paragraph{Integration into the Apriori Framework}
	Incorporating Rule Confidence Prediction involves augmenting the rule generation process with a predictive filtering step, leveraging LLM predictions to focus on high-potential candidates. This integration aims to conserve computational resources and streamline the overall process, enhancing the efficiency and effectiveness of association rule mining.
	
	\paragraph{Impact on Computational Efficiency and Insight Discovery}
	Focusing on rules with higher predicted confidence enhances the computational efficiency of the Apriori algorithm and the potential for discovering impactful insights. This approach aims to reduce the evaluation of low-potential rules, enabling quicker convergence to meaningful patterns and associations.
	
	\paragraph{Challenges and Adaptive Strategies}
	Implementing Rule Confidence Prediction presents challenges, including mitigating bias and ensuring the adaptability of LLM predictions. Addressing these challenges requires continuous refinement and the integration of adaptive strategies, aiming to enhance the predictive accuracy and utility of LLMs in association rule mining.
	
	\subsubsection{Semantic Analysis for Item Grouping}
	
	\paragraph{Elevating Itemset Generation with Semantic Insights}
	Semantic Analysis for Item Grouping enhances the Apriori algorithm by leveraging LLMs for semantic analysis of item descriptors, aiming to uncover meaningful patterns based on semantic relationships. This approach seeks to extend the algorithm's capability beyond transactional co-occurrence, enhancing the quality and relevance of mined association rules.
	
	\paragraph{Implementing Semantic Analysis}
	LLMs undertake deep semantic analysis of item descriptors, identifying semantic connections that enhance itemset generation. This semantic grounding aims to facilitate more meaningful association rule mining, enriching the interpretability and actionable insights derived from the analysis.
	
	\paragraph{Adapting Apriori to Semantic Insights}
	Integrating Semantic Analysis requires modifications to the Apriori algorithm, incorporating semantic groupings in the candidate itemset generation phase. This adaptation seeks to uncover nuanced relationships and enhance the algorithm's effectiveness in mining association rules from complex datasets.
	
	\paragraph{Benefits of Semantic-Driven Discoveries}
	Semantic analysis enriches association rule mining by uncovering nuanced patterns and facilitating serendipitous discoveries. This approach enhances the algorithm's capability to extract actionable insights, emphasizing the importance of semantic intelligence in data analysis.
	
	\paragraph{Navigating Semantic Complexity}
	Incorporating Semantic Analysis introduces complexity, necessitating accurate semantic interpretation and efficient data structures. Addressing these challenges involves leveraging advancements in natural language processing, aiming to enhance the Apriori algorithm's capability to navigate semantic complexity and uncover meaningful associations.
	
	\subsubsection{Adaptive Candidate Generation}
	
	\paragraph{Refining the Search for Meaningful Associations}
	Adaptive Candidate Generation, leveraging LLMs, aims to refine the Apriori algorithm's candidate generation process, focusing on item combinations with higher potential for significant insights. This method seeks to reduce computational overhead while enhancing pattern discovery, adapting the algorithm to the dataset's evolving analysis.
	
	\paragraph{Incorporating LLM Predictive Insights}
	Using LLMs to predict promising candidate itemsets before formal support evaluation aims to prioritize these combinations, potentially uncovering valuable associations more efficiently. This strategy seeks to enhance the Apriori algorithm's effectiveness in mining association rules by discerning subtle relationships within the dataset.
	
	\paragraph{Dynamic Adjustment to Candidate Generation}
	Implementing Adaptive Candidate Generation requires the Apriori algorithm to dynamically adjust its candidate generation strategy based on LLM recommendations, aiming to improve computational efficiency and adaptability. This adaptation focuses on optimizing the search process for frequent itemsets, enhancing the algorithm's performance across diverse datasets.
	
	\paragraph{Optimizing Computational Resources and Discovery}
	Focusing on promising candidate itemsets aims to optimize computational resource use, enhancing the efficiency of the Apriori algorithm and increasing the likelihood of discovering impactful insights. This targeted approach seeks to improve the scalability of the algorithm, enabling effective pattern discovery in large datasets.

	\subsubsection{Pseudocode for Algogenic Apriori Algorithm}
	The Algogenic Apriori Algorithm approach harnesses AI to enhance the traditional Apriori Algorithm by dynamically adjusting its parameters and strategies based on the observed behavior of the system and real-time error estimates. This pseudocode, available in \ref{fig:apriori-Algogen-pseudocode}, outlines an advanced framework incorporating AI-driven enhancements for adaptive parameter tuning, itemset generation, support threshold determination, and real-time optimization.
	
	\begin{algorithm}
		\caption{Algogenic Apriori Algorithm Pseudocode}
		\begin{algorithmic}[1]
			\Procedure{AlgogenicApriori}{TransactionDB, minSupport, minConfidence}
			
			\Comment{Preprocessing Phase}
			\State Analyze TransactionDB with LLM for initial insights.
			\State Identify frequent individual items based on minSupport.
			
			\Comment{Core Algorithm Phase}
			\State $L_1 \gets$ Apply Intelligent Itemset Pruning with LLM insights.
			\For{($k = 2; L_{k-1} \neq \emptyset; k++$)}
			\State $C_k \gets$ Generate Candidate Itemsets from $L_{k-1}$.
			\State $C_k \gets$ Apply Adaptive Candidate Generation with LLM.
			\State Adjust $minSupport$ dynamically with LLM insights.
			\State $L_k \gets$ Calculate Support and filter $C_k$ by $minSupport$.
			\EndFor
			\State $AllFrequentItemsets \gets$ Aggregate all $L_k$.
			\State $Rules \gets$ Generate Association Rules from $AllFrequentItemsets$.
			\State Predict Rule Confidence with LLM and filter by $minConfidence$.
			\State Refine $Rules$ using Semantic Analysis for Item Grouping with LLM.
			
			\Comment{Postprocessing Phase}
			\State Interpret Results with enhanced understanding from LLM.
			
			\State \Return $Rules$
			\EndProcedure
		\end{algorithmic}\label{fig:apriori-Algogen-pseudocode}
	\end{algorithm}

	\begin{figure}
		\centering
		\includegraphics[width=0.55\textwidth]{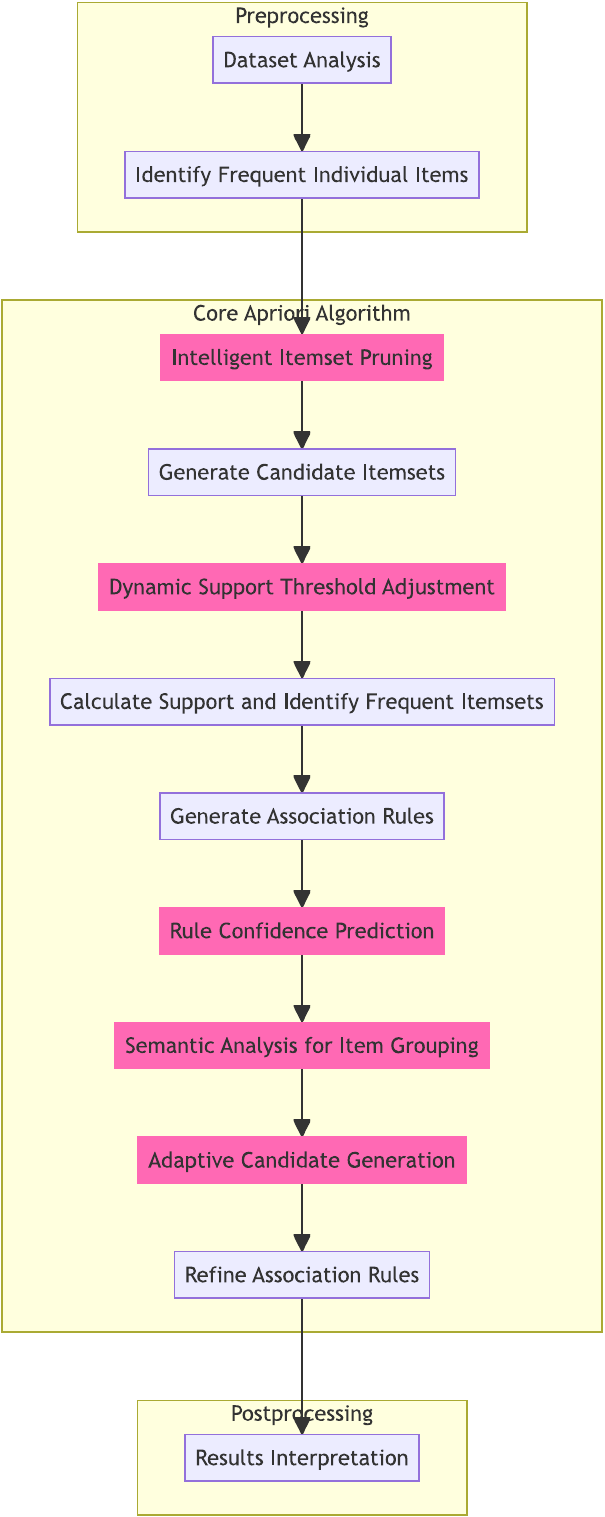}
		\caption{Enhancing the Apriori Algorithm with Generative AI: This figure illustrates the integration of generative AI enhancements within the Apriori algorithm for association rule mining. By incorporating AI-driven processes like intelligent itemset pruning, dynamic support threshold adjustment, rule confidence prediction, semantic analysis for item grouping, and adaptive candidate generation, the algorithm significantly improves in efficiency and discovery quality. These enhancements allow the Apriori algorithm to dynamically adapt to dataset characteristics and real-time insights, optimizing the search for meaningful patterns and associations in large datasets, thus enriching the data mining process with deeper insights and enhanced computational efficiency.}
		\label{fig:apriori}
	\end{figure}


	\part{General Algogens}

	
	\chapterimage{pngs/development_of_Algogens.png} 
	
	\chapter{Developing Algogens}\index{Developing Algogens}
	
	The development of Algogens marks a milestone in integrating generative AI with traditional algorithmic methods. This section delves into the multifaceted process of developing Algogen, detailing its conceptualization, design architecture, and the intricacies of its components. The journey from the initial idea to a fully realized framework illustrates the innovative approach to merging two distinct yet complementary technological realms. The development story of Algogens is not just about creating a new tool but crafting a novel methodology poised to transform the landscape of problem-solving across various industries.
	
	\section{Conceptualization of Algogens}\index{Conceptualization of Algogens}
	
	The conceptualization of Algogens represents a significant milestone in the evolution of computational problem-solving. This subsection delves into the foundational ideas behind Algogen, outlining its inception, the driving motivations for its development, and the initial challenges and objectives that shaped its design.
	
	\subsection{Origins and Foundational Ideas}
	The idea for Algogens emerged from a growing recognition of the limitations inherent in generative AI and traditional algorithmic methods when applied independently. Generative AI, while capable of producing novel solutions, often lacks the precision and structure required for certain problem domains. On the other hand, traditional algorithms excel in structured problem-solving but may struggle with complex, ambiguous tasks. The foundational concept behind Algogens was to create a framework that synergistically combines the creative problem-solving capabilities of AI with the structured, logical precision of algorithms. The aim was to harness the strengths of both approaches while mitigating their weaknesses.
	
	\paragraph{Integration of AI and Algorithms}
	One of the fundamental challenges in developing Algogens was finding ways to seamlessly integrate AI and algorithmic techniques. This integration required careful consideration of how to leverage the creative, exploratory nature of AI alongside the deterministic, rule-based nature of algorithms. Researchers explored various approaches, including hybrid models that combine machine learning algorithms with traditional algorithms, as well as techniques for incorporating AI-generated insights into algorithmic decision-making processes. This integration process involved extensive experimentation and iterative refinement to achieve optimal synergy between AI and algorithms. Moreover, it necessitated a deep understanding of both AI and algorithmic principles, as well as the ability to bridge the gap between them effectively. Overcoming these challenges required collaboration between experts from diverse backgrounds, including computer science, mathematics, and cognitive science, each bringing unique insights and expertise to the table. Through iterative experimentation and refinement, researchers honed their understanding of how AI and algorithms could complement each other, leading to breakthroughs in problem-solving methodologies and techniques.
	
	\paragraph{Hybridization Benefits}
	The hybridization of AI and algorithms in Algogens offers several distinct advantages. By combining the exploratory power of AI with the precision of algorithms, Algogens can tackle a wider range of problem domains than either approach alone. Additionally, the hybrid nature of Algogens enables it to adapt and learn from new data and experiences, continually improving its problem-solving capabilities over time. This adaptability is crucial in dynamic environments where challenges evolve rapidly, allowing Algogens to remain effective and relevant in the face of changing circumstances. Furthermore, the hybridization process fosters innovation by encouraging the cross-pollination of ideas and techniques from both AI and algorithmic fields, leading to novel approaches and solutions that may not have been possible otherwise. As Algogens evolves, it has the potential to revolutionize not only problem-solving but also the way we conceptualize and develop AI systems, paving the way for more integrated and versatile technologies in the future.
	
	\subsection{Motivations for Developing Algogen}
	The development of Algogens was motivated by the need for more adaptable, efficient, and intelligent problem-solving tools in various industries. Traditional approaches often lack flexibility and scalability in an era of rapidly evolving technology and increasingly complex challenges. Algogens was envisioned as a solution to bridge this gap, offering a dynamic and versatile tool capable of addressing a wide range of modern computational problems.
	
	\paragraph{Industry Applications}
	In industries ranging from finance to healthcare to manufacturing, there is a growing demand for intelligent problem-solving tools that can adapt to changing circumstances and provide actionable insights in real-time. Algogens offers a promising solution to this demand by combining the strengths of AI and algorithms to deliver more robust, efficient, and scalable problem-solving capabilities. This potential for widespread application across diverse industries highlights the significance of Algogens as a transformative technology with far-reaching implications. Moreover, the versatility of Algogens allows it to be tailored to specific industry needs, whether it's optimizing supply chain operations, predicting market trends, or personalizing healthcare treatments. The adoption of Algogens in various industries has the potential to drive significant productivity gains and innovation, leading to economic growth and improved quality of life for individuals and communities worldwide.
	
	\paragraph{Competitive Advantage}
	Companies that adopt Algogens stand to gain a significant competitive advantage in their respective industries. By leveraging advanced AI technologies alongside traditional algorithms, organizations can streamline their operations, optimize decision-making processes, and uncover new opportunities for innovation and growth. In an increasingly digital and data-driven world, Algogens represents a powerful tool for staying ahead of the curve. This competitive advantage stems from Algogens' ability to generate insights and solutions that are not only accurate but also timely, enabling organizations to make informed decisions faster than their competitors. Additionally, Algogens can help businesses adapt to changing market conditions more effectively by providing real-time analysis and predictive capabilities. As the adoption of Algogens becomes more widespread, it has the potential to reshape industry dynamics and redefine the benchmarks for success in the digital age.
	
	\subsection{Initial Challenges and Objectives}
	The initial phase of conceptualizing Algogens involved identifying and addressing several key challenges. One primary challenge was integrating the disparate methodologies of generative AI and algorithms in a harmonious and mutually beneficial way. Another was ensuring the framework was adaptable enough to be applied across various domains while remaining robust and reliable. The objectives were clear: to develop a framework that enhanced problem-solving capabilities and pushed the boundaries of what could be achieved by integrating AI and algorithms.
	
	\paragraph{Technical Complexity}
	One of the primary technical challenges in developing Algogens was the complexity of integrating AI and algorithmic techniques. Researchers faced difficulties in designing algorithms that could effectively leverage the outputs of AI models while maintaining computational efficiency and scalability. Additionally, ensuring the reliability and robustness of the framework required extensive testing and validation across a diverse range of problem domains. Overcoming these technical challenges required a multidisciplinary approach, with experts from AI, computer science, and mathematics collaborating to develop innovative solutions that could seamlessly integrate AI and algorithms. This intricate process involved not only developing advanced algorithms but also optimizing them to work in tandem with AI components efficiently. Moreover, it required the development of new methodologies and tools for evaluating and benchmarking the performance of Algogens across different tasks and domains. By pushing the boundaries of computational problem-solving, researchers paved the way for Algogens to excel in diverse applications, from data analysis to natural language processing.
	
	\paragraph{Ethical and Social Implications}
	As with any advanced technology, the development of Algogens raised important ethical and social considerations. Researchers grappled with questions surrounding data privacy, algorithmic bias, and the potential impact of AI-driven decision-making on society. Addressing these concerns required a multidisciplinary approach, involving experts from fields such as ethics, law, and sociology, to ensure that Algogens was developed and deployed responsibly. This emphasis on ethical considerations reflects a broader trend in the development of AI technologies, where ethical principles and societal impacts are given increasing importance alongside technical innovation. Moreover, the ethical implications of Algogens extend beyond its development phase and into its deployment, requiring ongoing monitoring and evaluation to mitigate potential risks and ensure its responsible use in society. As Algogens continues to evolve and proliferate, it will be essential to engage stakeholders and the broader public in discussions about its ethical implications and potential societal impacts. By fostering transparency and accountability, researchers can help build trust in Algogens and ensure its positive contribution to society.
	
	\subsection{Setting the Stage for Development}
	This conceptual phase set the stage for the subsequent development of Algogen. It involved extensive research into existing AI and algorithmic methods, consultations with experts in various fields, and a thorough analysis of potential applications and implications. The outcome was a blueprint for a framework that could transform computational problem-solving across multiple sectors.
	
	\paragraph{Research and Development Efforts}
	Building on the conceptual foundations laid out in this phase, researchers and engineers embarked on the development of Algogens in earnest. This involved designing and implementing algorithms, developing AI models, and integrating the two components into a cohesive framework. Throughout the development process, collaboration and communication were key, with interdisciplinary teams working together to overcome technical challenges and refine the capabilities of Algogens. This collaborative approach facilitated rapid progress and innovation, enabling Algogens to evolve from concept to reality in a relatively short period. Additionally, the research and development efforts laid the groundwork for future advancements in AI and algorithmic technologies, with insights and methodologies generated during the development of Algogens contributing to broader scientific knowledge and innovation. As Algogens continues to evolve, ongoing research and development efforts will be essential to unlock its full potential and address emerging challenges and opportunities.
	
	\paragraph{Proof-of-Concept Demonstrations}
	As development progressed, researchers conducted proof-of-concept demonstrations to validate the effectiveness of Algogens in real-world scenarios. These demonstrations served to showcase the capabilities of the framework, identify areas for improvement, and gather feedback from stakeholders. By iterating on these demonstrations and incorporating user feedback, researchers were able to iteratively refine and enhance Algogens, bringing it closer to its full potential. This iterative development process was essential in ensuring that Algogens met the diverse needs and requirements of its intended users, ultimately enhancing its usability and effectiveness in real-world applications. Moreover, the success of these proof-of-concept demonstrations bolstered confidence in Algogens' capabilities and paved the way for its broader adoption and implementation across various industries and domains. Moving forward, continued experimentation and validation will be crucial to further refine and optimize Algogens for a wide range of applications and use cases. Through ongoing collaboration and innovation, researchers can continue to push the boundaries of what is possible with Algogens, unlocking new opportunities for discovery and problem-solving.
	
	In conclusion, the conceptualization of Algogens was a process marked by innovation, foresight, and a deep understanding of the evolving landscape of technology and its applications. It laid the groundwork for a framework that promised to address existing challenges in computational problem-solving and open up new avenues for exploration and discovery. The successful integration of AI and algorithms in Algogens represents a significant advancement in the field of artificial intelligence, with far-reaching implications for industries, society, and the future of computing. As Algogens continues to evolve and mature, it has the potential to revolutionize how we approach problem-solving, drive innovation across diverse industries, and shape the trajectory of technological development in the years to come.

	\section{Design and Architecture of Algogens}\index{Design and Architecture of Algogens}
	
	The design and architecture of Algogens are fundamental to its functionality and effectiveness. This subsection provides a detailed overview of Algogen’s structural components, modular design, and the architectural choices that enable its robust and flexible problem-solving capabilities.
	
	\subsection{Overall Structure of Algogen}
	Algogens is structured as a cohesive system integrating two primary components: a generative AI module and an algorithmic processing module. These components are designed to interact seamlessly, with data and insights flowing bidirectionally to ensure that the creative insights from AI are grounded in the logical rigor of algorithms. This structure facilitates a balanced approach to problem-solving, leveraging the strengths of both AI and algorithmic methods.
	
	\paragraph{Synergistic Interaction between AI and Algorithms}
	The integration of the generative AI module and the algorithmic processing module within Algogens fosters a synergistic interaction that capitalizes on the unique strengths of each component. While the AI module excels in generating innovative solutions and exploring complex problem spaces, the algorithmic processing module provides the necessary structure and systematic approach to validate and refine these solutions. This bidirectional flow of data and insights ensures that Algogens can harness the creative potential of AI while maintaining the precision and reliability of algorithmic techniques.
	
	\paragraph{Adaptive Feedback Loop}
	An essential aspect of the overall structure of Algogens is the establishment of an adaptive feedback loop between its constituent modules. As the AI module generates solutions and insights, these are fed back into the algorithmic processing module for evaluation and refinement. Conversely, the algorithmic processing module provides feedback to the AI module, guiding its learning process and influencing the generation of future solutions. This adaptive feedback loop enables Algogens to continuously improve its problem-solving capabilities over time, adapting to new challenges and evolving problem domains.
	
	\paragraph{Integration of Cross-Domain Expertise}
	In addition to facilitating interaction between AI and algorithms, the overall structure of Algogens also allows for the integration of cross-domain expertise and knowledge. By incorporating insights and methodologies from diverse disciplines, such as cognitive science, mathematics, and engineering, Algogens can leverage a rich tapestry of perspectives to tackle complex problems from multiple angles. This interdisciplinary approach enhances the robustness and adaptability of Algogens, enabling it to address a wide range of challenges across different domains with sophistication and precision.
	
	\subsection{Modular Design for Flexibility}
	A key feature of Algogen’s architecture is its modular design. This allows individual components to be updated or replaced without disrupting the entire system, ensuring that Algogens remains adaptable and scalable. The modular nature also facilitates customization for specific industry applications, allowing components to be tailored to meet unique problem-solving requirements.
	
	\paragraph{Scalability and Extensibility}
	The modular design of Algogens not only enhances its flexibility but also ensures scalability and extensibility. New modules can be seamlessly integrated into the existing framework to incorporate advancements in AI or algorithmic techniques. Likewise, existing modules can be upgraded or replaced to accommodate changing requirements or technological advancements. This scalability and extensibility future-proof Algogens, allowing it to evolve and remain relevant in the face of emerging challenges and opportunities.
	
	\paragraph{Domain-Specific Modules}
	Another advantage of Algogen’s modular design is its ability to support domain-specific modules tailored to particular industries or problem domains. These specialized modules can encapsulate domain-specific knowledge and problem-solving strategies, enhancing Algogens’ effectiveness and relevance in diverse applications. By accommodating a wide range of domain-specific modules, Algogens can address the unique challenges faced by different industries, from financial forecasting to medical diagnosis, with tailored solutions optimized for each context.
	
	\paragraph{Flexibility in Component Interchangeability}
	Moreover, the modular design of Algogens facilitates flexibility in component interchangeability, allowing users to customize and optimize the system according to their specific needs and preferences. This granular control over individual components enables users to experiment with different algorithms, AI models, and data processing techniques to find the most effective combination for their problem-solving tasks. By empowering users with flexibility and control, Algogens enhances user engagement and satisfaction, fostering a collaborative and iterative problem-solving process that drives continuous improvement and innovation.
	
	\subsection{Algorithmic Backbone}
	The algorithmic backbone of Algogens consists of a suite of carefully selected and optimized algorithms chosen for their reliability, efficiency, and applicability across a wide range of problems. This backbone provides the structured, rule-based framework necessary for systematic problem-solving and serves as a stable foundation for the integration of AI.
	
	\paragraph{Robustness and Versatility}
	The algorithmic backbone of Algogens is characterized by its robustness and versatility, underpinned by a diverse set of algorithms capable of addressing various problem types and complexities. These algorithms are rigorously tested and optimized to ensure reliability and efficiency, enabling Algogens to tackle challenging problem domains with confidence. Moreover, the versatility of the algorithmic backbone allows Algogens to adapt to different problem-solving contexts, from optimization tasks to pattern recognition, with ease and efficiency.
	
	\paragraph{Continuous Improvement}
	Continuous improvement is a cornerstone of Algogens’ algorithmic backbone. Through ongoing research and development efforts, algorithms are continually refined and enhanced to improve their performance and effectiveness. This iterative process of optimization ensures that Algogens remains at the forefront of computational problem-solving, leveraging the latest advancements in algorithmic techniques to deliver superior results. Furthermore, the integration of feedback mechanisms enables Algogens to learn from past experiences and adapt its algorithmic strategies dynamically, further enhancing its problem-solving capabilities over time.
	
	\paragraph{Ethical Considerations in Algorithm Selection}
	In the selection and optimization of algorithms for Algogens’ backbone, ethical considerations play a crucial role. Algorithms are evaluated not only based on their technical performance but also on their ethical implications and societal impact. This holistic approach ensures that Algogens’ algorithmic backbone upholds ethical principles such as fairness, transparency, and accountability, mitigating the risk of bias or discrimination in decision-making processes. By prioritizing ethical considerations in algorithm selection, Algogens demonstrates its commitment to responsible and socially conscious problem-solving, fostering trust and confidence among users and stakeholders.
	
	\subsection{Integration of Generative AI}
	The generative AI component of Algogens is what sets it apart. It utilizes advanced machine learning models to generate creative solutions and scenarios. This AI module is designed to learn continuously from new data and experiences, ensuring that the proposed solutions are innovative, relevant, and practical.
	
	\paragraph{Creative Problem-Solving}
	The integration of generative AI into Algogens unlocks new possibilities for creative problem-solving. Unlike traditional algorithmic approaches, which rely on predefined rules and patterns, the AI module can explore novel solution spaces and generate innovative ideas that may not be immediately obvious to human designers. This creative capacity enables Algogens to tackle unconventional problems and discover non-intuitive solutions, pushing the boundaries of what is possible in computational problem-solving.
	
	\paragraph{Adaptive Learning Mechanisms}
	At the heart of Algogens’ generative AI module are adaptive learning mechanisms that enable it to continuously improve and adapt to new challenges. These mechanisms allow the AI module to analyze data, identify patterns, and refine its models based on feedback from the algorithmic processing module and real-world outcomes. By learning from both successes and failures, Algogens’ AI module can iteratively enhance its problem-solving capabilities, becoming increasingly adept at generating high-quality solutions over time.
	
	\paragraph{Human-Centric Design in AI Development}
	In the development and training of generative AI models for Algogens, a human-centric design approach is paramount. This approach emphasizes the importance of human involvement and oversight throughout the AI development lifecycle, from data collection and model training to deployment and evaluation. By involving domain experts, ethicists, and end-users in the AI development process, Algogens ensures that its AI models are aligned with human values and preferences, promoting trust and acceptance among users. Moreover, human-centric design fosters transparency and explainability in AI decision-making, enabling users to understand and critique the reasoning behind Algogens’ recommendations.
	
	\subsection{Data Processing and Communication Mechanisms}
	A critical aspect of Algogen’s architecture is the efficient data processing and communication between the AI and algorithmic modules. This involves sophisticated data handling and exchange protocols to ensure both modules can effectively share insights and contribute to problem-solving.
	
	\paragraph{Data Integration and Fusion}
	Algogens employs advanced data integration and fusion techniques to harmonize inputs from multiple sources and modalities. This enables the AI module to access diverse data streams, including structured and unstructured data, and extract meaningful insights to inform its decision-making process. Similarly, the algorithmic processing module can leverage these insights to refine its problem-solving strategies, creating a mutually beneficial exchange of information between the two modules.
	
	\paragraph{Real-Time Collaboration}
	To facilitate real-time collaboration between the AI and algorithmic modules, Algogens implements efficient communication mechanisms that enable rapid exchange of data and insights. This real-time collaboration is essential for dynamic problem-solving scenarios where timely decision-making is critical. By maintaining synchronized communication channels, Algogens ensures that both modules can work in concert to address emerging challenges and opportunities as they arise.
	
	\paragraph{Privacy-Preserving Data Handling}
	In handling sensitive or personal data, Algogens prioritizes privacy and data protection through robust encryption and anonymization techniques. These measures ensure that user data is safeguarded against unauthorized access or misuse, maintaining user trust and compliance with data privacy regulations. Additionally, Algogens implements data minimization strategies to collect and process only the necessary data for problem-solving tasks, reducing the risk of privacy breaches and enhancing user privacy and control over their data.
	
	\subsection{Interface and User Interaction}
	Algogens have an intuitive user interface, allowing users from various domains to interact with the system effectively. The interface provides insights into problem-solving, offers control over specific parameters, and presents solutions in an accessible format.
	
	\paragraph{User-Centric Design}
	The interface of Algogens is designed with a user-centric approach, prioritizing ease of use and accessibility. Through intuitive visualization tools and interactive features, users can gain valuable insights into the problem-solving process and understand the rationale behind Algogens’ recommendations. Moreover, the interface allows users to customize parameters and constraints according to their preferences, empowering them to tailor Algogens’ problem-solving approach to suit their specific needs and requirements.
	
	\paragraph{Transparency and Explainability}
	Transparency and explainability are core principles guiding the design of Algogens’ interface. By providing clear explanations of the underlying algorithms and decision-making processes, the interface fosters trust and confidence in Algogens’ recommendations. Users can gain a deeper understanding of how Algogens arrives at its solutions, enabling them to evaluate the reliability and relevance of the recommendations in the context of their domain expertise. Additionally, transparent interfaces promote collaboration and knowledge sharing among users, facilitating collective problem-solving efforts and fostering a sense of ownership and accountability.
	
	\paragraph{Interactive Feedback Mechanisms}
	To enhance user engagement and satisfaction, Algogens incorporates interactive feedback mechanisms into its interface, allowing users to provide input and feedback on recommended solutions. This two-way communication enables users to convey their preferences, constraints, and domain-specific knowledge to Algogens, helping to refine and personalize the problem-solving process. Moreover, interactive feedback mechanisms foster a sense of collaboration and partnership between users and Algogens, empowering users to actively participate in the problem-solving journey and contribute to the generation of innovative solutions.
	
	In summary, the design and architecture of Algogens are central to its success as an advanced problem-solving tool. The thoughtful integration of generative AI with a robust algorithmic backbone and a modular and flexible structure positions Algogens as a versatile and powerful framework capable of addressing many complex challenges. By leveraging synergies between AI and algorithms, Algogens enables creative and systematic problem-solving approaches that push the boundaries of computational intelligence. Through continuous innovation and user-centric design, Algogens strives to empower users across diverse domains to unlock new insights and drive meaningful outcomes.

	\section{Algorithmic Backbone of Algogens}\index{Algorithmic Backbone of Algogens}
	
	The algorithmic backbone of Algogens serves as its foundational layer, providing a structured and systematic approach to problem-solving. This subsection delves into the specifics of the algorithmic methods employed, illustrating how they contribute to the reliability, efficiency, and effectiveness of the framework.
	
	\subsection{Composition of the Algorithmic Backbone}
	The backbone consists of a carefully curated collection of algorithms, meticulously selected for their proven efficiency, reliability, and applicability across diverse problem domains. This ensemble encompasses a wide spectrum of algorithmic techniques, including but not limited to data processing, optimization, decision-making, pattern recognition, and simulation. Each algorithm is chosen based on its ability to address specific problem characteristics and constraints, ensuring that Algogens possesses the requisite tools to tackle a myriad of computational challenges with precision and efficacy.
	
	\paragraph{Diverse Algorithmic Techniques}
	The composition of the algorithmic backbone reflects a diverse array of algorithmic techniques drawn from various branches of mathematics, computer science, and artificial intelligence. From classic algorithms such as sorting and searching algorithms to more advanced techniques like genetic algorithms, neural networks, and evolutionary strategies, Algogens incorporates a rich tapestry of methodologies tailored to suit different problem-solving contexts. This diversity not only broadens the framework's problem-solving capabilities but also fosters innovation and creativity by offering multiple avenues for exploring and addressing complex problems.
	
	\paragraph{Scalability and Robustness}
	Moreover, the algorithms comprising Algogens' backbone are selected for their scalability and robustness, capable of handling large-scale computational tasks with efficiency and stability. Whether processing massive datasets, optimizing complex systems, or simulating intricate scenarios, the algorithms exhibit scalability in both computational complexity and resource utilization. Furthermore, their robustness ensures consistent performance across diverse problem domains, minimizing the risk of algorithmic failures or inaccuracies and instilling confidence in the reliability of Algogens' outputs.
	
	\paragraph{Adaptability to Varied Problem Contexts}
	Additionally, the algorithmic backbone of Algogens is characterized by its adaptability to varied problem contexts and requirements. Through the careful selection and customization of algorithms, Algogens can tailor its problem-solving approach to suit specific application domains, constraints, and objectives. Whether addressing optimization problems in logistics, classification tasks in machine learning, or simulation challenges in scientific research, Algogens' algorithmic backbone can be configured to accommodate diverse problem characteristics, ensuring the framework's relevance and effectiveness across a wide range of application scenarios.
	
	\subsection{Role in Structured Problem-Solving}
	At the heart of Algogens' problem-solving capability lies the pivotal role played by its algorithmic backbone in structuring and decomposing complex problems. The backbone serves as a scaffold upon which intricate problem spaces are partitioned into manageable components, facilitating systematic analysis, solution development, and validation. By breaking down complex problems into smaller, more tractable sub-problems, Algogens empowers users to approach problem-solving tasks methodically, ensuring that solutions are logically coherent, reproducible, and verifiable.
	
	\paragraph{Hierarchical Problem Decomposition}
	The structured problem-solving facilitated by Algogens' algorithmic backbone often employs a hierarchical decomposition approach, wherein complex problems are decomposed into hierarchies of sub-problems with increasing levels of granularity. This hierarchical organization not only aids in managing problem complexity but also enables modular solution development, allowing individual components of the problem to be tackled independently before being integrated into a cohesive whole. Such hierarchical problem decomposition fosters a systematic and disciplined problem-solving methodology, enhancing the efficiency and effectiveness of Algogens across a wide range of problem domains.
	
	\paragraph{Iterative Refinement Process}
	Furthermore, the role of the algorithmic backbone extends beyond initial problem decomposition to encompass iterative refinement processes aimed at optimizing solution quality and performance. Through iterative cycles of analysis, synthesis, evaluation, and adaptation, Algogens iteratively refines and enhances its solutions based on feedback from both internal evaluation metrics and user-defined criteria. This iterative refinement process ensures that Algogens' solutions evolve over time, becoming increasingly sophisticated, robust, and aligned with user expectations and domain requirements.
	
	\paragraph{Integration with Human Expertise}
	Additionally, the algorithmic backbone of Algogens is designed to integrate seamlessly with human expertise, leveraging the collective intelligence and problem-solving acumen of domain experts to augment and enrich algorithmic-driven solutions. By providing mechanisms for user interaction, feedback, and collaboration, Algogens empowers users to contribute domain-specific insights, constraints, and preferences to the problem-solving process, thereby enhancing the relevance, accuracy, and applicability of the generated solutions. This integration of human expertise with algorithmic methods fosters a symbiotic relationship that combines the analytical rigor of algorithms with the contextual understanding and intuition of human experts, leading to more informed, nuanced, and effective problem-solving outcomes.
	
	\subsection{Ensuring Reliability and Predictability}
	Within Algogens' algorithmic backbone, paramount importance is placed on ensuring the reliability and predictability of the employed algorithms. The algorithms are meticulously vetted and selected based on their adherence to established rules, logical processes, and mathematical principles, thereby guaranteeing consistent and dependable performance across diverse problem-solving scenarios. This emphasis on reliability and predictability is particularly critical in applications where decision accuracy and reproducibility are of paramount importance, such as in safety-critical systems, financial modeling, and healthcare diagnostics.
	
	\paragraph{Mathematical Rigor and Formalism}
	The reliability and predictability of Algogens' algorithms stem from their foundation in mathematical rigor and formalism, underpinned by well-defined mathematical models, principles, and methodologies. By grounding problem-solving processes in mathematical theory, Algogens ensures that its algorithms operate within a structured and logical framework, free from ambiguity, inconsistency, or subjective interpretation. This adherence to mathematical rigor not only enhances the trustworthiness of Algogens' solutions but also facilitates cross-disciplinary collaboration and knowledge transfer by providing a common language for expressing and communicating problem-solving concepts and methodologies.
	
	\paragraph{Verification and Validation Processes}
	Moreover, Algogens employs rigorous verification and validation processes to assess the reliability and predictability of its algorithms before deployment in real-world applications. Through extensive testing, benchmarking, and validation against ground truth or expert knowledge, Algogens verifies the correctness, robustness, and generalizability of its algorithms across a diverse range of test cases and scenarios. Additionally, ongoing monitoring and performance evaluation mechanisms ensure that Algogens' algorithms maintain their reliability and predictability over time, adapting to changing environmental conditions, data distributions, and user requirements while preserving their integrity and effectiveness.
	
	\paragraph{Risk Mitigation and Error Handling}
	To further enhance reliability and predictability, Algogens incorporates robust risk mitigation and error handling mechanisms into its algorithmic backbone. These mechanisms proactively identify, mitigate, and recover from potential algorithmic failures, errors, or anomalies that may arise during problem-solving processes. By anticipating and addressing potential sources of uncertainty, variability, or bias in algorithmic outputs, Algogens ensures that its solutions remain robust, dependable, and trustworthy in the face of unforeseen contingencies or adversities, thereby instilling confidence in users and stakeholders regarding the reliability and predictability of Algogens' problem-solving capabilities.
	
	\subsection{Integration with Generative AI}
	A critical aspect of Algogens' algorithmic backbone is its seamless integration with the generative AI component, fostering a symbiotic relationship between structured algorithmic methods and creative AI-driven approaches to problem-solving. This integration capitalizes on the complementary strengths of AI and algorithms, combining the structured, rule-based reasoning of algorithms with the creativity, adaptability, and pattern recognition capabilities of generative AI to produce innovative and practical solutions to complex problems.
	
	\paragraph{Harmonizing Creativity and Precision}
	The integration of generative AI with Algogens' algorithmic backbone enables the harmonization of creativity and precision in problem-solving processes. While algorithms provide a structured framework for logical reasoning, data processing, and decision-making, generative AI injects creativity and innovation into solution generation by exploring alternative problem spaces, generating novel ideas, and synthesizing unconventional solutions that may elude traditional algorithmic approaches. By leveraging the creative potential of AI within the structured problem-solving framework provided by algorithms, Algogens strikes a balance between exploratory creativity and systematic precision, leading to the emergence of novel, effective, and practical solutions to complex computational challenges.
	
	\paragraph{Feedback-driven Iterative Improvement}
	Furthermore, the integration of generative AI with Algogens' algorithmic backbone facilitates feedback-driven iterative improvement cycles, wherein AI-generated solutions are evaluated, refined, and validated using algorithmic methods, and vice versa. This iterative feedback loop enables continuous refinement and enhancement of both AI and algorithmic components, fostering mutual learning, adaptation, and improvement over time. Through collaborative problem-solving efforts facilitated by AI-algorithm synergy, Algogens evolves into a dynamic and adaptive problem-solving framework capable of tackling increasingly complex and diverse challenges with ingenuity and precision.
	
	\subsection{Adaptability to Diverse Applications}
	The versatility and adaptability of Algogens' algorithmic backbone are key attributes that underpin its utility across a myriad of industries, domains, and problem-solving contexts. Whether optimizing logistics in supply chain management, analyzing complex datasets in healthcare informatics, simulating environmental dynamics in climate science, or predicting market trends in finance, Algogens' algorithms can be tailored and customized to suit specific application requirements and constraints, thereby enabling the framework to address a wide spectrum of computational challenges with efficacy and versatility.
	
	\paragraph{Domain-specific Algorithm Customization}
	One of the primary mechanisms through which Algogens achieves adaptability to diverse applications is through domain-specific algorithm customization and optimization. By tailoring algorithmic parameters, configurations, and methodologies to align with the unique characteristics, constraints, and objectives of specific problem domains, Algogens can enhance the relevance, accuracy, and effectiveness of its solutions within those domains. This domain-specific customization not only improves solution quality but also fosters domain expertise integration, facilitating collaborative problem-solving efforts between Algogens and domain experts across various industries and disciplines.
	
	\paragraph{Cross-disciplinary Problem-solving}
	Furthermore, the adaptability of Algogens' algorithmic backbone extends beyond individual domains to encompass cross-disciplinary problem-solving scenarios, where computational challenges transcend traditional disciplinary boundaries and require interdisciplinary collaboration and expertise. By integrating insights, methodologies, and best practices from diverse domains such as mathematics, computer science, engineering, natural sciences, social sciences, and humanities, Algogens leverages the collective intelligence and problem-solving acumen of multidisciplinary teams to address complex, multifaceted challenges holistically and synergistically. This cross-disciplinary approach not only enriches problem-solving strategies but also fosters innovation, creativity, and resilience in the face of multifarious computational challenges.
	
	\paragraph{Dynamic Adaptation to Emerging Challenges}
	Moreover, Algogens' algorithmic backbone exhibits dynamic adaptation capabilities to respond effectively to emerging challenges, trends, and opportunities in rapidly evolving technological, societal, and environmental landscapes. Through continuous monitoring, analysis, and learning from real-world data and experiences, Algogens' algorithms can adapt and evolve their problem-solving strategies, heuristics, and models to accommodate shifting problem dynamics, user requirements, and external constraints. This dynamic adaptation ensures that Algogens remains at the forefront of computational problem-solving innovation, constantly pushing the boundaries of its capabilities to address novel and unprecedented challenges with agility, resilience, and efficacy.
	
	\subsection{Continuous Improvement and Evolution}
	In line with the principles of agile development and continuous improvement, the algorithmic backbone of Algogens is designed to undergo iterative refinement, enhancement, and evolution over time. This continuous improvement process encompasses a variety of mechanisms and practices aimed at enhancing algorithmic performance, robustness, efficiency, and relevance in response to evolving user needs, technological advancements, and domain-specific requirements.
	
	\paragraph{Iterative Algorithm Optimization}
	One of the primary mechanisms for continuous improvement within Algogens' algorithmic backbone is iterative algorithm optimization, wherein algorithms undergo iterative cycles of analysis, experimentation, tuning, and validation to improve their performance, efficiency, and effectiveness. Through rigorous experimentation with algorithmic parameters, configurations, and methodologies, Algogens identifies and implements optimizations that enhance algorithmic efficiency, reduce computational complexity, and improve solution quality, thereby ensuring that the framework remains at the cutting edge of computational problem-solving capabilities.
	
	\paragraph{Feedback-driven Learning and Adaptation}
	Additionally, Algogens leverages feedback-driven learning and adaptation mechanisms to enhance algorithmic performance and relevance in response to real-world feedback, user interactions, and problem-solving outcomes. By soliciting feedback from users, domain experts, and stakeholders, Algogens captures valuable insights, preferences, and domain-specific knowledge that can be used to iteratively refine and adapt its algorithms to better align with user expectations, domain requirements, and application contexts. This feedback-driven learning process enables Algogens to continuously learn from past experiences, identify areas for improvement, and adapt its algorithmic strategies dynamically to address emerging challenges and opportunities effectively.
	
	\paragraph{Integration of State-of-the-art Techniques}
	Furthermore, Algogens remains at the forefront of computational problem-solving innovation by integrating state-of-the-art algorithmic techniques, methodologies, and paradigms into its algorithmic backbone. Through ongoing research, development, and collaboration with academic and industrial partners, Algogens identifies and incorporates cutting-edge algorithms, heuristics, and optimization strategies that push the boundaries of computational intelligence, enabling the framework to tackle increasingly complex and diverse problem domains with sophistication and efficacy. This integration of state-of-the-art techniques ensures that Algogens remains a dynamic and adaptive problem-solving framework capable of addressing the evolving needs and challenges of the modern world.
	
	In conclusion, the algorithmic backbone serves as a critical foundation for Algogens, providing the structural framework, stability, and adaptability necessary for effective problem-solving across diverse domains and applications. By integrating structured algorithmic methods with creative generative AI approaches, Algogens strikes a balance between systematic precision and exploratory creativity, enabling the emergence of innovative, practical, and impactful solutions to complex computational challenges. Through continuous improvement, dynamic adaptation, and integration of state-of-the-art techniques, Algogens evolves into a versatile and resilient problem-solving framework that empowers users to unlock new insights, drive meaningful outcomes, and address the complex challenges of the digital age with confidence and ingenuity.

	\section{Role of Generative AI in Algogens}\index{Role of Generative AI in Algogens}
	
	Integrating generative AI within Algogens represents a key innovation, bringing creativity and adaptability that enhances problem-solving. This subsection examines the specific role of generative AI in Algogen, its functionalities, and the benefits it offers.
	
	\subsection{Functionality of Generative AI in Algogen}
	Generative AI in Algogens serves as the engine of innovation, tasked with the generation of novel solutions, ideas, and scenarios that may not be immediately discernible through traditional problem-solving methods alone. Leveraging advanced machine learning models, particularly those adept at pattern recognition, predictive analysis, and scenario generation, generative AI within Algogens pioneers the exploration of uncharted problem spaces, pushing the boundaries of what is possible in computational problem-solving. By harnessing the creative potential of AI, Algogens unlocks new avenues for solution discovery, offering fresh perspectives and insights that complement and enrich traditional algorithmic approaches.
	
	\paragraph{Innovative Solution Generation}
	Generative AI within Algogens operates as a catalyst for innovation, driving the generation of solutions that transcend conventional problem-solving paradigms. Through its ability to explore vast problem spaces, identify non-obvious patterns, and synthesize novel concepts, generative AI expands the horizons of problem-solving, enabling Algogens to tackle complex challenges with ingenuity and creativity. By fostering a culture of exploration and experimentation, generative AI inspires breakthroughs and discoveries that redefine the boundaries of what is considered possible in computational problem-solving.
	
	\paragraph{Exploration of Uncharted Problem Spaces}
	Moreover, generative AI within Algogens facilitates the exploration of uncharted problem spaces, where conventional problem-solving methods may fall short. By leveraging sophisticated machine learning techniques, such as deep learning and reinforcement learning, generative AI navigates complex problem landscapes, uncovering hidden relationships, emergent phenomena, and novel solution trajectories that elude deterministic approaches. This exploratory capacity enables Algogens to address previously intractable challenges, opening new frontiers of inquiry and innovation in diverse domains ranging from scientific research to creative arts.
	
	\paragraph{Synthesis of Multifaceted Solutions}
	Additionally, generative AI excels in the synthesis of multifaceted solutions that integrate diverse perspectives, constraints, and objectives. Through its ability to analyze and synthesize heterogeneous data sources, generative AI within Algogens generates solutions that are holistic, contextually grounded, and aligned with user preferences and domain-specific requirements. Whether optimizing complex systems, designing novel materials, or generating artistic creations, generative AI fosters the emergence of solutions that embody creativity, feasibility, and relevance, enriching the problem-solving process and driving impactful outcomes.
	
	\subsection{Learning, Adaptation, and Evolution}
	A hallmark feature of generative AI within Algogens is its capacity for continual learning, adaptation, and evolution. Unlike static algorithms, the AI component of Algogens possesses the inherent ability to learn from vast datasets, refine its understanding, and adapt its solution generation strategies over time. Through iterative exposure to diverse data sources and problem contexts, generative AI evolves its capabilities, becoming increasingly adept at generating sophisticated, contextually relevant solutions that align with user preferences, domain-specific constraints, and evolving problem dynamics. This dynamic learning and adaptation process imbues Algogens with resilience and agility, enabling it to navigate complex problem landscapes with confidence and efficacy.
	
	\paragraph{Continuous Learning from Diverse Data Sources}
	The learning process of generative AI within Algogens is characterized by its continuous engagement with diverse data sources, ranging from structured datasets to unstructured multimedia content. By ingesting, processing, and analyzing heterogeneous data streams, generative AI acquires a nuanced understanding of the problem domain, discerning subtle patterns, correlations, and dependencies that inform solution generation. This continuous learning paradigm enables Algogens to adapt to evolving problem dynamics, emerging trends, and user preferences, ensuring that its solutions remain relevant, accurate, and effective over time.
	
	\paragraph{Adaptive Solution Generation Strategies}
	Furthermore, generative AI within Algogens employs adaptive solution generation strategies that evolve in response to changing problem contexts and constraints. Through dynamic adjustment of model parameters, exploration-exploitation trade-offs, and feedback-driven refinement mechanisms, generative AI fine-tunes its solution generation processes to optimize performance, efficacy, and relevance. This adaptive capacity enables Algogens to navigate uncertainty, variability, and complexity inherent in real-world problem-solving scenarios, tailoring its solutions to meet evolving user needs and domain-specific requirements with precision and agility.
	
	\paragraph{Emergence of Novel Solution Modalities}
	Moreover, the continual evolution of generative AI within Algogens gives rise to the emergence of novel solution modalities that transcend traditional problem-solving approaches. By synthesizing insights from disparate data sources, leveraging generative adversarial networks, and exploring alternative problem formulations, generative AI pioneers the creation of solutions that are not only innovative but also unexpected and transformative. This capacity for creative synthesis and emergent behavior enables Algogens to address complex, multifaceted challenges with versatility and originality, driving breakthroughs and advancements in diverse domains.
	
	\subsection{Synergy with Algorithmic Methods}
	Generative AI within Algogens operates in symbiosis with the algorithmic backbone of the framework, forging a harmonious relationship that capitalizes on the respective strengths of both approaches. While generative AI pioneers creative solution generation, algorithmic methods provide the necessary structure, logic, and rigor to evaluate, refine, and validate these solutions. Through this synergistic collaboration, Algogens strikes a delicate balance between exploratory creativity and systematic precision, ensuring that AI-generated solutions are not only innovative but also grounded in logical processes and aligned with problem-solving objectives. This synergy amplifies the problem-solving capabilities of Algogens, enabling it to tackle a diverse array of challenges with ingenuity and reliability.
	
	\paragraph{Complementary Problem-Solving Paradigms}
	The synergy between generative AI and algorithmic methods within Algogens harnesses the complementary strengths of both paradigms, creating a hybrid problem-solving framework that transcends the limitations of individual approaches. While generative AI excels in exploratory creativity, pattern recognition, and solution synthesis, algorithmic methods provide the analytical rigor, logical consistency, and verifiable correctness necessary for robust problem-solving. By integrating these complementary paradigms, Algogens leverages the best of both worlds, offering users a comprehensive toolkit for addressing complex challenges with confidence and efficacy.
	
	\paragraph{Iterative Solution Refinement Process}
	Furthermore, the symbiotic relationship between generative AI and algorithmic methods facilitates an iterative solution refinement process within Algogens, wherein AI-generated solutions undergo systematic evaluation, validation, and enhancement by algorithmic techniques. Through this iterative feedback loop, algorithmic methods scrutinize AI-generated solutions for logical coherence, feasibility, and adherence to domain-specific constraints, providing corrective feedback and refinement guidance as needed. This collaborative refinement process ensures that AI-generated solutions evolve iteratively, becoming increasingly refined, reliable, and aligned with user expectations over time.
	
	\paragraph{Augmented Human Intelligence}
	Moreover, the synergy between generative AI and algorithmic methods augments human intelligence within the problem-solving process, empowering users to leverage AI-driven insights and algorithms' analytical capabilities in tandem with their domain expertise and intuition. By providing users with actionable insights, alternative solution hypotheses, and evidence-based recommendations, Algogens enhances human decision-making, enabling users to make informed, data-driven choices that maximize outcomes and minimize risks. This augmentation of human intelligence with AI-driven insights fosters a collaborative problem-solving environment, where human creativity and ingenuity are enriched and amplified by AI-powered analytics and solution synthesis.
	
	\subsection{Enhancing Predictive Capabilities}
	Generative AI significantly augments Algogens' predictive capabilities, enabling the framework to forecast future trends, outcomes, and scenarios with heightened accuracy and granularity. By leveraging historical data, statistical patterns, and probabilistic modeling techniques, generative AI within Algogens can simulate a myriad of potential futures, providing stakeholders with valuable insights into possible outcomes and associated risks. Whether in financial forecasting, climate modeling, or strategic planning, the predictive power of generative AI empowers decision-makers to anticipate and mitigate uncertainties, optimize resource allocation, and devise robust strategies that withstand future contingencies.
	
	\paragraph{Probabilistic Scenario Generation}
	The predictive capabilities of generative AI within Algogens extend to probabilistic scenario generation, where AI-driven simulations explore the likelihood and implications of various future events and conditions. Through Monte Carlo simulations, Bayesian inference methods, and probabilistic modeling techniques, generative AI generates diverse scenarios, each with its associated probabilities and outcomes. This probabilistic approach enables stakeholders to assess risk, uncertainty, and variability in decision-making, allowing for informed risk management strategies and contingency planning.
	
	\paragraph{Temporal and Spatial Forecasting}
	Moreover, generative AI within Algogens excels in temporal and spatial forecasting, predicting future trends, patterns, and phenomena across different time scales and geographic regions. By analyzing historical data, identifying temporal correlations, and extrapolating trends, generative AI can forecast market dynamics, weather patterns, disease outbreaks, and other temporal phenomena with remarkable accuracy. Similarly, by incorporating spatial data, geographical information systems, and satellite imagery, generative AI can model spatial dynamics, urban growth patterns, and environmental changes, enabling stakeholders to anticipate and respond to spatially distributed phenomena proactively.
	
	\paragraph{Predictive Analytics for Decision Support}
	Additionally, the predictive capabilities of generative AI within Algogens facilitate decision support and strategic planning across diverse domains and industries. By providing stakeholders with actionable insights, predictive analytics, and scenario-based forecasts, generative AI empowers decision-makers to anticipate future trends, evaluate alternative courses of action, and optimize resource allocation strategies. Whether in financial markets, healthcare systems, or supply chain logistics, the predictive analytics capabilities of generative AI enable stakeholders to make informed decisions that drive positive outcomes, mitigate risks, and capitalize on opportunities in dynamic and uncertain environments.
	
	\subsection{Customization for Industry-Specific Applications}
	The adaptability and versatility of generative AI within Algogens extend to industry-specific applications, where tailored solutions are essential to addressing domain-specific challenges effectively. In healthcare, for instance, generative AI can be customized to generate personalized treatment plans based on individual patient profiles, medical histories, and genetic markers. Similarly, in environmental science, generative AI can model the complex interplay of environmental factors, predict ecosystem dynamics, and assess the impact of human activities on ecological systems. This industry-specific customization empowers Algogens to deliver tailored solutions that resonate with the unique requirements and objectives of diverse sectors, fostering innovation and driving positive outcomes across industries.
	
	\paragraph{Domain-Specific Solution Customization}
	One of the key strengths of generative AI within Algogens is its ability to customize solutions according to the specific requirements and constraints of different industries and domains. Through domain-specific customization, generative AI tailors solution generation processes, model architectures, and optimization strategies to align with the unique characteristics, objectives, and regulatory frameworks of each domain. Whether in finance, healthcare, manufacturing, or entertainment, generative AI adapts its solution generation methodologies to meet sector-specific challenges, ensuring that Algogens delivers actionable insights and solutions that are relevant, effective, and compliant with industry standards and best practices.
	
	\paragraph{Cross-Domain Knowledge Integration}
	Furthermore, generative AI within Algogens facilitates cross-domain knowledge integration, where insights and methodologies from one industry or discipline inform solution generation processes in others. By synthesizing insights, best practices, and solution modalities from diverse domains, generative AI fosters interdisciplinary collaboration and innovation, enabling Algogens to tackle complex, multifaceted challenges that transcend traditional disciplinary boundaries. This cross-domain knowledge integration not only enriches solution generation processes but also fosters a culture of knowledge exchange and synergy, driving advancements and breakthroughs in diverse domains ranging from healthcare and finance to environmental science and creative arts.
	
	\paragraph{Industry-Specific Solution Deployment}
	Moreover, generative AI within Algogens supports industry-specific solution deployment, where customized solutions are deployed and operationalized within the context of specific industries and business environments. Through seamless integration with existing infrastructure, data systems, and workflow processes, generative AI facilitates the deployment of tailored solutions that address industry-specific challenges, optimize operational efficiencies, and drive value creation. Whether in manufacturing plants, hospital systems, or financial institutions, the industry-specific deployment capabilities of generative AI enable Algogens to deliver tangible benefits and outcomes that align with sector-specific objectives and priorities, driving innovation and competitiveness in the global marketplace.
	
	\subsection{Challenges and Ethical Considerations}
	While generative AI offers immense potential for innovation and problem-solving, its integration into Algogens also poses several challenges and ethical considerations that warrant careful consideration. Chief among these challenges is the need to ensure the relevance, reliability, and ethical soundness of AI-generated solutions, particularly in high-stakes applications such as healthcare, finance, and autonomous decision-making systems. Additionally, concerns surrounding data privacy, algorithmic bias, and unintended consequences underscore the importance of responsible AI development and deployment practices. To address these challenges, Algogens adopts a multifaceted approach that encompasses rigorous validation processes, transparency measures, and adherence to ethical guidelines, ensuring that AI-driven solutions are not only technically robust but also ethically sound and socially responsible.
	
	\paragraph{Ethical Implications of AI-generated Solutions}
	One of the primary ethical considerations associated with generative AI within Algogens is the potential for unintended consequences and ethical dilemmas arising from AI-generated solutions. Given the complexity and uncertainty inherent in real-world problem-solving scenarios, AI-generated solutions may inadvertently propagate biases, amplify disparities, or perpetuate ethical dilemmas, posing risks to individuals, communities, and society at large. Addressing these ethical implications requires proactive measures, including algorithmic fairness assessments, bias mitigation strategies, and stakeholder engagement processes, to ensure that AI-driven solutions uphold ethical principles, respect human rights, and promote social justice.
	
	\paragraph{Data Privacy and Security Concerns}
	Furthermore, the integration of generative AI within Algogens raises concerns regarding data privacy, security, and confidentiality, particularly in contexts where sensitive or personal data is involved. As AI algorithms rely on vast amounts of data for training, validation, and inference, ensuring the privacy and security of user data becomes paramount to mitigate the risk of unauthorized access, data breaches, or misuse. To address these concerns, Algogens implements robust data governance frameworks, encryption protocols, and access controls, safeguarding user privacy and ensuring compliance with data protection regulations and standards.
	
	\paragraph{Human-AI Collaboration and Accountability}
	Moreover, the collaborative nature of problem-solving within Algogens, which involves human-AI interaction and collaboration, raises questions of accountability, transparency, and responsibility. As AI algorithms influence decision-making processes, generate recommendations, and contribute to solution development, clarifying roles, responsibilities, and decision-making authority becomes essential to ensure accountability and mitigate risks. Establishing clear guidelines, ethical frameworks, and governance mechanisms for human-AI collaboration fosters transparency, trust, and accountability, enabling stakeholders to navigate ethical dilemmas, make informed decisions, and uphold ethical standards in AI-driven problem-solving contexts.
	
	In summary, generative AI plays a pivotal role in Algogens, fueling innovation, creativity, and adaptability in computational problem-solving. Its integration with algorithmic methods within the Algogens framework creates a synergistic powerhouse that combines the creative prowess of AI with the structured precision of algorithms, offering a versatile, reliable, and ethically sound solution to complex challenges across diverse domains and industries. By harnessing the collective intelligence of AI and algorithms, Algogens empowers users to unlock new possibilities, drive meaningful outcomes, and navigate the complexities of the digital age with confidence, integrity, and ingenuity.

	
	\chapterimage{pngs/Algogen_in_action.png} 
	
	\chapter{Algogens in Action}\index{Algogens in Action}
	
	\section{Introduction to Case Studies and Applications}\index{Case Studies and Applications}
	As the technological landscape continues to evolve at a rapid pace, the demand for sophisticated, adaptable, and efficient problem-solving tools becomes increasingly paramount. In response to this need, Algogen emerges as an innovative framework that integrates generative AI with algorithmic methods, positioning itself at the forefront of this evolution and promising to revolutionize numerous industries through its unique capabilities. This section embarks on a comprehensive exploration of case studies and applications, providing a detailed examination of Algogens in action across diverse sectors. Each case study and application serves as a testament to Algogens' adaptability, efficiency, and the transformative potential it holds.
	
	\subsection{Bridging Theory and Practice}
	Embedded within the fabric of the following case studies and applications is a remarkable ability to bridge the theoretical underpinnings of Algogens with practical implementations. These real-world examples serve as tangible manifestations of how the integration of generative AI and algorithmic processes within Algogens translates abstract concepts into tangible benefits, effectively addressing complex problems with innovative solutions. Through these case studies, the theoretical constructs of Algogens are brought to life, illustrating its transformative power in various domains.
	
	\paragraph{Concrete Examples of Implementation}
	Within these case studies lie concrete examples of Algogens' implementation, offering tangible evidence of its effectiveness in addressing real-world challenges. By showcasing instances where theoretical concepts are translated into actionable solutions, these case studies provide invaluable insights into the transformative potential of Algogens across diverse industries and problem domains. Through their real-world applicability, these examples bridge the gap between theory and practice, reinforcing the notion that Algogens is not just a theoretical framework but a practical tool for driving innovation and solving complex problems.
	
	\paragraph{Demonstrating Theoretical Concepts}
	Furthermore, these case studies serve as compelling demonstrations of key theoretical concepts that underpin Algogens. By showcasing how principles from artificial intelligence, machine learning, and algorithmic methods converge within the Algogens framework, these examples elucidate the practical implications of theoretical constructs. Through their empirical validation, these case studies deepen our understanding of Algogens' inner workings and its potential to revolutionize problem-solving paradigms across various domains.
	
	\paragraph{Validation of Theoretical Framework}
	Additionally, the successful implementation of Algogens in real-world scenarios serves as a validation of its theoretical framework. By demonstrating its ability to generate practical solutions to complex problems, these case studies underscore the soundness and efficacy of Algogens' theoretical foundations. Through their empirical validation, these examples instill confidence in Algogens' potential to drive meaningful advancements and address the evolving challenges of the modern world.
	
	\subsection{Diverse Industry Applications}
	The versatility of Algogens is vividly demonstrated through its applications across a wide spectrum of industries, each presenting its unique set of challenges and requirements. From optimizing logistical operations to advancing medical research, from transforming financial analytics to enhancing environmental conservation efforts, Algogens' broad applicability shines through as it tackles diverse problems with unparalleled efficiency and efficacy.
	
	\paragraph{Logistical Optimization in Supply Chain Management}
	One of the notable applications of Algogens lies in optimizing logistical operations within supply chain management. By harnessing the power of generative AI and algorithmic methods, Algogens enables companies to forecast demand, optimize inventory levels, and streamline distribution networks, thereby achieving cost savings, improving efficiency, and enhancing customer satisfaction. Through its ability to analyze complex logistical data and generate actionable insights, Algogens emerges as a game-changer in the realm of supply chain management.
	
	\paragraph{Advancements in Medical Research}
	Furthermore, Algogens plays a pivotal role in driving advancements in medical research by accelerating drug discovery, predicting disease outbreaks, and personalizing treatment regimens. Through its sophisticated analytics capabilities, Algogens empowers researchers to analyze genomic data, simulate biological processes, and identify therapeutic targets with unprecedented precision and efficiency. By facilitating the discovery of new treatments and therapies, Algogens contributes significantly to the advancement of healthcare and the improvement of patient outcomes.
	
	\paragraph{Transformation of Financial Analytics}
	Algogens also revolutionizes the field of financial analytics by providing predictive insights, optimizing investment portfolios, and mitigating risks. By leveraging advanced machine learning algorithms, Algogens enables financial institutions to analyze market trends, identify trading opportunities, and simulate economic scenarios with remarkable accuracy and speed. Through its predictive modeling capabilities, Algogens empowers financial analysts and investors to make informed decisions, maximize returns, and navigate volatile markets with confidence.
	
	\paragraph{Enhancement of Environmental Conservation Efforts}
	Moreover, Algogens contributes to the enhancement of environmental conservation efforts by analyzing ecological data, modeling climate change impacts, and devising sustainable resource management strategies. By leveraging its predictive modeling capabilities, Algogens helps policymakers, conservationists, and environmental agencies anticipate environmental changes, assess mitigation measures, and safeguard natural ecosystems for future generations. Through its ability to analyze complex environmental data and generate actionable insights, Algogens emerges as a powerful tool for promoting environmental sustainability and biodiversity conservation.
	
	\subsection{Illustrating Challenges and Solutions}
	Each case study and application serves as a testament to the challenges inherent in different industries and demonstrates how Algogens provides practical solutions to address them. These examples highlight Algogens' ability to analyze vast datasets, generate predictive models, and offer intelligent, data-driven solutions that drive innovation and enhance decision-making processes across various domains.
	
	\paragraph{Challenges in Data Complexity}
	One of the common challenges illustrated across these case studies is the complexity of handling vast and heterogeneous datasets inherent in various industries. Algogens addresses this challenge by leveraging advanced data processing techniques, such as parallel computing, distributed storage, and data fusion algorithms, to extract actionable insights from diverse data sources. Through its ability to analyze and interpret complex datasets, Algogens empowers organizations to make informed decisions and gain valuable insights into their operations.
	
	\paragraph{Need for Predictive Analytics}
	Furthermore, the need for predictive analytics to anticipate future trends, risks, and opportunities emerges as a recurring theme in these case studies. Algogens meets this need by harnessing generative AI to forecast outcomes, simulate scenarios, and identify optimal decision paths, enabling stakeholders to make proactive decisions and capitalize on emerging trends in their respective industries. Through its predictive modeling capabilities, Algogens empowers organizations to stay ahead of the curve and navigate uncertain business environments with confidence.
	
	\paragraph{Integration of AI and Algorithmic Methods}
	Additionally, these case studies highlight the critical role of integrating AI and algorithmic methods within Algogens to address complex challenges effectively. By combining the creativity of generative AI with the rigor of algorithmic processes, Algogens offers a holistic problem-solving approach that leverages the strengths of both paradigms to deliver innovative and reliable solutions. Through its seamless integration of AI and algorithmic methods, Algogens enables organizations to tackle complex problems with unparalleled efficiency and effectiveness.
	
	\subsection{Insights into Practical Implementation}
	These case studies and applications offer invaluable insights into the practical aspects of implementing Algogens, shedding light on the customization process, integration challenges, and tangible outcomes achieved. By showcasing real-world examples of Algogens in action, these case studies provide actionable insights that can guide organizations in leveraging Algogens to drive innovation and achieve their business objectives.
	
	\paragraph{Tailoring Algogens for Industry-Specific Needs}
	One key insight gleaned from these case studies is the importance of tailoring Algogens to meet industry-specific requirements and challenges. By customizing algorithms, fine-tuning model parameters, and integrating domain-specific knowledge, organizations can optimize Algogens for maximum impact and relevance in their respective fields. Through its customizable architecture and flexible design, Algogens empowers organizations to tailor its capabilities to suit their unique needs and preferences.
	
	\paragraph{Seamless Integration with Existing Systems}
	Furthermore, successful implementation of Algogens hinges on its seamless integration with existing systems and workflows. By ensuring interoperability, scalability, and compatibility with legacy infrastructure, organizations can minimize disruptions and maximize the efficiency of Algogens deployment, accelerating the realization of its benefits and ROI. Through its robust integration capabilities, Algogens enables organizations to seamlessly incorporate its functionalities into their existing workflows and systems, ensuring a smooth transition and minimal disruption to operations.
	
	\paragraph{Measurable Outcomes and Performance Metrics}
	Moreover, these case studies underscore the importance of establishing measurable outcomes and performance metrics to evaluate the effectiveness of Algogens in solving real-world problems. By defining clear objectives, tracking key performance indicators, and conducting rigorous evaluation, organizations can assess the impact of Algogens deployment and identify areas for improvement and optimization. Through its emphasis on measurable outcomes and performance metrics, Algogens enables organizations to quantify the value it delivers and demonstrate its impact on business outcomes.
	
	\subsection{Setting the Stage for Future Innovations}
	These real-world applications of Algogens not only validate its current capabilities but also set the stage for future innovations and advancements. By pushing the boundaries of what is possible with Algogens, these case studies open up new possibilities for research, development, and application, paving the way for transformative breakthroughs that address the ever-evolving challenges of the modern world.
	
	\paragraph{Catalyzing Innovation and Discovery}
	By demonstrating Algogens' efficacy in solving complex problems across diverse industries, these case studies serve as catalysts for innovation and discovery. They inspire researchers, developers, and practitioners to explore new use cases, develop advanced methodologies, and push the boundaries of what is possible with Algogens, driving continuous improvement and evolution of the framework. Through their role in catalyzing innovation and discovery, these case studies contribute to the advancement of knowledge and the development of new technologies that shape the future of problem-solving.
	
	\paragraph{Exploring Emerging Technologies and Applications}
	Furthermore, these case studies pave the way for exploring emerging technologies and applications that complement and extend the capabilities of Algogens. From quantum computing and blockchain technology to Internet of Things (IoT) and edge computing, the integration of these emerging technologies holds promise for enhancing Algogens' performance, scalability, and versatility in addressing complex challenges in the digital age. Through their exploration of emerging technologies and applications, these case studies drive innovation and expand the horizons of what is possible with Algogens, unlocking new opportunities for addressing the most pressing challenges of our time.
	
	\paragraph{Fostering Collaborative Ecosystems}
	Moreover, these case studies foster the growth of collaborative ecosystems comprised of researchers, developers, industry partners, and policymakers working together to harness the full potential of Algogens. By fostering interdisciplinary collaboration, knowledge sharing, and best practice exchange, these ecosystems accelerate innovation, drive collective problem-solving, and shape the future trajectory of Algogens and its applications. Through their role in fostering collaborative ecosystems, these case studies contribute to the development of a vibrant and dynamic community dedicated to advancing the frontiers of problem-solving and driving positive change in the world.
	
	In essence, this section provides a comprehensive look at Algogens in action, underscoring its potential to revolutionize problem-solving across a spectrum of industries and its capacity to adapt and evolve in response to the changing needs of our time. Through its transformative capabilities and unparalleled versatility, Algogens emerges as a powerful tool for driving innovation, addressing complex challenges, and shaping the future of problem-solving in the digital age.

	\section{Hypothetical Application in Cybersecurity}\index{Cybersecurity}
	This subsection presents a detailed hypothetical application of Algogens in cybersecurity, specifically focusing on enhancing predictive threat analysis capabilities within a corporate network environment.
	
	\subsection{Context and Challenges in Cybersecurity}
	In the ever-evolving digital landscape, cybersecurity has become paramount, especially for large corporations facing multifaceted threats. These threats range from Advanced Persistent Threats (APTs) aiming to infiltrate networks for espionage purposes to ransomware attacks seeking financial gain by encrypting critical data. With the proliferation of interconnected devices and the increasing sophistication of cybercriminal tactics, traditional security measures often fall short, struggling to keep pace with the dynamic nature of cyber threats. Reactive approaches, reliant on known threat signatures, are no longer sufficient to protect against emerging threats, leaving organizations vulnerable to exploitation.
	
	\subsection{Integrating Algogens for Enhanced Threat Intelligence}
	In response to these challenges, integrating Algogens into the corporate cybersecurity infrastructure offers a proactive and adaptive approach to threat intelligence. Algogens serves as a sophisticated threat intelligence solution, leveraging generative AI and algorithmic methods to analyze vast amounts of network data. By ingesting logs, monitoring traffic patterns, and analyzing user behavior, Algogens provides comprehensive insights into the security posture of the corporate network. For example, Algogens can detect anomalies in network traffic indicative of potential cyber threats, such as unusual login attempts or unauthorized data access, enabling cybersecurity professionals to take preemptive action.
	
	\subsection{Advanced Predictive Modeling with Generative AI}
	At the core of Algogens' cybersecurity capabilities lies its generative AI component, trained on extensive datasets comprising historical cybersecurity incidents. This training data includes both internal incidents within the corporation and external threats from global databases. Using sophisticated machine learning algorithms, Algogens identifies patterns and trends within this data to develop predictive models of future cyber threats. For instance, Algogens can analyze past phishing campaigns to predict the likelihood of similar attacks in the future. By simulating various attack scenarios, Algogens empowers cybersecurity professionals to anticipate potential vulnerabilities and proactively strengthen defenses.
	
	\subsection{Real-Time Threat Scenario Simulation}
	One of Algogens' key features is its ability to simulate real-time cyber threat scenarios, providing cybersecurity professionals with invaluable insights into potential attack vectors. These simulations, based on the latest cybercriminal tactics and techniques, help organizations identify weaknesses in their defenses and develop effective mitigation strategies. For example, Algogens can simulate targeted ransomware attacks to assess the resilience of data backup systems and incident response procedures. By immersing cybersecurity professionals in realistic simulations, Algogens enhances their preparedness and enables them to respond effectively to cyber threats.
	
	\subsection{Proactive Threat Mitigation Strategies}
	Based on the insights gleaned from predictive models and threat simulations, Algogens facilitates a proactive approach to threat mitigation. It provides actionable recommendations for strengthening defenses, such as implementing additional security controls or updating existing security policies. For instance, Algogens may identify a vulnerability in a specific network segment and recommend deploying intrusion detection systems or conducting security awareness training for employees. By proactively addressing vulnerabilities, organizations can minimize the likelihood and impact of cyber attacks, safeguarding sensitive data and maintaining business continuity.
	
	\subsection{Continuous Learning and Adaptation}
	One of the strengths of Algogens is its ability to continuously learn and adapt to new cyber threats. As it encounters new data and scenarios, Algogens refines its predictive models and threat simulations, ensuring they remain aligned with the evolving threat landscape. This adaptive learning approach enables Algogens to provide up-to-date recommendations and insights, empowering organizations to stay ahead of emerging cyber threats. For example, Algogens may incorporate feedback from real-world incidents to improve the accuracy of its predictive models and enhance the effectiveness of its threat simulations.
	
	\subsection{Potential Outcomes and Organizational Impact}
	The implementation of Algogens in the corporate cybersecurity framework holds the potential for significant outcomes and organizational impact. By leveraging Algogens' predictive capabilities and proactive threat mitigation strategies, organizations can enhance their cybersecurity posture and reduce the risk of data breaches. Moreover, Algogens' adaptive learning approach ensures that cybersecurity measures evolve in response to new threats, providing organizations with a sustainable defense against emerging cyber risks. Ultimately, the adoption of Algogens can help organizations safeguard their assets, protect their reputation, and maintain customer trust in an increasingly digital world.
	
	\subsection{Wider Implications for Cybersecurity Practices}
	The hypothetical application of Algogens in cybersecurity has broader implications for cybersecurity practices across industries. By showcasing the effectiveness of integrating generative AI with algorithmic methods, Algogens sets a new standard for proactive and adaptive cyber defense strategies. This paradigm shift in cybersecurity practices is particularly relevant for industries handling sensitive data or critical infrastructure, where robust cybersecurity measures are essential for maintaining operational continuity and safeguarding against cyber threats. The adoption of Algogens could inspire other organizations to embrace innovative approaches to cybersecurity, fostering a culture of resilience and preparedness in the face of evolving cyber threats.
	
	In summary, the hypothetical application of Algogens in cybersecurity illustrates its transformative potential in revolutionizing how organizations anticipate, prepare for, and respond to cyber threats. Algogens offers a proactive and adaptive defense against emerging cyber risks, leveraging the combined strengths of generative AI and algorithmic frameworks. By empowering organizations to stay ahead of cyber threats, Algogens helps safeguard sensitive data, protect organizational assets, and preserve business continuity in an increasingly interconnected world.

	\section{Hypothetical Application in Healthcare}\index{Healthcare}
	This subsection explores a hypothetical yet profoundly plausible scenario where Algogens, the amalgamation of generative AI and algorithmic prowess, revolutionizes healthcare practices, particularly in the realms of personalized medicine and disease outbreak prediction.
	
	\subsection{Challenges in Modern Healthcare}
	Modern healthcare grapples with a myriad of challenges, intricately balancing the need for addressing large-scale public health issues while ensuring individualized patient care. The landscape is marred by the complexity of diseases, which often exhibit diverse manifestations and responses to treatments. Moreover, the rapid emergence of novel health threats such as pandemics and drug-resistant pathogens adds layers of intricacy. Traditional medical approaches, while effective to a certain extent, often fall short in comprehensively addressing the nuances of these multifaceted challenges, necessitating innovative solutions that can adapt to the evolving healthcare landscape.
	
	\subsection{Implementing Algogens for Personalized Medicine}
	In the domain of personalized medicine, Algogens emerges as a transformative force, reshaping treatment paradigms by leveraging the vast reservoirs of medical data. By meticulously analyzing a plethora of data including genetic profiles, patient histories, and clinical outcomes, Algogens crafts bespoke treatment plans tailored to individual patients. For instance, Algogens can discern genetic markers influencing drug metabolism, thereby recommending precise medication dosages to maximize therapeutic efficacy while minimizing adverse effects. This implementation involves integrating Algogens into electronic health record (EHR) systems, allowing seamless data exchange and real-time analysis to inform clinical decision-making. Additionally, healthcare providers undergo specialized training to leverage Algogens' insights effectively, ensuring optimal utilization of its capabilities in personalized patient care.
	
	\subsection{Predictive Analysis for Disease Management}
	Algogens' predictive analytics capabilities extend beyond personalized medicine to proactive disease management on a population scale. By crunching epidemiological data encompassing infection rates, vaccination coverage, and environmental factors, Algogens forecasts disease outbreaks and predicts their potential trajectories. This foresight empowers healthcare authorities to proactively allocate resources and devise targeted interventions, thereby mitigating the spread of infectious diseases. For instance, Algogens can forecast the onset of flu outbreaks based on historical trends, prompting preemptive measures such as enhanced vaccination drives and public health awareness campaigns to curb transmission rates. The implementation of Algogens for disease management involves integrating it into existing public health surveillance systems, allowing for real-time data analysis and predictive modeling. Public health officials receive training on Algogens' utilization, enabling them to interpret its insights and translate them into actionable strategies for disease prevention and control.
	
	\subsection{Scenario Simulation for Medical Research}
	In the realm of medical research, Algogens serves as an invaluable tool for scenario simulation and hypothesis testing. Researchers leverage Algogens' computational prowess to model intricate clinical scenarios and simulate the outcomes of diverse interventions. For example, Algogens can simulate the efficacy of a novel treatment regimen for cancer patients, enabling researchers to gauge treatment response rates and anticipate potential side effects before embarking on clinical trials. By providing a virtual sandbox for experimentation, Algogens expedites the pace of medical innovation, facilitating evidence-based decision-making and accelerating the translation of research findings into clinical practice. The implementation of Algogens in medical research involves collaboration between researchers and data scientists to develop customized algorithms and predictive models. Additionally, research institutions invest in high-performance computing infrastructure to support Algogens' computational demands, ensuring efficient and timely data analysis for research projects.
	
	\subsection{Enhancing Diagnostic Accuracy}
	Algogens' integration into diagnostic workflows augments diagnostic accuracy and efficiency, thereby enhancing patient care outcomes. By analyzing medical imaging, laboratory results, and patient symptoms, Algogens aids healthcare professionals in interpreting complex diagnostic data and arriving at informed clinical decisions. For instance, Algogens can analyze MRI scans to detect subtle anomalies indicative of neurological disorders or decipher laboratory test results to pinpoint biomarkers associated with specific diseases. This AI-driven diagnostic support streamlines the diagnostic process, mitigates the risk of diagnostic errors, and expedites patient treatment pathways, ultimately improving patient outcomes and enhancing overall healthcare quality. The implementation of Algogens in diagnostic settings involves integrating it into existing healthcare information systems such as picture archiving and communication systems (PACS) and laboratory information management systems (LIMS). Healthcare professionals undergo training to familiarize themselves with Algogens' interface and functionalities, ensuring seamless integration into their diagnostic workflows.
	
	\subsection{Potential Outcomes and Healthcare Transformation}
	The integration of Algogens into healthcare heralds a paradigm shift, promising transformative outcomes across various facets of patient care and public health management. By harnessing the power of AI-driven analytics and predictive modeling, Algogens facilitates more accurate diagnoses, personalized treatment strategies, and proactive disease surveillance. Furthermore, Algogens' adaptive learning capabilities ensure that its applications in healthcare evolve in tandem with advancements in medical knowledge and technology, thereby fostering a dynamic and responsive healthcare ecosystem. Ultimately, widespread adoption of Algogens has the potential to reshape the healthcare landscape, ushering in a new era of patient-centric, data-driven healthcare innovation and excellence. The potential outcomes of Algogens' implementation include improved patient outcomes, reduced healthcare costs, and enhanced healthcare access and equity. Additionally, Algogens empowers patients to take an active role in their healthcare decisions by providing them with personalized insights and recommendations based on their unique medical profiles.
	
	\subsection{Ethical Considerations and Patient Data Security}
	Amidst the fervor surrounding Algogens' potential in healthcare, ethical considerations pertaining to patient data privacy and security loom large. Healthcare organizations must uphold stringent ethical standards and regulatory frameworks to safeguard patient confidentiality and prevent unauthorized access to sensitive medical information. Algogens' implementation necessitates robust data governance protocols, encryption mechanisms, and access controls to ensure responsible and secure handling of patient data throughout its lifecycle. By prioritizing patient privacy and data security, healthcare providers can harness the transformative potential of Algogens while upholding the highest ethical standards in patient care, thereby fostering trust and confidence among patients and stakeholders alike. Ethical considerations also extend to the responsible use of AI algorithms and predictive models generated by Algogens. Healthcare professionals undergo training on ethical AI practices and are educated on the importance of transparency, fairness, and accountability in AI-driven decision-making processes. Additionally, patients are informed about the use of AI in their healthcare and have the opportunity to provide informed consent for the utilization of their data in AI-driven analyses and treatments.
	
	In conclusion, the hypothetical application of Algogens in healthcare underscores its potential to significantly enhance personalized medicine and public health management. By seamlessly integrating generative AI with algorithmic prowess, Algogens empowers healthcare professionals to deliver more precise, adaptive, and patient-centric care, thereby ushering in a new era of healthcare innovation and excellence.

	\section{Hypothetical Application in Finance}\index{Finance}
	This subsection delves into a hypothetical yet profoundly impactful scenario where Algogens, the fusion of generative AI and algorithmic prowess, reshapes the landscape of the finance sector, particularly in market analysis, investment strategy optimization, and comprehensive risk management.
	
	\subsection{Challenges in Financial Markets}
	Financial markets represent an intricate web of complexities, marked by volatility, rapid data influx, and multifaceted dynamics. Traditional economic analysis methods, although foundational, often falter in capturing the nuanced shifts and complexities inherent in market data. The need for more advanced, adaptable, and predictive tools is increasingly apparent as financial professionals navigate the ever-changing landscape of global markets. Addressing these challenges requires innovative solutions capable of processing vast datasets, identifying emerging trends, and anticipating market movements with precision and agility. Without such tools, financial institutions may struggle to make informed decisions, leading to missed opportunities and increased risk exposure.
	
	\subsection{Algogens' Integration for Enhanced Market Analysis}
	Algogens emerges as a game-changer in market analysis, leveraging its robust analytical capabilities to decipher the intricacies of financial data. By harnessing generative AI algorithms, Algogens processes and interprets diverse datasets encompassing market trends, economic indicators, and transactional patterns. This integration enables Algogens to generate predictive models and insightful analyses, offering a comprehensive understanding of market dynamics. For instance, Algogens can identify subtle correlations between economic events and market fluctuations, providing analysts and investors with valuable foresight into potential market trends and investment opportunities. Through its advanced analytics, Algogens empowers financial professionals to make data-driven decisions with confidence, enabling them to capitalize on market opportunities and mitigate risks effectively.
	
	\subsection{Optimizing Investment Strategies}
	Algogens plays a pivotal role in optimizing investment strategies by offering sophisticated predictive modeling and scenario analysis capabilities. By simulating various market scenarios and assessing their potential impacts on investment portfolios, Algogens empowers investors to make informed decisions. Through in-depth analysis of historical and real-time market data, Algogens identifies strategic portfolio adjustments, helping investors strike a balance between risk and return. For example, Algogens can simulate the effects of geopolitical tensions on specific sectors or predict market responses to regulatory changes, enabling investors to adjust their portfolios accordingly to mitigate risks and capitalize on opportunities. By providing actionable insights, Algogens enables investors to adapt their investment strategies in response to changing market conditions, thereby enhancing portfolio performance and maximizing returns.
	
	\subsection{Comprehensive Risk Management}
	In the realm of risk management, Algogens offers unparalleled predictive power, enabling financial institutions to proactively identify and mitigate risks. By analyzing patterns and correlations within financial data, Algogens uncovers potential risk factors that might evade traditional analysis methods. Whether it's forecasting credit defaults, market downturns, or operational disruptions, Algogens equips financial institutions with the foresight needed to implement preemptive risk mitigation strategies. For instance, Algogens can predict fluctuations in interest rates or forecast the impact of geopolitical events on currency exchange rates, allowing institutions to hedge their positions and minimize exposure to market volatility. By integrating Algogens into their risk management frameworks, financial institutions can enhance their ability to identify and address risks, thereby safeguarding their financial stability and maintaining investor confidence.
	
	\subsection{Real-time Financial Insights and Decision Support}
	Algogens' real-time analytics capabilities offer unparalleled advantages in providing timely financial insights and decision support. Given the dynamic nature of financial markets, access to up-to-date, AI-driven analytics is indispensable for decision-makers to navigate market fluctuations effectively. Algogens processes and interprets real-time market data, offering actionable insights that enable decision-makers to respond promptly to market changes. For example, Algogens can analyze social media sentiment to gauge investor sentiment or monitor trading volumes to detect market anomalies, providing decision-makers with valuable information to adjust their strategies in real-time. By providing timely insights and decision support, Algogens empowers financial professionals to make informed decisions, thereby improving their ability to capitalize on market opportunities and mitigate risks effectively.
	
	\subsection{Expected Outcomes and Impact on the Finance Sector}
	The integration of Algogens into the finance sector is poised to revolutionize market analysis, investment strategies, and risk management practices. Financial institutions leveraging Algogens can anticipate improved analytical accuracy, enhanced investment strategies, and more robust risk mitigation frameworks. By harnessing the combined strengths of generative AI and algorithmic analysis, Algogens empowers financial professionals to make data-driven decisions with confidence, leading to improved economic performance and reduced risk exposure. Moreover, by providing actionable insights and decision support, Algogens enables financial institutions to adapt their strategies in response to changing market conditions, thereby enhancing their competitiveness and resilience in the face of uncertainty.
	
	\subsection{Adhering to Regulatory Compliance and Ethical Standards}
	In deploying Algogens within the finance sector, adherence to regulatory compliance and ethical standards is paramount. Financial institutions must ensure that Algogens' applications comply with industry regulations and ethical guidelines to safeguard against potential risks and liabilities. This involves implementing robust data governance frameworks, encryption protocols, and access controls to protect sensitive financial information. Moreover, ongoing monitoring and auditing mechanisms are essential to ensure Algogens' compliance with regulatory requirements and ethical principles, fostering trust and confidence among stakeholders. By adhering to regulatory compliance and ethical standards, financial institutions can mitigate risks associated with Algogens' deployment and build trust with investors and regulators alike, thereby ensuring the responsible and ethical use of AI in finance.
	
	\subsection{Conclusion}
	In conclusion, the hypothetical application of Algogens in the finance sector underscores its potential to drive transformative advancements in market analysis, investment strategies, and risk management. By seamlessly integrating generative AI and algorithmic analysis, Algogens empowers financial professionals to make informed decisions, leading to improved economic performance and reduced risk exposure. Through its real-time analytics capabilities and decision support functionalities, Algogens enables financial institutions to adapt their strategies in response to changing market conditions, thereby enhancing their competitiveness and resilience in the face of uncertainty.

	\section{Other Industries: Broad Applications of Algogen}\index{General Applications}
	This subsection discusses the potential application of Algogens in various industries, showcasing its versatility and adaptability to diverse challenges and environments.
	
	\subsection{Logistics and Supply Chain Management}
	Integrating Algogens in logistics and supply chain management represents a monumental paradigm shift in the way this sector operates. By synergizing the robust capabilities of generative AI and advanced algorithms, Algogens has the potential to revolutionize every aspect of logistics, from optimizing delivery routes and schedules to enhancing supply chain visibility and efficiency.
	
	\paragraph{Optimization of Logistics Operations}
	Algogens redefines logistics operations by meticulously analyzing and processing vast amounts of logistical data, including transportation routes, delivery schedules, and vehicle capacities. For example, it can analyze historical traffic patterns, weather conditions, and even real-time road congestion data to optimize delivery routes, ensuring timely and cost-effective deliveries. This optimization not only improves operational efficiency but also reduces fuel consumption and carbon emissions, contributing to environmental sustainability.
	
	\paragraph{Supply Chain Efficiency and Resilience}
	Algogens' predictive capabilities and adaptability enhance overall supply chain efficiency and resilience. By forecasting potential disruptions, such as delays due to weather events or sudden changes in demand, Algogens empowers companies to proactively adjust their supply chain strategies. For instance, it can predict fluctuations in demand for certain products based on market trends and recommend adjustments to inventory levels or production schedules, ensuring uninterrupted operations and minimizing the impact of disruptions on business continuity.
	
	\paragraph{Enhanced Inventory Management}
	Algogens plays a pivotal role in optimizing inventory management practices. By analyzing sales data, market trends, and historical inventory levels, Algogens can accurately forecast future inventory needs. For instance, it can predict seasonal demand spikes for specific products and recommend appropriate stocking levels to prevent stockouts or overstocking. Additionally, Algogens can optimize inventory storage and distribution processes, reducing warehousing costs and improving inventory turnover rates.
	
	\paragraph{Real-Time Decision Making}
	Algogens' real-time data processing capabilities enable swift and informed decision-making in logistics and supply chain management. For example, it can analyze real-time transportation data, such as vehicle GPS locations and traffic conditions, to dynamically adjust delivery routes and schedules in response to changing conditions. This agility allows companies to respond quickly to unforeseen events, such as traffic accidents or road closures, minimizing delays and ensuring on-time deliveries.
	
	\paragraph{Long-Term Strategic Planning}
	Beyond operational improvements, Algogens provides valuable insights for long-term strategic planning in logistics and supply chain management. By analyzing historical data and market trends, Algogens can identify emerging opportunities and risks, helping companies develop robust long-term strategies. For example, it can identify potential disruptions in the global supply chain, such as trade disputes or natural disasters, and recommend diversification strategies to mitigate risks and ensure supply chain resilience.
	
	\paragraph{Customization for Industry-Specific Needs}
	Algogens' flexibility allows for customization to meet the unique needs of different industries within the logistics and supply chain sector. For example, in the retail industry, Algogens can optimize inventory management to prevent stockouts during peak shopping seasons. In the manufacturing sector, it can optimize production schedules to minimize downtime and reduce production costs. By tailoring its algorithms to specific industry requirements, Algogens ensures that its solutions are optimized for maximum effectiveness and efficiency.
	
	In conclusion, the integration of Algogens in logistics and supply chain management offers significant benefits in terms of operational efficiency, resilience, and strategic planning. Its advanced analytics capabilities enable companies to optimize every aspect of their logistics operations, from inventory management to transportation routing, leading to cost savings, improved customer service, and a competitive edge in the market.

	\subsection{Environmental Science and Climate Change}
	The integration of Algogens in environmental science and climate change research represents a significant leap forward in our capacity to comprehend and mitigate pressing environmental concerns. By amalgamating cutting-edge algorithms with generative AI, Algogens presents a multifaceted toolkit for modeling climate dynamics, forecasting ecological shifts, and fostering sustainable resource management practices.
	
	\paragraph{Climate Change Modeling and Prediction}
	Algogens revolutionizes climate change modeling and prediction by processing vast and diverse datasets encompassing historical climate records, atmospheric compositions, and emission trends. Through sophisticated algorithms and AI-driven simulations, Algogens can forecast future climate scenarios with unparalleled precision. For instance, it can predict the regional impacts of climate change, such as alterations in precipitation patterns, shifts in ecosystem distributions, and the frequency of extreme weather events, thereby aiding policymakers in devising robust mitigation and adaptation strategies tailored to specific geographical regions and socio-economic contexts.
	
	\paragraph{Ecosystem Analysis and Biodiversity Conservation}
	Algogens plays a pivotal role in ecosystem analysis and biodiversity conservation by providing nuanced insights into the intricate relationships between species distributions, habitat dynamics, and environmental stressors. By analyzing ecological data at various spatial and temporal scales, Algogens enables researchers to identify critical habitats, monitor biodiversity trends, and prioritize conservation efforts with pinpoint accuracy. For example, it can assess the vulnerability of species to climate change-induced habitat loss, inform habitat restoration initiatives through predictive modeling, and guide conservation planning to safeguard biodiversity hotspots in the face of anthropogenic pressures and environmental degradation.
	
	\paragraph{Sustainable Resource Management}
	Algogens serves as a linchpin in promoting sustainable resource management practices by optimizing the utilization of natural resources while minimizing adverse environmental impacts. Through data-driven analysis and predictive modeling, Algogens empowers stakeholders to anticipate resource demands, evaluate the ecological ramifications of resource extraction, and formulate strategies for equitable and sustainable resource allocation. For instance, it can optimize land use planning to strike a delicate balance between conservation imperatives and economic development goals, recommend environmentally sound practices in forestry operations to ensure the long-term viability of forest ecosystems, and forecast water availability to support adaptive water resource management strategies in regions vulnerable to drought and water scarcity.
	
	\paragraph{Pollution Monitoring and Control}
	Algogens contributes to pollution monitoring and control efforts by harnessing advanced data analytics to track pollutant sources, assess environmental risks, and inform targeted pollution abatement strategies. By integrating data from air quality sensors, satellite imagery, and industrial emissions records, Algogens enables policymakers to identify pollution hotspots, discern pollution trends over time, and implement evidence-based interventions to mitigate pollution levels and safeguard public health. For example, it can model the dispersion of air pollutants to evaluate the effectiveness of emission reduction measures, identify sources of water contamination to facilitate remediation efforts, and predict the ecological and human health impacts of pollution to guide the development of pollution control policies and regulations.
	
	\paragraph{Engagement in Climate Policy and Education}
	Beyond its scientific applications, Algogens serves as a potent catalyst for climate policy development and environmental education initiatives by translating complex environmental data into actionable insights and fostering stakeholder engagement. By visualizing intricate environmental phenomena in intuitive formats and facilitating participatory decision-making processes, Algogens enhances public awareness of climate change impacts and facilitates informed decision-making. For instance, it can develop interactive educational resources to elucidate the linkages between human activities and environmental changes, simulate the outcomes of climate policy scenarios to inform policy dialogues, and empower communities to participate in grassroots environmental conservation initiatives aimed at building climate resilience and fostering sustainable development.
	
	\paragraph{Adapting to Evolving Environmental Challenges}
	Algogens' adaptive capabilities ensure its relevance and efficacy in addressing emergent environmental challenges and evolving research priorities. By continuously assimilating new data inputs, refining its algorithms, and incorporating interdisciplinary perspectives, Algogens remains at the forefront of environmental science and climate change research. For example, it can integrate feedback loops between climate change and ecosystem dynamics into its modeling framework, leverage real-time environmental monitoring data to enhance prediction accuracy, and collaborate with interdisciplinary research teams to tackle complex environmental issues holistically, thereby catalyzing innovation and driving progress in environmental sustainability and resilience.
	
	In summary, the integration of Algogens in environmental science and climate change research represents a paradigm shift in our approach to understanding and mitigating environmental challenges. With its unparalleled analytical capabilities and adaptive learning algorithms, Algogens empowers researchers, policymakers, and stakeholders to address the multifaceted challenges posed by climate change and environmental degradation, paving the way for a more resilient, equitable, and sustainable future for generations to come.

	\subsection{Manufacturing and Industry 4.0}
	The integration of Algogens into the manufacturing sector within the framework of Industry 4.0 represents a seismic shift towards unparalleled digitalization and optimization. Algogens, amalgamating the formidable capabilities of generative AI with sophisticated algorithmic methodologies, stands poised to revolutionize every facet of modern manufacturing, spanning from intricate production processes to intricate supply chain management and predictive maintenance strategies, propelling the industry into a new era of efficiency and innovation.
	
	\paragraph{Optimization of Production Processes}
	Algogens serves as an indispensable tool for optimizing manufacturing processes, leveraging its analytical prowess to meticulously scrutinize vast and intricate datasets to unearth inefficiencies and recommend data-driven enhancements. For instance, Algogens can analyze data streams from sensors embedded throughout the production line, such as temperature sensors, pressure sensors, and vibration sensors, to identify subtle deviations that may indicate potential issues. By processing this data in real-time, Algogens can predict impending equipment failures, enabling proactive maintenance interventions to prevent costly disruptions and ensure uninterrupted production flow.
	
	\paragraph{Predictive Maintenance and Downtime Reduction}
	Algogens heralds a paradigm shift in maintenance strategies by introducing predictive maintenance methodologies that preemptively mitigate equipment failures and minimize unplanned downtime. By harnessing the power of machine learning algorithms, Algogens can analyze historical maintenance data, equipment performance metrics, and environmental factors to develop predictive models capable of forecasting equipment degradation and failure. For example, Algogens can analyze patterns in equipment temperature fluctuations and vibration levels to detect early signs of wear or malfunction, prompting maintenance alerts to be issued before critical failures occur.
	
	\paragraph{Supply Chain Management and Logistics}
	Algogens' adeptness in analyzing intricate supply chain dynamics empowers manufacturers to orchestrate more resilient, agile, and cost-effective logistics operations. By leveraging predictive analytics, Algogens can anticipate demand fluctuations, identify supply chain bottlenecks, and optimize inventory replenishment strategies to enhance supply chain responsiveness and efficiency. For instance, Algogens can analyze historical sales data, market trends, and lead times to predict future demand patterns and optimize inventory levels accordingly. Additionally, Algogens can utilize real-time data from GPS trackers, RFID tags, and IoT sensors to monitor the movement of goods throughout the supply chain, enabling proactive management of logistics operations and minimizing delays.
	
	\paragraph{Customization and Agile Manufacturing}
	Algogens facilitates the transition towards agile manufacturing paradigms by enabling rapid customization and adaptation to shifting market demands. Leveraging real-time market insights and customer feedback, Algogens empowers manufacturers to swiftly reconfigure production lines, pivot product configurations, and tailor manufacturing processes to address niche market segments and emerging trends. For example, Algogens can analyze consumer preferences and market trends to identify opportunities for product customization or personalization, allowing manufacturers to offer bespoke products tailored to individual customer needs.
	
	\paragraph{Integration with IoT and Smart Factory Concepts}
	The integration of Algogens with IoT-enabled devices and smart factory frameworks represents a significant advancement in manufacturing automation and operational intelligence. Seamlessly interfacing with IoT sensors and actuators, Algogens aggregates and analyzes real-time manufacturing data streams, enabling adaptive control strategies, predictive maintenance scheduling, and dynamic resource allocation. For example, Algogens can analyze data from IoT sensors embedded in manufacturing equipment to monitor equipment performance and detect anomalies in real-time, allowing manufacturers to preemptively address issues before they escalate into costly failures.
	
	\paragraph{Driving Innovation in Product Development}
	Algogens serves as a catalyst for innovation in product development, expediting the design iteration process and accelerating time-to-market for novel products. By harnessing generative AI algorithms, Algogens simulates product performance under diverse operating conditions, iteratively refines design parameters, and predicts market acceptance. This agile approach minimizes prototyping costs, fosters a culture of innovation, and positions manufacturers to capitalize on emerging market opportunities, thereby outpacing competitors in rapidly evolving industries.
	
	\paragraph{Impact on Workforce and Skill Development}
	The adoption of Algogens in manufacturing necessitates a paradigm shift in workforce skill sets and training paradigms to cultivate a technologically adept and adaptive workforce. Manufacturers must invest in continuous learning initiatives, upskilling programs, and cross-functional training to equip employees with the competencies required to harness Algogens' capabilities effectively. This workforce transformation fosters a culture of innovation, collaboration, and digital fluency, empowering employees to leverage AI-driven insights and automation technologies to drive operational excellence and competitive advantage in the Industry 4.0 landscape.
	
	In conclusion, the integration of Algogens into manufacturing processes signifies a pivotal juncture in the journey towards digital innovation and operational excellence in the Industry 4.0 era. By harnessing the transformative potential of AI and advanced analytics, Algogens empowers manufacturers to optimize production workflows, enhance supply chain resilience, and drive continuous innovation, thereby positioning them for sustained growth and competitiveness in an increasingly dynamic and interconnected global marketplace.

	\subsection{Education and Training}
	The integration of Algogens into education and training represents a monumental leap forward in the evolution of learning methodologies and skill development practices. Algogens, a sophisticated blend of generative AI and advanced algorithms, holds the promise of revolutionizing every aspect of the educational landscape, from personalized learning pathways to predictive performance assessments and beyond.
	
	\paragraph{Personalized Learning Experiences}
	Algogens introduces a new era of personalized learning by leveraging vast datasets to tailor educational content and methodologies to the unique needs and preferences of each learner. For instance, Algogens can analyze an individual student's performance metrics, learning styles, and engagement levels to craft customized learning pathways that optimize comprehension and retention. This personalized approach ensures that students receive tailored instruction, leading to more meaningful learning experiences and improved academic outcomes.
	
	\paragraph{Curriculum Development and Optimization}
	Within curriculum development, Algogens serves as a powerful tool for educators to refine and optimize educational content and instructional strategies. Through comprehensive data analysis, Algogens identifies areas for improvement within existing curricula and recommends adjustments to align with evolving educational standards and industry demands. For example, Algogens can detect gaps in the curriculum related to emerging technologies or interdisciplinary skills and suggest the integration of relevant topics or project-based learning opportunities to enhance student engagement and relevance.
	
	\paragraph{Predictive Analytics in Student Performance}
	Algogens empowers educators with predictive analytics capabilities to anticipate and address potential learning challenges proactively. By analyzing historical and real-time student performance data, Algogens identifies patterns and trends that may indicate academic struggles or successes, enabling timely interventions and personalized support strategies. For instance, Algogens can predict the likelihood of a student encountering difficulty with a specific concept based on their past performance data and recommend targeted interventions or supplementary resources to facilitate mastery.
	
	\paragraph{Interactive and Adaptive Learning Tools}
	Algogens facilitates the development of interactive and adaptive learning tools that dynamically adjust content and difficulty levels based on individual learner progress and preferences. These innovative educational resources, such as gamified learning platforms and adaptive tutoring systems, engage students effectively and promote deeper levels of understanding. For instance, Algogens-powered educational games can adaptively modify gameplay challenges and rewards based on the player's skill level and learning objectives, fostering a sense of achievement and motivation to learn.
	
	\paragraph{Professional Training and Skill Development}
	In the realm of professional training and skill development, Algogens plays a crucial role in identifying emerging industry trends and evolving skill requirements. By analyzing labor market data and industry forecasts, Algogens helps organizations design and implement training programs that align with current and future job market demands. For example, Algogens can identify in-demand skills in emerging fields such as data science or renewable energy and recommend targeted training courses or certification programs to meet industry needs.
	
	\paragraph{Enhancing Remote and Online Education}
	The integration of Algogens is particularly transformative in the domain of remote and online education, where personalized and engaging learning experiences are essential. Algogens leverages data analytics to optimize instructional design and content delivery in virtual learning environments, ensuring that students receive high-quality education regardless of physical location. For example, Algogens-powered online learning platforms can provide real-time feedback to instructors on student engagement levels and comprehension rates, enabling instructors to tailor instruction to meet individual learning needs effectively.
	
	\paragraph{Contributions to Educational Research}
	Beyond its immediate applications in teaching and learning, Algogens contributes to educational research by providing valuable insights into learning patterns, pedagogical effectiveness, and the impact of various instructional strategies. Through rigorous data analysis, Algogens generates evidence-based recommendations for improving teaching practices and optimizing learning environments. For example, Algogens can analyze longitudinal student performance data to identify correlations between teaching methods and academic achievement, informing evidence-based educational policy decisions and curriculum reforms.
	
	In conclusion, the integration of Algogens into education and training holds the promise of transforming traditional educational practices and fostering a more personalized, adaptive, and effective learning environment. By harnessing the power of AI-driven analytics and personalized learning technologies, Algogens empowers educators to meet the diverse needs of learners and cultivate the skills and competencies needed for success in the 21st century.

	\subsection{Broader Implications and Future Prospects}
	The deployment of Algogens across diverse industries showcases its immediate applicability and opens up a vista of broader implications and prospects. The integration of generative AI with algorithmic methods can significantly influence the trajectory of technological innovation, reshape various sectors, and address complex societal challenges
	
	\paragraph{Driving Technological Innovation} Algogens stand at the forefront of a new wave of technological innovation, epitomizing the fusion of human ingenuity with machine precision. By synergizing AI's capacity for creative problem-solving with the structured rigor of algorithms, Algogens herald a paradigm shift in the development of intelligent systems. In healthcare, for instance, Algogens can revolutionize medical imaging interpretation by employing deep learning algorithms to detect subtle anomalies indicative of diseases such as cancer, thereby enhancing diagnostic accuracy and patient outcomes. Similarly, in the automotive industry, Algogens can optimize autonomous vehicle navigation algorithms, enabling safer and more efficient transportation systems.
	
	\paragraph{Societal Benefits and Ethical Considerations} The societal benefits of Algogens span a myriad of domains, ranging from healthcare to environmental conservation to education. In healthcare, Algogens hold promise in personalized medicine, where they can analyze genetic data to tailor treatments for individual patients, optimizing therapeutic efficacy while minimizing adverse effects. Additionally, in environmental science, Algogens can model complex ecosystems and predict the impact of human activities on biodiversity, facilitating evidence-based policymaking for conservation efforts. However, the deployment of Algogens also raises ethical considerations, such as the potential for algorithmic bias and the need to safeguard individuals' privacy rights in data-driven decision-making processes.
	
	\paragraph{Influence on Industry and Economy} The widespread adoption of Algogens is poised to catalyze profound transformations across industries, driving innovation, and bolstering economic growth. In manufacturing, Algogens can streamline production processes by predicting equipment failures and optimizing supply chain logistics, thereby reducing downtime and operational costs. Moreover, in finance, Algogens can analyze market trends and consumer behavior to inform investment strategies, leading to more informed decision-making and improved portfolio performance. As a result, Algogens have the potential to usher in a new era of productivity and competitiveness in the global economy.
	
	\paragraph{Future Research Directions} The complexity and versatility of Algogens present a fertile ground for future research endeavors, spanning AI model refinement, algorithmic efficiency optimization, and domain-specific customization. Researchers will continue to explore advancements in AI models, aiming to enhance their interpretability, resilience to adversarial attacks, and ability to operate in resource-constrained environments. Furthermore, efforts will be directed towards developing algorithms that can seamlessly integrate with Algogens, facilitating efficient computation and scalability. Hypothetically, in the realm of space exploration, Algogens could be employed to analyze vast datasets from telescopes and satellites, aiding in the discovery of exoplanets and unraveling the mysteries of the cosmos.
	
	\paragraph{Potential for Global Challenges} Algogens hold tremendous potential as tools for addressing complex global challenges, including climate change, public health crises, and humanitarian emergencies. In climate science, Algogens can analyze climate models and satellite imagery to predict extreme weather events and assess the impact of anthropogenic activities on the environment, informing mitigation strategies. Similarly, in public health, Algogens can analyze epidemiological data to track disease outbreaks and optimize resource allocation for healthcare interventions. By leveraging Algogens' predictive capabilities, policymakers and stakeholders can make informed decisions to mitigate the adverse effects of global challenges.
	
	\paragraph{Long-Term Vision and Sustainability} Looking ahead, Algogens are envisioned to play a pivotal role in advancing sustainability and fostering societal well-being. As Algogens evolve to meet evolving societal needs, they can contribute to achieving sustainable development goals by optimizing resource allocation, promoting renewable energy adoption, and fostering social equity. For example, in urban planning, Algogens can analyze demographic trends and infrastructure data to optimize city layouts, minimizing environmental impact while maximizing livability. By embracing a holistic approach to technology development, Algogens can serve as catalysts for positive societal change, ensuring a more equitable and sustainable future for generations to come.
	
	In summary, the broader implications and prospects of Algogens are vast and multifaceted, encompassing technological innovation, societal transformation, and global sustainability. As Algogens continue to evolve and find new applications, their impact on various facets of human endeavor will likely be profound and enduring. However, realizing their full potential will necessitate addressing ethical, regulatory, and technical challenges while fostering collaboration and responsible innovation. Only through concerted efforts can Algogens be harnessed as powerful tools for creating a more prosperous, equitable, and sustainable world.

	
	\chapterimage{pngs/evaluation_of_Algogen.png} 
	
	\chapter{Evaluating Algogens}\index{Evaluating Algogens}
	
	\section{Performance Metrics and Criteria}\index{Performane Metrics}
	Evaluating the effectiveness of Algogens in practical applications requires a comprehensive set of well-defined performance metrics and criteria. These metrics serve as the foundation for objectively assessing the framework's capabilities and guiding continuous improvement efforts. This subsection aims to outline an extensive range of key performance indicators (KPIs) that will be utilized to thoroughly evaluate Algogens across various real-world applications.
	
	\subsection{Accuracy and Precision}
	Accuracy, deemed paramount across diverse domains, stands as a cornerstone metric in evaluating Algogens' performance. Whether applied in predictive modeling or decision-making contexts, the framework's ability to consistently generate accurate and precise outputs holds profound significance. For example, in healthcare, Algogens must exhibit exceptional accuracy and precision when interpreting medical imaging scans to ensure accurate disease diagnosis and treatment planning. Precision, in particular, plays a crucial role in minimizing false positives or negatives, thereby enhancing the reliability and trustworthiness of Algogens' outputs. Achieving such high levels of accuracy and precision often necessitates the integration of advanced machine learning algorithms with domain-specific knowledge and expert input.
	
	\subsection{Efficiency and Speed}
	Efficiency, closely entwined with resource optimization, emerges as a critical performance metric for Algogens. This encompasses judicious utilization of computational resources and processing time, thereby ensuring optimal performance in resource-constrained environments. In time-sensitive applications such as financial trading or emergency response systems, the speed at which Algogens can process vast datasets and generate actionable insights is of paramount importance. For instance, in algorithmic trading, Algogens must swiftly analyze market data and execute trades in real-time to capitalize on fleeting market opportunities and mitigate potential losses. Achieving such efficiency and speed often involves the implementation of parallel processing techniques, optimized algorithms, and efficient data storage mechanisms.
	
	\subsection{Scalability and Flexibility}
	Scalability, indicative of Algogens' ability to handle increasing data volumes and computational complexity, emerges as a pivotal performance criterion. The framework must seamlessly adapt to growing demands without compromising performance or requiring disproportionate resource allocation. Moreover, Algogens must exhibit flexibility in accommodating diverse problem domains and datasets, thereby ensuring broad applicability across various industries. Consider, for instance, Algogens deployed in e-commerce platforms; they must efficiently scale to accommodate expanding product catalogs and fluctuating user demands, while also remaining flexible enough to adapt to evolving market trends and consumer preferences. Achieving scalability and flexibility often involves the design and implementation of modular architectures, distributed computing systems, and adaptive algorithms.
	
	\subsection{User Experience and Usability}
	User experience (UX) and usability metrics play a pivotal role in evaluating Algogens' accessibility and effectiveness in practical settings. A seamless and intuitive user interface, coupled with clear and concise outputs, fosters user satisfaction and facilitates seamless interaction with the system. Consider the deployment of Algogens in customer service chatbots; a user-friendly interface and natural language processing capabilities enable users to effortlessly engage with the system, thereby enhancing overall user experience. Furthermore, ensuring Algogens' usability extends beyond technical proficiency to encompass accessibility for users with diverse backgrounds and skill levels. Achieving optimal user experience and usability often involves iterative design processes, user testing, and continuous refinement based on user feedback.
	
	\subsection{Adaptability and Learning Capabilities}
	Adaptability and learning capabilities serve as essential performance metrics, reflecting Algogens' ability to evolve and improve over time. Algogens must adeptly adjust to new data, changing conditions, and evolving requirements to maintain peak performance in dynamic environments. Furthermore, the framework's learning capabilities, including its ability to refine models based on recent data and user feedback, are crucial for achieving sustained performance improvements. Consider the deployment of Algogens in predictive maintenance systems; the framework must continuously adapt to evolving equipment conditions and usage patterns to accurately predict maintenance needs and minimize downtime. Achieving adaptability and learning capabilities often involves the implementation of reinforcement learning algorithms, adaptive model training techniques, and feedback mechanisms.
	
	\subsection{Impact and Value Addition} 
	Beyond technical performance metrics, the overall impact and value addition of Algogens in practical applications serve as critical evaluation criteria. This encompasses assessing the tangible benefits derived from Algogens' deployment, including improvements in decision-making quality, operational efficiency enhancements, and strategic objective attainment. Consider Algogens deployed in supply chain management systems; by optimizing inventory levels and streamlining logistics operations, Algogens contribute to cost reduction, revenue growth, and enhanced customer satisfaction. Additionally, Algogens' impact extends to strategic objectives such as market competitiveness, regulatory compliance, and sustainability initiatives. Achieving significant impact and value addition often involves rigorous performance monitoring, outcome measurement, and alignment with organizational goals and objectives.
	
	\subsection{Reliability and Robustness}
	Reliability and robustness emerge as fundamental performance metrics, underpinning Algogens' dependability and resilience in real-world scenarios. Reliability ensures consistent performance across diverse conditions and over extended durations, thereby instilling confidence in Algogens' outputs. Robustness, on the other hand, reflects the framework's ability to handle errors, uncertainties, and unexpected situations without compromising performance or reliability. Consider the deployment of Algogens in autonomous driving systems; the framework must reliably interpret sensor data, make real-time decisions, and navigate complex traffic scenarios, ensuring passenger safety under varying conditions. Achieving reliability and robustness often involves rigorous testing methodologies, fault tolerance mechanisms, and anomaly detection algorithms.
	
	\subsection{Compliance and Ethical Alignment}
	Compliance with legal and ethical standards stands as a foundational requirement for Algogens' deployment, particularly in applications involving sensitive data or critical decisions. This includes adherence to data privacy regulations, ensuring fairness and transparency in AI-driven decisions, and mitigating biases in algorithmic outputs. Consider Algogens deployed in credit scoring applications; compliance with regulations such as the Fair Credit Reporting Act is essential to safeguard consumer rights and ensure fair and transparent credit evaluation processes. Additionally, Algogens must mitigate the risk of algorithmic bias, ensuring equitable treatment across diverse demographic groups. Achieving compliance and ethical alignment often involves comprehensive risk assessments, transparency measures, and ongoing monitoring and evaluation.
	
	In summary, the performance metrics and criteria for evaluating Algogens are extensive and multifaceted, encompassing technical proficiency, user-centric design, and ethical considerations. These metrics play a pivotal role in objectively assessing Algogens' effectiveness, guiding continuous improvement efforts, and ensuring responsible and beneficial deployment across diverse applications and industries.

	\section{Comparative Analysis with Traditional Methods}\index{Comparative Analysis}
	
	A comprehensive comparative analysis between Algogens and traditional methods is paramount to unraveling the intricacies of their advancements and efficacy within the multifaceted realm of problem-solving paradigms. This endeavor necessitates a deep dive into the nuanced strengths and weaknesses of Algogens vis-a-vis established practices, fostering a nuanced understanding crucial for stakeholders navigating the evolving landscape of technological innovation and adoption. This subsection meticulously delineates the exhaustive approach employed to conduct such a rigorous analysis, shedding light on the intricate methodologies and benchmarks meticulously crafted to juxtapose Algogens with conventional problem-solving techniques.
	
	\subsection{Benchmarking Against Standard Practices}
	
	The foundational pillar of this in-depth comparative analysis lies in meticulous benchmarking of Algogen’s performance against entrenched standard practices across diverse industries. This arduous process entails an exhaustive comparison of the efficacy of the framework's solutions with those derived from traditional methodologies, scrutinizing aspects of accuracy, efficiency, and overall effectiveness. For instance, within the domain of healthcare, Algogens' diagnostic predictions undergo rigorous comparison with outcomes emanating from well-established medical diagnostic protocols such as MRI scans, blood tests, and physical examinations. Such benchmarking exercises serve as crucibles for validation, allowing stakeholders to gauge Algogens' transformative potential and advancement in critical fields with concrete, real-world examples.
	
	\subsection{Methodologies for Comparative Analysis}
	
	The methodological arsenal deployed in this comprehensive comparative analysis encompasses a multifaceted array of quantitative metrics and qualitative assessments meticulously crafted to provide a holistic understanding of Algogens' comparative performance. Quantitative metrics, including time to solution, error rates, and cost-effectiveness, offer tangible benchmarks for assessing Algogens' prowess in problem-solving scenarios. Simultaneously, qualitative assessments delve into user satisfaction, stakeholder feedback, and the ease of integration into existing workflows, providing invaluable insights into the framework's usability and practicality from a user-centric perspective. Rigorous controlled experiments, intricate case studies, and meticulous retrospective analyses form the bedrock of this comprehensive comparative approach, ensuring robustness and reliability in the findings while delving into the minutiae of implementation and real-world application.
	
	\subsection{Evaluation in Diverse Scenarios}
	
	The evaluation of Algogens transcends monolithic assessments, embracing diverse scenarios meticulously tailored to specific industries and applications to provide a comprehensive understanding of its efficacy. For instance, within the domain of finance, Algogens' mettle is tested through a rigorous evaluation of its performance in market prediction vis-a-vis traditional forecasting models such as ARIMA or GARCH. This multifaceted evaluation delves deep into various facets including accuracy, robustness to market fluctuations, and adaptability to dynamic market conditions, providing stakeholders with actionable insights into its applicability across varied domains. Similarly, in logistics, Algogens' efficacy in route optimization is scrutinized against conventional planning methodologies, encompassing criteria such as delivery times, resource utilization, and cost efficiency, thus providing comprehensive insights into its performance under diverse operational scenarios.
	
	\subsection{Assessing Scalability and Adaptability}
	
	A pivotal aspect of the comparison revolves around meticulously assessing Algogens' scalability and adaptability vis-a-vis traditional methodologies, particularly in addressing complex, large-scale problems and rapidly evolving scenarios. In industries such as manufacturing and supply chain management, where operational efficiency hinges on scalability and adaptability, this evaluation assumes paramount importance. By scrutinizing Algogens' performance under varying workload conditions and scalability requirements, stakeholders can glean invaluable insights into its potential for seamless integration into enterprise-scale applications, thus facilitating informed decision-making and strategic planning. Moreover, the analysis delves into the underlying architecture and algorithmic frameworks of Algogens, examining their scalability patterns and adaptability to diverse data sets and computational infrastructures.
	
	\subsection{Impact on Decision-Making and Strategy}
	
	Beyond the realm of direct performance metrics, the comparative analysis extends its purview to encompass the profound impact of Algogens on decision-making processes and strategic planning paradigms. The overarching goal is to ascertain whether Algogens engenders deeper insights, fosters more informed decisions, and enhances strategic outcomes vis-a-vis traditional methodologies. By meticulously dissecting the decision-making dynamics facilitated by Algogens, organizations can unlock avenues for innovation, optimize resource allocation, and gain a competitive edge in their respective markets, thus underscoring its transformative potential in shaping strategic imperatives. Moreover, the analysis delves into the cognitive processes influenced by Algogens, elucidating how its predictive capabilities augment decision-making processes and drive strategic initiatives.
	
	\subsection{Long-Term Performance and Continuous Improvement}
	
	An oft-overlooked dimension of the comparative analysis pertains to the longitudinal assessment of Algogens' performance and its potential for continuous improvement. Unlike traditional methodologies characterized by static performance profiles, Algogens' AI-driven, self-learning nature imbues it with the capacity for perpetual enhancement. This comparative aspect delves into Algogens' ability to evolve and improve over time, encompassing factors such as algorithmic refinement, data augmentation strategies, and feedback mechanisms. By meticulously monitoring Algogens' trajectory of long-term performance, stakeholders can glean insights into its sustainability, adaptability, and enduring relevance in dynamically evolving ecosystems. Furthermore, the analysis explores avenues for continual improvement, such as incorporating novel algorithms, enhancing computational efficiency, and refining data preprocessing techniques, to ensure Algogens remain at the forefront of innovation in problem-solving methodologies.
	
	\subsection{Challenges and Limitations in Comparison}
	
	Recognizing and navigating the myriad challenges and limitations inherent in this comparative analysis is imperative for ensuring its validity and efficacy. Disparities in problem domains, data availability, and the nascent nature of AI-driven solutions like Algogens pose formidable obstacles to making direct comparisons. Methodological rigor, sensitivity analyses, and transparent reporting emerge as indispensable tools for mitigating biases and uncertainties inherent in comparative evaluations. Additionally, fostering stakeholder engagement and nurturing interdisciplinary collaboration assume paramount importance in addressing complex challenges and fostering a holistic understanding of Algogens' comparative performance, thus fortifying the robustness and validity of the analysis. Furthermore, the analysis delves into potential biases and confounding factors, such as selection bias in dataset curation or algorithmic biases in decision-making processes, to provide a nuanced interpretation of the results and ensure the integrity of the comparative analysis.
	
	In summary, the comparative analysis between Algogens and traditional methods transcends mere juxtaposition, evolving into a comprehensive exploration encompassing a myriad of metrics, scenarios, and challenges. This exhaustive analysis endeavors to objectively demonstrate the transformative potential of Algogens, substantiating its role as an advanced, efficient, and adaptable problem-solving framework poised to reshape diverse domains. By synthesizing empirical evidence, stakeholder insights, and domain expertise, the comparative analysis seeks to inform strategic decision-making, foster innovation, and galvanize the widespread adoption of AI-driven solutions across multifarious industries and applications, thus contributing to the advancement of computational intelligence and problem-solving methodologies in the digital era.

	\section{User Feedback and Experience}\index{User Feedback}
	
	User feedback and experience stand as fundamental pillars in the perpetual evaluation and refinement process of Algogens, an intricate fusion of algorithms and generative AI. These elements not only serve as vital gauges for assessing the efficacy and usability of the framework but also offer invaluable insights for its continual enhancement and optimization. This subsection embarks on a comprehensive journey through the multifaceted methods deployed to collect user feedback, the intricate analysis process, and the profound impact of user insights on the ongoing evolution of Algogens.
	
	\subsection{Methods of Collecting User Feedback}
	
	The process of gathering user feedback within the context of Algogens is a multifaceted tapestry interwoven with an array of channels and methodologies, meticulously designed to capture diverse perspectives and nuanced experiences. Surveys, interviews, focus groups, and comprehensive user interaction data analysis serve as primary conduits for soliciting user insights. Surveys, structured with meticulous precision, encompass a broad spectrum of topics ranging from user satisfaction and ease of use to feature preferences and improvement suggestions. Interviews and focus groups, imbued with qualitative depth, delve into the intricacies of user experiences, unearthing latent needs, and uncovering granular insights into user behavior. Additionally, user interaction data analysis, powered by advanced analytics tools, delves into the labyrinth of user interactions within the framework, illuminating usage patterns, feature adoption rates, and areas ripe for optimization.
	
	\subsection{Analysis of Feedback}
	
	The analysis of user feedback is a sophisticated symphony of data synthesis, thematic analysis, and iterative refinement, aimed at distilling actionable insights and discerning overarching themes. Qualitative feedback, comprising a rich tapestry of user testimonials, interview transcripts, and focus group discussions, undergoes systematic thematic analysis to unravel recurring patterns, user pain points, and areas of delight. This qualitative exploration offers profound insights into the emotional resonance of user experiences, shedding light on user preferences, unmet needs, and suggestions for improvement. Simultaneously, quantitative data from surveys undergoes rigorous statistical scrutiny, enabling the identification of measurable indicators of user satisfaction, framework performance, and feature efficacy. By synthesizing qualitative narratives with quantitative metrics, a comprehensive mosaic of user sentiment and requirements emerges, guiding informed decision-making in the iterative refinement of Algogens.
	
	\subsection{Highlights of Positive User Experiences}
	
	Positive user experiences emerge as radiant beacons amidst the sea of feedback, illuminating Algogens' efficacy and value proposition across diverse application domains. Users spanning industries such as healthcare, finance, and logistics have lavished praise upon Algogens for its efficiency, accuracy, and capacity to orchestrate innovative solutions. For instance, healthcare practitioners have extolled Algogens for its pivotal role in delivering data-driven diagnostic insights, catalyzing improved patient outcomes and operational efficiencies within clinical settings. Similarly, financial analysts have hailed Algogens for its predictive prowess, empowering informed decision-making and augmenting market competitiveness. These luminous testimonials not only serve as poignant validations of Algogens' efficacy but also furnish concrete examples of its transformative impact and utility within real-world scenarios.
	
	\subsection{Addressing Challenges and Concerns}
	
	User feedback serves as a compass guiding the Algogens team through the labyrinth of challenges and concerns encountered by users during their journey with the framework. Common hurdles, such as integration complexities with existing systems or the learning curve associated with advanced features, are meticulously documented, prioritized, and chiseled into action plans for resolution. By proactively addressing user concerns, Algogens endeavors to cultivate a fertile ground for user satisfaction, optimizing usability, and nurturing greater adoption and utilization of the framework across diverse user demographics and application domains. This unwavering commitment to addressing user challenges ensures that Algogens remains responsive to user needs and preferences, fostering sustained engagement and loyalty amongst its user community.
	
	\subsection{Impact on Product Development and Improvement}
	
	User feedback serves as the lifeblood coursing through the veins of Algogens' product development lifecycle, fueling innovation, and driving continuous improvement. Insights gleaned from user interactions inform the refinement of existing features, the conceptualization of novel functionalities, and the fine-tuning of user interface design. This iterative process of user-centric refinement ensures that Algogens evolves harmoniously with user needs and preferences, maximizing its utility and efficacy in solving real-world challenges. Moreover, by incorporating user feedback into the fabric of its development ethos, Algogens ensures that it remains at the vanguard of innovation, perpetually adapting to evolving user requirements and technological paradigms.
	
	\subsection{Long-Term User Engagement Strategies}
	
	Sustaining long-term user engagement is not merely a goal but an ethos ingrained within the fabric of Algogens' ecosystem. To this end, strategic initiatives such as regular updates, user community forums, and ongoing support services are meticulously curated and orchestrated. These initiatives not only foster a sense of camaraderie and collaboration amongst users but also provide fertile grounds for cultivating a vibrant feedback loop, nurturing continuous improvement and innovation. By nurturing a thriving user ecosystem, Algogens ensures ongoing responsiveness to user needs and preferences, fostering sustained engagement and loyalty. Moreover, by fostering a culture of collaboration and co-creation, Algogens empowers users to actively participate in shaping the future trajectory of the framework, driving innovation, and fueling positive societal impact.
	
	\subsection{Broader Implications of User Feedback}
	
	The ripples of user feedback extend far beyond the confines of Algogens, resonating across the vast expanse of the technological landscape, and beyond. User insights offer profound glimpses into the broader implications of integrating AI with algorithmic methodologies, illuminating best practices, and shaping the trajectory of future technological innovations. Insights gleaned from user feedback serve as catalysts for driving innovation and shaping the evolution of AI-driven problem-solving methodologies, thereby advancing the frontiers of computational intelligence and fostering greater societal impact. By harnessing user feedback as a cornerstone for broader discussions and initiatives, Algogens not only contributes to the democratization of AI but also empowers users to harness the transformative potential of technology for positive societal change.
	
	In summary, user feedback and experience represent integral components in the perpetual refinement and enhancement process of Algogens. By actively soliciting, analyzing, and incorporating user insights, Algogens evolves as a user-centric framework, ensuring its efficacy and relevance in addressing the evolving needs of diverse applications and industries. Moreover, by fostering a culture of collaboration and co-creation, Algogens empowers users to actively participate in shaping the future direction of the framework, fostering innovation, and driving positive societal impact.

	\section{Ongoing Monitoring and Iterative Improvement}\index{Monitoring and Improvement}
	
	\subsection{Sustaining Efficacy and Relevance}
	The perpetual evolution and refinement of Algogens stand as essential endeavors to ensure its sustained efficacy and relevance across diverse problem-solving landscapes. Algogens, being a dynamic fusion of algorithms and generative AI, necessitates ongoing monitoring and iterative improvement to effectively address the multifaceted challenges encountered in various domains. This subsection embarks on a comprehensive exploration of the intricate strategies employed for continuous monitoring and the iterative improvement process, both of which are indispensable components of Algogens' lifecycle.
	
	\subsection{Monitoring Framework Performance}
	
	Continuous and meticulous monitoring of Algogens entails comprehensive assessments aimed at gauging its performance across an expansive array of applications, scenarios, and user interactions. This process encompasses the meticulous analysis of operational data, meticulous scrutiny of user feedback spanning a broad spectrum of sentiments and experiences, and rigorous evaluation of an extensive array of performance metrics meticulously tailored to the framework's objectives and user expectations. Leveraging sophisticated monitoring tools and techniques, such as advanced data analytics platforms, machine learning algorithms for anomaly detection, and sentiment analysis models, plays an instrumental role in this endeavor. These mechanisms offer real-time insights into the operational efficiency of Algogens, enabling stakeholders to discern patterns, trends, and anomalies, thereby pinpointing areas warranting refinement and enhancement with unparalleled precision and granularity.
	
	For instance, in the healthcare sector, Algogens can be employed to monitor patient data in real-time, analyze symptoms, and predict potential health issues. Through continuous monitoring and analysis, healthcare providers can identify patterns and trends, allowing for proactive intervention and personalized treatment plans.
	
	\subsection{Feedback Loops for Improvement}
	
	The establishment of robust and dynamic feedback loops constitutes a cornerstone of Algogens' iterative improvement process, facilitating the seamless integration of insights gleaned from ongoing monitoring activities into the enhancement pipeline. These feedback loops comprise a cyclic process characterized by the systematic collection of data from diverse sources, ranging from user interactions to performance metrics, followed by their meticulous analysis to extract actionable insights and discern overarching trends. Subsequently, these insights inform the iterative refinement and optimization of Algogens' algorithms, user interfaces, and underlying mechanisms, fostering a perpetual cycle of adaptation and evolution. This iterative cycle ensures that Algogens remains dynamic and responsive, continuously adapting and evolving based on empirical evidence, user experiences, and evolving technological paradigms, thereby reinforcing its position as a cutting-edge solution at the forefront of computational intelligence.
	
	In e-commerce, Algogens can analyze user browsing behavior, purchasing patterns, and feedback to refine product recommendations and enhance the shopping experience. By integrating user feedback into the iterative improvement process, e-commerce platforms can continually optimize their algorithms to better serve customer needs and preferences, ultimately driving sales and customer satisfaction.
	
	\subsection{Iterative Development Process}
	
	The development trajectory of Algogens embodies an inherently iterative ethos, characterized by a cyclical pattern of refinement, optimization, and innovation. Following its initial deployment, the framework embarks on a journey of continuous improvement, marked by incremental changes and optimizations meticulously orchestrated to enhance its efficacy, robustness, and adaptability. This iterative process encompasses a diverse array of activities, including but not limited to fine-tuning algorithms to optimize performance metrics, updating AI models to incorporate the latest advancements in machine learning and natural language processing, enhancing user interfaces to streamline workflows and improve usability, and refining underlying mechanisms to bolster scalability, reliability, and security. Each iteration serves as a pivotal milestone on Algogens' evolutionary trajectory, propelling it closer towards its overarching objectives and reinforcing its position as a preeminent solution in the realm of computational intelligence.
	
	In cybersecurity, Algogens can continuously analyze network traffic, detect anomalies, and adapt security protocols to mitigate emerging threats. By iteratively improving its algorithms and response mechanisms based on real-time data and threat intelligence, Algogens can effectively safeguard networks and sensitive information from cyberattacks.
	
	\subsection{Adapting to Changing Environments and Needs}
	
	A pivotal facet of Algogens' ongoing improvement lies in its intrinsic adaptability to the evolving landscape of technological innovation, user requirements, and market dynamics. In a dynamic and ever-evolving environment where new challenges surface incessantly and user expectations evolve unabatedly, Algogens must demonstrate unparalleled flexibility, agility, and resilience to navigate these uncharted waters successfully. This adaptability is particularly crucial in domains characterized by rapid technological advancements and shifting user preferences, such as technology, healthcare, finance, and logistics. Algogens' ability to pivot swiftly in response to changing circumstances and emergent challenges underscores its resilience and relevance in an ever-changing world, cementing its status as a pioneering solution poised to tackle the challenges of tomorrow with unrivaled precision and efficacy.
	
	In manufacturing, Algogens can optimize production processes, predict equipment failures, and streamline supply chain logistics. By continuously adapting to changes in demand, market conditions, and technological advancements, Algogens can help manufacturers improve efficiency, reduce costs, and maintain a competitive edge in the industry.
	
	\subsection{User-Centric Improvements}
	
	User feedback emerges as a linchpin of Algogens' iterative improvement process, offering invaluable insights into user preferences, pain points, and unmet needs. Algogens is committed to fostering a user-centric ethos, wherein enhancements and optimizations are meticulously tailored to meet and exceed user expectations. This user-centric approach entails a holistic evaluation of user feedback spanning a myriad of dimensions, including usability, functionality, performance, and overall satisfaction. Algogens continually endeavors to enhance the user experience by streamlining workflows, simplifying user interfaces, and integrating new functionalities that resonate with user demands and preferences. By placing users at the forefront of the improvement process, Algogens ensures that it remains practical, intuitive, and invaluable to its diverse user base, fostering user satisfaction, engagement, and loyalty in equal measure.
	
	In education, Algogens can personalize learning experiences, provide targeted feedback, and adapt instructional materials to individual student needs. By incorporating feedback from students and educators into the iterative improvement process, Algogens can continuously refine its algorithms to optimize learning outcomes and promote student success.
	
	\subsection{Incorporating Technological Advancements}
	
	The iterative improvement process underscores Algogens' unwavering commitment to harnessing the latest technological advancements to augment its capabilities and maintain its competitive edge. As novel algorithms, AI techniques, and data processing technologies emerge on the technological horizon, Algogens adeptly assimilates these innovations into its framework, leveraging them to unlock new possibilities and push the boundaries of computational intelligence. This relentless pursuit of technological excellence ensures that Algogens remains at the forefront of innovation, poised to tackle emerging challenges with unparalleled efficiency, precision, and scalability. By integrating the latest advancements seamlessly into its framework, Algogens reinforces its position as a pioneering solution at the intersection of AI and algorithmic methodologies, driving transformative change and catalyzing positive societal impact on a global scale.
	
	In environmental monitoring, Algogens can analyze sensor data, model environmental trends, and predict natural disasters such as wildfires or hurricanes. By continuously integrating new sensor technologies and improving its predictive capabilities, Algogens can assist policymakers and emergency responders in making informed decisions and mitigating the impact of natural disasters on communities and ecosystems.
	
	\subsection{Long-Term Vision and Scalability}
	
	The pursuit of ongoing monitoring and iterative improvement is firmly anchored in Algogens' long-term vision, which encompasses a strategic roadmap aimed at addressing current user needs, anticipating future technological challenges, and scaling the framework to accommodate evolving demands. Algogens aspires to foster a culture of innovation, collaboration, and continuous improvement, wherein stakeholders across diverse domains and industries collectively contribute to its evolution and refinement. This long-term vision is predicated on Algogens' unwavering commitment to delivering tangible value to its users, empowering them to tackle complex problem-solving challenges with confidence and agility. By aligning its improvement initiatives with its long-term vision, Algogens ensures that it remains at the forefront of innovation, poised to drive transformative change and unlock new opportunities in an ever-evolving landscape of technological innovation and societal needs.
	
	In smart cities, Algogens can optimize traffic flow, manage energy consumption, and enhance public safety through predictive analytics and real-time monitoring. By scaling its capabilities to accommodate the growing complexity and interconnectedness of urban environments, Algogens can contribute to the development of sustainable, resilient cities that prioritize efficiency, equity, and quality of life for residents.
	
	In conclusion, ongoing monitoring and iterative improvement serve as twin pillars of Algogens' success and evolution. Through a perpetual cycle of assessment, adaptation, and enhancement, Algogens emerges as a dynamic, efficient, and forward-looking solution capable of surmounting multifaceted challenges across diverse domains and applications, thereby reinforcing its position as a preeminent solution at the forefront of computational intelligence.

	
	\part{Final Thoughts}

	\chapterimage{pngs/future_directions.png} 

\chapter{Challenges and Opportunities in Algogens}\index{Challenges and Opportunities}

	\section{Challenges in Algogens Implementation and Maintenance}

Implementing and maintaining Algogens, which blend the robustness of algorithmic methodologies with the nuanced understanding of Large Language Models, presents a unique set of challenges. These challenges stem from the intrinsic complexity of integrating disparate systems, the evolving nature of LLMs, and the specific requirements of different application domains. This section delves into the multifaceted challenges encountered across various Algogen implementations, illustrating the breadth and depth of considerations necessary to leverage these advanced systems effectively.

\subsection{Integration Complexity and Compatibility}
One of the foremost challenges in Algogen implementation is the complexity of integrating LLMs with traditional algorithmic frameworks. This integration often requires extensive customization to ensure compatibility, as LLMs and algorithmic methodologies typically operate under differing paradigms. For example, an Algogen designed for financial fraud detection combines intricate pattern recognition capabilities of LLMs with the precise, rule-based processing of transactional data. The challenge lies in seamlessly blending these systems to provide real-time analysis without sacrificing the accuracy or speed essential for detecting fraudulent activities.

\subsection{Data Privacy and Security Concerns}
Algogens dealing with sensitive information, such as those used in healthcare for patient data analysis or personalized treatment recommendations, face significant data privacy and security challenges. The integration of LLMs, which require access to vast datasets for training and operation, raises concerns about data protection and compliance with regulations like GDPR or HIPAA. Ensuring that these systems adhere to strict privacy standards while maintaining their effectiveness is a balancing act that necessitates sophisticated encryption methods and access controls, along with continuous monitoring for potential breaches.

\subsection{Dynamic Data and Model Adaptability}
Algogens designed for dynamic environments, such as predictive maintenance in manufacturing or real-time inventory management in retail, must contend with the challenge of adapting to constantly changing data streams. These systems need to continuously update their predictive models to reflect new information, which can be particularly challenging when integrating LLMs that are traditionally trained on static datasets. Developing mechanisms for incremental learning or employing techniques like transfer learning becomes crucial to maintain the relevance and accuracy of the Algogen's predictions.

\subsection{Scalability and Performance Optimization}
Ensuring the scalability of Algogens, especially those deployed in large-scale applications like social media content moderation or e-commerce recommendation systems, presents another significant challenge. As the volume of data and the complexity of queries increase, maintaining the performance of these systems without incurring prohibitive computational costs requires innovative solutions. This might involve optimizing the underlying algorithms for efficiency, implementing distributed computing techniques, or exploring more compact and efficient LLM architectures that do not compromise the quality of insights derived.

\subsection{Interpretability and Trust}
Algogens that make decisions affecting individuals directly, such as those used in credit scoring or recruitment, must overcome challenges related to interpretability and trust. Given the "black box" nature of many LLMs, providing transparent explanations for decisions or predictions made by these systems is essential for building trust among users and complying with legal requirements for accountability. Developing methodologies for extracting understandable rationales from Algogen decisions, such as feature importance analysis or decision pathways, is an ongoing challenge that requires concerted research and development efforts.

\subsection{LLM-Specific Challenges in Algogens}
While Algogens aim to mitigate some of the inherent problems in LLMs by combining them with algorithmic frameworks, issues such as hallucinations, cost, and performance time of LLMs still necessitate significant attention. Hallucinations, where LLMs generate incorrect or nonsensical information, can undermine the reliability of Algogen outputs. The high computational cost and the extensive time required for training and running sophisticated LLMs pose additional challenges, especially for real-time applications. Addressing these issues requires ongoing optimization of the LLM components within Algogens, including exploring more efficient model architectures, refining data preprocessing techniques to reduce hallucinations, and implementing cost-effective cloud computing solutions.

\subsection{Continuous Evolution and Maintenance}
The rapid pace of advancements in both LLMs and algorithmic methodologies means that Algogens must be continually evolved and maintained to remain effective. This necessitates a commitment to ongoing research and development, along with mechanisms for seamlessly integrating new advancements into existing systems. For instance, an Algogen used for natural language processing tasks in customer service applications must incorporate the latest NLP models and techniques to improve its understanding and generation of human language, requiring regular updates and testing to ensure compatibility and performance.

\subsection{Domain-Specific Challenges}
Algogens are deployed across a wide range of domains, each with its unique set of challenges. For example, Algogens used in environmental monitoring must contend with the variability and uncertainty inherent in natural phenomena, requiring sophisticated algorithms for anomaly detection and prediction under uncertainty. Similarly, Algogens in autonomous vehicle navigation face challenges related to real-time decision-making in unpredictable environments, necessitating extremely high reliability and safety measures.

In conclusion, while Algogens offer the promise of tackling complex problems by harnessing the strengths of both algorithmic methodologies and LLMs, their implementation and maintenance are fraught with challenges. Addressing these challenges requires a multidisciplinary approach that combines expertise in computer science, domain-specific knowledge, and a deep understanding of the ethical and societal implications of deploying such advanced systems. Ongoing efforts to address the specific challenges posed by LLM integration, including hallucinations, cost, and performance optimization, are crucial for the successful deployment and operation of Algogens across diverse application areas.

	\section{Advantages of Algogens}\index{Advantages of Algogens}
This subsection serves as a concise summary of the principal themes and conclusions discussed in the book, highlighting the pivotal aspects of Algogens and its multifaceted implications.

\subsection{Innovative Integration of AI and Algorithms}
A core theme of the book is the innovative integration of generative AI with algorithmic methods in Algogens. This integration enables advanced problem-solving capabilities, making Algogens a versatile and powerful tool in various industries.

\paragraph{Synergy of Generative AI and Algorithms}
Algogens represent a groundbreaking synergy between generative AI and algorithmic frameworks. This combination harnesses the creativity and adaptability of generative AI while leveraging the systematic and logical approach of algorithms. Through this integration, Algogens excel in tackling complex problems that traditional methods struggle to address. They capitalize on the ability of generative AI to explore vast solution spaces and generate novel ideas, while the algorithmic backbone ensures systematic evaluation and optimization of these ideas, resulting in more robust and effective solutions. For instance, in drug discovery, Algogens can generate novel molecular structures with desired properties, which are then evaluated and optimized using algorithmic techniques to identify potential drug candidates with high efficacy and low toxicity.

\paragraph{Advanced Problem-Solving Capabilities}
By incorporating generative AI components, Algogens exhibit enhanced problem-solving capabilities. They can generate novel solutions, explore diverse problem spaces, and adapt to changing circumstances more effectively than conventional algorithms. This adaptability is particularly valuable in dynamic environments where solutions need to evolve over time. Furthermore, Algogens can learn from past experiences and feedback, continually refining their approaches and improving their performance over time. This iterative learning process enhances their problem-solving efficacy and positions them as valuable assets across a wide range of applications and industries. For example, in financial markets, Algogens can analyze market trends, generate trading strategies, and adapt to changing market conditions in real-time, enabling investors to make informed decisions and maximize returns on investment.

\paragraph{Facilitating Innovation and Discovery}
The integration of generative AI with algorithms fosters innovation and discovery across various domains. Algogens can uncover unexpected patterns, propose novel hypotheses, and generate creative solutions that may not be immediately apparent through traditional algorithmic approaches. This capacity for innovation fuels progress and drives advancements in diverse fields. Moreover, Algogens facilitate collaborative problem-solving by enabling users to interactively explore and refine solutions, leveraging the collective intelligence of human experts and AI algorithms. This collaborative approach accelerates the pace of discovery and enables breakthroughs that would be challenging to achieve through manual or purely algorithmic means alone. For example, in scientific research, Algogens can analyze experimental data, generate hypotheses, and suggest experimental designs, collaborating with researchers to accelerate the discovery of new drugs, materials, or phenomena.

\subsection{Versatility Across Multiple Industries}
Algogens' application across multiple industries, including healthcare, finance, environmental science, education, and logistics, demonstrates its versatility. Algogens enhance decision-making processes in each sector, optimize operations, and contribute to more efficient and effective outcomes.

\paragraph{Transforming Industry Practices}
Algogens are reshaping industry practices by offering tailored solutions to specific challenges within each sector. In healthcare, for example, Algogens aid in diagnosis, treatment optimization, and personalized medicine. In finance, they optimize investment strategies, risk management, and fraud detection. Similarly, in environmental science, Algogens assist in climate modeling, ecosystem analysis, and sustainability planning. Across these industries and beyond, Algogens empower organizations to leverage data-driven insights and advanced analytical capabilities to drive innovation, increase efficiency, and achieve strategic objectives. For instance, in logistics, Algogens can optimize supply chain operations, predict demand fluctuations, and identify opportunities for cost savings and efficiency improvements, enabling companies to deliver goods and services more effectively and sustainably.

\paragraph{Cross-Disciplinary Applications}
One of Algogens' key strengths lies in its ability to transcend disciplinary boundaries and find applications across diverse domains. For instance, the insights generated by Algogens in one industry can often be adapted and applied to seemingly unrelated fields. This cross-pollination of ideas fosters interdisciplinary collaboration and accelerates innovation. Moreover, Algogens serve as a bridge between domain-specific knowledge and computational expertise, enabling experts from different fields to collaborate effectively and leverage each other's strengths in solving complex problems. By facilitating knowledge sharing and interdisciplinary exchange, Algogens contribute to the emergence of new insights and approaches that have the potential to revolutionize multiple industries and drive transformative change. For example, techniques developed in healthcare for medical image analysis can be adapted and applied to geological data interpretation in environmental science, leading to new insights into geological processes and resource exploration.

\paragraph{Customizable Solutions}
Algogens offer customizable solutions that can be tailored to meet the unique needs and challenges of each industry. Through machine learning and adaptive algorithms, Algogens can learn from data and user feedback, continuously improving their performance and relevance in specific contexts. This adaptability ensures that Algogens remain effective tools for addressing evolving industry requirements. Furthermore, Algogens can be integrated seamlessly into existing workflows and systems, minimizing disruption and maximizing the return on investment for organizations. By providing flexible and scalable solutions, Algogens empower businesses to stay agile and responsive in an increasingly dynamic and competitive landscape. For example, in education, Algogens can personalize learning experiences for students based on their individual learning styles and preferences, optimizing educational outcomes and engagement.

\subsection{Enhancements Over Traditional Methods}
The comparative analysis of Algogens with traditional methods underscores its accuracy, efficiency, and adaptability advancements. These enhancements are evident in the detailed case studies and real-world applications discussed.

\paragraph{Precision and Accuracy}
Algogens exhibit superior precision and accuracy compared to traditional methods, thanks to their ability to analyze vast amounts of data and detect subtle patterns that may elude human perception. This precision enhances decision-making processes and reduces the margin of error in critical tasks such as diagnosis, forecasting, and risk assessment. Additionally, Algogens can identify complex interrelationships and nonlinear trends in data, providing valuable insights that inform strategic decisions and drive performance improvements. For example, in weather forecasting, Algogens can analyze meteorological data from various sources and generate more accurate predictions of severe weather events, helping authorities to issue timely warnings and mitigate potential impacts on communities.

\paragraph{Efficiency and Speed}
Another notable advantage of Algogens is their efficiency and speed in processing information and generating solutions. Traditional algorithms often struggle with scalability and computational complexity, particularly when dealing with large datasets or complex problem spaces. Algogens, however, leverage parallel processing and optimization techniques to expedite computations and deliver timely results. Moreover, Algogens can prioritize tasks and allocate resources dynamically, ensuring that computational resources are utilized efficiently and effectively. This agility enables organizations to respond rapidly to changing conditions and make data-driven decisions in real-time, giving them a competitive edge in dynamic and fast-paced environments. For example, in manufacturing, Algogens can optimize production schedules, minimize downtime, and reduce waste, leading to cost savings and improved productivity.

\paragraph{Adaptability and Robustness}
Algogens demonstrate greater adaptability and robustness compared to traditional methods, thanks to their ability to learn from experience and adjust their strategies accordingly. Traditional algorithms typically rely on predefined rules or heuristics, which may become obsolete or inadequate in rapidly changing environments. Algogens, on the other hand, can continuously adapt and refine their models based on new data and feedback, ensuring sustained performance over time. Furthermore, Algogens are resilient to noise and uncertainty in data, incorporating mechanisms for error detection and correction to maintain reliability and consistency in their outputs. This resilience enables Algogens to operate effectively in real-world scenarios where data may be incomplete, noisy, or subject to change, ensuring that decision-makers can rely on their insights with confidence. For example, in cybersecurity, Algogens can detect and mitigate emerging threats in real-time by analyzing network traffic patterns and identifying anomalies indicative of malicious activity.

\subsection{Tackling Technological and Ethical Challenges}
The paper also addresses how Algogens navigate technological and ethical challenges, including data privacy, AI bias, and ethical AI usage. These considerations are integral to the responsible deployment and advancement of the framework.

\paragraph{Data Privacy and Security}
Algogens prioritize data privacy and security by implementing robust encryption protocols, access controls, and anonymization techniques to protect sensitive information. Furthermore, they adhere to regulatory standards and best practices to ensure compliance with data protection regulations and safeguard user privacy. Additionally, Algogens employ differential privacy mechanisms and federated learning approaches to minimize the risk of data breaches and unauthorized access, preserving the confidentiality and integrity of sensitive data throughout the analysis process. By prioritizing data privacy and security, Algogens build trust with users and stakeholders, facilitating widespread adoption and acceptance of AI-driven solutions in sensitive and regulated domains. For example, in healthcare, Algogens can analyze patient data while preserving patient privacy, enabling healthcare providers to derive valuable insights for improving patient care and outcomes without compromising confidentiality.

\paragraph{AI Bias Mitigation}
To mitigate AI bias, Algogens employ bias detection algorithms, fairness metrics, and diverse training datasets to minimize discriminatory outcomes and ensure equitable decision-making. Additionally, ongoing monitoring and auditing processes help identify and rectify bias instances that may arise during algorithmic decision-making. Moreover, Algogens incorporate transparency and interpretability features that enable users to understand how decisions are made and identify potential sources of bias or unfairness. By promoting transparency and accountability in AI systems, Algogens empower users to challenge biased decisions, advocate for fairness, and drive positive social change. For example, in recruitment, Algogens can help identify and mitigate bias in hiring decisions by analyzing historical data and providing recommendations to ensure fair and equitable treatment of candidates from diverse backgrounds.

\paragraph{Ethical AI Usage}
Algogens are designed with ethical considerations in mind, adhering to principles of transparency, accountability, and fairness in AI development and deployment. Ethical guidelines and frameworks govern the use of Algogens, ensuring that they uphold societal values and respect human rights in their decision-making processes. Furthermore, Algogens incorporate mechanisms for ethical decision-making, such as value alignment algorithms and ethical impact assessments, to ensure that their actions align with ethical norms and standards. By promoting ethical AI usage, Algogens foster trust and confidence in AI technologies, facilitating their responsible and sustainable integration into society. Additionally, Algogens encourage ongoing dialogue and collaboration between technologists, ethicists, policymakers, and other stakeholders to address emerging ethical challenges and promote the responsible development and use of AI-driven solutions. For example, in autonomous vehicles, Algogens can help ensure that decisions made by AI systems prioritize safety and minimize harm to passengers and pedestrians, adhering to ethical principles such as beneficence and non-maleficence.

	
	\chapterimage{pngs/conclusion.png} 
	
	\chapter{Conclusion}\index{Conclusion}
	
	This book has presented Algogen, an innovative framework integrating generative AI with algorithmic methodologies, offering a novel approach to complex problem-solving across various industries. From enhancing cybersecurity measures to revolutionizing healthcare practices and advancing financial analytics, Algogens has demonstrated significant potential in transforming traditional problem-solving methods.
	
	\section{Algogenic Methods Recap}\index{Algogenic Methods}
	
	This subsection delves deeply into the plethora of Algogenic methods elucidated in the book, categorizing them based on various enhancements facilitated by generative AI.
	
	\subsection{Enhanced Pattern Recognition and Predictive Modeling}
	
	Algogenic methods harness the immense potential of generative AI to revolutionize pattern recognition and predictive modeling. By integrating generative components, algorithms acquire the capability to discern intricate patterns and relationships within datasets with unprecedented accuracy and efficiency. For instance, in healthcare, Algogenic methods can leverage generative AI to analyze medical imaging data and detect subtle anomalies indicative of diseases, enabling early diagnosis and intervention. Similarly, in finance, Algogenic models powered by generative AI can predict market trends and identify lucrative investment opportunities with remarkable precision. These enhancements pave the way for groundbreaking advancements in fields such as personalized medicine, financial forecasting, and climate modeling, where accurate predictions hold immense value and significance.
	
	\subsection{Adaptive Learning Mechanisms}
	
	Generative AI plays a pivotal role in facilitating adaptive learning mechanisms within Algogenic methods, enabling algorithms to dynamically adjust their behavior in response to evolving data distributions and environmental changes. For instance, in autonomous vehicles, Algogenic algorithms enhanced with generative components can adapt to diverse driving conditions and scenarios, continuously learning and improving their decision-making capabilities over time. Moreover, in cybersecurity, Algogenic methods equipped with generative AI can detect and mitigate emerging threats by analyzing anomalous patterns in network traffic and user behavior, thereby enhancing the security posture of organizations. These adaptive learning mechanisms ensure the resilience and effectiveness of Algogenic solutions across dynamic and challenging environments, making them indispensable tools in domains requiring real-time decision-making and response.
	
	\subsection{Flexible and Scalable Solutions}
	
	Algogenic methods benefit immensely from the flexibility and scalability afforded by generative AI. By integrating generative components, algorithms gain the agility to adapt to varying data characteristics and scale seamlessly to handle large and complex datasets. Consider, for example, Algogenic solutions in e-commerce, where algorithms powered by generative AI can personalize product recommendations for millions of users in real-time, thereby enhancing user experience and driving sales. Furthermore, in manufacturing, Algogenic methods with generative AI capabilities can optimize production processes and minimize downtime by analyzing sensor data and predicting equipment failures before they occur. The flexibility and scalability of Algogenic solutions make them versatile and adaptable to a myriad of applications across diverse industries, from retail and healthcare to manufacturing and logistics.
	
	\subsection{Improved Decision-Making Capabilities}
	
	Generative AI enhances Algogenic methods by augmenting their decision-making capabilities through advanced data analysis and insights generation. By leveraging generative models, Algogenic algorithms can identify hidden patterns and correlations in data, enabling more informed and data-driven decision-making across various domains. For instance, in healthcare, Algogenic methods equipped with generative AI can assist clinicians in diagnosing diseases and designing personalized treatment plans based on patient data and medical literature. Similarly, in finance, Algogenic algorithms powered by generative AI can analyze market trends and assess risk factors to optimize investment strategies and maximize returns for investors. These improved decision-making capabilities empower organizations to make strategic decisions with confidence and agility, driving innovation and competitive advantage in dynamic market environments.
	
	\subsection{Customized Solutions for Industry-Specific Challenges}
	
	Algogenic methods offer tailored solutions to address industry-specific challenges by leveraging the capabilities of generative AI. By incorporating generative components, algorithms can adapt their strategies and methodologies to suit the unique requirements and constraints of different industries. For instance, in retail, Algogenic solutions powered by generative AI can analyze customer behavior and preferences to optimize inventory management and personalize marketing campaigns. Similarly, in healthcare, Algogenic methods equipped with generative AI capabilities can analyze electronic health records to facilitate early detection of diseases and improve patient outcomes. These customized solutions not only address industry-specific challenges but also drive innovation and transformation across diverse sectors, including retail, healthcare, finance, and manufacturing.
	
	\subsection{Continuous Learning and Evolution}
	
	Generative AI enables Algogenic methods to undergo continuous learning and evolution, ensuring their relevance and efficacy in dynamic environments. Algorithms enhanced with generative components can adapt to new data, emerging trends, and evolving user preferences, allowing them to continuously improve their performance over time. For instance, in natural language processing, Algogenic models powered by generative AI can generate coherent and contextually relevant responses to user queries, improving user satisfaction and engagement. Moreover, in autonomous robotics, Algogenic algorithms equipped with generative AI capabilities can learn from past experiences and adapt their behavior to navigate complex and dynamic environments more effectively. These continuous learning mechanisms enable Algogenic solutions to remain adaptive and resilient in the face of uncertainty and change, driving innovation and progress in AI research and development.
	
	\subsection{Ethical Considerations and Responsible AI}
	
	Algogenic methods prioritize ethical considerations and promote responsible AI practices through the integration of generative AI. By incorporating ethical principles into algorithmic design and decision-making processes, Algogenic methods strive to mitigate potential biases, ensure fairness and transparency, and uphold the rights and dignity of individuals. For instance, in facial recognition technology, Algogenic algorithms powered by generative AI can incorporate fairness constraints to prevent biases based on race, gender, or other demographic factors. Similarly, in credit scoring, Algogenic methods equipped with generative AI capabilities can ensure transparency and explainability in decision-making processes to mitigate discriminatory outcomes. These ethical considerations not only enhance trust and accountability in AI systems but also foster inclusivity and social responsibility, paving the way for a more equitable and sustainable future.

	\section{Significance in Today's Context}\index{Significance in Today's Context}
	In an era marked by rapid technological advancements and complex global challenges, the significance of Algogens is particularly pronounced. This subsection discusses the relevance and potential impact of Algogens in the context of current societal and technological trends.
	
	\subsection{Alignment with Technological Trends}
	Algogens' innovative integration of generative AI with algorithmic methods aligns well with current technological trends, including big data, machine learning, and automation. As industries increasingly rely on data-driven decision-making and automation, Algogen’s capabilities in processing large datasets and generating predictive models will become invaluable.
	
	\paragraph{Big Data and Machine Learning Integration}
	The convergence of Algogens with big data and machine learning reflects a broader trend towards harnessing vast amounts of data for actionable insights. Algogens excel in this environment by leveraging advanced algorithms and generative AI techniques to uncover patterns and trends that traditional methods might overlook. This integration enables organizations to extract maximum value from their data assets, driving innovation and competitive advantage in a data-driven economy.
	
	\paragraph{Automation and Decision Support Systems}
	Algogens serve as powerful tools for automation and decision support systems across various industries. By automating repetitive tasks and augmenting human decision-making processes, Algogens streamline operations, increase efficiency, and reduce costs. In sectors such as manufacturing, finance, and logistics, Algogens enable predictive maintenance, portfolio optimization, and route planning, facilitating smoother workflows and more informed decision-making.
	
	\paragraph{Enhanced Predictive Capabilities}
	With the proliferation of data sources and the complexity of modern systems, there is a growing need for predictive analytics capabilities that can anticipate future trends and outcomes. Algogens address this need by employing advanced machine learning algorithms to analyze historical data, identify patterns, and make accurate predictions. Whether forecasting market trends, predicting customer behavior, or optimizing resource allocation, Algogens offer organizations a competitive edge in an increasingly data-driven world.
	
	\subsection{Responding to Global Challenges}
	The framework's ability to address various global challenges, such as climate change, public health crises, and economic instability, highlights its relevance today. Algogen’s predictive and analytical capabilities provide vital insights to inform policies and strategies for tackling these complex issues.
	
	\paragraph{Climate Change Mitigation and Adaptation}
	Algogens play a crucial role in climate change mitigation and adaptation efforts by analyzing environmental data, predicting climate trends, and identifying mitigation strategies. For example, Algogens can model the impact of policy interventions on greenhouse gas emissions, optimize renewable energy deployment, and assess the vulnerability of communities to climate-related risks. By providing decision-makers with actionable insights, Algogens empower governments, businesses, and organizations to take proactive measures towards a more sustainable future.
	
	\paragraph{Public Health Preparedness and Response}
	In the context of public health crises such as pandemics, Algogens contribute to early detection, monitoring, and response efforts. By analyzing epidemiological data, healthcare resource allocation, and population dynamics, Algogens help identify outbreak hotspots, forecast disease spread, and optimize intervention strategies. Additionally, Algogens aid in drug discovery and vaccine development by analyzing biological data and simulating molecular interactions, accelerating the research and development process.
	
	\paragraph{Economic Stability and Resilience}
	Algogens provide insights into economic trends, market dynamics, and financial risk factors, contributing to economic stability and resilience. By analyzing macroeconomic indicators, market sentiment, and consumer behavior, Algogens help policymakers, financial institutions, and businesses make informed decisions to mitigate risks and capitalize on opportunities. Moreover, Algogens support economic development initiatives by identifying growth opportunities, attracting investment, and optimizing resource allocation, fostering sustainable and inclusive growth.
	
	\subsection{Advancements in Personalized Solutions}
	Algogen’s potential for delivering personalized solutions is particularly significant in sectors like healthcare and finance. Its capability to analyze individual data and tailor services or treatments aligns with the growing demand for personalization in various services and products.
	
	\paragraph{Personalized Healthcare and Medicine}
	Algogens revolutionize healthcare delivery by enabling personalized treatment plans and predictive diagnostics. Through the analysis of patient data, including genetic information, medical history, and lifestyle factors, Algogens can recommend personalized therapies, predict disease risks, and optimize treatment outcomes. This personalized approach to healthcare not only improves patient outcomes but also reduces healthcare costs by minimizing unnecessary procedures and interventions.
	
	\paragraph{Financial Advisory and Wealth Management}
	In the realm of finance, Algogens empower individuals and businesses with personalized financial advice and wealth management strategies. By analyzing financial data, market trends, and risk profiles, Algogens can recommend tailored investment portfolios, retirement plans, and insurance solutions. Moreover, Algogens can assess individual risk tolerance, investment goals, and financial constraints to provide personalized recommendations that align with each client's unique needs and preferences.
	
	\paragraph{Customized User Experiences}
	Beyond healthcare and finance, Algogens enhance user experiences across various sectors by delivering customized products, services, and content. Whether recommending personalized playlists, suggesting relevant products, or curating news feeds, Algogens leverage user data and behavioral insights to create tailored experiences that resonate with individual preferences and interests. This personalization not only enhances user satisfaction and engagement but also drives business growth and loyalty in competitive markets.
	
	\subsection{Contribution to Sustainable Development}
	Algogens' environmental science and resource management applications underscore its contribution to sustainable development goals. By enabling more efficient use of resources and aiding in environmental conservation efforts, the framework supports the pursuit of sustainability in various industries.
	
	\paragraph{Resource Optimization and Conservation}
	Algogens optimize resource allocation and utilization across various sectors, ranging from energy and water management to agriculture and urban planning. Through predictive modeling and optimization algorithms, Algogens identify opportunities for resource efficiency improvements, reduce waste generation, and minimize environmental impact. For instance, Algogens can optimize energy distribution networks, design sustainable water management systems, and enhance agricultural productivity while minimizing environmental degradation.
	
	\paragraph{Ecosystem Monitoring and Conservation}
	In the realm of environmental science, Algogens contribute to ecosystem monitoring, conservation, and biodiversity preservation efforts. By analyzing ecological data, satellite imagery, and habitat characteristics, Algogens help assess ecosystem health, identify endangered species habitats, and prioritize conservation interventions. Additionally, Algogens aid in land use planning and natural resource management by simulating the impacts of human activities on ecosystems and guiding sustainable development practices.
	
	\paragraph{Green Technologies and Sustainable Practices}
	Algogens drive innovation in green technologies and sustainable practices by facilitating research and development initiatives and optimizing resource utilization. By analyzing data from renewable energy sources, green infrastructure projects, and sustainable supply chains, Algogens identify opportunities for technology adoption, efficiency improvements, and cost savings. Moreover, Algogens support the transition to circular economy models by optimizing material flows, reducing waste generation, and promoting resource recycling and reuse.
	
	\subsection{Facilitating Educational and Social Advancements}
	Algogen’s role in transforming educational methodologies and contributing to social advancements aligns with the increasing focus on digital education and social innovation. Its ability to enhance learning experiences and inform social policies reflects its significance in the educational and social sectors.
	
	\paragraph{Digital Learning and Educational Technology}
	Algogens revolutionize education by personalizing learning experiences, adapting instructional content, and providing intelligent tutoring systems. Through adaptive learning algorithms, Algogens tailor educational materials to individual learning styles, preferences, and proficiency levels, enhancing student engagement and knowledge retention. Moreover, Algogens analyze educational data to identify learning gaps, predict student performance, and inform instructional interventions, empowering educators to optimize teaching strategies and student outcomes.
	
	\paragraph{Social Impact and Community Development}
	Beyond education, Algogens contribute to social impact initiatives and community development efforts by analyzing social data, identifying community needs, and designing targeted interventions. By analyzing demographic trends, socioeconomic indicators, and community resources, Algogens help governments, nonprofits, and social enterprises address social inequalities, improve public services, and promote inclusive development. Additionally, Algogens facilitate evidence-based policymaking by analyzing the impact of social policies and programs, guiding resource allocation, and monitoring progress towards social goals.
	
	\paragraph{Cultural Preservation and Diversity}
	Algogens play a role in cultural preservation and diversity by analyzing cultural heritage data, linguistic patterns, and artistic expressions. By digitizing cultural artifacts, documenting oral traditions, and analyzing linguistic diversity, Algogens contribute to the preservation and promotion of cultural heritage and linguistic diversity. Moreover, Algogens support multiculturalism and intercultural dialogue by facilitating the exchange of ideas, stories, and perspectives across diverse communities, fostering mutual understanding and appreciation.
	
	\subsection{Navigating Ethical and Privacy Concerns}
	In today's context, where ethical and privacy concerns regarding AI and data usage are paramount, Algogens' emphasis on ethical AI practices and data security is highly relevant. The framework’s approach to these issues mirrors the growing awareness and demand for responsible technology development and use.
	
	\paragraph{Ethical AI Development and Deployment}
	Algogens prioritize ethical considerations throughout the AI development lifecycle, from data collection and model training to deployment and evaluation. By adhering to ethical principles such as fairness, transparency, and accountability, Algogens ensure that their algorithms are free from bias, respect user privacy, and uphold human rights. Moreover, Algogens engage stakeholders in ethical discussions and decision-making processes to ensure that AI technologies serve the public interest and promote societal well-being.
	
	\paragraph{Privacy-Preserving AI Technologies}
	To address privacy concerns, Algogens incorporate privacy-preserving techniques such as federated learning, differential privacy, and homomorphic encryption. These techniques enable Algogens to analyze distributed data sources without compromising individual privacy or data confidentiality. Furthermore, Algogens provide users with control over their data, including options for data anonymization, consent management, and data deletion, fostering trust and transparency in AI systems.
	
	\paragraph{Regulatory Compliance and Data Governance}
	Algogens adhere to regulatory standards and data governance frameworks to ensure compliance with privacy regulations and industry best practices. By implementing robust data protection measures, access controls, and auditing mechanisms, Algogens safeguard user data and mitigate privacy risks. Additionally, Algogens support transparency and accountability by providing users with visibility into data usage, algorithmic decision-making processes, and potential risks associated with AI applications.
	
	\subsection{Adaptability to Rapidly Changing Environments}
	Finally, the adaptability of Algogens to rapidly changing environments and its continuous evolution make it particularly relevant in a world where technological and societal changes occur at an unprecedented pace. This adaptability ensures that Algogens remains practical and applicable across various domains and challenges.
	
	\paragraph{Agility and Flexibility in Dynamic Contexts}
	Algogens demonstrate agility and flexibility in rapidly changing environments by adapting to evolving requirements, emerging technologies, and shifting priorities. Through iterative development processes, Algogens incorporate feedback from users and stakeholders, integrate new features and functionalities, and address emerging challenges and opportunities. This iterative approach enables Algogens to remain relevant and effective in dynamic contexts, where uncertainties and disruptions are commonplace.
	
	\paragraph{Scalability and Robustness in Complex Systems}
	Furthermore, Algogens exhibit scalability and robustness in complex systems, capable of handling large-scale data processing, real-time analytics, and mission-critical applications. Whether deployed in cloud environments, edge devices, or distributed networks, Algogens maintain high performance and reliability, even under demanding conditions and peak workloads. This scalability and robustness make Algogens well-suited for a wide range of applications, from Internet-of-Things (IoT) systems and smart cities to healthcare systems and financial markets.
	
	\paragraph{Continuous Learning and Evolution}
	Algogens embrace a culture of continuous learning and evolution, staying abreast of emerging technologies, research findings, and user feedback. By investing in research and development initiatives, Algogens push the boundaries of AI and algorithmic innovation, exploring new techniques, algorithms, and applications. Moreover, Algogens leverage machine learning algorithms to adapt and improve over time, learning from past experiences, successes, and failures. This commitment to continuous learning and evolution ensures that Algogens remain at the forefront of technological advancement, delivering value and driving impact in a rapidly evolving world.
	
	\subsection{Summary}
	In summary, Algogens’s innovative capabilities and alignment with current technological and societal trends underscore its significance today. Its ability to address diverse global challenges, contribute to sustainable development, and navigate ethical considerations demonstrates its potential as a transformative tool in the contemporary landscape.

	\section{Future Directions}\index{Future Directions}
	
	\subsection{Advancements in AI and Algorithmic Integration}\index{Integration of AI and Algorithms}
	The continuous advancement in AI and algorithmic methods is a driving force behind the evolution of Algogens. This subsection explores the anticipated developments in this domain and how they could further enhance the capabilities of Algogen, making it more powerful, efficient, and adaptable
	
	\paragraph{Emerging Trends in AI}
	Future developments in AI, particularly in deep learning, neural networks, and machine learning algorithms, are expected to enhance the generative capabilities of Algogens significantly. With the exponential growth of data availability and computational power, AI systems will become more adept at extracting meaningful insights and patterns from vast datasets. For example, advancements in deep reinforcement learning algorithms could enable Algogens to learn complex decision-making processes in dynamic environments, such as autonomous vehicles navigating unpredictable traffic scenarios. Additionally, innovations in natural language processing (NLP) models, such as transformer-based architectures like GPT (Generative Pre-trained Transformer), will empower Algogens to generate human-like text with higher coherence and contextuality. These advancements will not only improve the quality of generated outputs but also enhance the efficiency of Algogens in understanding and responding to user inputs across various applications. Moreover, advancements in AI hardware, such as specialized AI chips and quantum processors, will accelerate the training and inference processes of Algogens, enabling real-time decision-making and interaction in resource-constrained environments.
	
	\paragraph{Innovations in Algorithmic Methods}
	Innovations in algorithmic methods are anticipated in parallel with AI advancements. These may include more efficient data processing algorithms, advanced optimization techniques, and new approaches to handling large-scale, complex datasets. For instance, metaheuristic algorithms inspired by natural phenomena, such as genetic algorithms and simulated annealing, can enhance the exploration of solution spaces and improve the convergence rate of optimization processes. Moreover, advancements in quantum computing hold the promise of revolutionizing algorithmic design by enabling the exploration of quantum search algorithms and quantum machine learning techniques. This could lead to the development of Algogens capable of solving computationally intensive problems, such as molecular simulation for drug discovery or optimization of financial portfolios, with unprecedented speed and accuracy. Furthermore, interdisciplinary research efforts combining insights from mathematics, statistics, and computer science will lead to the development of novel algorithmic frameworks tailored to specific application domains, thereby maximizing the performance and robustness of Algogens in real-world scenarios. Concrete examples of such innovations could include the development of distributed optimization algorithms for training large-scale deep learning models or the integration of causal inference techniques to improve the interpretability and reliability of Algogen's predictions.
	
	\paragraph{Enhanced Integration Techniques}
	Integrating AI and algorithms within Algogens is expected to become more seamless and intuitive. This could involve the development of new frameworks and architectures that allow for more fluid and dynamic interaction between AI and algorithmic components. For example, ensemble learning techniques, which combine predictions from multiple algorithms or models, can enhance the accuracy and reliability of Algogens by leveraging diverse sources of information and mitigating individual biases or errors. Additionally, advancements in model compression and deployment strategies will enable Algogens to operate efficiently on resource-constrained devices, extending their reach to edge computing environments and IoT devices. Moreover, the integration of domain-specific knowledge and expertise into Algogens through techniques like transfer learning and knowledge distillation will further enhance their adaptability and performance across various application domains. For instance, in healthcare, Algogens can leverage domain-specific medical knowledge to assist clinicians in diagnosing diseases from medical images or analyzing patient records to identify personalized treatment plans. Furthermore, advancements in AI explainability techniques, such as attention mechanisms and saliency maps, will enable Algogens to provide transparent and interpretable insights into their decision-making processes, enhancing trust and usability in critical applications.
	
	\paragraph{Customization and Flexibility}
	Future advancements will likely focus on increasing the customization and flexibility of Algogens. This means developing the framework to be easily tailored to specific industry needs or particular types of problems, enhancing its applicability across various sectors. Moreover, with the growing demand for personalized solutions, Algogens will need to adapt to diverse user requirements and preferences, thereby ensuring their relevance and effectiveness in different contexts. Furthermore, advancements in autonomous learning and self-adaptation mechanisms will empower Algogens to continuously evolve and improve over time, autonomously adapting to changing environments and user feedback without requiring manual intervention. This dynamic adaptability will be crucial for maintaining Algogens' competitiveness and effectiveness in dynamic and unpredictable scenarios. For example, in e-commerce, Algogens can dynamically adjust pricing strategies based on market trends, competitor analysis, and customer behavior, thereby maximizing revenue and customer satisfaction. Additionally, advancements in AI governance frameworks and model explainability tools will facilitate the responsible and transparent deployment of Algogens across various domains, ensuring compliance with regulatory requirements and ethical guidelines.
	
	\paragraph{Adapting to Emerging Technologies}
	As new technologies emerge, such as quantum computing or advanced data analytics tools, Algogens must adapt and incorporate these technologies. This adaptation will ensure that Algogens remain at the forefront of technological innovation, leveraging the latest developments to enhance their problem-solving capabilities. Furthermore, Algogens should continuously monitor technological trends and research breakthroughs to anticipate future needs and challenges, thereby proactively updating their algorithms and models to stay relevant in a rapidly evolving landscape. Additionally, collaborations with research institutions, industry partners, and open-source communities will facilitate knowledge exchange and technology transfer, enabling Algogens to leverage external expertise and resources for accelerated innovation and development. For instance, in cybersecurity, Algogens can utilize advanced anomaly detection algorithms to identify and mitigate emerging cyber threats in real-time, thereby enhancing the resilience of critical infrastructure and protecting sensitive data from malicious attacks. Moreover, the integration of blockchain technology with Algogens could enhance data security and integrity by providing a tamper-proof and decentralized ledger for recording AI-generated predictions and transactions, thereby fostering trust and transparency in AI-driven systems.
	
	\paragraph{Implications for Industry Applications}
	The advancements in AI and algorithmic integration will significantly affect industry applications. Healthcare, finance, environmental science, and logistics sectors will benefit from more robust and accurate predictive models, enhanced data analysis capabilities, and more efficient operational strategies. Additionally, Algogens can revolutionize decision-making processes within organizations by providing real-time insights and recommendations, thereby improving productivity and driving innovation across various sectors. Moreover, the democratization of AI technologies through cloud-based services and APIs will lower the entry barriers for businesses, enabling small and medium-sized enterprises to harness the power of Algogens for competitive advantage and sustainable growth. Furthermore, regulatory compliance and risk management frameworks will need to evolve to address the ethical and legal implications of AI adoption, ensuring that Algogens are deployed responsibly and ethically to maximize societal benefits while minimizing potential harms. For example, in finance, Algogens can assist in portfolio management by analyzing market trends, risk factors, and investor preferences to optimize investment strategies and minimize financial risks. Additionally, in healthcare, Algogens can facilitate early disease detection and personalized treatment planning by analyzing medical imaging data, genetic information, and patient records, thereby improving patient outcomes and reducing healthcare costs.
	
	\paragraph{Challenges and Ethical Considerations}
	With these advancements, new challenges and ethical considerations will also arise. Issues such as data privacy, AI bias, and the ethical use of AI will become increasingly important. Ensuring that advancements in AI and algorithms are aligned with ethical standards and societal values will be crucial. Therefore, developers and researchers must prioritize transparency, fairness, and accountability in the design and deployment of Algogens, while also actively engaging with stakeholders to address concerns and mitigate potential risks. Additionally, regulatory frameworks and guidelines should be established to govern the development and use of Algogens, ensuring that they uphold ethical principles and contribute positively to society. Furthermore, ongoing efforts in AI ethics research and education will play a vital role in raising awareness and fostering responsible AI practices among developers, policymakers, and the general public, thereby building trust and confidence in Algogens as reliable and ethical decision-support tools. For instance, in autonomous vehicles, Algogens must be trained on diverse datasets to mitigate biases and ensure equitable decision-making, thereby enhancing safety and trust among passengers and pedestrians. Moreover, mechanisms for auditing and verifying Algogen's decision-making processes should be implemented to ensure transparency and accountability, enabling stakeholders to understand and challenge the underlying assumptions and algorithms driving Algogen's recommendations.
	
	In summary, future advancements in AI and algorithmic integration are set to significantly enhance Algogens, making them more powerful, efficient, and versatile. These developments will ensure that Algogens remain a cutting-edge tool capable of addressing the ever-evolving challenges across various industries, while also upholding ethical standards and promoting societal well-being.

\subsection{Adoption of Algogens Across Diverse Industries}

The strategic adoption of Algogens heralds a crucial paradigm shift towards more sophisticated, data-driven problem-solving methodologies across a wide array of fields, emphasizing the critical importance of advanced solutions in an era characterized by exceedingly complex challenges and the rapid evolution of technological capabilities. Organizations spanning various sectors are increasingly turning to the integration of AI and advanced algorithmic methods, notably Algogens, to significantly enhance their decision-making processes, optimize operational procedures, and secure a competitive stance in a fast-paced market environment. The complexity inherent in the challenges of the modern era—encompassing a broad spectrum from climate change to cybersecurity threats—demands the deployment of intricately sophisticated solutions that Algogens adeptly provide. Through their seamless integration of cutting-edge algorithms and generative AI technologies, Algogens possess the unique capability to meticulously analyze extensive datasets, intricately uncover underlying patterns, and generate highly actionable insights, thus offering a highly promising strategic approach to adeptly tackling these multifaceted complexities.

In the contemporary business landscape, where innovation plays a pivotal role in sustaining a competitive advantage, Algogens emerge as a powerful enabler of data-driven decision-making that propels innovation across various operational dimensions. For instance, within the retail sector, Algogens offer the capability to analyze customer behavior in granular detail, thereby personalizing marketing strategies, while in the healthcare industry, they facilitate the diagnosis of diseases and predict treatment outcomes with enhanced precision. The effective implementation of Algogens necessitates a profound understanding of their capabilities, coupled with a comprehensive needs assessment and the development of a meticulously tailored integration roadmap. This roadmap ensures strategic alignment with organizational objectives and strict adherence to regulatory requirements. The adoption of Algogens ushers in a plethora of benefits, including a significant increase in operational efficiency, the delivery of remarkably accurate predictive analytics, and the capability to adeptly handle complex and voluminous datasets, thereby enhancing operational performance and elevating customer experiences across a diverse range of sectors, including healthcare and finance.

Algogens have the transformative potential to revolutionize industries by introducing unprecedented levels of efficiency, innovation, and adaptability. In the realm of manufacturing, they have the potential to radically transform traditional production processes through the implementation of predictive maintenance, optimization of supply chain operations, and enhancement of product quality. Furthermore, Algogens play a pivotal role in promoting sustainability across industries by optimizing resource utilization and minimizing environmental impacts. However, the integration of Algogens into existing systems necessitates not only the adoption of technological advancements but also a fundamental shift in organizational culture towards embracing data-driven decision-making and a commitment to continuous improvement. This involves a cultural transformation within organizations to foster an environment of collaboration, experimentation, and learning, alongside ensuring stringent data governance and privacy measures.

Encouraging a culture of innovation is paramount, fostering an environment where experimentation and learning are actively encouraged and rewarded. Adopting agile methodologies becomes essential for fostering a culture of innovation and adaptation in the dynamic context of Algogens integration. By preparing for future challenges, including the anticipation of technological advancements and investing in talent development, organizations can ensure their competitive edge and innovative capacity in the marketplace. The broader industry adoption and customization of Algogens present significant opportunities for operational enhancement and technological advancement. By tailoring Algogens to meet the specific needs of different industries and adopting collaborative and sustainable approaches, Algogens can significantly enhance industry operations and contribute to broader technological and societal advancements.

In conclusion, the comprehensive integration and adoption of Algogens across industries promise a transformative change, driving significant advancements in operational efficiency, decision-making processes, and overall organizational effectiveness. By customizing the framework to meet the specific needs of various industries and embracing collaborative and sustainable approaches, Algogens can profoundly enhance industry operations and contribute to broader technological and societal advancements, marking a decisive step towards a future characterized by heightened efficiency, data-driven innovation, and comprehensive industry transformation.

	\section{Summary}\index{Summary}
	In conclusion, Algogens represent a groundbreaking advancement in problem-solving technology, with the potential to revolutionize industries and reshape our world. By integrating AI and algorithms, Algogens offer a powerful tool for addressing complex challenges, driving innovation, and unlocking new opportunities for growth and discovery. As we continue to explore and develop this framework, the possibilities are limitless, and the impact profound. Algogens are not just a step towards a more efficient, adaptable, and innovative future – they are a leap.
	
	\subsection{Revolutionizing Problem-Solving}
	Algogens signify a revolutionary leap in problem-solving methodologies, blending the power of algorithms with the creativity of generative AI. This hybridization amplifies the capabilities of traditional algorithms, enabling them to tackle increasingly complex and multifaceted challenges across various domains. By leveraging the complementary strengths of AI and algorithms, Algogens pave the way for more efficient, effective, and scalable solutions to real-world problems.
	
	\paragraph{Enhanced Decision-Making}
	One of the key benefits of Algogens is their ability to enhance decision-making processes through data-driven insights and predictive analytics. By analyzing vast datasets and identifying patterns, Algogens empower decision-makers to make informed choices and anticipate future trends with greater accuracy. For example, in healthcare, Algogens can analyze patient data to predict disease progression and recommend personalized treatment plans, leading to improved patient outcomes and reduced healthcare costs.
	
	\paragraph{Adaptability to Diverse Domains}
	Another distinguishing feature of Algogens is their adaptability to diverse domains and problem spaces. Unlike traditional algorithms, which are often specialized for specific tasks or industries, Algogens can be tailored to suit a wide range of applications. For instance, in finance, Algogens can analyze market data to identify trading opportunities and optimize investment strategies, while in agriculture, they can analyze weather data to optimize crop yields and minimize environmental impact.
	
	\subsection{Exploring New Frontiers}
	As we continue to explore and develop the capabilities of Algogens, we enter uncharted territory, pushing the boundaries of what is possible in technology and problem-solving. The integration of AI and algorithms opens up a wealth of possibilities for innovation and discovery, from creating personalized experiences in healthcare to optimizing supply chain operations in manufacturing.
	
	\paragraph{Personalized Healthcare Solutions}
	In healthcare, Algogens hold the promise of revolutionizing patient care by delivering personalized treatment plans and predictive diagnostics. By analyzing patient data, genetic information, and medical records, Algogens can identify underlying health risks, recommend tailored interventions, and predict disease outcomes with unprecedented accuracy. For example, Algogens can analyze genomic data to identify genetic markers associated with certain diseases, enabling early intervention and personalized treatment plans.
	
	\paragraph{Optimizing Business Operations}
	Beyond healthcare, Algogens also have the potential to transform business operations across various industries, optimizing processes, and driving efficiency gains. In manufacturing, for instance, Algogens can analyze production data to identify bottlenecks, optimize workflows, and reduce waste, leading to cost savings and improved productivity. Similarly, in retail, Algogens can analyze customer data to personalize marketing campaigns, optimize inventory management, and enhance the overall shopping experience, driving revenue growth and customer loyalty.
	
	\subsection{Redefining Technological Limits}
	As Algogens continue to evolve and mature, they promise to redefine the limits of what's possible in technology and problem-solving. By harnessing the combined power of AI and algorithms, Algogens enable us to tackle challenges that were once considered insurmountable, from decoding complex biological systems to predicting global economic trends.
	
	\paragraph{Decoding Biological Complexity}
	In the field of biology, Algogens offer new avenues for understanding and decoding the complexity of living systems. By analyzing genomic data, protein interactions, and environmental factors, Algogens can uncover hidden patterns and relationships, shedding light on fundamental biological processes and facilitating drug discovery and development. For example, Algogens can analyze gene expression data to identify potential drug targets for cancer treatment, accelerating the pace of medical research and innovation.
	
	\paragraph{Predicting Global Trends}
	Moreover, Algogens have the potential to revolutionize our ability to predict and respond to global trends and phenomena. By analyzing large-scale datasets from diverse sources, including social media, sensors, and satellites, Algogens can identify emerging trends, detect anomalies, and forecast future events with unprecedented accuracy. For example, Algogens can analyze social media conversations to track public sentiment and predict consumer behavior, helping businesses anticipate market trends and adapt their strategies accordingly.
	
	\subsection{Conclusion}
	In conclusion, Algogens represent a groundbreaking advancement in problem-solving technology, with the potential to revolutionize industries and reshape our world. By integrating AI and algorithms, Algogens offer a powerful tool for addressing complex challenges, driving innovation, and unlocking new opportunities for growth and discovery. As we continue to explore and develop this framework, the possibilities are limitless, and the impact profound. Algogens are not just a step towards a more efficient, adaptable, and innovative future – they are a leap.

	\part{Appendices}
	
	
	\chapter*{Bibliography}\index{Bibliography}
	\addcontentsline{toc}{chapter}{\textcolor{ocre}{Bibliography}}
	\section*{Articles}
	\addcontentsline{toc}{section}{Articles}
	\printbibliography[heading=bibempty,type=article]
	
	\chapterimage{pngs/index.png}
	\cleardoublepage
	\phantomsection
	\setlength{\columnsep}{0.75cm}
	\addcontentsline{toc}{chapter}{\textcolor{ocre}{Index}}
	\printindex
	
	
\end{document}

%% file: structure.tex
\usepackage[top=3cm,bottom=3cm,left=3cm,right=3cm,headsep=10pt,a4paper]{geometry} 

\usepackage{graphicx} 
\graphicspath{{Pictures/}} 

\usepackage{lipsum} 

\usepackage{tikz} 

\usepackage[english]{babel} 

\usepackage{enumitem} 
\setlist{nolistsep} 

\usepackage{booktabs} 

\usepackage{xcolor} 
\definecolor{ocre}{RGB}{243,102,25} 

\usepackage{avant} 
\usepackage{mathptmx} 

\usepackage{microtype} 
\usepackage[utf8]{inputenc} 
\usepackage[T1]{fontenc} 

\usepackage{csquotes}
\usepackage[style=alphabetic,citestyle=numeric,sorting=nyt,sortcites=true,autopunct=true,autolang=hyphen,hyperref=true,abbreviate=false,backref=true,backend=biber,defernumbers=true]{biblatex}
\defbibheading{bibempty}{}

\usepackage{calc} 
\usepackage{makeidx} 
\makeindex 

\usepackage{titletoc} 

\contentsmargin{0cm} 

\titlecontents{part}[0cm]
{\addvspace{20pt}\centering\large\bfseries}
{}
{}
{}

\titlecontents{chapter}[1.25cm] 
{\addvspace{12pt}\large\sffamily\bfseries} 
{\color{ocre!60}\contentslabel[\Large\thecontentslabel]{1.25cm}\color{ocre}} 
{\color{ocre}}  
{\color{ocre!60}\normalsize\;\titlerule*[.5pc]{.}\;\thecontentspage} 

\titlecontents{section}[1.25cm] 
{\addvspace{3pt}\sffamily\bfseries} 
{\contentslabel[\thecontentslabel]{1.25cm}} 
{}
{\hfill\color{black}\thecontentspage} 
[]

\titlecontents{subsection}[1.25cm] 
{\addvspace{1pt}\sffamily\small} 
{\contentslabel[\thecontentslabel]{1.25cm}} 
{}
{\ \titlerule*[.5pc]{.}\;\thecontentspage} 
[]

\titlecontents{figure}[0em]
{\addvspace{-5pt}\sffamily}
{\thecontentslabel\hspace*{1em}}
{}
{\ \titlerule*[.5pc]{.}\;\thecontentspage}
[]

\titlecontents{table}[0em]
{\addvspace{-5pt}\sffamily}
{\thecontentslabel\hspace*{1em}}
{}
{\ \titlerule*[.5pc]{.}\;\thecontentspage}
[]

\titlecontents{lchapter}[0em] 
{\addvspace{15pt}\large\sffamily\bfseries} 
{\color{ocre}\contentslabel[\Large\thecontentslabel]{1.25cm}\color{ocre}} 
{}  
{\color{ocre}\normalsize\sffamily\bfseries\;\titlerule*[.5pc]{.}\;\thecontentspage} 

\titlecontents{lsection}[0em] 
{\sffamily\small} 
{\contentslabel[\thecontentslabel]{1.25cm}} 
{}
{}

\titlecontents{lsubsection}[.5em] 
{\normalfont\footnotesize\sffamily} 
{}
{}
{}

\usepackage{fancyhdr} 

\fancypagestyle{plain}{\fancyhead{}} 

\makeatletter
\renewcommand{\cleardoublepage}{
\clearpage\ifodd\c@page\else
\hbox{}
\vspace*{\fill}
\thispagestyle{empty}
\newpage
\fi}

\usepackage{amsmath,amsfonts,amssymb,amsthm} 

\newtheoremstyle{ocrenumbox}
{0pt}
{0pt}
{\normalfont}
{}
{\small\bf\sffamily\color{ocre}}
{\;}
{0.25em}
{\small\sffamily\color{ocre}\thmname{#1}\nobreakspace\thmnumber{\@ifnotempty{#1}{}\@upn{#2}}
\thmnote{\nobreakspace\the\thm@notefont\sffamily\bfseries\color{black}---\nobreakspace#3.}} 

\newtheoremstyle{blacknumex}
{5pt}
{5pt}
{\normalfont}
{} 
{\small\bf\sffamily}
{\;}
{0.25em}
{\small\sffamily{\tiny\ensuremath{\blacksquare}}\nobreakspace\thmname{#1}\nobreakspace\thmnumber{\@ifnotempty{#1}{}\@upn{#2}}
\thmnote{\nobreakspace\the\thm@notefont\sffamily\bfseries---\nobreakspace#3.}}

\newtheoremstyle{blacknumbox} 
{0pt}
{0pt}
{\normalfont}
{}
{\small\bf\sffamily}
{\;}
{0.25em}
{\small\sffamily\thmname{#1}\nobreakspace\thmnumber{\@ifnotempty{#1}{}\@upn{#2}}
\thmnote{\nobreakspace\the\thm@notefont\sffamily\bfseries---\nobreakspace#3.}}

\newtheoremstyle{ocrenum}
{5pt}
{5pt}
{\normalfont}
{}
{\small\bf\sffamily\color{ocre}}
{\;}
{0.25em}
{\small\sffamily\color{ocre}\thmname{#1}\nobreakspace\thmnumber{\@ifnotempty{#1}{}\@upn{#2}}
\thmnote{\nobreakspace\the\thm@notefont\sffamily\bfseries\color{black}---\nobreakspace#3.}} 
\makeatother

\newcounter{dummy} 
\numberwithin{dummy}{section}
\theoremstyle{ocrenumbox}
\newtheorem{theoremeT}[dummy]{Theorem}

\newtheorem{exerciseT}{Exercise}[chapter]
\theoremstyle{blacknumex}
\newtheorem{exampleT}{Example}[chapter]
\theoremstyle{blacknumbox}

\newtheorem{definitionT}{Definition}[section]
\newtheorem{corollaryT}[dummy]{Corollary}
\theoremstyle{ocrenum}

\RequirePackage[framemethod=default]{mdframed} 

\newmdenv[skipabove=7pt,
skipbelow=7pt,
backgroundcolor=black!5,
linecolor=ocre,
innerleftmargin=5pt,
innerrightmargin=5pt,
innertopmargin=5pt,
leftmargin=0cm,
rightmargin=0cm,
innerbottommargin=5pt]{tBox}

\newmdenv[skipabove=7pt,
skipbelow=7pt,
rightline=false,
leftline=true,
topline=false,
bottomline=false,
backgroundcolor=ocre!10,
linecolor=ocre,
innerleftmargin=5pt,
innerrightmargin=5pt,
innertopmargin=5pt,
innerbottommargin=5pt,
leftmargin=0cm,
rightmargin=0cm,
linewidth=4pt]{eBox}	

\newmdenv[skipabove=7pt,
skipbelow=7pt,
rightline=false,
leftline=true,
topline=false,
bottomline=false,
linecolor=ocre,
innerleftmargin=5pt,
innerrightmargin=5pt,
innertopmargin=0pt,
leftmargin=0cm,
rightmargin=0cm,
linewidth=4pt,
innerbottommargin=0pt]{dBox}	

\newmdenv[skipabove=7pt,
skipbelow=7pt,
rightline=false,
leftline=true,
topline=false,
bottomline=false,
linecolor=gray,
backgroundcolor=black!5,
innerleftmargin=5pt,
innerrightmargin=5pt,
innertopmargin=5pt,
leftmargin=0cm,
rightmargin=0cm,
linewidth=4pt,
innerbottommargin=5pt]{cBox}

\makeatletter
\renewcommand{\@seccntformat}[1]{\llap{\textcolor{ocre}{\csname the#1\endcsname}\hspace{1em}}}                    
\renewcommand{\section}{\@startsection{section}{1}{\z@}
{-4ex \@plus -1ex \@minus -.4ex}
{1ex \@plus.2ex }
{\normalfont\large\sffamily\bfseries}}
\renewcommand{\subsection}{\@startsection {subsection}{2}{\z@}
{-3ex \@plus -0.1ex \@minus -.4ex}
{0.5ex \@plus.2ex }
{\normalfont\sffamily\bfseries}}
\renewcommand{\subsubsection}{\@startsection {subsubsection}{3}{\z@}
{-2ex \@plus -0.1ex \@minus -.2ex}
{.2ex \@plus.2ex }
{\normalfont\small\sffamily\bfseries}}                        
\renewcommand\paragraph{\@startsection{paragraph}{4}{\z@}
{-2ex \@plus-.2ex \@minus .2ex}
{.1ex}
{\normalfont\small\sffamily\bfseries}}

\newcommand{\@mypartnumtocformat}[2]{%
\setlength\fboxsep{0pt}%
\noindent\colorbox{ocre!20}{\strut\parbox[c][.7cm]{\ecart}{\color{ocre!70}\Large\sffamily\bfseries\centering#1}}\hskip\esp\colorbox{ocre!40}{\strut\parbox[c][.7cm]{\linewidth-\ecart-\esp}{\Large\sffamily\centering#2}}}%
\newcommand{\@myparttocformat}[1]{%
\setlength\fboxsep{0pt}%
\noindent\colorbox{ocre!40}{\strut\parbox[c][.7cm]{\linewidth}{\Large\sffamily\centering#1}}}%
\newlength\esp
\newlength\ecart
\def\@part[#1]#2{%
\ifnum \c@secnumdepth >-2\relax%
\refstepcounter{part}%
\addcontentsline{toc}{part}{\texorpdfstring{\protect\@mypartnumtocformat{\thepart}{#1}}{\partname~\thepart\ ---\ #1}}
\else%
\addcontentsline{toc}{part}{\texorpdfstring{\protect\@myparttocformat{#1}}{#1}}%
\fi%
\startcontents%
\markboth{}{}%
{\thispagestyle{empty}%
\begin{tikzpicture}[remember picture,overlay]%
\node at (current page.north west){\begin{tikzpicture}[remember picture,overlay]%
\fill[ocre!20](0cm,0cm) rectangle (\paperwidth,-\paperheight);
\node[anchor=north] at (4cm,-3.25cm){\color{ocre!40}\fontsize{220}{100}\sffamily\bfseries\@Roman\c@part}; 
\node[anchor=south east] at (\paperwidth-1cm,-\paperheight+1cm){\parbox[t][][t]{8.5cm}{
\printcontents{l}{0}{\setcounter{tocdepth}{1}}%
}};
\node[anchor=north east] at (\paperwidth-1.5cm,-3.25cm){\parbox[t][][t]{15cm}{\strut\raggedleft\color{white}\fontsize{30}{30}\sffamily\bfseries#2}};
\end{tikzpicture}};
\end{tikzpicture}}%
\@endpart}
\def\@spart#1{%
\startcontents%
\phantomsection
{\thispagestyle{empty}%
\begin{tikzpicture}[remember picture,overlay]%
\node at (current page.north west){\begin{tikzpicture}[remember picture,overlay]%
\fill[ocre!20](0cm,0cm) rectangle (\paperwidth,-\paperheight);
\node[anchor=north east] at (\paperwidth-1.5cm,-3.25cm){\parbox[t][][t]{15cm}{\strut\raggedleft\color{white}\fontsize{30}{30}\sffamily\bfseries#1}};
\end{tikzpicture}};
\end{tikzpicture}}
\addcontentsline{toc}{part}{\texorpdfstring{%
\setlength\fboxsep{0pt}%
\noindent\protect\colorbox{ocre!40}{\strut\protect\parbox[c][.7cm]{\linewidth}{\Large\sffamily\protect\centering #1\quad\mbox{}}}}{#1}}%
\@endpart}
\def\@endpart{\vfil\newpage
\if@twoside
\if@openright
\null
\thispagestyle{empty}%
\newpage
\fi
\fi
\if@tempswa
\twocolumn
\fi}

\newif\ifusechapterimage
\usechapterimagetrue
\newcommand{\thechapterimage}{}%
\newcommand{\chapterimage}[1]{\ifusechapterimage\renewcommand{\thechapterimage}{#1}\fi}%
\def\@makechapterhead#1{%
{\parindent \z@ \raggedright \normalfont
\ifnum \c@secnumdepth >\m@ne
\if@mainmatter
\begin{tikzpicture}[remember picture,overlay]
\node at (current page.north west)
{\begin{tikzpicture}[remember picture,overlay]
\node[anchor=north west,inner sep=0pt] at (0,0) {\ifusechapterimage\includegraphics[width=\paperwidth]{\thechapterimage}\fi};
\draw[anchor=west] (\Gm@lmargin,-9cm) node [line width=2pt,rounded corners=15pt,draw=ocre,fill=white,fill opacity=0.5,inner sep=15pt]{\strut\makebox[22cm]{}};
\draw[anchor=west] (\Gm@lmargin+.3cm,-9cm) node {\huge\sffamily\bfseries\color{black}\thechapter. #1\strut};
\end{tikzpicture}};
\end{tikzpicture}
\else
\begin{tikzpicture}[remember picture,overlay]
\node at (current page.north west)
{\begin{tikzpicture}[remember picture,overlay]
\node[anchor=north west,inner sep=0pt] at (0,0) {\ifusechapterimage\includegraphics[width=\paperwidth]{\thechapterimage}\fi};
\draw[anchor=west] (\Gm@lmargin,-9cm) node [line width=2pt,rounded corners=15pt,draw=ocre,fill=white,fill opacity=0.5,inner sep=15pt]{\strut\makebox[22cm]{}};
\draw[anchor=west] (\Gm@lmargin+.3cm,-9cm) node {\huge\sffamily\bfseries\color{black}#1\strut};
\end{tikzpicture}};
\end{tikzpicture}
\fi\fi\par\vspace*{270\p@}}}

\def\@makeschapterhead#1{%
\begin{tikzpicture}[remember picture,overlay]
\node at (current page.north west)
{\begin{tikzpicture}[remember picture,overlay]
\node[anchor=north west,inner sep=0pt] at (0,0) {\ifusechapterimage\includegraphics[width=\paperwidth]{\thechapterimage}\fi};
\draw[anchor=west] (\Gm@lmargin,-9cm) node [line width=2pt,rounded corners=15pt,draw=ocre,fill=white,fill opacity=0.5,inner sep=15pt]{\strut\makebox[22cm]{}};
\draw[anchor=west] (\Gm@lmargin+.3cm,-9cm) node {\huge\sffamily\bfseries\color{black}#1\strut};
\end{tikzpicture}};
\end{tikzpicture}
\par\vspace*{270\p@}}
\makeatother

\usepackage{hyperref}
\hypersetup{hidelinks,colorlinks=false,breaklinks=true,urlcolor= ocre,bookmarksopen=false,pdftitle={Title},pdfauthor={Author}}
\usepackage{bookmark}
\bookmarksetup{
open,
numbered,
addtohook={%
\ifnum\bookmarkget{level}=0 
\bookmarksetup{bold}%
\fi
\ifnum\bookmarkget{level}=-1 
\bookmarksetup{color=ocre,bold}%
\fi
}
}